\documentclass[10pt,twocolumn,letterpaper]{article}
\usepackage{arxiv_preprint}
\ifdefined\pdfinfoomitdate \pdfinfoomitdate=1 \fi
\ifdefined\pdftrailerid \pdftrailerid{} \fi
\ifdefined\pdfsuppressptexinfo \pdfsuppressptexinfo=-1 \fi

\usepackage{cuted} 
\usepackage{currfile}
\usepackage{caption}

\newcommand{\todo}[1]{{\color{red}#1}}

\newif\ifaudithold \auditholdtrue

\usepackage[T1]{fontenc}    
\usepackage{tikz}
\usetikzlibrary{arrows.meta,positioning,calc,fit,backgrounds}
\usepackage{adjustbox}
\usepackage{multirow}
\usepackage{array}
\usepackage{pifont}
\usepackage{microtype}      

\makeatletter
\DeclareRobustCommand{\R}[1]{\@ifundefined{R@#1}{\todo{[undefined key #1]}}{\@nameuse{R@#1}}}
\makeatother
\makeatletter
\newcommand{\resIf}[3]{\@ifundefined{R@#1}{#3}{#2}}
\newcommand{\resIfPos}[3]{\@ifundefined{R@#1}{#3}{%
  \ifnum\numexpr0\csname R@#1\endcsname\relax>0 #2\else #3\fi}}
\def\res@pendmacro{\pend}
\newcommand{\resLanded}[3]{\@ifundefined{R@#1}{#3}{\expandafter\let\expandafter\res@tmp\csname R@#1\endcsname\ifx\res@tmp\res@pendmacro #3\else #2\fi}}
\newcommand{\bd@eval}[3]{%
  \@ifundefined{R@#1}{\gdef\bd@state{p}}{%
    \expandafter\ifx\csname R@#1\endcsname\res@pendmacro\gdef\bd@state{p}\else
      \begingroup\let\ensuremath\@firstofone
      \edef\bd@val{\csname R@#1\endcsname}%
      \pgfmathparse{ifthenelse(\bd@val<#2,2,ifthenelse(\bd@val>#3,1,0))}%
      \ifdim\pgfmathresult pt=2pt \xdef\bd@state{b}\else\ifdim\pgfmathresult pt=1pt \xdef\bd@state{a}\else\xdef\bd@state{i}\fi\fi
      \endgroup
    \fi}}
\newcommand{\bd@evalK}[3]{%
  \begingroup\let\ensuremath\@firstofone
  \@ifundefined{R@#2}{\edef\bd@lo{0}}{\edef\bd@lo{\csname R@#2\endcsname}}%
  \@ifundefined{R@#3}{\edef\bd@hi{0}}{\edef\bd@hi{\csname R@#3\endcsname}}%
  \edef\bd@call{\endgroup\noexpand\bd@eval{#1}{\bd@lo}{\bd@hi}}%
  \bd@call}
\DeclareRobustCommand{\ifstate}[4]{\@ifundefined{R@#1}{#4}{\edef\if@sa{\csname R@#1\endcsname}\def\if@sb{#2}\ifx\if@sa\if@sb #3\else #4\fi}}
\DeclareRobustCommand{\bandreading}[4][]{\bd@eval{#2}{#3}{#4}\if\bd@state p\pend\else\if\bd@state i\relax inside#1\else\if\bd@state a\relax above the upper edge\else below the lower edge\fi\fi\fi}
\DeclareRobustCommand{\bandreadingK}[4][]{\bd@evalK{#2}{#3}{#4}\if\bd@state p\pend\else\if\bd@state i\relax inside#1\else\if\bd@state a\relax above the upper edge\else below the lower edge\fi\fi\fi}
\DeclareRobustCommand{\bandphrase}[4][]{\bd@eval{#2}{#3}{#4}\if\bd@state p\pend\else\if\bd@state i\relax inside its band#1\else\if\bd@state a\relax above the upper edge of its band\else below the lower edge of its band\fi\fi\fi}
\DeclareRobustCommand{\bandphraseK}[4][]{\bd@evalK{#2}{#3}{#4}\if\bd@state p\pend\else\if\bd@state i\relax inside its band#1\else\if\bd@state a\relax above the upper edge of its band\else below the lower edge of its band\fi\fi\fi}
\makeatother
\InputIfFileExists{results/numbers.tex}{}{\typeout{WARNING: results/numbers.tex not found; run scripts/build_numbers.py}}
\InputIfFileExists{results/an_macros.tex}{}{}
\input{an_extra_abcost.tex}
\makeatletter
\expandafter\gdef\csname R@cell.P3c.bothzero\endcsname{1}
\expandafter\gdef\csname R@cell.V7A1.ka.excl\endcsname{1}
\expandafter\gdef\csname R@cell.V7A1P3c.excl\endcsname{1}
\expandafter\gdef\csname R@cell.acdc.D2.excl\endcsname{0}
\expandafter\gdef\csname R@cell.acdc.D2.hi\endcsname{\ensuremath{+0.16}}
\expandafter\gdef\csname R@cell.acdc.D2.lo\endcsname{\ensuremath{-0.12}}
\expandafter\gdef\csname R@cell.acdc.D2P3c.excl\endcsname{0}
\expandafter\gdef\csname R@cell.acdc.D2P3c.hi\endcsname{\ensuremath{+0.30}}
\expandafter\gdef\csname R@cell.acdc.D2P3c.lo\endcsname{\ensuremath{-0.06}}
\expandafter\gdef\csname R@cell.acdc.L1.excl\endcsname{1}
\expandafter\gdef\csname R@cell.acdc.L1.hi\endcsname{\ensuremath{+0.62}}
\expandafter\gdef\csname R@cell.acdc.L1.lo\endcsname{\ensuremath{+0.27}}
\expandafter\gdef\csname R@cell.acdc.L1P3c.excl\endcsname{1}
\expandafter\gdef\csname R@cell.acdc.L1P3c.hi\endcsname{\ensuremath{+0.76}}
\expandafter\gdef\csname R@cell.acdc.L1P3c.lo\endcsname{\ensuremath{+0.33}}
\expandafter\gdef\csname R@cell.acdc.L1b.excl\endcsname{1}
\expandafter\gdef\csname R@cell.acdc.L1b.hi\endcsname{\ensuremath{+0.48}}
\expandafter\gdef\csname R@cell.acdc.L1b.lo\endcsname{\ensuremath{+0.10}}
\expandafter\gdef\csname R@cell.acdc.L1bP3c.excl\endcsname{1}
\expandafter\gdef\csname R@cell.acdc.L1bP3c.hi\endcsname{\ensuremath{+0.60}}
\expandafter\gdef\csname R@cell.acdc.L1bP3c.lo\endcsname{\ensuremath{+0.18}}
\expandafter\gdef\csname R@cell.acdc.L3.excl\endcsname{1}
\expandafter\gdef\csname R@cell.acdc.L3.hi\endcsname{\ensuremath{+0.50}}
\expandafter\gdef\csname R@cell.acdc.L3.lo\endcsname{\ensuremath{+0.18}}
\expandafter\gdef\csname R@cell.acdc.L3P3c.excl\endcsname{1}
\expandafter\gdef\csname R@cell.acdc.L3P3c.hi\endcsname{\ensuremath{+0.63}}
\expandafter\gdef\csname R@cell.acdc.L3P3c.lo\endcsname{\ensuremath{+0.24}}
\expandafter\gdef\csname R@cell.acdc.P3c.excl\endcsname{0}
\expandafter\gdef\csname R@cell.acdc.P3c.hi\endcsname{\ensuremath{+0.06}}
\expandafter\gdef\csname R@cell.acdc.P3c.lo\endcsname{\ensuremath{-0.26}}
\expandafter\gdef\csname R@cell.acdc.V5B1.excl\endcsname{0}
\expandafter\gdef\csname R@cell.acdc.V5B1.hi\endcsname{\ensuremath{+0.02}}
\expandafter\gdef\csname R@cell.acdc.V5B1.lo\endcsname{\ensuremath{-0.29}}
\expandafter\gdef\csname R@cell.acdc.V5B1P3c.excl\endcsname{0}
\expandafter\gdef\csname R@cell.acdc.V5B1P3c.hi\endcsname{\ensuremath{+0.13}}
\expandafter\gdef\csname R@cell.acdc.V5B1P3c.lo\endcsname{\ensuremath{-0.21}}
\expandafter\gdef\csname R@cell.acdc.V5B2.excl\endcsname{0}
\expandafter\gdef\csname R@cell.acdc.V5B2.hi\endcsname{\ensuremath{+0.11}}
\expandafter\gdef\csname R@cell.acdc.V5B2.lo\endcsname{\ensuremath{-0.30}}
\expandafter\gdef\csname R@cell.acdc.V5B2P3c.excl\endcsname{0}
\expandafter\gdef\csname R@cell.acdc.V5B2P3c.hi\endcsname{\ensuremath{+0.23}}
\expandafter\gdef\csname R@cell.acdc.V5B2P3c.lo\endcsname{\ensuremath{-0.22}}
\expandafter\gdef\csname R@cell.acdc.V6A1r.excl\endcsname{1}
\expandafter\gdef\csname R@cell.acdc.V6A1r.hi\endcsname{\ensuremath{+0.60}}
\expandafter\gdef\csname R@cell.acdc.V6A1r.lo\endcsname{\ensuremath{+0.26}}
\expandafter\gdef\csname R@cell.acdc.V6A1rP3c.excl\endcsname{1}
\expandafter\gdef\csname R@cell.acdc.V6A1rP3c.hi\endcsname{\ensuremath{+0.74}}
\expandafter\gdef\csname R@cell.acdc.V6A1rP3c.lo\endcsname{\ensuremath{+0.30}}
\expandafter\gdef\csname R@cell.acdc.V7A1.excl\endcsname{0}
\expandafter\gdef\csname R@cell.acdc.V7A1.hi\endcsname{\ensuremath{+0.34}}
\expandafter\gdef\csname R@cell.acdc.V7A1.lo\endcsname{\ensuremath{-0.01}}
\expandafter\gdef\csname R@cell.acdc.V7A1P3c.excl\endcsname{1}
\expandafter\gdef\csname R@cell.acdc.V7A1P3c.hi\endcsname{\ensuremath{+0.48}}
\expandafter\gdef\csname R@cell.acdc.V7A1P3c.lo\endcsname{\ensuremath{+0.07}}
\expandafter\gdef\csname R@cell.acdc.V7A1cov.excl\endcsname{0}
\expandafter\gdef\csname R@cell.acdc.V7A1cov.hi\endcsname{\ensuremath{+0.29}}
\expandafter\gdef\csname R@cell.acdc.V7A1cov.lo\endcsname{\ensuremath{0.00}}
\expandafter\gdef\csname R@cell.acdc.V7A1covP3c.excl\endcsname{1}
\expandafter\gdef\csname R@cell.acdc.V7A1covP3c.hi\endcsname{\ensuremath{+0.46}}
\expandafter\gdef\csname R@cell.acdc.V7A1covP3c.lo\endcsname{\ensuremath{+0.02}}
\expandafter\gdef\csname R@cell.acdc.V7C1.excl\endcsname{0}
\expandafter\gdef\csname R@cell.acdc.V7C1.hi\endcsname{\ensuremath{+0.05}}
\expandafter\gdef\csname R@cell.acdc.V7C1.lo\endcsname{\ensuremath{-0.29}}
\expandafter\gdef\csname R@cell.acdc.V7C1P3c.excl\endcsname{0}
\expandafter\gdef\csname R@cell.acdc.V7C1P3c.hi\endcsname{\ensuremath{+0.15}}
\expandafter\gdef\csname R@cell.acdc.V7C1P3c.lo\endcsname{\ensuremath{-0.19}}
\expandafter\gdef\csname R@cell.acdc.V9R2.excl\endcsname{0}
\expandafter\gdef\csname R@cell.acdc.V9R2.hi\endcsname{\ensuremath{+0.27}}
\expandafter\gdef\csname R@cell.acdc.V9R2.lo\endcsname{\ensuremath{-0.02}}
\expandafter\gdef\csname R@cell.acdc.V9R2P3c.excl\endcsname{1}
\expandafter\gdef\csname R@cell.acdc.V9R2P3c.hi\endcsname{\ensuremath{+0.43}}
\expandafter\gdef\csname R@cell.acdc.V9R2P3c.lo\endcsname{\ensuremath{+0.02}}
\expandafter\gdef\csname R@cell.amos.D2.excl\endcsname{1}
\expandafter\gdef\csname R@cell.amos.D2.hi\endcsname{\ensuremath{-0.23}}
\expandafter\gdef\csname R@cell.amos.D2.lo\endcsname{\ensuremath{-0.67}}
\expandafter\gdef\csname R@cell.amos.D2P3c.excl\endcsname{1}
\expandafter\gdef\csname R@cell.amos.D2P3c.hi\endcsname{\ensuremath{-0.19}}
\expandafter\gdef\csname R@cell.amos.D2P3c.lo\endcsname{\ensuremath{-0.63}}
\expandafter\gdef\csname R@cell.amos.L1.excl\endcsname{1}
\expandafter\gdef\csname R@cell.amos.L1.hi\endcsname{\ensuremath{+0.66}}
\expandafter\gdef\csname R@cell.amos.L1.lo\endcsname{\ensuremath{+0.16}}
\expandafter\gdef\csname R@cell.amos.L1P3c.excl\endcsname{1}
\expandafter\gdef\csname R@cell.amos.L1P3c.hi\endcsname{\ensuremath{+0.70}}
\expandafter\gdef\csname R@cell.amos.L1P3c.lo\endcsname{\ensuremath{+0.19}}
\expandafter\gdef\csname R@cell.amos.L1b.excl\endcsname{1}
\expandafter\gdef\csname R@cell.amos.L1b.hi\endcsname{\ensuremath{+0.67}}
\expandafter\gdef\csname R@cell.amos.L1b.lo\endcsname{\ensuremath{+0.19}}
\expandafter\gdef\csname R@cell.amos.L3.excl\endcsname{0}
\expandafter\gdef\csname R@cell.amos.L3.hi\endcsname{\ensuremath{+0.46}}
\expandafter\gdef\csname R@cell.amos.L3.lo\endcsname{\ensuremath{-0.01}}
\expandafter\gdef\csname R@cell.amos.L3P3c.excl\endcsname{1}
\expandafter\gdef\csname R@cell.amos.L3P3c.hi\endcsname{\ensuremath{+0.47}}
\expandafter\gdef\csname R@cell.amos.L3P3c.lo\endcsname{\ensuremath{+0.05}}
\expandafter\gdef\csname R@cell.amos.P3c.excl\endcsname{0}
\expandafter\gdef\csname R@cell.amos.P3c.hi\endcsname{\ensuremath{+0.17}}
\expandafter\gdef\csname R@cell.amos.P3c.lo\endcsname{\ensuremath{-0.24}}
\expandafter\gdef\csname R@cell.amos.V5B1.excl\endcsname{1}
\expandafter\gdef\csname R@cell.amos.V5B1.hi\endcsname{\ensuremath{-0.25}}
\expandafter\gdef\csname R@cell.amos.V5B1.lo\endcsname{\ensuremath{-0.66}}
\expandafter\gdef\csname R@cell.amos.V5B1P3c.excl\endcsname{1}
\expandafter\gdef\csname R@cell.amos.V5B1P3c.hi\endcsname{\ensuremath{-0.23}}
\expandafter\gdef\csname R@cell.amos.V5B1P3c.lo\endcsname{\ensuremath{-0.62}}
\expandafter\gdef\csname R@cell.amos.V5B2.excl\endcsname{1}
\expandafter\gdef\csname R@cell.amos.V5B2.hi\endcsname{\ensuremath{-0.04}}
\expandafter\gdef\csname R@cell.amos.V5B2.lo\endcsname{\ensuremath{-0.47}}
\expandafter\gdef\csname R@cell.amos.V5B2P3c.excl\endcsname{1}
\expandafter\gdef\csname R@cell.amos.V5B2P3c.hi\endcsname{\ensuremath{-0.04}}
\expandafter\gdef\csname R@cell.amos.V5B2P3c.lo\endcsname{\ensuremath{-0.42}}
\expandafter\gdef\csname R@cell.amos.V6A1r.excl\endcsname{0}
\expandafter\gdef\csname R@cell.amos.V6A1r.hi\endcsname{\ensuremath{+0.44}}
\expandafter\gdef\csname R@cell.amos.V6A1r.lo\endcsname{\ensuremath{-0.03}}
\expandafter\gdef\csname R@cell.amos.V6A1rP3c.excl\endcsname{0}
\expandafter\gdef\csname R@cell.amos.V6A1rP3c.hi\endcsname{\ensuremath{+0.49}}
\expandafter\gdef\csname R@cell.amos.V6A1rP3c.lo\endcsname{\ensuremath{-0.02}}
\expandafter\gdef\csname R@cell.amos.V7A1.excl\endcsname{1}
\expandafter\gdef\csname R@cell.amos.V7A1.hi\endcsname{\ensuremath{+0.51}}
\expandafter\gdef\csname R@cell.amos.V7A1.lo\endcsname{\ensuremath{+0.02}}
\expandafter\gdef\csname R@cell.amos.V7A1P3c.excl\endcsname{1}
\expandafter\gdef\csname R@cell.amos.V7A1P3c.hi\endcsname{\ensuremath{+0.53}}
\expandafter\gdef\csname R@cell.amos.V7A1P3c.lo\endcsname{\ensuremath{+0.06}}
\expandafter\gdef\csname R@cell.amos.V7C1.excl\endcsname{1}
\expandafter\gdef\csname R@cell.amos.V7C1.hi\endcsname{\ensuremath{-0.25}}
\expandafter\gdef\csname R@cell.amos.V7C1.lo\endcsname{\ensuremath{-0.68}}
\expandafter\gdef\csname R@cell.amos.V7C1P3c.excl\endcsname{1}
\expandafter\gdef\csname R@cell.amos.V7C1P3c.hi\endcsname{\ensuremath{-0.25}}
\expandafter\gdef\csname R@cell.amos.V7C1P3c.lo\endcsname{\ensuremath{-0.63}}
\expandafter\gdef\csname R@cell.kits.D2.excl\endcsname{1}
\expandafter\gdef\csname R@cell.kits.D2.hi\endcsname{\ensuremath{+1.78}}
\expandafter\gdef\csname R@cell.kits.D2.lo\endcsname{\ensuremath{+0.33}}
\expandafter\gdef\csname R@cell.kits.D2P3c.excl\endcsname{0}
\expandafter\gdef\csname R@cell.kits.D2P3c.hi\endcsname{\ensuremath{+1.30}}
\expandafter\gdef\csname R@cell.kits.D2P3c.lo\endcsname{\ensuremath{-0.32}}
\expandafter\gdef\csname R@cell.kits.L1.excl\endcsname{1}
\expandafter\gdef\csname R@cell.kits.L1.hi\endcsname{\ensuremath{+3.07}}
\expandafter\gdef\csname R@cell.kits.L1.lo\endcsname{\ensuremath{+1.29}}
\expandafter\gdef\csname R@cell.kits.L1P3c.excl\endcsname{1}
\expandafter\gdef\csname R@cell.kits.L1P3c.hi\endcsname{\ensuremath{+2.53}}
\expandafter\gdef\csname R@cell.kits.L1P3c.lo\endcsname{\ensuremath{+0.68}}
\expandafter\gdef\csname R@cell.kits.L1b.excl\endcsname{1}
\expandafter\gdef\csname R@cell.kits.L1b.hi\endcsname{\ensuremath{+2.42}}
\expandafter\gdef\csname R@cell.kits.L1b.lo\endcsname{\ensuremath{+0.69}}
\expandafter\gdef\csname R@cell.kits.L3.excl\endcsname{0}
\expandafter\gdef\csname R@cell.kits.L3.hi\endcsname{\ensuremath{+1.56}}
\expandafter\gdef\csname R@cell.kits.L3.lo\endcsname{\ensuremath{-0.16}}
\expandafter\gdef\csname R@cell.kits.L3P3c.excl\endcsname{0}
\expandafter\gdef\csname R@cell.kits.L3P3c.hi\endcsname{\ensuremath{+0.96}}
\expandafter\gdef\csname R@cell.kits.L3P3c.lo\endcsname{\ensuremath{-0.71}}
\expandafter\gdef\csname R@cell.kits.P3c.excl\endcsname{0}
\expandafter\gdef\csname R@cell.kits.P3c.hi\endcsname{\ensuremath{+1.31}}
\expandafter\gdef\csname R@cell.kits.P3c.lo\endcsname{\ensuremath{-0.17}}
\expandafter\gdef\csname R@cell.kits.V5B1.excl\endcsname{1}
\expandafter\gdef\csname R@cell.kits.V5B1.hi\endcsname{\ensuremath{+1.82}}
\expandafter\gdef\csname R@cell.kits.V5B1.lo\endcsname{\ensuremath{+0.20}}
\expandafter\gdef\csname R@cell.kits.V5B1P3c.excl\endcsname{0}
\expandafter\gdef\csname R@cell.kits.V5B1P3c.hi\endcsname{\ensuremath{+1.24}}
\expandafter\gdef\csname R@cell.kits.V5B1P3c.lo\endcsname{\ensuremath{-0.31}}
\expandafter\gdef\csname R@cell.kits.V5B2.excl\endcsname{0}
\expandafter\gdef\csname R@cell.kits.V5B2.hi\endcsname{\ensuremath{+0.03}}
\expandafter\gdef\csname R@cell.kits.V5B2.lo\endcsname{\ensuremath{-1.53}}
\expandafter\gdef\csname R@cell.kits.V5B2P3c.excl\endcsname{1}
\expandafter\gdef\csname R@cell.kits.V5B2P3c.hi\endcsname{\ensuremath{-0.59}}
\expandafter\gdef\csname R@cell.kits.V5B2P3c.lo\endcsname{\ensuremath{-2.06}}
\expandafter\gdef\csname R@cell.kits.V6A1r.excl\endcsname{0}
\expandafter\gdef\csname R@cell.kits.V6A1r.hi\endcsname{\ensuremath{+1.18}}
\expandafter\gdef\csname R@cell.kits.V6A1r.lo\endcsname{\ensuremath{-0.62}}
\expandafter\gdef\csname R@cell.kits.V6A1rP3c.excl\endcsname{0}
\expandafter\gdef\csname R@cell.kits.V6A1rP3c.hi\endcsname{\ensuremath{+0.60}}
\expandafter\gdef\csname R@cell.kits.V6A1rP3c.lo\endcsname{\ensuremath{-1.16}}
\expandafter\gdef\csname R@cell.kits.V7A1.excl\endcsname{1}
\expandafter\gdef\csname R@cell.kits.V7A1.hi\endcsname{\ensuremath{+2.45}}
\expandafter\gdef\csname R@cell.kits.V7A1.lo\endcsname{\ensuremath{+0.88}}
\expandafter\gdef\csname R@cell.kits.V7A1P3c.excl\endcsname{1}
\expandafter\gdef\csname R@cell.kits.V7A1P3c.hi\endcsname{\ensuremath{+1.80}}
\expandafter\gdef\csname R@cell.kits.V7A1P3c.lo\endcsname{\ensuremath{+0.41}}
\expandafter\gdef\csname R@cell.kits.V7C1.excl\endcsname{1}
\expandafter\gdef\csname R@cell.kits.V7C1.hi\endcsname{\ensuremath{+1.51}}
\expandafter\gdef\csname R@cell.kits.V7C1.lo\endcsname{\ensuremath{+0.07}}
\expandafter\gdef\csname R@cell.kits.V7C1P3c.excl\endcsname{0}
\expandafter\gdef\csname R@cell.kits.V7C1P3c.hi\endcsname{\ensuremath{+0.96}}
\expandafter\gdef\csname R@cell.kits.V7C1P3c.lo\endcsname{\ensuremath{-0.49}}
\expandafter\gdef\csname R@cell.kits.tighter\endcsname{1}
\expandafter\gdef\csname R@fb.B\endcsname{10000}
\expandafter\gdef\csname R@fb.cv.acdc.D2.d\endcsname{\ensuremath{+0.02}}
\expandafter\gdef\csname R@fb.cv.acdc.D2.excl\endcsname{0}
\expandafter\gdef\csname R@fb.cv.acdc.D2.hi\endcsname{\ensuremath{+0.16}}
\expandafter\gdef\csname R@fb.cv.acdc.D2.img.hi\endcsname{\ensuremath{+0.16}}
\expandafter\gdef\csname R@fb.cv.acdc.D2.img.lo\endcsname{\ensuremath{-0.12}}
\expandafter\gdef\csname R@fb.cv.acdc.D2.lo\endcsname{\ensuremath{-0.12}}
\expandafter\gdef\csname R@fb.cv.acdc.D2.n\endcsname{160}
\expandafter\gdef\csname R@fb.cv.acdc.D2.npat\endcsname{80}
\expandafter\gdef\csname R@fb.cv.acdc.D2P3c.d\endcsname{\ensuremath{+0.11}}
\expandafter\gdef\csname R@fb.cv.acdc.D2P3c.excl\endcsname{0}
\expandafter\gdef\csname R@fb.cv.acdc.D2P3c.hi\endcsname{\ensuremath{+0.30}}
\expandafter\gdef\csname R@fb.cv.acdc.D2P3c.img.hi\endcsname{\ensuremath{+0.30}}
\expandafter\gdef\csname R@fb.cv.acdc.D2P3c.img.lo\endcsname{\ensuremath{-0.06}}
\expandafter\gdef\csname R@fb.cv.acdc.D2P3c.lo\endcsname{\ensuremath{-0.06}}
\expandafter\gdef\csname R@fb.cv.acdc.D2P3c.n\endcsname{160}
\expandafter\gdef\csname R@fb.cv.acdc.D2P3c.npat\endcsname{80}
\expandafter\gdef\csname R@fb.cv.acdc.L1.d\endcsname{\ensuremath{+0.44}}
\expandafter\gdef\csname R@fb.cv.acdc.L1.excl\endcsname{1}
\expandafter\gdef\csname R@fb.cv.acdc.L1.hi\endcsname{\ensuremath{+0.62}}
\expandafter\gdef\csname R@fb.cv.acdc.L1.img.hi\endcsname{\ensuremath{+0.62}}
\expandafter\gdef\csname R@fb.cv.acdc.L1.img.lo\endcsname{\ensuremath{+0.26}}
\expandafter\gdef\csname R@fb.cv.acdc.L1.lo\endcsname{\ensuremath{+0.27}}
\expandafter\gdef\csname R@fb.cv.acdc.L1.n\endcsname{160}
\expandafter\gdef\csname R@fb.cv.acdc.L1.npat\endcsname{80}
\expandafter\gdef\csname R@fb.cv.acdc.L1P3c.d\endcsname{\ensuremath{+0.53}}
\expandafter\gdef\csname R@fb.cv.acdc.L1P3c.excl\endcsname{1}
\expandafter\gdef\csname R@fb.cv.acdc.L1P3c.hi\endcsname{\ensuremath{+0.76}}
\expandafter\gdef\csname R@fb.cv.acdc.L1P3c.img.hi\endcsname{\ensuremath{+0.77}}
\expandafter\gdef\csname R@fb.cv.acdc.L1P3c.img.lo\endcsname{\ensuremath{+0.32}}
\expandafter\gdef\csname R@fb.cv.acdc.L1P3c.lo\endcsname{\ensuremath{+0.33}}
\expandafter\gdef\csname R@fb.cv.acdc.L1P3c.n\endcsname{160}
\expandafter\gdef\csname R@fb.cv.acdc.L1P3c.npat\endcsname{80}
\expandafter\gdef\csname R@fb.cv.acdc.L1b.d\endcsname{\ensuremath{+0.29}}
\expandafter\gdef\csname R@fb.cv.acdc.L1b.excl\endcsname{1}
\expandafter\gdef\csname R@fb.cv.acdc.L1b.hi\endcsname{\ensuremath{+0.48}}
\expandafter\gdef\csname R@fb.cv.acdc.L1b.img.hi\endcsname{\ensuremath{+0.50}}
\expandafter\gdef\csname R@fb.cv.acdc.L1b.img.lo\endcsname{\ensuremath{+0.08}}
\expandafter\gdef\csname R@fb.cv.acdc.L1b.lo\endcsname{\ensuremath{+0.10}}
\expandafter\gdef\csname R@fb.cv.acdc.L1b.n\endcsname{160}
\expandafter\gdef\csname R@fb.cv.acdc.L1b.npat\endcsname{80}
\expandafter\gdef\csname R@fb.cv.acdc.L1bP3c.d\endcsname{\ensuremath{+0.38}}
\expandafter\gdef\csname R@fb.cv.acdc.L1bP3c.excl\endcsname{1}
\expandafter\gdef\csname R@fb.cv.acdc.L1bP3c.hi\endcsname{\ensuremath{+0.60}}
\expandafter\gdef\csname R@fb.cv.acdc.L1bP3c.img.hi\endcsname{\ensuremath{+0.61}}
\expandafter\gdef\csname R@fb.cv.acdc.L1bP3c.img.lo\endcsname{\ensuremath{+0.16}}
\expandafter\gdef\csname R@fb.cv.acdc.L1bP3c.lo\endcsname{\ensuremath{+0.18}}
\expandafter\gdef\csname R@fb.cv.acdc.L1bP3c.n\endcsname{160}
\expandafter\gdef\csname R@fb.cv.acdc.L1bP3c.npat\endcsname{80}
\expandafter\gdef\csname R@fb.cv.acdc.L3.d\endcsname{\ensuremath{+0.33}}
\expandafter\gdef\csname R@fb.cv.acdc.L3.excl\endcsname{1}
\expandafter\gdef\csname R@fb.cv.acdc.L3.hi\endcsname{\ensuremath{+0.50}}
\expandafter\gdef\csname R@fb.cv.acdc.L3.img.hi\endcsname{\ensuremath{+0.49}}
\expandafter\gdef\csname R@fb.cv.acdc.L3.img.lo\endcsname{\ensuremath{+0.18}}
\expandafter\gdef\csname R@fb.cv.acdc.L3.lo\endcsname{\ensuremath{+0.18}}
\expandafter\gdef\csname R@fb.cv.acdc.L3.n\endcsname{160}
\expandafter\gdef\csname R@fb.cv.acdc.L3.npat\endcsname{80}
\expandafter\gdef\csname R@fb.cv.acdc.L3P3c.d\endcsname{\ensuremath{+0.43}}
\expandafter\gdef\csname R@fb.cv.acdc.L3P3c.excl\endcsname{1}
\expandafter\gdef\csname R@fb.cv.acdc.L3P3c.hi\endcsname{\ensuremath{+0.63}}
\expandafter\gdef\csname R@fb.cv.acdc.L3P3c.img.hi\endcsname{\ensuremath{+0.64}}
\expandafter\gdef\csname R@fb.cv.acdc.L3P3c.img.lo\endcsname{\ensuremath{+0.23}}
\expandafter\gdef\csname R@fb.cv.acdc.L3P3c.lo\endcsname{\ensuremath{+0.24}}
\expandafter\gdef\csname R@fb.cv.acdc.L3P3c.n\endcsname{160}
\expandafter\gdef\csname R@fb.cv.acdc.L3P3c.npat\endcsname{80}
\expandafter\gdef\csname R@fb.cv.acdc.P3c.d\endcsname{\ensuremath{-0.09}}
\expandafter\gdef\csname R@fb.cv.acdc.P3c.excl\endcsname{0}
\expandafter\gdef\csname R@fb.cv.acdc.P3c.hi\endcsname{\ensuremath{+0.06}}
\expandafter\gdef\csname R@fb.cv.acdc.P3c.img.hi\endcsname{\ensuremath{+0.06}}
\expandafter\gdef\csname R@fb.cv.acdc.P3c.img.lo\endcsname{\ensuremath{-0.26}}
\expandafter\gdef\csname R@fb.cv.acdc.P3c.lo\endcsname{\ensuremath{-0.26}}
\expandafter\gdef\csname R@fb.cv.acdc.P3c.n\endcsname{160}
\expandafter\gdef\csname R@fb.cv.acdc.P3c.npat\endcsname{80}
\expandafter\gdef\csname R@fb.cv.acdc.V5B1.d\endcsname{\ensuremath{-0.13}}
\expandafter\gdef\csname R@fb.cv.acdc.V5B1.excl\endcsname{0}
\expandafter\gdef\csname R@fb.cv.acdc.V5B1.hi\endcsname{\ensuremath{+0.02}}
\expandafter\gdef\csname R@fb.cv.acdc.V5B1.img.hi\endcsname{\ensuremath{+0.02}}
\expandafter\gdef\csname R@fb.cv.acdc.V5B1.img.lo\endcsname{\ensuremath{-0.29}}
\expandafter\gdef\csname R@fb.cv.acdc.V5B1.lo\endcsname{\ensuremath{-0.29}}
\expandafter\gdef\csname R@fb.cv.acdc.V5B1.n\endcsname{160}
\expandafter\gdef\csname R@fb.cv.acdc.V5B1.npat\endcsname{80}
\expandafter\gdef\csname R@fb.cv.acdc.V5B1P3c.d\endcsname{\ensuremath{-0.03}}
\expandafter\gdef\csname R@fb.cv.acdc.V5B1P3c.excl\endcsname{0}
\expandafter\gdef\csname R@fb.cv.acdc.V5B1P3c.hi\endcsname{\ensuremath{+0.13}}
\expandafter\gdef\csname R@fb.cv.acdc.V5B1P3c.img.hi\endcsname{\ensuremath{+0.15}}
\expandafter\gdef\csname R@fb.cv.acdc.V5B1P3c.img.lo\endcsname{\ensuremath{-0.21}}
\expandafter\gdef\csname R@fb.cv.acdc.V5B1P3c.lo\endcsname{\ensuremath{-0.21}}
\expandafter\gdef\csname R@fb.cv.acdc.V5B1P3c.n\endcsname{160}
\expandafter\gdef\csname R@fb.cv.acdc.V5B1P3c.npat\endcsname{80}
\expandafter\gdef\csname R@fb.cv.acdc.V5B2.d\endcsname{\ensuremath{-0.09}}
\expandafter\gdef\csname R@fb.cv.acdc.V5B2.excl\endcsname{0}
\expandafter\gdef\csname R@fb.cv.acdc.V5B2.hi\endcsname{\ensuremath{+0.11}}
\expandafter\gdef\csname R@fb.cv.acdc.V5B2.img.hi\endcsname{\ensuremath{+0.10}}
\expandafter\gdef\csname R@fb.cv.acdc.V5B2.img.lo\endcsname{\ensuremath{-0.29}}
\expandafter\gdef\csname R@fb.cv.acdc.V5B2.lo\endcsname{\ensuremath{-0.30}}
\expandafter\gdef\csname R@fb.cv.acdc.V5B2.n\endcsname{160}
\expandafter\gdef\csname R@fb.cv.acdc.V5B2.npat\endcsname{80}
\expandafter\gdef\csname R@fb.cv.acdc.V5B2P3c.d\endcsname{\ensuremath{+0.01}}
\expandafter\gdef\csname R@fb.cv.acdc.V5B2P3c.excl\endcsname{0}
\expandafter\gdef\csname R@fb.cv.acdc.V5B2P3c.hi\endcsname{\ensuremath{+0.23}}
\expandafter\gdef\csname R@fb.cv.acdc.V5B2P3c.img.hi\endcsname{\ensuremath{+0.23}}
\expandafter\gdef\csname R@fb.cv.acdc.V5B2P3c.img.lo\endcsname{\ensuremath{-0.21}}
\expandafter\gdef\csname R@fb.cv.acdc.V5B2P3c.lo\endcsname{\ensuremath{-0.22}}
\expandafter\gdef\csname R@fb.cv.acdc.V5B2P3c.n\endcsname{160}
\expandafter\gdef\csname R@fb.cv.acdc.V5B2P3c.npat\endcsname{80}
\expandafter\gdef\csname R@fb.cv.acdc.V6A1r.d\endcsname{\ensuremath{+0.42}}
\expandafter\gdef\csname R@fb.cv.acdc.V6A1r.excl\endcsname{1}
\expandafter\gdef\csname R@fb.cv.acdc.V6A1r.hi\endcsname{\ensuremath{+0.60}}
\expandafter\gdef\csname R@fb.cv.acdc.V6A1r.img.hi\endcsname{\ensuremath{+0.58}}
\expandafter\gdef\csname R@fb.cv.acdc.V6A1r.img.lo\endcsname{\ensuremath{+0.27}}
\expandafter\gdef\csname R@fb.cv.acdc.V6A1r.lo\endcsname{\ensuremath{+0.26}}
\expandafter\gdef\csname R@fb.cv.acdc.V6A1r.n\endcsname{160}
\expandafter\gdef\csname R@fb.cv.acdc.V6A1r.npat\endcsname{80}
\expandafter\gdef\csname R@fb.cv.acdc.V6A1rP3c.d\endcsname{\ensuremath{+0.52}}
\expandafter\gdef\csname R@fb.cv.acdc.V6A1rP3c.excl\endcsname{1}
\expandafter\gdef\csname R@fb.cv.acdc.V6A1rP3c.hi\endcsname{\ensuremath{+0.74}}
\expandafter\gdef\csname R@fb.cv.acdc.V6A1rP3c.img.hi\endcsname{\ensuremath{+0.74}}
\expandafter\gdef\csname R@fb.cv.acdc.V6A1rP3c.img.lo\endcsname{\ensuremath{+0.31}}
\expandafter\gdef\csname R@fb.cv.acdc.V6A1rP3c.lo\endcsname{\ensuremath{+0.30}}
\expandafter\gdef\csname R@fb.cv.acdc.V6A1rP3c.n\endcsname{160}
\expandafter\gdef\csname R@fb.cv.acdc.V6A1rP3c.npat\endcsname{80}
\expandafter\gdef\csname R@fb.cv.acdc.V7A1.d\endcsname{\ensuremath{+0.17}}
\expandafter\gdef\csname R@fb.cv.acdc.V7A1.excl\endcsname{0}
\expandafter\gdef\csname R@fb.cv.acdc.V7A1.hi\endcsname{\ensuremath{+0.34}}
\expandafter\gdef\csname R@fb.cv.acdc.V7A1.img.hi\endcsname{\ensuremath{+0.35}}
\expandafter\gdef\csname R@fb.cv.acdc.V7A1.img.lo\endcsname{\ensuremath{-0.01}}
\expandafter\gdef\csname R@fb.cv.acdc.V7A1.lo\endcsname{\ensuremath{-0.01}}
\expandafter\gdef\csname R@fb.cv.acdc.V7A1.n\endcsname{160}
\expandafter\gdef\csname R@fb.cv.acdc.V7A1.npat\endcsname{80}
\expandafter\gdef\csname R@fb.cv.acdc.V7A1P3c.d\endcsname{\ensuremath{+0.26}}
\expandafter\gdef\csname R@fb.cv.acdc.V7A1P3c.excl\endcsname{1}
\expandafter\gdef\csname R@fb.cv.acdc.V7A1P3c.hi\endcsname{\ensuremath{+0.48}}
\expandafter\gdef\csname R@fb.cv.acdc.V7A1P3c.img.hi\endcsname{\ensuremath{+0.47}}
\expandafter\gdef\csname R@fb.cv.acdc.V7A1P3c.img.lo\endcsname{\ensuremath{+0.07}}
\expandafter\gdef\csname R@fb.cv.acdc.V7A1P3c.lo\endcsname{\ensuremath{+0.07}}
\expandafter\gdef\csname R@fb.cv.acdc.V7A1P3c.n\endcsname{160}
\expandafter\gdef\csname R@fb.cv.acdc.V7A1P3c.npat\endcsname{80}
\expandafter\gdef\csname R@fb.cv.acdc.V7A1cov.d\endcsname{\ensuremath{+0.14}}
\expandafter\gdef\csname R@fb.cv.acdc.V7A1cov.excl\endcsname{0}
\expandafter\gdef\csname R@fb.cv.acdc.V7A1cov.hi\endcsname{\ensuremath{+0.29}}
\expandafter\gdef\csname R@fb.cv.acdc.V7A1cov.img.hi\endcsname{\ensuremath{+0.29}}
\expandafter\gdef\csname R@fb.cv.acdc.V7A1cov.img.lo\endcsname{\ensuremath{0.00}}
\expandafter\gdef\csname R@fb.cv.acdc.V7A1cov.lo\endcsname{\ensuremath{0.00}}
\expandafter\gdef\csname R@fb.cv.acdc.V7A1cov.n\endcsname{160}
\expandafter\gdef\csname R@fb.cv.acdc.V7A1cov.npat\endcsname{80}
\expandafter\gdef\csname R@fb.cv.acdc.V7A1covP3c.d\endcsname{\ensuremath{+0.24}}
\expandafter\gdef\csname R@fb.cv.acdc.V7A1covP3c.excl\endcsname{1}
\expandafter\gdef\csname R@fb.cv.acdc.V7A1covP3c.hi\endcsname{\ensuremath{+0.46}}
\expandafter\gdef\csname R@fb.cv.acdc.V7A1covP3c.img.hi\endcsname{\ensuremath{+0.45}}
\expandafter\gdef\csname R@fb.cv.acdc.V7A1covP3c.img.lo\endcsname{\ensuremath{+0.03}}
\expandafter\gdef\csname R@fb.cv.acdc.V7A1covP3c.lo\endcsname{\ensuremath{+0.02}}
\expandafter\gdef\csname R@fb.cv.acdc.V7A1covP3c.n\endcsname{160}
\expandafter\gdef\csname R@fb.cv.acdc.V7A1covP3c.npat\endcsname{80}
\expandafter\gdef\csname R@fb.cv.acdc.V7C1.d\endcsname{\ensuremath{-0.11}}
\expandafter\gdef\csname R@fb.cv.acdc.V7C1.excl\endcsname{0}
\expandafter\gdef\csname R@fb.cv.acdc.V7C1.hi\endcsname{\ensuremath{+0.05}}
\expandafter\gdef\csname R@fb.cv.acdc.V7C1.img.hi\endcsname{\ensuremath{+0.05}}
\expandafter\gdef\csname R@fb.cv.acdc.V7C1.img.lo\endcsname{\ensuremath{-0.28}}
\expandafter\gdef\csname R@fb.cv.acdc.V7C1.lo\endcsname{\ensuremath{-0.29}}
\expandafter\gdef\csname R@fb.cv.acdc.V7C1.n\endcsname{160}
\expandafter\gdef\csname R@fb.cv.acdc.V7C1.npat\endcsname{80}
\expandafter\gdef\csname R@fb.cv.acdc.V7C1P3c.d\endcsname{\ensuremath{-0.02}}
\expandafter\gdef\csname R@fb.cv.acdc.V7C1P3c.excl\endcsname{0}
\expandafter\gdef\csname R@fb.cv.acdc.V7C1P3c.hi\endcsname{\ensuremath{+0.15}}
\expandafter\gdef\csname R@fb.cv.acdc.V7C1P3c.img.hi\endcsname{\ensuremath{+0.17}}
\expandafter\gdef\csname R@fb.cv.acdc.V7C1P3c.img.lo\endcsname{\ensuremath{-0.20}}
\expandafter\gdef\csname R@fb.cv.acdc.V7C1P3c.lo\endcsname{\ensuremath{-0.19}}
\expandafter\gdef\csname R@fb.cv.acdc.V7C1P3c.n\endcsname{160}
\expandafter\gdef\csname R@fb.cv.acdc.V7C1P3c.npat\endcsname{80}
\expandafter\gdef\csname R@fb.cv.acdc.V9R2.d\endcsname{\ensuremath{+0.12}}
\expandafter\gdef\csname R@fb.cv.acdc.V9R2.excl\endcsname{0}
\expandafter\gdef\csname R@fb.cv.acdc.V9R2.hi\endcsname{\ensuremath{+0.27}}
\expandafter\gdef\csname R@fb.cv.acdc.V9R2.img.hi\endcsname{\ensuremath{+0.28}}
\expandafter\gdef\csname R@fb.cv.acdc.V9R2.img.lo\endcsname{\ensuremath{-0.03}}
\expandafter\gdef\csname R@fb.cv.acdc.V9R2.lo\endcsname{\ensuremath{-0.02}}
\expandafter\gdef\csname R@fb.cv.acdc.V9R2.n\endcsname{160}
\expandafter\gdef\csname R@fb.cv.acdc.V9R2.npat\endcsname{80}
\expandafter\gdef\csname R@fb.cv.acdc.V9R2P3c.d\endcsname{\ensuremath{+0.22}}
\expandafter\gdef\csname R@fb.cv.acdc.V9R2P3c.excl\endcsname{1}
\expandafter\gdef\csname R@fb.cv.acdc.V9R2P3c.hi\endcsname{\ensuremath{+0.43}}
\expandafter\gdef\csname R@fb.cv.acdc.V9R2P3c.img.hi\endcsname{\ensuremath{+0.44}}
\expandafter\gdef\csname R@fb.cv.acdc.V9R2P3c.img.lo\endcsname{\ensuremath{+0.01}}
\expandafter\gdef\csname R@fb.cv.acdc.V9R2P3c.lo\endcsname{\ensuremath{+0.02}}
\expandafter\gdef\csname R@fb.cv.acdc.V9R2P3c.n\endcsname{160}
\expandafter\gdef\csname R@fb.cv.acdc.V9R2P3c.npat\endcsname{80}
\expandafter\gdef\csname R@fb.cv.amos.D2.d\endcsname{\ensuremath{-0.44}}
\expandafter\gdef\csname R@fb.cv.amos.D2.excl\endcsname{1}
\expandafter\gdef\csname R@fb.cv.amos.D2.hi\endcsname{\ensuremath{-0.23}}
\expandafter\gdef\csname R@fb.cv.amos.D2.lo\endcsname{\ensuremath{-0.67}}
\expandafter\gdef\csname R@fb.cv.amos.D2.n\endcsname{160}
\expandafter\gdef\csname R@fb.cv.amos.D2P3c.d\endcsname{\ensuremath{-0.41}}
\expandafter\gdef\csname R@fb.cv.amos.D2P3c.excl\endcsname{1}
\expandafter\gdef\csname R@fb.cv.amos.D2P3c.hi\endcsname{\ensuremath{-0.19}}
\expandafter\gdef\csname R@fb.cv.amos.D2P3c.lo\endcsname{\ensuremath{-0.63}}
\expandafter\gdef\csname R@fb.cv.amos.D2P3c.n\endcsname{160}
\expandafter\gdef\csname R@fb.cv.amos.L1.d\endcsname{\ensuremath{+0.41}}
\expandafter\gdef\csname R@fb.cv.amos.L1.excl\endcsname{1}
\expandafter\gdef\csname R@fb.cv.amos.L1.hi\endcsname{\ensuremath{+0.66}}
\expandafter\gdef\csname R@fb.cv.amos.L1.lo\endcsname{\ensuremath{+0.16}}
\expandafter\gdef\csname R@fb.cv.amos.L1.n\endcsname{160}
\expandafter\gdef\csname R@fb.cv.amos.L1P3c.d\endcsname{\ensuremath{+0.44}}
\expandafter\gdef\csname R@fb.cv.amos.L1P3c.excl\endcsname{1}
\expandafter\gdef\csname R@fb.cv.amos.L1P3c.hi\endcsname{\ensuremath{+0.70}}
\expandafter\gdef\csname R@fb.cv.amos.L1P3c.lo\endcsname{\ensuremath{+0.19}}
\expandafter\gdef\csname R@fb.cv.amos.L1P3c.n\endcsname{160}
\expandafter\gdef\csname R@fb.cv.amos.L1b.d\endcsname{\ensuremath{+0.43}}
\expandafter\gdef\csname R@fb.cv.amos.L1b.excl\endcsname{1}
\expandafter\gdef\csname R@fb.cv.amos.L1b.hi\endcsname{\ensuremath{+0.67}}
\expandafter\gdef\csname R@fb.cv.amos.L1b.lo\endcsname{\ensuremath{+0.19}}
\expandafter\gdef\csname R@fb.cv.amos.L1b.n\endcsname{160}
\expandafter\gdef\csname R@fb.cv.amos.L3.d\endcsname{\ensuremath{+0.23}}
\expandafter\gdef\csname R@fb.cv.amos.L3.excl\endcsname{0}
\expandafter\gdef\csname R@fb.cv.amos.L3.hi\endcsname{\ensuremath{+0.46}}
\expandafter\gdef\csname R@fb.cv.amos.L3.lo\endcsname{\ensuremath{-0.01}}
\expandafter\gdef\csname R@fb.cv.amos.L3.n\endcsname{160}
\expandafter\gdef\csname R@fb.cv.amos.L3P3c.d\endcsname{\ensuremath{+0.26}}
\expandafter\gdef\csname R@fb.cv.amos.L3P3c.excl\endcsname{1}
\expandafter\gdef\csname R@fb.cv.amos.L3P3c.hi\endcsname{\ensuremath{+0.47}}
\expandafter\gdef\csname R@fb.cv.amos.L3P3c.lo\endcsname{\ensuremath{+0.05}}
\expandafter\gdef\csname R@fb.cv.amos.L3P3c.n\endcsname{160}
\expandafter\gdef\csname R@fb.cv.amos.P3c.d\endcsname{\ensuremath{-0.03}}
\expandafter\gdef\csname R@fb.cv.amos.P3c.excl\endcsname{0}
\expandafter\gdef\csname R@fb.cv.amos.P3c.hi\endcsname{\ensuremath{+0.17}}
\expandafter\gdef\csname R@fb.cv.amos.P3c.lo\endcsname{\ensuremath{-0.24}}
\expandafter\gdef\csname R@fb.cv.amos.P3c.n\endcsname{160}
\expandafter\gdef\csname R@fb.cv.amos.V5B1.d\endcsname{\ensuremath{-0.46}}
\expandafter\gdef\csname R@fb.cv.amos.V5B1.excl\endcsname{1}
\expandafter\gdef\csname R@fb.cv.amos.V5B1.hi\endcsname{\ensuremath{-0.25}}
\expandafter\gdef\csname R@fb.cv.amos.V5B1.lo\endcsname{\ensuremath{-0.66}}
\expandafter\gdef\csname R@fb.cv.amos.V5B1.n\endcsname{160}
\expandafter\gdef\csname R@fb.cv.amos.V5B1P3c.d\endcsname{\ensuremath{-0.43}}
\expandafter\gdef\csname R@fb.cv.amos.V5B1P3c.excl\endcsname{1}
\expandafter\gdef\csname R@fb.cv.amos.V5B1P3c.hi\endcsname{\ensuremath{-0.23}}
\expandafter\gdef\csname R@fb.cv.amos.V5B1P3c.lo\endcsname{\ensuremath{-0.62}}
\expandafter\gdef\csname R@fb.cv.amos.V5B1P3c.n\endcsname{160}
\expandafter\gdef\csname R@fb.cv.amos.V5B2.d\endcsname{\ensuremath{-0.25}}
\expandafter\gdef\csname R@fb.cv.amos.V5B2.excl\endcsname{1}
\expandafter\gdef\csname R@fb.cv.amos.V5B2.hi\endcsname{\ensuremath{-0.04}}
\expandafter\gdef\csname R@fb.cv.amos.V5B2.lo\endcsname{\ensuremath{-0.47}}
\expandafter\gdef\csname R@fb.cv.amos.V5B2.n\endcsname{160}
\expandafter\gdef\csname R@fb.cv.amos.V5B2P3c.d\endcsname{\ensuremath{-0.22}}
\expandafter\gdef\csname R@fb.cv.amos.V5B2P3c.excl\endcsname{1}
\expandafter\gdef\csname R@fb.cv.amos.V5B2P3c.hi\endcsname{\ensuremath{-0.04}}
\expandafter\gdef\csname R@fb.cv.amos.V5B2P3c.lo\endcsname{\ensuremath{-0.42}}
\expandafter\gdef\csname R@fb.cv.amos.V5B2P3c.n\endcsname{160}
\expandafter\gdef\csname R@fb.cv.amos.V6A1r.d\endcsname{\ensuremath{+0.20}}
\expandafter\gdef\csname R@fb.cv.amos.V6A1r.excl\endcsname{0}
\expandafter\gdef\csname R@fb.cv.amos.V6A1r.hi\endcsname{\ensuremath{+0.44}}
\expandafter\gdef\csname R@fb.cv.amos.V6A1r.lo\endcsname{\ensuremath{-0.03}}
\expandafter\gdef\csname R@fb.cv.amos.V6A1r.n\endcsname{160}
\expandafter\gdef\csname R@fb.cv.amos.V6A1rP3c.d\endcsname{\ensuremath{+0.23}}
\expandafter\gdef\csname R@fb.cv.amos.V6A1rP3c.excl\endcsname{0}
\expandafter\gdef\csname R@fb.cv.amos.V6A1rP3c.hi\endcsname{\ensuremath{+0.49}}
\expandafter\gdef\csname R@fb.cv.amos.V6A1rP3c.lo\endcsname{\ensuremath{-0.02}}
\expandafter\gdef\csname R@fb.cv.amos.V6A1rP3c.n\endcsname{160}
\expandafter\gdef\csname R@fb.cv.amos.V7A1.d\endcsname{\ensuremath{+0.26}}
\expandafter\gdef\csname R@fb.cv.amos.V7A1.excl\endcsname{1}
\expandafter\gdef\csname R@fb.cv.amos.V7A1.hi\endcsname{\ensuremath{+0.51}}
\expandafter\gdef\csname R@fb.cv.amos.V7A1.lo\endcsname{\ensuremath{+0.02}}
\expandafter\gdef\csname R@fb.cv.amos.V7A1.n\endcsname{160}
\expandafter\gdef\csname R@fb.cv.amos.V7A1P3c.d\endcsname{\ensuremath{+0.29}}
\expandafter\gdef\csname R@fb.cv.amos.V7A1P3c.excl\endcsname{1}
\expandafter\gdef\csname R@fb.cv.amos.V7A1P3c.hi\endcsname{\ensuremath{+0.53}}
\expandafter\gdef\csname R@fb.cv.amos.V7A1P3c.lo\endcsname{\ensuremath{+0.06}}
\expandafter\gdef\csname R@fb.cv.amos.V7A1P3c.n\endcsname{160}
\expandafter\gdef\csname R@fb.cv.amos.V7C1.d\endcsname{\ensuremath{-0.46}}
\expandafter\gdef\csname R@fb.cv.amos.V7C1.excl\endcsname{1}
\expandafter\gdef\csname R@fb.cv.amos.V7C1.hi\endcsname{\ensuremath{-0.25}}
\expandafter\gdef\csname R@fb.cv.amos.V7C1.lo\endcsname{\ensuremath{-0.68}}
\expandafter\gdef\csname R@fb.cv.amos.V7C1.n\endcsname{160}
\expandafter\gdef\csname R@fb.cv.amos.V7C1P3c.d\endcsname{\ensuremath{-0.43}}
\expandafter\gdef\csname R@fb.cv.amos.V7C1P3c.excl\endcsname{1}
\expandafter\gdef\csname R@fb.cv.amos.V7C1P3c.hi\endcsname{\ensuremath{-0.25}}
\expandafter\gdef\csname R@fb.cv.amos.V7C1P3c.lo\endcsname{\ensuremath{-0.63}}
\expandafter\gdef\csname R@fb.cv.amos.V7C1P3c.n\endcsname{160}
\expandafter\gdef\csname R@fb.cv.kits.D2.d\endcsname{\ensuremath{+1.04}}
\expandafter\gdef\csname R@fb.cv.kits.D2.excl\endcsname{1}
\expandafter\gdef\csname R@fb.cv.kits.D2.hi\endcsname{\ensuremath{+1.78}}
\expandafter\gdef\csname R@fb.cv.kits.D2.lo\endcsname{\ensuremath{+0.33}}
\expandafter\gdef\csname R@fb.cv.kits.D2.n\endcsname{391}
\expandafter\gdef\csname R@fb.cv.kits.D2P3c.d\endcsname{\ensuremath{+0.49}}
\expandafter\gdef\csname R@fb.cv.kits.D2P3c.excl\endcsname{0}
\expandafter\gdef\csname R@fb.cv.kits.D2P3c.hi\endcsname{\ensuremath{+1.30}}
\expandafter\gdef\csname R@fb.cv.kits.D2P3c.lo\endcsname{\ensuremath{-0.32}}
\expandafter\gdef\csname R@fb.cv.kits.D2P3c.n\endcsname{391}
\expandafter\gdef\csname R@fb.cv.kits.L1.d\endcsname{\ensuremath{+2.15}}
\expandafter\gdef\csname R@fb.cv.kits.L1.excl\endcsname{1}
\expandafter\gdef\csname R@fb.cv.kits.L1.hi\endcsname{\ensuremath{+3.07}}
\expandafter\gdef\csname R@fb.cv.kits.L1.lo\endcsname{\ensuremath{+1.29}}
\expandafter\gdef\csname R@fb.cv.kits.L1.n\endcsname{391}
\expandafter\gdef\csname R@fb.cv.kits.L1P3c.d\endcsname{\ensuremath{+1.59}}
\expandafter\gdef\csname R@fb.cv.kits.L1P3c.excl\endcsname{1}
\expandafter\gdef\csname R@fb.cv.kits.L1P3c.hi\endcsname{\ensuremath{+2.53}}
\expandafter\gdef\csname R@fb.cv.kits.L1P3c.lo\endcsname{\ensuremath{+0.68}}
\expandafter\gdef\csname R@fb.cv.kits.L1P3c.n\endcsname{391}
\expandafter\gdef\csname R@fb.cv.kits.L1b.d\endcsname{\ensuremath{+1.54}}
\expandafter\gdef\csname R@fb.cv.kits.L1b.excl\endcsname{1}
\expandafter\gdef\csname R@fb.cv.kits.L1b.hi\endcsname{\ensuremath{+2.42}}
\expandafter\gdef\csname R@fb.cv.kits.L1b.lo\endcsname{\ensuremath{+0.69}}
\expandafter\gdef\csname R@fb.cv.kits.L1b.n\endcsname{391}
\expandafter\gdef\csname R@fb.cv.kits.L3.d\endcsname{\ensuremath{+0.68}}
\expandafter\gdef\csname R@fb.cv.kits.L3.excl\endcsname{0}
\expandafter\gdef\csname R@fb.cv.kits.L3.hi\endcsname{\ensuremath{+1.56}}
\expandafter\gdef\csname R@fb.cv.kits.L3.lo\endcsname{\ensuremath{-0.16}}
\expandafter\gdef\csname R@fb.cv.kits.L3.n\endcsname{391}
\expandafter\gdef\csname R@fb.cv.kits.L3P3c.d\endcsname{\ensuremath{+0.12}}
\expandafter\gdef\csname R@fb.cv.kits.L3P3c.excl\endcsname{0}
\expandafter\gdef\csname R@fb.cv.kits.L3P3c.hi\endcsname{\ensuremath{+0.96}}
\expandafter\gdef\csname R@fb.cv.kits.L3P3c.lo\endcsname{\ensuremath{-0.71}}
\expandafter\gdef\csname R@fb.cv.kits.L3P3c.n\endcsname{391}
\expandafter\gdef\csname R@fb.cv.kits.P3c.d\endcsname{\ensuremath{+0.56}}
\expandafter\gdef\csname R@fb.cv.kits.P3c.excl\endcsname{0}
\expandafter\gdef\csname R@fb.cv.kits.P3c.hi\endcsname{\ensuremath{+1.31}}
\expandafter\gdef\csname R@fb.cv.kits.P3c.lo\endcsname{\ensuremath{-0.17}}
\expandafter\gdef\csname R@fb.cv.kits.P3c.n\endcsname{391}
\expandafter\gdef\csname R@fb.cv.kits.V5B1.d\endcsname{\ensuremath{+0.99}}
\expandafter\gdef\csname R@fb.cv.kits.V5B1.excl\endcsname{1}
\expandafter\gdef\csname R@fb.cv.kits.V5B1.hi\endcsname{\ensuremath{+1.82}}
\expandafter\gdef\csname R@fb.cv.kits.V5B1.lo\endcsname{\ensuremath{+0.20}}
\expandafter\gdef\csname R@fb.cv.kits.V5B1.n\endcsname{391}
\expandafter\gdef\csname R@fb.cv.kits.V5B1P3c.d\endcsname{\ensuremath{+0.44}}
\expandafter\gdef\csname R@fb.cv.kits.V5B1P3c.excl\endcsname{0}
\expandafter\gdef\csname R@fb.cv.kits.V5B1P3c.hi\endcsname{\ensuremath{+1.24}}
\expandafter\gdef\csname R@fb.cv.kits.V5B1P3c.lo\endcsname{\ensuremath{-0.31}}
\expandafter\gdef\csname R@fb.cv.kits.V5B1P3c.n\endcsname{391}
\expandafter\gdef\csname R@fb.cv.kits.V5B2.d\endcsname{\ensuremath{-0.75}}
\expandafter\gdef\csname R@fb.cv.kits.V5B2.excl\endcsname{0}
\expandafter\gdef\csname R@fb.cv.kits.V5B2.hi\endcsname{\ensuremath{+0.03}}
\expandafter\gdef\csname R@fb.cv.kits.V5B2.lo\endcsname{\ensuremath{-1.53}}
\expandafter\gdef\csname R@fb.cv.kits.V5B2.n\endcsname{391}
\expandafter\gdef\csname R@fb.cv.kits.V5B2P3c.d\endcsname{\ensuremath{-1.31}}
\expandafter\gdef\csname R@fb.cv.kits.V5B2P3c.excl\endcsname{1}
\expandafter\gdef\csname R@fb.cv.kits.V5B2P3c.hi\endcsname{\ensuremath{-0.59}}
\expandafter\gdef\csname R@fb.cv.kits.V5B2P3c.lo\endcsname{\ensuremath{-2.06}}
\expandafter\gdef\csname R@fb.cv.kits.V5B2P3c.n\endcsname{391}
\expandafter\gdef\csname R@fb.cv.kits.V6A1r.d\endcsname{\ensuremath{+0.28}}
\expandafter\gdef\csname R@fb.cv.kits.V6A1r.excl\endcsname{0}
\expandafter\gdef\csname R@fb.cv.kits.V6A1r.hi\endcsname{\ensuremath{+1.18}}
\expandafter\gdef\csname R@fb.cv.kits.V6A1r.lo\endcsname{\ensuremath{-0.62}}
\expandafter\gdef\csname R@fb.cv.kits.V6A1r.n\endcsname{391}
\expandafter\gdef\csname R@fb.cv.kits.V6A1rP3c.d\endcsname{\ensuremath{-0.28}}
\expandafter\gdef\csname R@fb.cv.kits.V6A1rP3c.excl\endcsname{0}
\expandafter\gdef\csname R@fb.cv.kits.V6A1rP3c.hi\endcsname{\ensuremath{+0.60}}
\expandafter\gdef\csname R@fb.cv.kits.V6A1rP3c.lo\endcsname{\ensuremath{-1.16}}
\expandafter\gdef\csname R@fb.cv.kits.V6A1rP3c.n\endcsname{391}
\expandafter\gdef\csname R@fb.cv.kits.V7A1.d\endcsname{\ensuremath{+1.63}}
\expandafter\gdef\csname R@fb.cv.kits.V7A1.excl\endcsname{1}
\expandafter\gdef\csname R@fb.cv.kits.V7A1.hi\endcsname{\ensuremath{+2.45}}
\expandafter\gdef\csname R@fb.cv.kits.V7A1.lo\endcsname{\ensuremath{+0.88}}
\expandafter\gdef\csname R@fb.cv.kits.V7A1.n\endcsname{391}
\expandafter\gdef\csname R@fb.cv.kits.V7A1P3c.d\endcsname{\ensuremath{+1.08}}
\expandafter\gdef\csname R@fb.cv.kits.V7A1P3c.excl\endcsname{1}
\expandafter\gdef\csname R@fb.cv.kits.V7A1P3c.hi\endcsname{\ensuremath{+1.80}}
\expandafter\gdef\csname R@fb.cv.kits.V7A1P3c.lo\endcsname{\ensuremath{+0.41}}
\expandafter\gdef\csname R@fb.cv.kits.V7A1P3c.n\endcsname{391}
\expandafter\gdef\csname R@fb.cv.kits.V7C1.d\endcsname{\ensuremath{+0.78}}
\expandafter\gdef\csname R@fb.cv.kits.V7C1.excl\endcsname{1}
\expandafter\gdef\csname R@fb.cv.kits.V7C1.hi\endcsname{\ensuremath{+1.51}}
\expandafter\gdef\csname R@fb.cv.kits.V7C1.lo\endcsname{\ensuremath{+0.07}}
\expandafter\gdef\csname R@fb.cv.kits.V7C1.n\endcsname{391}
\expandafter\gdef\csname R@fb.cv.kits.V7C1P3c.d\endcsname{\ensuremath{+0.23}}
\expandafter\gdef\csname R@fb.cv.kits.V7C1P3c.excl\endcsname{0}
\expandafter\gdef\csname R@fb.cv.kits.V7C1P3c.hi\endcsname{\ensuremath{+0.96}}
\expandafter\gdef\csname R@fb.cv.kits.V7C1P3c.lo\endcsname{\ensuremath{-0.49}}
\expandafter\gdef\csname R@fb.cv.kits.V7C1P3c.n\endcsname{391}
\expandafter\gdef\csname R@fb.on\endcsname{1}
\expandafter\gdef\csname R@fb.r18.acdc.d\endcsname{\ensuremath{+0.22}}
\expandafter\gdef\csname R@fb.r18.acdc.d3\endcsname{\ensuremath{+0.216}}
\expandafter\gdef\csname R@fb.r18.acdc.excl\endcsname{1}
\expandafter\gdef\csname R@fb.r18.acdc.hi\endcsname{\ensuremath{+0.43}}
\expandafter\gdef\csname R@fb.r18.acdc.hi3\endcsname{\ensuremath{+0.429}}
\expandafter\gdef\csname R@fb.r18.acdc.img.hi\endcsname{\ensuremath{+0.42}}
\expandafter\gdef\csname R@fb.r18.acdc.img.lo\endcsname{\ensuremath{+0.02}}
\expandafter\gdef\csname R@fb.r18.acdc.lo\endcsname{\ensuremath{0.00}}
\expandafter\gdef\csname R@fb.r18.acdc.lo3\endcsname{\ensuremath{+0.002}}
\expandafter\gdef\csname R@fb.r18.acdc.n\endcsname{100}
\expandafter\gdef\csname R@fb.r18.acdc.npat\endcsname{50}
\expandafter\gdef\csname R@fb.r18.acdc.s0.d3\endcsname{\ensuremath{+0.028}}
\expandafter\gdef\csname R@fb.r18.acdc.s1.d3\endcsname{\ensuremath{+0.326}}
\expandafter\gdef\csname R@fb.r18.acdc.s2.d3\endcsname{\ensuremath{+0.293}}
\expandafter\gdef\csname R@fb.r18.acdc.verdict\endcsname{CONFIRMED}
\expandafter\gdef\csname R@fb.r18.amos.d\endcsname{\ensuremath{+0.17}}
\expandafter\gdef\csname R@fb.r18.amos.d3\endcsname{\ensuremath{+0.167}}
\expandafter\gdef\csname R@fb.r18.amos.excl\endcsname{1}
\expandafter\gdef\csname R@fb.r18.amos.hi\endcsname{\ensuremath{+0.31}}
\expandafter\gdef\csname R@fb.r18.amos.hi3\endcsname{\ensuremath{+0.308}}
\expandafter\gdef\csname R@fb.r18.amos.lo\endcsname{\ensuremath{+0.03}}
\expandafter\gdef\csname R@fb.r18.amos.lo3\endcsname{\ensuremath{+0.027}}
\expandafter\gdef\csname R@fb.r18.amos.n\endcsname{100}
\expandafter\gdef\csname R@fb.r18.amos.s0.d3\endcsname{\ensuremath{+0.076}}
\expandafter\gdef\csname R@fb.r18.amos.s1.d3\endcsname{\ensuremath{+0.334}}
\expandafter\gdef\csname R@fb.r18.amos.s2.d3\endcsname{\ensuremath{+0.092}}
\expandafter\gdef\csname R@fb.r18.amos.verdict\endcsname{CONFIRMED}
\expandafter\gdef\csname R@fb.r18.both.excl\endcsname{1}
\expandafter\gdef\csname R@fb.r19.r2.acdc.d\endcsname{\ensuremath{+0.25}}
\expandafter\gdef\csname R@fb.r19.r2.acdc.excl\endcsname{1}
\expandafter\gdef\csname R@fb.r19.r2.acdc.hi\endcsname{\ensuremath{+0.43}}
\expandafter\gdef\csname R@fb.r19.r2.acdc.img.hi\endcsname{\ensuremath{+0.43}}
\expandafter\gdef\csname R@fb.r19.r2.acdc.img.lo\endcsname{\ensuremath{+0.06}}
\expandafter\gdef\csname R@fb.r19.r2.acdc.lo\endcsname{\ensuremath{+0.07}}
\expandafter\gdef\csname R@fb.r19.r2.acdc.n\endcsname{160}
\expandafter\gdef\csname R@fb.r19.r2.acdc.nfold\endcsname{4}
\expandafter\gdef\csname R@fb.r19.r2.acdc.npat\endcsname{80}
\expandafter\gdef\csname R@fb.r19.r2.acdc.nup\endcsname{4}
\expandafter\gdef\csname R@fb.r19.r2.acdc.verdict\endcsname{CONFIRMED}
\expandafter\gdef\csname R@fb.r19.r2.amos.d\endcsname{\ensuremath{+0.62}}
\expandafter\gdef\csname R@fb.r19.r2.amos.excl\endcsname{1}
\expandafter\gdef\csname R@fb.r19.r2.amos.hi\endcsname{\ensuremath{+0.85}}
\expandafter\gdef\csname R@fb.r19.r2.amos.lo\endcsname{\ensuremath{+0.41}}
\expandafter\gdef\csname R@fb.r19.r2.amos.n\endcsname{160}
\expandafter\gdef\csname R@fb.r19.r2.amos.nfold\endcsname{4}
\expandafter\gdef\csname R@fb.r19.r2.amos.nup\endcsname{4}
\expandafter\gdef\csname R@fb.r19.r2.amos.verdict\endcsname{CONFIRMED}
\expandafter\gdef\csname R@fb.r19.r2.held\endcsname{1}
\expandafter\gdef\csname R@fb.r19.r2.kits.d\endcsname{\ensuremath{+0.34}}
\expandafter\gdef\csname R@fb.r19.r2.kits.excl\endcsname{0}
\expandafter\gdef\csname R@fb.r19.r2.kits.hi\endcsname{\ensuremath{+1.02}}
\expandafter\gdef\csname R@fb.r19.r2.kits.lo\endcsname{\ensuremath{-0.36}}
\expandafter\gdef\csname R@fb.r19.r2.kits.n\endcsname{391}
\expandafter\gdef\csname R@fb.r19.r2.kits.nfold\endcsname{4}
\expandafter\gdef\csname R@fb.r19.r2.kits.nup\endcsname{2}
\expandafter\gdef\csname R@fb.r19.r2.kits.verdict\endcsname{UNDERPOWERED}
\expandafter\gdef\csname R@fb.r19.r2.sign.held\endcsname{1}
\expandafter\gdef\csname R@fb.r19.r2.sign.n\endcsname{8}
\expandafter\gdef\csname R@fb.r19.r2.sign.npos\endcsname{8}
\expandafter\gdef\csname R@fb.r19.r2.sign.p\endcsname{0.004}
\expandafter\gdef\csname R@fb.r19.r3inc.acdc.d\endcsname{\ensuremath{+0.05}}
\expandafter\gdef\csname R@fb.r19.r3inc.acdc.excl\endcsname{0}
\expandafter\gdef\csname R@fb.r19.r3inc.acdc.hi\endcsname{\ensuremath{+0.22}}
\expandafter\gdef\csname R@fb.r19.r3inc.acdc.img.hi\endcsname{\ensuremath{+0.20}}
\expandafter\gdef\csname R@fb.r19.r3inc.acdc.img.lo\endcsname{\ensuremath{-0.11}}
\expandafter\gdef\csname R@fb.r19.r3inc.acdc.lo\endcsname{\ensuremath{-0.13}}
\expandafter\gdef\csname R@fb.r19.r3inc.acdc.n\endcsname{160}
\expandafter\gdef\csname R@fb.r19.r3inc.acdc.nfold\endcsname{4}
\expandafter\gdef\csname R@fb.r19.r3inc.acdc.npat\endcsname{80}
\expandafter\gdef\csname R@fb.r19.r3inc.acdc.nup\endcsname{2}
\expandafter\gdef\csname R@fb.r19.r3inc.acdc.verdict\endcsname{UNDERPOWERED}
\expandafter\gdef\csname R@fb.r19.r3inc.amos.d\endcsname{\ensuremath{+0.10}}
\expandafter\gdef\csname R@fb.r19.r3inc.amos.excl\endcsname{0}
\expandafter\gdef\csname R@fb.r19.r3inc.amos.hi\endcsname{\ensuremath{+0.32}}
\expandafter\gdef\csname R@fb.r19.r3inc.amos.lo\endcsname{\ensuremath{-0.13}}
\expandafter\gdef\csname R@fb.r19.r3inc.amos.n\endcsname{160}
\expandafter\gdef\csname R@fb.r19.r3inc.amos.nfold\endcsname{4}
\expandafter\gdef\csname R@fb.r19.r3inc.amos.nup\endcsname{2}
\expandafter\gdef\csname R@fb.r19.r3inc.amos.verdict\endcsname{UNDERPOWERED}
\expandafter\gdef\csname R@fb.r19.r3inc.kits.d\endcsname{\ensuremath{+0.30}}
\expandafter\gdef\csname R@fb.r19.r3inc.kits.excl\endcsname{0}
\expandafter\gdef\csname R@fb.r19.r3inc.kits.hi\endcsname{\ensuremath{+1.01}}
\expandafter\gdef\csname R@fb.r19.r3inc.kits.lo\endcsname{\ensuremath{-0.34}}
\expandafter\gdef\csname R@fb.r19.r3inc.kits.n\endcsname{391}
\expandafter\gdef\csname R@fb.r19.r3inc.kits.nfold\endcsname{4}
\expandafter\gdef\csname R@fb.r19.r3inc.kits.nup\endcsname{3}
\expandafter\gdef\csname R@fb.r19.r3inc.kits.verdict\endcsname{UNDERPOWERED}
\expandafter\gdef\csname R@fb.r19.r3off.acdc.d\endcsname{\ensuremath{-0.05}}
\expandafter\gdef\csname R@fb.r19.r3off.acdc.excl\endcsname{0}
\expandafter\gdef\csname R@fb.r19.r3off.acdc.hi\endcsname{\ensuremath{+0.13}}
\expandafter\gdef\csname R@fb.r19.r3off.acdc.img.hi\endcsname{\ensuremath{+0.11}}
\expandafter\gdef\csname R@fb.r19.r3off.acdc.img.lo\endcsname{\ensuremath{-0.20}}
\expandafter\gdef\csname R@fb.r19.r3off.acdc.lo\endcsname{\ensuremath{-0.22}}
\expandafter\gdef\csname R@fb.r19.r3off.acdc.n\endcsname{160}
\expandafter\gdef\csname R@fb.r19.r3off.acdc.nfold\endcsname{4}
\expandafter\gdef\csname R@fb.r19.r3off.acdc.npat\endcsname{80}
\expandafter\gdef\csname R@fb.r19.r3off.acdc.nup\endcsname{2}
\expandafter\gdef\csname R@fb.r19.r3off.acdc.verdict\endcsname{UNDERPOWERED}
\expandafter\gdef\csname R@fb.r19.r3off.amos.d\endcsname{\ensuremath{-0.10}}
\expandafter\gdef\csname R@fb.r19.r3off.amos.excl\endcsname{0}
\expandafter\gdef\csname R@fb.r19.r3off.amos.hi\endcsname{\ensuremath{+0.13}}
\expandafter\gdef\csname R@fb.r19.r3off.amos.lo\endcsname{\ensuremath{-0.32}}
\expandafter\gdef\csname R@fb.r19.r3off.amos.n\endcsname{160}
\expandafter\gdef\csname R@fb.r19.r3off.amos.nfold\endcsname{4}
\expandafter\gdef\csname R@fb.r19.r3off.amos.nup\endcsname{2}
\expandafter\gdef\csname R@fb.r19.r3off.amos.verdict\endcsname{UNDERPOWERED}
\expandafter\gdef\csname R@fb.r19.r3off.kits.d\endcsname{\ensuremath{-0.30}}
\expandafter\gdef\csname R@fb.r19.r3off.kits.excl\endcsname{0}
\expandafter\gdef\csname R@fb.r19.r3off.kits.hi\endcsname{\ensuremath{+0.34}}
\expandafter\gdef\csname R@fb.r19.r3off.kits.lo\endcsname{\ensuremath{-1.01}}
\expandafter\gdef\csname R@fb.r19.r3off.kits.n\endcsname{391}
\expandafter\gdef\csname R@fb.r19.r3off.kits.nfold\endcsname{4}
\expandafter\gdef\csname R@fb.r19.r3off.kits.nup\endcsname{1}
\expandafter\gdef\csname R@fb.r19.r3off.kits.verdict\endcsname{UNDERPOWERED}
\expandafter\gdef\csname R@fb.r20.boot2.d\endcsname{\ensuremath{+1.63}}
\expandafter\gdef\csname R@fb.r20.boot2.excl\endcsname{1}
\expandafter\gdef\csname R@fb.r20.boot2.hi\endcsname{\ensuremath{+2.27}}
\expandafter\gdef\csname R@fb.r20.boot2.lo\endcsname{\ensuremath{+1.04}}
\expandafter\gdef\csname R@fb.r20.boot2.n\endcsname{391}
\expandafter\gdef\csname R@fb.r2iso.amos.abs\endcsname{\ensuremath{+0.07}}
\expandafter\gdef\csname R@fb.r2iso.amos.abs.hi\endcsname{\ensuremath{+0.22}}
\expandafter\gdef\csname R@fb.r2iso.amos.abs.lo\endcsname{\ensuremath{-0.08}}
\expandafter\gdef\csname R@fb.r2iso.amos.abspct\endcsname{11}
\expandafter\gdef\csname R@fb.r2iso.amos.full\endcsname{\ensuremath{+0.62}}
\expandafter\gdef\csname R@fb.r2iso.amos.full.hi\endcsname{\ensuremath{+0.85}}
\expandafter\gdef\csname R@fb.r2iso.amos.full.lo\endcsname{\ensuremath{+0.41}}
\expandafter\gdef\csname R@fb.r2iso.amos.pres\endcsname{\ensuremath{+0.55}}
\expandafter\gdef\csname R@fb.r2iso.amos.pres.hi\endcsname{\ensuremath{+0.70}}
\expandafter\gdef\csname R@fb.r2iso.amos.pres.lo\endcsname{\ensuremath{+0.41}}
\expandafter\gdef\csname R@fb.r8.a.h1v7a1.d\endcsname{\ensuremath{-0.88}}
\expandafter\gdef\csname R@fb.r8.a.h1v7a1.excl\endcsname{0}
\expandafter\gdef\csname R@fb.r8.a.h1v7a1.hi\endcsname{\ensuremath{+3.13}}
\expandafter\gdef\csname R@fb.r8.a.h1v7a1.lo\endcsname{\ensuremath{-5.41}}
\expandafter\gdef\csname R@fb.r8.a.h1v7a1.n\endcsname{27}
\expandafter\gdef\csname R@fb.r8.a.h1v7a1.nseed\endcsname{5/3}
\expandafter\gdef\csname R@fb.r8.a.h1y2.d\endcsname{\ensuremath{-1.21}}
\expandafter\gdef\csname R@fb.r8.a.h1y2.excl\endcsname{0}
\expandafter\gdef\csname R@fb.r8.a.h1y2.hi\endcsname{\ensuremath{+1.96}}
\expandafter\gdef\csname R@fb.r8.a.h1y2.lo\endcsname{\ensuremath{-5.57}}
\expandafter\gdef\csname R@fb.r8.a.h1y2.n\endcsname{27}
\expandafter\gdef\csname R@fb.r8.a.h1y2.nseed\endcsname{5/5}
\expandafter\gdef\csname R@fb.r8.b.d\endcsname{\ensuremath{+2.56}}
\expandafter\gdef\csname R@fb.r8.b.excl\endcsname{0}
\expandafter\gdef\csname R@fb.r8.b.hi\endcsname{\ensuremath{+5.28}}
\expandafter\gdef\csname R@fb.r8.b.lo\endcsname{\ensuremath{-0.16}}
\expandafter\gdef\csname R@fb.r8.b.n\endcsname{180}
\expandafter\gdef\csname R@fb.seed\endcsname{20261013}
\makeatother

\input{an_extra_counts.tex}
\input{an_extra_cv.tex}
\makeatletter
\expandafter\gdef\csname R@an.extrfp.L1.f1\endcsname{11.69}
\expandafter\gdef\csname R@an.extrfp.L1.f2\endcsname{2.09}
\expandafter\gdef\csname R@an.extrfp.L1.f3\endcsname{10.37}
\expandafter\gdef\csname R@an.extrfp.L1.f4\endcsname{3.04}
\expandafter\gdef\csname R@an.extrfp.L1.mL\endcsname{6.80}
\expandafter\gdef\csname R@an.extrfp.V7A1.f1\endcsname{14.57}
\expandafter\gdef\csname R@an.extrfp.V7A1.f2\endcsname{4.36}
\expandafter\gdef\csname R@an.extrfp.V7A1.f3\endcsname{5.89}
\expandafter\gdef\csname R@an.extrfp.V7A1.f4\endcsname{11.25}
\expandafter\gdef\csname R@an.extrfp.V7A1.mL\endcsname{9.02}
\expandafter\gdef\csname R@an.extrfp.Y2.f1\endcsname{11.60}
\expandafter\gdef\csname R@an.extrfp.Y2.f2\endcsname{3.20}
\expandafter\gdef\csname R@an.extrfp.Y2.f3\endcsname{4.04}
\expandafter\gdef\csname R@an.extrfp.Y2.f4\endcsname{0.09}
\expandafter\gdef\csname R@an.extrfp.Y2.mL\endcsname{4.73}
\expandafter\gdef\csname R@an.extrfp.Y2.vs.L1.exceptfold\endcsname{2}
\expandafter\gdef\csname R@an.extrfp.Y2.vs.L1.folds\endcsname{3}
\expandafter\gdef\csname R@an.extrfp.Y2.vs.V7A1.folds\endcsname{4}
\expandafter\gdef\csname R@an.extrfp.Y2.vs.V7A1.topfold\endcsname{4}
\expandafter\gdef\csname R@an.extrfp.Y2.vs.V7A1.topshare\endcsname{65}
\expandafter\gdef\csname R@an.extrfp.Y2.vs.base.folds\endcsname{4}
\expandafter\gdef\csname R@an.extrfp.base.f1\endcsname{21.15}
\expandafter\gdef\csname R@an.extrfp.base.f2\endcsname{14.65}
\expandafter\gdef\csname R@an.extrfp.base.f3\endcsname{11.11}
\expandafter\gdef\csname R@an.extrfp.base.f4\endcsname{13.09}
\expandafter\gdef\csname R@an.extrfp.base.mL\endcsname{15.00}
\expandafter\gdef\csname R@an.extrfp.drop.pct\endcsname{0.49}
\expandafter\gdef\csname R@an.extrfp.ncases\endcsname{391}
\expandafter\gdef\csname R@an.extrfp.nfolds\endcsname{4}
\expandafter\gdef\csname R@an.extrfp.thr.inside\endcsname{50}
\expandafter\gdef\csname R@an.extrfp.thr.minvol\endcsname{0.05}
\makeatother

\makeatletter
\expandafter\gdef\csname R@fnd.cyst.order.n\endcsname{4}
\expandafter\gdef\csname R@fnd.cyst.order.tot\endcsname{4}
\expandafter\gdef\csname R@fnd.cyst.order.word\endcsname{four}
\expandafter\gdef\csname R@fnd.r4.acdc.mimic\endcsname{0}
\expandafter\gdef\csname R@fnd.r4.amos.mimic\endcsname{0}
\expandafter\gdef\csname R@fnd.r4.brats.mimic\endcsname{2}
\expandafter\gdef\csname R@fnd.r4.btcv.mimic\endcsname{0}
\expandafter\gdef\csname R@fnd.r4.kits.mimic\endcsname{2}
\expandafter\gdef\csname R@fnd.r4.lits.mimic\endcsname{0}
\expandafter\gdef\csname R@fnd.r4.nfire\endcsname{2}
\expandafter\gdef\csname R@fnd.w9.amos.dL1\endcsname{\ensuremath{-0.18}}
\expandafter\gdef\csname R@fnd.w9.amos.dL1.nup\endcsname{0}
\expandafter\gdef\csname R@fnd.w9.amos.dM\endcsname{\ensuremath{-0.03}}
\expandafter\gdef\csname R@fnd.w9.amos.dM.nup\endcsname{2}
\expandafter\gdef\csname R@fnd.w9.amos.dbase\endcsname{\ensuremath{+0.23}}
\expandafter\gdef\csname R@fnd.w9.amos.dbase.nup\endcsname{2}
\expandafter\gdef\csname R@fnd.w9.amos.mean\endcsname{88.57}
\expandafter\gdef\csname R@fnd.w9.cystcov\endcsname{0.92}
\expandafter\gdef\csname R@fnd.w9.cystcov.M\endcsname{0.81}
\expandafter\gdef\csname R@fnd.w9.dtum\endcsname{79.84}
\expandafter\gdef\csname R@fnd.w9.dtum.dM\endcsname{\ensuremath{+1.01}}
\expandafter\gdef\csname R@fnd.w9.f0gate.dM\endcsname{\ensuremath{+0.043}}
\expandafter\gdef\csname R@fnd.w9.kits.dL1\endcsname{\ensuremath{+0.56}}
\expandafter\gdef\csname R@fnd.w9.kits.dL1.nfold\endcsname{4}
\expandafter\gdef\csname R@fnd.w9.kits.dL1.nup\endcsname{4}
\expandafter\gdef\csname R@fnd.w9.kits.dM\endcsname{\ensuremath{+1.07}}
\expandafter\gdef\csname R@fnd.w9.kits.dM.nfold\endcsname{4}
\expandafter\gdef\csname R@fnd.w9.kits.dM.nup\endcsname{4}
\expandafter\gdef\csname R@fnd.w9.kits.dbase\endcsname{\ensuremath{+2.71}}
\expandafter\gdef\csname R@fnd.w9.kits.dbase.nfold\endcsname{4}
\expandafter\gdef\csname R@fnd.w9.kits.dbase.nup\endcsname{4}
\expandafter\gdef\csname R@fnd.w9.kits.mean\endcsname{86.70}
\expandafter\gdef\csname R@fnd.w9.mem\endcsname{7.77}
\expandafter\gdef\csname R@fnd.w9.nclause\endcsname{6}
\expandafter\gdef\csname R@fnd.w9.nclause.word\endcsname{six}
\expandafter\gdef\csname R@fnd.w9.on\endcsname{1}
\expandafter\gdef\csname R@fnd.w9.par\endcsname{31.04}
\expandafter\gdef\csname R@fnd.w9.par.extra\endcsname{0}
\expandafter\gdef\csname R@fnd.w9.par.vsL1\endcsname{0.30}
\expandafter\gdef\csname R@fnd.w9.prec\endcsname{0.780}
\expandafter\gdef\csname R@fnd.w9.prec.M\endcsname{0.758}
\expandafter\gdef\csname R@fnd.w9.prec.bear.dM\endcsname{\ensuremath{+0.022}}
\expandafter\gdef\csname R@fnd.w9.prec.dM\endcsname{\ensuremath{+0.022}}
\expandafter\gdef\csname R@fnd.w9.prec.free.dM\endcsname{\ensuremath{+0.018}}
\expandafter\gdef\csname R@fnd.w9.prec.nup\endcsname{3}
\expandafter\gdef\csname R@fnd.w9.rec.bear.dM\endcsname{\ensuremath{-0.013}}
\expandafter\gdef\csname R@fnd.w9.rec.dM\endcsname{\ensuremath{-0.006}}
\expandafter\gdef\csname R@fnd.w9.rec.free.dM\endcsname{\ensuremath{+0.000}}
\expandafter\gdef\csname R@fnd.w9.slope\endcsname{0.170}
\expandafter\gdef\csname R@fnd.w9.slope.M\endcsname{0.397}
\expandafter\gdef\csname R@fnd.w9.slope.rel\endcsname{57}
\expandafter\gdef\csname R@fnd.w9.step.vsbase\endcsname{0.95}
\expandafter\gdef\csname R@fnd.w9.thr.amos\endcsname{0.20}
\expandafter\gdef\csname R@fnd.w9.thr.dtum\endcsname{+0.80}
\expandafter\gdef\csname R@fnd.w9.thr.f0prec\endcsname{+0.032}
\expandafter\gdef\csname R@fnd.w9.thr.prec\endcsname{+0.020}
\expandafter\gdef\csname R@fnd.w9.thr.rec\endcsname{0.015}
\makeatother

\input{an_extra_hyp.tex}
\makeatletter
\expandafter\gdef\csname R@ins.corr.n\endcsname{22}
\expandafter\gdef\csname R@ins.corr.narch\endcsname{12}
\expandafter\gdef\csname R@ins.corr.nonbase.n\endcsname{13}
\expandafter\gdef\csname R@ins.corr.nonbase.prec\endcsname{0.914}
\expandafter\gdef\csname R@ins.corr.nonbase.rec\endcsname{0.034}
\expandafter\gdef\csname R@ins.corr.prec\endcsname{0.951}
\expandafter\gdef\csname R@ins.corr.rec\endcsname{0.084}
\expandafter\gdef\csname R@ins.corr.volbias\endcsname{\ensuremath{-0.801}}
\expandafter\gdef\csname R@ins.cyst.bear.L1\endcsname{76.06}
\expandafter\gdef\csname R@ins.cyst.bear.L3\endcsname{73.10}
\expandafter\gdef\csname R@ins.cyst.bear.V7A1\endcsname{74.89}
\expandafter\gdef\csname R@ins.cyst.bear.V7C1\endcsname{72.70}
\expandafter\gdef\csname R@ins.cyst.bear.base\endcsname{72.22}
\expandafter\gdef\csname R@ins.cyst.free.L1\endcsname{82.76}
\expandafter\gdef\csname R@ins.cyst.free.L3\endcsname{82.18}
\expandafter\gdef\csname R@ins.cyst.free.V7A1\endcsname{82.89}
\expandafter\gdef\csname R@ins.cyst.free.V7C1\endcsname{81.75}
\expandafter\gdef\csname R@ins.cyst.free.base\endcsname{81.50}
\expandafter\gdef\csname R@ins.cyst.n.all\endcsname{391}
\expandafter\gdef\csname R@ins.cyst.n.bear\endcsname{198}
\expandafter\gdef\csname R@ins.cyst.n.free\endcsname{193}
\expandafter\gdef\csname R@ins.cyst.pen.L1\endcsname{\ensuremath{-6.69}}
\expandafter\gdef\csname R@ins.cyst.pen.L3\endcsname{\ensuremath{-9.08}}
\expandafter\gdef\csname R@ins.cyst.pen.V7A1\endcsname{\ensuremath{-8.00}}
\expandafter\gdef\csname R@ins.cyst.pen.V7C1\endcsname{\ensuremath{-9.05}}
\expandafter\gdef\csname R@ins.cyst.pen.base\endcsname{\ensuremath{-9.28}}
\expandafter\gdef\csname R@ins.cyst.prec.bear.L1\endcsname{0.776}
\expandafter\gdef\csname R@ins.cyst.prec.bear.L3\endcsname{0.725}
\expandafter\gdef\csname R@ins.cyst.prec.bear.V7A1\endcsname{0.758}
\expandafter\gdef\csname R@ins.cyst.prec.bear.V7C1\endcsname{0.739}
\expandafter\gdef\csname R@ins.cyst.prec.bear.base\endcsname{0.712}
\expandafter\gdef\csname R@ins.cyst.prec.free.L1\endcsname{0.856}
\expandafter\gdef\csname R@ins.cyst.prec.free.L3\endcsname{0.843}
\expandafter\gdef\csname R@ins.cyst.prec.free.V7A1\endcsname{0.855}
\expandafter\gdef\csname R@ins.cyst.prec.free.V7C1\endcsname{0.841}
\expandafter\gdef\csname R@ins.cyst.prec.free.base\endcsname{0.819}
\expandafter\gdef\csname R@ins.cyst.rec.bear.L1\endcsname{0.808}
\expandafter\gdef\csname R@ins.cyst.rec.bear.L3\endcsname{0.821}
\expandafter\gdef\csname R@ins.cyst.rec.bear.V7A1\endcsname{0.815}
\expandafter\gdef\csname R@ins.cyst.rec.bear.V7C1\endcsname{0.804}
\expandafter\gdef\csname R@ins.cyst.rec.bear.base\endcsname{0.831}
\expandafter\gdef\csname R@ins.cyst.rec.free.L1\endcsname{0.849}
\expandafter\gdef\csname R@ins.cyst.rec.free.L3\endcsname{0.857}
\expandafter\gdef\csname R@ins.cyst.rec.free.V7A1\endcsname{0.852}
\expandafter\gdef\csname R@ins.cyst.rec.free.V7C1\endcsname{0.852}
\expandafter\gdef\csname R@ins.cyst.rec.free.base\endcsname{0.862}
\expandafter\gdef\csname R@ins.cyst.signs.neg\endcsname{20}
\expandafter\gdef\csname R@ins.cyst.signs.tot\endcsname{20}
\expandafter\gdef\csname R@ins.f0.base.dice.hi\endcsname{76.44}
\expandafter\gdef\csname R@ins.f0.base.dice.lo\endcsname{73.78}
\expandafter\gdef\csname R@ins.f0.base.dice.sd\endcsname{0.95}
\expandafter\gdef\csname R@ins.f0.base.n\endcsname{9}
\expandafter\gdef\csname R@ins.f0.base.prec.hi\endcsname{0.749}
\expandafter\gdef\csname R@ins.f0.base.prec.lo\endcsname{0.720}
\expandafter\gdef\csname R@ins.f0.base.prec.sd\endcsname{0.011}
\expandafter\gdef\csname R@ins.f0.pen.mean\endcsname{\ensuremath{-14.36}}
\expandafter\gdef\csname R@ins.f0.pen.sd\endcsname{0.97}
\expandafter\gdef\csname R@ins.gain.all.L1\endcsname{\ensuremath{+2.57}}
\expandafter\gdef\csname R@ins.gain.all.L3\endcsname{\ensuremath{+0.78}}
\expandafter\gdef\csname R@ins.gain.all.V7A1\endcsname{\ensuremath{+2.04}}
\expandafter\gdef\csname R@ins.gain.all.V7C1\endcsname{\ensuremath{+0.37}}
\expandafter\gdef\csname R@ins.gain.bear.L1\endcsname{\ensuremath{+3.84}}
\expandafter\gdef\csname R@ins.gain.bear.L3\endcsname{\ensuremath{+0.88}}
\expandafter\gdef\csname R@ins.gain.bear.V7A1\endcsname{\ensuremath{+2.67}}
\expandafter\gdef\csname R@ins.gain.bear.V7C1\endcsname{\ensuremath{+0.48}}
\expandafter\gdef\csname R@ins.gain.free.L1\endcsname{\ensuremath{+1.26}}
\expandafter\gdef\csname R@ins.gain.free.L3\endcsname{\ensuremath{+0.68}}
\expandafter\gdef\csname R@ins.gain.free.V7A1\endcsname{\ensuremath{+1.39}}
\expandafter\gdef\csname R@ins.gain.free.V7C1\endcsname{\ensuremath{+0.25}}
\expandafter\gdef\csname R@ins.gross.deficit.L1\endcsname{49.6}
\expandafter\gdef\csname R@ins.gross.deficit.L3\endcsname{52.8}
\expandafter\gdef\csname R@ins.gross.deficit.V7A1\endcsname{48.6}
\expandafter\gdef\csname R@ins.gross.deficit.V7C1\endcsname{50.7}
\expandafter\gdef\csname R@ins.gross.deficit.base\endcsname{60.1}
\expandafter\gdef\csname R@ins.gross.pairs.L1\endcsname{20.5}
\expandafter\gdef\csname R@ins.gross.pairs.L3\endcsname{23.8}
\expandafter\gdef\csname R@ins.gross.pairs.V7A1\endcsname{22.3}
\expandafter\gdef\csname R@ins.gross.pairs.V7C1\endcsname{22.3}
\expandafter\gdef\csname R@ins.gross.pairs.base\endcsname{27.4}
\expandafter\gdef\csname R@ins.gross.thr\endcsname{2}
\expandafter\gdef\csname R@ins.share.cases.bear\endcsname{50.6}
\expandafter\gdef\csname R@ins.share.cases.both\endcsname{19.4}
\expandafter\gdef\csname R@ins.share.cases.free\endcsname{49.4}
\expandafter\gdef\csname R@ins.share.cases.small\endcsname{34.3}
\expandafter\gdef\csname R@ins.share.deficit.bear\endcsname{60.6}
\expandafter\gdef\csname R@ins.share.deficit.both\endcsname{33.3}
\expandafter\gdef\csname R@ins.share.deficit.free\endcsname{39.4}
\expandafter\gdef\csname R@ins.share.deficit.small\endcsname{49.5}
\expandafter\gdef\csname R@ins.share.n.bear\endcsname{198}
\expandafter\gdef\csname R@ins.share.n.both\endcsname{76}
\expandafter\gdef\csname R@ins.share.n.free\endcsname{193}
\expandafter\gdef\csname R@ins.share.n.small\endcsname{134}
\expandafter\gdef\csname R@ins.share.small.ml\endcsname{15}
\expandafter\gdef\csname R@ins.slope.ctrl.L1\endcsname{0.056}
\expandafter\gdef\csname R@ins.slope.ctrl.L3\endcsname{0.066}
\expandafter\gdef\csname R@ins.slope.ctrl.V7A1\endcsname{0.057}
\expandafter\gdef\csname R@ins.slope.ctrl.V7C1\endcsname{0.054}
\expandafter\gdef\csname R@ins.slope.ctrl.base\endcsname{0.055}
\expandafter\gdef\csname R@ins.slope.cyst.L1\endcsname{0.191}
\expandafter\gdef\csname R@ins.slope.cyst.L3\endcsname{0.281}
\expandafter\gdef\csname R@ins.slope.cyst.V7A1\endcsname{0.283}
\expandafter\gdef\csname R@ins.slope.cyst.V7C1\endcsname{0.503}
\expandafter\gdef\csname R@ins.slope.cyst.base\endcsname{0.279}
\expandafter\gdef\csname R@ins.slope.signs.pos\endcsname{20}
\expandafter\gdef\csname R@ins.slope.signs.tot\endcsname{20}
\expandafter\gdef\csname R@ins.slope.tum.L1\endcsname{0.052}
\expandafter\gdef\csname R@ins.slope.tum.L3\endcsname{0.048}
\expandafter\gdef\csname R@ins.slope.tum.V7A1\endcsname{0.051}
\expandafter\gdef\csname R@ins.slope.tum.V7C1\endcsname{0.049}
\expandafter\gdef\csname R@ins.slope.tum.base\endcsname{0.051}
\expandafter\gdef\csname R@ins.small.dice.L1\endcsname{69.97}
\expandafter\gdef\csname R@ins.small.dice.L3\endcsname{68.31}
\expandafter\gdef\csname R@ins.small.dice.V7A1\endcsname{70.67}
\expandafter\gdef\csname R@ins.small.dice.V7C1\endcsname{68.10}
\expandafter\gdef\csname R@ins.small.dice.base\endcsname{66.49}
\expandafter\gdef\csname R@ins.small.fnml.L1\endcsname{1.59}
\expandafter\gdef\csname R@ins.small.fnml.L3\endcsname{1.45}
\expandafter\gdef\csname R@ins.small.fnml.V7A1\endcsname{1.53}
\expandafter\gdef\csname R@ins.small.fnml.V7C1\endcsname{1.58}
\expandafter\gdef\csname R@ins.small.fnml.base\endcsname{1.36}
\expandafter\gdef\csname R@ins.small.fpml.L1\endcsname{5.21}
\expandafter\gdef\csname R@ins.small.fpml.L3\endcsname{11.25}
\expandafter\gdef\csname R@ins.small.fpml.V7A1\endcsname{6.77}
\expandafter\gdef\csname R@ins.small.fpml.V7C1\endcsname{9.08}
\expandafter\gdef\csname R@ins.small.fpml.base\endcsname{9.28}
\expandafter\gdef\csname R@ins.small.hd20.L1\endcsname{39.6}
\expandafter\gdef\csname R@ins.small.hd20.L3\endcsname{44.0}
\expandafter\gdef\csname R@ins.small.hd20.V7A1\endcsname{41.8}
\expandafter\gdef\csname R@ins.small.hd20.V7C1\endcsname{45.1}
\expandafter\gdef\csname R@ins.small.hd20.base\endcsname{52.2}
\expandafter\gdef\csname R@ins.small.layers.L1\endcsname{2.76}
\expandafter\gdef\csname R@ins.small.layers.L3\endcsname{5.85}
\expandafter\gdef\csname R@ins.small.layers.V7A1\endcsname{2.97}
\expandafter\gdef\csname R@ins.small.layers.V7C1\endcsname{3.92}
\expandafter\gdef\csname R@ins.small.layers.base\endcsname{3.99}
\expandafter\gdef\csname R@ins.small.n\endcsname{134}
\expandafter\gdef\csname R@ins.small.refml\endcsname{6.93}
\expandafter\gdef\csname R@ins.small.volbias.L1\endcsname{\ensuremath{+0.78}}
\expandafter\gdef\csname R@ins.small.volbias.L3\endcsname{\ensuremath{+2.42}}
\expandafter\gdef\csname R@ins.small.volbias.V7A1\endcsname{\ensuremath{+0.91}}
\expandafter\gdef\csname R@ins.small.volbias.V7C1\endcsname{\ensuremath{+1.44}}
\expandafter\gdef\csname R@ins.small.volbias.base\endcsname{\ensuremath{+1.31}}
\expandafter\gdef\csname R@ins.thick.gap.L1\endcsname{\ensuremath{+2.56}}
\expandafter\gdef\csname R@ins.thick.gap.L3\endcsname{\ensuremath{+2.19}}
\expandafter\gdef\csname R@ins.thick.gap.V7A1\endcsname{\ensuremath{+2.96}}
\expandafter\gdef\csname R@ins.thick.gap.V7C1\endcsname{\ensuremath{+2.43}}
\expandafter\gdef\csname R@ins.thick.gap.base\endcsname{\ensuremath{+0.64}}
\expandafter\gdef\csname R@ins.thick.mm\endcsname{4}
\expandafter\gdef\csname R@ins.thick.n.thick\endcsname{183}
\expandafter\gdef\csname R@ins.thick.n.thin\endcsname{208}
\expandafter\gdef\csname R@ins.thick.thick.L1\endcsname{78.00}
\expandafter\gdef\csname R@ins.thick.thick.L3\endcsname{76.41}
\expandafter\gdef\csname R@ins.thick.thick.V7A1\endcsname{77.26}
\expandafter\gdef\csname R@ins.thick.thick.V7C1\endcsname{75.87}
\expandafter\gdef\csname R@ins.thick.thick.base\endcsname{76.46}
\expandafter\gdef\csname R@ins.thick.thin.L1\endcsname{80.57}
\expandafter\gdef\csname R@ins.thick.thin.L3\endcsname{78.60}
\expandafter\gdef\csname R@ins.thick.thin.V7A1\endcsname{80.23}
\expandafter\gdef\csname R@ins.thick.thin.V7C1\endcsname{78.30}
\expandafter\gdef\csname R@ins.thick.thin.base\endcsname{77.10}
\expandafter\gdef\csname R@ins.tum.dice.L1\endcsname{79.37}
\expandafter\gdef\csname R@ins.tum.dice.L3\endcsname{77.58}
\expandafter\gdef\csname R@ins.tum.dice.V7A1\endcsname{78.84}
\expandafter\gdef\csname R@ins.tum.dice.V7C1\endcsname{77.17}
\expandafter\gdef\csname R@ins.tum.dice.base\endcsname{76.80}
\expandafter\gdef\csname R@ins.tum.prec.L1\endcsname{0.815}
\expandafter\gdef\csname R@ins.tum.prec.L3\endcsname{0.783}
\expandafter\gdef\csname R@ins.tum.prec.V7A1\endcsname{0.806}
\expandafter\gdef\csname R@ins.tum.prec.V7C1\endcsname{0.789}
\expandafter\gdef\csname R@ins.tum.prec.base\endcsname{0.765}
\expandafter\gdef\csname R@ins.tum.rec.L1\endcsname{0.828}
\expandafter\gdef\csname R@ins.tum.rec.L3\endcsname{0.839}
\expandafter\gdef\csname R@ins.tum.rec.V7A1\endcsname{0.833}
\expandafter\gdef\csname R@ins.tum.rec.V7C1\endcsname{0.828}
\expandafter\gdef\csname R@ins.tum.rec.base\endcsname{0.846}
\expandafter\gdef\csname R@ins.tum.volbias.L1\endcsname{\ensuremath{+0.38}}
\expandafter\gdef\csname R@ins.tum.volbias.L3\endcsname{\ensuremath{+1.08}}
\expandafter\gdef\csname R@ins.tum.volbias.V7A1\endcsname{\ensuremath{+0.53}}
\expandafter\gdef\csname R@ins.tum.volbias.V7C1\endcsname{\ensuremath{+0.79}}
\expandafter\gdef\csname R@ins.tum.volbias.base\endcsname{\ensuremath{+0.77}}
\expandafter\gdef\csname R@ins.w9.L1.absent\endcsname{12}
\expandafter\gdef\csname R@ins.w9.L1.amos\endcsname{88.75}
\expandafter\gdef\csname R@ins.w9.L1.dtum\endcsname{79.36}
\expandafter\gdef\csname R@ins.w9.L1.gross\endcsname{20.7}
\expandafter\gdef\csname R@ins.w9.L1.hd20\endcsname{36.4}
\expandafter\gdef\csname R@ins.w9.L1.kits\endcsname{86.14}
\expandafter\gdef\csname R@ins.w9.L1.prec\endcsname{0.776}
\expandafter\gdef\csname R@ins.w9.L1.rec\endcsname{0.828}
\expandafter\gdef\csname R@ins.w9.L1.smallvb\endcsname{0.76}
\expandafter\gdef\csname R@ins.w9.V7A1.absent\endcsname{10}
\expandafter\gdef\csname R@ins.w9.V7A1.amos\endcsname{88.61}
\expandafter\gdef\csname R@ins.w9.V7A1.dtum\endcsname{78.83}
\expandafter\gdef\csname R@ins.w9.V7A1.gross\endcsname{22.5}
\expandafter\gdef\csname R@ins.w9.V7A1.hd20\endcsname{38.5}
\expandafter\gdef\csname R@ins.w9.V7A1.kits\endcsname{85.63}
\expandafter\gdef\csname R@ins.w9.V7A1.prec\endcsname{0.758}
\expandafter\gdef\csname R@ins.w9.V7A1.rec\endcsname{0.833}
\expandafter\gdef\csname R@ins.w9.V7A1.smallvb\endcsname{0.89}
\expandafter\gdef\csname R@ins.w9.Y2.absent\endcsname{13}
\expandafter\gdef\csname R@ins.w9.Y2.amos\endcsname{88.57}
\expandafter\gdef\csname R@ins.w9.Y2.amos.d\endcsname{\ensuremath{-0.03}}
\expandafter\gdef\csname R@ins.w9.Y2.amos.dB\endcsname{\ensuremath{+0.23}}
\expandafter\gdef\csname R@ins.w9.Y2.amosok\endcsname{1}
\expandafter\gdef\csname R@ins.w9.Y2.diceok\endcsname{1}
\expandafter\gdef\csname R@ins.w9.Y2.dtum\endcsname{79.84}
\expandafter\gdef\csname R@ins.w9.Y2.dtum.d\endcsname{\ensuremath{+1.01}}
\expandafter\gdef\csname R@ins.w9.Y2.f0gate\endcsname{1}
\expandafter\gdef\csname R@ins.w9.Y2.f0gate.d\endcsname{\ensuremath{+0.043}}
\expandafter\gdef\csname R@ins.w9.Y2.gross\endcsname{18.4}
\expandafter\gdef\csname R@ins.w9.Y2.gross.rel\endcsname{\ensuremath{-18}}
\expandafter\gdef\csname R@ins.w9.Y2.grossok\endcsname{1}
\expandafter\gdef\csname R@ins.w9.Y2.guardok\endcsname{1}
\expandafter\gdef\csname R@ins.w9.Y2.hd20\endcsname{32.9}
\expandafter\gdef\csname R@ins.w9.Y2.hd20.rel\endcsname{\ensuremath{-14}}
\expandafter\gdef\csname R@ins.w9.Y2.hd20ok\endcsname{1}
\expandafter\gdef\csname R@ins.w9.Y2.kits\endcsname{86.70}
\expandafter\gdef\csname R@ins.w9.Y2.kits.d\endcsname{\ensuremath{+1.07}}
\expandafter\gdef\csname R@ins.w9.Y2.kits.dB\endcsname{\ensuremath{+2.71}}
\expandafter\gdef\csname R@ins.w9.Y2.kitsok\endcsname{1}
\expandafter\gdef\csname R@ins.w9.Y2.landed\endcsname{1}
\expandafter\gdef\csname R@ins.w9.Y2.on\endcsname{1}
\expandafter\gdef\csname R@ins.w9.Y2.par\endcsname{1.00}
\expandafter\gdef\csname R@ins.w9.Y2.prec\endcsname{0.780}
\expandafter\gdef\csname R@ins.w9.Y2.prec.d\endcsname{\ensuremath{+0.022}}
\expandafter\gdef\csname R@ins.w9.Y2.prec.nup\endcsname{3}
\expandafter\gdef\csname R@ins.w9.Y2.precok\endcsname{1}
\expandafter\gdef\csname R@ins.w9.Y2.rec\endcsname{0.827}
\expandafter\gdef\csname R@ins.w9.Y2.rec.d\endcsname{\ensuremath{-0.006}}
\expandafter\gdef\csname R@ins.w9.Y2.recok\endcsname{1}
\expandafter\gdef\csname R@ins.w9.Y2.smallok\endcsname{1}
\expandafter\gdef\csname R@ins.w9.Y2.smallvb\endcsname{0.47}
\expandafter\gdef\csname R@ins.w9.Y2.smallvb.d\endcsname{\ensuremath{-0.42}}
\expandafter\gdef\csname R@ins.w9.Y2.works\endcsname{1}
\expandafter\gdef\csname R@ins.w9.any.unrepl\endcsname{1}
\expandafter\gdef\csname R@ins.w9.any.works\endcsname{1}
\expandafter\gdef\csname R@ins.w9.base.absent\endcsname{13}
\expandafter\gdef\csname R@ins.w9.base.amos\endcsname{88.35}
\expandafter\gdef\csname R@ins.w9.base.dtum\endcsname{76.80}
\expandafter\gdef\csname R@ins.w9.base.gross\endcsname{27.4}
\expandafter\gdef\csname R@ins.w9.base.hd20\endcsname{45.5}
\expandafter\gdef\csname R@ins.w9.base.kits\endcsname{83.99}
\expandafter\gdef\csname R@ins.w9.base.prec\endcsname{0.714}
\expandafter\gdef\csname R@ins.w9.base.rec\endcsname{0.846}
\expandafter\gdef\csname R@ins.w9.base.smallvb\endcsname{1.30}
\expandafter\gdef\csname R@ins.w9.thr.absent\endcsname{10}
\expandafter\gdef\csname R@ins.w9.thr.amos\endcsname{0.20}
\expandafter\gdef\csname R@ins.w9.thr.dtum\endcsname{0.80}
\expandafter\gdef\csname R@ins.w9.thr.f0hd20\endcsname{15}
\expandafter\gdef\csname R@ins.w9.thr.f0prec\endcsname{0.032}
\expandafter\gdef\csname R@ins.w9.thr.gross\endcsname{12}
\expandafter\gdef\csname R@ins.w9.thr.hd20\endcsname{12}
\expandafter\gdef\csname R@ins.w9.thr.hdmm\endcsname{20}
\expandafter\gdef\csname R@ins.w9.thr.kits\endcsname{0.00}
\expandafter\gdef\csname R@ins.w9.thr.layers\endcsname{2}
\expandafter\gdef\csname R@ins.w9.thr.prec\endcsname{0.020}
\expandafter\gdef\csname R@ins.w9.thr.rec\endcsname{0.015}
\expandafter\gdef\csname R@ins.w9.thr.smallml\endcsname{15}
\expandafter\gdef\csname R@ins.w9.thr.smallvb\endcsname{0.25}
\makeatother

\makeatletter
\expandafter\gdef\csname R@an.bos.auc.max\endcsname{0.516}
\expandafter\gdef\csname R@an.bos.auc.mean\endcsname{0.159}
\expandafter\gdef\csname R@an.bos.auc.vol\endcsname{0.782}
\expandafter\gdef\csname R@an.bos.falseflag\endcsname{90}
\expandafter\gdef\csname R@an.bos.n.neg\endcsname{596}
\expandafter\gdef\csname R@an.bos.n.pos\endcsname{25}
\expandafter\gdef\csname R@an.bos.sens\endcsname{80}
\expandafter\gdef\csname R@an.bos.thr.hu\endcsname{40}
\expandafter\gdef\csname R@an.bos.thr.mm\endcsname{4}
\expandafter\gdef\csname R@an.hu.L1.flip\endcsname{35.2}
\expandafter\gdef\csname R@an.hu.L1.flip.hi\endcsname{40.5}
\expandafter\gdef\csname R@an.hu.L1.flip.lo\endcsname{30.1}
\expandafter\gdef\csname R@an.hu.V7A1.flip\endcsname{37.1}
\expandafter\gdef\csname R@an.hu.V7A1.flip.hi\endcsname{42.9}
\expandafter\gdef\csname R@an.hu.V7A1.flip.lo\endcsname{32.6}
\expandafter\gdef\csname R@an.hu.Y2.flip\endcsname{38.8}
\expandafter\gdef\csname R@an.hu.Y2.flip.hi\endcsname{45.4}
\expandafter\gdef\csname R@an.hu.Y2.flip.lo\endcsname{33.5}
\expandafter\gdef\csname R@an.hu.base.flip\endcsname{28.0}
\expandafter\gdef\csname R@an.hu.base.flip.hi\endcsname{32.9}
\expandafter\gdef\csname R@an.hu.base.flip.lo\endcsname{23.9}
\expandafter\gdef\csname R@an.hu.boot\endcsname{2000}
\expandafter\gdef\csname R@an.hu.bosniak.hi\endcsname{40}
\expandafter\gdef\csname R@an.hu.bosniak.lo\endcsname{20}
\expandafter\gdef\csname R@an.hu.cyst.L1.phi\endcsname{37.7}
\expandafter\gdef\csname R@an.hu.cyst.V7A1.phi\endcsname{36.9}
\expandafter\gdef\csname R@an.hu.cyst.Y2.phi\endcsname{35.0}
\expandafter\gdef\csname R@an.hu.cyst.base.phi\endcsname{36.1}
\expandafter\gdef\csname R@an.hu.cyst.p.max\endcsname{38}
\expandafter\gdef\csname R@an.hu.cyst.p.min\endcsname{35}
\expandafter\gdef\csname R@an.hu.flip.max\endcsname{38.8}
\expandafter\gdef\csname R@an.hu.flip.min\endcsname{28.0}
\expandafter\gdef\csname R@an.hu.grid.hi\endcsname{160}
\expandafter\gdef\csname R@an.hu.n\endcsname{1033}
\expandafter\gdef\csname R@an.hu.n.cyst\endcsname{593}
\expandafter\gdef\csname R@an.hu.n.tumour\endcsname{440}
\expandafter\gdef\csname R@an.hu.narms\endcsname{4}
\expandafter\gdef\csname R@an.hu.nin\endcsname{4}
\expandafter\gdef\csname R@an.hu.tumour.L1.flip\endcsname{-0.4}
\expandafter\gdef\csname R@an.hu.tumour.V7A1.flip\endcsname{1.9}
\expandafter\gdef\csname R@an.hu.tumour.Y2.flip\endcsname{2.8}
\expandafter\gdef\csname R@an.hu.tumour.base.flip\endcsname{-3.5}
\expandafter\gdef\csname R@an.hu.tumour.flip.absmax\endcsname{4}
\expandafter\gdef\csname R@an.hu.tumour.flip.max\endcsname{2.8}
\expandafter\gdef\csname R@an.hu.tumour.flip.min\endcsname{-3.5}
\expandafter\gdef\csname R@an.med2.cyst.dice.best\endcsname{0.54}
\expandafter\gdef\csname R@an.med2.cyst.hu\endcsname{20}
\expandafter\gdef\csname R@an.med2.cyst.n\endcsname{25}
\expandafter\gdef\csname R@an.med2.cyst.share\endcsname{50}
\expandafter\gdef\csname R@an.med2.inst.n\endcsname{441}
\makeatother

\makeatletter
\expandafter\gdef\csname R@med.abs.D2.all\endcsname{15}
\expandafter\gdef\csname R@med.abs.D2.bl\endcsname{2}
\expandafter\gdef\csname R@med.abs.D2.es\endcsname{0}
\expandafter\gdef\csname R@med.abs.D2.gb\endcsname{8}
\expandafter\gdef\csname R@med.abs.D2.lk\endcsname{1}
\expandafter\gdef\csname R@med.abs.D2.pu\endcsname{2}
\expandafter\gdef\csname R@med.abs.D2.st\endcsname{2}
\expandafter\gdef\csname R@med.abs.H1.all\endcsname{13}
\expandafter\gdef\csname R@med.abs.H1.bl\endcsname{2}
\expandafter\gdef\csname R@med.abs.H1.es\endcsname{0}
\expandafter\gdef\csname R@med.abs.H1.gb\endcsname{8}
\expandafter\gdef\csname R@med.abs.H1.lk\endcsname{0}
\expandafter\gdef\csname R@med.abs.H1.pu\endcsname{1}
\expandafter\gdef\csname R@med.abs.H1.st\endcsname{2}
\expandafter\gdef\csname R@med.abs.H2.all\endcsname{13}
\expandafter\gdef\csname R@med.abs.H2.bl\endcsname{0}
\expandafter\gdef\csname R@med.abs.H2.es\endcsname{0}
\expandafter\gdef\csname R@med.abs.H2.gb\endcsname{7}
\expandafter\gdef\csname R@med.abs.H2.lk\endcsname{1}
\expandafter\gdef\csname R@med.abs.H2.pu\endcsname{3}
\expandafter\gdef\csname R@med.abs.H2.st\endcsname{2}
\expandafter\gdef\csname R@med.abs.L1.all\endcsname{12}
\expandafter\gdef\csname R@med.abs.L1.bl\endcsname{1}
\expandafter\gdef\csname R@med.abs.L1.es\endcsname{0}
\expandafter\gdef\csname R@med.abs.L1.gb\endcsname{6}
\expandafter\gdef\csname R@med.abs.L1.lk\endcsname{0}
\expandafter\gdef\csname R@med.abs.L1.lt1ml\endcsname{7}
\expandafter\gdef\csname R@med.abs.L1.pu\endcsname{3}
\expandafter\gdef\csname R@med.abs.L1.st\endcsname{2}
\expandafter\gdef\csname R@med.abs.L3.all\endcsname{15}
\expandafter\gdef\csname R@med.abs.L3.bl\endcsname{2}
\expandafter\gdef\csname R@med.abs.L3.es\endcsname{0}
\expandafter\gdef\csname R@med.abs.L3.gb\endcsname{6}
\expandafter\gdef\csname R@med.abs.L3.lk\endcsname{1}
\expandafter\gdef\csname R@med.abs.L3.lt1ml\endcsname{5}
\expandafter\gdef\csname R@med.abs.L3.pu\endcsname{4}
\expandafter\gdef\csname R@med.abs.L3.st\endcsname{2}
\expandafter\gdef\csname R@med.abs.V5B1.all\endcsname{14}
\expandafter\gdef\csname R@med.abs.V5B1.bl\endcsname{1}
\expandafter\gdef\csname R@med.abs.V5B1.es\endcsname{0}
\expandafter\gdef\csname R@med.abs.V5B1.gb\endcsname{7}
\expandafter\gdef\csname R@med.abs.V5B1.lk\endcsname{1}
\expandafter\gdef\csname R@med.abs.V5B1.pu\endcsname{3}
\expandafter\gdef\csname R@med.abs.V5B1.st\endcsname{2}
\expandafter\gdef\csname R@med.abs.V5B2.all\endcsname{15}
\expandafter\gdef\csname R@med.abs.V5B2.bl\endcsname{2}
\expandafter\gdef\csname R@med.abs.V5B2.es\endcsname{0}
\expandafter\gdef\csname R@med.abs.V5B2.gb\endcsname{7}
\expandafter\gdef\csname R@med.abs.V5B2.lk\endcsname{1}
\expandafter\gdef\csname R@med.abs.V5B2.pu\endcsname{3}
\expandafter\gdef\csname R@med.abs.V5B2.st\endcsname{2}
\expandafter\gdef\csname R@med.abs.V7A1.all\endcsname{10}
\expandafter\gdef\csname R@med.abs.V7A1.bl\endcsname{1}
\expandafter\gdef\csname R@med.abs.V7A1.es\endcsname{0}
\expandafter\gdef\csname R@med.abs.V7A1.gb\endcsname{5}
\expandafter\gdef\csname R@med.abs.V7A1.lk\endcsname{1}
\expandafter\gdef\csname R@med.abs.V7A1.lt1ml\endcsname{3}
\expandafter\gdef\csname R@med.abs.V7A1.pu\endcsname{1}
\expandafter\gdef\csname R@med.abs.V7A1.st\endcsname{2}
\expandafter\gdef\csname R@med.abs.V7C1.all\endcsname{15}
\expandafter\gdef\csname R@med.abs.V7C1.bl\endcsname{1}
\expandafter\gdef\csname R@med.abs.V7C1.es\endcsname{0}
\expandafter\gdef\csname R@med.abs.V7C1.gb\endcsname{7}
\expandafter\gdef\csname R@med.abs.V7C1.lk\endcsname{1}
\expandafter\gdef\csname R@med.abs.V7C1.lt1ml\endcsname{6}
\expandafter\gdef\csname R@med.abs.V7C1.pu\endcsname{4}
\expandafter\gdef\csname R@med.abs.V7C1.st\endcsname{2}
\expandafter\gdef\csname R@med.abs.Y2.all\endcsname{13}
\expandafter\gdef\csname R@med.abs.Y2.bl\endcsname{1}
\expandafter\gdef\csname R@med.abs.Y2.es\endcsname{0}
\expandafter\gdef\csname R@med.abs.Y2.gb\endcsname{7}
\expandafter\gdef\csname R@med.abs.Y2.lk\endcsname{1}
\expandafter\gdef\csname R@med.abs.Y2.pu\endcsname{2}
\expandafter\gdef\csname R@med.abs.Y2.st\endcsname{2}
\expandafter\gdef\csname R@med.abs.base.all\endcsname{13}
\expandafter\gdef\csname R@med.abs.base.bl\endcsname{1}
\expandafter\gdef\csname R@med.abs.base.es\endcsname{0}
\expandafter\gdef\csname R@med.abs.base.gb\endcsname{5}
\expandafter\gdef\csname R@med.abs.base.lk\endcsname{1}
\expandafter\gdef\csname R@med.abs.base.lt1ml\endcsname{5}
\expandafter\gdef\csname R@med.abs.base.pu\endcsname{4}
\expandafter\gdef\csname R@med.abs.base.pu.small\endcsname{2}
\expandafter\gdef\csname R@med.abs.base.pu.smallml\endcsname{0.14}
\expandafter\gdef\csname R@med.abs.base.pu.smallvox\endcsname{38}
\expandafter\gdef\csname R@med.abs.base.st\endcsname{2}
\expandafter\gdef\csname R@med.abs.gb.big.ml\endcsname{79}
\expandafter\gdef\csname R@med.abs.gb.big.narms\endcsname{8}
\expandafter\gdef\csname R@med.abs.n\endcsname{25}
\expandafter\gdef\csname R@med.abs.n.bl\endcsname{5}
\expandafter\gdef\csname R@med.abs.n.es\endcsname{1}
\expandafter\gdef\csname R@med.abs.n.gb\endcsname{9}
\expandafter\gdef\csname R@med.abs.n.lk\endcsname{1}
\expandafter\gdef\csname R@med.abs.n.pu\endcsname{7}
\expandafter\gdef\csname R@med.abs.n.st\endcsname{2}
\expandafter\gdef\csname R@med.abs.narms\endcsname{8}
\expandafter\gdef\csname R@med.abs.pres.bl\endcsname{155}
\expandafter\gdef\csname R@med.abs.pres.es\endcsname{159}
\expandafter\gdef\csname R@med.abs.pres.gb\endcsname{151}
\expandafter\gdef\csname R@med.abs.pres.lk\endcsname{159}
\expandafter\gdef\csname R@med.abs.pres.pu\endcsname{153}
\expandafter\gdef\csname R@med.abs.pres.st\endcsname{158}
\expandafter\gdef\csname R@med.abs.publ\endcsname{5}
\expandafter\gdef\csname R@med.acdc.far.max\endcsname{1}
\expandafter\gdef\csname R@med.acdc.n\endcsname{160}
\expandafter\gdef\csname R@med.acdc.narms\endcsname{11}
\expandafter\gdef\csname R@med.amos.n5mm\endcsname{129}
\expandafter\gdef\csname R@med.amos.ncases\endcsname{160}
\expandafter\gdef\csname R@med.amos.norg\endcsname{15}
\expandafter\gdef\csname R@med.amos.zt\endcsname{2.0}
\expandafter\gdef\csname R@med.dec.D2.bl.abs\endcsname{\ensuremath{-0.57}}
\expandafter\gdef\csname R@med.dec.D2.bl.full\endcsname{\ensuremath{-1.00}}
\expandafter\gdef\csname R@med.dec.D2.bl.pres\endcsname{\ensuremath{-0.43}}
\expandafter\gdef\csname R@med.dec.D2.es.abs\endcsname{\ensuremath{0.00}}
\expandafter\gdef\csname R@med.dec.D2.es.full\endcsname{\ensuremath{-1.05}}
\expandafter\gdef\csname R@med.dec.D2.es.pres\endcsname{\ensuremath{-1.05}}
\expandafter\gdef\csname R@med.dec.D2.gb.abs\endcsname{\ensuremath{-1.53}}
\expandafter\gdef\csname R@med.dec.D2.gb.full\endcsname{\ensuremath{-2.32}}
\expandafter\gdef\csname R@med.dec.D2.gb.pres\endcsname{\ensuremath{-0.80}}
\expandafter\gdef\csname R@med.dec.D2.lk.abs\endcsname{\ensuremath{0.00}}
\expandafter\gdef\csname R@med.dec.D2.lk.full\endcsname{\ensuremath{+0.03}}
\expandafter\gdef\csname R@med.dec.D2.lk.pres\endcsname{\ensuremath{+0.03}}
\expandafter\gdef\csname R@med.dec.D2.mean.abs\endcsname{\ensuremath{-0.07}}
\expandafter\gdef\csname R@med.dec.D2.mean.abspct\endcsname{15}
\expandafter\gdef\csname R@med.dec.D2.mean.full\endcsname{\ensuremath{-0.44}}
\expandafter\gdef\csname R@med.dec.D2.mean.pres\endcsname{\ensuremath{-0.37}}
\expandafter\gdef\csname R@med.dec.D2.pu.abs\endcsname{\ensuremath{+1.08}}
\expandafter\gdef\csname R@med.dec.D2.pu.full\endcsname{\ensuremath{+0.92}}
\expandafter\gdef\csname R@med.dec.D2.pu.pres\endcsname{\ensuremath{-0.17}}
\expandafter\gdef\csname R@med.dec.D2.st.abs\endcsname{\ensuremath{0.00}}
\expandafter\gdef\csname R@med.dec.D2.st.full\endcsname{\ensuremath{-0.09}}
\expandafter\gdef\csname R@med.dec.D2.st.pres\endcsname{\ensuremath{-0.09}}
\expandafter\gdef\csname R@med.dec.H1.bl.abs\endcsname{\ensuremath{-0.60}}
\expandafter\gdef\csname R@med.dec.H1.bl.full\endcsname{\ensuremath{+0.74}}
\expandafter\gdef\csname R@med.dec.H1.bl.pres\endcsname{\ensuremath{+1.34}}
\expandafter\gdef\csname R@med.dec.H1.es.abs\endcsname{\ensuremath{0.00}}
\expandafter\gdef\csname R@med.dec.H1.es.full\endcsname{\ensuremath{-0.04}}
\expandafter\gdef\csname R@med.dec.H1.es.pres\endcsname{\ensuremath{-0.04}}
\expandafter\gdef\csname R@med.dec.H1.gb.abs\endcsname{\ensuremath{-1.60}}
\expandafter\gdef\csname R@med.dec.H1.gb.full\endcsname{\ensuremath{-1.10}}
\expandafter\gdef\csname R@med.dec.H1.gb.pres\endcsname{\ensuremath{+0.50}}
\expandafter\gdef\csname R@med.dec.H1.lk.abs\endcsname{\ensuremath{+0.59}}
\expandafter\gdef\csname R@med.dec.H1.lk.full\endcsname{\ensuremath{+0.23}}
\expandafter\gdef\csname R@med.dec.H1.lk.pres\endcsname{\ensuremath{-0.36}}
\expandafter\gdef\csname R@med.dec.H1.mean.abs\endcsname{\ensuremath{0.00}}
\expandafter\gdef\csname R@med.dec.H1.mean.full\endcsname{\ensuremath{+0.28}}
\expandafter\gdef\csname R@med.dec.H1.mean.pres\endcsname{\ensuremath{+0.28}}
\expandafter\gdef\csname R@med.dec.H1.pu.abs\endcsname{\ensuremath{+1.60}}
\expandafter\gdef\csname R@med.dec.H1.pu.full\endcsname{\ensuremath{+3.11}}
\expandafter\gdef\csname R@med.dec.H1.pu.pres\endcsname{\ensuremath{+1.51}}
\expandafter\gdef\csname R@med.dec.H1.st.abs\endcsname{\ensuremath{-0.01}}
\expandafter\gdef\csname R@med.dec.H1.st.full\endcsname{\ensuremath{+0.70}}
\expandafter\gdef\csname R@med.dec.H1.st.pres\endcsname{\ensuremath{+0.71}}
\expandafter\gdef\csname R@med.dec.H2.bl.abs\endcsname{\ensuremath{+0.57}}
\expandafter\gdef\csname R@med.dec.H2.bl.full\endcsname{\ensuremath{+1.68}}
\expandafter\gdef\csname R@med.dec.H2.bl.pres\endcsname{\ensuremath{+1.11}}
\expandafter\gdef\csname R@med.dec.H2.es.abs\endcsname{\ensuremath{0.00}}
\expandafter\gdef\csname R@med.dec.H2.es.full\endcsname{\ensuremath{-0.14}}
\expandafter\gdef\csname R@med.dec.H2.es.pres\endcsname{\ensuremath{-0.14}}
\expandafter\gdef\csname R@med.dec.H2.gb.abs\endcsname{\ensuremath{-1.05}}
\expandafter\gdef\csname R@med.dec.H2.gb.full\endcsname{\ensuremath{-0.77}}
\expandafter\gdef\csname R@med.dec.H2.gb.pres\endcsname{\ensuremath{+0.27}}
\expandafter\gdef\csname R@med.dec.H2.lk.abs\endcsname{\ensuremath{0.00}}
\expandafter\gdef\csname R@med.dec.H2.lk.full\endcsname{\ensuremath{-0.22}}
\expandafter\gdef\csname R@med.dec.H2.lk.pres\endcsname{\ensuremath{-0.22}}
\expandafter\gdef\csname R@med.dec.H2.mean.abs\endcsname{\ensuremath{0.00}}
\expandafter\gdef\csname R@med.dec.H2.mean.full\endcsname{\ensuremath{+0.22}}
\expandafter\gdef\csname R@med.dec.H2.mean.pres\endcsname{\ensuremath{+0.22}}
\expandafter\gdef\csname R@med.dec.H2.pu.abs\endcsname{\ensuremath{+0.50}}
\expandafter\gdef\csname R@med.dec.H2.pu.full\endcsname{\ensuremath{+1.84}}
\expandafter\gdef\csname R@med.dec.H2.pu.pres\endcsname{\ensuremath{+1.34}}
\expandafter\gdef\csname R@med.dec.H2.st.abs\endcsname{\ensuremath{0.00}}
\expandafter\gdef\csname R@med.dec.H2.st.full\endcsname{\ensuremath{+0.56}}
\expandafter\gdef\csname R@med.dec.H2.st.pres\endcsname{\ensuremath{+0.56}}
\expandafter\gdef\csname R@med.dec.L1.bl.abs\endcsname{\ensuremath{-0.03}}
\expandafter\gdef\csname R@med.dec.L1.bl.full\endcsname{\ensuremath{+1.44}}
\expandafter\gdef\csname R@med.dec.L1.bl.pres\endcsname{\ensuremath{+1.48}}
\expandafter\gdef\csname R@med.dec.L1.es.abs\endcsname{\ensuremath{0.00}}
\expandafter\gdef\csname R@med.dec.L1.es.full\endcsname{\ensuremath{-0.11}}
\expandafter\gdef\csname R@med.dec.L1.es.pres\endcsname{\ensuremath{-0.11}}
\expandafter\gdef\csname R@med.dec.L1.gb.abs\endcsname{\ensuremath{-0.53}}
\expandafter\gdef\csname R@med.dec.L1.gb.full\endcsname{\ensuremath{+0.66}}
\expandafter\gdef\csname R@med.dec.L1.gb.pres\endcsname{\ensuremath{+1.18}}
\expandafter\gdef\csname R@med.dec.L1.lk.abs\endcsname{\ensuremath{+0.59}}
\expandafter\gdef\csname R@med.dec.L1.lk.full\endcsname{\ensuremath{+0.27}}
\expandafter\gdef\csname R@med.dec.L1.lk.pres\endcsname{\ensuremath{-0.32}}
\expandafter\gdef\csname R@med.dec.L1.mean.abs\endcsname{\ensuremath{+0.04}}
\expandafter\gdef\csname R@med.dec.L1.mean.abspct\endcsname{9}
\expandafter\gdef\csname R@med.dec.L1.mean.full\endcsname{\ensuremath{+0.41}}
\expandafter\gdef\csname R@med.dec.L1.mean.pres\endcsname{\ensuremath{+0.37}}
\expandafter\gdef\csname R@med.dec.L1.pu.abs\endcsname{\ensuremath{+0.50}}
\expandafter\gdef\csname R@med.dec.L1.pu.full\endcsname{\ensuremath{+2.01}}
\expandafter\gdef\csname R@med.dec.L1.pu.pres\endcsname{\ensuremath{+1.51}}
\expandafter\gdef\csname R@med.dec.L1.st.abs\endcsname{\ensuremath{0.00}}
\expandafter\gdef\csname R@med.dec.L1.st.full\endcsname{\ensuremath{+0.49}}
\expandafter\gdef\csname R@med.dec.L1.st.pres\endcsname{\ensuremath{+0.49}}
\expandafter\gdef\csname R@med.dec.L3.bl.abs\endcsname{\ensuremath{-0.60}}
\expandafter\gdef\csname R@med.dec.L3.bl.full\endcsname{\ensuremath{+0.78}}
\expandafter\gdef\csname R@med.dec.L3.bl.pres\endcsname{\ensuremath{+1.38}}
\expandafter\gdef\csname R@med.dec.L3.es.abs\endcsname{\ensuremath{0.00}}
\expandafter\gdef\csname R@med.dec.L3.es.full\endcsname{\ensuremath{+0.07}}
\expandafter\gdef\csname R@med.dec.L3.es.pres\endcsname{\ensuremath{+0.07}}
\expandafter\gdef\csname R@med.dec.L3.gb.abs\endcsname{\ensuremath{-0.59}}
\expandafter\gdef\csname R@med.dec.L3.gb.full\endcsname{\ensuremath{+0.54}}
\expandafter\gdef\csname R@med.dec.L3.gb.pres\endcsname{\ensuremath{+1.13}}
\expandafter\gdef\csname R@med.dec.L3.lk.abs\endcsname{\ensuremath{0.00}}
\expandafter\gdef\csname R@med.dec.L3.lk.full\endcsname{\ensuremath{-0.40}}
\expandafter\gdef\csname R@med.dec.L3.lk.pres\endcsname{\ensuremath{-0.40}}
\expandafter\gdef\csname R@med.dec.L3.mean.abs\endcsname{\ensuremath{-0.08}}
\expandafter\gdef\csname R@med.dec.L3.mean.abspct\endcsname{-35}
\expandafter\gdef\csname R@med.dec.L3.mean.full\endcsname{\ensuremath{+0.23}}
\expandafter\gdef\csname R@med.dec.L3.mean.pres\endcsname{\ensuremath{+0.31}}
\expandafter\gdef\csname R@med.dec.L3.pu.abs\endcsname{\ensuremath{-0.02}}
\expandafter\gdef\csname R@med.dec.L3.pu.full\endcsname{\ensuremath{+1.09}}
\expandafter\gdef\csname R@med.dec.L3.pu.pres\endcsname{\ensuremath{+1.11}}
\expandafter\gdef\csname R@med.dec.L3.st.abs\endcsname{\ensuremath{-0.01}}
\expandafter\gdef\csname R@med.dec.L3.st.full\endcsname{\ensuremath{+0.92}}
\expandafter\gdef\csname R@med.dec.L3.st.pres\endcsname{\ensuremath{+0.93}}
\expandafter\gdef\csname R@med.dec.V5B1.bl.abs\endcsname{\ensuremath{0.00}}
\expandafter\gdef\csname R@med.dec.V5B1.bl.full\endcsname{\ensuremath{-0.33}}
\expandafter\gdef\csname R@med.dec.V5B1.bl.pres\endcsname{\ensuremath{-0.33}}
\expandafter\gdef\csname R@med.dec.V5B1.es.abs\endcsname{\ensuremath{0.00}}
\expandafter\gdef\csname R@med.dec.V5B1.es.full\endcsname{\ensuremath{-1.29}}
\expandafter\gdef\csname R@med.dec.V5B1.es.pres\endcsname{\ensuremath{-1.29}}
\expandafter\gdef\csname R@med.dec.V5B1.gb.abs\endcsname{\ensuremath{-1.01}}
\expandafter\gdef\csname R@med.dec.V5B1.gb.full\endcsname{\ensuremath{-1.81}}
\expandafter\gdef\csname R@med.dec.V5B1.gb.pres\endcsname{\ensuremath{-0.80}}
\expandafter\gdef\csname R@med.dec.V5B1.lk.abs\endcsname{\ensuremath{0.00}}
\expandafter\gdef\csname R@med.dec.V5B1.lk.full\endcsname{\ensuremath{-0.24}}
\expandafter\gdef\csname R@med.dec.V5B1.lk.pres\endcsname{\ensuremath{-0.24}}
\expandafter\gdef\csname R@med.dec.V5B1.mean.abs\endcsname{\ensuremath{-0.03}}
\expandafter\gdef\csname R@med.dec.V5B1.mean.abspct\endcsname{7}
\expandafter\gdef\csname R@med.dec.V5B1.mean.full\endcsname{\ensuremath{-0.46}}
\expandafter\gdef\csname R@med.dec.V5B1.mean.pres\endcsname{\ensuremath{-0.43}}
\expandafter\gdef\csname R@med.dec.V5B1.pu.abs\endcsname{\ensuremath{+0.55}}
\expandafter\gdef\csname R@med.dec.V5B1.pu.full\endcsname{\ensuremath{+0.59}}
\expandafter\gdef\csname R@med.dec.V5B1.pu.pres\endcsname{\ensuremath{+0.03}}
\expandafter\gdef\csname R@med.dec.V5B1.st.abs\endcsname{\ensuremath{0.00}}
\expandafter\gdef\csname R@med.dec.V5B1.st.full\endcsname{\ensuremath{-0.25}}
\expandafter\gdef\csname R@med.dec.V5B1.st.pres\endcsname{\ensuremath{-0.26}}
\expandafter\gdef\csname R@med.dec.V5B2.bl.abs\endcsname{\ensuremath{-0.60}}
\expandafter\gdef\csname R@med.dec.V5B2.bl.full\endcsname{\ensuremath{+0.30}}
\expandafter\gdef\csname R@med.dec.V5B2.bl.pres\endcsname{\ensuremath{+0.90}}
\expandafter\gdef\csname R@med.dec.V5B2.es.abs\endcsname{\ensuremath{0.00}}
\expandafter\gdef\csname R@med.dec.V5B2.es.full\endcsname{\ensuremath{-0.96}}
\expandafter\gdef\csname R@med.dec.V5B2.es.pres\endcsname{\ensuremath{-0.96}}
\expandafter\gdef\csname R@med.dec.V5B2.gb.abs\endcsname{\ensuremath{-1.03}}
\expandafter\gdef\csname R@med.dec.V5B2.gb.full\endcsname{\ensuremath{-1.21}}
\expandafter\gdef\csname R@med.dec.V5B2.gb.pres\endcsname{\ensuremath{-0.17}}
\expandafter\gdef\csname R@med.dec.V5B2.lk.abs\endcsname{\ensuremath{0.00}}
\expandafter\gdef\csname R@med.dec.V5B2.lk.full\endcsname{\ensuremath{-0.24}}
\expandafter\gdef\csname R@med.dec.V5B2.lk.pres\endcsname{\ensuremath{-0.25}}
\expandafter\gdef\csname R@med.dec.V5B2.mean.abs\endcsname{\ensuremath{-0.07}}
\expandafter\gdef\csname R@med.dec.V5B2.mean.abspct\endcsname{29}
\expandafter\gdef\csname R@med.dec.V5B2.mean.full\endcsname{\ensuremath{-0.25}}
\expandafter\gdef\csname R@med.dec.V5B2.mean.pres\endcsname{\ensuremath{-0.18}}
\expandafter\gdef\csname R@med.dec.V5B2.pu.abs\endcsname{\ensuremath{+0.50}}
\expandafter\gdef\csname R@med.dec.V5B2.pu.full\endcsname{\ensuremath{+1.10}}
\expandafter\gdef\csname R@med.dec.V5B2.pu.pres\endcsname{\ensuremath{+0.60}}
\expandafter\gdef\csname R@med.dec.V5B2.st.abs\endcsname{\ensuremath{0.00}}
\expandafter\gdef\csname R@med.dec.V5B2.st.full\endcsname{\ensuremath{-0.18}}
\expandafter\gdef\csname R@med.dec.V5B2.st.pres\endcsname{\ensuremath{-0.18}}
\expandafter\gdef\csname R@med.dec.V7A1.bl.abs\endcsname{\ensuremath{-0.01}}
\expandafter\gdef\csname R@med.dec.V7A1.bl.full\endcsname{\ensuremath{+1.30}}
\expandafter\gdef\csname R@med.dec.V7A1.bl.pres\endcsname{\ensuremath{+1.30}}
\expandafter\gdef\csname R@med.dec.V7A1.es.abs\endcsname{\ensuremath{0.00}}
\expandafter\gdef\csname R@med.dec.V7A1.es.full\endcsname{\ensuremath{-0.59}}
\expandafter\gdef\csname R@med.dec.V7A1.es.pres\endcsname{\ensuremath{-0.59}}
\expandafter\gdef\csname R@med.dec.V7A1.gb.abs\endcsname{\ensuremath{+0.03}}
\expandafter\gdef\csname R@med.dec.V7A1.gb.full\endcsname{\ensuremath{-0.24}}
\expandafter\gdef\csname R@med.dec.V7A1.gb.pres\endcsname{\ensuremath{-0.27}}
\expandafter\gdef\csname R@med.dec.V7A1.lk.abs\endcsname{\ensuremath{0.00}}
\expandafter\gdef\csname R@med.dec.V7A1.lk.full\endcsname{\ensuremath{-0.40}}
\expandafter\gdef\csname R@med.dec.V7A1.lk.pres\endcsname{\ensuremath{-0.41}}
\expandafter\gdef\csname R@med.dec.V7A1.mean.abs\endcsname{\ensuremath{+0.11}}
\expandafter\gdef\csname R@med.dec.V7A1.mean.abspct\endcsname{42}
\expandafter\gdef\csname R@med.dec.V7A1.mean.full\endcsname{\ensuremath{+0.26}}
\expandafter\gdef\csname R@med.dec.V7A1.mean.pres\endcsname{\ensuremath{+0.15}}
\expandafter\gdef\csname R@med.dec.V7A1.pu.abs\endcsname{\ensuremath{+1.60}}
\expandafter\gdef\csname R@med.dec.V7A1.pu.full\endcsname{\ensuremath{+2.85}}
\expandafter\gdef\csname R@med.dec.V7A1.pu.pres\endcsname{\ensuremath{+1.25}}
\expandafter\gdef\csname R@med.dec.V7A1.st.abs\endcsname{\ensuremath{-0.01}}
\expandafter\gdef\csname R@med.dec.V7A1.st.full\endcsname{\ensuremath{+0.53}}
\expandafter\gdef\csname R@med.dec.V7A1.st.pres\endcsname{\ensuremath{+0.53}}
\expandafter\gdef\csname R@med.dec.V7C1.bl.abs\endcsname{\ensuremath{+0.01}}
\expandafter\gdef\csname R@med.dec.V7C1.bl.full\endcsname{\ensuremath{-0.25}}
\expandafter\gdef\csname R@med.dec.V7C1.bl.pres\endcsname{\ensuremath{-0.25}}
\expandafter\gdef\csname R@med.dec.V7C1.es.abs\endcsname{\ensuremath{0.00}}
\expandafter\gdef\csname R@med.dec.V7C1.es.full\endcsname{\ensuremath{-1.31}}
\expandafter\gdef\csname R@med.dec.V7C1.es.pres\endcsname{\ensuremath{-1.31}}
\expandafter\gdef\csname R@med.dec.V7C1.gb.abs\endcsname{\ensuremath{-1.03}}
\expandafter\gdef\csname R@med.dec.V7C1.gb.full\endcsname{\ensuremath{-1.44}}
\expandafter\gdef\csname R@med.dec.V7C1.gb.pres\endcsname{\ensuremath{-0.41}}
\expandafter\gdef\csname R@med.dec.V7C1.lk.abs\endcsname{\ensuremath{0.00}}
\expandafter\gdef\csname R@med.dec.V7C1.lk.full\endcsname{\ensuremath{-0.46}}
\expandafter\gdef\csname R@med.dec.V7C1.lk.pres\endcsname{\ensuremath{-0.46}}
\expandafter\gdef\csname R@med.dec.V7C1.mean.abs\endcsname{\ensuremath{-0.07}}
\expandafter\gdef\csname R@med.dec.V7C1.mean.abspct\endcsname{15}
\expandafter\gdef\csname R@med.dec.V7C1.mean.full\endcsname{\ensuremath{-0.46}}
\expandafter\gdef\csname R@med.dec.V7C1.mean.pres\endcsname{\ensuremath{-0.39}}
\expandafter\gdef\csname R@med.dec.V7C1.pu.abs\endcsname{\ensuremath{-0.01}}
\expandafter\gdef\csname R@med.dec.V7C1.pu.full\endcsname{\ensuremath{+0.50}}
\expandafter\gdef\csname R@med.dec.V7C1.pu.pres\endcsname{\ensuremath{+0.51}}
\expandafter\gdef\csname R@med.dec.V7C1.st.abs\endcsname{\ensuremath{+0.01}}
\expandafter\gdef\csname R@med.dec.V7C1.st.full\endcsname{\ensuremath{-0.24}}
\expandafter\gdef\csname R@med.dec.V7C1.st.pres\endcsname{\ensuremath{-0.25}}
\expandafter\gdef\csname R@med.dec.Y2.bl.abs\endcsname{\ensuremath{-0.01}}
\expandafter\gdef\csname R@med.dec.Y2.bl.full\endcsname{\ensuremath{+1.11}}
\expandafter\gdef\csname R@med.dec.Y2.bl.pres\endcsname{\ensuremath{+1.11}}
\expandafter\gdef\csname R@med.dec.Y2.es.abs\endcsname{\ensuremath{0.00}}
\expandafter\gdef\csname R@med.dec.Y2.es.full\endcsname{\ensuremath{-0.50}}
\expandafter\gdef\csname R@med.dec.Y2.es.pres\endcsname{\ensuremath{-0.50}}
\expandafter\gdef\csname R@med.dec.Y2.gb.abs\endcsname{\ensuremath{-1.05}}
\expandafter\gdef\csname R@med.dec.Y2.gb.full\endcsname{\ensuremath{-0.80}}
\expandafter\gdef\csname R@med.dec.Y2.gb.pres\endcsname{\ensuremath{+0.25}}
\expandafter\gdef\csname R@med.dec.Y2.lk.abs\endcsname{\ensuremath{-0.01}}
\expandafter\gdef\csname R@med.dec.Y2.lk.full\endcsname{\ensuremath{+0.13}}
\expandafter\gdef\csname R@med.dec.Y2.lk.pres\endcsname{\ensuremath{+0.13}}
\expandafter\gdef\csname R@med.dec.Y2.mean.abs\endcsname{\ensuremath{0.00}}
\expandafter\gdef\csname R@med.dec.Y2.mean.full\endcsname{\ensuremath{+0.23}}
\expandafter\gdef\csname R@med.dec.Y2.mean.pres\endcsname{\ensuremath{+0.23}}
\expandafter\gdef\csname R@med.dec.Y2.pu.abs\endcsname{\ensuremath{+1.06}}
\expandafter\gdef\csname R@med.dec.Y2.pu.full\endcsname{\ensuremath{+2.55}}
\expandafter\gdef\csname R@med.dec.Y2.pu.pres\endcsname{\ensuremath{+1.50}}
\expandafter\gdef\csname R@med.dec.Y2.st.abs\endcsname{\ensuremath{0.00}}
\expandafter\gdef\csname R@med.dec.Y2.st.full\endcsname{\ensuremath{+0.66}}
\expandafter\gdef\csname R@med.dec.Y2.st.pres\endcsname{\ensuremath{+0.67}}
\expandafter\gdef\csname R@med.erf.V7A1.r90\endcsname{64.2}
\expandafter\gdef\csname R@med.erf.V7C1.r90\endcsname{65.2}
\expandafter\gdef\csname R@med.erf.base.r90\endcsname{64.5}
\expandafter\gdef\csname R@med.erf.rin\endcsname{49.2}
\expandafter\gdef\csname R@med.kits.f0.far.V7A1\endcsname{5}
\expandafter\gdef\csname R@med.kits.f0.far.V7C1\endcsname{8}
\expandafter\gdef\csname R@med.kits.f0.far.base\endcsname{17}
\expandafter\gdef\csname R@med.kits.f0.ft.V7A1\endcsname{11}
\expandafter\gdef\csname R@med.kits.f0.ft.V7C1\endcsname{11}
\expandafter\gdef\csname R@med.kits.f0.ft.base\endcsname{14}
\expandafter\gdef\csname R@med.kits.f0.ftfp.V7A1\endcsname{8}
\expandafter\gdef\csname R@med.kits.f0.ftfp.V7C1\endcsname{8}
\expandafter\gdef\csname R@med.kits.f0.ftfp.base\endcsname{10}
\expandafter\gdef\csname R@med.kits.f0.n\endcsname{98}
\expandafter\gdef\csname R@med.kits.far.H1\endcsname{21}
\expandafter\gdef\csname R@med.kits.far.H2\endcsname{30}
\expandafter\gdef\csname R@med.kits.far.L1\endcsname{23}
\expandafter\gdef\csname R@med.kits.far.L3\endcsname{41}
\expandafter\gdef\csname R@med.kits.far.V5B2\endcsname{49}
\expandafter\gdef\csname R@med.kits.far.V7A1\endcsname{35}
\expandafter\gdef\csname R@med.kits.far.V7C1\endcsname{36}
\expandafter\gdef\csname R@med.kits.far.X1\endcsname{45}
\expandafter\gdef\csname R@med.kits.far.X4\endcsname{44}
\expandafter\gdef\csname R@med.kits.far.X6\endcsname{25}
\expandafter\gdef\csname R@med.kits.far.Y1\endcsname{25}
\expandafter\gdef\csname R@med.kits.far.Y2\endcsname{18}
\expandafter\gdef\csname R@med.kits.far.base\endcsname{52}
\expandafter\gdef\csname R@med.kits.far.fp.H1\endcsname{20}
\expandafter\gdef\csname R@med.kits.far.fp.H2\endcsname{30}
\expandafter\gdef\csname R@med.kits.far.fp.L1\endcsname{22}
\expandafter\gdef\csname R@med.kits.far.fp.L3\endcsname{39}
\expandafter\gdef\csname R@med.kits.far.fp.V5B2\endcsname{44}
\expandafter\gdef\csname R@med.kits.far.fp.V7A1\endcsname{34}
\expandafter\gdef\csname R@med.kits.far.fp.V7C1\endcsname{34}
\expandafter\gdef\csname R@med.kits.far.fp.X1\endcsname{43}
\expandafter\gdef\csname R@med.kits.far.fp.X4\endcsname{43}
\expandafter\gdef\csname R@med.kits.far.fp.X6\endcsname{25}
\expandafter\gdef\csname R@med.kits.far.fp.Y1\endcsname{23}
\expandafter\gdef\csname R@med.kits.far.fp.Y2\endcsname{17}
\expandafter\gdef\csname R@med.kits.far.fp.base\endcsname{49}
\expandafter\gdef\csname R@med.kits.far.hd.H1\endcsname{101}
\expandafter\gdef\csname R@med.kits.far.hd.H2\endcsname{120}
\expandafter\gdef\csname R@med.kits.far.hd.L1\endcsname{122}
\expandafter\gdef\csname R@med.kits.far.hd.L3\endcsname{143}
\expandafter\gdef\csname R@med.kits.far.hd.V5B2\endcsname{136}
\expandafter\gdef\csname R@med.kits.far.hd.V7A1\endcsname{174}
\expandafter\gdef\csname R@med.kits.far.hd.V7C1\endcsname{118}
\expandafter\gdef\csname R@med.kits.far.hd.X1\endcsname{128}
\expandafter\gdef\csname R@med.kits.far.hd.X4\endcsname{115}
\expandafter\gdef\csname R@med.kits.far.hd.X6\endcsname{138}
\expandafter\gdef\csname R@med.kits.far.hd.Y1\endcsname{129}
\expandafter\gdef\csname R@med.kits.far.hd.Y2\endcsname{75}
\expandafter\gdef\csname R@med.kits.far.hd.base\endcsname{99}
\expandafter\gdef\csname R@med.kits.ft.D2\endcsname{48}
\expandafter\gdef\csname R@med.kits.ft.H1\endcsname{47}
\expandafter\gdef\csname R@med.kits.ft.H2\endcsname{46}
\expandafter\gdef\csname R@med.kits.ft.L1\endcsname{47}
\expandafter\gdef\csname R@med.kits.ft.L3\endcsname{52}
\expandafter\gdef\csname R@med.kits.ft.V5B1\endcsname{50}
\expandafter\gdef\csname R@med.kits.ft.V5B2\endcsname{57}
\expandafter\gdef\csname R@med.kits.ft.V7A1\endcsname{42}
\expandafter\gdef\csname R@med.kits.ft.V7C1\endcsname{55}
\expandafter\gdef\csname R@med.kits.ft.X1\endcsname{49}
\expandafter\gdef\csname R@med.kits.ft.X4\endcsname{52}
\expandafter\gdef\csname R@med.kits.ft.X6\endcsname{45}
\expandafter\gdef\csname R@med.kits.ft.Y1\endcsname{46}
\expandafter\gdef\csname R@med.kits.ft.Y2\endcsname{42}
\expandafter\gdef\csname R@med.kits.ft.base\endcsname{55}
\expandafter\gdef\csname R@med.kits.ftfp.H1\endcsname{24}
\expandafter\gdef\csname R@med.kits.ftfp.H2\endcsname{31}
\expandafter\gdef\csname R@med.kits.ftfp.L1\endcsname{26}
\expandafter\gdef\csname R@med.kits.ftfp.L3\endcsname{35}
\expandafter\gdef\csname R@med.kits.ftfp.V5B2\endcsname{40}
\expandafter\gdef\csname R@med.kits.ftfp.V7A1\endcsname{26}
\expandafter\gdef\csname R@med.kits.ftfp.V7C1\endcsname{35}
\expandafter\gdef\csname R@med.kits.ftfp.X1\endcsname{31}
\expandafter\gdef\csname R@med.kits.ftfp.X4\endcsname{37}
\expandafter\gdef\csname R@med.kits.ftfp.X6\endcsname{27}
\expandafter\gdef\csname R@med.kits.ftfp.Y1\endcsname{30}
\expandafter\gdef\csname R@med.kits.ftfp.Y2\endcsname{21}
\expandafter\gdef\csname R@med.kits.ftfp.base\endcsname{40}
\expandafter\gdef\csname R@med.kits.n\endcsname{391}
\expandafter\gdef\csname R@med.kits.thr.dice\endcsname{0.85}
\expandafter\gdef\csname R@med.kits.thr.hd\endcsname{20}
\expandafter\gdef\csname R@med.kits.thr.tum\endcsname{0.5}
\expandafter\gdef\csname R@med.kits.zmed\endcsname{3.0}
\expandafter\gdef\csname R@med.kits.zt\endcsname{1.0}
\expandafter\gdef\csname R@med.nsd.base.pgap\endcsname{0.079}
\expandafter\gdef\csname R@med.nsd.base.rhogap\endcsname{\ensuremath{+0.47}}
\expandafter\gdef\csname R@med.nsd.f0.nthin\endcsname{5}
\expandafter\gdef\csname R@med.nsd.f0.p\endcsname{0.51}
\expandafter\gdef\csname R@med.nsd.f0.pq\endcsname{\ensuremath{p=0.51}}
\expandafter\gdef\csname R@med.nsd.f0.rest.d1\endcsname{\ensuremath{-0.69}}
\expandafter\gdef\csname R@med.nsd.f0.rest.d2\endcsname{\ensuremath{-0.43}}
\expandafter\gdef\csname R@med.nsd.f0.rest.dice\endcsname{\ensuremath{-0.06}}
\expandafter\gdef\csname R@med.nsd.f0.rho\endcsname{\ensuremath{-0.19}}
\expandafter\gdef\csname R@med.nsd.f0.thin.d1\endcsname{\ensuremath{-0.42}}
\expandafter\gdef\csname R@med.nsd.f0.thin.d2\endcsname{\ensuremath{-0.20}}
\expandafter\gdef\csname R@med.nsd.f0.thin.dice\endcsname{\ensuremath{-0.47}}
\expandafter\gdef\csname R@med.nsd.f0.thin.names\endcsname{adrenal glands, oesophagus, gallbladder and pancreas}
\expandafter\gdef\csname R@med.nw.amos.H1.abs\endcsname{5}
\expandafter\gdef\csname R@med.nw.amos.H1.abs.band\endcsname{inside}
\expandafter\gdef\csname R@med.nw.amos.H1.abs.d\endcsname{\ensuremath{-0.7}}
\expandafter\gdef\csname R@med.nw.amos.H1.abs.dM\endcsname{\ensuremath{-1.0}}
\expandafter\gdef\csname R@med.nw.amos.H1.abs.z\endcsname{\ensuremath{-1.2}}
\expandafter\gdef\csname R@med.nw.amos.H1.dice\endcsname{88.30}
\expandafter\gdef\csname R@med.nw.amos.H1.dice.band\endcsname{above}
\expandafter\gdef\csname R@med.nw.amos.H1.dice.d\endcsname{\ensuremath{+0.59}}
\expandafter\gdef\csname R@med.nw.amos.H1.dice.dM\endcsname{\ensuremath{+0.64}}
\expandafter\gdef\csname R@med.nw.amos.H1.dice.z\endcsname{\ensuremath{+4.7}}
\expandafter\gdef\csname R@med.nw.amos.H1.landed\endcsname{1}
\expandafter\gdef\csname R@med.nw.amos.H1.nabs\endcsname{6}
\expandafter\gdef\csname R@med.nw.amos.H1.rec\endcsname{1}
\expandafter\gdef\csname R@med.nw.amos.H1.thin\endcsname{83.06}
\expandafter\gdef\csname R@med.nw.amos.H1.thin.band\endcsname{inside}
\expandafter\gdef\csname R@med.nw.amos.H1.thin.d\endcsname{\ensuremath{-0.01}}
\expandafter\gdef\csname R@med.nw.amos.H1.thin.dM\endcsname{\ensuremath{+0.02}}
\expandafter\gdef\csname R@med.nw.amos.H1.thin.z\endcsname{\ensuremath{-0.2}}
\expandafter\gdef\csname R@med.nw.amos.H1.thin1\endcsname{80.34}
\expandafter\gdef\csname R@med.nw.amos.H1.thin1.band\endcsname{above}
\expandafter\gdef\csname R@med.nw.amos.H1.thin1.d\endcsname{\ensuremath{+0.45}}
\expandafter\gdef\csname R@med.nw.amos.H1.thin1.dM\endcsname{\ensuremath{+0.34}}
\expandafter\gdef\csname R@med.nw.amos.H1.thin1.z\endcsname{\ensuremath{+3.7}}
\expandafter\gdef\csname R@med.nw.amos.H1.thin2\endcsname{88.71}
\expandafter\gdef\csname R@med.nw.amos.H1.thin2.band\endcsname{above}
\expandafter\gdef\csname R@med.nw.amos.H1.thin2.d\endcsname{\ensuremath{+0.29}}
\expandafter\gdef\csname R@med.nw.amos.H1.thin2.dM\endcsname{\ensuremath{+0.08}}
\expandafter\gdef\csname R@med.nw.amos.H1.thin2.z\endcsname{\ensuremath{+2.7}}
\expandafter\gdef\csname R@med.nw.amos.H2.abs\endcsname{5}
\expandafter\gdef\csname R@med.nw.amos.H2.abs.band\endcsname{inside}
\expandafter\gdef\csname R@med.nw.amos.H2.abs.d\endcsname{\ensuremath{-0.7}}
\expandafter\gdef\csname R@med.nw.amos.H2.abs.dM\endcsname{\ensuremath{-1.0}}
\expandafter\gdef\csname R@med.nw.amos.H2.abs.z\endcsname{\ensuremath{-1.2}}
\expandafter\gdef\csname R@med.nw.amos.H2.dice\endcsname{88.11}
\expandafter\gdef\csname R@med.nw.amos.H2.dice.band\endcsname{above}
\expandafter\gdef\csname R@med.nw.amos.H2.dice.d\endcsname{\ensuremath{+0.39}}
\expandafter\gdef\csname R@med.nw.amos.H2.dice.dM\endcsname{\ensuremath{+0.45}}
\expandafter\gdef\csname R@med.nw.amos.H2.dice.z\endcsname{\ensuremath{+3.1}}
\expandafter\gdef\csname R@med.nw.amos.H2.landed\endcsname{1}
\expandafter\gdef\csname R@med.nw.amos.H2.nabs\endcsname{6}
\expandafter\gdef\csname R@med.nw.amos.H2.rec\endcsname{1}
\expandafter\gdef\csname R@med.nw.amos.H2.thin\endcsname{83.04}
\expandafter\gdef\csname R@med.nw.amos.H2.thin.band\endcsname{inside}
\expandafter\gdef\csname R@med.nw.amos.H2.thin.d\endcsname{\ensuremath{-0.03}}
\expandafter\gdef\csname R@med.nw.amos.H2.thin.dM\endcsname{\ensuremath{0.00}}
\expandafter\gdef\csname R@med.nw.amos.H2.thin.z\endcsname{\ensuremath{-0.5}}
\expandafter\gdef\csname R@med.nw.amos.H2.thin1\endcsname{80.30}
\expandafter\gdef\csname R@med.nw.amos.H2.thin1.band\endcsname{above}
\expandafter\gdef\csname R@med.nw.amos.H2.thin1.d\endcsname{\ensuremath{+0.41}}
\expandafter\gdef\csname R@med.nw.amos.H2.thin1.dM\endcsname{\ensuremath{+0.30}}
\expandafter\gdef\csname R@med.nw.amos.H2.thin1.z\endcsname{\ensuremath{+3.3}}
\expandafter\gdef\csname R@med.nw.amos.H2.thin2\endcsname{88.83}
\expandafter\gdef\csname R@med.nw.amos.H2.thin2.band\endcsname{above}
\expandafter\gdef\csname R@med.nw.amos.H2.thin2.d\endcsname{\ensuremath{+0.41}}
\expandafter\gdef\csname R@med.nw.amos.H2.thin2.dM\endcsname{\ensuremath{+0.21}}
\expandafter\gdef\csname R@med.nw.amos.H2.thin2.z\endcsname{\ensuremath{+3.8}}
\expandafter\gdef\csname R@med.nw.amos.V7A1.abs.mean\endcsname{6}
\expandafter\gdef\csname R@med.nw.amos.V7A1.abs.n\endcsname{1}
\expandafter\gdef\csname R@med.nw.amos.V7A1.abs.s0\endcsname{6}
\expandafter\gdef\csname R@med.nw.amos.V7A1.dice.mean\endcsname{87.66}
\expandafter\gdef\csname R@med.nw.amos.V7A1.dice.n\endcsname{1}
\expandafter\gdef\csname R@med.nw.amos.V7A1.dice.s0\endcsname{87.66}
\expandafter\gdef\csname R@med.nw.amos.V7A1.gmac.mean\endcsname{513.5}
\expandafter\gdef\csname R@med.nw.amos.V7A1.gmac.n\endcsname{1}
\expandafter\gdef\csname R@med.nw.amos.V7A1.gmac.s0\endcsname{513.5}
\expandafter\gdef\csname R@med.nw.amos.V7A1.nabs.mean\endcsname{6}
\expandafter\gdef\csname R@med.nw.amos.V7A1.nabs.n\endcsname{1}
\expandafter\gdef\csname R@med.nw.amos.V7A1.nabs.s0\endcsname{6}
\expandafter\gdef\csname R@med.nw.amos.V7A1.thin.mean\endcsname{83.04}
\expandafter\gdef\csname R@med.nw.amos.V7A1.thin.n\endcsname{1}
\expandafter\gdef\csname R@med.nw.amos.V7A1.thin.s0\endcsname{83.04}
\expandafter\gdef\csname R@med.nw.amos.V7A1.thin1.mean\endcsname{80.00}
\expandafter\gdef\csname R@med.nw.amos.V7A1.thin1.n\endcsname{1}
\expandafter\gdef\csname R@med.nw.amos.V7A1.thin1.s0\endcsname{80.00}
\expandafter\gdef\csname R@med.nw.amos.V7A1.thin2.mean\endcsname{88.63}
\expandafter\gdef\csname R@med.nw.amos.V7A1.thin2.n\endcsname{1}
\expandafter\gdef\csname R@med.nw.amos.V7A1.thin2.s0\endcsname{88.63}
\expandafter\gdef\csname R@med.nw.amos.X1.abs\endcsname{5}
\expandafter\gdef\csname R@med.nw.amos.X1.abs.band\endcsname{inside}
\expandafter\gdef\csname R@med.nw.amos.X1.abs.d\endcsname{\ensuremath{-0.7}}
\expandafter\gdef\csname R@med.nw.amos.X1.abs.dM\endcsname{\ensuremath{-1.0}}
\expandafter\gdef\csname R@med.nw.amos.X1.abs.z\endcsname{\ensuremath{-1.2}}
\expandafter\gdef\csname R@med.nw.amos.X1.dice\endcsname{87.32}
\expandafter\gdef\csname R@med.nw.amos.X1.dice.band\endcsname{below}
\expandafter\gdef\csname R@med.nw.amos.X1.dice.d\endcsname{\ensuremath{-0.40}}
\expandafter\gdef\csname R@med.nw.amos.X1.dice.dM\endcsname{\ensuremath{-0.34}}
\expandafter\gdef\csname R@med.nw.amos.X1.dice.z\endcsname{\ensuremath{-3.2}}
\expandafter\gdef\csname R@med.nw.amos.X1.landed\endcsname{1}
\expandafter\gdef\csname R@med.nw.amos.X1.nabs\endcsname{6}
\expandafter\gdef\csname R@med.nw.amos.X1.rec\endcsname{1}
\expandafter\gdef\csname R@med.nw.amos.X1.thin\endcsname{82.17}
\expandafter\gdef\csname R@med.nw.amos.X1.thin.band\endcsname{below}
\expandafter\gdef\csname R@med.nw.amos.X1.thin.d\endcsname{\ensuremath{-0.90}}
\expandafter\gdef\csname R@med.nw.amos.X1.thin.dM\endcsname{\ensuremath{-0.87}}
\expandafter\gdef\csname R@med.nw.amos.X1.thin.z\endcsname{\ensuremath{-15.3}}
\expandafter\gdef\csname R@med.nw.amos.X1.thin1\endcsname{78.61}
\expandafter\gdef\csname R@med.nw.amos.X1.thin1.band\endcsname{below}
\expandafter\gdef\csname R@med.nw.amos.X1.thin1.d\endcsname{\ensuremath{-1.28}}
\expandafter\gdef\csname R@med.nw.amos.X1.thin1.dM\endcsname{\ensuremath{-1.39}}
\expandafter\gdef\csname R@med.nw.amos.X1.thin1.z\endcsname{\ensuremath{-10.6}}
\expandafter\gdef\csname R@med.nw.amos.X1.thin2\endcsname{87.53}
\expandafter\gdef\csname R@med.nw.amos.X1.thin2.band\endcsname{below}
\expandafter\gdef\csname R@med.nw.amos.X1.thin2.d\endcsname{\ensuremath{-0.89}}
\expandafter\gdef\csname R@med.nw.amos.X1.thin2.dM\endcsname{\ensuremath{-1.10}}
\expandafter\gdef\csname R@med.nw.amos.X1.thin2.z\endcsname{\ensuremath{-8.2}}
\expandafter\gdef\csname R@med.nw.amos.X2.abs\endcsname{6}
\expandafter\gdef\csname R@med.nw.amos.X2.abs.band\endcsname{inside}
\expandafter\gdef\csname R@med.nw.amos.X2.abs.d\endcsname{\ensuremath{+0.3}}
\expandafter\gdef\csname R@med.nw.amos.X2.abs.dM\endcsname{\ensuremath{0.0}}
\expandafter\gdef\csname R@med.nw.amos.X2.abs.z\endcsname{\ensuremath{+0.6}}
\expandafter\gdef\csname R@med.nw.amos.X2.dice\endcsname{87.80}
\expandafter\gdef\csname R@med.nw.amos.X2.dice.band\endcsname{inside}
\expandafter\gdef\csname R@med.nw.amos.X2.dice.d\endcsname{\ensuremath{+0.09}}
\expandafter\gdef\csname R@med.nw.amos.X2.dice.dM\endcsname{\ensuremath{+0.14}}
\expandafter\gdef\csname R@med.nw.amos.X2.dice.z\endcsname{\ensuremath{+0.7}}
\expandafter\gdef\csname R@med.nw.amos.X2.landed\endcsname{1}
\expandafter\gdef\csname R@med.nw.amos.X2.nabs\endcsname{6}
\expandafter\gdef\csname R@med.nw.amos.X2.rec\endcsname{1}
\expandafter\gdef\csname R@med.nw.amos.X2.thin\endcsname{82.99}
\expandafter\gdef\csname R@med.nw.amos.X2.thin.band\endcsname{inside}
\expandafter\gdef\csname R@med.nw.amos.X2.thin.d\endcsname{\ensuremath{-0.08}}
\expandafter\gdef\csname R@med.nw.amos.X2.thin.dM\endcsname{\ensuremath{-0.05}}
\expandafter\gdef\csname R@med.nw.amos.X2.thin.z\endcsname{\ensuremath{-1.3}}
\expandafter\gdef\csname R@med.nw.amos.X2.thin1\endcsname{79.79}
\expandafter\gdef\csname R@med.nw.amos.X2.thin1.band\endcsname{inside}
\expandafter\gdef\csname R@med.nw.amos.X2.thin1.d\endcsname{\ensuremath{-0.10}}
\expandafter\gdef\csname R@med.nw.amos.X2.thin1.dM\endcsname{\ensuremath{-0.21}}
\expandafter\gdef\csname R@med.nw.amos.X2.thin1.z\endcsname{\ensuremath{-0.8}}
\expandafter\gdef\csname R@med.nw.amos.X2.thin2\endcsname{88.43}
\expandafter\gdef\csname R@med.nw.amos.X2.thin2.band\endcsname{inside}
\expandafter\gdef\csname R@med.nw.amos.X2.thin2.d\endcsname{\ensuremath{+0.01}}
\expandafter\gdef\csname R@med.nw.amos.X2.thin2.dM\endcsname{\ensuremath{-0.20}}
\expandafter\gdef\csname R@med.nw.amos.X2.thin2.z\endcsname{\ensuremath{+0.1}}
\expandafter\gdef\csname R@med.nw.amos.X3.abs\endcsname{5}
\expandafter\gdef\csname R@med.nw.amos.X3.abs.band\endcsname{inside}
\expandafter\gdef\csname R@med.nw.amos.X3.abs.d\endcsname{\ensuremath{-0.7}}
\expandafter\gdef\csname R@med.nw.amos.X3.abs.dM\endcsname{\ensuremath{-1.0}}
\expandafter\gdef\csname R@med.nw.amos.X3.abs.z\endcsname{\ensuremath{-1.2}}
\expandafter\gdef\csname R@med.nw.amos.X3.dice\endcsname{88.11}
\expandafter\gdef\csname R@med.nw.amos.X3.dice.band\endcsname{above}
\expandafter\gdef\csname R@med.nw.amos.X3.dice.d\endcsname{\ensuremath{+0.40}}
\expandafter\gdef\csname R@med.nw.amos.X3.dice.dM\endcsname{\ensuremath{+0.46}}
\expandafter\gdef\csname R@med.nw.amos.X3.dice.z\endcsname{\ensuremath{+3.2}}
\expandafter\gdef\csname R@med.nw.amos.X3.landed\endcsname{1}
\expandafter\gdef\csname R@med.nw.amos.X3.nabs\endcsname{6}
\expandafter\gdef\csname R@med.nw.amos.X3.rec\endcsname{1}
\expandafter\gdef\csname R@med.nw.amos.X3.thin\endcsname{83.00}
\expandafter\gdef\csname R@med.nw.amos.X3.thin.band\endcsname{inside}
\expandafter\gdef\csname R@med.nw.amos.X3.thin.d\endcsname{\ensuremath{-0.07}}
\expandafter\gdef\csname R@med.nw.amos.X3.thin.dM\endcsname{\ensuremath{-0.04}}
\expandafter\gdef\csname R@med.nw.amos.X3.thin.z\endcsname{\ensuremath{-1.2}}
\expandafter\gdef\csname R@med.nw.amos.X3.thin1\endcsname{79.84}
\expandafter\gdef\csname R@med.nw.amos.X3.thin1.band\endcsname{inside}
\expandafter\gdef\csname R@med.nw.amos.X3.thin1.d\endcsname{\ensuremath{-0.05}}
\expandafter\gdef\csname R@med.nw.amos.X3.thin1.dM\endcsname{\ensuremath{-0.16}}
\expandafter\gdef\csname R@med.nw.amos.X3.thin1.z\endcsname{\ensuremath{-0.4}}
\expandafter\gdef\csname R@med.nw.amos.X3.thin2\endcsname{88.45}
\expandafter\gdef\csname R@med.nw.amos.X3.thin2.band\endcsname{inside}
\expandafter\gdef\csname R@med.nw.amos.X3.thin2.d\endcsname{\ensuremath{+0.03}}
\expandafter\gdef\csname R@med.nw.amos.X3.thin2.dM\endcsname{\ensuremath{-0.17}}
\expandafter\gdef\csname R@med.nw.amos.X3.thin2.z\endcsname{\ensuremath{+0.3}}
\expandafter\gdef\csname R@med.nw.amos.X4.abs\endcsname{4}
\expandafter\gdef\csname R@med.nw.amos.X4.abs.band\endcsname{below}
\expandafter\gdef\csname R@med.nw.amos.X4.abs.d\endcsname{\ensuremath{-1.7}}
\expandafter\gdef\csname R@med.nw.amos.X4.abs.dM\endcsname{\ensuremath{-2.0}}
\expandafter\gdef\csname R@med.nw.amos.X4.abs.z\endcsname{\ensuremath{-2.9}}
\expandafter\gdef\csname R@med.nw.amos.X4.dice\endcsname{87.56}
\expandafter\gdef\csname R@med.nw.amos.X4.dice.band\endcsname{inside}
\expandafter\gdef\csname R@med.nw.amos.X4.dice.d\endcsname{\ensuremath{-0.15}}
\expandafter\gdef\csname R@med.nw.amos.X4.dice.dM\endcsname{\ensuremath{-0.10}}
\expandafter\gdef\csname R@med.nw.amos.X4.dice.z\endcsname{\ensuremath{-1.2}}
\expandafter\gdef\csname R@med.nw.amos.X4.landed\endcsname{1}
\expandafter\gdef\csname R@med.nw.amos.X4.nabs\endcsname{6}
\expandafter\gdef\csname R@med.nw.amos.X4.rec\endcsname{1}
\expandafter\gdef\csname R@med.nw.amos.X4.thin\endcsname{82.12}
\expandafter\gdef\csname R@med.nw.amos.X4.thin.band\endcsname{below}
\expandafter\gdef\csname R@med.nw.amos.X4.thin.d\endcsname{\ensuremath{-0.95}}
\expandafter\gdef\csname R@med.nw.amos.X4.thin.dM\endcsname{\ensuremath{-0.92}}
\expandafter\gdef\csname R@med.nw.amos.X4.thin.z\endcsname{\ensuremath{-16.1}}
\expandafter\gdef\csname R@med.nw.amos.X4.thin1\endcsname{78.80}
\expandafter\gdef\csname R@med.nw.amos.X4.thin1.band\endcsname{below}
\expandafter\gdef\csname R@med.nw.amos.X4.thin1.d\endcsname{\ensuremath{-1.09}}
\expandafter\gdef\csname R@med.nw.amos.X4.thin1.dM\endcsname{\ensuremath{-1.20}}
\expandafter\gdef\csname R@med.nw.amos.X4.thin1.z\endcsname{\ensuremath{-9.0}}
\expandafter\gdef\csname R@med.nw.amos.X4.thin2\endcsname{87.75}
\expandafter\gdef\csname R@med.nw.amos.X4.thin2.band\endcsname{below}
\expandafter\gdef\csname R@med.nw.amos.X4.thin2.d\endcsname{\ensuremath{-0.67}}
\expandafter\gdef\csname R@med.nw.amos.X4.thin2.dM\endcsname{\ensuremath{-0.88}}
\expandafter\gdef\csname R@med.nw.amos.X4.thin2.z\endcsname{\ensuremath{-6.2}}
\expandafter\gdef\csname R@med.nw.amos.X5.abs\endcsname{4}
\expandafter\gdef\csname R@med.nw.amos.X5.abs.band\endcsname{below}
\expandafter\gdef\csname R@med.nw.amos.X5.abs.d\endcsname{\ensuremath{-1.7}}
\expandafter\gdef\csname R@med.nw.amos.X5.abs.dM\endcsname{\ensuremath{-2.0}}
\expandafter\gdef\csname R@med.nw.amos.X5.abs.z\endcsname{\ensuremath{-2.9}}
\expandafter\gdef\csname R@med.nw.amos.X5.dice\endcsname{87.81}
\expandafter\gdef\csname R@med.nw.amos.X5.dice.band\endcsname{inside}
\expandafter\gdef\csname R@med.nw.amos.X5.dice.d\endcsname{\ensuremath{+0.10}}
\expandafter\gdef\csname R@med.nw.amos.X5.dice.dM\endcsname{\ensuremath{+0.15}}
\expandafter\gdef\csname R@med.nw.amos.X5.dice.z\endcsname{\ensuremath{+0.8}}
\expandafter\gdef\csname R@med.nw.amos.X5.landed\endcsname{1}
\expandafter\gdef\csname R@med.nw.amos.X5.nabs\endcsname{6}
\expandafter\gdef\csname R@med.nw.amos.X5.rec\endcsname{1}
\expandafter\gdef\csname R@med.nw.amos.X5.thin\endcsname{82.22}
\expandafter\gdef\csname R@med.nw.amos.X5.thin.band\endcsname{below}
\expandafter\gdef\csname R@med.nw.amos.X5.thin.d\endcsname{\ensuremath{-0.85}}
\expandafter\gdef\csname R@med.nw.amos.X5.thin.dM\endcsname{\ensuremath{-0.82}}
\expandafter\gdef\csname R@med.nw.amos.X5.thin.z\endcsname{\ensuremath{-14.5}}
\expandafter\gdef\csname R@med.nw.amos.X5.thin1\endcsname{78.89}
\expandafter\gdef\csname R@med.nw.amos.X5.thin1.band\endcsname{below}
\expandafter\gdef\csname R@med.nw.amos.X5.thin1.d\endcsname{\ensuremath{-1.00}}
\expandafter\gdef\csname R@med.nw.amos.X5.thin1.dM\endcsname{\ensuremath{-1.11}}
\expandafter\gdef\csname R@med.nw.amos.X5.thin1.z\endcsname{\ensuremath{-8.2}}
\expandafter\gdef\csname R@med.nw.amos.X5.thin2\endcsname{87.96}
\expandafter\gdef\csname R@med.nw.amos.X5.thin2.band\endcsname{below}
\expandafter\gdef\csname R@med.nw.amos.X5.thin2.d\endcsname{\ensuremath{-0.46}}
\expandafter\gdef\csname R@med.nw.amos.X5.thin2.dM\endcsname{\ensuremath{-0.67}}
\expandafter\gdef\csname R@med.nw.amos.X5.thin2.z\endcsname{\ensuremath{-4.3}}
\expandafter\gdef\csname R@med.nw.amos.X6.abs\endcsname{5}
\expandafter\gdef\csname R@med.nw.amos.X6.abs.band\endcsname{inside}
\expandafter\gdef\csname R@med.nw.amos.X6.abs.d\endcsname{\ensuremath{-0.7}}
\expandafter\gdef\csname R@med.nw.amos.X6.abs.dM\endcsname{\ensuremath{-1.0}}
\expandafter\gdef\csname R@med.nw.amos.X6.abs.z\endcsname{\ensuremath{-1.2}}
\expandafter\gdef\csname R@med.nw.amos.X6.dice\endcsname{87.98}
\expandafter\gdef\csname R@med.nw.amos.X6.dice.band\endcsname{above}
\expandafter\gdef\csname R@med.nw.amos.X6.dice.d\endcsname{\ensuremath{+0.26}}
\expandafter\gdef\csname R@med.nw.amos.X6.dice.dM\endcsname{\ensuremath{+0.32}}
\expandafter\gdef\csname R@med.nw.amos.X6.dice.z\endcsname{\ensuremath{+2.1}}
\expandafter\gdef\csname R@med.nw.amos.X6.landed\endcsname{1}
\expandafter\gdef\csname R@med.nw.amos.X6.nabs\endcsname{6}
\expandafter\gdef\csname R@med.nw.amos.X6.rec\endcsname{1}
\expandafter\gdef\csname R@med.nw.amos.X6.thin\endcsname{82.86}
\expandafter\gdef\csname R@med.nw.amos.X6.thin.band\endcsname{below}
\expandafter\gdef\csname R@med.nw.amos.X6.thin.d\endcsname{\ensuremath{-0.21}}
\expandafter\gdef\csname R@med.nw.amos.X6.thin.dM\endcsname{\ensuremath{-0.18}}
\expandafter\gdef\csname R@med.nw.amos.X6.thin.z\endcsname{\ensuremath{-3.5}}
\expandafter\gdef\csname R@med.nw.amos.X6.thin1\endcsname{79.86}
\expandafter\gdef\csname R@med.nw.amos.X6.thin1.band\endcsname{inside}
\expandafter\gdef\csname R@med.nw.amos.X6.thin1.d\endcsname{\ensuremath{-0.03}}
\expandafter\gdef\csname R@med.nw.amos.X6.thin1.dM\endcsname{\ensuremath{-0.14}}
\expandafter\gdef\csname R@med.nw.amos.X6.thin1.z\endcsname{\ensuremath{-0.3}}
\expandafter\gdef\csname R@med.nw.amos.X6.thin2\endcsname{88.38}
\expandafter\gdef\csname R@med.nw.amos.X6.thin2.band\endcsname{inside}
\expandafter\gdef\csname R@med.nw.amos.X6.thin2.d\endcsname{\ensuremath{-0.04}}
\expandafter\gdef\csname R@med.nw.amos.X6.thin2.dM\endcsname{\ensuremath{-0.24}}
\expandafter\gdef\csname R@med.nw.amos.X6.thin2.z\endcsname{\ensuremath{-0.4}}
\expandafter\gdef\csname R@med.nw.amos.Y1.abs\endcsname{6}
\expandafter\gdef\csname R@med.nw.amos.Y1.abs.band\endcsname{inside}
\expandafter\gdef\csname R@med.nw.amos.Y1.abs.d\endcsname{\ensuremath{+0.3}}
\expandafter\gdef\csname R@med.nw.amos.Y1.abs.dM\endcsname{\ensuremath{0.0}}
\expandafter\gdef\csname R@med.nw.amos.Y1.abs.z\endcsname{\ensuremath{+0.6}}
\expandafter\gdef\csname R@med.nw.amos.Y1.dice\endcsname{87.90}
\expandafter\gdef\csname R@med.nw.amos.Y1.dice.band\endcsname{inside}
\expandafter\gdef\csname R@med.nw.amos.Y1.dice.d\endcsname{\ensuremath{+0.19}}
\expandafter\gdef\csname R@med.nw.amos.Y1.dice.dM\endcsname{\ensuremath{+0.24}}
\expandafter\gdef\csname R@med.nw.amos.Y1.dice.z\endcsname{\ensuremath{+1.5}}
\expandafter\gdef\csname R@med.nw.amos.Y1.landed\endcsname{1}
\expandafter\gdef\csname R@med.nw.amos.Y1.nabs\endcsname{6}
\expandafter\gdef\csname R@med.nw.amos.Y1.rec\endcsname{1}
\expandafter\gdef\csname R@med.nw.amos.Y1.thin\endcsname{82.94}
\expandafter\gdef\csname R@med.nw.amos.Y1.thin.band\endcsname{below}
\expandafter\gdef\csname R@med.nw.amos.Y1.thin.d\endcsname{\ensuremath{-0.13}}
\expandafter\gdef\csname R@med.nw.amos.Y1.thin.dM\endcsname{\ensuremath{-0.10}}
\expandafter\gdef\csname R@med.nw.amos.Y1.thin.z\endcsname{\ensuremath{-2.3}}
\expandafter\gdef\csname R@med.nw.amos.Y1.thin1\endcsname{79.95}
\expandafter\gdef\csname R@med.nw.amos.Y1.thin1.band\endcsname{inside}
\expandafter\gdef\csname R@med.nw.amos.Y1.thin1.d\endcsname{\ensuremath{+0.06}}
\expandafter\gdef\csname R@med.nw.amos.Y1.thin1.dM\endcsname{\ensuremath{-0.05}}
\expandafter\gdef\csname R@med.nw.amos.Y1.thin1.z\endcsname{\ensuremath{+0.5}}
\expandafter\gdef\csname R@med.nw.amos.Y1.thin2\endcsname{88.55}
\expandafter\gdef\csname R@med.nw.amos.Y1.thin2.band\endcsname{inside}
\expandafter\gdef\csname R@med.nw.amos.Y1.thin2.d\endcsname{\ensuremath{+0.12}}
\expandafter\gdef\csname R@med.nw.amos.Y1.thin2.dM\endcsname{\ensuremath{-0.08}}
\expandafter\gdef\csname R@med.nw.amos.Y1.thin2.z\endcsname{\ensuremath{+1.2}}
\expandafter\gdef\csname R@med.nw.amos.Y2.abs\endcsname{3}
\expandafter\gdef\csname R@med.nw.amos.Y2.abs.band\endcsname{below}
\expandafter\gdef\csname R@med.nw.amos.Y2.abs.d\endcsname{\ensuremath{-2.7}}
\expandafter\gdef\csname R@med.nw.amos.Y2.abs.dM\endcsname{\ensuremath{-3.0}}
\expandafter\gdef\csname R@med.nw.amos.Y2.abs.z\endcsname{\ensuremath{-4.6}}
\expandafter\gdef\csname R@med.nw.amos.Y2.dice\endcsname{88.36}
\expandafter\gdef\csname R@med.nw.amos.Y2.dice.band\endcsname{above}
\expandafter\gdef\csname R@med.nw.amos.Y2.dice.d\endcsname{\ensuremath{+0.65}}
\expandafter\gdef\csname R@med.nw.amos.Y2.dice.dM\endcsname{\ensuremath{+0.70}}
\expandafter\gdef\csname R@med.nw.amos.Y2.dice.z\endcsname{\ensuremath{+5.2}}
\expandafter\gdef\csname R@med.nw.amos.Y2.landed\endcsname{1}
\expandafter\gdef\csname R@med.nw.amos.Y2.nabs\endcsname{6}
\expandafter\gdef\csname R@med.nw.amos.Y2.rec\endcsname{1}
\expandafter\gdef\csname R@med.nw.amos.Y2.thin\endcsname{83.16}
\expandafter\gdef\csname R@med.nw.amos.Y2.thin.band\endcsname{inside}
\expandafter\gdef\csname R@med.nw.amos.Y2.thin.d\endcsname{\ensuremath{+0.09}}
\expandafter\gdef\csname R@med.nw.amos.Y2.thin.dM\endcsname{\ensuremath{+0.12}}
\expandafter\gdef\csname R@med.nw.amos.Y2.thin.z\endcsname{\ensuremath{+1.5}}
\expandafter\gdef\csname R@med.nw.amos.Y2.thin1\endcsname{80.05}
\expandafter\gdef\csname R@med.nw.amos.Y2.thin1.band\endcsname{inside}
\expandafter\gdef\csname R@med.nw.amos.Y2.thin1.d\endcsname{\ensuremath{+0.15}}
\expandafter\gdef\csname R@med.nw.amos.Y2.thin1.dM\endcsname{\ensuremath{+0.05}}
\expandafter\gdef\csname R@med.nw.amos.Y2.thin1.z\endcsname{\ensuremath{+1.3}}
\expandafter\gdef\csname R@med.nw.amos.Y2.thin2\endcsname{88.53}
\expandafter\gdef\csname R@med.nw.amos.Y2.thin2.band\endcsname{inside}
\expandafter\gdef\csname R@med.nw.amos.Y2.thin2.d\endcsname{\ensuremath{+0.11}}
\expandafter\gdef\csname R@med.nw.amos.Y2.thin2.dM\endcsname{\ensuremath{-0.10}}
\expandafter\gdef\csname R@med.nw.amos.Y2.thin2.z\endcsname{\ensuremath{+1.0}}
\expandafter\gdef\csname R@med.nw.amos.base.abs.hi\endcsname{6.82}
\expandafter\gdef\csname R@med.nw.amos.base.abs.lo\endcsname{4.51}
\expandafter\gdef\csname R@med.nw.amos.base.abs.mean\endcsname{5.7}
\expandafter\gdef\csname R@med.nw.amos.base.abs.n\endcsname{3}
\expandafter\gdef\csname R@med.nw.amos.base.abs.s3\endcsname{5}
\expandafter\gdef\csname R@med.nw.amos.base.abs.s4\endcsname{6}
\expandafter\gdef\csname R@med.nw.amos.base.abs.s5\endcsname{6}
\expandafter\gdef\csname R@med.nw.amos.base.abs.sd\endcsname{0.58}
\expandafter\gdef\csname R@med.nw.amos.base.dice.hi\endcsname{87.96}
\expandafter\gdef\csname R@med.nw.amos.base.dice.lo\endcsname{87.46}
\expandafter\gdef\csname R@med.nw.amos.base.dice.mean\endcsname{87.71}
\expandafter\gdef\csname R@med.nw.amos.base.dice.n\endcsname{3}
\expandafter\gdef\csname R@med.nw.amos.base.dice.s3\endcsname{87.79}
\expandafter\gdef\csname R@med.nw.amos.base.dice.s4\endcsname{87.77}
\expandafter\gdef\csname R@med.nw.amos.base.dice.s5\endcsname{87.57}
\expandafter\gdef\csname R@med.nw.amos.base.dice.sd\endcsname{0.12}
\expandafter\gdef\csname R@med.nw.amos.base.gmac.hi\endcsname{527.00}
\expandafter\gdef\csname R@med.nw.amos.base.gmac.lo\endcsname{527.00}
\expandafter\gdef\csname R@med.nw.amos.base.gmac.mean\endcsname{527.0}
\expandafter\gdef\csname R@med.nw.amos.base.gmac.n\endcsname{3}
\expandafter\gdef\csname R@med.nw.amos.base.gmac.s3\endcsname{527.0}
\expandafter\gdef\csname R@med.nw.amos.base.gmac.s4\endcsname{527.0}
\expandafter\gdef\csname R@med.nw.amos.base.gmac.s5\endcsname{527.0}
\expandafter\gdef\csname R@med.nw.amos.base.gmac.sd\endcsname{0.00}
\expandafter\gdef\csname R@med.nw.amos.base.nabs.hi\endcsname{6.00}
\expandafter\gdef\csname R@med.nw.amos.base.nabs.lo\endcsname{6.00}
\expandafter\gdef\csname R@med.nw.amos.base.nabs.mean\endcsname{6.0}
\expandafter\gdef\csname R@med.nw.amos.base.nabs.n\endcsname{3}
\expandafter\gdef\csname R@med.nw.amos.base.nabs.s3\endcsname{6}
\expandafter\gdef\csname R@med.nw.amos.base.nabs.s4\endcsname{6}
\expandafter\gdef\csname R@med.nw.amos.base.nabs.s5\endcsname{6}
\expandafter\gdef\csname R@med.nw.amos.base.nabs.sd\endcsname{0.00}
\expandafter\gdef\csname R@med.nw.amos.base.thin.hi\endcsname{83.19}
\expandafter\gdef\csname R@med.nw.amos.base.thin.lo\endcsname{82.95}
\expandafter\gdef\csname R@med.nw.amos.base.thin.mean\endcsname{83.07}
\expandafter\gdef\csname R@med.nw.amos.base.thin.n\endcsname{3}
\expandafter\gdef\csname R@med.nw.amos.base.thin.s3\endcsname{83.01}
\expandafter\gdef\csname R@med.nw.amos.base.thin.s4\endcsname{83.13}
\expandafter\gdef\csname R@med.nw.amos.base.thin.s5\endcsname{83.08}
\expandafter\gdef\csname R@med.nw.amos.base.thin.sd\endcsname{0.06}
\expandafter\gdef\csname R@med.nw.amos.base.thin1.hi\endcsname{80.14}
\expandafter\gdef\csname R@med.nw.amos.base.thin1.lo\endcsname{79.65}
\expandafter\gdef\csname R@med.nw.amos.base.thin1.mean\endcsname{79.89}
\expandafter\gdef\csname R@med.nw.amos.base.thin1.n\endcsname{3}
\expandafter\gdef\csname R@med.nw.amos.base.thin1.s3\endcsname{79.79}
\expandafter\gdef\csname R@med.nw.amos.base.thin1.s4\endcsname{79.86}
\expandafter\gdef\csname R@med.nw.amos.base.thin1.s5\endcsname{80.03}
\expandafter\gdef\csname R@med.nw.amos.base.thin1.sd\endcsname{0.12}
\expandafter\gdef\csname R@med.nw.amos.base.thin2.hi\endcsname{88.64}
\expandafter\gdef\csname R@med.nw.amos.base.thin2.lo\endcsname{88.21}
\expandafter\gdef\csname R@med.nw.amos.base.thin2.mean\endcsname{88.42}
\expandafter\gdef\csname R@med.nw.amos.base.thin2.n\endcsname{3}
\expandafter\gdef\csname R@med.nw.amos.base.thin2.s3\endcsname{88.35}
\expandafter\gdef\csname R@med.nw.amos.base.thin2.s4\endcsname{88.37}
\expandafter\gdef\csname R@med.nw.amos.base.thin2.s5\endcsname{88.55}
\expandafter\gdef\csname R@med.nw.amos.base.thin2.sd\endcsname{0.11}
\expandafter\gdef\csname R@med.nw.kits.H1.dice\endcsname{86.15}
\expandafter\gdef\csname R@med.nw.kits.H1.dice.band\endcsname{above}
\expandafter\gdef\csname R@med.nw.kits.H1.dice.d\endcsname{\ensuremath{+1.89}}
\expandafter\gdef\csname R@med.nw.kits.H1.dice.dM\endcsname{\ensuremath{+0.15}}
\expandafter\gdef\csname R@med.nw.kits.H1.dice.z\endcsname{\ensuremath{+3.0}}
\expandafter\gdef\csname R@med.nw.kits.H1.far\endcsname{6}
\expandafter\gdef\csname R@med.nw.kits.H1.far.band\endcsname{below}
\expandafter\gdef\csname R@med.nw.kits.H1.far.d\endcsname{\ensuremath{-6.1}}
\expandafter\gdef\csname R@med.nw.kits.H1.far.dM\endcsname{\ensuremath{-0.3}}
\expandafter\gdef\csname R@med.nw.kits.H1.far.z\endcsname{\ensuremath{-2.1}}
\expandafter\gdef\csname R@med.nw.kits.H1.farfp\endcsname{6}
\expandafter\gdef\csname R@med.nw.kits.H1.farfp.band\endcsname{below}
\expandafter\gdef\csname R@med.nw.kits.H1.farfp.d\endcsname{\ensuremath{-5.9}}
\expandafter\gdef\csname R@med.nw.kits.H1.farfp.dM\endcsname{\ensuremath{-0.3}}
\expandafter\gdef\csname R@med.nw.kits.H1.farfp.z\endcsname{\ensuremath{-2.0}}
\expandafter\gdef\csname R@med.nw.kits.H1.ft\endcsname{12}
\expandafter\gdef\csname R@med.nw.kits.H1.ft.band\endcsname{inside}
\expandafter\gdef\csname R@med.nw.kits.H1.ft.d\endcsname{\ensuremath{-2.1}}
\expandafter\gdef\csname R@med.nw.kits.H1.ft.dM\endcsname{\ensuremath{+1.0}}
\expandafter\gdef\csname R@med.nw.kits.H1.ft.z\endcsname{\ensuremath{-1.7}}
\expandafter\gdef\csname R@med.nw.kits.H1.ftfp\endcsname{7}
\expandafter\gdef\csname R@med.nw.kits.H1.ftfp.band\endcsname{below}
\expandafter\gdef\csname R@med.nw.kits.H1.ftfp.d\endcsname{\ensuremath{-3.1}}
\expandafter\gdef\csname R@med.nw.kits.H1.ftfp.dM\endcsname{\ensuremath{0.0}}
\expandafter\gdef\csname R@med.nw.kits.H1.ftfp.z\endcsname{\ensuremath{-2.1}}
\expandafter\gdef\csname R@med.nw.kits.H1.landed\endcsname{1}
\expandafter\gdef\csname R@med.nw.kits.H1.rec\endcsname{1}
\expandafter\gdef\csname R@med.nw.kits.H1.tc\endcsname{4.76}
\expandafter\gdef\csname R@med.nw.kits.H1.tc.band\endcsname{inside}
\expandafter\gdef\csname R@med.nw.kits.H1.tc.d\endcsname{\ensuremath{-0.75}}
\expandafter\gdef\csname R@med.nw.kits.H1.tc.dM\endcsname{\ensuremath{-0.85}}
\expandafter\gdef\csname R@med.nw.kits.H1.tc.z\endcsname{\ensuremath{-1.2}}
\expandafter\gdef\csname R@med.nw.kits.H1.tcn\endcsname{5}
\expandafter\gdef\csname R@med.nw.kits.H1.tcn.band\endcsname{inside}
\expandafter\gdef\csname R@med.nw.kits.H1.tcn.d\endcsname{\ensuremath{+0.7}}
\expandafter\gdef\csname R@med.nw.kits.H1.tcn.dM\endcsname{\ensuremath{-0.3}}
\expandafter\gdef\csname R@med.nw.kits.H1.tcn.z\endcsname{\ensuremath{+0.7}}
\expandafter\gdef\csname R@med.nw.kits.H1.zthick\endcsname{85.78}
\expandafter\gdef\csname R@med.nw.kits.H1.zthick.band\endcsname{inside}
\expandafter\gdef\csname R@med.nw.kits.H1.zthick.d\endcsname{\ensuremath{+1.94}}
\expandafter\gdef\csname R@med.nw.kits.H1.zthick.dM\endcsname{\ensuremath{-0.24}}
\expandafter\gdef\csname R@med.nw.kits.H1.zthick.z\endcsname{\ensuremath{+2.0}}
\expandafter\gdef\csname R@med.nw.kits.H1.zthin\endcsname{86.48}
\expandafter\gdef\csname R@med.nw.kits.H1.zthin.band\endcsname{above}
\expandafter\gdef\csname R@med.nw.kits.H1.zthin.d\endcsname{\ensuremath{+1.85}}
\expandafter\gdef\csname R@med.nw.kits.H1.zthin.dM\endcsname{\ensuremath{+0.50}}
\expandafter\gdef\csname R@med.nw.kits.H1.zthin.z\endcsname{\ensuremath{+2.1}}
\expandafter\gdef\csname R@med.nw.kits.H2.dice\endcsname{86.32}
\expandafter\gdef\csname R@med.nw.kits.H2.dice.band\endcsname{above}
\expandafter\gdef\csname R@med.nw.kits.H2.dice.d\endcsname{\ensuremath{+2.07}}
\expandafter\gdef\csname R@med.nw.kits.H2.dice.dM\endcsname{\ensuremath{+0.32}}
\expandafter\gdef\csname R@med.nw.kits.H2.dice.z\endcsname{\ensuremath{+3.3}}
\expandafter\gdef\csname R@med.nw.kits.H2.far\endcsname{8}
\expandafter\gdef\csname R@med.nw.kits.H2.far.band\endcsname{inside}
\expandafter\gdef\csname R@med.nw.kits.H2.far.d\endcsname{\ensuremath{-4.1}}
\expandafter\gdef\csname R@med.nw.kits.H2.far.dM\endcsname{\ensuremath{+1.7}}
\expandafter\gdef\csname R@med.nw.kits.H2.far.z\endcsname{\ensuremath{-1.4}}
\expandafter\gdef\csname R@med.nw.kits.H2.farfp\endcsname{8}
\expandafter\gdef\csname R@med.nw.kits.H2.farfp.band\endcsname{inside}
\expandafter\gdef\csname R@med.nw.kits.H2.farfp.d\endcsname{\ensuremath{-3.9}}
\expandafter\gdef\csname R@med.nw.kits.H2.farfp.dM\endcsname{\ensuremath{+1.7}}
\expandafter\gdef\csname R@med.nw.kits.H2.farfp.z\endcsname{\ensuremath{-1.3}}
\expandafter\gdef\csname R@med.nw.kits.H2.ft\endcsname{13}
\expandafter\gdef\csname R@med.nw.kits.H2.ft.band\endcsname{inside}
\expandafter\gdef\csname R@med.nw.kits.H2.ft.d\endcsname{\ensuremath{-1.1}}
\expandafter\gdef\csname R@med.nw.kits.H2.ft.dM\endcsname{\ensuremath{+2.0}}
\expandafter\gdef\csname R@med.nw.kits.H2.ft.z\endcsname{\ensuremath{-0.9}}
\expandafter\gdef\csname R@med.nw.kits.H2.ftfp\endcsname{7}
\expandafter\gdef\csname R@med.nw.kits.H2.ftfp.band\endcsname{below}
\expandafter\gdef\csname R@med.nw.kits.H2.ftfp.d\endcsname{\ensuremath{-3.1}}
\expandafter\gdef\csname R@med.nw.kits.H2.ftfp.dM\endcsname{\ensuremath{0.0}}
\expandafter\gdef\csname R@med.nw.kits.H2.ftfp.z\endcsname{\ensuremath{-2.1}}
\expandafter\gdef\csname R@med.nw.kits.H2.landed\endcsname{1}
\expandafter\gdef\csname R@med.nw.kits.H2.rec\endcsname{1}
\expandafter\gdef\csname R@med.nw.kits.H2.tc\endcsname{5.92}
\expandafter\gdef\csname R@med.nw.kits.H2.tc.band\endcsname{inside}
\expandafter\gdef\csname R@med.nw.kits.H2.tc.d\endcsname{\ensuremath{+0.41}}
\expandafter\gdef\csname R@med.nw.kits.H2.tc.dM\endcsname{\ensuremath{+0.31}}
\expandafter\gdef\csname R@med.nw.kits.H2.tc.z\endcsname{\ensuremath{+0.6}}
\expandafter\gdef\csname R@med.nw.kits.H2.tcn\endcsname{6}
\expandafter\gdef\csname R@med.nw.kits.H2.tcn.band\endcsname{inside}
\expandafter\gdef\csname R@med.nw.kits.H2.tcn.d\endcsname{\ensuremath{+1.7}}
\expandafter\gdef\csname R@med.nw.kits.H2.tcn.dM\endcsname{\ensuremath{+0.7}}
\expandafter\gdef\csname R@med.nw.kits.H2.tcn.z\endcsname{\ensuremath{+1.7}}
\expandafter\gdef\csname R@med.nw.kits.H2.zthick\endcsname{86.45}
\expandafter\gdef\csname R@med.nw.kits.H2.zthick.band\endcsname{above}
\expandafter\gdef\csname R@med.nw.kits.H2.zthick.d\endcsname{\ensuremath{+2.61}}
\expandafter\gdef\csname R@med.nw.kits.H2.zthick.dM\endcsname{\ensuremath{+0.43}}
\expandafter\gdef\csname R@med.nw.kits.H2.zthick.z\endcsname{\ensuremath{+2.7}}
\expandafter\gdef\csname R@med.nw.kits.H2.zthin\endcsname{86.21}
\expandafter\gdef\csname R@med.nw.kits.H2.zthin.band\endcsname{inside}
\expandafter\gdef\csname R@med.nw.kits.H2.zthin.d\endcsname{\ensuremath{+1.59}}
\expandafter\gdef\csname R@med.nw.kits.H2.zthin.dM\endcsname{\ensuremath{+0.23}}
\expandafter\gdef\csname R@med.nw.kits.H2.zthin.z\endcsname{\ensuremath{+1.8}}
\expandafter\gdef\csname R@med.nw.kits.V7A1.dice.hi\endcsname{86.69}
\expandafter\gdef\csname R@med.nw.kits.V7A1.dice.lo\endcsname{85.32}
\expandafter\gdef\csname R@med.nw.kits.V7A1.dice.mean\endcsname{86.00}
\expandafter\gdef\csname R@med.nw.kits.V7A1.dice.n\endcsname{3}
\expandafter\gdef\csname R@med.nw.kits.V7A1.dice.s0\endcsname{85.93}
\expandafter\gdef\csname R@med.nw.kits.V7A1.dice.s1\endcsname{85.70}
\expandafter\gdef\csname R@med.nw.kits.V7A1.dice.s2\endcsname{86.38}
\expandafter\gdef\csname R@med.nw.kits.V7A1.dice.sd\endcsname{0.34}
\expandafter\gdef\csname R@med.nw.kits.V7A1.far.hi\endcsname{9.39}
\expandafter\gdef\csname R@med.nw.kits.V7A1.far.lo\endcsname{3.28}
\expandafter\gdef\csname R@med.nw.kits.V7A1.far.mean\endcsname{6.3}
\expandafter\gdef\csname R@med.nw.kits.V7A1.far.n\endcsname{3}
\expandafter\gdef\csname R@med.nw.kits.V7A1.far.s0\endcsname{5}
\expandafter\gdef\csname R@med.nw.kits.V7A1.far.s1\endcsname{8}
\expandafter\gdef\csname R@med.nw.kits.V7A1.far.s2\endcsname{6}
\expandafter\gdef\csname R@med.nw.kits.V7A1.far.sd\endcsname{1.53}
\expandafter\gdef\csname R@med.nw.kits.V7A1.farfp.hi\endcsname{9.39}
\expandafter\gdef\csname R@med.nw.kits.V7A1.farfp.lo\endcsname{3.28}
\expandafter\gdef\csname R@med.nw.kits.V7A1.farfp.mean\endcsname{6.3}
\expandafter\gdef\csname R@med.nw.kits.V7A1.farfp.n\endcsname{3}
\expandafter\gdef\csname R@med.nw.kits.V7A1.farfp.s0\endcsname{5}
\expandafter\gdef\csname R@med.nw.kits.V7A1.farfp.s1\endcsname{8}
\expandafter\gdef\csname R@med.nw.kits.V7A1.farfp.s2\endcsname{6}
\expandafter\gdef\csname R@med.nw.kits.V7A1.farfp.sd\endcsname{1.53}
\expandafter\gdef\csname R@med.nw.kits.V7A1.ft.hi\endcsname{13.00}
\expandafter\gdef\csname R@med.nw.kits.V7A1.ft.lo\endcsname{9.00}
\expandafter\gdef\csname R@med.nw.kits.V7A1.ft.mean\endcsname{11.0}
\expandafter\gdef\csname R@med.nw.kits.V7A1.ft.n\endcsname{3}
\expandafter\gdef\csname R@med.nw.kits.V7A1.ft.s0\endcsname{11}
\expandafter\gdef\csname R@med.nw.kits.V7A1.ft.s1\endcsname{12}
\expandafter\gdef\csname R@med.nw.kits.V7A1.ft.s2\endcsname{10}
\expandafter\gdef\csname R@med.nw.kits.V7A1.ft.sd\endcsname{1.00}
\expandafter\gdef\csname R@med.nw.kits.V7A1.ftfp.hi\endcsname{9.00}
\expandafter\gdef\csname R@med.nw.kits.V7A1.ftfp.lo\endcsname{5.00}
\expandafter\gdef\csname R@med.nw.kits.V7A1.ftfp.mean\endcsname{7.0}
\expandafter\gdef\csname R@med.nw.kits.V7A1.ftfp.n\endcsname{3}
\expandafter\gdef\csname R@med.nw.kits.V7A1.ftfp.s0\endcsname{8}
\expandafter\gdef\csname R@med.nw.kits.V7A1.ftfp.s1\endcsname{6}
\expandafter\gdef\csname R@med.nw.kits.V7A1.ftfp.s2\endcsname{7}
\expandafter\gdef\csname R@med.nw.kits.V7A1.ftfp.sd\endcsname{1.00}
\expandafter\gdef\csname R@med.nw.kits.V7A1.gmac.hi\endcsname{373.50}
\expandafter\gdef\csname R@med.nw.kits.V7A1.gmac.lo\endcsname{373.50}
\expandafter\gdef\csname R@med.nw.kits.V7A1.gmac.mean\endcsname{373.5}
\expandafter\gdef\csname R@med.nw.kits.V7A1.gmac.n\endcsname{3}
\expandafter\gdef\csname R@med.nw.kits.V7A1.gmac.s0\endcsname{373.5}
\expandafter\gdef\csname R@med.nw.kits.V7A1.gmac.s1\endcsname{373.5}
\expandafter\gdef\csname R@med.nw.kits.V7A1.gmac.s2\endcsname{373.5}
\expandafter\gdef\csname R@med.nw.kits.V7A1.gmac.sd\endcsname{0.00}
\expandafter\gdef\csname R@med.nw.kits.V7A1.tc.hi\endcsname{5.88}
\expandafter\gdef\csname R@med.nw.kits.V7A1.tc.lo\endcsname{5.34}
\expandafter\gdef\csname R@med.nw.kits.V7A1.tc.mean\endcsname{5.61}
\expandafter\gdef\csname R@med.nw.kits.V7A1.tc.n\endcsname{3}
\expandafter\gdef\csname R@med.nw.kits.V7A1.tc.s0\endcsname{5.60}
\expandafter\gdef\csname R@med.nw.kits.V7A1.tc.s1\endcsname{5.75}
\expandafter\gdef\csname R@med.nw.kits.V7A1.tc.s2\endcsname{5.48}
\expandafter\gdef\csname R@med.nw.kits.V7A1.tc.sd\endcsname{0.14}
\expandafter\gdef\csname R@med.nw.kits.V7A1.tcn.hi\endcsname{6.49}
\expandafter\gdef\csname R@med.nw.kits.V7A1.tcn.lo\endcsname{4.18}
\expandafter\gdef\csname R@med.nw.kits.V7A1.tcn.mean\endcsname{5.3}
\expandafter\gdef\csname R@med.nw.kits.V7A1.tcn.n\endcsname{3}
\expandafter\gdef\csname R@med.nw.kits.V7A1.tcn.s0\endcsname{6}
\expandafter\gdef\csname R@med.nw.kits.V7A1.tcn.s1\endcsname{5}
\expandafter\gdef\csname R@med.nw.kits.V7A1.tcn.s2\endcsname{5}
\expandafter\gdef\csname R@med.nw.kits.V7A1.tcn.sd\endcsname{0.58}
\expandafter\gdef\csname R@med.nw.kits.V7A1.zthick.hi\endcsname{86.92}
\expandafter\gdef\csname R@med.nw.kits.V7A1.zthick.lo\endcsname{85.14}
\expandafter\gdef\csname R@med.nw.kits.V7A1.zthick.mean\endcsname{86.03}
\expandafter\gdef\csname R@med.nw.kits.V7A1.zthick.n\endcsname{3}
\expandafter\gdef\csname R@med.nw.kits.V7A1.zthick.s0\endcsname{85.95}
\expandafter\gdef\csname R@med.nw.kits.V7A1.zthick.s1\endcsname{85.63}
\expandafter\gdef\csname R@med.nw.kits.V7A1.zthick.s2\endcsname{86.51}
\expandafter\gdef\csname R@med.nw.kits.V7A1.zthick.sd\endcsname{0.44}
\expandafter\gdef\csname R@med.nw.kits.V7A1.zthin.hi\endcsname{86.48}
\expandafter\gdef\csname R@med.nw.kits.V7A1.zthin.lo\endcsname{85.48}
\expandafter\gdef\csname R@med.nw.kits.V7A1.zthin.mean\endcsname{85.98}
\expandafter\gdef\csname R@med.nw.kits.V7A1.zthin.n\endcsname{3}
\expandafter\gdef\csname R@med.nw.kits.V7A1.zthin.s0\endcsname{85.91}
\expandafter\gdef\csname R@med.nw.kits.V7A1.zthin.s1\endcsname{85.77}
\expandafter\gdef\csname R@med.nw.kits.V7A1.zthin.s2\endcsname{86.26}
\expandafter\gdef\csname R@med.nw.kits.V7A1.zthin.sd\endcsname{0.25}
\expandafter\gdef\csname R@med.nw.kits.X1.dice\endcsname{86.51}
\expandafter\gdef\csname R@med.nw.kits.X1.dice.band\endcsname{above}
\expandafter\gdef\csname R@med.nw.kits.X1.dice.d\endcsname{\ensuremath{+2.26}}
\expandafter\gdef\csname R@med.nw.kits.X1.dice.dM\endcsname{\ensuremath{+0.51}}
\expandafter\gdef\csname R@med.nw.kits.X1.dice.z\endcsname{\ensuremath{+3.6}}
\expandafter\gdef\csname R@med.nw.kits.X1.far\endcsname{4}
\expandafter\gdef\csname R@med.nw.kits.X1.far.band\endcsname{below}
\expandafter\gdef\csname R@med.nw.kits.X1.far.d\endcsname{\ensuremath{-8.1}}
\expandafter\gdef\csname R@med.nw.kits.X1.far.dM\endcsname{\ensuremath{-2.3}}
\expandafter\gdef\csname R@med.nw.kits.X1.far.z\endcsname{\ensuremath{-2.8}}
\expandafter\gdef\csname R@med.nw.kits.X1.farfp\endcsname{4}
\expandafter\gdef\csname R@med.nw.kits.X1.farfp.band\endcsname{below}
\expandafter\gdef\csname R@med.nw.kits.X1.farfp.d\endcsname{\ensuremath{-7.9}}
\expandafter\gdef\csname R@med.nw.kits.X1.farfp.dM\endcsname{\ensuremath{-2.3}}
\expandafter\gdef\csname R@med.nw.kits.X1.farfp.z\endcsname{\ensuremath{-2.7}}
\expandafter\gdef\csname R@med.nw.kits.X1.ft\endcsname{13}
\expandafter\gdef\csname R@med.nw.kits.X1.ft.band\endcsname{inside}
\expandafter\gdef\csname R@med.nw.kits.X1.ft.d\endcsname{\ensuremath{-1.1}}
\expandafter\gdef\csname R@med.nw.kits.X1.ft.dM\endcsname{\ensuremath{+2.0}}
\expandafter\gdef\csname R@med.nw.kits.X1.ft.z\endcsname{\ensuremath{-0.9}}
\expandafter\gdef\csname R@med.nw.kits.X1.ftfp\endcsname{7}
\expandafter\gdef\csname R@med.nw.kits.X1.ftfp.band\endcsname{below}
\expandafter\gdef\csname R@med.nw.kits.X1.ftfp.d\endcsname{\ensuremath{-3.1}}
\expandafter\gdef\csname R@med.nw.kits.X1.ftfp.dM\endcsname{\ensuremath{0.0}}
\expandafter\gdef\csname R@med.nw.kits.X1.ftfp.z\endcsname{\ensuremath{-2.1}}
\expandafter\gdef\csname R@med.nw.kits.X1.landed\endcsname{1}
\expandafter\gdef\csname R@med.nw.kits.X1.rec\endcsname{1}
\expandafter\gdef\csname R@med.nw.kits.X1.tc\endcsname{4.88}
\expandafter\gdef\csname R@med.nw.kits.X1.tc.band\endcsname{inside}
\expandafter\gdef\csname R@med.nw.kits.X1.tc.d\endcsname{\ensuremath{-0.63}}
\expandafter\gdef\csname R@med.nw.kits.X1.tc.dM\endcsname{\ensuremath{-0.73}}
\expandafter\gdef\csname R@med.nw.kits.X1.tc.z\endcsname{\ensuremath{-1.0}}
\expandafter\gdef\csname R@med.nw.kits.X1.tcn\endcsname{6}
\expandafter\gdef\csname R@med.nw.kits.X1.tcn.band\endcsname{inside}
\expandafter\gdef\csname R@med.nw.kits.X1.tcn.d\endcsname{\ensuremath{+1.7}}
\expandafter\gdef\csname R@med.nw.kits.X1.tcn.dM\endcsname{\ensuremath{+0.7}}
\expandafter\gdef\csname R@med.nw.kits.X1.tcn.z\endcsname{\ensuremath{+1.7}}
\expandafter\gdef\csname R@med.nw.kits.X1.zthick\endcsname{85.80}
\expandafter\gdef\csname R@med.nw.kits.X1.zthick.band\endcsname{above}
\expandafter\gdef\csname R@med.nw.kits.X1.zthick.d\endcsname{\ensuremath{+1.95}}
\expandafter\gdef\csname R@med.nw.kits.X1.zthick.dM\endcsname{\ensuremath{-0.23}}
\expandafter\gdef\csname R@med.nw.kits.X1.zthick.z\endcsname{\ensuremath{+2.0}}
\expandafter\gdef\csname R@med.nw.kits.X1.zthin\endcsname{87.15}
\expandafter\gdef\csname R@med.nw.kits.X1.zthin.band\endcsname{above}
\expandafter\gdef\csname R@med.nw.kits.X1.zthin.d\endcsname{\ensuremath{+2.52}}
\expandafter\gdef\csname R@med.nw.kits.X1.zthin.dM\endcsname{\ensuremath{+1.17}}
\expandafter\gdef\csname R@med.nw.kits.X1.zthin.z\endcsname{\ensuremath{+2.9}}
\expandafter\gdef\csname R@med.nw.kits.X2.dice\endcsname{86.62}
\expandafter\gdef\csname R@med.nw.kits.X2.dice.band\endcsname{above}
\expandafter\gdef\csname R@med.nw.kits.X2.dice.d\endcsname{\ensuremath{+2.36}}
\expandafter\gdef\csname R@med.nw.kits.X2.dice.dM\endcsname{\ensuremath{+0.61}}
\expandafter\gdef\csname R@med.nw.kits.X2.dice.z\endcsname{\ensuremath{+3.7}}
\expandafter\gdef\csname R@med.nw.kits.X2.far\endcsname{8}
\expandafter\gdef\csname R@med.nw.kits.X2.far.band\endcsname{inside}
\expandafter\gdef\csname R@med.nw.kits.X2.far.d\endcsname{\ensuremath{-4.1}}
\expandafter\gdef\csname R@med.nw.kits.X2.far.dM\endcsname{\ensuremath{+1.7}}
\expandafter\gdef\csname R@med.nw.kits.X2.far.z\endcsname{\ensuremath{-1.4}}
\expandafter\gdef\csname R@med.nw.kits.X2.farfp\endcsname{8}
\expandafter\gdef\csname R@med.nw.kits.X2.farfp.band\endcsname{inside}
\expandafter\gdef\csname R@med.nw.kits.X2.farfp.d\endcsname{\ensuremath{-3.9}}
\expandafter\gdef\csname R@med.nw.kits.X2.farfp.dM\endcsname{\ensuremath{+1.7}}
\expandafter\gdef\csname R@med.nw.kits.X2.farfp.z\endcsname{\ensuremath{-1.3}}
\expandafter\gdef\csname R@med.nw.kits.X2.ft\endcsname{10}
\expandafter\gdef\csname R@med.nw.kits.X2.ft.band\endcsname{below}
\expandafter\gdef\csname R@med.nw.kits.X2.ft.d\endcsname{\ensuremath{-4.1}}
\expandafter\gdef\csname R@med.nw.kits.X2.ft.dM\endcsname{\ensuremath{-1.0}}
\expandafter\gdef\csname R@med.nw.kits.X2.ft.z\endcsname{\ensuremath{-3.2}}
\expandafter\gdef\csname R@med.nw.kits.X2.ftfp\endcsname{5}
\expandafter\gdef\csname R@med.nw.kits.X2.ftfp.band\endcsname{below}
\expandafter\gdef\csname R@med.nw.kits.X2.ftfp.d\endcsname{\ensuremath{-5.1}}
\expandafter\gdef\csname R@med.nw.kits.X2.ftfp.dM\endcsname{\ensuremath{-2.0}}
\expandafter\gdef\csname R@med.nw.kits.X2.ftfp.z\endcsname{\ensuremath{-3.5}}
\expandafter\gdef\csname R@med.nw.kits.X2.landed\endcsname{1}
\expandafter\gdef\csname R@med.nw.kits.X2.rec\endcsname{1}
\expandafter\gdef\csname R@med.nw.kits.X2.tc\endcsname{5.22}
\expandafter\gdef\csname R@med.nw.kits.X2.tc.band\endcsname{inside}
\expandafter\gdef\csname R@med.nw.kits.X2.tc.d\endcsname{\ensuremath{-0.28}}
\expandafter\gdef\csname R@med.nw.kits.X2.tc.dM\endcsname{\ensuremath{-0.38}}
\expandafter\gdef\csname R@med.nw.kits.X2.tc.z\endcsname{\ensuremath{-0.4}}
\expandafter\gdef\csname R@med.nw.kits.X2.tcn\endcsname{6}
\expandafter\gdef\csname R@med.nw.kits.X2.tcn.band\endcsname{inside}
\expandafter\gdef\csname R@med.nw.kits.X2.tcn.d\endcsname{\ensuremath{+1.7}}
\expandafter\gdef\csname R@med.nw.kits.X2.tcn.dM\endcsname{\ensuremath{+0.7}}
\expandafter\gdef\csname R@med.nw.kits.X2.tcn.z\endcsname{\ensuremath{+1.7}}
\expandafter\gdef\csname R@med.nw.kits.X2.zthick\endcsname{85.72}
\expandafter\gdef\csname R@med.nw.kits.X2.zthick.band\endcsname{inside}
\expandafter\gdef\csname R@med.nw.kits.X2.zthick.d\endcsname{\ensuremath{+1.87}}
\expandafter\gdef\csname R@med.nw.kits.X2.zthick.dM\endcsname{\ensuremath{-0.31}}
\expandafter\gdef\csname R@med.nw.kits.X2.zthick.z\endcsname{\ensuremath{+1.9}}
\expandafter\gdef\csname R@med.nw.kits.X2.zthin\endcsname{87.41}
\expandafter\gdef\csname R@med.nw.kits.X2.zthin.band\endcsname{above}
\expandafter\gdef\csname R@med.nw.kits.X2.zthin.d\endcsname{\ensuremath{+2.79}}
\expandafter\gdef\csname R@med.nw.kits.X2.zthin.dM\endcsname{\ensuremath{+1.43}}
\expandafter\gdef\csname R@med.nw.kits.X2.zthin.z\endcsname{\ensuremath{+3.2}}
\expandafter\gdef\csname R@med.nw.kits.X3.dice\endcsname{85.18}
\expandafter\gdef\csname R@med.nw.kits.X3.dice.band\endcsname{inside}
\expandafter\gdef\csname R@med.nw.kits.X3.dice.d\endcsname{\ensuremath{+0.92}}
\expandafter\gdef\csname R@med.nw.kits.X3.dice.dM\endcsname{\ensuremath{-0.83}}
\expandafter\gdef\csname R@med.nw.kits.X3.dice.z\endcsname{\ensuremath{+1.5}}
\expandafter\gdef\csname R@med.nw.kits.X3.far\endcsname{9}
\expandafter\gdef\csname R@med.nw.kits.X3.far.band\endcsname{inside}
\expandafter\gdef\csname R@med.nw.kits.X3.far.d\endcsname{\ensuremath{-3.1}}
\expandafter\gdef\csname R@med.nw.kits.X3.far.dM\endcsname{\ensuremath{+2.7}}
\expandafter\gdef\csname R@med.nw.kits.X3.far.z\endcsname{\ensuremath{-1.1}}
\expandafter\gdef\csname R@med.nw.kits.X3.farfp\endcsname{8}
\expandafter\gdef\csname R@med.nw.kits.X3.farfp.band\endcsname{inside}
\expandafter\gdef\csname R@med.nw.kits.X3.farfp.d\endcsname{\ensuremath{-3.9}}
\expandafter\gdef\csname R@med.nw.kits.X3.farfp.dM\endcsname{\ensuremath{+1.7}}
\expandafter\gdef\csname R@med.nw.kits.X3.farfp.z\endcsname{\ensuremath{-1.3}}
\expandafter\gdef\csname R@med.nw.kits.X3.ft\endcsname{13}
\expandafter\gdef\csname R@med.nw.kits.X3.ft.band\endcsname{inside}
\expandafter\gdef\csname R@med.nw.kits.X3.ft.d\endcsname{\ensuremath{-1.1}}
\expandafter\gdef\csname R@med.nw.kits.X3.ft.dM\endcsname{\ensuremath{+2.0}}
\expandafter\gdef\csname R@med.nw.kits.X3.ft.z\endcsname{\ensuremath{-0.9}}
\expandafter\gdef\csname R@med.nw.kits.X3.ftfp\endcsname{10}
\expandafter\gdef\csname R@med.nw.kits.X3.ftfp.band\endcsname{inside}
\expandafter\gdef\csname R@med.nw.kits.X3.ftfp.d\endcsname{\ensuremath{-0.1}}
\expandafter\gdef\csname R@med.nw.kits.X3.ftfp.dM\endcsname{\ensuremath{+3.0}}
\expandafter\gdef\csname R@med.nw.kits.X3.ftfp.z\endcsname{\ensuremath{-0.1}}
\expandafter\gdef\csname R@med.nw.kits.X3.landed\endcsname{1}
\expandafter\gdef\csname R@med.nw.kits.X3.rec\endcsname{1}
\expandafter\gdef\csname R@med.nw.kits.X3.tc\endcsname{6.10}
\expandafter\gdef\csname R@med.nw.kits.X3.tc.band\endcsname{inside}
\expandafter\gdef\csname R@med.nw.kits.X3.tc.d\endcsname{\ensuremath{+0.59}}
\expandafter\gdef\csname R@med.nw.kits.X3.tc.dM\endcsname{\ensuremath{+0.49}}
\expandafter\gdef\csname R@med.nw.kits.X3.tc.z\endcsname{\ensuremath{+0.9}}
\expandafter\gdef\csname R@med.nw.kits.X3.tcn\endcsname{6}
\expandafter\gdef\csname R@med.nw.kits.X3.tcn.band\endcsname{inside}
\expandafter\gdef\csname R@med.nw.kits.X3.tcn.d\endcsname{\ensuremath{+1.7}}
\expandafter\gdef\csname R@med.nw.kits.X3.tcn.dM\endcsname{\ensuremath{+0.7}}
\expandafter\gdef\csname R@med.nw.kits.X3.tcn.z\endcsname{\ensuremath{+1.7}}
\expandafter\gdef\csname R@med.nw.kits.X3.zthick\endcsname{85.01}
\expandafter\gdef\csname R@med.nw.kits.X3.zthick.band\endcsname{inside}
\expandafter\gdef\csname R@med.nw.kits.X3.zthick.d\endcsname{\ensuremath{+1.16}}
\expandafter\gdef\csname R@med.nw.kits.X3.zthick.dM\endcsname{\ensuremath{-1.02}}
\expandafter\gdef\csname R@med.nw.kits.X3.zthick.z\endcsname{\ensuremath{+1.2}}
\expandafter\gdef\csname R@med.nw.kits.X3.zthin\endcsname{85.32}
\expandafter\gdef\csname R@med.nw.kits.X3.zthin.band\endcsname{inside}
\expandafter\gdef\csname R@med.nw.kits.X3.zthin.d\endcsname{\ensuremath{+0.70}}
\expandafter\gdef\csname R@med.nw.kits.X3.zthin.dM\endcsname{\ensuremath{-0.66}}
\expandafter\gdef\csname R@med.nw.kits.X3.zthin.z\endcsname{\ensuremath{+0.8}}
\expandafter\gdef\csname R@med.nw.kits.X4.dice\endcsname{86.58}
\expandafter\gdef\csname R@med.nw.kits.X4.dice.band\endcsname{above}
\expandafter\gdef\csname R@med.nw.kits.X4.dice.d\endcsname{\ensuremath{+2.32}}
\expandafter\gdef\csname R@med.nw.kits.X4.dice.dM\endcsname{\ensuremath{+0.57}}
\expandafter\gdef\csname R@med.nw.kits.X4.dice.z\endcsname{\ensuremath{+3.7}}
\expandafter\gdef\csname R@med.nw.kits.X4.far\endcsname{6}
\expandafter\gdef\csname R@med.nw.kits.X4.far.band\endcsname{below}
\expandafter\gdef\csname R@med.nw.kits.X4.far.d\endcsname{\ensuremath{-6.1}}
\expandafter\gdef\csname R@med.nw.kits.X4.far.dM\endcsname{\ensuremath{-0.3}}
\expandafter\gdef\csname R@med.nw.kits.X4.far.z\endcsname{\ensuremath{-2.1}}
\expandafter\gdef\csname R@med.nw.kits.X4.farfp\endcsname{6}
\expandafter\gdef\csname R@med.nw.kits.X4.farfp.band\endcsname{below}
\expandafter\gdef\csname R@med.nw.kits.X4.farfp.d\endcsname{\ensuremath{-5.9}}
\expandafter\gdef\csname R@med.nw.kits.X4.farfp.dM\endcsname{\ensuremath{-0.3}}
\expandafter\gdef\csname R@med.nw.kits.X4.farfp.z\endcsname{\ensuremath{-2.0}}
\expandafter\gdef\csname R@med.nw.kits.X4.ft\endcsname{10}
\expandafter\gdef\csname R@med.nw.kits.X4.ft.band\endcsname{below}
\expandafter\gdef\csname R@med.nw.kits.X4.ft.d\endcsname{\ensuremath{-4.1}}
\expandafter\gdef\csname R@med.nw.kits.X4.ft.dM\endcsname{\ensuremath{-1.0}}
\expandafter\gdef\csname R@med.nw.kits.X4.ft.z\endcsname{\ensuremath{-3.2}}
\expandafter\gdef\csname R@med.nw.kits.X4.ftfp\endcsname{7}
\expandafter\gdef\csname R@med.nw.kits.X4.ftfp.band\endcsname{below}
\expandafter\gdef\csname R@med.nw.kits.X4.ftfp.d\endcsname{\ensuremath{-3.1}}
\expandafter\gdef\csname R@med.nw.kits.X4.ftfp.dM\endcsname{\ensuremath{0.0}}
\expandafter\gdef\csname R@med.nw.kits.X4.ftfp.z\endcsname{\ensuremath{-2.1}}
\expandafter\gdef\csname R@med.nw.kits.X4.landed\endcsname{1}
\expandafter\gdef\csname R@med.nw.kits.X4.rec\endcsname{1}
\expandafter\gdef\csname R@med.nw.kits.X4.tc\endcsname{4.96}
\expandafter\gdef\csname R@med.nw.kits.X4.tc.band\endcsname{inside}
\expandafter\gdef\csname R@med.nw.kits.X4.tc.d\endcsname{\ensuremath{-0.55}}
\expandafter\gdef\csname R@med.nw.kits.X4.tc.dM\endcsname{\ensuremath{-0.65}}
\expandafter\gdef\csname R@med.nw.kits.X4.tc.z\endcsname{\ensuremath{-0.9}}
\expandafter\gdef\csname R@med.nw.kits.X4.tcn\endcsname{4}
\expandafter\gdef\csname R@med.nw.kits.X4.tcn.band\endcsname{inside}
\expandafter\gdef\csname R@med.nw.kits.X4.tcn.d\endcsname{\ensuremath{-0.3}}
\expandafter\gdef\csname R@med.nw.kits.X4.tcn.dM\endcsname{\ensuremath{-1.3}}
\expandafter\gdef\csname R@med.nw.kits.X4.tcn.z\endcsname{\ensuremath{-0.3}}
\expandafter\gdef\csname R@med.nw.kits.X4.zthick\endcsname{86.28}
\expandafter\gdef\csname R@med.nw.kits.X4.zthick.band\endcsname{above}
\expandafter\gdef\csname R@med.nw.kits.X4.zthick.d\endcsname{\ensuremath{+2.43}}
\expandafter\gdef\csname R@med.nw.kits.X4.zthick.dM\endcsname{\ensuremath{+0.25}}
\expandafter\gdef\csname R@med.nw.kits.X4.zthick.z\endcsname{\ensuremath{+2.5}}
\expandafter\gdef\csname R@med.nw.kits.X4.zthin\endcsname{86.83}
\expandafter\gdef\csname R@med.nw.kits.X4.zthin.band\endcsname{above}
\expandafter\gdef\csname R@med.nw.kits.X4.zthin.d\endcsname{\ensuremath{+2.21}}
\expandafter\gdef\csname R@med.nw.kits.X4.zthin.dM\endcsname{\ensuremath{+0.85}}
\expandafter\gdef\csname R@med.nw.kits.X4.zthin.z\endcsname{\ensuremath{+2.6}}
\expandafter\gdef\csname R@med.nw.kits.X5.dice\endcsname{86.47}
\expandafter\gdef\csname R@med.nw.kits.X5.dice.band\endcsname{above}
\expandafter\gdef\csname R@med.nw.kits.X5.dice.d\endcsname{\ensuremath{+2.21}}
\expandafter\gdef\csname R@med.nw.kits.X5.dice.dM\endcsname{\ensuremath{+0.47}}
\expandafter\gdef\csname R@med.nw.kits.X5.dice.z\endcsname{\ensuremath{+3.5}}
\expandafter\gdef\csname R@med.nw.kits.X5.far\endcsname{7}
\expandafter\gdef\csname R@med.nw.kits.X5.far.band\endcsname{inside}
\expandafter\gdef\csname R@med.nw.kits.X5.far.d\endcsname{\ensuremath{-5.1}}
\expandafter\gdef\csname R@med.nw.kits.X5.far.dM\endcsname{\ensuremath{+0.7}}
\expandafter\gdef\csname R@med.nw.kits.X5.far.z\endcsname{\ensuremath{-1.8}}
\expandafter\gdef\csname R@med.nw.kits.X5.farfp\endcsname{6}
\expandafter\gdef\csname R@med.nw.kits.X5.farfp.band\endcsname{below}
\expandafter\gdef\csname R@med.nw.kits.X5.farfp.d\endcsname{\ensuremath{-5.9}}
\expandafter\gdef\csname R@med.nw.kits.X5.farfp.dM\endcsname{\ensuremath{-0.3}}
\expandafter\gdef\csname R@med.nw.kits.X5.farfp.z\endcsname{\ensuremath{-2.0}}
\expandafter\gdef\csname R@med.nw.kits.X5.ft\endcsname{12}
\expandafter\gdef\csname R@med.nw.kits.X5.ft.band\endcsname{inside}
\expandafter\gdef\csname R@med.nw.kits.X5.ft.d\endcsname{\ensuremath{-2.1}}
\expandafter\gdef\csname R@med.nw.kits.X5.ft.dM\endcsname{\ensuremath{+1.0}}
\expandafter\gdef\csname R@med.nw.kits.X5.ft.z\endcsname{\ensuremath{-1.7}}
\expandafter\gdef\csname R@med.nw.kits.X5.ftfp\endcsname{8}
\expandafter\gdef\csname R@med.nw.kits.X5.ftfp.band\endcsname{inside}
\expandafter\gdef\csname R@med.nw.kits.X5.ftfp.d\endcsname{\ensuremath{-2.1}}
\expandafter\gdef\csname R@med.nw.kits.X5.ftfp.dM\endcsname{\ensuremath{+1.0}}
\expandafter\gdef\csname R@med.nw.kits.X5.ftfp.z\endcsname{\ensuremath{-1.5}}
\expandafter\gdef\csname R@med.nw.kits.X5.landed\endcsname{1}
\expandafter\gdef\csname R@med.nw.kits.X5.rec\endcsname{1}
\expandafter\gdef\csname R@med.nw.kits.X5.tc\endcsname{5.19}
\expandafter\gdef\csname R@med.nw.kits.X5.tc.band\endcsname{inside}
\expandafter\gdef\csname R@med.nw.kits.X5.tc.d\endcsname{\ensuremath{-0.32}}
\expandafter\gdef\csname R@med.nw.kits.X5.tc.dM\endcsname{\ensuremath{-0.42}}
\expandafter\gdef\csname R@med.nw.kits.X5.tc.z\endcsname{\ensuremath{-0.5}}
\expandafter\gdef\csname R@med.nw.kits.X5.tcn\endcsname{5}
\expandafter\gdef\csname R@med.nw.kits.X5.tcn.band\endcsname{inside}
\expandafter\gdef\csname R@med.nw.kits.X5.tcn.d\endcsname{\ensuremath{+0.7}}
\expandafter\gdef\csname R@med.nw.kits.X5.tcn.dM\endcsname{\ensuremath{-0.3}}
\expandafter\gdef\csname R@med.nw.kits.X5.tcn.z\endcsname{\ensuremath{+0.7}}
\expandafter\gdef\csname R@med.nw.kits.X5.zthick\endcsname{86.57}
\expandafter\gdef\csname R@med.nw.kits.X5.zthick.band\endcsname{above}
\expandafter\gdef\csname R@med.nw.kits.X5.zthick.d\endcsname{\ensuremath{+2.72}}
\expandafter\gdef\csname R@med.nw.kits.X5.zthick.dM\endcsname{\ensuremath{+0.54}}
\expandafter\gdef\csname R@med.nw.kits.X5.zthick.z\endcsname{\ensuremath{+2.8}}
\expandafter\gdef\csname R@med.nw.kits.X5.zthin\endcsname{86.38}
\expandafter\gdef\csname R@med.nw.kits.X5.zthin.band\endcsname{above}
\expandafter\gdef\csname R@med.nw.kits.X5.zthin.d\endcsname{\ensuremath{+1.76}}
\expandafter\gdef\csname R@med.nw.kits.X5.zthin.dM\endcsname{\ensuremath{+0.40}}
\expandafter\gdef\csname R@med.nw.kits.X5.zthin.z\endcsname{\ensuremath{+2.0}}
\expandafter\gdef\csname R@med.nw.kits.X6.dice\endcsname{86.36}
\expandafter\gdef\csname R@med.nw.kits.X6.dice.band\endcsname{above}
\expandafter\gdef\csname R@med.nw.kits.X6.dice.d\endcsname{\ensuremath{+2.10}}
\expandafter\gdef\csname R@med.nw.kits.X6.dice.dM\endcsname{\ensuremath{+0.36}}
\expandafter\gdef\csname R@med.nw.kits.X6.dice.z\endcsname{\ensuremath{+3.3}}
\expandafter\gdef\csname R@med.nw.kits.X6.far\endcsname{9}
\expandafter\gdef\csname R@med.nw.kits.X6.far.band\endcsname{inside}
\expandafter\gdef\csname R@med.nw.kits.X6.far.d\endcsname{\ensuremath{-3.1}}
\expandafter\gdef\csname R@med.nw.kits.X6.far.dM\endcsname{\ensuremath{+2.7}}
\expandafter\gdef\csname R@med.nw.kits.X6.far.z\endcsname{\ensuremath{-1.1}}
\expandafter\gdef\csname R@med.nw.kits.X6.farfp\endcsname{9}
\expandafter\gdef\csname R@med.nw.kits.X6.farfp.band\endcsname{inside}
\expandafter\gdef\csname R@med.nw.kits.X6.farfp.d\endcsname{\ensuremath{-2.9}}
\expandafter\gdef\csname R@med.nw.kits.X6.farfp.dM\endcsname{\ensuremath{+2.7}}
\expandafter\gdef\csname R@med.nw.kits.X6.farfp.z\endcsname{\ensuremath{-1.0}}
\expandafter\gdef\csname R@med.nw.kits.X6.ft\endcsname{11}
\expandafter\gdef\csname R@med.nw.kits.X6.ft.band\endcsname{below}
\expandafter\gdef\csname R@med.nw.kits.X6.ft.d\endcsname{\ensuremath{-3.1}}
\expandafter\gdef\csname R@med.nw.kits.X6.ft.dM\endcsname{\ensuremath{0.0}}
\expandafter\gdef\csname R@med.nw.kits.X6.ft.z\endcsname{\ensuremath{-2.5}}
\expandafter\gdef\csname R@med.nw.kits.X6.ftfp\endcsname{8}
\expandafter\gdef\csname R@med.nw.kits.X6.ftfp.band\endcsname{inside}
\expandafter\gdef\csname R@med.nw.kits.X6.ftfp.d\endcsname{\ensuremath{-2.1}}
\expandafter\gdef\csname R@med.nw.kits.X6.ftfp.dM\endcsname{\ensuremath{+1.0}}
\expandafter\gdef\csname R@med.nw.kits.X6.ftfp.z\endcsname{\ensuremath{-1.5}}
\expandafter\gdef\csname R@med.nw.kits.X6.landed\endcsname{1}
\expandafter\gdef\csname R@med.nw.kits.X6.rec\endcsname{1}
\expandafter\gdef\csname R@med.nw.kits.X6.tc\endcsname{5.54}
\expandafter\gdef\csname R@med.nw.kits.X6.tc.band\endcsname{inside}
\expandafter\gdef\csname R@med.nw.kits.X6.tc.d\endcsname{\ensuremath{+0.03}}
\expandafter\gdef\csname R@med.nw.kits.X6.tc.dM\endcsname{\ensuremath{-0.07}}
\expandafter\gdef\csname R@med.nw.kits.X6.tc.z\endcsname{\ensuremath{0.0}}
\expandafter\gdef\csname R@med.nw.kits.X6.tcn\endcsname{6}
\expandafter\gdef\csname R@med.nw.kits.X6.tcn.band\endcsname{inside}
\expandafter\gdef\csname R@med.nw.kits.X6.tcn.d\endcsname{\ensuremath{+1.7}}
\expandafter\gdef\csname R@med.nw.kits.X6.tcn.dM\endcsname{\ensuremath{+0.7}}
\expandafter\gdef\csname R@med.nw.kits.X6.tcn.z\endcsname{\ensuremath{+1.7}}
\expandafter\gdef\csname R@med.nw.kits.X6.zthick\endcsname{86.90}
\expandafter\gdef\csname R@med.nw.kits.X6.zthick.band\endcsname{above}
\expandafter\gdef\csname R@med.nw.kits.X6.zthick.d\endcsname{\ensuremath{+3.05}}
\expandafter\gdef\csname R@med.nw.kits.X6.zthick.dM\endcsname{\ensuremath{+0.87}}
\expandafter\gdef\csname R@med.nw.kits.X6.zthick.z\endcsname{\ensuremath{+3.1}}
\expandafter\gdef\csname R@med.nw.kits.X6.zthin\endcsname{85.89}
\expandafter\gdef\csname R@med.nw.kits.X6.zthin.band\endcsname{inside}
\expandafter\gdef\csname R@med.nw.kits.X6.zthin.d\endcsname{\ensuremath{+1.27}}
\expandafter\gdef\csname R@med.nw.kits.X6.zthin.dM\endcsname{\ensuremath{-0.09}}
\expandafter\gdef\csname R@med.nw.kits.X6.zthin.z\endcsname{\ensuremath{+1.5}}
\expandafter\gdef\csname R@med.nw.kits.Y1.dice\endcsname{86.04}
\expandafter\gdef\csname R@med.nw.kits.Y1.dice.band\endcsname{above}
\expandafter\gdef\csname R@med.nw.kits.Y1.dice.d\endcsname{\ensuremath{+1.78}}
\expandafter\gdef\csname R@med.nw.kits.Y1.dice.dM\endcsname{\ensuremath{+0.03}}
\expandafter\gdef\csname R@med.nw.kits.Y1.dice.z\endcsname{\ensuremath{+2.8}}
\expandafter\gdef\csname R@med.nw.kits.Y1.far\endcsname{7}
\expandafter\gdef\csname R@med.nw.kits.Y1.far.band\endcsname{inside}
\expandafter\gdef\csname R@med.nw.kits.Y1.far.d\endcsname{\ensuremath{-5.1}}
\expandafter\gdef\csname R@med.nw.kits.Y1.far.dM\endcsname{\ensuremath{+0.7}}
\expandafter\gdef\csname R@med.nw.kits.Y1.far.z\endcsname{\ensuremath{-1.8}}
\expandafter\gdef\csname R@med.nw.kits.Y1.farfp\endcsname{7}
\expandafter\gdef\csname R@med.nw.kits.Y1.farfp.band\endcsname{inside}
\expandafter\gdef\csname R@med.nw.kits.Y1.farfp.d\endcsname{\ensuremath{-4.9}}
\expandafter\gdef\csname R@med.nw.kits.Y1.farfp.dM\endcsname{\ensuremath{+0.7}}
\expandafter\gdef\csname R@med.nw.kits.Y1.farfp.z\endcsname{\ensuremath{-1.7}}
\expandafter\gdef\csname R@med.nw.kits.Y1.ft\endcsname{12}
\expandafter\gdef\csname R@med.nw.kits.Y1.ft.band\endcsname{inside}
\expandafter\gdef\csname R@med.nw.kits.Y1.ft.d\endcsname{\ensuremath{-2.1}}
\expandafter\gdef\csname R@med.nw.kits.Y1.ft.dM\endcsname{\ensuremath{+1.0}}
\expandafter\gdef\csname R@med.nw.kits.Y1.ft.z\endcsname{\ensuremath{-1.7}}
\expandafter\gdef\csname R@med.nw.kits.Y1.ftfp\endcsname{6}
\expandafter\gdef\csname R@med.nw.kits.Y1.ftfp.band\endcsname{below}
\expandafter\gdef\csname R@med.nw.kits.Y1.ftfp.d\endcsname{\ensuremath{-4.1}}
\expandafter\gdef\csname R@med.nw.kits.Y1.ftfp.dM\endcsname{\ensuremath{-1.0}}
\expandafter\gdef\csname R@med.nw.kits.Y1.ftfp.z\endcsname{\ensuremath{-2.8}}
\expandafter\gdef\csname R@med.nw.kits.Y1.landed\endcsname{1}
\expandafter\gdef\csname R@med.nw.kits.Y1.rec\endcsname{1}
\expandafter\gdef\csname R@med.nw.kits.Y1.tc\endcsname{5.81}
\expandafter\gdef\csname R@med.nw.kits.Y1.tc.band\endcsname{inside}
\expandafter\gdef\csname R@med.nw.kits.Y1.tc.d\endcsname{\ensuremath{+0.31}}
\expandafter\gdef\csname R@med.nw.kits.Y1.tc.dM\endcsname{\ensuremath{+0.21}}
\expandafter\gdef\csname R@med.nw.kits.Y1.tc.z\endcsname{\ensuremath{+0.5}}
\expandafter\gdef\csname R@med.nw.kits.Y1.tcn\endcsname{5}
\expandafter\gdef\csname R@med.nw.kits.Y1.tcn.band\endcsname{inside}
\expandafter\gdef\csname R@med.nw.kits.Y1.tcn.d\endcsname{\ensuremath{+0.7}}
\expandafter\gdef\csname R@med.nw.kits.Y1.tcn.dM\endcsname{\ensuremath{-0.3}}
\expandafter\gdef\csname R@med.nw.kits.Y1.tcn.z\endcsname{\ensuremath{+0.7}}
\expandafter\gdef\csname R@med.nw.kits.Y1.zthick\endcsname{86.20}
\expandafter\gdef\csname R@med.nw.kits.Y1.zthick.band\endcsname{above}
\expandafter\gdef\csname R@med.nw.kits.Y1.zthick.d\endcsname{\ensuremath{+2.35}}
\expandafter\gdef\csname R@med.nw.kits.Y1.zthick.dM\endcsname{\ensuremath{+0.17}}
\expandafter\gdef\csname R@med.nw.kits.Y1.zthick.z\endcsname{\ensuremath{+2.4}}
\expandafter\gdef\csname R@med.nw.kits.Y1.zthin\endcsname{85.89}
\expandafter\gdef\csname R@med.nw.kits.Y1.zthin.band\endcsname{inside}
\expandafter\gdef\csname R@med.nw.kits.Y1.zthin.d\endcsname{\ensuremath{+1.27}}
\expandafter\gdef\csname R@med.nw.kits.Y1.zthin.dM\endcsname{\ensuremath{-0.09}}
\expandafter\gdef\csname R@med.nw.kits.Y1.zthin.z\endcsname{\ensuremath{+1.5}}
\expandafter\gdef\csname R@med.nw.kits.Y2.dice\endcsname{86.43}
\expandafter\gdef\csname R@med.nw.kits.Y2.dice.band\endcsname{above}
\expandafter\gdef\csname R@med.nw.kits.Y2.dice.d\endcsname{\ensuremath{+2.17}}
\expandafter\gdef\csname R@med.nw.kits.Y2.dice.dM\endcsname{\ensuremath{+0.43}}
\expandafter\gdef\csname R@med.nw.kits.Y2.dice.z\endcsname{\ensuremath{+3.4}}
\expandafter\gdef\csname R@med.nw.kits.Y2.far\endcsname{6}
\expandafter\gdef\csname R@med.nw.kits.Y2.far.band\endcsname{below}
\expandafter\gdef\csname R@med.nw.kits.Y2.far.d\endcsname{\ensuremath{-6.1}}
\expandafter\gdef\csname R@med.nw.kits.Y2.far.dM\endcsname{\ensuremath{-0.3}}
\expandafter\gdef\csname R@med.nw.kits.Y2.far.z\endcsname{\ensuremath{-2.1}}
\expandafter\gdef\csname R@med.nw.kits.Y2.farfp\endcsname{6}
\expandafter\gdef\csname R@med.nw.kits.Y2.farfp.band\endcsname{below}
\expandafter\gdef\csname R@med.nw.kits.Y2.farfp.d\endcsname{\ensuremath{-5.9}}
\expandafter\gdef\csname R@med.nw.kits.Y2.farfp.dM\endcsname{\ensuremath{-0.3}}
\expandafter\gdef\csname R@med.nw.kits.Y2.farfp.z\endcsname{\ensuremath{-2.0}}
\expandafter\gdef\csname R@med.nw.kits.Y2.ft\endcsname{13}
\expandafter\gdef\csname R@med.nw.kits.Y2.ft.band\endcsname{inside}
\expandafter\gdef\csname R@med.nw.kits.Y2.ft.d\endcsname{\ensuremath{-1.1}}
\expandafter\gdef\csname R@med.nw.kits.Y2.ft.dM\endcsname{\ensuremath{+2.0}}
\expandafter\gdef\csname R@med.nw.kits.Y2.ft.z\endcsname{\ensuremath{-0.9}}
\expandafter\gdef\csname R@med.nw.kits.Y2.ftfp\endcsname{7}
\expandafter\gdef\csname R@med.nw.kits.Y2.ftfp.band\endcsname{below}
\expandafter\gdef\csname R@med.nw.kits.Y2.ftfp.d\endcsname{\ensuremath{-3.1}}
\expandafter\gdef\csname R@med.nw.kits.Y2.ftfp.dM\endcsname{\ensuremath{0.0}}
\expandafter\gdef\csname R@med.nw.kits.Y2.ftfp.z\endcsname{\ensuremath{-2.1}}
\expandafter\gdef\csname R@med.nw.kits.Y2.landed\endcsname{1}
\expandafter\gdef\csname R@med.nw.kits.Y2.rec\endcsname{1}
\expandafter\gdef\csname R@med.nw.kits.Y2.tc\endcsname{5.48}
\expandafter\gdef\csname R@med.nw.kits.Y2.tc.band\endcsname{inside}
\expandafter\gdef\csname R@med.nw.kits.Y2.tc.d\endcsname{\ensuremath{-0.03}}
\expandafter\gdef\csname R@med.nw.kits.Y2.tc.dM\endcsname{\ensuremath{-0.13}}
\expandafter\gdef\csname R@med.nw.kits.Y2.tc.z\endcsname{\ensuremath{0.0}}
\expandafter\gdef\csname R@med.nw.kits.Y2.tcn\endcsname{6}
\expandafter\gdef\csname R@med.nw.kits.Y2.tcn.band\endcsname{inside}
\expandafter\gdef\csname R@med.nw.kits.Y2.tcn.d\endcsname{\ensuremath{+1.7}}
\expandafter\gdef\csname R@med.nw.kits.Y2.tcn.dM\endcsname{\ensuremath{+0.7}}
\expandafter\gdef\csname R@med.nw.kits.Y2.tcn.z\endcsname{\ensuremath{+1.7}}
\expandafter\gdef\csname R@med.nw.kits.Y2.zthick\endcsname{85.31}
\expandafter\gdef\csname R@med.nw.kits.Y2.zthick.band\endcsname{inside}
\expandafter\gdef\csname R@med.nw.kits.Y2.zthick.d\endcsname{\ensuremath{+1.46}}
\expandafter\gdef\csname R@med.nw.kits.Y2.zthick.dM\endcsname{\ensuremath{-0.72}}
\expandafter\gdef\csname R@med.nw.kits.Y2.zthick.z\endcsname{\ensuremath{+1.5}}
\expandafter\gdef\csname R@med.nw.kits.Y2.zthin\endcsname{87.42}
\expandafter\gdef\csname R@med.nw.kits.Y2.zthin.band\endcsname{above}
\expandafter\gdef\csname R@med.nw.kits.Y2.zthin.d\endcsname{\ensuremath{+2.80}}
\expandafter\gdef\csname R@med.nw.kits.Y2.zthin.dM\endcsname{\ensuremath{+1.44}}
\expandafter\gdef\csname R@med.nw.kits.Y2.zthin.z\endcsname{\ensuremath{+3.2}}
\expandafter\gdef\csname R@med.nw.kits.base.dice.hi\endcsname{85.52}
\expandafter\gdef\csname R@med.nw.kits.base.dice.lo\endcsname{82.99}
\expandafter\gdef\csname R@med.nw.kits.base.dice.mean\endcsname{84.26}
\expandafter\gdef\csname R@med.nw.kits.base.dice.n\endcsname{9}
\expandafter\gdef\csname R@med.nw.kits.base.dice.s10\endcsname{85.11}
\expandafter\gdef\csname R@med.nw.kits.base.dice.s11\endcsname{84.80}
\expandafter\gdef\csname R@med.nw.kits.base.dice.s3\endcsname{83.75}
\expandafter\gdef\csname R@med.nw.kits.base.dice.s4\endcsname{84.62}
\expandafter\gdef\csname R@med.nw.kits.base.dice.s5\endcsname{83.23}
\expandafter\gdef\csname R@med.nw.kits.base.dice.s6\endcsname{84.79}
\expandafter\gdef\csname R@med.nw.kits.base.dice.s7\endcsname{83.93}
\expandafter\gdef\csname R@med.nw.kits.base.dice.s8\endcsname{84.40}
\expandafter\gdef\csname R@med.nw.kits.base.dice.s9\endcsname{83.70}
\expandafter\gdef\csname R@med.nw.kits.base.dice.sd\endcsname{0.63}
\expandafter\gdef\csname R@med.nw.kits.base.far.hi\endcsname{17.89}
\expandafter\gdef\csname R@med.nw.kits.base.far.lo\endcsname{6.33}
\expandafter\gdef\csname R@med.nw.kits.base.far.mean\endcsname{12.1}
\expandafter\gdef\csname R@med.nw.kits.base.far.n\endcsname{9}
\expandafter\gdef\csname R@med.nw.kits.base.far.s10\endcsname{11}
\expandafter\gdef\csname R@med.nw.kits.base.far.s11\endcsname{10}
\expandafter\gdef\csname R@med.nw.kits.base.far.s3\endcsname{17}
\expandafter\gdef\csname R@med.nw.kits.base.far.s4\endcsname{15}
\expandafter\gdef\csname R@med.nw.kits.base.far.s5\endcsname{11}
\expandafter\gdef\csname R@med.nw.kits.base.far.s6\endcsname{11}
\expandafter\gdef\csname R@med.nw.kits.base.far.s7\endcsname{15}
\expandafter\gdef\csname R@med.nw.kits.base.far.s8\endcsname{8}
\expandafter\gdef\csname R@med.nw.kits.base.far.s9\endcsname{11}
\expandafter\gdef\csname R@med.nw.kits.base.far.sd\endcsname{2.89}
\expandafter\gdef\csname R@med.nw.kits.base.farfp.hi\endcsname{17.67}
\expandafter\gdef\csname R@med.nw.kits.base.farfp.lo\endcsname{6.11}
\expandafter\gdef\csname R@med.nw.kits.base.farfp.mean\endcsname{11.9}
\expandafter\gdef\csname R@med.nw.kits.base.farfp.n\endcsname{9}
\expandafter\gdef\csname R@med.nw.kits.base.farfp.s10\endcsname{11}
\expandafter\gdef\csname R@med.nw.kits.base.farfp.s11\endcsname{9}
\expandafter\gdef\csname R@med.nw.kits.base.farfp.s3\endcsname{17}
\expandafter\gdef\csname R@med.nw.kits.base.farfp.s4\endcsname{15}
\expandafter\gdef\csname R@med.nw.kits.base.farfp.s5\endcsname{11}
\expandafter\gdef\csname R@med.nw.kits.base.farfp.s6\endcsname{11}
\expandafter\gdef\csname R@med.nw.kits.base.farfp.s7\endcsname{14}
\expandafter\gdef\csname R@med.nw.kits.base.farfp.s8\endcsname{8}
\expandafter\gdef\csname R@med.nw.kits.base.farfp.s9\endcsname{11}
\expandafter\gdef\csname R@med.nw.kits.base.farfp.sd\endcsname{2.89}
\expandafter\gdef\csname R@med.nw.kits.base.ft.hi\endcsname{16.65}
\expandafter\gdef\csname R@med.nw.kits.base.ft.lo\endcsname{11.57}
\expandafter\gdef\csname R@med.nw.kits.base.ft.mean\endcsname{14.1}
\expandafter\gdef\csname R@med.nw.kits.base.ft.n\endcsname{9}
\expandafter\gdef\csname R@med.nw.kits.base.ft.s10\endcsname{13}
\expandafter\gdef\csname R@med.nw.kits.base.ft.s11\endcsname{15}
\expandafter\gdef\csname R@med.nw.kits.base.ft.s3\endcsname{14}
\expandafter\gdef\csname R@med.nw.kits.base.ft.s4\endcsname{14}
\expandafter\gdef\csname R@med.nw.kits.base.ft.s5\endcsname{16}
\expandafter\gdef\csname R@med.nw.kits.base.ft.s6\endcsname{13}
\expandafter\gdef\csname R@med.nw.kits.base.ft.s7\endcsname{12}
\expandafter\gdef\csname R@med.nw.kits.base.ft.s8\endcsname{15}
\expandafter\gdef\csname R@med.nw.kits.base.ft.s9\endcsname{15}
\expandafter\gdef\csname R@med.nw.kits.base.ft.sd\endcsname{1.27}
\expandafter\gdef\csname R@med.nw.kits.base.ftfp.hi\endcsname{13.02}
\expandafter\gdef\csname R@med.nw.kits.base.ftfp.lo\endcsname{7.21}
\expandafter\gdef\csname R@med.nw.kits.base.ftfp.mean\endcsname{10.1}
\expandafter\gdef\csname R@med.nw.kits.base.ftfp.n\endcsname{9}
\expandafter\gdef\csname R@med.nw.kits.base.ftfp.s10\endcsname{10}
\expandafter\gdef\csname R@med.nw.kits.base.ftfp.s11\endcsname{11}
\expandafter\gdef\csname R@med.nw.kits.base.ftfp.s3\endcsname{10}
\expandafter\gdef\csname R@med.nw.kits.base.ftfp.s4\endcsname{10}
\expandafter\gdef\csname R@med.nw.kits.base.ftfp.s5\endcsname{11}
\expandafter\gdef\csname R@med.nw.kits.base.ftfp.s6\endcsname{9}
\expandafter\gdef\csname R@med.nw.kits.base.ftfp.s7\endcsname{7}
\expandafter\gdef\csname R@med.nw.kits.base.ftfp.s8\endcsname{11}
\expandafter\gdef\csname R@med.nw.kits.base.ftfp.s9\endcsname{12}
\expandafter\gdef\csname R@med.nw.kits.base.ftfp.sd\endcsname{1.45}
\expandafter\gdef\csname R@med.nw.kits.base.gmac.hi\endcsname{477.40}
\expandafter\gdef\csname R@med.nw.kits.base.gmac.lo\endcsname{477.40}
\expandafter\gdef\csname R@med.nw.kits.base.gmac.mean\endcsname{477.4}
\expandafter\gdef\csname R@med.nw.kits.base.gmac.n\endcsname{9}
\expandafter\gdef\csname R@med.nw.kits.base.gmac.s10\endcsname{477.4}
\expandafter\gdef\csname R@med.nw.kits.base.gmac.s11\endcsname{477.4}
\expandafter\gdef\csname R@med.nw.kits.base.gmac.s3\endcsname{477.4}
\expandafter\gdef\csname R@med.nw.kits.base.gmac.s4\endcsname{477.4}
\expandafter\gdef\csname R@med.nw.kits.base.gmac.s5\endcsname{477.4}
\expandafter\gdef\csname R@med.nw.kits.base.gmac.s6\endcsname{477.4}
\expandafter\gdef\csname R@med.nw.kits.base.gmac.s7\endcsname{477.4}
\expandafter\gdef\csname R@med.nw.kits.base.gmac.s8\endcsname{477.4}
\expandafter\gdef\csname R@med.nw.kits.base.gmac.s9\endcsname{477.4}
\expandafter\gdef\csname R@med.nw.kits.base.gmac.sd\endcsname{0.00}
\expandafter\gdef\csname R@med.nw.kits.base.tc.hi\endcsname{6.79}
\expandafter\gdef\csname R@med.nw.kits.base.tc.lo\endcsname{4.23}
\expandafter\gdef\csname R@med.nw.kits.base.tc.mean\endcsname{5.51}
\expandafter\gdef\csname R@med.nw.kits.base.tc.n\endcsname{9}
\expandafter\gdef\csname R@med.nw.kits.base.tc.s10\endcsname{6.13}
\expandafter\gdef\csname R@med.nw.kits.base.tc.s11\endcsname{5.21}
\expandafter\gdef\csname R@med.nw.kits.base.tc.s3\endcsname{6.22}
\expandafter\gdef\csname R@med.nw.kits.base.tc.s4\endcsname{4.92}
\expandafter\gdef\csname R@med.nw.kits.base.tc.s5\endcsname{5.74}
\expandafter\gdef\csname R@med.nw.kits.base.tc.s6\endcsname{5.97}
\expandafter\gdef\csname R@med.nw.kits.base.tc.s7\endcsname{4.22}
\expandafter\gdef\csname R@med.nw.kits.base.tc.s8\endcsname{5.60}
\expandafter\gdef\csname R@med.nw.kits.base.tc.s9\endcsname{5.58}
\expandafter\gdef\csname R@med.nw.kits.base.tc.sd\endcsname{0.64}
\expandafter\gdef\csname R@med.nw.kits.base.tcn.hi\endcsname{6.33}
\expandafter\gdef\csname R@med.nw.kits.base.tcn.lo\endcsname{2.33}
\expandafter\gdef\csname R@med.nw.kits.base.tcn.mean\endcsname{4.3}
\expandafter\gdef\csname R@med.nw.kits.base.tcn.n\endcsname{9}
\expandafter\gdef\csname R@med.nw.kits.base.tcn.s10\endcsname{5}
\expandafter\gdef\csname R@med.nw.kits.base.tcn.s11\endcsname{4}
\expandafter\gdef\csname R@med.nw.kits.base.tcn.s3\endcsname{5}
\expandafter\gdef\csname R@med.nw.kits.base.tcn.s4\endcsname{3}
\expandafter\gdef\csname R@med.nw.kits.base.tcn.s5\endcsname{5}
\expandafter\gdef\csname R@med.nw.kits.base.tcn.s6\endcsname{6}
\expandafter\gdef\csname R@med.nw.kits.base.tcn.s7\endcsname{3}
\expandafter\gdef\csname R@med.nw.kits.base.tcn.s8\endcsname{4}
\expandafter\gdef\csname R@med.nw.kits.base.tcn.s9\endcsname{4}
\expandafter\gdef\csname R@med.nw.kits.base.tcn.sd\endcsname{1.00}
\expandafter\gdef\csname R@med.nw.kits.base.zthick.hi\endcsname{85.79}
\expandafter\gdef\csname R@med.nw.kits.base.zthick.lo\endcsname{81.90}
\expandafter\gdef\csname R@med.nw.kits.base.zthick.mean\endcsname{83.85}
\expandafter\gdef\csname R@med.nw.kits.base.zthick.n\endcsname{9}
\expandafter\gdef\csname R@med.nw.kits.base.zthick.s10\endcsname{85.26}
\expandafter\gdef\csname R@med.nw.kits.base.zthick.s11\endcsname{84.20}
\expandafter\gdef\csname R@med.nw.kits.base.zthick.s3\endcsname{84.05}
\expandafter\gdef\csname R@med.nw.kits.base.zthick.s4\endcsname{83.67}
\expandafter\gdef\csname R@med.nw.kits.base.zthick.s5\endcsname{82.85}
\expandafter\gdef\csname R@med.nw.kits.base.zthick.s6\endcsname{83.93}
\expandafter\gdef\csname R@med.nw.kits.base.zthick.s7\endcsname{82.45}
\expandafter\gdef\csname R@med.nw.kits.base.zthick.s8\endcsname{85.18}
\expandafter\gdef\csname R@med.nw.kits.base.zthick.s9\endcsname{83.04}
\expandafter\gdef\csname R@med.nw.kits.base.zthick.sd\endcsname{0.97}
\expandafter\gdef\csname R@med.nw.kits.base.zthin.hi\endcsname{86.35}
\expandafter\gdef\csname R@med.nw.kits.base.zthin.lo\endcsname{82.89}
\expandafter\gdef\csname R@med.nw.kits.base.zthin.mean\endcsname{84.62}
\expandafter\gdef\csname R@med.nw.kits.base.zthin.n\endcsname{9}
\expandafter\gdef\csname R@med.nw.kits.base.zthin.s10\endcsname{84.97}
\expandafter\gdef\csname R@med.nw.kits.base.zthin.s11\endcsname{85.34}
\expandafter\gdef\csname R@med.nw.kits.base.zthin.s3\endcsname{83.47}
\expandafter\gdef\csname R@med.nw.kits.base.zthin.s4\endcsname{85.46}
\expandafter\gdef\csname R@med.nw.kits.base.zthin.s5\endcsname{83.57}
\expandafter\gdef\csname R@med.nw.kits.base.zthin.s6\endcsname{85.56}
\expandafter\gdef\csname R@med.nw.kits.base.zthin.s7\endcsname{85.23}
\expandafter\gdef\csname R@med.nw.kits.base.zthin.s8\endcsname{83.70}
\expandafter\gdef\csname R@med.nw.kits.base.zthin.s9\endcsname{84.28}
\expandafter\gdef\csname R@med.nw.kits.base.zthin.sd\endcsname{0.86}
\expandafter\gdef\csname R@med.nw.n.cv\endcsname{7}
\expandafter\gdef\csname R@med.nw.n.cvlanded.amos\endcsname{3}
\expandafter\gdef\csname R@med.nw.n.cvlanded.kits\endcsname{7}
\expandafter\gdef\csname R@med.nw.n.landed.amos\endcsname{10}
\expandafter\gdef\csname R@med.nw.n.landed.kits\endcsname{10}
\expandafter\gdef\csname R@med.nw.n.screens\endcsname{10}
\expandafter\gdef\csname R@med.nw.name.H1\endcsname{HRA}
\expandafter\gdef\csname R@med.nw.name.H2\endcsname{CIG}
\expandafter\gdef\csname R@med.nw.name.X1\endcsname{AnaTok}
\expandafter\gdef\csname R@med.nw.name.X2\endcsname{BandRef}
\expandafter\gdef\csname R@med.nw.name.X3\endcsname{Proto}
\expandafter\gdef\csname R@med.nw.name.X4\endcsname{AnatoMix}
\expandafter\gdef\csname R@med.nw.name.X5\endcsname{AnatoMix-P}
\expandafter\gdef\csname R@med.nw.name.X6\endcsname{SliceRule}
\expandafter\gdef\csname R@med.nw.name.Y1\endcsname{AcqZ}
\expandafter\gdef\csname R@med.nw.name.Y2\endcsname{MassStat}
\expandafter\gdef\csname R@med.nwcv.amos.D2.abs\endcsname{15}
\expandafter\gdef\csname R@med.nwcv.amos.D2.abs.d\endcsname{\ensuremath{+2}}
\expandafter\gdef\csname R@med.nwcv.amos.D2.dice\endcsname{87.91}
\expandafter\gdef\csname R@med.nwcv.amos.D2.dice.d\endcsname{\ensuremath{-0.44}}
\expandafter\gdef\csname R@med.nwcv.amos.D2.dice.hi\endcsname{\ensuremath{-0.12}}
\expandafter\gdef\csname R@med.nwcv.amos.D2.dice.lo\endcsname{\ensuremath{-0.75}}
\expandafter\gdef\csname R@med.nwcv.amos.D2.gmac\endcsname{411.1}
\expandafter\gdef\csname R@med.nwcv.amos.D2.gmac.d\endcsname{\ensuremath{-115.9}}
\expandafter\gdef\csname R@med.nwcv.amos.D2.landed\endcsname{1}
\expandafter\gdef\csname R@med.nwcv.amos.D2.nabs\endcsname{25}
\expandafter\gdef\csname R@med.nwcv.amos.D2.thin\endcsname{82.15}
\expandafter\gdef\csname R@med.nwcv.amos.D2.thin.d\endcsname{\ensuremath{-0.74}}
\expandafter\gdef\csname R@med.nwcv.amos.H1.abs\endcsname{13}
\expandafter\gdef\csname R@med.nwcv.amos.H1.abs.d\endcsname{\ensuremath{0}}
\expandafter\gdef\csname R@med.nwcv.amos.H1.dice\endcsname{88.63}
\expandafter\gdef\csname R@med.nwcv.amos.H1.dice.d\endcsname{\ensuremath{+0.28}}
\expandafter\gdef\csname R@med.nwcv.amos.H1.dice.hi\endcsname{\ensuremath{+1.04}}
\expandafter\gdef\csname R@med.nwcv.amos.H1.dice.lo\endcsname{\ensuremath{-0.48}}
\expandafter\gdef\csname R@med.nwcv.amos.H1.landed\endcsname{1}
\expandafter\gdef\csname R@med.nwcv.amos.H1.nabs\endcsname{25}
\expandafter\gdef\csname R@med.nwcv.amos.H1.thin\endcsname{83.02}
\expandafter\gdef\csname R@med.nwcv.amos.H1.thin.d\endcsname{\ensuremath{+0.13}}
\expandafter\gdef\csname R@med.nwcv.amos.H1.thin1\endcsname{80.79}
\expandafter\gdef\csname R@med.nwcv.amos.H1.thin1.d\endcsname{\ensuremath{+0.37}}
\expandafter\gdef\csname R@med.nwcv.amos.H1.thin2\endcsname{89.37}
\expandafter\gdef\csname R@med.nwcv.amos.H1.thin2.d\endcsname{\ensuremath{+0.32}}
\expandafter\gdef\csname R@med.nwcv.amos.H2.abs\endcsname{13}
\expandafter\gdef\csname R@med.nwcv.amos.H2.abs.d\endcsname{\ensuremath{0}}
\expandafter\gdef\csname R@med.nwcv.amos.H2.dice\endcsname{88.57}
\expandafter\gdef\csname R@med.nwcv.amos.H2.dice.d\endcsname{\ensuremath{+0.22}}
\expandafter\gdef\csname R@med.nwcv.amos.H2.dice.hi\endcsname{\ensuremath{+0.78}}
\expandafter\gdef\csname R@med.nwcv.amos.H2.dice.lo\endcsname{\ensuremath{-0.35}}
\expandafter\gdef\csname R@med.nwcv.amos.H2.landed\endcsname{1}
\expandafter\gdef\csname R@med.nwcv.amos.H2.nabs\endcsname{25}
\expandafter\gdef\csname R@med.nwcv.amos.H2.thin\endcsname{82.92}
\expandafter\gdef\csname R@med.nwcv.amos.H2.thin.d\endcsname{\ensuremath{+0.03}}
\expandafter\gdef\csname R@med.nwcv.amos.H2.thin1\endcsname{80.62}
\expandafter\gdef\csname R@med.nwcv.amos.H2.thin1.d\endcsname{\ensuremath{+0.20}}
\expandafter\gdef\csname R@med.nwcv.amos.H2.thin2\endcsname{89.32}
\expandafter\gdef\csname R@med.nwcv.amos.H2.thin2.d\endcsname{\ensuremath{+0.27}}
\expandafter\gdef\csname R@med.nwcv.amos.L1.abs\endcsname{12}
\expandafter\gdef\csname R@med.nwcv.amos.L1.abs.d\endcsname{\ensuremath{-1}}
\expandafter\gdef\csname R@med.nwcv.amos.L1.dice\endcsname{88.75}
\expandafter\gdef\csname R@med.nwcv.amos.L1.dice.d\endcsname{\ensuremath{+0.41}}
\expandafter\gdef\csname R@med.nwcv.amos.L1.dice.hi\endcsname{\ensuremath{+1.09}}
\expandafter\gdef\csname R@med.nwcv.amos.L1.dice.lo\endcsname{\ensuremath{-0.27}}
\expandafter\gdef\csname R@med.nwcv.amos.L1.gmac\endcsname{977.3}
\expandafter\gdef\csname R@med.nwcv.amos.L1.gmac.d\endcsname{\ensuremath{+450.3}}
\expandafter\gdef\csname R@med.nwcv.amos.L1.landed\endcsname{1}
\expandafter\gdef\csname R@med.nwcv.amos.L1.nabs\endcsname{25}
\expandafter\gdef\csname R@med.nwcv.amos.L1.thin\endcsname{83.07}
\expandafter\gdef\csname R@med.nwcv.amos.L1.thin.d\endcsname{\ensuremath{+0.18}}
\expandafter\gdef\csname R@med.nwcv.amos.L1.thin1\endcsname{80.91}
\expandafter\gdef\csname R@med.nwcv.amos.L1.thin1.d\endcsname{\ensuremath{+0.48}}
\expandafter\gdef\csname R@med.nwcv.amos.L1.thin2\endcsname{89.47}
\expandafter\gdef\csname R@med.nwcv.amos.L1.thin2.d\endcsname{\ensuremath{+0.42}}
\expandafter\gdef\csname R@med.nwcv.amos.L3.abs\endcsname{15}
\expandafter\gdef\csname R@med.nwcv.amos.L3.abs.d\endcsname{\ensuremath{+2}}
\expandafter\gdef\csname R@med.nwcv.amos.L3.dice\endcsname{88.58}
\expandafter\gdef\csname R@med.nwcv.amos.L3.dice.d\endcsname{\ensuremath{+0.23}}
\expandafter\gdef\csname R@med.nwcv.amos.L3.dice.hi\endcsname{\ensuremath{+0.54}}
\expandafter\gdef\csname R@med.nwcv.amos.L3.dice.lo\endcsname{\ensuremath{-0.08}}
\expandafter\gdef\csname R@med.nwcv.amos.L3.gmac\endcsname{732.1}
\expandafter\gdef\csname R@med.nwcv.amos.L3.gmac.d\endcsname{\ensuremath{+205.1}}
\expandafter\gdef\csname R@med.nwcv.amos.L3.landed\endcsname{1}
\expandafter\gdef\csname R@med.nwcv.amos.L3.nabs\endcsname{25}
\expandafter\gdef\csname R@med.nwcv.amos.L3.thin\endcsname{83.10}
\expandafter\gdef\csname R@med.nwcv.amos.L3.thin.d\endcsname{\ensuremath{+0.21}}
\expandafter\gdef\csname R@med.nwcv.amos.L3.thin1\endcsname{80.83}
\expandafter\gdef\csname R@med.nwcv.amos.L3.thin1.d\endcsname{\ensuremath{+0.40}}
\expandafter\gdef\csname R@med.nwcv.amos.L3.thin2\endcsname{89.37}
\expandafter\gdef\csname R@med.nwcv.amos.L3.thin2.d\endcsname{\ensuremath{+0.32}}
\expandafter\gdef\csname R@med.nwcv.amos.V5B1.abs\endcsname{14}
\expandafter\gdef\csname R@med.nwcv.amos.V5B1.abs.d\endcsname{\ensuremath{+1}}
\expandafter\gdef\csname R@med.nwcv.amos.V5B1.dice\endcsname{87.89}
\expandafter\gdef\csname R@med.nwcv.amos.V5B1.dice.d\endcsname{\ensuremath{-0.46}}
\expandafter\gdef\csname R@med.nwcv.amos.V5B1.dice.hi\endcsname{\ensuremath{-0.10}}
\expandafter\gdef\csname R@med.nwcv.amos.V5B1.dice.lo\endcsname{\ensuremath{-0.82}}
\expandafter\gdef\csname R@med.nwcv.amos.V5B1.gmac\endcsname{295.2}
\expandafter\gdef\csname R@med.nwcv.amos.V5B1.gmac.d\endcsname{\ensuremath{-231.8}}
\expandafter\gdef\csname R@med.nwcv.amos.V5B1.landed\endcsname{1}
\expandafter\gdef\csname R@med.nwcv.amos.V5B1.nabs\endcsname{25}
\expandafter\gdef\csname R@med.nwcv.amos.V5B1.thin\endcsname{82.02}
\expandafter\gdef\csname R@med.nwcv.amos.V5B1.thin.d\endcsname{\ensuremath{-0.87}}
\expandafter\gdef\csname R@med.nwcv.amos.V5B2.abs\endcsname{15}
\expandafter\gdef\csname R@med.nwcv.amos.V5B2.abs.d\endcsname{\ensuremath{+2}}
\expandafter\gdef\csname R@med.nwcv.amos.V5B2.dice\endcsname{88.09}
\expandafter\gdef\csname R@med.nwcv.amos.V5B2.dice.d\endcsname{\ensuremath{-0.25}}
\expandafter\gdef\csname R@med.nwcv.amos.V5B2.dice.hi\endcsname{\ensuremath{+0.32}}
\expandafter\gdef\csname R@med.nwcv.amos.V5B2.dice.lo\endcsname{\ensuremath{-0.83}}
\expandafter\gdef\csname R@med.nwcv.amos.V5B2.gmac\endcsname{295.3}
\expandafter\gdef\csname R@med.nwcv.amos.V5B2.gmac.d\endcsname{\ensuremath{-231.7}}
\expandafter\gdef\csname R@med.nwcv.amos.V5B2.landed\endcsname{1}
\expandafter\gdef\csname R@med.nwcv.amos.V5B2.nabs\endcsname{25}
\expandafter\gdef\csname R@med.nwcv.amos.V5B2.thin\endcsname{82.39}
\expandafter\gdef\csname R@med.nwcv.amos.V5B2.thin.d\endcsname{\ensuremath{-0.50}}
\expandafter\gdef\csname R@med.nwcv.amos.V7A1.abs\endcsname{10}
\expandafter\gdef\csname R@med.nwcv.amos.V7A1.abs.d\endcsname{\ensuremath{-3}}
\expandafter\gdef\csname R@med.nwcv.amos.V7A1.dice\endcsname{88.61}
\expandafter\gdef\csname R@med.nwcv.amos.V7A1.dice.d\endcsname{\ensuremath{+0.26}}
\expandafter\gdef\csname R@med.nwcv.amos.V7A1.dice.hi\endcsname{\ensuremath{+1.02}}
\expandafter\gdef\csname R@med.nwcv.amos.V7A1.dice.lo\endcsname{\ensuremath{-0.50}}
\expandafter\gdef\csname R@med.nwcv.amos.V7A1.gmac\endcsname{513.5}
\expandafter\gdef\csname R@med.nwcv.amos.V7A1.gmac.d\endcsname{\ensuremath{-13.5}}
\expandafter\gdef\csname R@med.nwcv.amos.V7A1.landed\endcsname{1}
\expandafter\gdef\csname R@med.nwcv.amos.V7A1.nabs\endcsname{25}
\expandafter\gdef\csname R@med.nwcv.amos.V7A1.thin\endcsname{82.67}
\expandafter\gdef\csname R@med.nwcv.amos.V7A1.thin.d\endcsname{\ensuremath{-0.22}}
\expandafter\gdef\csname R@med.nwcv.amos.V7A1.thin1\endcsname{80.33}
\expandafter\gdef\csname R@med.nwcv.amos.V7A1.thin1.d\endcsname{\ensuremath{-0.09}}
\expandafter\gdef\csname R@med.nwcv.amos.V7A1.thin2\endcsname{89.08}
\expandafter\gdef\csname R@med.nwcv.amos.V7A1.thin2.d\endcsname{\ensuremath{+0.03}}
\expandafter\gdef\csname R@med.nwcv.amos.V7C1.abs\endcsname{15}
\expandafter\gdef\csname R@med.nwcv.amos.V7C1.abs.d\endcsname{\ensuremath{+2}}
\expandafter\gdef\csname R@med.nwcv.amos.V7C1.dice\endcsname{87.89}
\expandafter\gdef\csname R@med.nwcv.amos.V7C1.dice.d\endcsname{\ensuremath{-0.46}}
\expandafter\gdef\csname R@med.nwcv.amos.V7C1.dice.hi\endcsname{\ensuremath{-0.11}}
\expandafter\gdef\csname R@med.nwcv.amos.V7C1.dice.lo\endcsname{\ensuremath{-0.81}}
\expandafter\gdef\csname R@med.nwcv.amos.V7C1.gmac\endcsname{295.2}
\expandafter\gdef\csname R@med.nwcv.amos.V7C1.gmac.d\endcsname{\ensuremath{-231.8}}
\expandafter\gdef\csname R@med.nwcv.amos.V7C1.landed\endcsname{1}
\expandafter\gdef\csname R@med.nwcv.amos.V7C1.nabs\endcsname{25}
\expandafter\gdef\csname R@med.nwcv.amos.V7C1.thin\endcsname{82.13}
\expandafter\gdef\csname R@med.nwcv.amos.V7C1.thin.d\endcsname{\ensuremath{-0.75}}
\expandafter\gdef\csname R@med.nwcv.amos.V7C1.thin1\endcsname{79.68}
\expandafter\gdef\csname R@med.nwcv.amos.V7C1.thin1.d\endcsname{\ensuremath{-0.74}}
\expandafter\gdef\csname R@med.nwcv.amos.V7C1.thin2\endcsname{88.43}
\expandafter\gdef\csname R@med.nwcv.amos.V7C1.thin2.d\endcsname{\ensuremath{-0.61}}
\expandafter\gdef\csname R@med.nwcv.amos.Y2.abs\endcsname{13}
\expandafter\gdef\csname R@med.nwcv.amos.Y2.abs.d\endcsname{\ensuremath{0}}
\expandafter\gdef\csname R@med.nwcv.amos.Y2.dice\endcsname{88.57}
\expandafter\gdef\csname R@med.nwcv.amos.Y2.dice.d\endcsname{\ensuremath{+0.23}}
\expandafter\gdef\csname R@med.nwcv.amos.Y2.dice.hi\endcsname{\ensuremath{+0.93}}
\expandafter\gdef\csname R@med.nwcv.amos.Y2.dice.lo\endcsname{\ensuremath{-0.47}}
\expandafter\gdef\csname R@med.nwcv.amos.Y2.landed\endcsname{1}
\expandafter\gdef\csname R@med.nwcv.amos.Y2.nabs\endcsname{25}
\expandafter\gdef\csname R@med.nwcv.amos.Y2.thin\endcsname{82.77}
\expandafter\gdef\csname R@med.nwcv.amos.Y2.thin.d\endcsname{\ensuremath{-0.12}}
\expandafter\gdef\csname R@med.nwcv.amos.Y2.thin1\endcsname{80.41}
\expandafter\gdef\csname R@med.nwcv.amos.Y2.thin1.d\endcsname{\ensuremath{-0.01}}
\expandafter\gdef\csname R@med.nwcv.amos.Y2.thin2\endcsname{89.06}
\expandafter\gdef\csname R@med.nwcv.amos.Y2.thin2.d\endcsname{\ensuremath{+0.02}}
\expandafter\gdef\csname R@med.nwcv.amos.base.abs\endcsname{13}
\expandafter\gdef\csname R@med.nwcv.amos.base.dice\endcsname{88.35}
\expandafter\gdef\csname R@med.nwcv.amos.base.gmac\endcsname{527.0}
\expandafter\gdef\csname R@med.nwcv.amos.base.landed\endcsname{1}
\expandafter\gdef\csname R@med.nwcv.amos.base.nabs\endcsname{25}
\expandafter\gdef\csname R@med.nwcv.amos.base.thin\endcsname{82.89}
\expandafter\gdef\csname R@med.nwcv.amos.base.thin1\endcsname{80.42}
\expandafter\gdef\csname R@med.nwcv.amos.base.thin2\endcsname{89.04}
\expandafter\gdef\csname R@med.nwcv.kits.D2.dice\endcsname{85.04}
\expandafter\gdef\csname R@med.nwcv.kits.D2.dice.d\endcsname{\ensuremath{+1.04}}
\expandafter\gdef\csname R@med.nwcv.kits.D2.dice.hi\endcsname{\ensuremath{+2.12}}
\expandafter\gdef\csname R@med.nwcv.kits.D2.dice.lo\endcsname{\ensuremath{-0.03}}
\expandafter\gdef\csname R@med.nwcv.kits.D2.ft\endcsname{48}
\expandafter\gdef\csname R@med.nwcv.kits.D2.ft.d\endcsname{\ensuremath{-7}}
\expandafter\gdef\csname R@med.nwcv.kits.D2.gmac\endcsname{367.2}
\expandafter\gdef\csname R@med.nwcv.kits.D2.gmac.d\endcsname{\ensuremath{-110.2}}
\expandafter\gdef\csname R@med.nwcv.kits.D2.landed\endcsname{1}
\expandafter\gdef\csname R@med.nwcv.kits.D2.tc\endcsname{4.07}
\expandafter\gdef\csname R@med.nwcv.kits.D2.tc.d\endcsname{\ensuremath{+0.91}}
\expandafter\gdef\csname R@med.nwcv.kits.D2.tcn\endcsname{14}
\expandafter\gdef\csname R@med.nwcv.kits.D2.tcn.d\endcsname{\ensuremath{0}}
\expandafter\gdef\csname R@med.nwcv.kits.D2.zthick\endcsname{85.15}
\expandafter\gdef\csname R@med.nwcv.kits.D2.zthick.d\endcsname{\ensuremath{+0.94}}
\expandafter\gdef\csname R@med.nwcv.kits.D2.zthin\endcsname{84.95}
\expandafter\gdef\csname R@med.nwcv.kits.D2.zthin.d\endcsname{\ensuremath{+1.14}}
\expandafter\gdef\csname R@med.nwcv.kits.H1.dice\endcsname{86.00}
\expandafter\gdef\csname R@med.nwcv.kits.H1.dice.d\endcsname{\ensuremath{+2.01}}
\expandafter\gdef\csname R@med.nwcv.kits.H1.dice.hi\endcsname{\ensuremath{+4.00}}
\expandafter\gdef\csname R@med.nwcv.kits.H1.dice.lo\endcsname{\ensuremath{+0.02}}
\expandafter\gdef\csname R@med.nwcv.kits.H1.far\endcsname{21}
\expandafter\gdef\csname R@med.nwcv.kits.H1.far.d\endcsname{\ensuremath{-31}}
\expandafter\gdef\csname R@med.nwcv.kits.H1.farfp\endcsname{20}
\expandafter\gdef\csname R@med.nwcv.kits.H1.farfp.d\endcsname{\ensuremath{-29}}
\expandafter\gdef\csname R@med.nwcv.kits.H1.ft\endcsname{47}
\expandafter\gdef\csname R@med.nwcv.kits.H1.ft.d\endcsname{\ensuremath{-8}}
\expandafter\gdef\csname R@med.nwcv.kits.H1.ftfp\endcsname{24}
\expandafter\gdef\csname R@med.nwcv.kits.H1.ftfp.d\endcsname{\ensuremath{-16}}
\expandafter\gdef\csname R@med.nwcv.kits.H1.landed\endcsname{1}
\expandafter\gdef\csname R@med.nwcv.kits.H1.tc\endcsname{3.60}
\expandafter\gdef\csname R@med.nwcv.kits.H1.tc.d\endcsname{\ensuremath{+0.44}}
\expandafter\gdef\csname R@med.nwcv.kits.H1.tcn\endcsname{13}
\expandafter\gdef\csname R@med.nwcv.kits.H1.tcn.d\endcsname{\ensuremath{-1}}
\expandafter\gdef\csname R@med.nwcv.kits.H1.zthick\endcsname{86.04}
\expandafter\gdef\csname R@med.nwcv.kits.H1.zthick.d\endcsname{\ensuremath{+1.83}}
\expandafter\gdef\csname R@med.nwcv.kits.H1.zthin\endcsname{85.98}
\expandafter\gdef\csname R@med.nwcv.kits.H1.zthin.d\endcsname{\ensuremath{+2.17}}
\expandafter\gdef\csname R@med.nwcv.kits.H2.dice\endcsname{84.95}
\expandafter\gdef\csname R@med.nwcv.kits.H2.dice.d\endcsname{\ensuremath{+0.95}}
\expandafter\gdef\csname R@med.nwcv.kits.H2.dice.hi\endcsname{\ensuremath{+2.95}}
\expandafter\gdef\csname R@med.nwcv.kits.H2.dice.lo\endcsname{\ensuremath{-1.05}}
\expandafter\gdef\csname R@med.nwcv.kits.H2.far\endcsname{30}
\expandafter\gdef\csname R@med.nwcv.kits.H2.far.d\endcsname{\ensuremath{-22}}
\expandafter\gdef\csname R@med.nwcv.kits.H2.farfp\endcsname{30}
\expandafter\gdef\csname R@med.nwcv.kits.H2.farfp.d\endcsname{\ensuremath{-19}}
\expandafter\gdef\csname R@med.nwcv.kits.H2.ft\endcsname{46}
\expandafter\gdef\csname R@med.nwcv.kits.H2.ft.d\endcsname{\ensuremath{-9}}
\expandafter\gdef\csname R@med.nwcv.kits.H2.ftfp\endcsname{31}
\expandafter\gdef\csname R@med.nwcv.kits.H2.ftfp.d\endcsname{\ensuremath{-9}}
\expandafter\gdef\csname R@med.nwcv.kits.H2.landed\endcsname{1}
\expandafter\gdef\csname R@med.nwcv.kits.H2.tc\endcsname{3.78}
\expandafter\gdef\csname R@med.nwcv.kits.H2.tc.d\endcsname{\ensuremath{+0.62}}
\expandafter\gdef\csname R@med.nwcv.kits.H2.tcn\endcsname{8}
\expandafter\gdef\csname R@med.nwcv.kits.H2.tcn.d\endcsname{\ensuremath{-6}}
\expandafter\gdef\csname R@med.nwcv.kits.H2.zthick\endcsname{85.10}
\expandafter\gdef\csname R@med.nwcv.kits.H2.zthick.d\endcsname{\ensuremath{+0.89}}
\expandafter\gdef\csname R@med.nwcv.kits.H2.zthin\endcsname{84.83}
\expandafter\gdef\csname R@med.nwcv.kits.H2.zthin.d\endcsname{\ensuremath{+1.02}}
\expandafter\gdef\csname R@med.nwcv.kits.L1.dice\endcsname{86.14}
\expandafter\gdef\csname R@med.nwcv.kits.L1.dice.d\endcsname{\ensuremath{+2.15}}
\expandafter\gdef\csname R@med.nwcv.kits.L1.dice.hi\endcsname{\ensuremath{+3.32}}
\expandafter\gdef\csname R@med.nwcv.kits.L1.dice.lo\endcsname{\ensuremath{+0.98}}
\expandafter\gdef\csname R@med.nwcv.kits.L1.far\endcsname{23}
\expandafter\gdef\csname R@med.nwcv.kits.L1.far.d\endcsname{\ensuremath{-29}}
\expandafter\gdef\csname R@med.nwcv.kits.L1.farfp\endcsname{22}
\expandafter\gdef\csname R@med.nwcv.kits.L1.farfp.d\endcsname{\ensuremath{-27}}
\expandafter\gdef\csname R@med.nwcv.kits.L1.ft\endcsname{47}
\expandafter\gdef\csname R@med.nwcv.kits.L1.ft.d\endcsname{\ensuremath{-8}}
\expandafter\gdef\csname R@med.nwcv.kits.L1.ftfp\endcsname{26}
\expandafter\gdef\csname R@med.nwcv.kits.L1.ftfp.d\endcsname{\ensuremath{-14}}
\expandafter\gdef\csname R@med.nwcv.kits.L1.gmac\endcsname{717.5}
\expandafter\gdef\csname R@med.nwcv.kits.L1.gmac.d\endcsname{\ensuremath{+240.1}}
\expandafter\gdef\csname R@med.nwcv.kits.L1.landed\endcsname{1}
\expandafter\gdef\csname R@med.nwcv.kits.L1.tc\endcsname{3.51}
\expandafter\gdef\csname R@med.nwcv.kits.L1.tc.d\endcsname{\ensuremath{+0.35}}
\expandafter\gdef\csname R@med.nwcv.kits.L1.tcn\endcsname{11}
\expandafter\gdef\csname R@med.nwcv.kits.L1.tcn.d\endcsname{\ensuremath{-3}}
\expandafter\gdef\csname R@med.nwcv.kits.L1.zthick\endcsname{85.73}
\expandafter\gdef\csname R@med.nwcv.kits.L1.zthick.d\endcsname{\ensuremath{+1.52}}
\expandafter\gdef\csname R@med.nwcv.kits.L1.zthin\endcsname{86.51}
\expandafter\gdef\csname R@med.nwcv.kits.L1.zthin.d\endcsname{\ensuremath{+2.70}}
\expandafter\gdef\csname R@med.nwcv.kits.L3.dice\endcsname{84.67}
\expandafter\gdef\csname R@med.nwcv.kits.L3.dice.d\endcsname{\ensuremath{+0.68}}
\expandafter\gdef\csname R@med.nwcv.kits.L3.dice.hi\endcsname{\ensuremath{+2.44}}
\expandafter\gdef\csname R@med.nwcv.kits.L3.dice.lo\endcsname{\ensuremath{-1.08}}
\expandafter\gdef\csname R@med.nwcv.kits.L3.far\endcsname{41}
\expandafter\gdef\csname R@med.nwcv.kits.L3.far.d\endcsname{\ensuremath{-11}}
\expandafter\gdef\csname R@med.nwcv.kits.L3.farfp\endcsname{39}
\expandafter\gdef\csname R@med.nwcv.kits.L3.farfp.d\endcsname{\ensuremath{-10}}
\expandafter\gdef\csname R@med.nwcv.kits.L3.ft\endcsname{52}
\expandafter\gdef\csname R@med.nwcv.kits.L3.ft.d\endcsname{\ensuremath{-3}}
\expandafter\gdef\csname R@med.nwcv.kits.L3.ftfp\endcsname{35}
\expandafter\gdef\csname R@med.nwcv.kits.L3.ftfp.d\endcsname{\ensuremath{-5}}
\expandafter\gdef\csname R@med.nwcv.kits.L3.gmac\endcsname{586.8}
\expandafter\gdef\csname R@med.nwcv.kits.L3.gmac.d\endcsname{\ensuremath{+109.4}}
\expandafter\gdef\csname R@med.nwcv.kits.L3.landed\endcsname{1}
\expandafter\gdef\csname R@med.nwcv.kits.L3.tc\endcsname{3.37}
\expandafter\gdef\csname R@med.nwcv.kits.L3.tc.d\endcsname{\ensuremath{+0.21}}
\expandafter\gdef\csname R@med.nwcv.kits.L3.tcn\endcsname{14}
\expandafter\gdef\csname R@med.nwcv.kits.L3.tcn.d\endcsname{\ensuremath{0}}
\expandafter\gdef\csname R@med.nwcv.kits.L3.zthick\endcsname{84.58}
\expandafter\gdef\csname R@med.nwcv.kits.L3.zthick.d\endcsname{\ensuremath{+0.38}}
\expandafter\gdef\csname R@med.nwcv.kits.L3.zthin\endcsname{84.76}
\expandafter\gdef\csname R@med.nwcv.kits.L3.zthin.d\endcsname{\ensuremath{+0.96}}
\expandafter\gdef\csname R@med.nwcv.kits.V5B1.dice\endcsname{84.99}
\expandafter\gdef\csname R@med.nwcv.kits.V5B1.dice.d\endcsname{\ensuremath{+0.99}}
\expandafter\gdef\csname R@med.nwcv.kits.V5B1.dice.hi\endcsname{\ensuremath{+2.58}}
\expandafter\gdef\csname R@med.nwcv.kits.V5B1.dice.lo\endcsname{\ensuremath{-0.59}}
\expandafter\gdef\csname R@med.nwcv.kits.V5B1.ft\endcsname{50}
\expandafter\gdef\csname R@med.nwcv.kits.V5B1.ft.d\endcsname{\ensuremath{-5}}
\expandafter\gdef\csname R@med.nwcv.kits.V5B1.gmac\endcsname{257.1}
\expandafter\gdef\csname R@med.nwcv.kits.V5B1.gmac.d\endcsname{\ensuremath{-220.3}}
\expandafter\gdef\csname R@med.nwcv.kits.V5B1.landed\endcsname{1}
\expandafter\gdef\csname R@med.nwcv.kits.V5B1.tc\endcsname{4.31}
\expandafter\gdef\csname R@med.nwcv.kits.V5B1.tc.d\endcsname{\ensuremath{+1.15}}
\expandafter\gdef\csname R@med.nwcv.kits.V5B1.tcn\endcsname{12}
\expandafter\gdef\csname R@med.nwcv.kits.V5B1.tcn.d\endcsname{\ensuremath{-2}}
\expandafter\gdef\csname R@med.nwcv.kits.V5B1.zthick\endcsname{84.60}
\expandafter\gdef\csname R@med.nwcv.kits.V5B1.zthick.d\endcsname{\ensuremath{+0.39}}
\expandafter\gdef\csname R@med.nwcv.kits.V5B1.zthin\endcsname{85.34}
\expandafter\gdef\csname R@med.nwcv.kits.V5B1.zthin.d\endcsname{\ensuremath{+1.53}}
\expandafter\gdef\csname R@med.nwcv.kits.V5B2.dice\endcsname{83.24}
\expandafter\gdef\csname R@med.nwcv.kits.V5B2.dice.d\endcsname{\ensuremath{-0.75}}
\expandafter\gdef\csname R@med.nwcv.kits.V5B2.dice.hi\endcsname{\ensuremath{+1.18}}
\expandafter\gdef\csname R@med.nwcv.kits.V5B2.dice.lo\endcsname{\ensuremath{-2.69}}
\expandafter\gdef\csname R@med.nwcv.kits.V5B2.far\endcsname{49}
\expandafter\gdef\csname R@med.nwcv.kits.V5B2.far.d\endcsname{\ensuremath{-3}}
\expandafter\gdef\csname R@med.nwcv.kits.V5B2.farfp\endcsname{44}
\expandafter\gdef\csname R@med.nwcv.kits.V5B2.farfp.d\endcsname{\ensuremath{-5}}
\expandafter\gdef\csname R@med.nwcv.kits.V5B2.ft\endcsname{57}
\expandafter\gdef\csname R@med.nwcv.kits.V5B2.ft.d\endcsname{\ensuremath{+2}}
\expandafter\gdef\csname R@med.nwcv.kits.V5B2.ftfp\endcsname{40}
\expandafter\gdef\csname R@med.nwcv.kits.V5B2.ftfp.d\endcsname{\ensuremath{0}}
\expandafter\gdef\csname R@med.nwcv.kits.V5B2.gmac\endcsname{257.2}
\expandafter\gdef\csname R@med.nwcv.kits.V5B2.gmac.d\endcsname{\ensuremath{-220.2}}
\expandafter\gdef\csname R@med.nwcv.kits.V5B2.landed\endcsname{1}
\expandafter\gdef\csname R@med.nwcv.kits.V5B2.tc\endcsname{4.37}
\expandafter\gdef\csname R@med.nwcv.kits.V5B2.tc.d\endcsname{\ensuremath{+1.21}}
\expandafter\gdef\csname R@med.nwcv.kits.V5B2.tcn\endcsname{11}
\expandafter\gdef\csname R@med.nwcv.kits.V5B2.tcn.d\endcsname{\ensuremath{-3}}
\expandafter\gdef\csname R@med.nwcv.kits.V5B2.zthick\endcsname{83.55}
\expandafter\gdef\csname R@med.nwcv.kits.V5B2.zthick.d\endcsname{\ensuremath{-0.65}}
\expandafter\gdef\csname R@med.nwcv.kits.V5B2.zthin\endcsname{82.98}
\expandafter\gdef\csname R@med.nwcv.kits.V5B2.zthin.d\endcsname{\ensuremath{-0.83}}
\expandafter\gdef\csname R@med.nwcv.kits.V7A1.dice\endcsname{85.63}
\expandafter\gdef\csname R@med.nwcv.kits.V7A1.dice.d\endcsname{\ensuremath{+1.63}}
\expandafter\gdef\csname R@med.nwcv.kits.V7A1.dice.hi\endcsname{\ensuremath{+3.27}}
\expandafter\gdef\csname R@med.nwcv.kits.V7A1.dice.lo\endcsname{\ensuremath{0.00}}
\expandafter\gdef\csname R@med.nwcv.kits.V7A1.far\endcsname{35}
\expandafter\gdef\csname R@med.nwcv.kits.V7A1.far.d\endcsname{\ensuremath{-17}}
\expandafter\gdef\csname R@med.nwcv.kits.V7A1.farfp\endcsname{34}
\expandafter\gdef\csname R@med.nwcv.kits.V7A1.farfp.d\endcsname{\ensuremath{-15}}
\expandafter\gdef\csname R@med.nwcv.kits.V7A1.ft\endcsname{42}
\expandafter\gdef\csname R@med.nwcv.kits.V7A1.ft.d\endcsname{\ensuremath{-13}}
\expandafter\gdef\csname R@med.nwcv.kits.V7A1.ftfp\endcsname{26}
\expandafter\gdef\csname R@med.nwcv.kits.V7A1.ftfp.d\endcsname{\ensuremath{-14}}
\expandafter\gdef\csname R@med.nwcv.kits.V7A1.gmac\endcsname{373.5}
\expandafter\gdef\csname R@med.nwcv.kits.V7A1.gmac.d\endcsname{\ensuremath{-103.9}}
\expandafter\gdef\csname R@med.nwcv.kits.V7A1.landed\endcsname{1}
\expandafter\gdef\csname R@med.nwcv.kits.V7A1.tc\endcsname{3.28}
\expandafter\gdef\csname R@med.nwcv.kits.V7A1.tc.d\endcsname{\ensuremath{+0.12}}
\expandafter\gdef\csname R@med.nwcv.kits.V7A1.tcn\endcsname{11}
\expandafter\gdef\csname R@med.nwcv.kits.V7A1.tcn.d\endcsname{\ensuremath{-3}}
\expandafter\gdef\csname R@med.nwcv.kits.V7A1.zthick\endcsname{85.13}
\expandafter\gdef\csname R@med.nwcv.kits.V7A1.zthick.d\endcsname{\ensuremath{+0.92}}
\expandafter\gdef\csname R@med.nwcv.kits.V7A1.zthin\endcsname{86.08}
\expandafter\gdef\csname R@med.nwcv.kits.V7A1.zthin.d\endcsname{\ensuremath{+2.27}}
\expandafter\gdef\csname R@med.nwcv.kits.V7C1.dice\endcsname{84.78}
\expandafter\gdef\csname R@med.nwcv.kits.V7C1.dice.d\endcsname{\ensuremath{+0.78}}
\expandafter\gdef\csname R@med.nwcv.kits.V7C1.dice.hi\endcsname{\ensuremath{+2.48}}
\expandafter\gdef\csname R@med.nwcv.kits.V7C1.dice.lo\endcsname{\ensuremath{-0.92}}
\expandafter\gdef\csname R@med.nwcv.kits.V7C1.far\endcsname{36}
\expandafter\gdef\csname R@med.nwcv.kits.V7C1.far.d\endcsname{\ensuremath{-16}}
\expandafter\gdef\csname R@med.nwcv.kits.V7C1.farfp\endcsname{34}
\expandafter\gdef\csname R@med.nwcv.kits.V7C1.farfp.d\endcsname{\ensuremath{-15}}
\expandafter\gdef\csname R@med.nwcv.kits.V7C1.ft\endcsname{55}
\expandafter\gdef\csname R@med.nwcv.kits.V7C1.ft.d\endcsname{\ensuremath{0}}
\expandafter\gdef\csname R@med.nwcv.kits.V7C1.ftfp\endcsname{35}
\expandafter\gdef\csname R@med.nwcv.kits.V7C1.ftfp.d\endcsname{\ensuremath{-5}}
\expandafter\gdef\csname R@med.nwcv.kits.V7C1.gmac\endcsname{257.1}
\expandafter\gdef\csname R@med.nwcv.kits.V7C1.gmac.d\endcsname{\ensuremath{-220.3}}
\expandafter\gdef\csname R@med.nwcv.kits.V7C1.landed\endcsname{1}
\expandafter\gdef\csname R@med.nwcv.kits.V7C1.tc\endcsname{4.27}
\expandafter\gdef\csname R@med.nwcv.kits.V7C1.tc.d\endcsname{\ensuremath{+1.11}}
\expandafter\gdef\csname R@med.nwcv.kits.V7C1.tcn\endcsname{14}
\expandafter\gdef\csname R@med.nwcv.kits.V7C1.tcn.d\endcsname{\ensuremath{0}}
\expandafter\gdef\csname R@med.nwcv.kits.V7C1.zthick\endcsname{84.67}
\expandafter\gdef\csname R@med.nwcv.kits.V7C1.zthick.d\endcsname{\ensuremath{+0.46}}
\expandafter\gdef\csname R@med.nwcv.kits.V7C1.zthin\endcsname{84.88}
\expandafter\gdef\csname R@med.nwcv.kits.V7C1.zthin.d\endcsname{\ensuremath{+1.07}}
\expandafter\gdef\csname R@med.nwcv.kits.X1.dice\endcsname{84.96}
\expandafter\gdef\csname R@med.nwcv.kits.X1.dice.d\endcsname{\ensuremath{+0.97}}
\expandafter\gdef\csname R@med.nwcv.kits.X1.dice.hi\endcsname{\ensuremath{+1.91}}
\expandafter\gdef\csname R@med.nwcv.kits.X1.dice.lo\endcsname{\ensuremath{+0.03}}
\expandafter\gdef\csname R@med.nwcv.kits.X1.far\endcsname{45}
\expandafter\gdef\csname R@med.nwcv.kits.X1.far.d\endcsname{\ensuremath{-7}}
\expandafter\gdef\csname R@med.nwcv.kits.X1.farfp\endcsname{43}
\expandafter\gdef\csname R@med.nwcv.kits.X1.farfp.d\endcsname{\ensuremath{-6}}
\expandafter\gdef\csname R@med.nwcv.kits.X1.ft\endcsname{49}
\expandafter\gdef\csname R@med.nwcv.kits.X1.ft.d\endcsname{\ensuremath{-6}}
\expandafter\gdef\csname R@med.nwcv.kits.X1.ftfp\endcsname{31}
\expandafter\gdef\csname R@med.nwcv.kits.X1.ftfp.d\endcsname{\ensuremath{-9}}
\expandafter\gdef\csname R@med.nwcv.kits.X1.landed\endcsname{1}
\expandafter\gdef\csname R@med.nwcv.kits.X1.tc\endcsname{3.54}
\expandafter\gdef\csname R@med.nwcv.kits.X1.tc.d\endcsname{\ensuremath{+0.38}}
\expandafter\gdef\csname R@med.nwcv.kits.X1.tcn\endcsname{11}
\expandafter\gdef\csname R@med.nwcv.kits.X1.tcn.d\endcsname{\ensuremath{-3}}
\expandafter\gdef\csname R@med.nwcv.kits.X1.zthick\endcsname{84.80}
\expandafter\gdef\csname R@med.nwcv.kits.X1.zthick.d\endcsname{\ensuremath{+0.59}}
\expandafter\gdef\csname R@med.nwcv.kits.X1.zthin\endcsname{85.12}
\expandafter\gdef\csname R@med.nwcv.kits.X1.zthin.d\endcsname{\ensuremath{+1.31}}
\expandafter\gdef\csname R@med.nwcv.kits.X4.dice\endcsname{84.42}
\expandafter\gdef\csname R@med.nwcv.kits.X4.dice.d\endcsname{\ensuremath{+0.43}}
\expandafter\gdef\csname R@med.nwcv.kits.X4.dice.hi\endcsname{\ensuremath{+1.23}}
\expandafter\gdef\csname R@med.nwcv.kits.X4.dice.lo\endcsname{\ensuremath{-0.37}}
\expandafter\gdef\csname R@med.nwcv.kits.X4.far\endcsname{44}
\expandafter\gdef\csname R@med.nwcv.kits.X4.far.d\endcsname{\ensuremath{-8}}
\expandafter\gdef\csname R@med.nwcv.kits.X4.farfp\endcsname{43}
\expandafter\gdef\csname R@med.nwcv.kits.X4.farfp.d\endcsname{\ensuremath{-6}}
\expandafter\gdef\csname R@med.nwcv.kits.X4.ft\endcsname{52}
\expandafter\gdef\csname R@med.nwcv.kits.X4.ft.d\endcsname{\ensuremath{-3}}
\expandafter\gdef\csname R@med.nwcv.kits.X4.ftfp\endcsname{37}
\expandafter\gdef\csname R@med.nwcv.kits.X4.ftfp.d\endcsname{\ensuremath{-3}}
\expandafter\gdef\csname R@med.nwcv.kits.X4.landed\endcsname{1}
\expandafter\gdef\csname R@med.nwcv.kits.X4.tc\endcsname{3.71}
\expandafter\gdef\csname R@med.nwcv.kits.X4.tc.d\endcsname{\ensuremath{+0.55}}
\expandafter\gdef\csname R@med.nwcv.kits.X4.tcn\endcsname{12}
\expandafter\gdef\csname R@med.nwcv.kits.X4.tcn.d\endcsname{\ensuremath{-2}}
\expandafter\gdef\csname R@med.nwcv.kits.X4.zthick\endcsname{84.15}
\expandafter\gdef\csname R@med.nwcv.kits.X4.zthick.d\endcsname{\ensuremath{-0.06}}
\expandafter\gdef\csname R@med.nwcv.kits.X4.zthin\endcsname{84.67}
\expandafter\gdef\csname R@med.nwcv.kits.X4.zthin.d\endcsname{\ensuremath{+0.86}}
\expandafter\gdef\csname R@med.nwcv.kits.X6.dice\endcsname{85.26}
\expandafter\gdef\csname R@med.nwcv.kits.X6.dice.d\endcsname{\ensuremath{+1.27}}
\expandafter\gdef\csname R@med.nwcv.kits.X6.dice.hi\endcsname{\ensuremath{+3.95}}
\expandafter\gdef\csname R@med.nwcv.kits.X6.dice.lo\endcsname{\ensuremath{-1.41}}
\expandafter\gdef\csname R@med.nwcv.kits.X6.far\endcsname{25}
\expandafter\gdef\csname R@med.nwcv.kits.X6.far.d\endcsname{\ensuremath{-27}}
\expandafter\gdef\csname R@med.nwcv.kits.X6.farfp\endcsname{25}
\expandafter\gdef\csname R@med.nwcv.kits.X6.farfp.d\endcsname{\ensuremath{-24}}
\expandafter\gdef\csname R@med.nwcv.kits.X6.ft\endcsname{45}
\expandafter\gdef\csname R@med.nwcv.kits.X6.ft.d\endcsname{\ensuremath{-10}}
\expandafter\gdef\csname R@med.nwcv.kits.X6.ftfp\endcsname{27}
\expandafter\gdef\csname R@med.nwcv.kits.X6.ftfp.d\endcsname{\ensuremath{-13}}
\expandafter\gdef\csname R@med.nwcv.kits.X6.landed\endcsname{1}
\expandafter\gdef\csname R@med.nwcv.kits.X6.tc\endcsname{3.38}
\expandafter\gdef\csname R@med.nwcv.kits.X6.tc.d\endcsname{\ensuremath{+0.22}}
\expandafter\gdef\csname R@med.nwcv.kits.X6.tcn\endcsname{12}
\expandafter\gdef\csname R@med.nwcv.kits.X6.tcn.d\endcsname{\ensuremath{-2}}
\expandafter\gdef\csname R@med.nwcv.kits.X6.zthick\endcsname{85.38}
\expandafter\gdef\csname R@med.nwcv.kits.X6.zthick.d\endcsname{\ensuremath{+1.17}}
\expandafter\gdef\csname R@med.nwcv.kits.X6.zthin\endcsname{85.17}
\expandafter\gdef\csname R@med.nwcv.kits.X6.zthin.d\endcsname{\ensuremath{+1.37}}
\expandafter\gdef\csname R@med.nwcv.kits.Y1.dice\endcsname{85.24}
\expandafter\gdef\csname R@med.nwcv.kits.Y1.dice.d\endcsname{\ensuremath{+1.24}}
\expandafter\gdef\csname R@med.nwcv.kits.Y1.dice.hi\endcsname{\ensuremath{+2.09}}
\expandafter\gdef\csname R@med.nwcv.kits.Y1.dice.lo\endcsname{\ensuremath{+0.40}}
\expandafter\gdef\csname R@med.nwcv.kits.Y1.far\endcsname{25}
\expandafter\gdef\csname R@med.nwcv.kits.Y1.far.d\endcsname{\ensuremath{-27}}
\expandafter\gdef\csname R@med.nwcv.kits.Y1.farfp\endcsname{23}
\expandafter\gdef\csname R@med.nwcv.kits.Y1.farfp.d\endcsname{\ensuremath{-26}}
\expandafter\gdef\csname R@med.nwcv.kits.Y1.ft\endcsname{46}
\expandafter\gdef\csname R@med.nwcv.kits.Y1.ft.d\endcsname{\ensuremath{-9}}
\expandafter\gdef\csname R@med.nwcv.kits.Y1.ftfp\endcsname{30}
\expandafter\gdef\csname R@med.nwcv.kits.Y1.ftfp.d\endcsname{\ensuremath{-10}}
\expandafter\gdef\csname R@med.nwcv.kits.Y1.landed\endcsname{1}
\expandafter\gdef\csname R@med.nwcv.kits.Y1.tc\endcsname{3.45}
\expandafter\gdef\csname R@med.nwcv.kits.Y1.tc.d\endcsname{\ensuremath{+0.29}}
\expandafter\gdef\csname R@med.nwcv.kits.Y1.tcn\endcsname{11}
\expandafter\gdef\csname R@med.nwcv.kits.Y1.tcn.d\endcsname{\ensuremath{-3}}
\expandafter\gdef\csname R@med.nwcv.kits.Y1.zthick\endcsname{85.22}
\expandafter\gdef\csname R@med.nwcv.kits.Y1.zthick.d\endcsname{\ensuremath{+1.01}}
\expandafter\gdef\csname R@med.nwcv.kits.Y1.zthin\endcsname{85.26}
\expandafter\gdef\csname R@med.nwcv.kits.Y1.zthin.d\endcsname{\ensuremath{+1.45}}
\expandafter\gdef\csname R@med.nwcv.kits.Y2.dice\endcsname{86.70}
\expandafter\gdef\csname R@med.nwcv.kits.Y2.dice.d\endcsname{\ensuremath{+2.71}}
\expandafter\gdef\csname R@med.nwcv.kits.Y2.dice.hi\endcsname{\ensuremath{+3.73}}
\expandafter\gdef\csname R@med.nwcv.kits.Y2.dice.lo\endcsname{\ensuremath{+1.68}}
\expandafter\gdef\csname R@med.nwcv.kits.Y2.far\endcsname{18}
\expandafter\gdef\csname R@med.nwcv.kits.Y2.far.d\endcsname{\ensuremath{-34}}
\expandafter\gdef\csname R@med.nwcv.kits.Y2.farfp\endcsname{17}
\expandafter\gdef\csname R@med.nwcv.kits.Y2.farfp.d\endcsname{\ensuremath{-32}}
\expandafter\gdef\csname R@med.nwcv.kits.Y2.ft\endcsname{42}
\expandafter\gdef\csname R@med.nwcv.kits.Y2.ft.d\endcsname{\ensuremath{-13}}
\expandafter\gdef\csname R@med.nwcv.kits.Y2.ftfp\endcsname{21}
\expandafter\gdef\csname R@med.nwcv.kits.Y2.ftfp.d\endcsname{\ensuremath{-19}}
\expandafter\gdef\csname R@med.nwcv.kits.Y2.landed\endcsname{1}
\expandafter\gdef\csname R@med.nwcv.kits.Y2.tc\endcsname{4.01}
\expandafter\gdef\csname R@med.nwcv.kits.Y2.tc.d\endcsname{\ensuremath{+0.85}}
\expandafter\gdef\csname R@med.nwcv.kits.Y2.tcn\endcsname{15}
\expandafter\gdef\csname R@med.nwcv.kits.Y2.tcn.d\endcsname{\ensuremath{+1}}
\expandafter\gdef\csname R@med.nwcv.kits.Y2.zthick\endcsname{86.68}
\expandafter\gdef\csname R@med.nwcv.kits.Y2.zthick.d\endcsname{\ensuremath{+2.47}}
\expandafter\gdef\csname R@med.nwcv.kits.Y2.zthin\endcsname{86.73}
\expandafter\gdef\csname R@med.nwcv.kits.Y2.zthin.d\endcsname{\ensuremath{+2.92}}
\expandafter\gdef\csname R@med.nwcv.kits.base.dice\endcsname{83.99}
\expandafter\gdef\csname R@med.nwcv.kits.base.far\endcsname{52}
\expandafter\gdef\csname R@med.nwcv.kits.base.farfp\endcsname{49}
\expandafter\gdef\csname R@med.nwcv.kits.base.ft\endcsname{55}
\expandafter\gdef\csname R@med.nwcv.kits.base.ftfp\endcsname{40}
\expandafter\gdef\csname R@med.nwcv.kits.base.gmac\endcsname{477.4}
\expandafter\gdef\csname R@med.nwcv.kits.base.landed\endcsname{1}
\expandafter\gdef\csname R@med.nwcv.kits.base.tc\endcsname{3.16}
\expandafter\gdef\csname R@med.nwcv.kits.base.tcn\endcsname{14}
\expandafter\gdef\csname R@med.nwcv.kits.base.zthick\endcsname{84.21}
\expandafter\gdef\csname R@med.nwcv.kits.base.zthin\endcsname{83.81}
\expandafter\gdef\csname R@med.organ.D2.aorta\endcsname{\ensuremath{-0.13}}
\expandafter\gdef\csname R@med.organ.D2.bladder\endcsname{\ensuremath{-1.00}}
\expandafter\gdef\csname R@med.organ.D2.duodenum\endcsname{\ensuremath{-0.45}}
\expandafter\gdef\csname R@med.organ.D2.esophagus\endcsname{\ensuremath{-1.05}}
\expandafter\gdef\csname R@med.organ.D2.gallbladder\endcsname{\ensuremath{-2.32}}
\expandafter\gdef\csname R@med.organ.D2.ivc\endcsname{\ensuremath{-0.34}}
\expandafter\gdef\csname R@med.organ.D2.ladrenal\endcsname{\ensuremath{-0.44}}
\expandafter\gdef\csname R@med.organ.D2.liver\endcsname{\ensuremath{-0.02}}
\expandafter\gdef\csname R@med.organ.D2.lkidney\endcsname{\ensuremath{+0.03}}
\expandafter\gdef\csname R@med.organ.D2.pancreas\endcsname{\ensuremath{-0.71}}
\expandafter\gdef\csname R@med.organ.D2.prostuterus\endcsname{\ensuremath{+0.92}}
\expandafter\gdef\csname R@med.organ.D2.radrenal\endcsname{\ensuremath{-0.71}}
\expandafter\gdef\csname R@med.organ.D2.rkidney\endcsname{\ensuremath{-0.22}}
\expandafter\gdef\csname R@med.organ.D2.spleen\endcsname{\ensuremath{-0.04}}
\expandafter\gdef\csname R@med.organ.D2.stomach\endcsname{\ensuremath{-0.09}}
\expandafter\gdef\csname R@med.organ.H1.aorta\endcsname{\ensuremath{+0.18}}
\expandafter\gdef\csname R@med.organ.H1.bladder\endcsname{\ensuremath{+0.74}}
\expandafter\gdef\csname R@med.organ.H1.duodenum\endcsname{\ensuremath{+0.79}}
\expandafter\gdef\csname R@med.organ.H1.esophagus\endcsname{\ensuremath{-0.04}}
\expandafter\gdef\csname R@med.organ.H1.gallbladder\endcsname{\ensuremath{-1.10}}
\expandafter\gdef\csname R@med.organ.H1.ivc\endcsname{\ensuremath{-0.03}}
\expandafter\gdef\csname R@med.organ.H1.ladrenal\endcsname{\ensuremath{-0.01}}
\expandafter\gdef\csname R@med.organ.H1.liver\endcsname{\ensuremath{-0.11}}
\expandafter\gdef\csname R@med.organ.H1.lkidney\endcsname{\ensuremath{+0.23}}
\expandafter\gdef\csname R@med.organ.H1.pancreas\endcsname{\ensuremath{-0.17}}
\expandafter\gdef\csname R@med.organ.H1.prostuterus\endcsname{\ensuremath{+3.11}}
\expandafter\gdef\csname R@med.organ.H1.radrenal\endcsname{\ensuremath{+0.36}}
\expandafter\gdef\csname R@med.organ.H1.rkidney\endcsname{\ensuremath{-0.41}}
\expandafter\gdef\csname R@med.organ.H1.spleen\endcsname{\ensuremath{-0.03}}
\expandafter\gdef\csname R@med.organ.H1.stomach\endcsname{\ensuremath{+0.70}}
\expandafter\gdef\csname R@med.organ.H2.aorta\endcsname{\ensuremath{+0.11}}
\expandafter\gdef\csname R@med.organ.H2.bladder\endcsname{\ensuremath{+1.68}}
\expandafter\gdef\csname R@med.organ.H2.duodenum\endcsname{\ensuremath{+0.36}}
\expandafter\gdef\csname R@med.organ.H2.esophagus\endcsname{\ensuremath{-0.14}}
\expandafter\gdef\csname R@med.organ.H2.gallbladder\endcsname{\ensuremath{-0.77}}
\expandafter\gdef\csname R@med.organ.H2.ivc\endcsname{\ensuremath{+0.02}}
\expandafter\gdef\csname R@med.organ.H2.ladrenal\endcsname{\ensuremath{+0.12}}
\expandafter\gdef\csname R@med.organ.H2.liver\endcsname{\ensuremath{+0.13}}
\expandafter\gdef\csname R@med.organ.H2.lkidney\endcsname{\ensuremath{-0.22}}
\expandafter\gdef\csname R@med.organ.H2.pancreas\endcsname{\ensuremath{-0.19}}
\expandafter\gdef\csname R@med.organ.H2.prostuterus\endcsname{\ensuremath{+1.84}}
\expandafter\gdef\csname R@med.organ.H2.radrenal\endcsname{\ensuremath{+0.08}}
\expandafter\gdef\csname R@med.organ.H2.rkidney\endcsname{\ensuremath{-0.20}}
\expandafter\gdef\csname R@med.organ.H2.spleen\endcsname{\ensuremath{-0.07}}
\expandafter\gdef\csname R@med.organ.H2.stomach\endcsname{\ensuremath{+0.56}}
\expandafter\gdef\csname R@med.organ.L1.aorta\endcsname{\ensuremath{+0.18}}
\expandafter\gdef\csname R@med.organ.L1.bladder\endcsname{\ensuremath{+1.44}}
\expandafter\gdef\csname R@med.organ.L1.duodenum\endcsname{\ensuremath{+0.90}}
\expandafter\gdef\csname R@med.organ.L1.esophagus\endcsname{\ensuremath{-0.11}}
\expandafter\gdef\csname R@med.organ.L1.gallbladder\endcsname{\ensuremath{+0.66}}
\expandafter\gdef\csname R@med.organ.L1.ivc\endcsname{\ensuremath{+0.05}}
\expandafter\gdef\csname R@med.organ.L1.ladrenal\endcsname{\ensuremath{-0.17}}
\expandafter\gdef\csname R@med.organ.L1.liver\endcsname{\ensuremath{+0.11}}
\expandafter\gdef\csname R@med.organ.L1.lkidney\endcsname{\ensuremath{+0.27}}
\expandafter\gdef\csname R@med.organ.L1.pancreas\endcsname{\ensuremath{-0.18}}
\expandafter\gdef\csname R@med.organ.L1.prostuterus\endcsname{\ensuremath{+2.01}}
\expandafter\gdef\csname R@med.organ.L1.radrenal\endcsname{\ensuremath{+0.17}}
\expandafter\gdef\csname R@med.organ.L1.rkidney\endcsname{\ensuremath{+0.24}}
\expandafter\gdef\csname R@med.organ.L1.spleen\endcsname{\ensuremath{+0.05}}
\expandafter\gdef\csname R@med.organ.L1.stomach\endcsname{\ensuremath{+0.49}}
\expandafter\gdef\csname R@med.organ.L3.aorta\endcsname{\ensuremath{+0.14}}
\expandafter\gdef\csname R@med.organ.L3.bladder\endcsname{\ensuremath{+0.78}}
\expandafter\gdef\csname R@med.organ.L3.duodenum\endcsname{\ensuremath{+0.51}}
\expandafter\gdef\csname R@med.organ.L3.esophagus\endcsname{\ensuremath{+0.07}}
\expandafter\gdef\csname R@med.organ.L3.gallbladder\endcsname{\ensuremath{+0.54}}
\expandafter\gdef\csname R@med.organ.L3.ivc\endcsname{\ensuremath{-0.07}}
\expandafter\gdef\csname R@med.organ.L3.ladrenal\endcsname{\ensuremath{-0.05}}
\expandafter\gdef\csname R@med.organ.L3.liver\endcsname{\ensuremath{+0.08}}
\expandafter\gdef\csname R@med.organ.L3.lkidney\endcsname{\ensuremath{-0.40}}
\expandafter\gdef\csname R@med.organ.L3.pancreas\endcsname{\ensuremath{-0.56}}
\expandafter\gdef\csname R@med.organ.L3.prostuterus\endcsname{\ensuremath{+1.09}}
\expandafter\gdef\csname R@med.organ.L3.radrenal\endcsname{\ensuremath{+0.46}}
\expandafter\gdef\csname R@med.organ.L3.rkidney\endcsname{\ensuremath{-0.12}}
\expandafter\gdef\csname R@med.organ.L3.spleen\endcsname{\ensuremath{+0.03}}
\expandafter\gdef\csname R@med.organ.L3.stomach\endcsname{\ensuremath{+0.92}}
\expandafter\gdef\csname R@med.organ.V5B1.aorta\endcsname{\ensuremath{-0.11}}
\expandafter\gdef\csname R@med.organ.V5B1.bladder\endcsname{\ensuremath{-0.33}}
\expandafter\gdef\csname R@med.organ.V5B1.duodenum\endcsname{\ensuremath{-0.29}}
\expandafter\gdef\csname R@med.organ.V5B1.esophagus\endcsname{\ensuremath{-1.29}}
\expandafter\gdef\csname R@med.organ.V5B1.gallbladder\endcsname{\ensuremath{-1.81}}
\expandafter\gdef\csname R@med.organ.V5B1.ivc\endcsname{\ensuremath{-0.43}}
\expandafter\gdef\csname R@med.organ.V5B1.ladrenal\endcsname{\ensuremath{-0.86}}
\expandafter\gdef\csname R@med.organ.V5B1.liver\endcsname{\ensuremath{-0.10}}
\expandafter\gdef\csname R@med.organ.V5B1.lkidney\endcsname{\ensuremath{-0.24}}
\expandafter\gdef\csname R@med.organ.V5B1.pancreas\endcsname{\ensuremath{-0.64}}
\expandafter\gdef\csname R@med.organ.V5B1.prostuterus\endcsname{\ensuremath{+0.59}}
\expandafter\gdef\csname R@med.organ.V5B1.radrenal\endcsname{\ensuremath{-0.76}}
\expandafter\gdef\csname R@med.organ.V5B1.rkidney\endcsname{\ensuremath{-0.18}}
\expandafter\gdef\csname R@med.organ.V5B1.spleen\endcsname{\ensuremath{-0.20}}
\expandafter\gdef\csname R@med.organ.V5B1.stomach\endcsname{\ensuremath{-0.25}}
\expandafter\gdef\csname R@med.organ.V5B2.aorta\endcsname{\ensuremath{-0.04}}
\expandafter\gdef\csname R@med.organ.V5B2.bladder\endcsname{\ensuremath{+0.30}}
\expandafter\gdef\csname R@med.organ.V5B2.duodenum\endcsname{\ensuremath{-0.59}}
\expandafter\gdef\csname R@med.organ.V5B2.esophagus\endcsname{\ensuremath{-0.96}}
\expandafter\gdef\csname R@med.organ.V5B2.gallbladder\endcsname{\ensuremath{-1.21}}
\expandafter\gdef\csname R@med.organ.V5B2.ivc\endcsname{\ensuremath{-0.42}}
\expandafter\gdef\csname R@med.organ.V5B2.ladrenal\endcsname{\ensuremath{-0.69}}
\expandafter\gdef\csname R@med.organ.V5B2.liver\endcsname{\ensuremath{-0.03}}
\expandafter\gdef\csname R@med.organ.V5B2.lkidney\endcsname{\ensuremath{-0.24}}
\expandafter\gdef\csname R@med.organ.V5B2.pancreas\endcsname{\ensuremath{-0.51}}
\expandafter\gdef\csname R@med.organ.V5B2.prostuterus\endcsname{\ensuremath{+1.10}}
\expandafter\gdef\csname R@med.organ.V5B2.radrenal\endcsname{\ensuremath{-0.17}}
\expandafter\gdef\csname R@med.organ.V5B2.rkidney\endcsname{\ensuremath{-0.17}}
\expandafter\gdef\csname R@med.organ.V5B2.spleen\endcsname{\ensuremath{-0.03}}
\expandafter\gdef\csname R@med.organ.V5B2.stomach\endcsname{\ensuremath{-0.18}}
\expandafter\gdef\csname R@med.organ.V7A1.aorta\endcsname{\ensuremath{+0.13}}
\expandafter\gdef\csname R@med.organ.V7A1.bladder\endcsname{\ensuremath{+1.30}}
\expandafter\gdef\csname R@med.organ.V7A1.duodenum\endcsname{\ensuremath{+0.42}}
\expandafter\gdef\csname R@med.organ.V7A1.esophagus\endcsname{\ensuremath{-0.59}}
\expandafter\gdef\csname R@med.organ.V7A1.gallbladder\endcsname{\ensuremath{-0.24}}
\expandafter\gdef\csname R@med.organ.V7A1.ivc\endcsname{\ensuremath{-0.09}}
\expandafter\gdef\csname R@med.organ.V7A1.ladrenal\endcsname{\ensuremath{-0.06}}
\expandafter\gdef\csname R@med.organ.V7A1.liver\endcsname{\ensuremath{+0.09}}
\expandafter\gdef\csname R@med.organ.V7A1.lkidney\endcsname{\ensuremath{-0.40}}
\expandafter\gdef\csname R@med.organ.V7A1.pancreas\endcsname{\ensuremath{-0.28}}
\expandafter\gdef\csname R@med.organ.V7A1.prostuterus\endcsname{\ensuremath{+2.85}}
\expandafter\gdef\csname R@med.organ.V7A1.radrenal\endcsname{\ensuremath{+0.09}}
\expandafter\gdef\csname R@med.organ.V7A1.rkidney\endcsname{\ensuremath{+0.01}}
\expandafter\gdef\csname R@med.organ.V7A1.spleen\endcsname{\ensuremath{+0.14}}
\expandafter\gdef\csname R@med.organ.V7A1.stomach\endcsname{\ensuremath{+0.53}}
\expandafter\gdef\csname R@med.organ.V7C1.aorta\endcsname{\ensuremath{-0.07}}
\expandafter\gdef\csname R@med.organ.V7C1.bladder\endcsname{\ensuremath{-0.25}}
\expandafter\gdef\csname R@med.organ.V7C1.duodenum\endcsname{\ensuremath{-0.74}}
\expandafter\gdef\csname R@med.organ.V7C1.esophagus\endcsname{\ensuremath{-1.31}}
\expandafter\gdef\csname R@med.organ.V7C1.gallbladder\endcsname{\ensuremath{-1.44}}
\expandafter\gdef\csname R@med.organ.V7C1.ivc\endcsname{\ensuremath{-0.28}}
\expandafter\gdef\csname R@med.organ.V7C1.ladrenal\endcsname{\ensuremath{-0.67}}
\expandafter\gdef\csname R@med.organ.V7C1.liver\endcsname{\ensuremath{+0.06}}
\expandafter\gdef\csname R@med.organ.V7C1.lkidney\endcsname{\ensuremath{-0.46}}
\expandafter\gdef\csname R@med.organ.V7C1.pancreas\endcsname{\ensuremath{-0.64}}
\expandafter\gdef\csname R@med.organ.V7C1.prostuterus\endcsname{\ensuremath{+0.50}}
\expandafter\gdef\csname R@med.organ.V7C1.radrenal\endcsname{\ensuremath{-0.75}}
\expandafter\gdef\csname R@med.organ.V7C1.rkidney\endcsname{\ensuremath{-0.51}}
\expandafter\gdef\csname R@med.organ.V7C1.spleen\endcsname{\ensuremath{-0.09}}
\expandafter\gdef\csname R@med.organ.V7C1.stomach\endcsname{\ensuremath{-0.24}}
\expandafter\gdef\csname R@med.organ.Y2.aorta\endcsname{\ensuremath{+0.06}}
\expandafter\gdef\csname R@med.organ.Y2.bladder\endcsname{\ensuremath{+1.11}}
\expandafter\gdef\csname R@med.organ.Y2.duodenum\endcsname{\ensuremath{+0.58}}
\expandafter\gdef\csname R@med.organ.Y2.esophagus\endcsname{\ensuremath{-0.50}}
\expandafter\gdef\csname R@med.organ.Y2.gallbladder\endcsname{\ensuremath{-0.80}}
\expandafter\gdef\csname R@med.organ.Y2.ivc\endcsname{\ensuremath{-0.16}}
\expandafter\gdef\csname R@med.organ.Y2.ladrenal\endcsname{\ensuremath{-0.02}}
\expandafter\gdef\csname R@med.organ.Y2.liver\endcsname{\ensuremath{+0.06}}
\expandafter\gdef\csname R@med.organ.Y2.lkidney\endcsname{\ensuremath{+0.13}}
\expandafter\gdef\csname R@med.organ.Y2.pancreas\endcsname{\ensuremath{-0.56}}
\expandafter\gdef\csname R@med.organ.Y2.prostuterus\endcsname{\ensuremath{+2.55}}
\expandafter\gdef\csname R@med.organ.Y2.radrenal\endcsname{\ensuremath{+0.19}}
\expandafter\gdef\csname R@med.organ.Y2.rkidney\endcsname{\ensuremath{0.00}}
\expandafter\gdef\csname R@med.organ.Y2.spleen\endcsname{\ensuremath{+0.10}}
\expandafter\gdef\csname R@med.organ.Y2.stomach\endcsname{\ensuremath{+0.66}}
\expandafter\gdef\csname R@med.shape.b.aorta\endcsname{0.23}
\expandafter\gdef\csname R@med.shape.b.bladder\endcsname{0.44}
\expandafter\gdef\csname R@med.shape.b.duodenum\endcsname{0.43}
\expandafter\gdef\csname R@med.shape.b.esophagus\endcsname{0.52}
\expandafter\gdef\csname R@med.shape.b.gallbladder\endcsname{0.47}
\expandafter\gdef\csname R@med.shape.b.ivc\endcsname{0.27}
\expandafter\gdef\csname R@med.shape.b.ladrenal\endcsname{0.75}
\expandafter\gdef\csname R@med.shape.b.liver\endcsname{0.15}
\expandafter\gdef\csname R@med.shape.b.lkidney\endcsname{0.25}
\expandafter\gdef\csname R@med.shape.b.max\endcsname{0.78}
\expandafter\gdef\csname R@med.shape.b.min\endcsname{0.15}
\expandafter\gdef\csname R@med.shape.b.pancreas\endcsname{0.46}
\expandafter\gdef\csname R@med.shape.b.prostuterus\endcsname{0.39}
\expandafter\gdef\csname R@med.shape.b.radrenal\endcsname{0.78}
\expandafter\gdef\csname R@med.shape.b.rkidney\endcsname{0.25}
\expandafter\gdef\csname R@med.shape.b.spleen\endcsname{0.24}
\expandafter\gdef\csname R@med.shape.b.stomach\endcsname{0.28}
\expandafter\gdef\csname R@med.shape.d.D2.aorta\endcsname{\ensuremath{-0.13}}
\expandafter\gdef\csname R@med.shape.d.D2.bladder\endcsname{\ensuremath{-0.43}}
\expandafter\gdef\csname R@med.shape.d.D2.duodenum\endcsname{\ensuremath{-0.45}}
\expandafter\gdef\csname R@med.shape.d.D2.esophagus\endcsname{\ensuremath{-1.05}}
\expandafter\gdef\csname R@med.shape.d.D2.gallbladder\endcsname{\ensuremath{-0.80}}
\expandafter\gdef\csname R@med.shape.d.D2.ivc\endcsname{\ensuremath{-0.34}}
\expandafter\gdef\csname R@med.shape.d.D2.ladrenal\endcsname{\ensuremath{-0.44}}
\expandafter\gdef\csname R@med.shape.d.D2.liver\endcsname{\ensuremath{-0.02}}
\expandafter\gdef\csname R@med.shape.d.D2.lkidney\endcsname{\ensuremath{+0.03}}
\expandafter\gdef\csname R@med.shape.d.D2.pancreas\endcsname{\ensuremath{-0.71}}
\expandafter\gdef\csname R@med.shape.d.D2.prostuterus\endcsname{\ensuremath{-0.17}}
\expandafter\gdef\csname R@med.shape.d.D2.radrenal\endcsname{\ensuremath{-0.71}}
\expandafter\gdef\csname R@med.shape.d.D2.rkidney\endcsname{\ensuremath{-0.22}}
\expandafter\gdef\csname R@med.shape.d.D2.spleen\endcsname{\ensuremath{-0.04}}
\expandafter\gdef\csname R@med.shape.d.D2.stomach\endcsname{\ensuremath{-0.09}}
\expandafter\gdef\csname R@med.shape.d.H1.aorta\endcsname{\ensuremath{+0.18}}
\expandafter\gdef\csname R@med.shape.d.H1.bladder\endcsname{\ensuremath{+1.34}}
\expandafter\gdef\csname R@med.shape.d.H1.duodenum\endcsname{\ensuremath{+0.79}}
\expandafter\gdef\csname R@med.shape.d.H1.esophagus\endcsname{\ensuremath{-0.04}}
\expandafter\gdef\csname R@med.shape.d.H1.gallbladder\endcsname{\ensuremath{+0.50}}
\expandafter\gdef\csname R@med.shape.d.H1.ivc\endcsname{\ensuremath{-0.03}}
\expandafter\gdef\csname R@med.shape.d.H1.ladrenal\endcsname{\ensuremath{-0.01}}
\expandafter\gdef\csname R@med.shape.d.H1.liver\endcsname{\ensuremath{-0.11}}
\expandafter\gdef\csname R@med.shape.d.H1.lkidney\endcsname{\ensuremath{-0.36}}
\expandafter\gdef\csname R@med.shape.d.H1.pancreas\endcsname{\ensuremath{-0.17}}
\expandafter\gdef\csname R@med.shape.d.H1.prostuterus\endcsname{\ensuremath{+1.51}}
\expandafter\gdef\csname R@med.shape.d.H1.radrenal\endcsname{\ensuremath{+0.36}}
\expandafter\gdef\csname R@med.shape.d.H1.rkidney\endcsname{\ensuremath{-0.41}}
\expandafter\gdef\csname R@med.shape.d.H1.spleen\endcsname{\ensuremath{-0.03}}
\expandafter\gdef\csname R@med.shape.d.H1.stomach\endcsname{\ensuremath{+0.71}}
\expandafter\gdef\csname R@med.shape.d.H2.aorta\endcsname{\ensuremath{+0.11}}
\expandafter\gdef\csname R@med.shape.d.H2.bladder\endcsname{\ensuremath{+1.11}}
\expandafter\gdef\csname R@med.shape.d.H2.duodenum\endcsname{\ensuremath{+0.36}}
\expandafter\gdef\csname R@med.shape.d.H2.esophagus\endcsname{\ensuremath{-0.14}}
\expandafter\gdef\csname R@med.shape.d.H2.gallbladder\endcsname{\ensuremath{+0.27}}
\expandafter\gdef\csname R@med.shape.d.H2.ivc\endcsname{\ensuremath{+0.02}}
\expandafter\gdef\csname R@med.shape.d.H2.ladrenal\endcsname{\ensuremath{+0.12}}
\expandafter\gdef\csname R@med.shape.d.H2.liver\endcsname{\ensuremath{+0.13}}
\expandafter\gdef\csname R@med.shape.d.H2.lkidney\endcsname{\ensuremath{-0.22}}
\expandafter\gdef\csname R@med.shape.d.H2.pancreas\endcsname{\ensuremath{-0.19}}
\expandafter\gdef\csname R@med.shape.d.H2.prostuterus\endcsname{\ensuremath{+1.34}}
\expandafter\gdef\csname R@med.shape.d.H2.radrenal\endcsname{\ensuremath{+0.08}}
\expandafter\gdef\csname R@med.shape.d.H2.rkidney\endcsname{\ensuremath{-0.20}}
\expandafter\gdef\csname R@med.shape.d.H2.spleen\endcsname{\ensuremath{-0.07}}
\expandafter\gdef\csname R@med.shape.d.H2.stomach\endcsname{\ensuremath{+0.56}}
\expandafter\gdef\csname R@med.shape.d.L1.aorta\endcsname{\ensuremath{+0.18}}
\expandafter\gdef\csname R@med.shape.d.L1.bladder\endcsname{\ensuremath{+1.48}}
\expandafter\gdef\csname R@med.shape.d.L1.duodenum\endcsname{\ensuremath{+0.90}}
\expandafter\gdef\csname R@med.shape.d.L1.esophagus\endcsname{\ensuremath{-0.11}}
\expandafter\gdef\csname R@med.shape.d.L1.gallbladder\endcsname{\ensuremath{+1.18}}
\expandafter\gdef\csname R@med.shape.d.L1.ivc\endcsname{\ensuremath{+0.05}}
\expandafter\gdef\csname R@med.shape.d.L1.ladrenal\endcsname{\ensuremath{-0.17}}
\expandafter\gdef\csname R@med.shape.d.L1.liver\endcsname{\ensuremath{+0.11}}
\expandafter\gdef\csname R@med.shape.d.L1.lkidney\endcsname{\ensuremath{-0.32}}
\expandafter\gdef\csname R@med.shape.d.L1.pancreas\endcsname{\ensuremath{-0.18}}
\expandafter\gdef\csname R@med.shape.d.L1.prostuterus\endcsname{\ensuremath{+1.51}}
\expandafter\gdef\csname R@med.shape.d.L1.radrenal\endcsname{\ensuremath{+0.17}}
\expandafter\gdef\csname R@med.shape.d.L1.rkidney\endcsname{\ensuremath{+0.24}}
\expandafter\gdef\csname R@med.shape.d.L1.spleen\endcsname{\ensuremath{+0.05}}
\expandafter\gdef\csname R@med.shape.d.L1.stomach\endcsname{\ensuremath{+0.49}}
\expandafter\gdef\csname R@med.shape.d.L3.aorta\endcsname{\ensuremath{+0.14}}
\expandafter\gdef\csname R@med.shape.d.L3.bladder\endcsname{\ensuremath{+1.38}}
\expandafter\gdef\csname R@med.shape.d.L3.duodenum\endcsname{\ensuremath{+0.51}}
\expandafter\gdef\csname R@med.shape.d.L3.esophagus\endcsname{\ensuremath{+0.07}}
\expandafter\gdef\csname R@med.shape.d.L3.gallbladder\endcsname{\ensuremath{+1.13}}
\expandafter\gdef\csname R@med.shape.d.L3.ivc\endcsname{\ensuremath{-0.07}}
\expandafter\gdef\csname R@med.shape.d.L3.ladrenal\endcsname{\ensuremath{-0.05}}
\expandafter\gdef\csname R@med.shape.d.L3.liver\endcsname{\ensuremath{+0.08}}
\expandafter\gdef\csname R@med.shape.d.L3.lkidney\endcsname{\ensuremath{-0.40}}
\expandafter\gdef\csname R@med.shape.d.L3.pancreas\endcsname{\ensuremath{-0.56}}
\expandafter\gdef\csname R@med.shape.d.L3.prostuterus\endcsname{\ensuremath{+1.11}}
\expandafter\gdef\csname R@med.shape.d.L3.radrenal\endcsname{\ensuremath{+0.46}}
\expandafter\gdef\csname R@med.shape.d.L3.rkidney\endcsname{\ensuremath{-0.12}}
\expandafter\gdef\csname R@med.shape.d.L3.spleen\endcsname{\ensuremath{+0.03}}
\expandafter\gdef\csname R@med.shape.d.L3.stomach\endcsname{\ensuremath{+0.93}}
\expandafter\gdef\csname R@med.shape.d.V5B1.aorta\endcsname{\ensuremath{-0.11}}
\expandafter\gdef\csname R@med.shape.d.V5B1.bladder\endcsname{\ensuremath{-0.33}}
\expandafter\gdef\csname R@med.shape.d.V5B1.duodenum\endcsname{\ensuremath{-0.29}}
\expandafter\gdef\csname R@med.shape.d.V5B1.esophagus\endcsname{\ensuremath{-1.29}}
\expandafter\gdef\csname R@med.shape.d.V5B1.gallbladder\endcsname{\ensuremath{-0.80}}
\expandafter\gdef\csname R@med.shape.d.V5B1.ivc\endcsname{\ensuremath{-0.43}}
\expandafter\gdef\csname R@med.shape.d.V5B1.ladrenal\endcsname{\ensuremath{-0.86}}
\expandafter\gdef\csname R@med.shape.d.V5B1.liver\endcsname{\ensuremath{-0.10}}
\expandafter\gdef\csname R@med.shape.d.V5B1.lkidney\endcsname{\ensuremath{-0.24}}
\expandafter\gdef\csname R@med.shape.d.V5B1.pancreas\endcsname{\ensuremath{-0.64}}
\expandafter\gdef\csname R@med.shape.d.V5B1.prostuterus\endcsname{\ensuremath{+0.03}}
\expandafter\gdef\csname R@med.shape.d.V5B1.radrenal\endcsname{\ensuremath{-0.76}}
\expandafter\gdef\csname R@med.shape.d.V5B1.rkidney\endcsname{\ensuremath{-0.18}}
\expandafter\gdef\csname R@med.shape.d.V5B1.spleen\endcsname{\ensuremath{-0.20}}
\expandafter\gdef\csname R@med.shape.d.V5B1.stomach\endcsname{\ensuremath{-0.26}}
\expandafter\gdef\csname R@med.shape.d.V5B2.aorta\endcsname{\ensuremath{-0.04}}
\expandafter\gdef\csname R@med.shape.d.V5B2.bladder\endcsname{\ensuremath{+0.90}}
\expandafter\gdef\csname R@med.shape.d.V5B2.duodenum\endcsname{\ensuremath{-0.59}}
\expandafter\gdef\csname R@med.shape.d.V5B2.esophagus\endcsname{\ensuremath{-0.96}}
\expandafter\gdef\csname R@med.shape.d.V5B2.gallbladder\endcsname{\ensuremath{-0.17}}
\expandafter\gdef\csname R@med.shape.d.V5B2.ivc\endcsname{\ensuremath{-0.42}}
\expandafter\gdef\csname R@med.shape.d.V5B2.ladrenal\endcsname{\ensuremath{-0.69}}
\expandafter\gdef\csname R@med.shape.d.V5B2.liver\endcsname{\ensuremath{-0.03}}
\expandafter\gdef\csname R@med.shape.d.V5B2.lkidney\endcsname{\ensuremath{-0.25}}
\expandafter\gdef\csname R@med.shape.d.V5B2.pancreas\endcsname{\ensuremath{-0.51}}
\expandafter\gdef\csname R@med.shape.d.V5B2.prostuterus\endcsname{\ensuremath{+0.60}}
\expandafter\gdef\csname R@med.shape.d.V5B2.radrenal\endcsname{\ensuremath{-0.17}}
\expandafter\gdef\csname R@med.shape.d.V5B2.rkidney\endcsname{\ensuremath{-0.17}}
\expandafter\gdef\csname R@med.shape.d.V5B2.spleen\endcsname{\ensuremath{-0.03}}
\expandafter\gdef\csname R@med.shape.d.V5B2.stomach\endcsname{\ensuremath{-0.18}}
\expandafter\gdef\csname R@med.shape.d.V7A1.aorta\endcsname{\ensuremath{+0.13}}
\expandafter\gdef\csname R@med.shape.d.V7A1.bladder\endcsname{\ensuremath{+1.30}}
\expandafter\gdef\csname R@med.shape.d.V7A1.duodenum\endcsname{\ensuremath{+0.42}}
\expandafter\gdef\csname R@med.shape.d.V7A1.esophagus\endcsname{\ensuremath{-0.59}}
\expandafter\gdef\csname R@med.shape.d.V7A1.gallbladder\endcsname{\ensuremath{-0.27}}
\expandafter\gdef\csname R@med.shape.d.V7A1.ivc\endcsname{\ensuremath{-0.09}}
\expandafter\gdef\csname R@med.shape.d.V7A1.ladrenal\endcsname{\ensuremath{-0.06}}
\expandafter\gdef\csname R@med.shape.d.V7A1.liver\endcsname{\ensuremath{+0.09}}
\expandafter\gdef\csname R@med.shape.d.V7A1.lkidney\endcsname{\ensuremath{-0.41}}
\expandafter\gdef\csname R@med.shape.d.V7A1.pancreas\endcsname{\ensuremath{-0.28}}
\expandafter\gdef\csname R@med.shape.d.V7A1.prostuterus\endcsname{\ensuremath{+1.25}}
\expandafter\gdef\csname R@med.shape.d.V7A1.radrenal\endcsname{\ensuremath{+0.09}}
\expandafter\gdef\csname R@med.shape.d.V7A1.rkidney\endcsname{\ensuremath{+0.01}}
\expandafter\gdef\csname R@med.shape.d.V7A1.spleen\endcsname{\ensuremath{+0.14}}
\expandafter\gdef\csname R@med.shape.d.V7A1.stomach\endcsname{\ensuremath{+0.53}}
\expandafter\gdef\csname R@med.shape.d.V7C1.aorta\endcsname{\ensuremath{-0.07}}
\expandafter\gdef\csname R@med.shape.d.V7C1.bladder\endcsname{\ensuremath{-0.25}}
\expandafter\gdef\csname R@med.shape.d.V7C1.duodenum\endcsname{\ensuremath{-0.74}}
\expandafter\gdef\csname R@med.shape.d.V7C1.esophagus\endcsname{\ensuremath{-1.31}}
\expandafter\gdef\csname R@med.shape.d.V7C1.gallbladder\endcsname{\ensuremath{-0.41}}
\expandafter\gdef\csname R@med.shape.d.V7C1.ivc\endcsname{\ensuremath{-0.28}}
\expandafter\gdef\csname R@med.shape.d.V7C1.ladrenal\endcsname{\ensuremath{-0.67}}
\expandafter\gdef\csname R@med.shape.d.V7C1.liver\endcsname{\ensuremath{+0.06}}
\expandafter\gdef\csname R@med.shape.d.V7C1.lkidney\endcsname{\ensuremath{-0.46}}
\expandafter\gdef\csname R@med.shape.d.V7C1.pancreas\endcsname{\ensuremath{-0.64}}
\expandafter\gdef\csname R@med.shape.d.V7C1.prostuterus\endcsname{\ensuremath{+0.51}}
\expandafter\gdef\csname R@med.shape.d.V7C1.radrenal\endcsname{\ensuremath{-0.75}}
\expandafter\gdef\csname R@med.shape.d.V7C1.rkidney\endcsname{\ensuremath{-0.51}}
\expandafter\gdef\csname R@med.shape.d.V7C1.spleen\endcsname{\ensuremath{-0.09}}
\expandafter\gdef\csname R@med.shape.d.V7C1.stomach\endcsname{\ensuremath{-0.25}}
\expandafter\gdef\csname R@med.shape.d.Y2.aorta\endcsname{\ensuremath{+0.06}}
\expandafter\gdef\csname R@med.shape.d.Y2.bladder\endcsname{\ensuremath{+1.11}}
\expandafter\gdef\csname R@med.shape.d.Y2.duodenum\endcsname{\ensuremath{+0.58}}
\expandafter\gdef\csname R@med.shape.d.Y2.esophagus\endcsname{\ensuremath{-0.50}}
\expandafter\gdef\csname R@med.shape.d.Y2.gallbladder\endcsname{\ensuremath{+0.25}}
\expandafter\gdef\csname R@med.shape.d.Y2.ivc\endcsname{\ensuremath{-0.16}}
\expandafter\gdef\csname R@med.shape.d.Y2.ladrenal\endcsname{\ensuremath{-0.02}}
\expandafter\gdef\csname R@med.shape.d.Y2.liver\endcsname{\ensuremath{+0.06}}
\expandafter\gdef\csname R@med.shape.d.Y2.lkidney\endcsname{\ensuremath{+0.13}}
\expandafter\gdef\csname R@med.shape.d.Y2.pancreas\endcsname{\ensuremath{-0.56}}
\expandafter\gdef\csname R@med.shape.d.Y2.prostuterus\endcsname{\ensuremath{+1.50}}
\expandafter\gdef\csname R@med.shape.d.Y2.radrenal\endcsname{\ensuremath{+0.19}}
\expandafter\gdef\csname R@med.shape.d.Y2.rkidney\endcsname{\ensuremath{0.00}}
\expandafter\gdef\csname R@med.shape.d.Y2.spleen\endcsname{\ensuremath{+0.10}}
\expandafter\gdef\csname R@med.shape.d.Y2.stomach\endcsname{\ensuremath{+0.67}}
\expandafter\gdef\csname R@med.shape.p.D2\endcsname{\ensuremath{<}0.001}
\expandafter\gdef\csname R@med.shape.p.H1\endcsname{0.25}
\expandafter\gdef\csname R@med.shape.p.H2\endcsname{0.64}
\expandafter\gdef\csname R@med.shape.p.L1\endcsname{0.98}
\expandafter\gdef\csname R@med.shape.p.L3\endcsname{0.47}
\expandafter\gdef\csname R@med.shape.p.V5B1\endcsname{\ensuremath{<}0.001}
\expandafter\gdef\csname R@med.shape.p.V5B2\endcsname{0.074}
\expandafter\gdef\csname R@med.shape.p.V7A1\endcsname{0.36}
\expandafter\gdef\csname R@med.shape.p.V7C1\endcsname{0.0034}
\expandafter\gdef\csname R@med.shape.p.Y2\endcsname{0.99}
\expandafter\gdef\csname R@med.shape.pall.D2\endcsname{0.0026}
\expandafter\gdef\csname R@med.shape.pall.L1\endcsname{0.71}
\expandafter\gdef\csname R@med.shape.pall.L3\endcsname{0.58}
\expandafter\gdef\csname R@med.shape.pall.V5B1\endcsname{\ensuremath{<}0.001}
\expandafter\gdef\csname R@med.shape.pall.V5B2\endcsname{0.034}
\expandafter\gdef\csname R@med.shape.pall.V7A1\endcsname{0.35}
\expandafter\gdef\csname R@med.shape.pall.V7C1\endcsname{0.0013}
\expandafter\gdef\csname R@med.shape.pq.D2\endcsname{\ensuremath{p<0.001}}
\expandafter\gdef\csname R@med.shape.pq.H1\endcsname{\ensuremath{p=0.25}}
\expandafter\gdef\csname R@med.shape.pq.H2\endcsname{\ensuremath{p=0.64}}
\expandafter\gdef\csname R@med.shape.pq.L1\endcsname{\ensuremath{p=0.98}}
\expandafter\gdef\csname R@med.shape.pq.L3\endcsname{\ensuremath{p=0.47}}
\expandafter\gdef\csname R@med.shape.pq.V5B1\endcsname{\ensuremath{p<0.001}}
\expandafter\gdef\csname R@med.shape.pq.V5B2\endcsname{\ensuremath{p=0.074}}
\expandafter\gdef\csname R@med.shape.pq.V7A1\endcsname{\ensuremath{p=0.36}}
\expandafter\gdef\csname R@med.shape.pq.V7C1\endcsname{\ensuremath{p=0.0034}}
\expandafter\gdef\csname R@med.shape.pq.Y2\endcsname{\ensuremath{p=0.99}}
\expandafter\gdef\csname R@med.shape.pqall.D2\endcsname{\ensuremath{p=0.0026}}
\expandafter\gdef\csname R@med.shape.pqall.H1\endcsname{\ensuremath{p=0.73}}
\expandafter\gdef\csname R@med.shape.pqall.H2\endcsname{\ensuremath{p=0.91}}
\expandafter\gdef\csname R@med.shape.pqall.L1\endcsname{\ensuremath{p=0.71}}
\expandafter\gdef\csname R@med.shape.pqall.L3\endcsname{\ensuremath{p=0.58}}
\expandafter\gdef\csname R@med.shape.pqall.V5B1\endcsname{\ensuremath{p<0.001}}
\expandafter\gdef\csname R@med.shape.pqall.V5B2\endcsname{\ensuremath{p=0.034}}
\expandafter\gdef\csname R@med.shape.pqall.V7A1\endcsname{\ensuremath{p=0.35}}
\expandafter\gdef\csname R@med.shape.pqall.V7C1\endcsname{\ensuremath{p=0.0013}}
\expandafter\gdef\csname R@med.shape.pqall.Y2\endcsname{\ensuremath{p=0.57}}
\expandafter\gdef\csname R@med.shape.pvol.D2\endcsname{\ensuremath{<}0.001}
\expandafter\gdef\csname R@med.shape.pvol.L1\endcsname{0.85}
\expandafter\gdef\csname R@med.shape.pvol.L3\endcsname{0.54}
\expandafter\gdef\csname R@med.shape.pvol.V5B1\endcsname{0.0048}
\expandafter\gdef\csname R@med.shape.pvol.V5B2\endcsname{0.12}
\expandafter\gdef\csname R@med.shape.pvol.V7A1\endcsname{0.33}
\expandafter\gdef\csname R@med.shape.pvol.V7C1\endcsname{0.016}
\expandafter\gdef\csname R@med.shape.rho.D2\endcsname{\ensuremath{-0.83}}
\expandafter\gdef\csname R@med.shape.rho.H1\endcsname{\ensuremath{+0.32}}
\expandafter\gdef\csname R@med.shape.rho.H2\endcsname{\ensuremath{+0.13}}
\expandafter\gdef\csname R@med.shape.rho.L1\endcsname{\ensuremath{-0.01}}
\expandafter\gdef\csname R@med.shape.rho.L3\endcsname{\ensuremath{+0.20}}
\expandafter\gdef\csname R@med.shape.rho.V5B1\endcsname{\ensuremath{-0.84}}
\expandafter\gdef\csname R@med.shape.rho.V5B2\endcsname{\ensuremath{-0.47}}
\expandafter\gdef\csname R@med.shape.rho.V7A1\endcsname{\ensuremath{-0.25}}
\expandafter\gdef\csname R@med.shape.rho.V7C1\endcsname{\ensuremath{-0.70}}
\expandafter\gdef\csname R@med.shape.rho.Y2\endcsname{\ensuremath{0.00}}
\expandafter\gdef\csname R@med.shape.rho.bbase\endcsname{\ensuremath{-0.92}}
\expandafter\gdef\csname R@med.shape.rho.bdice\endcsname{\ensuremath{-0.92}}
\expandafter\gdef\csname R@med.shape.rho.bvol\endcsname{\ensuremath{-0.86}}
\expandafter\gdef\csname R@med.shape.rhoall.D2\endcsname{\ensuremath{-0.72}}
\expandafter\gdef\csname R@med.shape.rhoall.H1\endcsname{\ensuremath{+0.10}}
\expandafter\gdef\csname R@med.shape.rhoall.H2\endcsname{\ensuremath{-0.03}}
\expandafter\gdef\csname R@med.shape.rhoall.L1\endcsname{\ensuremath{-0.10}}
\expandafter\gdef\csname R@med.shape.rhoall.L3\endcsname{\ensuremath{+0.15}}
\expandafter\gdef\csname R@med.shape.rhoall.V5B1\endcsname{\ensuremath{-0.83}}
\expandafter\gdef\csname R@med.shape.rhoall.V5B2\endcsname{\ensuremath{-0.55}}
\expandafter\gdef\csname R@med.shape.rhoall.V7A1\endcsname{\ensuremath{-0.26}}
\expandafter\gdef\csname R@med.shape.rhoall.V7C1\endcsname{\ensuremath{-0.75}}
\expandafter\gdef\csname R@med.shape.rhoall.Y2\endcsname{\ensuremath{-0.16}}
\expandafter\gdef\csname R@med.shape.rhomed.D2\endcsname{\ensuremath{-0.83}}
\expandafter\gdef\csname R@med.shape.rhomed.L1\endcsname{\ensuremath{-0.02}}
\expandafter\gdef\csname R@med.shape.rhomed.L3\endcsname{\ensuremath{+0.14}}
\expandafter\gdef\csname R@med.shape.rhomed.V5B1\endcsname{\ensuremath{-0.79}}
\expandafter\gdef\csname R@med.shape.rhomed.V5B2\endcsname{\ensuremath{-0.50}}
\expandafter\gdef\csname R@med.shape.rhomed.V7A1\endcsname{\ensuremath{-0.25}}
\expandafter\gdef\csname R@med.shape.rhomed.V7C1\endcsname{\ensuremath{-0.76}}
\expandafter\gdef\csname R@med.shape.rhosv.D2\endcsname{\ensuremath{-0.83}}
\expandafter\gdef\csname R@med.shape.rhosv.L1\endcsname{\ensuremath{-0.01}}
\expandafter\gdef\csname R@med.shape.rhosv.L3\endcsname{\ensuremath{+0.20}}
\expandafter\gdef\csname R@med.shape.rhosv.V5B1\endcsname{\ensuremath{-0.84}}
\expandafter\gdef\csname R@med.shape.rhosv.V5B2\endcsname{\ensuremath{-0.47}}
\expandafter\gdef\csname R@med.shape.rhosv.V7A1\endcsname{\ensuremath{-0.25}}
\expandafter\gdef\csname R@med.shape.rhosv.V7C1\endcsname{\ensuremath{-0.70}}
\expandafter\gdef\csname R@med.shape.rhovol.D2\endcsname{\ensuremath{+0.81}}
\expandafter\gdef\csname R@med.shape.rhovol.L1\endcsname{\ensuremath{-0.05}}
\expandafter\gdef\csname R@med.shape.rhovol.L3\endcsname{\ensuremath{-0.17}}
\expandafter\gdef\csname R@med.shape.rhovol.V5B1\endcsname{\ensuremath{+0.69}}
\expandafter\gdef\csname R@med.shape.rhovol.V5B2\endcsname{\ensuremath{+0.42}}
\expandafter\gdef\csname R@med.shape.rhovol.V7A1\endcsname{\ensuremath{+0.27}}
\expandafter\gdef\csname R@med.shape.rhovol.V7C1\endcsname{\ensuremath{+0.61}}
\expandafter\gdef\csname R@med.shape.vol.aorta\endcsname{91}
\expandafter\gdef\csname R@med.shape.vol.bladder\endcsname{83}
\expandafter\gdef\csname R@med.shape.vol.duodenum\endcsname{52}
\expandafter\gdef\csname R@med.shape.vol.esophagus\endcsname{12}
\expandafter\gdef\csname R@med.shape.vol.gallbladder\endcsname{24}
\expandafter\gdef\csname R@med.shape.vol.ivc\endcsname{69}
\expandafter\gdef\csname R@med.shape.vol.ladrenal\endcsname{3.8}
\expandafter\gdef\csname R@med.shape.vol.liver\endcsname{1256}
\expandafter\gdef\csname R@med.shape.vol.lkidney\endcsname{143}
\expandafter\gdef\csname R@med.shape.vol.pancreas\endcsname{71}
\expandafter\gdef\csname R@med.shape.vol.prostuterus\endcsname{31}
\expandafter\gdef\csname R@med.shape.vol.radrenal\endcsname{3.2}
\expandafter\gdef\csname R@med.shape.vol.rkidney\endcsname{137}
\expandafter\gdef\csname R@med.shape.vol.spleen\endcsname{179}
\expandafter\gdef\csname R@med.shape.vol.stomach\endcsname{249}
\expandafter\gdef\csname R@med.thick.D2.gap\endcsname{\ensuremath{+0.20}}
\expandafter\gdef\csname R@med.thick.D2.hi\endcsname{\ensuremath{+1.64}}
\expandafter\gdef\csname R@med.thick.D2.lo\endcsname{\ensuremath{-1.28}}
\expandafter\gdef\csname R@med.thick.D2.p\endcsname{0.78}
\expandafter\gdef\csname R@med.thick.D2.pq\endcsname{\ensuremath{p=0.78}}
\expandafter\gdef\csname R@med.thick.D2.thick\endcsname{\ensuremath{+0.94}}
\expandafter\gdef\csname R@med.thick.D2.thin\endcsname{\ensuremath{+1.14}}
\expandafter\gdef\csname R@med.thick.H1.gap\endcsname{\ensuremath{+0.34}}
\expandafter\gdef\csname R@med.thick.H1.hi\endcsname{\ensuremath{+2.23}}
\expandafter\gdef\csname R@med.thick.H1.lo\endcsname{\ensuremath{-1.56}}
\expandafter\gdef\csname R@med.thick.H1.p\endcsname{0.74}
\expandafter\gdef\csname R@med.thick.H1.pq\endcsname{\ensuremath{p=0.74}}
\expandafter\gdef\csname R@med.thick.H1.thick\endcsname{\ensuremath{+1.83}}
\expandafter\gdef\csname R@med.thick.H1.thin\endcsname{\ensuremath{+2.17}}
\expandafter\gdef\csname R@med.thick.H2.gap\endcsname{\ensuremath{+0.12}}
\expandafter\gdef\csname R@med.thick.H2.hi\endcsname{\ensuremath{+1.81}}
\expandafter\gdef\csname R@med.thick.H2.lo\endcsname{\ensuremath{-1.66}}
\expandafter\gdef\csname R@med.thick.H2.p\endcsname{0.89}
\expandafter\gdef\csname R@med.thick.H2.pq\endcsname{\ensuremath{p=0.89}}
\expandafter\gdef\csname R@med.thick.H2.thick\endcsname{\ensuremath{+0.89}}
\expandafter\gdef\csname R@med.thick.H2.thin\endcsname{\ensuremath{+1.02}}
\expandafter\gdef\csname R@med.thick.L1.gap\endcsname{\ensuremath{+1.18}}
\expandafter\gdef\csname R@med.thick.L1.hi\endcsname{\ensuremath{+2.93}}
\expandafter\gdef\csname R@med.thick.L1.lo\endcsname{\ensuremath{-0.43}}
\expandafter\gdef\csname R@med.thick.L1.p\endcsname{0.20}
\expandafter\gdef\csname R@med.thick.L1.pq\endcsname{\ensuremath{p=0.20}}
\expandafter\gdef\csname R@med.thick.L1.thick\endcsname{\ensuremath{+1.52}}
\expandafter\gdef\csname R@med.thick.L1.thin\endcsname{\ensuremath{+2.70}}
\expandafter\gdef\csname R@med.thick.V7A1.fb.hi\endcsname{\ensuremath{+2.91}}
\expandafter\gdef\csname R@med.thick.V7A1.fb.lo\endcsname{\ensuremath{-0.16}}
\expandafter\gdef\csname R@med.thick.V7A1.gap\endcsname{\ensuremath{+1.35}}
\expandafter\gdef\csname R@med.thick.V7A1.gap2\endcsname{\ensuremath{+1.58}}
\expandafter\gdef\csname R@med.thick.V7A1.gap3\endcsname{\ensuremath{+0.97}}
\expandafter\gdef\csname R@med.thick.V7A1.hi\endcsname{\ensuremath{+2.82}}
\expandafter\gdef\csname R@med.thick.V7A1.hi2\endcsname{\ensuremath{+3.41}}
\expandafter\gdef\csname R@med.thick.V7A1.hi3\endcsname{\ensuremath{+2.51}}
\expandafter\gdef\csname R@med.thick.V7A1.lo\endcsname{\ensuremath{-0.28}}
\expandafter\gdef\csname R@med.thick.V7A1.lo2\endcsname{\ensuremath{-0.22}}
\expandafter\gdef\csname R@med.thick.V7A1.lo3\endcsname{\ensuremath{-0.58}}
\expandafter\gdef\csname R@med.thick.V7A1.p\endcsname{0.094}
\expandafter\gdef\csname R@med.thick.V7A1.p2\endcsname{0.11}
\expandafter\gdef\csname R@med.thick.V7A1.p3\endcsname{0.26}
\expandafter\gdef\csname R@med.thick.V7A1.pq\endcsname{\ensuremath{p=0.094}}
\expandafter\gdef\csname R@med.thick.V7A1.pq2\endcsname{\ensuremath{p=0.11}}
\expandafter\gdef\csname R@med.thick.V7A1.pq3\endcsname{\ensuremath{p=0.26}}
\expandafter\gdef\csname R@med.thick.V7A1.thick\endcsname{\ensuremath{+0.92}}
\expandafter\gdef\csname R@med.thick.V7A1.thick2\endcsname{\ensuremath{+1.30}}
\expandafter\gdef\csname R@med.thick.V7A1.thick3\endcsname{\ensuremath{+1.30}}
\expandafter\gdef\csname R@med.thick.V7A1.thin\endcsname{\ensuremath{+2.27}}
\expandafter\gdef\csname R@med.thick.V7A1.thin2\endcsname{\ensuremath{+2.88}}
\expandafter\gdef\csname R@med.thick.V7A1.thin3\endcsname{\ensuremath{+2.27}}
\expandafter\gdef\csname R@med.thick.V7C1.gap\endcsname{\ensuremath{+0.61}}
\expandafter\gdef\csname R@med.thick.V7C1.hi\endcsname{\ensuremath{+2.09}}
\expandafter\gdef\csname R@med.thick.V7C1.lo\endcsname{\ensuremath{-0.82}}
\expandafter\gdef\csname R@med.thick.V7C1.p\endcsname{0.42}
\expandafter\gdef\csname R@med.thick.V7C1.pq\endcsname{\ensuremath{p=0.42}}
\expandafter\gdef\csname R@med.thick.V7C1.thick\endcsname{\ensuremath{+0.46}}
\expandafter\gdef\csname R@med.thick.V7C1.thin\endcsname{\ensuremath{+1.07}}
\expandafter\gdef\csname R@med.thick.X1.gap\endcsname{\ensuremath{+0.72}}
\expandafter\gdef\csname R@med.thick.X1.hi\endcsname{\ensuremath{+2.35}}
\expandafter\gdef\csname R@med.thick.X1.lo\endcsname{\ensuremath{-0.76}}
\expandafter\gdef\csname R@med.thick.X1.p\endcsname{0.36}
\expandafter\gdef\csname R@med.thick.X1.pq\endcsname{\ensuremath{p=0.36}}
\expandafter\gdef\csname R@med.thick.X1.thick\endcsname{\ensuremath{+0.59}}
\expandafter\gdef\csname R@med.thick.X1.thin\endcsname{\ensuremath{+1.31}}
\expandafter\gdef\csname R@med.thick.X4.gap\endcsname{\ensuremath{+0.92}}
\expandafter\gdef\csname R@med.thick.X4.hi\endcsname{\ensuremath{+2.46}}
\expandafter\gdef\csname R@med.thick.X4.lo\endcsname{\ensuremath{-0.66}}
\expandafter\gdef\csname R@med.thick.X4.p\endcsname{0.26}
\expandafter\gdef\csname R@med.thick.X4.pq\endcsname{\ensuremath{p=0.26}}
\expandafter\gdef\csname R@med.thick.X4.thick\endcsname{\ensuremath{-0.06}}
\expandafter\gdef\csname R@med.thick.X4.thin\endcsname{\ensuremath{+0.86}}
\expandafter\gdef\csname R@med.thick.X6.gap\endcsname{\ensuremath{+0.19}}
\expandafter\gdef\csname R@med.thick.X6.hi\endcsname{\ensuremath{+2.10}}
\expandafter\gdef\csname R@med.thick.X6.lo\endcsname{\ensuremath{-1.70}}
\expandafter\gdef\csname R@med.thick.X6.p\endcsname{0.84}
\expandafter\gdef\csname R@med.thick.X6.pq\endcsname{\ensuremath{p=0.84}}
\expandafter\gdef\csname R@med.thick.X6.thick\endcsname{\ensuremath{+1.17}}
\expandafter\gdef\csname R@med.thick.X6.thin\endcsname{\ensuremath{+1.37}}
\expandafter\gdef\csname R@med.thick.Y1.gap\endcsname{\ensuremath{+0.43}}
\expandafter\gdef\csname R@med.thick.Y1.hi\endcsname{\ensuremath{+2.01}}
\expandafter\gdef\csname R@med.thick.Y1.lo\endcsname{\ensuremath{-0.98}}
\expandafter\gdef\csname R@med.thick.Y1.p\endcsname{0.59}
\expandafter\gdef\csname R@med.thick.Y1.pq\endcsname{\ensuremath{p=0.59}}
\expandafter\gdef\csname R@med.thick.Y1.thick\endcsname{\ensuremath{+1.01}}
\expandafter\gdef\csname R@med.thick.Y1.thin\endcsname{\ensuremath{+1.45}}
\expandafter\gdef\csname R@med.thick.Y2.gap\endcsname{\ensuremath{+0.45}}
\expandafter\gdef\csname R@med.thick.Y2.hi\endcsname{\ensuremath{+2.58}}
\expandafter\gdef\csname R@med.thick.Y2.lo\endcsname{\ensuremath{-1.37}}
\expandafter\gdef\csname R@med.thick.Y2.p\endcsname{0.67}
\expandafter\gdef\csname R@med.thick.Y2.pq\endcsname{\ensuremath{p=0.67}}
\expandafter\gdef\csname R@med.thick.Y2.thick\endcsname{\ensuremath{+2.47}}
\expandafter\gdef\csname R@med.thick.Y2.thin\endcsname{\ensuremath{+2.92}}
\expandafter\gdef\csname R@med.thick.agree\endcsname{391}
\expandafter\gdef\csname R@med.thick.n5\endcsname{179}
\expandafter\gdef\csname R@med.thick.nthick\endcsname{183}
\expandafter\gdef\csname R@med.thick.nthin\endcsname{208}
\expandafter\gdef\csname R@med.thick.thr\endcsname{4}
\expandafter\gdef\csname R@med.thick.thr2\endcsname{2}
\expandafter\gdef\csname R@med.thick.thr3\endcsname{3}
\expandafter\gdef\csname R@med.thick.zmax\endcsname{5}
\expandafter\gdef\csname R@med.within.gap.D2\endcsname{\ensuremath{+0.02}}
\expandafter\gdef\csname R@med.within.gap.V7A1\endcsname{\ensuremath{+0.01}}
\expandafter\gdef\csname R@med.within.gap.V7C1\endcsname{\ensuremath{-0.12}}
\expandafter\gdef\csname R@med.within.hi.D2\endcsname{\ensuremath{+0.32}}
\expandafter\gdef\csname R@med.within.hi.V7A1\endcsname{\ensuremath{+0.42}}
\expandafter\gdef\csname R@med.within.hi.V7C1\endcsname{\ensuremath{+0.23}}
\expandafter\gdef\csname R@med.within.lo.D2\endcsname{\ensuremath{-0.25}}
\expandafter\gdef\csname R@med.within.lo.V7A1\endcsname{\ensuremath{-0.35}}
\expandafter\gdef\csname R@med.within.lo.V7C1\endcsname{\ensuremath{-0.47}}
\expandafter\gdef\csname R@med.within.n\endcsname{2375}
\expandafter\gdef\csname R@med.within.p.D2\endcsname{0.80}
\expandafter\gdef\csname R@med.within.p.V7A1\endcsname{0.20}
\expandafter\gdef\csname R@med.within.p.V7C1\endcsname{0.71}
\expandafter\gdef\csname R@med.within.pq.D2\endcsname{\ensuremath{p=0.80}}
\expandafter\gdef\csname R@med.within.pq.V7A1\endcsname{\ensuremath{p=0.20}}
\expandafter\gdef\csname R@med.within.pq.V7C1\endcsname{\ensuremath{p=0.71}}
\expandafter\gdef\csname R@med.within.pred.D2\endcsname{\ensuremath{-0.20}}
\expandafter\gdef\csname R@med.within.pred.V7A1\endcsname{\ensuremath{-0.03}}
\expandafter\gdef\csname R@med.within.pred.V7C1\endcsname{\ensuremath{-0.20}}
\expandafter\gdef\csname R@med.within.rho.D2\endcsname{\ensuremath{-0.005}}
\expandafter\gdef\csname R@med.within.rho.V7A1\endcsname{\ensuremath{+0.026}}
\expandafter\gdef\csname R@med.within.rho.V7C1\endcsname{\ensuremath{-0.008}}
\makeatother

\makeatletter
\expandafter\gdef\csname R@mr.r1.acdc.fine2gmac\endcsname{132}
\expandafter\gdef\csname R@mr.r1.acdc.gmac\endcsname{41.7}
\expandafter\gdef\csname R@mr.r1.acdc.params\endcsname{1.4}
\expandafter\gdef\csname R@mr.r1.acdc.ratio\endcsname{29}
\expandafter\gdef\csname R@mr.r1.acdc.totgmac\endcsname{315}
\expandafter\gdef\csname R@mr.r1.acdc.totparams\endcsname{30.4}
\expandafter\gdef\csname R@mr.r1.amos.fine2gmac\endcsname{225}
\expandafter\gdef\csname R@mr.r1.amos.gmac\endcsname{42.8}
\expandafter\gdef\csname R@mr.r1.amos.params\endcsname{1.4}
\expandafter\gdef\csname R@mr.r1.amos.ratio\endcsname{31}
\expandafter\gdef\csname R@mr.r1.amos.totgmac\endcsname{527}
\expandafter\gdef\csname R@mr.r1.amos.totparams\endcsname{31.1}
\expandafter\gdef\csname R@mr.r1.brats.fine2gmac\endcsname{267}
\expandafter\gdef\csname R@mr.r1.brats.gmac\endcsname{55.4}
\expandafter\gdef\csname R@mr.r1.brats.params\endcsname{1.6}
\expandafter\gdef\csname R@mr.r1.brats.ratio\endcsname{35}
\expandafter\gdef\csname R@mr.r1.brats.totgmac\endcsname{483}
\expandafter\gdef\csname R@mr.r1.brats.totparams\endcsname{31.2}
\expandafter\gdef\csname R@mr.r1.btcv.fine2gmac\endcsname{203}
\expandafter\gdef\csname R@mr.r1.btcv.gmac\endcsname{42.8}
\expandafter\gdef\csname R@mr.r1.btcv.params\endcsname{1.4}
\expandafter\gdef\csname R@mr.r1.btcv.ratio\endcsname{30}
\expandafter\gdef\csname R@mr.r1.btcv.totgmac\endcsname{475}
\expandafter\gdef\csname R@mr.r1.btcv.totparams\endcsname{30.7}
\expandafter\gdef\csname R@mr.r1.gmac.max\endcsname{56.0}
\expandafter\gdef\csname R@mr.r1.gmac.min\endcsname{41.7}
\expandafter\gdef\csname R@mr.r1.kits.fine2gmac\endcsname{267}
\expandafter\gdef\csname R@mr.r1.kits.gmac\endcsname{56.0}
\expandafter\gdef\csname R@mr.r1.kits.params\endcsname{1.6}
\expandafter\gdef\csname R@mr.r1.kits.ratio\endcsname{35}
\expandafter\gdef\csname R@mr.r1.kits.totgmac\endcsname{477}
\expandafter\gdef\csname R@mr.r1.kits.totparams\endcsname{31.2}
\expandafter\gdef\csname R@mr.r1.lits.fine2gmac\endcsname{267}
\expandafter\gdef\csname R@mr.r1.lits.gmac\endcsname{56.0}
\expandafter\gdef\csname R@mr.r1.lits.params\endcsname{1.6}
\expandafter\gdef\csname R@mr.r1.lits.ratio\endcsname{35}
\expandafter\gdef\csname R@mr.r1.lits.totgmac\endcsname{477}
\expandafter\gdef\csname R@mr.r1.lits.totparams\endcsname{31.2}
\expandafter\gdef\csname R@mr.r1.params.max\endcsname{1.6}
\expandafter\gdef\csname R@mr.r1.params.min\endcsname{1.4}
\expandafter\gdef\csname R@mr.r2.acdc.above\endcsname{RV, myocardium, LV cavity}
\expandafter\gdef\csname R@mr.r2.acdc.bfmax\endcsname{0.68}
\expandafter\gdef\csname R@mr.r2.acdc.bfmed\endcsname{0.63}
\expandafter\gdef\csname R@mr.r2.acdc.inst\endcsname{92}
\expandafter\gdef\csname R@mr.r2.acdc.lv.bf\endcsname{0.42}
\expandafter\gdef\csname R@mr.r2.acdc.lv.inst\endcsname{80}
\expandafter\gdef\csname R@mr.r2.acdc.lv.n\endcsname{160}
\expandafter\gdef\csname R@mr.r2.acdc.lv.vol\endcsname{125}
\expandafter\gdef\csname R@mr.r2.acdc.myo.bf\endcsname{0.68}
\expandafter\gdef\csname R@mr.r2.acdc.myo.inst\endcsname{100}
\expandafter\gdef\csname R@mr.r2.acdc.myo.n\endcsname{160}
\expandafter\gdef\csname R@mr.r2.acdc.myo.vol\endcsname{128}
\expandafter\gdef\csname R@mr.r2.acdc.nabove\endcsname{3}
\expandafter\gdef\csname R@mr.r2.acdc.ncases\endcsname{160}
\expandafter\gdef\csname R@mr.r2.acdc.nclass\endcsname{3}
\expandafter\gdef\csname R@mr.r2.acdc.rv.bf\endcsname{0.63}
\expandafter\gdef\csname R@mr.r2.acdc.rv.inst\endcsname{97}
\expandafter\gdef\csname R@mr.r2.acdc.rv.n\endcsname{160}
\expandafter\gdef\csname R@mr.r2.acdc.rv.vol\endcsname{111}
\expandafter\gdef\csname R@mr.r2.acdc.shabove\endcsname{100}
\expandafter\gdef\csname R@mr.r2.acdc.volmed\endcsname{125}
\expandafter\gdef\csname R@mr.r2.amos.above\endcsname{gallbladder, oesophagus, pancreas, R adrenal, L adrenal, duodenum, bladder, prostate/uterus}
\expandafter\gdef\csname R@mr.r2.amos.aorta.bf\endcsname{0.23}
\expandafter\gdef\csname R@mr.r2.amos.aorta.inst\endcsname{2}
\expandafter\gdef\csname R@mr.r2.amos.aorta.n\endcsname{160}
\expandafter\gdef\csname R@mr.r2.amos.aorta.vol\endcsname{91}
\expandafter\gdef\csname R@mr.r2.amos.bfmax\endcsname{0.78}
\expandafter\gdef\csname R@mr.r2.amos.bfmed\endcsname{0.39}
\expandafter\gdef\csname R@mr.r2.amos.bladder.bf\endcsname{0.44}
\expandafter\gdef\csname R@mr.r2.amos.bladder.inst\endcsname{61}
\expandafter\gdef\csname R@mr.r2.amos.bladder.n\endcsname{155}
\expandafter\gdef\csname R@mr.r2.amos.bladder.vol\endcsname{83}
\expandafter\gdef\csname R@mr.r2.amos.duodenum.bf\endcsname{0.43}
\expandafter\gdef\csname R@mr.r2.amos.duodenum.inst\endcsname{80}
\expandafter\gdef\csname R@mr.r2.amos.duodenum.n\endcsname{160}
\expandafter\gdef\csname R@mr.r2.amos.duodenum.vol\endcsname{52}
\expandafter\gdef\csname R@mr.r2.amos.esophagus.bf\endcsname{0.52}
\expandafter\gdef\csname R@mr.r2.amos.esophagus.inst\endcsname{85}
\expandafter\gdef\csname R@mr.r2.amos.esophagus.n\endcsname{159}
\expandafter\gdef\csname R@mr.r2.amos.esophagus.vol\endcsname{12}
\expandafter\gdef\csname R@mr.r2.amos.gallbladder.bf\endcsname{0.47}
\expandafter\gdef\csname R@mr.r2.amos.gallbladder.inst\endcsname{78}
\expandafter\gdef\csname R@mr.r2.amos.gallbladder.n\endcsname{151}
\expandafter\gdef\csname R@mr.r2.amos.gallbladder.vol\endcsname{24}
\expandafter\gdef\csname R@mr.r2.amos.inst\endcsname{48}
\expandafter\gdef\csname R@mr.r2.amos.ivc.bf\endcsname{0.27}
\expandafter\gdef\csname R@mr.r2.amos.ivc.inst\endcsname{17}
\expandafter\gdef\csname R@mr.r2.amos.ivc.n\endcsname{160}
\expandafter\gdef\csname R@mr.r2.amos.ivc.vol\endcsname{69}
\expandafter\gdef\csname R@mr.r2.amos.ladrenal.bf\endcsname{0.75}
\expandafter\gdef\csname R@mr.r2.amos.ladrenal.inst\endcsname{100}
\expandafter\gdef\csname R@mr.r2.amos.ladrenal.n\endcsname{160}
\expandafter\gdef\csname R@mr.r2.amos.ladrenal.vol\endcsname{3.8}
\expandafter\gdef\csname R@mr.r2.amos.liver.bf\endcsname{0.15}
\expandafter\gdef\csname R@mr.r2.amos.liver.inst\endcsname{0}
\expandafter\gdef\csname R@mr.r2.amos.liver.n\endcsname{160}
\expandafter\gdef\csname R@mr.r2.amos.liver.vol\endcsname{1256}
\expandafter\gdef\csname R@mr.r2.amos.lkidney.bf\endcsname{0.25}
\expandafter\gdef\csname R@mr.r2.amos.lkidney.inst\endcsname{6}
\expandafter\gdef\csname R@mr.r2.amos.lkidney.n\endcsname{159}
\expandafter\gdef\csname R@mr.r2.amos.lkidney.vol\endcsname{143}
\expandafter\gdef\csname R@mr.r2.amos.nabove\endcsname{8}
\expandafter\gdef\csname R@mr.r2.amos.ncases\endcsname{160}
\expandafter\gdef\csname R@mr.r2.amos.nclass\endcsname{15}
\expandafter\gdef\csname R@mr.r2.amos.pancreas.bf\endcsname{0.46}
\expandafter\gdef\csname R@mr.r2.amos.pancreas.inst\endcsname{82}
\expandafter\gdef\csname R@mr.r2.amos.pancreas.n\endcsname{160}
\expandafter\gdef\csname R@mr.r2.amos.pancreas.vol\endcsname{71}
\expandafter\gdef\csname R@mr.r2.amos.prostuterus.bf\endcsname{0.39}
\expandafter\gdef\csname R@mr.r2.amos.prostuterus.inst\endcsname{67}
\expandafter\gdef\csname R@mr.r2.amos.prostuterus.n\endcsname{153}
\expandafter\gdef\csname R@mr.r2.amos.prostuterus.vol\endcsname{31}
\expandafter\gdef\csname R@mr.r2.amos.radrenal.bf\endcsname{0.78}
\expandafter\gdef\csname R@mr.r2.amos.radrenal.inst\endcsname{100}
\expandafter\gdef\csname R@mr.r2.amos.radrenal.n\endcsname{160}
\expandafter\gdef\csname R@mr.r2.amos.radrenal.vol\endcsname{3.2}
\expandafter\gdef\csname R@mr.r2.amos.rkidney.bf\endcsname{0.25}
\expandafter\gdef\csname R@mr.r2.amos.rkidney.inst\endcsname{3}
\expandafter\gdef\csname R@mr.r2.amos.rkidney.n\endcsname{160}
\expandafter\gdef\csname R@mr.r2.amos.rkidney.vol\endcsname{137}
\expandafter\gdef\csname R@mr.r2.amos.shabove\endcsname{53}
\expandafter\gdef\csname R@mr.r2.amos.spleen.bf\endcsname{0.24}
\expandafter\gdef\csname R@mr.r2.amos.spleen.inst\endcsname{11}
\expandafter\gdef\csname R@mr.r2.amos.spleen.n\endcsname{160}
\expandafter\gdef\csname R@mr.r2.amos.spleen.vol\endcsname{179}
\expandafter\gdef\csname R@mr.r2.amos.stomach.bf\endcsname{0.28}
\expandafter\gdef\csname R@mr.r2.amos.stomach.inst\endcsname{26}
\expandafter\gdef\csname R@mr.r2.amos.stomach.n\endcsname{158}
\expandafter\gdef\csname R@mr.r2.amos.stomach.vol\endcsname{249}
\expandafter\gdef\csname R@mr.r2.amos.volmed\endcsname{71}
\expandafter\gdef\csname R@mr.r2.brats.above\endcsname{enhancing tumour}
\expandafter\gdef\csname R@mr.r2.brats.bfmax\endcsname{0.44}
\expandafter\gdef\csname R@mr.r2.brats.bfmed\endcsname{0.18}
\expandafter\gdef\csname R@mr.r2.brats.et.bf\endcsname{0.44}
\expandafter\gdef\csname R@mr.r2.brats.et.inst\endcsname{80}
\expandafter\gdef\csname R@mr.r2.brats.et.n\endcsname{20}
\expandafter\gdef\csname R@mr.r2.brats.et.vol\endcsname{21}
\expandafter\gdef\csname R@mr.r2.brats.inst\endcsname{27}
\expandafter\gdef\csname R@mr.r2.brats.nabove\endcsname{1}
\expandafter\gdef\csname R@mr.r2.brats.ncases\endcsname{20}
\expandafter\gdef\csname R@mr.r2.brats.nclass\endcsname{3}
\expandafter\gdef\csname R@mr.r2.brats.shabove\endcsname{33}
\expandafter\gdef\csname R@mr.r2.brats.tc.bf\endcsname{0.18}
\expandafter\gdef\csname R@mr.r2.brats.tc.inst\endcsname{0}
\expandafter\gdef\csname R@mr.r2.brats.tc.n\endcsname{20}
\expandafter\gdef\csname R@mr.r2.brats.tc.vol\endcsname{44}
\expandafter\gdef\csname R@mr.r2.brats.volmed\endcsname{44}
\expandafter\gdef\csname R@mr.r2.brats.wt.bf\endcsname{0.16}
\expandafter\gdef\csname R@mr.r2.brats.wt.inst\endcsname{0}
\expandafter\gdef\csname R@mr.r2.brats.wt.n\endcsname{20}
\expandafter\gdef\csname R@mr.r2.brats.wt.vol\endcsname{116}
\expandafter\gdef\csname R@mr.r2.btcv.above\endcsname{gallbladder, oesophagus, portal/splenic vein, pancreas, R adrenal, L adrenal}
\expandafter\gdef\csname R@mr.r2.btcv.aorta.bf\endcsname{0.23}
\expandafter\gdef\csname R@mr.r2.btcv.aorta.inst\endcsname{0}
\expandafter\gdef\csname R@mr.r2.btcv.aorta.n\endcsname{20}
\expandafter\gdef\csname R@mr.r2.btcv.aorta.vol\endcsname{83}
\expandafter\gdef\csname R@mr.r2.btcv.bfmax\endcsname{0.73}
\expandafter\gdef\csname R@mr.r2.btcv.bfmed\endcsname{0.26}
\expandafter\gdef\csname R@mr.r2.btcv.esophagus.bf\endcsname{0.47}
\expandafter\gdef\csname R@mr.r2.btcv.esophagus.inst\endcsname{95}
\expandafter\gdef\csname R@mr.r2.btcv.esophagus.n\endcsname{20}
\expandafter\gdef\csname R@mr.r2.btcv.esophagus.vol\endcsname{15}
\expandafter\gdef\csname R@mr.r2.btcv.gallbladder.bf\endcsname{0.47}
\expandafter\gdef\csname R@mr.r2.btcv.gallbladder.inst\endcsname{84}
\expandafter\gdef\csname R@mr.r2.btcv.gallbladder.n\endcsname{19}
\expandafter\gdef\csname R@mr.r2.btcv.gallbladder.vol\endcsname{17}
\expandafter\gdef\csname R@mr.r2.btcv.inst\endcsname{46}
\expandafter\gdef\csname R@mr.r2.btcv.ivc.bf\endcsname{0.26}
\expandafter\gdef\csname R@mr.r2.btcv.ivc.inst\endcsname{10}
\expandafter\gdef\csname R@mr.r2.btcv.ivc.n\endcsname{20}
\expandafter\gdef\csname R@mr.r2.btcv.ivc.vol\endcsname{81}
\expandafter\gdef\csname R@mr.r2.btcv.ladrenal.bf\endcsname{0.71}
\expandafter\gdef\csname R@mr.r2.btcv.ladrenal.inst\endcsname{100}
\expandafter\gdef\csname R@mr.r2.btcv.ladrenal.n\endcsname{20}
\expandafter\gdef\csname R@mr.r2.btcv.ladrenal.vol\endcsname{5.3}
\expandafter\gdef\csname R@mr.r2.btcv.liver.bf\endcsname{0.13}
\expandafter\gdef\csname R@mr.r2.btcv.liver.inst\endcsname{0}
\expandafter\gdef\csname R@mr.r2.btcv.liver.n\endcsname{20}
\expandafter\gdef\csname R@mr.r2.btcv.liver.vol\endcsname{1668}
\expandafter\gdef\csname R@mr.r2.btcv.lkidney.bf\endcsname{0.26}
\expandafter\gdef\csname R@mr.r2.btcv.lkidney.inst\endcsname{10}
\expandafter\gdef\csname R@mr.r2.btcv.lkidney.n\endcsname{20}
\expandafter\gdef\csname R@mr.r2.btcv.lkidney.vol\endcsname{158}
\expandafter\gdef\csname R@mr.r2.btcv.nabove\endcsname{6}
\expandafter\gdef\csname R@mr.r2.btcv.ncases\endcsname{20}
\expandafter\gdef\csname R@mr.r2.btcv.nclass\endcsname{13}
\expandafter\gdef\csname R@mr.r2.btcv.pancreas.bf\endcsname{0.43}
\expandafter\gdef\csname R@mr.r2.btcv.pancreas.inst\endcsname{80}
\expandafter\gdef\csname R@mr.r2.btcv.pancreas.n\endcsname{20}
\expandafter\gdef\csname R@mr.r2.btcv.pancreas.vol\endcsname{79}
\expandafter\gdef\csname R@mr.r2.btcv.psv.bf\endcsname{0.71}
\expandafter\gdef\csname R@mr.r2.btcv.psv.inst\endcsname{100}
\expandafter\gdef\csname R@mr.r2.btcv.psv.n\endcsname{20}
\expandafter\gdef\csname R@mr.r2.btcv.psv.vol\endcsname{29}
\expandafter\gdef\csname R@mr.r2.btcv.radrenal.bf\endcsname{0.73}
\expandafter\gdef\csname R@mr.r2.btcv.radrenal.inst\endcsname{100}
\expandafter\gdef\csname R@mr.r2.btcv.radrenal.n\endcsname{20}
\expandafter\gdef\csname R@mr.r2.btcv.radrenal.vol\endcsname{4.2}
\expandafter\gdef\csname R@mr.r2.btcv.rkidney.bf\endcsname{0.26}
\expandafter\gdef\csname R@mr.r2.btcv.rkidney.inst\endcsname{15}
\expandafter\gdef\csname R@mr.r2.btcv.rkidney.n\endcsname{20}
\expandafter\gdef\csname R@mr.r2.btcv.rkidney.vol\endcsname{151}
\expandafter\gdef\csname R@mr.r2.btcv.shabove\endcsname{46}
\expandafter\gdef\csname R@mr.r2.btcv.spleen.bf\endcsname{0.20}
\expandafter\gdef\csname R@mr.r2.btcv.spleen.inst\endcsname{0}
\expandafter\gdef\csname R@mr.r2.btcv.spleen.n\endcsname{20}
\expandafter\gdef\csname R@mr.r2.btcv.spleen.vol\endcsname{214}
\expandafter\gdef\csname R@mr.r2.btcv.stomach.bf\endcsname{0.22}
\expandafter\gdef\csname R@mr.r2.btcv.stomach.inst\endcsname{0}
\expandafter\gdef\csname R@mr.r2.btcv.stomach.n\endcsname{20}
\expandafter\gdef\csname R@mr.r2.btcv.stomach.vol\endcsname{402}
\expandafter\gdef\csname R@mr.r2.btcv.volmed\endcsname{81}
\expandafter\gdef\csname R@mr.r2.kits.above\endcsname{none}
\expandafter\gdef\csname R@mr.r2.kits.bfmax\endcsname{0.31}
\expandafter\gdef\csname R@mr.r2.kits.bfmed\endcsname{0.31}
\expandafter\gdef\csname R@mr.r2.kits.inst\endcsname{25}
\expandafter\gdef\csname R@mr.r2.kits.kidney.bf\endcsname{0.17}
\expandafter\gdef\csname R@mr.r2.kits.kidney.inst\endcsname{0}
\expandafter\gdef\csname R@mr.r2.kits.kidney.n\endcsname{391}
\expandafter\gdef\csname R@mr.r2.kits.kidney.vol\endcsname{473}
\expandafter\gdef\csname R@mr.r2.kits.masses.bf\endcsname{0.31}
\expandafter\gdef\csname R@mr.r2.kits.masses.inst\endcsname{39}
\expandafter\gdef\csname R@mr.r2.kits.masses.n\endcsname{391}
\expandafter\gdef\csname R@mr.r2.kits.masses.vol\endcsname{47}
\expandafter\gdef\csname R@mr.r2.kits.nabove\endcsname{0}
\expandafter\gdef\csname R@mr.r2.kits.ncases\endcsname{391}
\expandafter\gdef\csname R@mr.r2.kits.nclass\endcsname{3}
\expandafter\gdef\csname R@mr.r2.kits.shabove\endcsname{0}
\expandafter\gdef\csname R@mr.r2.kits.tumour.bf\endcsname{0.31}
\expandafter\gdef\csname R@mr.r2.kits.tumour.inst\endcsname{37}
\expandafter\gdef\csname R@mr.r2.kits.tumour.n\endcsname{391}
\expandafter\gdef\csname R@mr.r2.kits.tumour.vol\endcsname{34}
\expandafter\gdef\csname R@mr.r2.kits.volmed\endcsname{47}
\expandafter\gdef\csname R@mr.r2.lits.above\endcsname{tumour}
\expandafter\gdef\csname R@mr.r2.lits.bfmax\endcsname{0.36}
\expandafter\gdef\csname R@mr.r2.lits.bfmed\endcsname{0.22}
\expandafter\gdef\csname R@mr.r2.lits.inst\endcsname{21}
\expandafter\gdef\csname R@mr.r2.lits.liver.bf\endcsname{0.08}
\expandafter\gdef\csname R@mr.r2.lits.liver.inst\endcsname{0}
\expandafter\gdef\csname R@mr.r2.lits.liver.n\endcsname{20}
\expandafter\gdef\csname R@mr.r2.lits.liver.vol\endcsname{1591}
\expandafter\gdef\csname R@mr.r2.lits.nabove\endcsname{1}
\expandafter\gdef\csname R@mr.r2.lits.ncases\endcsname{20}
\expandafter\gdef\csname R@mr.r2.lits.nclass\endcsname{2}
\expandafter\gdef\csname R@mr.r2.lits.shabove\endcsname{50}
\expandafter\gdef\csname R@mr.r2.lits.tumour.bf\endcsname{0.36}
\expandafter\gdef\csname R@mr.r2.lits.tumour.inst\endcsname{44}
\expandafter\gdef\csname R@mr.r2.lits.tumour.n\endcsname{18}
\expandafter\gdef\csname R@mr.r2.lits.tumour.vol\endcsname{18}
\expandafter\gdef\csname R@mr.r2.lits.volmed\endcsname{805}
\expandafter\gdef\csname R@mr.r2.nabove.all\endcsname{19}
\expandafter\gdef\csname R@mr.r2.nabove.med\endcsname{19}
\expandafter\gdef\csname R@mr.r2.nclass.all\endcsname{39}
\expandafter\gdef\csname R@mr.r2.samp.maxdev\endcsname{0.04}
\expandafter\gdef\csname R@mr.r2.samp.meandev\endcsname{0.014}
\expandafter\gdef\csname R@mr.r2.samp.n\endcsname{20}
\expandafter\gdef\csname R@mr.r2.samp.nflip\endcsname{0}
\expandafter\gdef\csname R@mr.r2.thr\endcsname{$1/3$}
\expandafter\gdef\csname R@mr.r2.thr.dec\endcsname{0.33}
\expandafter\gdef\csname R@mr.r2.thr.med\endcsname{0.31}
\expandafter\gdef\csname R@mr.r2test.V5B2.all\endcsname{75.13}
\expandafter\gdef\csname R@mr.r2test.V5B2.bear\endcsname{70.18}
\expandafter\gdef\csname R@mr.r2test.V5B2.free\endcsname{80.20}
\expandafter\gdef\csname R@mr.r2test.V5B2.pen\endcsname{\ensuremath{-10.02}}
\expandafter\gdef\csname R@mr.r2test.V7A1.all\endcsname{78.84}
\expandafter\gdef\csname R@mr.r2test.V7A1.bear\endcsname{74.89}
\expandafter\gdef\csname R@mr.r2test.V7A1.free\endcsname{82.89}
\expandafter\gdef\csname R@mr.r2test.V7A1.pen\endcsname{\ensuremath{-8.00}}
\expandafter\gdef\csname R@mr.r2test.base.all\endcsname{76.80}
\expandafter\gdef\csname R@mr.r2test.base.bear\endcsname{72.22}
\expandafter\gdef\csname R@mr.r2test.base.free\endcsname{81.50}
\expandafter\gdef\csname R@mr.r2test.base.pen\endcsname{\ensuremath{-9.28}}
\expandafter\gdef\csname R@mr.r2test.cystml.med\endcsname{4.4}
\expandafter\gdef\csname R@mr.r2test.deep.all\endcsname{\ensuremath{+3.71}}
\expandafter\gdef\csname R@mr.r2test.deep.bear\endcsname{\ensuremath{+4.71}}
\expandafter\gdef\csname R@mr.r2test.deep.bear.folds\endcsname{4}
\expandafter\gdef\csname R@mr.r2test.deep.bear.hi\endcsname{\ensuremath{+6.56}}
\expandafter\gdef\csname R@mr.r2test.deep.bear.lo\endcsname{\ensuremath{+2.85}}
\expandafter\gdef\csname R@mr.r2test.deep.did\endcsname{\ensuremath{+2.03}}
\expandafter\gdef\csname R@mr.r2test.deep.did.fb.hi\endcsname{\ensuremath{+4.49}}
\expandafter\gdef\csname R@mr.r2test.deep.did.fb.lo\endcsname{\ensuremath{-0.49}}
\expandafter\gdef\csname R@mr.r2test.deep.did.hi\endcsname{\ensuremath{+4.50}}
\expandafter\gdef\csname R@mr.r2test.deep.did.lo\endcsname{\ensuremath{-0.47}}
\expandafter\gdef\csname R@mr.r2test.deep.free\endcsname{\ensuremath{+2.69}}
\expandafter\gdef\csname R@mr.r2test.deep.free.folds\endcsname{4}
\expandafter\gdef\csname R@mr.r2test.deep.free.hi\endcsname{\ensuremath{+4.44}}
\expandafter\gdef\csname R@mr.r2test.deep.free.lo\endcsname{\ensuremath{+1.04}}
\expandafter\gdef\csname R@mr.r2test.n.all\endcsname{391}
\expandafter\gdef\csname R@mr.r2test.n.bear\endcsname{198}
\expandafter\gdef\csname R@mr.r2test.n.free\endcsname{193}
\expandafter\gdef\csname R@mr.r2test.vsbase.V5B2.all\endcsname{\ensuremath{-1.67}}
\expandafter\gdef\csname R@mr.r2test.vsbase.V5B2.bear\endcsname{\ensuremath{-2.04}}
\expandafter\gdef\csname R@mr.r2test.vsbase.V5B2.bear.folds\endcsname{1}
\expandafter\gdef\csname R@mr.r2test.vsbase.V5B2.bear.hi\endcsname{\ensuremath{-0.35}}
\expandafter\gdef\csname R@mr.r2test.vsbase.V5B2.bear.lo\endcsname{\ensuremath{-3.79}}
\expandafter\gdef\csname R@mr.r2test.vsbase.V5B2.did\endcsname{\ensuremath{-0.74}}
\expandafter\gdef\csname R@mr.r2test.vsbase.V5B2.did.fb.hi\endcsname{\ensuremath{+1.63}}
\expandafter\gdef\csname R@mr.r2test.vsbase.V5B2.did.fb.lo\endcsname{\ensuremath{-3.14}}
\expandafter\gdef\csname R@mr.r2test.vsbase.V5B2.did.hi\endcsname{\ensuremath{+1.66}}
\expandafter\gdef\csname R@mr.r2test.vsbase.V5B2.did.lo\endcsname{\ensuremath{-3.14}}
\expandafter\gdef\csname R@mr.r2test.vsbase.V5B2.free\endcsname{\ensuremath{-1.29}}
\expandafter\gdef\csname R@mr.r2test.vsbase.V5B2.free.folds\endcsname{2}
\expandafter\gdef\csname R@mr.r2test.vsbase.V5B2.free.hi\endcsname{\ensuremath{+0.35}}
\expandafter\gdef\csname R@mr.r2test.vsbase.V5B2.free.lo\endcsname{\ensuremath{-3.04}}
\expandafter\gdef\csname R@mr.r2test.vsbase.V7A1.all\endcsname{\ensuremath{+2.04}}
\expandafter\gdef\csname R@mr.r2test.vsbase.V7A1.bear\endcsname{\ensuremath{+2.67}}
\expandafter\gdef\csname R@mr.r2test.vsbase.V7A1.bear.folds\endcsname{3}
\expandafter\gdef\csname R@mr.r2test.vsbase.V7A1.bear.hi\endcsname{\ensuremath{+4.76}}
\expandafter\gdef\csname R@mr.r2test.vsbase.V7A1.bear.lo\endcsname{\ensuremath{+0.65}}
\expandafter\gdef\csname R@mr.r2test.vsbase.V7A1.did\endcsname{\ensuremath{+1.28}}
\expandafter\gdef\csname R@mr.r2test.vsbase.V7A1.did.fb.hi\endcsname{\ensuremath{+3.87}}
\expandafter\gdef\csname R@mr.r2test.vsbase.V7A1.did.fb.lo\endcsname{\ensuremath{-1.45}}
\expandafter\gdef\csname R@mr.r2test.vsbase.V7A1.did.hi\endcsname{\ensuremath{+3.87}}
\expandafter\gdef\csname R@mr.r2test.vsbase.V7A1.did.lo\endcsname{\ensuremath{-1.33}}
\expandafter\gdef\csname R@mr.r2test.vsbase.V7A1.free\endcsname{\ensuremath{+1.39}}
\expandafter\gdef\csname R@mr.r2test.vsbase.V7A1.free.folds\endcsname{3}
\expandafter\gdef\csname R@mr.r2test.vsbase.V7A1.free.hi\endcsname{\ensuremath{+3.11}}
\expandafter\gdef\csname R@mr.r2test.vsbase.V7A1.free.lo\endcsname{\ensuremath{-0.20}}
\expandafter\gdef\csname R@mr.r3.acdc.cover\endcsname{no}
\expandafter\gdef\csname R@mr.r3.acdc.cumstride\endcsname{$4{\times}32{\times}32$}
\expandafter\gdef\csname R@mr.r3.acdc.kmin\endcsname{15}
\expandafter\gdef\csname R@mr.r3.acdc.map\endcsname{$5{\times}8{\times}7$}
\expandafter\gdef\csname R@mr.r3.acdc.nblk\endcsname{3}
\expandafter\gdef\csname R@mr.r3.acdc.patch\endcsname{$20{\times}256{\times}224$}
\expandafter\gdef\csname R@mr.r3.amos.cover\endcsname{no}
\expandafter\gdef\csname R@mr.r3.amos.cumstride\endcsname{$16{\times}32{\times}32$}
\expandafter\gdef\csname R@mr.r3.amos.kmin\endcsname{11}
\expandafter\gdef\csname R@mr.r3.amos.map\endcsname{$4{\times}6{\times}5$}
\expandafter\gdef\csname R@mr.r3.amos.nblk\endcsname{2}
\expandafter\gdef\csname R@mr.r3.amos.patch\endcsname{$64{\times}192{\times}160$}
\expandafter\gdef\csname R@mr.r3.brats.cover\endcsname{yes}
\expandafter\gdef\csname R@mr.r3.brats.cumstride\endcsname{$32^3$}
\expandafter\gdef\csname R@mr.r3.brats.kmin\endcsname{7}
\expandafter\gdef\csname R@mr.r3.brats.map\endcsname{$4^3$}
\expandafter\gdef\csname R@mr.r3.brats.nblk\endcsname{1}
\expandafter\gdef\csname R@mr.r3.brats.patch\endcsname{$128^3$}
\expandafter\gdef\csname R@mr.r3.btcv.cover\endcsname{no}
\expandafter\gdef\csname R@mr.r3.btcv.cumstride\endcsname{$8{\times}32{\times}32$}
\expandafter\gdef\csname R@mr.r3.btcv.kmin\endcsname{11}
\expandafter\gdef\csname R@mr.r3.btcv.map\endcsname{$6^3$}
\expandafter\gdef\csname R@mr.r3.btcv.nblk\endcsname{2}
\expandafter\gdef\csname R@mr.r3.btcv.patch\endcsname{$48{\times}192{\times}192$}
\expandafter\gdef\csname R@mr.r3.kits.cover\endcsname{yes}
\expandafter\gdef\csname R@mr.r3.kits.cumstride\endcsname{$32^3$}
\expandafter\gdef\csname R@mr.r3.kits.kmin\endcsname{7}
\expandafter\gdef\csname R@mr.r3.kits.map\endcsname{$4^3$}
\expandafter\gdef\csname R@mr.r3.kits.nblk\endcsname{1}
\expandafter\gdef\csname R@mr.r3.kits.patch\endcsname{$128^3$}
\expandafter\gdef\csname R@mr.r3.lits.cover\endcsname{yes}
\expandafter\gdef\csname R@mr.r3.lits.cumstride\endcsname{$32^3$}
\expandafter\gdef\csname R@mr.r3.lits.kmin\endcsname{7}
\expandafter\gdef\csname R@mr.r3.lits.map\endcsname{$4^3$}
\expandafter\gdef\csname R@mr.r3.lits.nblk\endcsname{1}
\expandafter\gdef\csname R@mr.r3.lits.patch\endcsname{$128^3$}
\expandafter\gdef\csname R@mr.r3.ncover\endcsname{3}
\expandafter\gdef\csname R@mr.r3.rellat.coarse\endcsname{5.21}
\expandafter\gdef\csname R@mr.r3.rellat.dw3.coarse\endcsname{0.68}
\expandafter\gdef\csname R@mr.r3.rellat.dw3.fine\endcsname{1.70}
\expandafter\gdef\csname R@mr.r3.rellat.fine\endcsname{22.95}
\expandafter\gdef\csname R@mr.r3.relmac.coarse\endcsname{0.040}
\expandafter\gdef\csname R@mr.r3.relmac.e1\endcsname{0.20}
\expandafter\gdef\csname R@mr.r3.relmac.e4\endcsname{0.040}
\expandafter\gdef\csname R@mr.r3.relmac.fine\endcsname{0.40}
\expandafter\gdef\csname R@mr.r4.acdc.chg\endcsname{1}
\expandafter\gdef\csname R@mr.r4.acdc.chgtex\endcsname{E1/D3}
\expandafter\gdef\csname R@mr.r4.acdc.eq\endcsname{14.0}
\expandafter\gdef\csname R@mr.r4.acdc.fires\endcsname{yes}
\expandafter\gdef\csname R@mr.r4.acdc.fivemm\endcsname{14.0}
\expandafter\gdef\csname R@mr.r4.acdc.gt\endcsname{86.0}
\expandafter\gdef\csname R@mr.r4.acdc.ncases\endcsname{300}
\expandafter\gdef\csname R@mr.r4.acdc.planz1\endcsname{0}
\expandafter\gdef\csname R@mr.r4.acdc.ratio\endcsname{3.28}
\expandafter\gdef\csname R@mr.r4.acdc.rep\endcsname{86.0}
\expandafter\gdef\csname R@mr.r4.acdc.zmed\endcsname{10.0}
\expandafter\gdef\csname R@mr.r4.acdc.zone\endcsname{0, 1}
\expandafter\gdef\csname R@mr.r4.acdc.zplan\endcsname{5.0}
\expandafter\gdef\csname R@mr.r4.acdc.zpten\endcsname{5}
\expandafter\gdef\csname R@mr.r4.amos.chg\endcsname{1}
\expandafter\gdef\csname R@mr.r4.amos.chgtex\endcsname{E1/D3}
\expandafter\gdef\csname R@mr.r4.amos.eq\endcsname{22.3}
\expandafter\gdef\csname R@mr.r4.amos.fires\endcsname{yes}
\expandafter\gdef\csname R@mr.r4.amos.fivemm\endcsname{73.7}
\expandafter\gdef\csname R@mr.r4.amos.gt\endcsname{76.0}
\expandafter\gdef\csname R@mr.r4.amos.ncases\endcsname{300}
\expandafter\gdef\csname R@mr.r4.amos.planz1\endcsname{0}
\expandafter\gdef\csname R@mr.r4.amos.ratio\endcsname{3.65}
\expandafter\gdef\csname R@mr.r4.amos.rep\endcsname{76.0}
\expandafter\gdef\csname R@mr.r4.amos.zmed\endcsname{5.0}
\expandafter\gdef\csname R@mr.r4.amos.zone\endcsname{0, 1}
\expandafter\gdef\csname R@mr.r4.amos.zplan\endcsname{2.0}
\expandafter\gdef\csname R@mr.r4.amos.zpten\endcsname{2}
\expandafter\gdef\csname R@mr.r4.brats.chg\endcsname{none}
\expandafter\gdef\csname R@mr.r4.brats.chgtex\endcsname{--}
\expandafter\gdef\csname R@mr.r4.brats.eq\endcsname{100.0}
\expandafter\gdef\csname R@mr.r4.brats.fires\endcsname{no}
\expandafter\gdef\csname R@mr.r4.brats.fivemm\endcsname{0.0}
\expandafter\gdef\csname R@mr.r4.brats.gt\endcsname{0.0}
\expandafter\gdef\csname R@mr.r4.brats.ncases\endcsname{484}
\expandafter\gdef\csname R@mr.r4.brats.planz1\endcsname{none}
\expandafter\gdef\csname R@mr.r4.brats.rep\endcsname{0.0}
\expandafter\gdef\csname R@mr.r4.brats.zmed\endcsname{1.0}
\expandafter\gdef\csname R@mr.r4.brats.zone\endcsname{none}
\expandafter\gdef\csname R@mr.r4.brats.zplan\endcsname{1.0}
\expandafter\gdef\csname R@mr.r4.brats.zpten\endcsname{1}
\expandafter\gdef\csname R@mr.r4.btcv.chg\endcsname{none}
\expandafter\gdef\csname R@mr.r4.btcv.chgtex\endcsname{--}
\expandafter\gdef\csname R@mr.r4.btcv.eq\endcsname{56.7}
\expandafter\gdef\csname R@mr.r4.btcv.fires\endcsname{no}
\expandafter\gdef\csname R@mr.r4.btcv.fivemm\endcsname{40.0}
\expandafter\gdef\csname R@mr.r4.btcv.gt\endcsname{40.0}
\expandafter\gdef\csname R@mr.r4.btcv.ncases\endcsname{30}
\expandafter\gdef\csname R@mr.r4.btcv.planz1\endcsname{0}
\expandafter\gdef\csname R@mr.r4.btcv.rep\endcsname{40.0}
\expandafter\gdef\csname R@mr.r4.btcv.zmed\endcsname{3.0}
\expandafter\gdef\csname R@mr.r4.btcv.zone\endcsname{0}
\expandafter\gdef\csname R@mr.r4.btcv.zplan\endcsname{3.0}
\expandafter\gdef\csname R@mr.r4.btcv.zpten\endcsname{3}
\expandafter\gdef\csname R@mr.r4.kits.chg\endcsname{0}
\expandafter\gdef\csname R@mr.r4.kits.chgtex\endcsname{E0/D4}
\expandafter\gdef\csname R@mr.r4.kits.eq\endcsname{11.5}
\expandafter\gdef\csname R@mr.r4.kits.fires\endcsname{yes}
\expandafter\gdef\csname R@mr.r4.kits.fivemm\endcsname{45.6}
\expandafter\gdef\csname R@mr.r4.kits.gt\endcsname{79.1}
\expandafter\gdef\csname R@mr.r4.kits.ncases\endcsname{489}
\expandafter\gdef\csname R@mr.r4.kits.planz1\endcsname{none}
\expandafter\gdef\csname R@mr.r4.kits.ratio\endcsname{3.84}
\expandafter\gdef\csname R@mr.r4.kits.ratioone\endcsname{1.92}
\expandafter\gdef\csname R@mr.r4.kits.rep\endcsname{71.8}
\expandafter\gdef\csname R@mr.r4.kits.zmed\endcsname{3.0}
\expandafter\gdef\csname R@mr.r4.kits.zone\endcsname{0}
\expandafter\gdef\csname R@mr.r4.kits.zplan\endcsname{1.0}
\expandafter\gdef\csname R@mr.r4.kits.zpten\endcsname{1}
\expandafter\gdef\csname R@mr.r4.lits.chg\endcsname{none}
\expandafter\gdef\csname R@mr.r4.lits.chgtex\endcsname{--}
\expandafter\gdef\csname R@mr.r4.lits.eq\endcsname{25.2}
\expandafter\gdef\csname R@mr.r4.lits.fires\endcsname{no}
\expandafter\gdef\csname R@mr.r4.lits.fivemm\endcsname{6.9}
\expandafter\gdef\csname R@mr.r4.lits.gt\endcsname{35.9}
\expandafter\gdef\csname R@mr.r4.lits.ncases\endcsname{131}
\expandafter\gdef\csname R@mr.r4.lits.planz1\endcsname{none}
\expandafter\gdef\csname R@mr.r4.lits.rep\endcsname{16.8}
\expandafter\gdef\csname R@mr.r4.lits.zmed\endcsname{1.0}
\expandafter\gdef\csname R@mr.r4.lits.zone\endcsname{none}
\expandafter\gdef\csname R@mr.r4.lits.zplan\endcsname{1.0}
\expandafter\gdef\csname R@mr.r4.lits.zpten\endcsname{0.7}
\expandafter\gdef\csname R@mr.r4.nfires\endcsname{3}
\makeatother

\makeatletter
\expandafter\gdef\csname R@mn.amos.dbase\endcsname{\ensuremath{+0.54}}
\expandafter\gdef\csname R@mn.amos.f1\endcsname{89.57}
\expandafter\gdef\csname R@mn.amos.f2\endcsname{89.24}
\expandafter\gdef\csname R@mn.amos.f3\endcsname{88.94}
\expandafter\gdef\csname R@mn.amos.f4\endcsname{87.82}
\expandafter\gdef\csname R@mn.amos.mean\endcsname{88.89}
\expandafter\gdef\csname R@mn.amos.verdict\endcsname{below}
\expandafter\gdef\csname R@mn.gmac\endcsname{243.6}
\expandafter\gdef\csname R@mn.kits.dbase\endcsname{\ensuremath{-0.75}}
\expandafter\gdef\csname R@mn.kits.f1\endcsname{82.43}
\expandafter\gdef\csname R@mn.kits.f2\endcsname{81.96}
\expandafter\gdef\csname R@mn.kits.f3\endcsname{86.52}
\expandafter\gdef\csname R@mn.kits.f4\endcsname{82.05}
\expandafter\gdef\csname R@mn.kits.mean\endcsname{83.24}
\expandafter\gdef\csname R@mn.kits.verdict\endcsname{above}
\expandafter\gdef\csname R@mn.mem\endcsname{11.14}
\expandafter\gdef\csname R@mn.on\endcsname{1}
\expandafter\gdef\csname R@mn.par\endcsname{17.55}
\expandafter\gdef\csname R@mn.preset\endcsname{M}
\expandafter\gdef\csname R@mn.slowest\endcsname{1}
\expandafter\gdef\csname R@mn.step\endcsname{672.3}
\expandafter\gdef\csname R@mn.step.amos.r\endcsname{5.95}
\expandafter\gdef\csname R@mn.step.kits.r\endcsname{5.45}
\expandafter\gdef\csname R@mn.top.amos\endcsname{1}
\expandafter\gdef\csname R@mn.undercut.S\endcsname{1}
\makeatother

\input{an_extra_perclass_stock.tex}
\makeatletter
\expandafter\gdef\csname R@qualpct.A.nzero\endcsname{68}
\expandafter\gdef\csname R@qualpct.A.pool\endcsname{198}
\expandafter\gdef\csname R@qualpct.A.top10.share\endcsname{109}
\expandafter\gdef\csname R@qualpct.A10.case\endcsname{00094}
\expandafter\gdef\csname R@qualpct.A10.d\endcsname{-0.42}
\expandafter\gdef\csname R@qualpct.A10.dice.V7A1\endcsname{81.4}
\expandafter\gdef\csname R@qualpct.A10.dice.Y2\endcsname{79.7}
\expandafter\gdef\csname R@qualpct.A10.dice.base\endcsname{65.8}
\expandafter\gdef\csname R@qualpct.A10.fold\endcsname{3}
\expandafter\gdef\csname R@qualpct.A10.fp.V7A1\endcsname{2.0}
\expandafter\gdef\csname R@qualpct.A10.fp.Y2\endcsname{2.4}
\expandafter\gdef\csname R@qualpct.A10.fp.base\endcsname{7.2}
\expandafter\gdef\csname R@qualpct.A10.n\endcsname{198}
\expandafter\gdef\csname R@qualpct.A10.nslice\endcsname{43}
\expandafter\gdef\csname R@qualpct.A10.ntie\endcsname{1}
\expandafter\gdef\csname R@qualpct.A10.pct\endcsname{10}
\expandafter\gdef\csname R@qualpct.A10.rank\endcsname{20}
\expandafter\gdef\csname R@qualpct.A10.slice\endcsname{10}
\expandafter\gdef\csname R@qualpct.A50.case\endcsname{00002}
\expandafter\gdef\csname R@qualpct.A50.d\endcsname{0.00}
\expandafter\gdef\csname R@qualpct.A50.dice.V7A1\endcsname{95.8}
\expandafter\gdef\csname R@qualpct.A50.dice.Y2\endcsname{95.7}
\expandafter\gdef\csname R@qualpct.A50.dice.base\endcsname{95.8}
\expandafter\gdef\csname R@qualpct.A50.fold\endcsname{3}
\expandafter\gdef\csname R@qualpct.A50.fp.V7A1\endcsname{0.0}
\expandafter\gdef\csname R@qualpct.A50.fp.Y2\endcsname{0.0}
\expandafter\gdef\csname R@qualpct.A50.fp.base\endcsname{0.2}
\expandafter\gdef\csname R@qualpct.A50.n\endcsname{198}
\expandafter\gdef\csname R@qualpct.A50.nslice\endcsname{261}
\expandafter\gdef\csname R@qualpct.A50.ntie\endcsname{68}
\expandafter\gdef\csname R@qualpct.A50.pct\endcsname{50}
\expandafter\gdef\csname R@qualpct.A50.rank\endcsname{99}
\expandafter\gdef\csname R@qualpct.A50.slice\endcsname{195}
\expandafter\gdef\csname R@qualpct.A90.case\endcsname{00145}
\expandafter\gdef\csname R@qualpct.A90.d\endcsname{2.86}
\expandafter\gdef\csname R@qualpct.A90.dice.V7A1\endcsname{95.8}
\expandafter\gdef\csname R@qualpct.A90.dice.Y2\endcsname{96.3}
\expandafter\gdef\csname R@qualpct.A90.dice.base\endcsname{92.0}
\expandafter\gdef\csname R@qualpct.A90.fold\endcsname{2}
\expandafter\gdef\csname R@qualpct.A90.fp.V7A1\endcsname{3.8}
\expandafter\gdef\csname R@qualpct.A90.fp.Y2\endcsname{0.9}
\expandafter\gdef\csname R@qualpct.A90.fp.base\endcsname{13.7}
\expandafter\gdef\csname R@qualpct.A90.n\endcsname{198}
\expandafter\gdef\csname R@qualpct.A90.nslice\endcsname{86}
\expandafter\gdef\csname R@qualpct.A90.ntie\endcsname{1}
\expandafter\gdef\csname R@qualpct.A90.pct\endcsname{90}
\expandafter\gdef\csname R@qualpct.A90.rank\endcsname{179}
\expandafter\gdef\csname R@qualpct.A90.slice\endcsname{33}
\expandafter\gdef\csname R@qualpct.B.nzero\endcsname{0}
\expandafter\gdef\csname R@qualpct.B.pool\endcsname{81}
\expandafter\gdef\csname R@qualpct.B.top10.share\endcsname{97}
\expandafter\gdef\csname R@qualpct.B10.case\endcsname{00583}
\expandafter\gdef\csname R@qualpct.B10.d\endcsname{-1.64}
\expandafter\gdef\csname R@qualpct.B10.dice.V7A1\endcsname{89.7}
\expandafter\gdef\csname R@qualpct.B10.dice.Y2\endcsname{87.0}
\expandafter\gdef\csname R@qualpct.B10.dice.base\endcsname{90.1}
\expandafter\gdef\csname R@qualpct.B10.extr.V7A1\endcsname{0.0}
\expandafter\gdef\csname R@qualpct.B10.extr.Y2\endcsname{1.6}
\expandafter\gdef\csname R@qualpct.B10.extr.base\endcsname{0.0}
\expandafter\gdef\csname R@qualpct.B10.fold\endcsname{3}
\expandafter\gdef\csname R@qualpct.B10.n\endcsname{81}
\expandafter\gdef\csname R@qualpct.B10.nslice\endcsname{256}
\expandafter\gdef\csname R@qualpct.B10.ntie\endcsname{1}
\expandafter\gdef\csname R@qualpct.B10.pct\endcsname{10}
\expandafter\gdef\csname R@qualpct.B10.rank\endcsname{9}
\expandafter\gdef\csname R@qualpct.B10.slice\endcsname{107}
\expandafter\gdef\csname R@qualpct.B50.case\endcsname{00209}
\expandafter\gdef\csname R@qualpct.B50.d\endcsname{0.96}
\expandafter\gdef\csname R@qualpct.B50.dice.V7A1\endcsname{75.5}
\expandafter\gdef\csname R@qualpct.B50.dice.Y2\endcsname{88.4}
\expandafter\gdef\csname R@qualpct.B50.dice.base\endcsname{87.5}
\expandafter\gdef\csname R@qualpct.B50.extr.V7A1\endcsname{1.0}
\expandafter\gdef\csname R@qualpct.B50.extr.Y2\endcsname{0.0}
\expandafter\gdef\csname R@qualpct.B50.extr.base\endcsname{0.0}
\expandafter\gdef\csname R@qualpct.B50.fold\endcsname{2}
\expandafter\gdef\csname R@qualpct.B50.n\endcsname{81}
\expandafter\gdef\csname R@qualpct.B50.nslice\endcsname{102}
\expandafter\gdef\csname R@qualpct.B50.ntie\endcsname{1}
\expandafter\gdef\csname R@qualpct.B50.pct\endcsname{50}
\expandafter\gdef\csname R@qualpct.B50.rank\endcsname{41}
\expandafter\gdef\csname R@qualpct.B50.slice\endcsname{4}
\expandafter\gdef\csname R@qualpct.B90.case\endcsname{00192}
\expandafter\gdef\csname R@qualpct.B90.d\endcsname{60.40}
\expandafter\gdef\csname R@qualpct.B90.dice.V7A1\endcsname{91.1}
\expandafter\gdef\csname R@qualpct.B90.dice.Y2\endcsname{94.8}
\expandafter\gdef\csname R@qualpct.B90.dice.base\endcsname{93.7}
\expandafter\gdef\csname R@qualpct.B90.extr.V7A1\endcsname{60.4}
\expandafter\gdef\csname R@qualpct.B90.extr.Y2\endcsname{0.0}
\expandafter\gdef\csname R@qualpct.B90.extr.base\endcsname{18.4}
\expandafter\gdef\csname R@qualpct.B90.fold\endcsname{3}
\expandafter\gdef\csname R@qualpct.B90.n\endcsname{81}
\expandafter\gdef\csname R@qualpct.B90.nslice\endcsname{133}
\expandafter\gdef\csname R@qualpct.B90.ntie\endcsname{1}
\expandafter\gdef\csname R@qualpct.B90.pct\endcsname{90}
\expandafter\gdef\csname R@qualpct.B90.rank\endcsname{73}
\expandafter\gdef\csname R@qualpct.B90.slice\endcsname{132}
\expandafter\gdef\csname R@qualpct.C.excl\endcsname{0}
\expandafter\gdef\csname R@qualpct.C.nzero\endcsname{1}
\expandafter\gdef\csname R@qualpct.C.pool\endcsname{479}
\expandafter\gdef\csname R@qualpct.C.top10.share\endcsname{122}
\expandafter\gdef\csname R@qualpct.C10.case\endcsname{0067}
\expandafter\gdef\csname R@qualpct.C10.d\endcsname{-2.2}
\expandafter\gdef\csname R@qualpct.C10.dice.D2\endcsname{59.0}
\expandafter\gdef\csname R@qualpct.C10.dice.V7A1\endcsname{56.8}
\expandafter\gdef\csname R@qualpct.C10.dice.base\endcsname{59.8}
\expandafter\gdef\csname R@qualpct.C10.fold\endcsname{3}
\expandafter\gdef\csname R@qualpct.C10.n\endcsname{479}
\expandafter\gdef\csname R@qualpct.C10.nslice\endcsname{122}
\expandafter\gdef\csname R@qualpct.C10.ntie\endcsname{1}
\expandafter\gdef\csname R@qualpct.C10.organ\endcsname{oesophagus}
\expandafter\gdef\csname R@qualpct.C10.pct\endcsname{10}
\expandafter\gdef\csname R@qualpct.C10.rank\endcsname{48}
\expandafter\gdef\csname R@qualpct.C10.slice\endcsname{85}
\expandafter\gdef\csname R@qualpct.C50.case\endcsname{0259}
\expandafter\gdef\csname R@qualpct.C50.d\endcsname{0.3}
\expandafter\gdef\csname R@qualpct.C50.dice.D2\endcsname{91.2}
\expandafter\gdef\csname R@qualpct.C50.dice.V7A1\endcsname{91.5}
\expandafter\gdef\csname R@qualpct.C50.dice.base\endcsname{86.5}
\expandafter\gdef\csname R@qualpct.C50.fold\endcsname{3}
\expandafter\gdef\csname R@qualpct.C50.n\endcsname{479}
\expandafter\gdef\csname R@qualpct.C50.nslice\endcsname{78}
\expandafter\gdef\csname R@qualpct.C50.ntie\endcsname{1}
\expandafter\gdef\csname R@qualpct.C50.organ\endcsname{oesophagus}
\expandafter\gdef\csname R@qualpct.C50.pct\endcsname{50}
\expandafter\gdef\csname R@qualpct.C50.rank\endcsname{240}
\expandafter\gdef\csname R@qualpct.C50.slice\endcsname{75}
\expandafter\gdef\csname R@qualpct.C90.case\endcsname{0049}
\expandafter\gdef\csname R@qualpct.C90.d\endcsname{3.3}
\expandafter\gdef\csname R@qualpct.C90.dice.D2\endcsname{81.3}
\expandafter\gdef\csname R@qualpct.C90.dice.V7A1\endcsname{84.6}
\expandafter\gdef\csname R@qualpct.C90.dice.base\endcsname{84.4}
\expandafter\gdef\csname R@qualpct.C90.fold\endcsname{1}
\expandafter\gdef\csname R@qualpct.C90.n\endcsname{479}
\expandafter\gdef\csname R@qualpct.C90.nslice\endcsname{96}
\expandafter\gdef\csname R@qualpct.C90.ntie\endcsname{1}
\expandafter\gdef\csname R@qualpct.C90.organ\endcsname{oesophagus}
\expandafter\gdef\csname R@qualpct.C90.pct\endcsname{90}
\expandafter\gdef\csname R@qualpct.C90.rank\endcsname{432}
\expandafter\gdef\csname R@qualpct.C90.slice\endcsname{85}
\expandafter\gdef\csname R@qualpct.gap.dice\endcsname{0.001}
\expandafter\gdef\csname R@qualpct.gap.dice.max\endcsname{0.00033}
\expandafter\gdef\csname R@qualpct.gap.ml\endcsname{0.01}
\expandafter\gdef\csname R@qualpct.gap.ml.max\endcsname{0.0061}
\expandafter\gdef\csname R@qualpct.gap.nprint\endcsname{1}
\expandafter\gdef\csname R@qualpct.on\endcsname{1}
\makeatother

\input{an_extra_qual.tex}
\makeatletter
\expandafter\gdef\csname R@r17.amos.M.d\endcsname{\ensuremath{-0.07}}
\expandafter\gdef\csname R@r17.amos.M.hi\endcsname{\ensuremath{+0.17}}
\expandafter\gdef\csname R@r17.amos.M.lo\endcsname{\ensuremath{-0.33}}
\expandafter\gdef\csname R@r17.amos.M.nfold\endcsname{4}
\expandafter\gdef\csname R@r17.amos.M.nup\endcsname{1}
\expandafter\gdef\csname R@r17.amos.M.verdict\endcsname{FAILS (bar not measured)}
\expandafter\gdef\csname R@r17.amos.base.d\endcsname{\ensuremath{+0.19}}
\expandafter\gdef\csname R@r17.amos.base.hi\endcsname{\ensuremath{+0.46}}
\expandafter\gdef\csname R@r17.amos.base.lo\endcsname{\ensuremath{-0.07}}
\expandafter\gdef\csname R@r17.amos.base.nfold\endcsname{4}
\expandafter\gdef\csname R@r17.amos.base.nup\endcsname{3}
\expandafter\gdef\csname R@r17.amos.base.verdict\endcsname{FAILS (bar not measured)}
\expandafter\gdef\csname R@r17.colon.M.d\endcsname{\ensuremath{-0.06}}
\expandafter\gdef\csname R@r17.colon.M.hi\endcsname{\ensuremath{+3.71}}
\expandafter\gdef\csname R@r17.colon.M.lo\endcsname{\ensuremath{-3.54}}
\expandafter\gdef\csname R@r17.colon.M.nfold\endcsname{4}
\expandafter\gdef\csname R@r17.colon.M.nup\endcsname{1}
\expandafter\gdef\csname R@r17.colon.M.verdict\endcsname{UNDERPOWERED}
\expandafter\gdef\csname R@r17.colon.base.d\endcsname{\ensuremath{-4.01}}
\expandafter\gdef\csname R@r17.colon.base.hi\endcsname{\ensuremath{-0.65}}
\expandafter\gdef\csname R@r17.colon.base.lo\endcsname{\ensuremath{-7.65}}
\expandafter\gdef\csname R@r17.colon.base.nfold\endcsname{4}
\expandafter\gdef\csname R@r17.colon.base.nup\endcsname{0}
\expandafter\gdef\csname R@r17.colon.base.verdict\endcsname{REVERSED}
\expandafter\gdef\csname R@r17.cov.acdc.nstage\endcsname{1}
\expandafter\gdef\csname R@r17.cov.acdc.sel\endcsname{E5}
\expandafter\gdef\csname R@r17.cov.amos.nstage\endcsname{1}
\expandafter\gdef\csname R@r17.cov.amos.sel\endcsname{E5}
\expandafter\gdef\csname R@r17.cov.brats.nstage\endcsname{2}
\expandafter\gdef\csname R@r17.cov.brats.sel\endcsname{E4--E5}
\expandafter\gdef\csname R@r17.cov.btcv.nstage\endcsname{1}
\expandafter\gdef\csname R@r17.cov.btcv.sel\endcsname{E5}
\expandafter\gdef\csname R@r17.cov.colon.nstage\endcsname{1}
\expandafter\gdef\csname R@r17.cov.colon.sel\endcsname{E5}
\expandafter\gdef\csname R@r17.cov.identity\endcsname{1}
\expandafter\gdef\csname R@r17.cov.kits.nstage\endcsname{2}
\expandafter\gdef\csname R@r17.cov.kits.sel\endcsname{E4--E5}
\expandafter\gdef\csname R@r17.cov.lits.nstage\endcsname{2}
\expandafter\gdef\csname R@r17.cov.lits.sel\endcsname{E4--E5}
\expandafter\gdef\csname R@r17.cov.lung.nstage\endcsname{1}
\expandafter\gdef\csname R@r17.cov.lung.sel\endcsname{E5}
\expandafter\gdef\csname R@r17.cov.maxaxis\endcsname{8}
\expandafter\gdef\csname R@r17.cov.panc.nstage\endcsname{1}
\expandafter\gdef\csname R@r17.cov.panc.sel\endcsname{E5}
\expandafter\gdef\csname R@r17.cov.six.one\endcsname{3}
\expandafter\gdef\csname R@r17.cov.six.two\endcsname{3}
\expandafter\gdef\csname R@r17.landed\endcsname{1}
\expandafter\gdef\csname R@r17.lung.M.d\endcsname{\ensuremath{-0.68}}
\expandafter\gdef\csname R@r17.lung.M.hi\endcsname{\ensuremath{+3.71}}
\expandafter\gdef\csname R@r17.lung.M.lo\endcsname{\ensuremath{-5.24}}
\expandafter\gdef\csname R@r17.lung.M.nfold\endcsname{4}
\expandafter\gdef\csname R@r17.lung.M.nup\endcsname{2}
\expandafter\gdef\csname R@r17.lung.M.verdict\endcsname{FAILS (bar not measured)}
\expandafter\gdef\csname R@r17.lung.base.d\endcsname{\ensuremath{-1.86}}
\expandafter\gdef\csname R@r17.lung.base.hi\endcsname{\ensuremath{+1.42}}
\expandafter\gdef\csname R@r17.lung.base.lo\endcsname{\ensuremath{-5.64}}
\expandafter\gdef\csname R@r17.lung.base.nfold\endcsname{4}
\expandafter\gdef\csname R@r17.lung.base.nup\endcsname{3}
\expandafter\gdef\csname R@r17.lung.base.verdict\endcsname{FAILS (bar not measured)}
\expandafter\gdef\csname R@r17.panc.M.d\endcsname{\ensuremath{-0.50}}
\expandafter\gdef\csname R@r17.panc.M.hi\endcsname{\ensuremath{+2.23}}
\expandafter\gdef\csname R@r17.panc.M.lo\endcsname{\ensuremath{-3.26}}
\expandafter\gdef\csname R@r17.panc.M.nfold\endcsname{4}
\expandafter\gdef\csname R@r17.panc.M.nup\endcsname{2}
\expandafter\gdef\csname R@r17.panc.M.verdict\endcsname{FAILS (bar not measured)}
\expandafter\gdef\csname R@r17.panc.base.d\endcsname{\ensuremath{+2.05}}
\expandafter\gdef\csname R@r17.panc.base.hi\endcsname{\ensuremath{+5.41}}
\expandafter\gdef\csname R@r17.panc.base.lo\endcsname{\ensuremath{-1.28}}
\expandafter\gdef\csname R@r17.panc.base.nfold\endcsname{4}
\expandafter\gdef\csname R@r17.panc.base.nup\endcsname{3}
\expandafter\gdef\csname R@r17.panc.base.verdict\endcsname{FAILS (bar not measured)}
\makeatother

\makeatletter
\expandafter\gdef\csname R@r18.acdc.M.mean\endcsname{91.58}
\expandafter\gdef\csname R@r18.acdc.base.mean\endcsname{91.37}
\expandafter\gdef\csname R@r18.acdc.d\endcsname{\ensuremath{+0.22}}
\expandafter\gdef\csname R@r18.acdc.hi\endcsname{\ensuremath{+0.42}}
\expandafter\gdef\csname R@r18.acdc.lo\endcsname{\ensuremath{+0.02}}
\expandafter\gdef\csname R@r18.acdc.n\endcsname{100}
\expandafter\gdef\csname R@r18.acdc.ntrain\endcsname{200}
\expandafter\gdef\csname R@r18.acdc.nup\endcsname{3}
\expandafter\gdef\csname R@r18.acdc.s0.d\endcsname{\ensuremath{+0.03}}
\expandafter\gdef\csname R@r18.acdc.s1.d\endcsname{\ensuremath{+0.33}}
\expandafter\gdef\csname R@r18.acdc.s2.d\endcsname{\ensuremath{+0.29}}
\expandafter\gdef\csname R@r18.amos.M.mean\endcsname{89.91}
\expandafter\gdef\csname R@r18.amos.base.mean\endcsname{89.74}
\expandafter\gdef\csname R@r18.amos.d\endcsname{\ensuremath{+0.17}}
\expandafter\gdef\csname R@r18.amos.hi\endcsname{\ensuremath{+0.31}}
\expandafter\gdef\csname R@r18.amos.lo\endcsname{\ensuremath{+0.03}}
\expandafter\gdef\csname R@r18.amos.n\endcsname{100}
\expandafter\gdef\csname R@r18.amos.ntrain\endcsname{200}
\expandafter\gdef\csname R@r18.amos.nup\endcsname{3}
\expandafter\gdef\csname R@r18.amos.s0.d\endcsname{\ensuremath{+0.08}}
\expandafter\gdef\csname R@r18.amos.s1.d\endcsname{\ensuremath{+0.33}}
\expandafter\gdef\csname R@r18.amos.s2.d\endcsname{\ensuremath{+0.09}}
\expandafter\gdef\csname R@r18.n\endcsname{100}
\expandafter\gdef\csname R@r18.nseed\endcsname{3}
\expandafter\gdef\csname R@r18.nseed.word\endcsname{three}
\expandafter\gdef\csname R@r18.ntrain\endcsname{200}
\expandafter\gdef\csname R@r18.on\endcsname{1}
\makeatother

\makeatletter
\expandafter\gdef\csname R@r19.acdc.protocol\endcsname{mirror off}
\expandafter\gdef\csname R@r19.bar.acdc\endcsname{0.20}
\expandafter\gdef\csname R@r19.bar.acdc.tta\endcsname{0.17}
\expandafter\gdef\csname R@r19.bar.amos\endcsname{0.34}
\expandafter\gdef\csname R@r19.bar.kits\endcsname{1.08}
\expandafter\gdef\csname R@r19.cost.V9R2.gmac\endcsname{0.78}
\expandafter\gdef\csname R@r19.cost.V9R2.params\endcsname{0.99}
\expandafter\gdef\csname R@r19.cost.V9R2.step\endcsname{0.81}
\expandafter\gdef\csname R@r19.on\endcsname{1}
\expandafter\gdef\csname R@r19.r2.acdc.boot\endcsname{\ensuremath{+0.25}}
\expandafter\gdef\csname R@r19.r2.acdc.d\endcsname{\ensuremath{+0.25}}
\expandafter\gdef\csname R@r19.r2.acdc.dabs\endcsname{0.25}
\expandafter\gdef\csname R@r19.r2.acdc.hi\endcsname{\ensuremath{+0.44}}
\expandafter\gdef\csname R@r19.r2.acdc.lo\endcsname{\ensuremath{+0.06}}
\expandafter\gdef\csname R@r19.r2.acdc.ncase\endcsname{160}
\expandafter\gdef\csname R@r19.r2.acdc.ndn\endcsname{0}
\expandafter\gdef\csname R@r19.r2.acdc.nfold\endcsname{4}
\expandafter\gdef\csname R@r19.r2.acdc.nup\endcsname{4}
\expandafter\gdef\csname R@r19.r2.acdc.tta.d\endcsname{\ensuremath{+0.24}}
\expandafter\gdef\csname R@r19.r2.acdc.tta.verdict\endcsname{CONFIRMED}
\expandafter\gdef\csname R@r19.r2.acdc.verdict\endcsname{CONFIRMED}
\expandafter\gdef\csname R@r19.r2.amos.boot\endcsname{\ensuremath{+0.55}}
\expandafter\gdef\csname R@r19.r2.amos.d\endcsname{\ensuremath{+0.62}}
\expandafter\gdef\csname R@r19.r2.amos.dabs\endcsname{0.62}
\expandafter\gdef\csname R@r19.r2.amos.hi\endcsname{\ensuremath{+0.69}}
\expandafter\gdef\csname R@r19.r2.amos.lo\endcsname{\ensuremath{+0.41}}
\expandafter\gdef\csname R@r19.r2.amos.ncase\endcsname{160}
\expandafter\gdef\csname R@r19.r2.amos.ndn\endcsname{0}
\expandafter\gdef\csname R@r19.r2.amos.nfold\endcsname{4}
\expandafter\gdef\csname R@r19.r2.amos.nup\endcsname{4}
\expandafter\gdef\csname R@r19.r2.amos.verdict\endcsname{CONFIRMED}
\expandafter\gdef\csname R@r19.r2.both.allup\endcsname{1}
\expandafter\gdef\csname R@r19.r2.kits.boot\endcsname{\ensuremath{+0.34}}
\expandafter\gdef\csname R@r19.r2.kits.d\endcsname{\ensuremath{+0.34}}
\expandafter\gdef\csname R@r19.r2.kits.dabs\endcsname{0.34}
\expandafter\gdef\csname R@r19.r2.kits.hi\endcsname{\ensuremath{+1.03}}
\expandafter\gdef\csname R@r19.r2.kits.lo\endcsname{\ensuremath{-0.36}}
\expandafter\gdef\csname R@r19.r2.kits.ncase\endcsname{391}
\expandafter\gdef\csname R@r19.r2.kits.ndn\endcsname{2}
\expandafter\gdef\csname R@r19.r2.kits.nfold\endcsname{4}
\expandafter\gdef\csname R@r19.r2.kits.nup\endcsname{2}
\expandafter\gdef\csname R@r19.r2.kits.verdict\endcsname{UNDERPOWERED}
\expandafter\gdef\csname R@r19.r2.nconf\endcsname{2}
\expandafter\gdef\csname R@r19.r3inc.acdc.boot\endcsname{\ensuremath{+0.05}}
\expandafter\gdef\csname R@r19.r3inc.acdc.d\endcsname{\ensuremath{+0.05}}
\expandafter\gdef\csname R@r19.r3inc.acdc.dabs\endcsname{0.05}
\expandafter\gdef\csname R@r19.r3inc.acdc.hi\endcsname{\ensuremath{+0.20}}
\expandafter\gdef\csname R@r19.r3inc.acdc.lo\endcsname{\ensuremath{-0.11}}
\expandafter\gdef\csname R@r19.r3inc.acdc.ncase\endcsname{160}
\expandafter\gdef\csname R@r19.r3inc.acdc.ndn\endcsname{2}
\expandafter\gdef\csname R@r19.r3inc.acdc.nfold\endcsname{4}
\expandafter\gdef\csname R@r19.r3inc.acdc.nup\endcsname{2}
\expandafter\gdef\csname R@r19.r3inc.acdc.verdict\endcsname{UNDERPOWERED}
\expandafter\gdef\csname R@r19.r3inc.amos.boot\endcsname{\ensuremath{+0.03}}
\expandafter\gdef\csname R@r19.r3inc.amos.d\endcsname{\ensuremath{+0.10}}
\expandafter\gdef\csname R@r19.r3inc.amos.dabs\endcsname{0.10}
\expandafter\gdef\csname R@r19.r3inc.amos.hi\endcsname{\ensuremath{+0.18}}
\expandafter\gdef\csname R@r19.r3inc.amos.lo\endcsname{\ensuremath{-0.13}}
\expandafter\gdef\csname R@r19.r3inc.amos.ncase\endcsname{160}
\expandafter\gdef\csname R@r19.r3inc.amos.ndn\endcsname{2}
\expandafter\gdef\csname R@r19.r3inc.amos.nfold\endcsname{4}
\expandafter\gdef\csname R@r19.r3inc.amos.nup\endcsname{2}
\expandafter\gdef\csname R@r19.r3inc.amos.verdict\endcsname{UNDERPOWERED}
\expandafter\gdef\csname R@r19.r3inc.kits.boot\endcsname{\ensuremath{+0.30}}
\expandafter\gdef\csname R@r19.r3inc.kits.d\endcsname{\ensuremath{+0.30}}
\expandafter\gdef\csname R@r19.r3inc.kits.dabs\endcsname{0.30}
\expandafter\gdef\csname R@r19.r3inc.kits.hi\endcsname{\ensuremath{+0.97}}
\expandafter\gdef\csname R@r19.r3inc.kits.lo\endcsname{\ensuremath{-0.36}}
\expandafter\gdef\csname R@r19.r3inc.kits.ncase\endcsname{391}
\expandafter\gdef\csname R@r19.r3inc.kits.ndn\endcsname{1}
\expandafter\gdef\csname R@r19.r3inc.kits.nfold\endcsname{4}
\expandafter\gdef\csname R@r19.r3inc.kits.nup\endcsname{3}
\expandafter\gdef\csname R@r19.r3inc.kits.verdict\endcsname{UNDERPOWERED}
\expandafter\gdef\csname R@r19.r3off.acdc.boot\endcsname{\ensuremath{-0.05}}
\expandafter\gdef\csname R@r19.r3off.acdc.d\endcsname{\ensuremath{-0.05}}
\expandafter\gdef\csname R@r19.r3off.acdc.dabs\endcsname{0.05}
\expandafter\gdef\csname R@r19.r3off.acdc.hi\endcsname{\ensuremath{+0.11}}
\expandafter\gdef\csname R@r19.r3off.acdc.lo\endcsname{\ensuremath{-0.20}}
\expandafter\gdef\csname R@r19.r3off.acdc.ncase\endcsname{160}
\expandafter\gdef\csname R@r19.r3off.acdc.ndn\endcsname{2}
\expandafter\gdef\csname R@r19.r3off.acdc.nfold\endcsname{4}
\expandafter\gdef\csname R@r19.r3off.acdc.nup\endcsname{2}
\expandafter\gdef\csname R@r19.r3off.acdc.tta.d\endcsname{\ensuremath{+0.04}}
\expandafter\gdef\csname R@r19.r3off.acdc.tta.verdict\endcsname{UNDERPOWERED}
\expandafter\gdef\csname R@r19.r3off.acdc.verdict\endcsname{UNDERPOWERED}
\expandafter\gdef\csname R@r19.r3off.amos.boot\endcsname{\ensuremath{-0.03}}
\expandafter\gdef\csname R@r19.r3off.amos.d\endcsname{\ensuremath{-0.10}}
\expandafter\gdef\csname R@r19.r3off.amos.dabs\endcsname{0.10}
\expandafter\gdef\csname R@r19.r3off.amos.hi\endcsname{\ensuremath{+0.13}}
\expandafter\gdef\csname R@r19.r3off.amos.lo\endcsname{\ensuremath{-0.18}}
\expandafter\gdef\csname R@r19.r3off.amos.ncase\endcsname{160}
\expandafter\gdef\csname R@r19.r3off.amos.ndn\endcsname{2}
\expandafter\gdef\csname R@r19.r3off.amos.nfold\endcsname{4}
\expandafter\gdef\csname R@r19.r3off.amos.nup\endcsname{2}
\expandafter\gdef\csname R@r19.r3off.amos.verdict\endcsname{UNDERPOWERED}
\expandafter\gdef\csname R@r19.r3off.kits.boot\endcsname{\ensuremath{-0.30}}
\expandafter\gdef\csname R@r19.r3off.kits.d\endcsname{\ensuremath{-0.30}}
\expandafter\gdef\csname R@r19.r3off.kits.dabs\endcsname{0.30}
\expandafter\gdef\csname R@r19.r3off.kits.hi\endcsname{\ensuremath{+0.36}}
\expandafter\gdef\csname R@r19.r3off.kits.lo\endcsname{\ensuremath{-0.97}}
\expandafter\gdef\csname R@r19.r3off.kits.ncase\endcsname{391}
\expandafter\gdef\csname R@r19.r3off.kits.ndn\endcsname{3}
\expandafter\gdef\csname R@r19.r3off.kits.nfold\endcsname{4}
\expandafter\gdef\csname R@r19.r3off.kits.nup\endcsname{1}
\expandafter\gdef\csname R@r19.r3off.kits.verdict\endcsname{UNDERPOWERED}
\makeatother

\makeatletter
\expandafter\gdef\csname R@r20.M.ds.absmax\endcsname{0.59}
\expandafter\gdef\csname R@r20.M.f1.ds\endcsname{\ensuremath{-0.12}}
\expandafter\gdef\csname R@r20.M.f1.m2\endcsname{84.84}
\expandafter\gdef\csname R@r20.M.f1.s1\endcsname{84.78}
\expandafter\gdef\csname R@r20.M.f2.ds\endcsname{\ensuremath{+0.11}}
\expandafter\gdef\csname R@r20.M.f2.m2\endcsname{85.77}
\expandafter\gdef\csname R@r20.M.f2.s1\endcsname{85.83}
\expandafter\gdef\csname R@r20.M.f3.ds\endcsname{\ensuremath{-0.59}}
\expandafter\gdef\csname R@r20.M.f3.m2\endcsname{88.59}
\expandafter\gdef\csname R@r20.M.f3.s1\endcsname{88.30}
\expandafter\gdef\csname R@r20.M.f4.ds\endcsname{\ensuremath{+0.08}}
\expandafter\gdef\csname R@r20.M.f4.m2\endcsname{83.05}
\expandafter\gdef\csname R@r20.M.f4.s1\endcsname{83.09}
\expandafter\gdef\csname R@r20.M.mean2\endcsname{85.56}
\expandafter\gdef\csname R@r20.M.s1.mean\endcsname{85.50}
\expandafter\gdef\csname R@r20.base.ds.absmax\endcsname{1.56}
\expandafter\gdef\csname R@r20.base.f1.ds\endcsname{\ensuremath{+1.56}}
\expandafter\gdef\csname R@r20.base.f1.m2\endcsname{83.64}
\expandafter\gdef\csname R@r20.base.f1.s1\endcsname{84.42}
\expandafter\gdef\csname R@r20.base.f2.ds\endcsname{\ensuremath{-1.03}}
\expandafter\gdef\csname R@r20.base.f2.m2\endcsname{82.80}
\expandafter\gdef\csname R@r20.base.f2.s1\endcsname{82.28}
\expandafter\gdef\csname R@r20.base.f3.ds\endcsname{\ensuremath{+0.35}}
\expandafter\gdef\csname R@r20.base.f3.m2\endcsname{87.09}
\expandafter\gdef\csname R@r20.base.f3.s1\endcsname{87.26}
\expandafter\gdef\csname R@r20.base.f4.ds\endcsname{\ensuremath{-1.40}}
\expandafter\gdef\csname R@r20.base.f4.m2\endcsname{82.19}
\expandafter\gdef\csname R@r20.base.f4.s1\endcsname{81.50}
\expandafter\gdef\csname R@r20.base.mean2\endcsname{83.93}
\expandafter\gdef\csname R@r20.base.s1.mean\endcsname{83.86}
\expandafter\gdef\csname R@r20.boot2.d\endcsname{\ensuremath{+1.64}}
\expandafter\gdef\csname R@r20.boot2.hi\endcsname{\ensuremath{+2.28}}
\expandafter\gdef\csname R@r20.boot2.lo\endcsname{\ensuremath{+1.04}}
\expandafter\gdef\csname R@r20.boot2.n\endcsname{391}
\expandafter\gdef\csname R@r20.cons.hi\endcsname{\ensuremath{+2.66}}
\expandafter\gdef\csname R@r20.cons.lo\endcsname{\ensuremath{+0.61}}
\expandafter\gdef\csname R@r20.d1\endcsname{\ensuremath{+1.63}}
\expandafter\gdef\csname R@r20.d1.hi\endcsname{\ensuremath{+3.27}}
\expandafter\gdef\csname R@r20.d1.lo\endcsname{\ensuremath{0.00}}
\expandafter\gdef\csname R@r20.d2\endcsname{\ensuremath{+1.63}}
\expandafter\gdef\csname R@r20.d2.excl\endcsname{1}
\expandafter\gdef\csname R@r20.d2.hi\endcsname{\ensuremath{+3.12}}
\expandafter\gdef\csname R@r20.d2.lo\endcsname{\ensuremath{+0.15}}
\expandafter\gdef\csname R@r20.d2.nup\endcsname{4}
\expandafter\gdef\csname R@r20.ds1\endcsname{\ensuremath{+1.63}}
\expandafter\gdef\csname R@r20.ds1.nup\endcsname{4}
\expandafter\gdef\csname R@r20.nseed\endcsname{2}
\expandafter\gdef\csname R@r20.on\endcsname{1}
\expandafter\gdef\csname R@r20.sd.df\endcsname{8}
\expandafter\gdef\csname R@r20.sd.res\endcsname{1.22}
\expandafter\gdef\csname R@r20.sd.run\endcsname{0.61}
\makeatother

\makeatletter
\expandafter\gdef\csname R@r24.D.sign.n\endcsname{6}
\expandafter\gdef\csname R@r24.D.sign.n.word\endcsname{six}
\expandafter\gdef\csname R@r24.D.sign.nneg\endcsname{5}
\expandafter\gdef\csname R@r24.D.sign.npos\endcsname{1}
\expandafter\gdef\csname R@r24.D.sign.p\endcsname{0.984}
\expandafter\gdef\csname R@r24.D.sign.pneg\endcsname{0.11}
\expandafter\gdef\csname R@r24.L1.params.acdc\endcsname{3.34}
\expandafter\gdef\csname R@r24.L1.params.amos\endcsname{3.29}
\expandafter\gdef\csname R@r24.L1.params.r1\endcsname{3.3}
\expandafter\gdef\csname R@r24.acdc.D.est\endcsname{\ensuremath{-0.31}}
\expandafter\gdef\csname R@r24.acdc.D.hi\endcsname{\ensuremath{-0.15}}
\expandafter\gdef\csname R@r24.acdc.D.lo\endcsname{\ensuremath{-0.48}}
\expandafter\gdef\csname R@r24.acdc.D.nup\endcsname{0}
\expandafter\gdef\csname R@r24.acdc.D.s0\endcsname{\ensuremath{-0.404}}
\expandafter\gdef\csname R@r24.acdc.D.s1\endcsname{\ensuremath{-0.238}}
\expandafter\gdef\csname R@r24.acdc.D.s2\endcsname{\ensuremath{-0.297}}
\expandafter\gdef\csname R@r24.acdc.D.verdict\endcsname{REVERSED}
\expandafter\gdef\csname R@r24.acdc.L1.mean\endcsname{91.90}
\expandafter\gdef\csname R@r24.acdc.L1.s0\endcsname{91.90}
\expandafter\gdef\csname R@r24.acdc.L1.s0.d\endcsname{\ensuremath{+0.43}}
\expandafter\gdef\csname R@r24.acdc.L1.s1\endcsname{91.91}
\expandafter\gdef\csname R@r24.acdc.L1.s1.d\endcsname{\ensuremath{+0.56}}
\expandafter\gdef\csname R@r24.acdc.L1.s2\endcsname{91.87}
\expandafter\gdef\csname R@r24.acdc.L1.s2.d\endcsname{\ensuremath{+0.59}}
\expandafter\gdef\csname R@r24.acdc.X\endcsname{41}
\expandafter\gdef\csname R@r24.acdc.X.hi\endcsname{71}
\expandafter\gdef\csname R@r24.acdc.X.lo\endcsname{0}
\expandafter\gdef\csname R@r24.acdc.X.within\endcsname{0}
\expandafter\gdef\csname R@r24.acdc.d\endcsname{\ensuremath{+0.53}}
\expandafter\gdef\csname R@r24.acdc.gl.est\endcsname{\ensuremath{+0.53}}
\expandafter\gdef\csname R@r24.acdc.gl.hi\endcsname{\ensuremath{+0.71}}
\expandafter\gdef\csname R@r24.acdc.gl.lo\endcsname{\ensuremath{+0.37}}
\expandafter\gdef\csname R@r24.acdc.gl.nup\endcsname{3}
\expandafter\gdef\csname R@r24.acdc.gl.s0\endcsname{\ensuremath{+0.432}}
\expandafter\gdef\csname R@r24.acdc.gl.s1\endcsname{\ensuremath{+0.565}}
\expandafter\gdef\csname R@r24.acdc.gl.s2\endcsname{\ensuremath{+0.590}}
\expandafter\gdef\csname R@r24.acdc.gl.verdict\endcsname{CONFIRMED}
\expandafter\gdef\csname R@r24.acdc.gm.est\endcsname{\ensuremath{+0.22}}
\expandafter\gdef\csname R@r24.acdc.gm.hi\endcsname{\ensuremath{+0.43}}
\expandafter\gdef\csname R@r24.acdc.gm.lo\endcsname{\ensuremath{0.00}}
\expandafter\gdef\csname R@r24.acdc.gm.nup\endcsname{3}
\expandafter\gdef\csname R@r24.acdc.gm.s0\endcsname{\ensuremath{+0.028}}
\expandafter\gdef\csname R@r24.acdc.gm.s1\endcsname{\ensuremath{+0.326}}
\expandafter\gdef\csname R@r24.acdc.gm.s2\endcsname{\ensuremath{+0.293}}
\expandafter\gdef\csname R@r24.acdc.gm.verdict\endcsname{CONFIRMED}
\expandafter\gdef\csname R@r24.acdc.row\endcsname{4}
\expandafter\gdef\csname R@r24.amos.D.est\endcsname{\ensuremath{-0.11}}
\expandafter\gdef\csname R@r24.amos.D.hi\endcsname{\ensuremath{+0.03}}
\expandafter\gdef\csname R@r24.amos.D.lo\endcsname{\ensuremath{-0.25}}
\expandafter\gdef\csname R@r24.amos.D.nup\endcsname{1}
\expandafter\gdef\csname R@r24.amos.D.s0\endcsname{\ensuremath{-0.133}}
\expandafter\gdef\csname R@r24.amos.D.s1\endcsname{\ensuremath{+0.038}}
\expandafter\gdef\csname R@r24.amos.D.s2\endcsname{\ensuremath{-0.248}}
\expandafter\gdef\csname R@r24.amos.D.verdict\endcsname{NOT CONFIRMED}
\expandafter\gdef\csname R@r24.amos.L1.mean\endcsname{90.02}
\expandafter\gdef\csname R@r24.amos.L1.s0\endcsname{90.01}
\expandafter\gdef\csname R@r24.amos.L1.s0.d\endcsname{\ensuremath{+0.21}}
\expandafter\gdef\csname R@r24.amos.L1.s1\endcsname{90.02}
\expandafter\gdef\csname R@r24.amos.L1.s1.d\endcsname{\ensuremath{+0.30}}
\expandafter\gdef\csname R@r24.amos.L1.s2\endcsname{90.04}
\expandafter\gdef\csname R@r24.amos.L1.s2.d\endcsname{\ensuremath{+0.34}}
\expandafter\gdef\csname R@r24.amos.X\endcsname{59}
\expandafter\gdef\csname R@r24.amos.X.hi\endcsname{121}
\expandafter\gdef\csname R@r24.amos.X.lo\endcsname{17}
\expandafter\gdef\csname R@r24.amos.X.within\endcsname{1}
\expandafter\gdef\csname R@r24.amos.d\endcsname{\ensuremath{+0.28}}
\expandafter\gdef\csname R@r24.amos.gl.est\endcsname{\ensuremath{+0.28}}
\expandafter\gdef\csname R@r24.amos.gl.hi\endcsname{\ensuremath{+0.46}}
\expandafter\gdef\csname R@r24.amos.gl.lo\endcsname{\ensuremath{+0.09}}
\expandafter\gdef\csname R@r24.amos.gl.nup\endcsname{3}
\expandafter\gdef\csname R@r24.amos.gl.s0\endcsname{\ensuremath{+0.209}}
\expandafter\gdef\csname R@r24.amos.gl.s1\endcsname{\ensuremath{+0.296}}
\expandafter\gdef\csname R@r24.amos.gl.s2\endcsname{\ensuremath{+0.339}}
\expandafter\gdef\csname R@r24.amos.gl.verdict\endcsname{CONFIRMED}
\expandafter\gdef\csname R@r24.amos.gm.est\endcsname{\ensuremath{+0.17}}
\expandafter\gdef\csname R@r24.amos.gm.hi\endcsname{\ensuremath{+0.31}}
\expandafter\gdef\csname R@r24.amos.gm.lo\endcsname{\ensuremath{+0.03}}
\expandafter\gdef\csname R@r24.amos.gm.nup\endcsname{3}
\expandafter\gdef\csname R@r24.amos.gm.s0\endcsname{\ensuremath{+0.076}}
\expandafter\gdef\csname R@r24.amos.gm.s1\endcsname{\ensuremath{+0.334}}
\expandafter\gdef\csname R@r24.amos.gm.s2\endcsname{\ensuremath{+0.092}}
\expandafter\gdef\csname R@r24.amos.gm.verdict\endcsname{CONFIRMED}
\expandafter\gdef\csname R@r24.amos.row\endcsname{2}
\expandafter\gdef\csname R@r24.cost.acdc.macs\endcsname{0.51}
\expandafter\gdef\csname R@r24.cost.acdc.params\endcsname{0.30}
\expandafter\gdef\csname R@r24.cost.acdc.step\endcsname{0.71}
\expandafter\gdef\csname R@r24.cost.amos.macs\endcsname{0.53}
\expandafter\gdef\csname R@r24.cost.amos.params\endcsname{0.30}
\expandafter\gdef\csname R@r24.cost.amos.step\endcsname{0.68}
\expandafter\gdef\csname R@r24.cost.params\endcsname{0.30}
\expandafter\gdef\csname R@r24.cost.step.r1\endcsname{0.7}
\expandafter\gdef\csname R@r24.gl.sign.n\endcsname{6}
\expandafter\gdef\csname R@r24.gl.sign.n.word\endcsname{six}
\expandafter\gdef\csname R@r24.gl.sign.nneg\endcsname{0}
\expandafter\gdef\csname R@r24.gl.sign.npos\endcsname{6}
\expandafter\gdef\csname R@r24.gl.sign.p\endcsname{0.016}
\expandafter\gdef\csname R@r24.gl.sign.pneg\endcsname{1.00}
\expandafter\gdef\csname R@r24.on\endcsname{1}
\makeatother

\makeatletter
\expandafter\gdef\csname R@r30.ab.11.d\endcsname{\ensuremath{-0.46}}
\expandafter\gdef\csname R@r30.ab.7.d\endcsname{\ensuremath{-0.68}}
\expandafter\gdef\csname R@r30.ab.8.d\endcsname{\ensuremath{+0.05}}
\expandafter\gdef\csname R@r30.bar\endcsname{1.15}
\expandafter\gdef\csname R@r30.nseed\endcsname{3}
\expandafter\gdef\csname R@r30.on\endcsname{1}
\expandafter\gdef\csname R@r30.sec48\endcsname{Narrowed to LeanNeXt-S's multiply-adds or to its parameter count, nnU-Net falls below LeanNeXt-S on fold 0 by more than two standard errors of the three-seed mean, and an EffiDec3D-style decoder at the same multiply-adds also stays below it.}
\expandafter\gdef\csname R@r30.sec48.branch\endcsname{1}
\makeatother

\makeatletter
\expandafter\gdef\csname R@r31.ctx.d\endcsname{\ensuremath{+0.07}}
\expandafter\gdef\csname R@r31.d\endcsname{\ensuremath{+1.00}}
\expandafter\gdef\csname R@r31.hi\endcsname{\ensuremath{+1.78}}
\expandafter\gdef\csname R@r31.lo\endcsname{\ensuremath{+0.27}}
\expandafter\gdef\csname R@r31.nfold\endcsname{4}
\expandafter\gdef\csname R@r31.nup\endcsname{4}
\expandafter\gdef\csname R@r31.on\endcsname{1}
\makeatother

\makeatletter
\expandafter\gdef\csname R@r32.acdc.L1.d\endcsname{\ensuremath{+0.36}}
\expandafter\gdef\csname R@r32.acdc.L1.hi\endcsname{\ensuremath{+0.55}}
\expandafter\gdef\csname R@r32.acdc.L1.lo\endcsname{\ensuremath{+0.19}}
\expandafter\gdef\csname R@r32.acdc.L1.mean\endcsname{91.75}
\expandafter\gdef\csname R@r32.acdc.V7A1.d\endcsname{\ensuremath{+0.08}}
\expandafter\gdef\csname R@r32.acdc.V7A1.hi\endcsname{\ensuremath{+0.24}}
\expandafter\gdef\csname R@r32.acdc.V7A1.lo\endcsname{\ensuremath{-0.08}}
\expandafter\gdef\csname R@r32.acdc.V7A1.mean\endcsname{91.47}
\expandafter\gdef\csname R@r32.acdc.V7C1.d\endcsname{\ensuremath{-0.19}}
\expandafter\gdef\csname R@r32.acdc.V7C1.hi\endcsname{\ensuremath{-0.05}}
\expandafter\gdef\csname R@r32.acdc.V7C1.lo\endcsname{\ensuremath{-0.33}}
\expandafter\gdef\csname R@r32.acdc.V7C1.mean\endcsname{91.20}
\expandafter\gdef\csname R@r32.acdc.base.mean\endcsname{91.39}
\expandafter\gdef\csname R@r32.amos.L1.d\endcsname{\ensuremath{+0.41}}
\expandafter\gdef\csname R@r32.amos.L1.hi\endcsname{\ensuremath{+0.63}}
\expandafter\gdef\csname R@r32.amos.L1.lo\endcsname{\ensuremath{+0.21}}
\expandafter\gdef\csname R@r32.amos.L1.mean\endcsname{89.27}
\expandafter\gdef\csname R@r32.amos.MedNeXtM.d\endcsname{\ensuremath{+0.45}}
\expandafter\gdef\csname R@r32.amos.MedNeXtM.hi\endcsname{\ensuremath{+0.67}}
\expandafter\gdef\csname R@r32.amos.MedNeXtM.lo\endcsname{\ensuremath{+0.26}}
\expandafter\gdef\csname R@r32.amos.MedNeXtM.mean\endcsname{89.32}
\expandafter\gdef\csname R@r32.amos.V7A1.d\endcsname{\ensuremath{+0.16}}
\expandafter\gdef\csname R@r32.amos.V7A1.hi\endcsname{\ensuremath{+0.36}}
\expandafter\gdef\csname R@r32.amos.V7A1.lo\endcsname{\ensuremath{-0.04}}
\expandafter\gdef\csname R@r32.amos.V7A1.mean\endcsname{89.03}
\expandafter\gdef\csname R@r32.amos.V7C1.d\endcsname{\ensuremath{-0.38}}
\expandafter\gdef\csname R@r32.amos.V7C1.hi\endcsname{\ensuremath{-0.20}}
\expandafter\gdef\csname R@r32.amos.V7C1.lo\endcsname{\ensuremath{-0.56}}
\expandafter\gdef\csname R@r32.amos.V7C1.mean\endcsname{88.49}
\expandafter\gdef\csname R@r32.amos.Y2.d\endcsname{\ensuremath{+0.09}}
\expandafter\gdef\csname R@r32.amos.Y2.hi\endcsname{\ensuremath{+0.28}}
\expandafter\gdef\csname R@r32.amos.Y2.lo\endcsname{\ensuremath{-0.10}}
\expandafter\gdef\csname R@r32.amos.Y2.mean\endcsname{88.96}
\expandafter\gdef\csname R@r32.amos.base.mean\endcsname{88.87}
\expandafter\gdef\csname R@r32.ctx.kits.Y2.d\endcsname{\ensuremath{+0.91}}
\expandafter\gdef\csname R@r32.ctx.kits.Y2.hi\endcsname{\ensuremath{+1.59}}
\expandafter\gdef\csname R@r32.ctx.kits.Y2.lo\endcsname{\ensuremath{+0.28}}
\expandafter\gdef\csname R@r32.ctx.kits.Y2.nup\endcsname{4}
\expandafter\gdef\csname R@r32.kits.L1.d\endcsname{\ensuremath{+1.49}}
\expandafter\gdef\csname R@r32.kits.L1.hi\endcsname{\ensuremath{+2.33}}
\expandafter\gdef\csname R@r32.kits.L1.lo\endcsname{\ensuremath{+0.68}}
\expandafter\gdef\csname R@r32.kits.L1.mean\endcsname{87.01}
\expandafter\gdef\csname R@r32.kits.MedNeXtM.d\endcsname{\ensuremath{-1.19}}
\expandafter\gdef\csname R@r32.kits.MedNeXtM.hi\endcsname{\ensuremath{-0.36}}
\expandafter\gdef\csname R@r32.kits.MedNeXtM.lo\endcsname{\ensuremath{-2.02}}
\expandafter\gdef\csname R@r32.kits.MedNeXtM.mean\endcsname{84.33}
\expandafter\gdef\csname R@r32.kits.V7A1.d\endcsname{\ensuremath{+1.06}}
\expandafter\gdef\csname R@r32.kits.V7A1.d2\endcsname{\ensuremath{+1.10}}
\expandafter\gdef\csname R@r32.kits.V7A1.hi\endcsname{\ensuremath{+1.74}}
\expandafter\gdef\csname R@r32.kits.V7A1.lo\endcsname{\ensuremath{+0.45}}
\expandafter\gdef\csname R@r32.kits.V7A1.mean\endcsname{86.58}
\expandafter\gdef\csname R@r32.kits.V7A1.mean2\endcsname{86.62}
\expandafter\gdef\csname R@r32.kits.V7A1.nup2\endcsname{4}
\expandafter\gdef\csname R@r32.kits.V7A1.ts.hi\endcsname{\ensuremath{+1.74}}
\expandafter\gdef\csname R@r32.kits.V7A1.ts.lo\endcsname{\ensuremath{+0.45}}
\expandafter\gdef\csname R@r32.kits.V7C1.d\endcsname{\ensuremath{+0.47}}
\expandafter\gdef\csname R@r32.kits.V7C1.hi\endcsname{\ensuremath{+1.02}}
\expandafter\gdef\csname R@r32.kits.V7C1.lo\endcsname{\ensuremath{-0.05}}
\expandafter\gdef\csname R@r32.kits.V7C1.mean\endcsname{85.99}
\expandafter\gdef\csname R@r32.kits.Y2.d\endcsname{\ensuremath{+1.97}}
\expandafter\gdef\csname R@r32.kits.Y2.hi\endcsname{\ensuremath{+2.85}}
\expandafter\gdef\csname R@r32.kits.Y2.lo\endcsname{\ensuremath{+1.15}}
\expandafter\gdef\csname R@r32.kits.Y2.mean\endcsname{87.48}
\expandafter\gdef\csname R@r32.kits.base.mean\endcsname{85.51}
\expandafter\gdef\csname R@r32.kits.base.mean2\endcsname{85.53}
\expandafter\gdef\csname R@r32.lacdc.L1.d\endcsname{\ensuremath{+0.45}}
\expandafter\gdef\csname R@r32.lacdc.L1.hi\endcsname{\ensuremath{+0.66}}
\expandafter\gdef\csname R@r32.lacdc.L1.lo\endcsname{\ensuremath{+0.26}}
\expandafter\gdef\csname R@r32.lacdc.L1.mean\endcsname{92.25}
\expandafter\gdef\csname R@r32.lacdc.V7A1.d\endcsname{\ensuremath{+0.28}}
\expandafter\gdef\csname R@r32.lacdc.V7A1.hi\endcsname{\ensuremath{+0.49}}
\expandafter\gdef\csname R@r32.lacdc.V7A1.lo\endcsname{\ensuremath{+0.08}}
\expandafter\gdef\csname R@r32.lacdc.V7A1.mean\endcsname{92.08}
\expandafter\gdef\csname R@r32.lacdc.base.mean\endcsname{91.80}
\expandafter\gdef\csname R@r32.lamos.L1.d\endcsname{\ensuremath{+0.27}}
\expandafter\gdef\csname R@r32.lamos.L1.hi\endcsname{\ensuremath{+0.46}}
\expandafter\gdef\csname R@r32.lamos.L1.lo\endcsname{\ensuremath{+0.09}}
\expandafter\gdef\csname R@r32.lamos.L1.mean\endcsname{90.48}
\expandafter\gdef\csname R@r32.lamos.V7A1.d\endcsname{\ensuremath{+0.10}}
\expandafter\gdef\csname R@r32.lamos.V7A1.hi\endcsname{\ensuremath{+0.28}}
\expandafter\gdef\csname R@r32.lamos.V7A1.lo\endcsname{\ensuremath{-0.06}}
\expandafter\gdef\csname R@r32.lamos.V7A1.mean\endcsname{90.32}
\expandafter\gdef\csname R@r32.lamos.base.mean\endcsname{90.21}
\expandafter\gdef\csname R@r32.lsign.n\endcsname{6}
\expandafter\gdef\csname R@r32.lsign.npos\endcsname{5}
\expandafter\gdef\csname R@r32.on\endcsname{1}
\makeatother

\makeatletter
\expandafter\gdef\csname R@r33.c.acdc.d\endcsname{\ensuremath{+0.06}}
\expandafter\gdef\csname R@r33.c.acdc.hi\endcsname{\ensuremath{+0.22}}
\expandafter\gdef\csname R@r33.c.acdc.lo\endcsname{\ensuremath{-0.10}}
\expandafter\gdef\csname R@r33.c.acdc.nup\endcsname{2}
\expandafter\gdef\csname R@r33.c.amos.d\endcsname{\ensuremath{-0.24}}
\expandafter\gdef\csname R@r33.c.amos.hi\endcsname{\ensuremath{-0.02}}
\expandafter\gdef\csname R@r33.c.amos.lo\endcsname{\ensuremath{-0.46}}
\expandafter\gdef\csname R@r33.c.amos.nup\endcsname{1}
\expandafter\gdef\csname R@r33.c.kits.d\endcsname{\ensuremath{+1.42}}
\expandafter\gdef\csname R@r33.c.kits.hi\endcsname{\ensuremath{+2.50}}
\expandafter\gdef\csname R@r33.c.kits.lo\endcsname{\ensuremath{+0.35}}
\expandafter\gdef\csname R@r33.c.kits.nup\endcsname{4}
\expandafter\gdef\csname R@r33.cost.acdc.gmac\endcsname{259.4}
\expandafter\gdef\csname R@r33.cost.acdc.mem\endcsname{10.37}
\expandafter\gdef\csname R@r33.cost.acdc.params\endcsname{30.95}
\expandafter\gdef\csname R@r33.cost.acdc.step\endcsname{138.0}
\expandafter\gdef\csname R@r33.cost.amos.gmac\endcsname{431.7}
\expandafter\gdef\csname R@r33.cost.amos.mem\endcsname{8.93}
\expandafter\gdef\csname R@r33.cost.amos.params\endcsname{31.69}
\expandafter\gdef\csname R@r33.cost.amos.step\endcsname{137.6}
\expandafter\gdef\csname R@r33.cost.kits.gmac\endcsname{426.6}
\expandafter\gdef\csname R@r33.cost.kits.mem\endcsname{8.46}
\expandafter\gdef\csname R@r33.cost.kits.params\endcsname{31.76}
\expandafter\gdef\csname R@r33.cost.kits.step\endcsname{121.7}
\expandafter\gdef\csname R@r33.mean.acdc.L1S\endcsname{91.01}
\expandafter\gdef\csname R@r33.mean.acdc.V7A1\endcsname{91.11}
\expandafter\gdef\csname R@r33.mean.acdc.base\endcsname{90.94}
\expandafter\gdef\csname R@r33.mean.amos.L1S\endcsname{88.11}
\expandafter\gdef\csname R@r33.mean.amos.V7A1\endcsname{88.61}
\expandafter\gdef\csname R@r33.mean.amos.base\endcsname{88.35}
\expandafter\gdef\csname R@r33.mean.kits.L1S\endcsname{85.35}
\expandafter\gdef\csname R@r33.mean.kits.L1S.s0\endcsname{85.53}
\expandafter\gdef\csname R@r33.mean.kits.L1S.s1\endcsname{85.17}
\expandafter\gdef\csname R@r33.mean.kits.V7A1\endcsname{85.56}
\expandafter\gdef\csname R@r33.mean.kits.V7A1.s0\endcsname{85.63}
\expandafter\gdef\csname R@r33.mean.kits.V7A1.s1\endcsname{85.50}
\expandafter\gdef\csname R@r33.mean.kits.base\endcsname{83.93}
\expandafter\gdef\csname R@r33.on\endcsname{1}
\expandafter\gdef\csname R@r33.p.acdc.d\endcsname{\ensuremath{+0.11}}
\expandafter\gdef\csname R@r33.p.acdc.fd.1\endcsname{\ensuremath{+0.21}}
\expandafter\gdef\csname R@r33.p.acdc.fd.2\endcsname{\ensuremath{+0.21}}
\expandafter\gdef\csname R@r33.p.acdc.fd.3\endcsname{\ensuremath{-0.40}}
\expandafter\gdef\csname R@r33.p.acdc.fd.4\endcsname{\ensuremath{+0.40}}
\expandafter\gdef\csname R@r33.p.acdc.hi\endcsname{\ensuremath{+0.31}}
\expandafter\gdef\csname R@r33.p.acdc.lo\endcsname{\ensuremath{-0.10}}
\expandafter\gdef\csname R@r33.p.acdc.nfold\endcsname{4}
\expandafter\gdef\csname R@r33.p.acdc.nup\endcsname{3}
\expandafter\gdef\csname R@r33.p.amos.d\endcsname{\ensuremath{+0.50}}
\expandafter\gdef\csname R@r33.p.amos.fd.1\endcsname{\ensuremath{+0.02}}
\expandafter\gdef\csname R@r33.p.amos.fd.2\endcsname{\ensuremath{+0.66}}
\expandafter\gdef\csname R@r33.p.amos.fd.3\endcsname{\ensuremath{+0.62}}
\expandafter\gdef\csname R@r33.p.amos.fd.4\endcsname{\ensuremath{+0.69}}
\expandafter\gdef\csname R@r33.p.amos.hi\endcsname{\ensuremath{+0.75}}
\expandafter\gdef\csname R@r33.p.amos.lo\endcsname{\ensuremath{+0.26}}
\expandafter\gdef\csname R@r33.p.amos.nfold\endcsname{4}
\expandafter\gdef\csname R@r33.p.amos.nup\endcsname{4}
\expandafter\gdef\csname R@r33.p.kits.d\endcsname{\ensuremath{+0.21}}
\expandafter\gdef\csname R@r33.p.kits.fd.1\endcsname{\ensuremath{-0.17}}
\expandafter\gdef\csname R@r33.p.kits.fd.2\endcsname{\ensuremath{+0.89}}
\expandafter\gdef\csname R@r33.p.kits.fd.3\endcsname{\ensuremath{+0.08}}
\expandafter\gdef\csname R@r33.p.kits.fd.4\endcsname{\ensuremath{+0.03}}
\expandafter\gdef\csname R@r33.p.kits.hi\endcsname{\ensuremath{+1.00}}
\expandafter\gdef\csname R@r33.p.kits.lo\endcsname{\ensuremath{-0.58}}
\expandafter\gdef\csname R@r33.p.kits.nfold\endcsname{4}
\expandafter\gdef\csname R@r33.p.kits.nup\endcsname{3}
\expandafter\gdef\csname R@r33.p.kits.pairs.dn\endcsname{5}
\expandafter\gdef\csname R@r33.p.kits.pairs.n\endcsname{8}
\expandafter\gdef\csname R@r33.p.kits.pairs.p\endcsname{0.3633}
\expandafter\gdef\csname R@r33.ratio.acdc.gmac\endcsname{0.82}
\expandafter\gdef\csname R@r33.ratio.acdc.params\endcsname{1.02}
\expandafter\gdef\csname R@r33.ratio.acdc.step\endcsname{0.97}
\expandafter\gdef\csname R@r33.ratio.amos.gmac\endcsname{0.82}
\expandafter\gdef\csname R@r33.ratio.amos.params\endcsname{1.02}
\expandafter\gdef\csname R@r33.ratio.amos.step\endcsname{0.98}
\expandafter\gdef\csname R@r33.ratio.kits.gmac\endcsname{0.89}
\expandafter\gdef\csname R@r33.ratio.kits.params\endcsname{1.02}
\expandafter\gdef\csname R@r33.ratio.kits.step\endcsname{0.99}
\expandafter\gdef\csname R@r33.sched.blocks\endcsname{7}
\expandafter\gdef\csname R@r33.sched.l1.n\endcsname{1, 3, 4, 6, 6, 6}
\expandafter\gdef\csname R@r33.sched.n\endcsname{1, 1, 1, 1, 1, 2}
\expandafter\gdef\csname R@r33.sched.params\endcsname{31.76}
\expandafter\gdef\csname R@r33.sched.ru.n\endcsname{1, 1, 1, 1, 1, 1}
\expandafter\gdef\csname R@r33.sched.ru.params\endcsname{26.23}
\expandafter\gdef\csname R@r33.sched.target\endcsname{31.20}
\makeatother

\makeatletter
\expandafter\gdef\csname R@r34.d.L1\endcsname{\ensuremath{+2.07}}
\expandafter\gdef\csname R@r34.d.V7A1\endcsname{\ensuremath{+1.63}}
\expandafter\gdef\csname R@r34.d.Y2\endcsname{\ensuremath{+2.57}}
\expandafter\gdef\csname R@r34.fold.L1.1\endcsname{\ensuremath{+1.75}}
\expandafter\gdef\csname R@r34.fold.L1.2\endcsname{\ensuremath{+3.26}}
\expandafter\gdef\csname R@r34.fold.L1.3\endcsname{\ensuremath{+1.35}}
\expandafter\gdef\csname R@r34.fold.L1.4\endcsname{\ensuremath{+1.92}}
\expandafter\gdef\csname R@r34.fold.V7A1.1\endcsname{\ensuremath{+1.20}}
\expandafter\gdef\csname R@r34.fold.V7A1.2\endcsname{\ensuremath{+2.98}}
\expandafter\gdef\csname R@r34.fold.V7A1.3\endcsname{\ensuremath{+1.50}}
\expandafter\gdef\csname R@r34.fold.V7A1.4\endcsname{\ensuremath{+0.86}}
\expandafter\gdef\csname R@r34.fold.Y2.1\endcsname{\ensuremath{+2.96}}
\expandafter\gdef\csname R@r34.fold.Y2.2\endcsname{\ensuremath{+3.02}}
\expandafter\gdef\csname R@r34.fold.Y2.3\endcsname{\ensuremath{+1.71}}
\expandafter\gdef\csname R@r34.fold.Y2.4\endcsname{\ensuremath{+2.61}}
\expandafter\gdef\csname R@r34.i.d\endcsname{\ensuremath{+0.50}}
\expandafter\gdef\csname R@r34.i.fd.1\endcsname{\ensuremath{+1.21}}
\expandafter\gdef\csname R@r34.i.fd.2\endcsname{\ensuremath{-0.24}}
\expandafter\gdef\csname R@r34.i.fd.3\endcsname{\ensuremath{+0.36}}
\expandafter\gdef\csname R@r34.i.fd.4\endcsname{\ensuremath{+0.68}}
\expandafter\gdef\csname R@r34.i.hi\endcsname{\ensuremath{+1.22}}
\expandafter\gdef\csname R@r34.i.k\endcsname{6}
\expandafter\gdef\csname R@r34.i.lo\endcsname{\ensuremath{-0.21}}
\expandafter\gdef\csname R@r34.i.n\endcsname{8}
\expandafter\gdef\csname R@r34.i.p\endcsname{0.1445}
\expandafter\gdef\csname R@r34.i.s0.d\endcsname{\ensuremath{+0.56}}
\expandafter\gdef\csname R@r34.i.s0.nup\endcsname{4}
\expandafter\gdef\csname R@r34.i.s1.d\endcsname{\ensuremath{+0.45}}
\expandafter\gdef\csname R@r34.i.s1.nup\endcsname{2}
\expandafter\gdef\csname R@r34.ii.d\endcsname{\ensuremath{+0.94}}
\expandafter\gdef\csname R@r34.ii.fd.1\endcsname{\ensuremath{+1.76}}
\expandafter\gdef\csname R@r34.ii.fd.2\endcsname{\ensuremath{+0.04}}
\expandafter\gdef\csname R@r34.ii.fd.3\endcsname{\ensuremath{+0.20}}
\expandafter\gdef\csname R@r34.ii.fd.4\endcsname{\ensuremath{+1.75}}
\expandafter\gdef\csname R@r34.ii.hi\endcsname{\ensuremath{+1.59}}
\expandafter\gdef\csname R@r34.ii.k\endcsname{7}
\expandafter\gdef\csname R@r34.ii.lo\endcsname{\ensuremath{+0.29}}
\expandafter\gdef\csname R@r34.ii.n\endcsname{8}
\expandafter\gdef\csname R@r34.ii.p\endcsname{0.0352}
\expandafter\gdef\csname R@r34.ii.s0.d\endcsname{\ensuremath{+1.07}}
\expandafter\gdef\csname R@r34.ii.s0.nup\endcsname{4}
\expandafter\gdef\csname R@r34.ii.s1.d\endcsname{\ensuremath{+0.81}}
\expandafter\gdef\csname R@r34.ii.s1.nup\endcsname{3}
\expandafter\gdef\csname R@r34.iiiL1.d\endcsname{\ensuremath{+2.07}}
\expandafter\gdef\csname R@r34.iiiL1.fd.1\endcsname{\ensuremath{+1.75}}
\expandafter\gdef\csname R@r34.iiiL1.fd.2\endcsname{\ensuremath{+3.26}}
\expandafter\gdef\csname R@r34.iiiL1.fd.3\endcsname{\ensuremath{+1.35}}
\expandafter\gdef\csname R@r34.iiiL1.fd.4\endcsname{\ensuremath{+1.92}}
\expandafter\gdef\csname R@r34.iiiL1.hi\endcsname{\ensuremath{+3.01}}
\expandafter\gdef\csname R@r34.iiiL1.k\endcsname{8}
\expandafter\gdef\csname R@r34.iiiL1.lo\endcsname{\ensuremath{+1.13}}
\expandafter\gdef\csname R@r34.iiiL1.n\endcsname{8}
\expandafter\gdef\csname R@r34.iiiL1.p\endcsname{0.0039}
\expandafter\gdef\csname R@r34.iiiL1.s0.d\endcsname{\ensuremath{+2.15}}
\expandafter\gdef\csname R@r34.iiiL1.s1.d\endcsname{\ensuremath{+1.99}}
\expandafter\gdef\csname R@r34.iiiY2.d\endcsname{\ensuremath{+2.57}}
\expandafter\gdef\csname R@r34.iiiY2.fd.1\endcsname{\ensuremath{+2.96}}
\expandafter\gdef\csname R@r34.iiiY2.fd.2\endcsname{\ensuremath{+3.02}}
\expandafter\gdef\csname R@r34.iiiY2.fd.3\endcsname{\ensuremath{+1.71}}
\expandafter\gdef\csname R@r34.iiiY2.fd.4\endcsname{\ensuremath{+2.61}}
\expandafter\gdef\csname R@r34.iiiY2.hi\endcsname{\ensuremath{+3.59}}
\expandafter\gdef\csname R@r34.iiiY2.k\endcsname{8}
\expandafter\gdef\csname R@r34.iiiY2.lo\endcsname{\ensuremath{+1.55}}
\expandafter\gdef\csname R@r34.iiiY2.n\endcsname{8}
\expandafter\gdef\csname R@r34.iiiY2.p\endcsname{0.0039}
\expandafter\gdef\csname R@r34.iiiY2.s0.d\endcsname{\ensuremath{+2.71}}
\expandafter\gdef\csname R@r34.iiiY2.s1.d\endcsname{\ensuremath{+2.44}}
\expandafter\gdef\csname R@r34.mean.L1\endcsname{86.00}
\expandafter\gdef\csname R@r34.mean.L1.s1\endcsname{85.86}
\expandafter\gdef\csname R@r34.mean.V7A1\endcsname{85.56}
\expandafter\gdef\csname R@r34.mean.V7A1.s1\endcsname{85.50}
\expandafter\gdef\csname R@r34.mean.Y2\endcsname{86.50}
\expandafter\gdef\csname R@r34.mean.Y2.s1\endcsname{86.31}
\expandafter\gdef\csname R@r34.mean.base\endcsname{83.93}
\expandafter\gdef\csname R@r34.mean.base.s1\endcsname{83.86}
\expandafter\gdef\csname R@r34.on\endcsname{1}
\expandafter\gdef\csname R@r34.prereg.file\endcsname{PREREG\_\allowbreak{}R34\_\allowbreak{}20261101.md}
\expandafter\gdef\csname R@r34.prereg.sha16\endcsname{eaa64aa4232a43b6}
\expandafter\gdef\csname R@r34.share.V7A1\endcsname{79}
\makeatother

\makeatletter
\expandafter\gdef\csname R@r35.a.n\endcsname{21}
\expandafter\gdef\csname R@r35.a.p\endcsname{0.06}
\expandafter\gdef\csname R@r35.a.rho\endcsname{\ensuremath{+0.35}}
\expandafter\gdef\csname R@r35.b.above\endcsname{\ensuremath{+0.69}}
\expandafter\gdef\csname R@r35.b.below\endcsname{\ensuremath{+0.35}}
\expandafter\gdef\csname R@r35.b.gap\endcsname{\ensuremath{+0.35}}
\expandafter\gdef\csname R@r35.b.nabove\endcsname{11}
\expandafter\gdef\csname R@r35.b.nbelow\endcsname{10}
\expandafter\gdef\csname R@r35.b.p\endcsname{0.08}
\expandafter\gdef\csname R@r35.c.acdc.above\endcsname{\ensuremath{+0.25}}
\expandafter\gdef\csname R@r35.c.acdc.below\endcsname{--}
\expandafter\gdef\csname R@r35.c.acdc.nabove\endcsname{3}
\expandafter\gdef\csname R@r35.c.acdc.nbelow\endcsname{0}
\expandafter\gdef\csname R@r35.c.amos.above\endcsname{\ensuremath{+0.86}}
\expandafter\gdef\csname R@r35.c.amos.below\endcsname{\ensuremath{+0.35}}
\expandafter\gdef\csname R@r35.c.amos.nabove\endcsname{8}
\expandafter\gdef\csname R@r35.c.amos.nbelow\endcsname{7}
\expandafter\gdef\csname R@r35.c.kits.above\endcsname{--}
\expandafter\gdef\csname R@r35.c.kits.below\endcsname{\ensuremath{+0.34}}
\expandafter\gdef\csname R@r35.c.kits.nabove\endcsname{0}
\expandafter\gdef\csname R@r35.c.kits.nbelow\endcsname{3}
\expandafter\gdef\csname R@r35.cls.acdc.lv.bf\endcsname{0.42}
\expandafter\gdef\csname R@r35.cls.acdc.lv.d\endcsname{\ensuremath{+0.24}}
\expandafter\gdef\csname R@r35.cls.acdc.lv.n\endcsname{160}
\expandafter\gdef\csname R@r35.cls.acdc.lv.name\endcsname{LV cavity}
\expandafter\gdef\csname R@r35.cls.acdc.lv.side\endcsname{$>$}
\expandafter\gdef\csname R@r35.cls.acdc.myo.bf\endcsname{0.68}
\expandafter\gdef\csname R@r35.cls.acdc.myo.d\endcsname{\ensuremath{+0.18}}
\expandafter\gdef\csname R@r35.cls.acdc.myo.n\endcsname{160}
\expandafter\gdef\csname R@r35.cls.acdc.myo.name\endcsname{myocardium}
\expandafter\gdef\csname R@r35.cls.acdc.myo.side\endcsname{$>$}
\expandafter\gdef\csname R@r35.cls.acdc.rv.bf\endcsname{0.63}
\expandafter\gdef\csname R@r35.cls.acdc.rv.d\endcsname{\ensuremath{+0.33}}
\expandafter\gdef\csname R@r35.cls.acdc.rv.n\endcsname{160}
\expandafter\gdef\csname R@r35.cls.acdc.rv.name\endcsname{RV}
\expandafter\gdef\csname R@r35.cls.acdc.rv.side\endcsname{$>$}
\expandafter\gdef\csname R@r35.cls.amos.aorta.bf\endcsname{0.23}
\expandafter\gdef\csname R@r35.cls.amos.aorta.d\endcsname{\ensuremath{+0.15}}
\expandafter\gdef\csname R@r35.cls.amos.aorta.n\endcsname{160}
\expandafter\gdef\csname R@r35.cls.amos.aorta.name\endcsname{aorta}
\expandafter\gdef\csname R@r35.cls.amos.aorta.side\endcsname{$\le$}
\expandafter\gdef\csname R@r35.cls.amos.bladder.bf\endcsname{0.44}
\expandafter\gdef\csname R@r35.cls.amos.bladder.d\endcsname{\ensuremath{+0.59}}
\expandafter\gdef\csname R@r35.cls.amos.bladder.n\endcsname{155}
\expandafter\gdef\csname R@r35.cls.amos.bladder.name\endcsname{bladder}
\expandafter\gdef\csname R@r35.cls.amos.bladder.side\endcsname{$>$}
\expandafter\gdef\csname R@r35.cls.amos.duodenum.bf\endcsname{0.43}
\expandafter\gdef\csname R@r35.cls.amos.duodenum.d\endcsname{\ensuremath{+1.03}}
\expandafter\gdef\csname R@r35.cls.amos.duodenum.n\endcsname{160}
\expandafter\gdef\csname R@r35.cls.amos.duodenum.name\endcsname{duodenum}
\expandafter\gdef\csname R@r35.cls.amos.duodenum.side\endcsname{$>$}
\expandafter\gdef\csname R@r35.cls.amos.esophagus.bf\endcsname{0.52}
\expandafter\gdef\csname R@r35.cls.amos.esophagus.d\endcsname{\ensuremath{+0.67}}
\expandafter\gdef\csname R@r35.cls.amos.esophagus.n\endcsname{159}
\expandafter\gdef\csname R@r35.cls.amos.esophagus.name\endcsname{oesophagus}
\expandafter\gdef\csname R@r35.cls.amos.esophagus.side\endcsname{$>$}
\expandafter\gdef\csname R@r35.cls.amos.gallbladder.bf\endcsname{0.47}
\expandafter\gdef\csname R@r35.cls.amos.gallbladder.d\endcsname{\ensuremath{+1.26}}
\expandafter\gdef\csname R@r35.cls.amos.gallbladder.n\endcsname{151}
\expandafter\gdef\csname R@r35.cls.amos.gallbladder.name\endcsname{gallbladder}
\expandafter\gdef\csname R@r35.cls.amos.gallbladder.side\endcsname{$>$}
\expandafter\gdef\csname R@r35.cls.amos.ivc.bf\endcsname{0.27}
\expandafter\gdef\csname R@r35.cls.amos.ivc.d\endcsname{\ensuremath{+0.28}}
\expandafter\gdef\csname R@r35.cls.amos.ivc.n\endcsname{160}
\expandafter\gdef\csname R@r35.cls.amos.ivc.name\endcsname{IVC}
\expandafter\gdef\csname R@r35.cls.amos.ivc.side\endcsname{$\le$}
\expandafter\gdef\csname R@r35.cls.amos.ladrenal.bf\endcsname{0.75}
\expandafter\gdef\csname R@r35.cls.amos.ladrenal.d\endcsname{\ensuremath{+0.68}}
\expandafter\gdef\csname R@r35.cls.amos.ladrenal.n\endcsname{160}
\expandafter\gdef\csname R@r35.cls.amos.ladrenal.name\endcsname{L adrenal}
\expandafter\gdef\csname R@r35.cls.amos.ladrenal.side\endcsname{$>$}
\expandafter\gdef\csname R@r35.cls.amos.liver.bf\endcsname{0.15}
\expandafter\gdef\csname R@r35.cls.amos.liver.d\endcsname{\ensuremath{+0.07}}
\expandafter\gdef\csname R@r35.cls.amos.liver.n\endcsname{160}
\expandafter\gdef\csname R@r35.cls.amos.liver.name\endcsname{liver}
\expandafter\gdef\csname R@r35.cls.amos.liver.side\endcsname{$\le$}
\expandafter\gdef\csname R@r35.cls.amos.lkidney.bf\endcsname{0.25}
\expandafter\gdef\csname R@r35.cls.amos.lkidney.d\endcsname{\ensuremath{+0.67}}
\expandafter\gdef\csname R@r35.cls.amos.lkidney.n\endcsname{159}
\expandafter\gdef\csname R@r35.cls.amos.lkidney.name\endcsname{L kidney}
\expandafter\gdef\csname R@r35.cls.amos.lkidney.side\endcsname{$\le$}
\expandafter\gdef\csname R@r35.cls.amos.pancreas.bf\endcsname{0.46}
\expandafter\gdef\csname R@r35.cls.amos.pancreas.d\endcsname{\ensuremath{-0.12}}
\expandafter\gdef\csname R@r35.cls.amos.pancreas.n\endcsname{160}
\expandafter\gdef\csname R@r35.cls.amos.pancreas.name\endcsname{pancreas}
\expandafter\gdef\csname R@r35.cls.amos.pancreas.side\endcsname{$>$}
\expandafter\gdef\csname R@r35.cls.amos.prostuterus.bf\endcsname{0.39}
\expandafter\gdef\csname R@r35.cls.amos.prostuterus.d\endcsname{\ensuremath{+2.15}}
\expandafter\gdef\csname R@r35.cls.amos.prostuterus.n\endcsname{153}
\expandafter\gdef\csname R@r35.cls.amos.prostuterus.name\endcsname{prostate/uterus}
\expandafter\gdef\csname R@r35.cls.amos.prostuterus.side\endcsname{$>$}
\expandafter\gdef\csname R@r35.cls.amos.radrenal.bf\endcsname{0.78}
\expandafter\gdef\csname R@r35.cls.amos.radrenal.d\endcsname{\ensuremath{+0.62}}
\expandafter\gdef\csname R@r35.cls.amos.radrenal.n\endcsname{160}
\expandafter\gdef\csname R@r35.cls.amos.radrenal.name\endcsname{R adrenal}
\expandafter\gdef\csname R@r35.cls.amos.radrenal.side\endcsname{$>$}
\expandafter\gdef\csname R@r35.cls.amos.rkidney.bf\endcsname{0.25}
\expandafter\gdef\csname R@r35.cls.amos.rkidney.d\endcsname{\ensuremath{+0.19}}
\expandafter\gdef\csname R@r35.cls.amos.rkidney.n\endcsname{160}
\expandafter\gdef\csname R@r35.cls.amos.rkidney.name\endcsname{R kidney}
\expandafter\gdef\csname R@r35.cls.amos.rkidney.side\endcsname{$\le$}
\expandafter\gdef\csname R@r35.cls.amos.spleen.bf\endcsname{0.24}
\expandafter\gdef\csname R@r35.cls.amos.spleen.d\endcsname{\ensuremath{+0.34}}
\expandafter\gdef\csname R@r35.cls.amos.spleen.n\endcsname{160}
\expandafter\gdef\csname R@r35.cls.amos.spleen.name\endcsname{spleen}
\expandafter\gdef\csname R@r35.cls.amos.spleen.side\endcsname{$\le$}
\expandafter\gdef\csname R@r35.cls.amos.stomach.bf\endcsname{0.28}
\expandafter\gdef\csname R@r35.cls.amos.stomach.d\endcsname{\ensuremath{+0.76}}
\expandafter\gdef\csname R@r35.cls.amos.stomach.n\endcsname{158}
\expandafter\gdef\csname R@r35.cls.amos.stomach.name\endcsname{stomach}
\expandafter\gdef\csname R@r35.cls.amos.stomach.side\endcsname{$\le$}
\expandafter\gdef\csname R@r35.cls.kits.kidney.bf\endcsname{0.17}
\expandafter\gdef\csname R@r35.cls.kits.kidney.d\endcsname{\ensuremath{-0.34}}
\expandafter\gdef\csname R@r35.cls.kits.kidney.n\endcsname{391}
\expandafter\gdef\csname R@r35.cls.kits.kidney.name\endcsname{kidney}
\expandafter\gdef\csname R@r35.cls.kits.kidney.side\endcsname{$\le$}
\expandafter\gdef\csname R@r35.cls.kits.masses.bf\endcsname{0.31}
\expandafter\gdef\csname R@r35.cls.kits.masses.d\endcsname{\ensuremath{+0.09}}
\expandafter\gdef\csname R@r35.cls.kits.masses.n\endcsname{391}
\expandafter\gdef\csname R@r35.cls.kits.masses.name\endcsname{masses}
\expandafter\gdef\csname R@r35.cls.kits.masses.side\endcsname{$\le$}
\expandafter\gdef\csname R@r35.cls.kits.tumour.bf\endcsname{0.31}
\expandafter\gdef\csname R@r35.cls.kits.tumour.d\endcsname{\ensuremath{+1.26}}
\expandafter\gdef\csname R@r35.cls.kits.tumour.n\endcsname{391}
\expandafter\gdef\csname R@r35.cls.kits.tumour.name\endcsname{tumour}
\expandafter\gdef\csname R@r35.cls.kits.tumour.side\endcsname{$\le$}
\expandafter\gdef\csname R@r35.ctx.rho\endcsname{\ensuremath{-0.83}}
\expandafter\gdef\csname R@r35.n\endcsname{15}
\expandafter\gdef\csname R@r35.nfold\endcsname{4}
\expandafter\gdef\csname R@r35.nperm\endcsname{100000}
\expandafter\gdef\csname R@r35.on\endcsname{1}
\expandafter\gdef\csname R@r35.p\endcsname{0.04}
\expandafter\gdef\csname R@r35.p.exact\endcsname{0.0398}
\expandafter\gdef\csname R@r35.prereg.file\endcsname{PREREG\_\allowbreak{}R35\_\allowbreak{}20261102.md}
\expandafter\gdef\csname R@r35.prereg.sha16\endcsname{d2918c292c80424e}
\expandafter\gdef\csname R@r35.rho\endcsname{\ensuremath{+0.47}}
\expandafter\gdef\csname R@r35.thr\endcsname{0.33}
\makeatother

\makeatletter
\expandafter\gdef\csname R@r37.hi\endcsname{\ensuremath{-0.08}}
\expandafter\gdef\csname R@r37.lo\endcsname{\ensuremath{-0.60}}
\expandafter\gdef\csname R@r37.m.035.n\endcsname{20}
\expandafter\gdef\csname R@r37.m.035.shrink\endcsname{\ensuremath{-0.33}}
\expandafter\gdef\csname R@r37.m.075.n\endcsname{16}
\expandafter\gdef\csname R@r37.m.075.shrink\endcsname{\ensuremath{-0.40}}
\expandafter\gdef\csname R@r37.m.125.n\endcsname{12}
\expandafter\gdef\csname R@r37.m.125.shrink\endcsname{\ensuremath{-0.57}}
\expandafter\gdef\csname R@r37.m.175.n\endcsname{6}
\expandafter\gdef\csname R@r37.m.175.shrink\endcsname{\ensuremath{-0.87}}
\expandafter\gdef\csname R@r37.mean\endcsname{\ensuremath{-0.34}}
\expandafter\gdef\csname R@r37.on\endcsname{1}
\makeatother

\makeatletter
\expandafter\gdef\csname R@r4.amos.stages\endcsname{D2 and D1}
\expandafter\gdef\csname R@r4.bands\endcsname{\ensuremath{-30}, 30, 80 and 200}
\expandafter\gdef\csname R@r4.gmac.kits\endcsname{373.6}
\expandafter\gdef\csname R@r4.kits.res\endcsname{$1/4$ and $1/8$}
\expandafter\gdef\csname R@r4.kits.stages\endcsname{D2 and D1}
\expandafter\gdef\csname R@r4.nstat\endcsname{6}
\expandafter\gdef\csname R@r4.par.extra\endcsname{2\,688}
\expandafter\gdef\csname R@r4.step.amos\endcsname{159.4}
\expandafter\gdef\csname R@r4.step.amos.v7a1\endcsname{140.3}
\expandafter\gdef\csname R@r4.step.kits\endcsname{117.1}
\expandafter\gdef\csname R@r4.step.kits.v7a1\endcsname{113.1}
\expandafter\gdef\csname R@r4.steps\endcsname{8}
\expandafter\gdef\csname R@r4.tau\endcsname{25}
\makeatother

\makeatletter
\expandafter\gdef\csname R@r8.a.V7A1.liver\endcsname{96.48}
\expandafter\gdef\csname R@r8.a.V7A1.mean\endcsname{82.44}
\expandafter\gdef\csname R@r8.a.V7A1.nseed\endcsname{3}
\expandafter\gdef\csname R@r8.a.V7A1.tumour\endcsname{68.40}
\expandafter\gdef\csname R@r8.a.Y2.liver\endcsname{96.52}
\expandafter\gdef\csname R@r8.a.Y2.mean\endcsname{82.63}
\expandafter\gdef\csname R@r8.a.Y2.nseed\endcsname{5}
\expandafter\gdef\csname R@r8.a.Y2.tumour\endcsname{68.73}
\expandafter\gdef\csname R@r8.a.base.liver\endcsname{96.47}
\expandafter\gdef\csname R@r8.a.base.mean\endcsname{80.79}
\expandafter\gdef\csname R@r8.a.base.nseed\endcsname{3}
\expandafter\gdef\csname R@r8.a.base.tumour\endcsname{65.12}
\expandafter\gdef\csname R@r8.a.boot.h1v7a1.d\endcsname{\ensuremath{-1.35}}
\expandafter\gdef\csname R@r8.a.boot.h1v7a1.hi\endcsname{\ensuremath{+2.08}}
\expandafter\gdef\csname R@r8.a.boot.h1v7a1.lo\endcsname{\ensuremath{-5.73}}
\expandafter\gdef\csname R@r8.a.boot.h1v7a1.n\endcsname{26}
\expandafter\gdef\csname R@r8.a.boot.h1y2.d\endcsname{\ensuremath{-2.26}}
\expandafter\gdef\csname R@r8.a.boot.h1y2.hi\endcsname{\ensuremath{+0.14}}
\expandafter\gdef\csname R@r8.a.boot.h1y2.lo\endcsname{\ensuremath{-6.17}}
\expandafter\gdef\csname R@r8.a.boot.h1y2.n\endcsname{26}
\expandafter\gdef\csname R@r8.a.cross\endcsname{not observed}
\expandafter\gdef\csname R@r8.a.d.h1v7a1\endcsname{\ensuremath{-0.88}}
\expandafter\gdef\csname R@r8.a.d.h1y2\endcsname{\ensuremath{-1.21}}
\expandafter\gdef\csname R@r8.a.guard.liver\endcsname{-0.10}
\expandafter\gdef\csname R@r8.a.guard.mean\endcsname{-0.50}
\expandafter\gdef\csname R@r8.a.h1.liver\endcsname{96.52}
\expandafter\gdef\csname R@r8.a.h1.mean\endcsname{82.02}
\expandafter\gdef\csname R@r8.a.h1.nseed\endcsname{5}
\expandafter\gdef\csname R@r8.a.h1.tumour\endcsname{67.52}
\expandafter\gdef\csname R@r8.a.thr.a1\endcsname{3.19}
\expandafter\gdef\csname R@r8.a.thr.rep\endcsname{1.59}
\expandafter\gdef\csname R@r8.a.verdict\endcsname{FAIL}
\expandafter\gdef\csname R@r8.b.V7A1.both\endcsname{66.39}
\expandafter\gdef\csname R@r8.b.V7A1.cancer\endcsname{51.67}
\expandafter\gdef\csname R@r8.b.V7A1.organ\endcsname{81.10}
\expandafter\gdef\csname R@r8.b.base.both\endcsname{64.99}
\expandafter\gdef\csname R@r8.b.base.cancer\endcsname{49.12}
\expandafter\gdef\csname R@r8.b.base.organ\endcsname{80.87}
\expandafter\gdef\csname R@r8.b.boot.d\endcsname{\ensuremath{+2.56}}
\expandafter\gdef\csname R@r8.b.boot.hi\endcsname{\ensuremath{+5.35}}
\expandafter\gdef\csname R@r8.b.boot.lo\endcsname{\ensuremath{-0.14}}
\expandafter\gdef\csname R@r8.b.boot.n\endcsname{180}
\expandafter\gdef\csname R@r8.b.both.d\endcsname{\ensuremath{+1.40}}
\expandafter\gdef\csname R@r8.b.both.f1\endcsname{\ensuremath{+2.01}}
\expandafter\gdef\csname R@r8.b.both.f2\endcsname{\ensuremath{+3.06}}
\expandafter\gdef\csname R@r8.b.both.f3\endcsname{\ensuremath{+0.38}}
\expandafter\gdef\csname R@r8.b.both.f4\endcsname{\ensuremath{+0.13}}
\expandafter\gdef\csname R@r8.b.both.nup\endcsname{4}
\expandafter\gdef\csname R@r8.b.d.f1\endcsname{\ensuremath{+3.29}}
\expandafter\gdef\csname R@r8.b.d.f2\endcsname{\ensuremath{+5.85}}
\expandafter\gdef\csname R@r8.b.d.f3\endcsname{\ensuremath{+0.60}}
\expandafter\gdef\csname R@r8.b.d.f4\endcsname{\ensuremath{+0.48}}
\expandafter\gdef\csname R@r8.b.d.mean\endcsname{\ensuremath{+2.56}}
\expandafter\gdef\csname R@r8.b.guard.organ\endcsname{-1.00}
\expandafter\gdef\csname R@r8.b.nfold\endcsname{4}
\expandafter\gdef\csname R@r8.b.nup\endcsname{4}
\expandafter\gdef\csname R@r8.b.organ.d\endcsname{\ensuremath{+0.23}}
\expandafter\gdef\csname R@r8.b.organ.f1\endcsname{\ensuremath{+0.73}}
\expandafter\gdef\csname R@r8.b.organ.f2\endcsname{\ensuremath{+0.27}}
\expandafter\gdef\csname R@r8.b.organ.f3\endcsname{\ensuremath{+0.16}}
\expandafter\gdef\csname R@r8.b.organ.f4\endcsname{\ensuremath{-0.22}}
\expandafter\gdef\csname R@r8.b.organ.nup\endcsname{3}
\expandafter\gdef\csname R@r8.b.thr.folds\endcsname{3}
\expandafter\gdef\csname R@r8.b.verdict\endcsname{FAIL}
\expandafter\gdef\csname R@r8.b.verdict.prereg\endcsname{NOT CONFIRMED}
\expandafter\gdef\csname R@r8.boot.n\endcsname{10000}
\expandafter\gdef\csname R@r8.kits.h1.d\endcsname{\ensuremath{-0.743}}
\expandafter\gdef\csname R@r8.sd.pooled\endcsname{2.520}
\expandafter\gdef\csname R@r8.trig.delta\endcsname{10}
\expandafter\gdef\csname R@r8.trig.kits\endcsname{0.666}
\expandafter\gdef\csname R@r8.trig.kits.n\endcsname{440}
\expandafter\gdef\csname R@r8.trig.lits\endcsname{0.962}
\expandafter\gdef\csname R@r8.trig.lits.n\endcsname{745}
\expandafter\gdef\csname R@r8.trig.nds\endcsname{three}
\expandafter\gdef\csname R@r8.trig.panc\endcsname{0.504}
\expandafter\gdef\csname R@r8.trig.panc.n\endcsname{226}
\makeatother

\input{an_extra_rscost.tex}
\input{an_extra_s6.tex}
\makeatletter
\expandafter\gdef\csname R@an.sat.amos.L1.boundary.pp\endcsname{9.46}
\expandafter\gdef\csname R@an.sat.amos.L1.boundary.share\endcsname{87.7}
\expandafter\gdef\csname R@an.sat.amos.L1.deficit\endcsname{10.79}
\expandafter\gdef\csname R@an.sat.amos.L1.emptypair.pp\endcsname{0.46}
\expandafter\gdef\csname R@an.sat.amos.L1.nbelow\endcsname{43}
\expandafter\gdef\csname R@an.sat.amos.L1.wholeobject.pp\endcsname{1.33}
\expandafter\gdef\csname R@an.sat.amos.L1.wholeobject.share\endcsname{12.3}
\expandafter\gdef\csname R@an.sat.amos.V7A1.boundary.pp\endcsname{9.54}
\expandafter\gdef\csname R@an.sat.amos.V7A1.boundary.share\endcsname{86.8}
\expandafter\gdef\csname R@an.sat.amos.V7A1.deficit\endcsname{11.00}
\expandafter\gdef\csname R@an.sat.amos.V7A1.emptypair.pp\endcsname{0.39}
\expandafter\gdef\csname R@an.sat.amos.V7A1.nbelow\endcsname{48}
\expandafter\gdef\csname R@an.sat.amos.V7A1.wholeobject.pp\endcsname{1.46}
\expandafter\gdef\csname R@an.sat.amos.V7A1.wholeobject.share\endcsname{13.2}
\expandafter\gdef\csname R@an.sat.amos.V7C1.boundary.pp\endcsname{10.05}
\expandafter\gdef\csname R@an.sat.amos.V7C1.boundary.share\endcsname{87.0}
\expandafter\gdef\csname R@an.sat.amos.V7C1.deficit\endcsname{11.54}
\expandafter\gdef\csname R@an.sat.amos.V7C1.emptypair.pp\endcsname{0.57}
\expandafter\gdef\csname R@an.sat.amos.V7C1.nbelow\endcsname{49}
\expandafter\gdef\csname R@an.sat.amos.V7C1.wholeobject.pp\endcsname{1.50}
\expandafter\gdef\csname R@an.sat.amos.V7C1.wholeobject.share\endcsname{13.0}
\expandafter\gdef\csname R@an.sat.amos.Y2.boundary.pp\endcsname{9.56}
\expandafter\gdef\csname R@an.sat.amos.Y2.boundary.share\endcsname{87.5}
\expandafter\gdef\csname R@an.sat.amos.Y2.cut03.npairs\endcsname{21}
\expandafter\gdef\csname R@an.sat.amos.Y2.cut03.share\endcsname{7.2}
\expandafter\gdef\csname R@an.sat.amos.Y2.cut07.npairs\endcsname{137}
\expandafter\gdef\csname R@an.sat.amos.Y2.cut07.share\endcsname{25.6}
\expandafter\gdef\csname R@an.sat.amos.Y2.deficit\endcsname{10.93}
\expandafter\gdef\csname R@an.sat.amos.Y2.emptypair.pp\endcsname{0.50}
\expandafter\gdef\csname R@an.sat.amos.Y2.nbelow\endcsname{45}
\expandafter\gdef\csname R@an.sat.amos.Y2.wholeobject.pp\endcsname{1.37}
\expandafter\gdef\csname R@an.sat.amos.Y2.wholeobject.share\endcsname{12.5}
\expandafter\gdef\csname R@an.sat.amos.base.boundary.pp\endcsname{9.78}
\expandafter\gdef\csname R@an.sat.amos.base.boundary.share\endcsname{87.7}
\expandafter\gdef\csname R@an.sat.amos.base.deficit\endcsname{11.15}
\expandafter\gdef\csname R@an.sat.amos.base.emptypair.pp\endcsname{0.50}
\expandafter\gdef\csname R@an.sat.amos.base.nbelow\endcsname{43}
\expandafter\gdef\csname R@an.sat.amos.base.wholeobject.pp\endcsname{1.37}
\expandafter\gdef\csname R@an.sat.amos.base.wholeobject.share\endcsname{12.3}
\expandafter\gdef\csname R@an.sat.amos.boundary.share.max\endcsname{87.7}
\expandafter\gdef\csname R@an.sat.amos.boundary.share.min\endcsname{86.8}
\expandafter\gdef\csname R@an.sat.amos.cut\endcsname{0.5}
\expandafter\gdef\csname R@an.sat.amos.cut03\endcsname{0.3}
\expandafter\gdef\csname R@an.sat.amos.cut07\endcsname{0.7}
\expandafter\gdef\csname R@an.sat.amos.deficit.max\endcsname{11.54}
\expandafter\gdef\csname R@an.sat.amos.deficit.min\endcsname{10.79}
\expandafter\gdef\csname R@an.sat.amos.emptypair.max\endcsname{0.57}
\expandafter\gdef\csname R@an.sat.amos.emptypair.min\endcsname{0.39}
\expandafter\gdef\csname R@an.sat.amos.narms\endcsname{5}
\expandafter\gdef\csname R@an.sat.amos.narms.word\endcsname{five}
\expandafter\gdef\csname R@an.sat.amos.npairs\endcsname{2375}
\expandafter\gdef\csname R@an.sat.amos.wholeobject.share.max\endcsname{13.2}
\expandafter\gdef\csname R@an.sat.amos.wholeobject.share.min\endcsname{12.3}
\expandafter\gdef\csname R@an.sat.bin1.label\endcsname{1--1.25}
\expandafter\gdef\csname R@an.sat.bin1.n\endcsname{153}
\expandafter\gdef\csname R@an.sat.bin1.spread.mm\endcsname{0.040}
\expandafter\gdef\csname R@an.sat.bin1.spread.vox\endcsname{0.029}
\expandafter\gdef\csname R@an.sat.bin1.tmm\endcsname{1.55}
\expandafter\gdef\csname R@an.sat.bin2.label\endcsname{1.25--1.5}
\expandafter\gdef\csname R@an.sat.bin2.n\endcsname{155}
\expandafter\gdef\csname R@an.sat.bin2.spread.mm\endcsname{0.040}
\expandafter\gdef\csname R@an.sat.bin2.spread.vox\endcsname{0.030}
\expandafter\gdef\csname R@an.sat.bin2.tmm\endcsname{1.80}
\expandafter\gdef\csname R@an.sat.bin3.label\endcsname{1.5--2}
\expandafter\gdef\csname R@an.sat.bin3.n\endcsname{331}
\expandafter\gdef\csname R@an.sat.bin3.spread.mm\endcsname{0.048}
\expandafter\gdef\csname R@an.sat.bin3.spread.vox\endcsname{0.036}
\expandafter\gdef\csname R@an.sat.bin3.tmm\endcsname{2.36}
\expandafter\gdef\csname R@an.sat.bin4.label\endcsname{2--2.5}
\expandafter\gdef\csname R@an.sat.bin4.n\endcsname{295}
\expandafter\gdef\csname R@an.sat.bin4.spread.mm\endcsname{0.060}
\expandafter\gdef\csname R@an.sat.bin4.spread.vox\endcsname{0.044}
\expandafter\gdef\csname R@an.sat.bin4.tmm\endcsname{2.99}
\expandafter\gdef\csname R@an.sat.bin5.label\endcsname{2.5--4}
\expandafter\gdef\csname R@an.sat.bin5.n\endcsname{679}
\expandafter\gdef\csname R@an.sat.bin5.spread.mm\endcsname{0.043}
\expandafter\gdef\csname R@an.sat.bin5.spread.vox\endcsname{0.032}
\expandafter\gdef\csname R@an.sat.bin5.tmm\endcsname{4.36}
\expandafter\gdef\csname R@an.sat.bin6.label\endcsname{4--7}
\expandafter\gdef\csname R@an.sat.bin6.n\endcsname{586}
\expandafter\gdef\csname R@an.sat.bin6.spread.mm\endcsname{0.029}
\expandafter\gdef\csname R@an.sat.bin6.spread.vox\endcsname{0.022}
\expandafter\gdef\csname R@an.sat.bin6.tmm\endcsname{6.24}
\expandafter\gdef\csname R@an.sat.bin7.label\endcsname{$>$7}
\expandafter\gdef\csname R@an.sat.bin7.n\endcsname{133}
\expandafter\gdef\csname R@an.sat.bin7.spread.mm\endcsname{0.045}
\expandafter\gdef\csname R@an.sat.bin7.spread.vox\endcsname{0.048}
\expandafter\gdef\csname R@an.sat.bin7.tmm\endcsname{9.18}
\expandafter\gdef\csname R@an.sat.e.L1.b1.vox\endcsname{0.473}
\expandafter\gdef\csname R@an.sat.e.L1.b2.vox\endcsname{0.459}
\expandafter\gdef\csname R@an.sat.e.L1.b3.vox\endcsname{0.572}
\expandafter\gdef\csname R@an.sat.e.L1.b4.vox\endcsname{0.533}
\expandafter\gdef\csname R@an.sat.e.L1.b5.vox\endcsname{0.470}
\expandafter\gdef\csname R@an.sat.e.L1.b6.vox\endcsname{0.423}
\expandafter\gdef\csname R@an.sat.e.L1.b7.vox\endcsname{0.597}
\expandafter\gdef\csname R@an.sat.e.L1.mm\endcsname{0.605}
\expandafter\gdef\csname R@an.sat.e.L1.vox\endcsname{0.487}
\expandafter\gdef\csname R@an.sat.e.V7A1.b1.vox\endcsname{0.468}
\expandafter\gdef\csname R@an.sat.e.V7A1.b2.vox\endcsname{0.471}
\expandafter\gdef\csname R@an.sat.e.V7A1.b3.vox\endcsname{0.581}
\expandafter\gdef\csname R@an.sat.e.V7A1.b4.vox\endcsname{0.540}
\expandafter\gdef\csname R@an.sat.e.V7A1.b5.vox\endcsname{0.475}
\expandafter\gdef\csname R@an.sat.e.V7A1.b6.vox\endcsname{0.432}
\expandafter\gdef\csname R@an.sat.e.V7A1.b7.vox\endcsname{0.601}
\expandafter\gdef\csname R@an.sat.e.V7A1.mm\endcsname{0.614}
\expandafter\gdef\csname R@an.sat.e.V7A1.vox\endcsname{0.494}
\expandafter\gdef\csname R@an.sat.e.V7C1.b1.vox\endcsname{0.497}
\expandafter\gdef\csname R@an.sat.e.V7C1.b2.vox\endcsname{0.489}
\expandafter\gdef\csname R@an.sat.e.V7C1.b3.vox\endcsname{0.608}
\expandafter\gdef\csname R@an.sat.e.V7C1.b4.vox\endcsname{0.577}
\expandafter\gdef\csname R@an.sat.e.V7C1.b5.vox\endcsname{0.502}
\expandafter\gdef\csname R@an.sat.e.V7C1.b6.vox\endcsname{0.445}
\expandafter\gdef\csname R@an.sat.e.V7C1.b7.vox\endcsname{0.633}
\expandafter\gdef\csname R@an.sat.e.V7C1.mm\endcsname{0.646}
\expandafter\gdef\csname R@an.sat.e.V7C1.vox\endcsname{0.518}
\expandafter\gdef\csname R@an.sat.e.Y2.b1.vox\endcsname{0.470}
\expandafter\gdef\csname R@an.sat.e.Y2.b2.vox\endcsname{0.463}
\expandafter\gdef\csname R@an.sat.e.Y2.b3.vox\endcsname{0.582}
\expandafter\gdef\csname R@an.sat.e.Y2.b4.vox\endcsname{0.542}
\expandafter\gdef\csname R@an.sat.e.Y2.b5.vox\endcsname{0.476}
\expandafter\gdef\csname R@an.sat.e.Y2.b6.vox\endcsname{0.430}
\expandafter\gdef\csname R@an.sat.e.Y2.b7.vox\endcsname{0.611}
\expandafter\gdef\csname R@an.sat.e.Y2.mm\endcsname{0.614}
\expandafter\gdef\csname R@an.sat.e.Y2.vox\endcsname{0.494}
\expandafter\gdef\csname R@an.sat.e.base.b1.vox\endcsname{0.474}
\expandafter\gdef\csname R@an.sat.e.base.b2.vox\endcsname{0.478}
\expandafter\gdef\csname R@an.sat.e.base.b3.vox\endcsname{0.594}
\expandafter\gdef\csname R@an.sat.e.base.b4.vox\endcsname{0.553}
\expandafter\gdef\csname R@an.sat.e.base.b5.vox\endcsname{0.488}
\expandafter\gdef\csname R@an.sat.e.base.b6.vox\endcsname{0.423}
\expandafter\gdef\csname R@an.sat.e.base.b7.vox\endcsname{0.645}
\expandafter\gdef\csname R@an.sat.e.base.binmax\endcsname{0.65}
\expandafter\gdef\csname R@an.sat.e.base.binmin\endcsname{0.42}
\expandafter\gdef\csname R@an.sat.e.base.mm\endcsname{0.623}
\expandafter\gdef\csname R@an.sat.e.base.vox\endcsname{0.502}
\expandafter\gdef\csname R@an.sat.e.binwise.spread.mm\endcsname{0.060}
\expandafter\gdef\csname R@an.sat.e.binwise.spread.vox\endcsname{0.05}
\expandafter\gdef\csname R@an.sat.e.binwise.vox.max\endcsname{0.65}
\expandafter\gdef\csname R@an.sat.e.binwise.vox.min\endcsname{0.42}
\expandafter\gdef\csname R@an.sat.e.mm.max\endcsname{0.646}
\expandafter\gdef\csname R@an.sat.e.mm.min\endcsname{0.605}
\expandafter\gdef\csname R@an.sat.e.narms\endcsname{5}
\expandafter\gdef\csname R@an.sat.e.vox.max\endcsname{0.518}
\expandafter\gdef\csname R@an.sat.e.vox.min\endcsname{0.487}
\expandafter\gdef\csname R@an.sat.e.vox.spread\endcsname{0.031}
\expandafter\gdef\csname R@an.sat.grid.native.edge\endcsname{1.278}
\expandafter\gdef\csname R@an.sat.grid.native.xy\endcsname{0.689}
\expandafter\gdef\csname R@an.sat.grid.native.z\endcsname{5}
\expandafter\gdef\csname R@an.sat.grid.native.zmax\endcsname{5}
\expandafter\gdef\csname R@an.sat.grid.native.zmin\endcsname{1.25}
\expandafter\gdef\csname R@an.sat.grid.ncases\endcsname{160}
\expandafter\gdef\csname R@an.sat.grid.plan.xy\endcsname{0.685}
\expandafter\gdef\csname R@an.sat.grid.plan.z\endcsname{2}
\expandafter\gdef\csname R@an.sat.kits.L1.bound.dice\endcsname{86.31}
\expandafter\gdef\csname R@an.sat.kits.L1.deficit\endcsname{20.63}
\expandafter\gdef\csname R@an.sat.kits.L1.gross.dice\endcsname{60.95}
\expandafter\gdef\csname R@an.sat.kits.L1.gross.share\endcsname{51.8}
\expandafter\gdef\csname R@an.sat.kits.L1.tumour.dice\endcsname{79.36}
\expandafter\gdef\csname R@an.sat.kits.V7A1.bound.dice\endcsname{86.17}
\expandafter\gdef\csname R@an.sat.kits.V7A1.deficit\endcsname{21.16}
\expandafter\gdef\csname R@an.sat.kits.V7A1.gross.dice\endcsname{59.38}
\expandafter\gdef\csname R@an.sat.kits.V7A1.gross.share\endcsname{52.5}
\expandafter\gdef\csname R@an.sat.kits.V7A1.tumour.dice\endcsname{78.83}
\expandafter\gdef\csname R@an.sat.kits.Y2.bound.dice\endcsname{85.79}
\expandafter\gdef\csname R@an.sat.kits.Y2.deficit\endcsname{20.15}
\expandafter\gdef\csname R@an.sat.kits.Y2.gross.dice\endcsname{64.07}
\expandafter\gdef\csname R@an.sat.kits.Y2.gross.share\endcsname{48.8}
\expandafter\gdef\csname R@an.sat.kits.Y2.tumour.dice\endcsname{79.84}
\expandafter\gdef\csname R@an.sat.kits.base.bound.dice\endcsname{87.26}
\expandafter\gdef\csname R@an.sat.kits.base.deficit\endcsname{23.20}
\expandafter\gdef\csname R@an.sat.kits.base.gross.dice\endcsname{49.04}
\expandafter\gdef\csname R@an.sat.kits.base.gross.share\endcsname{60.1}
\expandafter\gdef\csname R@an.sat.kits.base.tumour.dice\endcsname{76.80}
\expandafter\gdef\csname R@an.sat.kits.bound.n\endcsname{284}
\expandafter\gdef\csname R@an.sat.kits.gross.n\endcsname{107}
\expandafter\gdef\csname R@an.sat.kits.layers.thr\endcsname{2}
\expandafter\gdef\csname R@an.sat.kits.ncases\endcsname{391}
\expandafter\gdef\csname R@an.sat.surf.Ladrenal.pct\endcsname{74.6}
\expandafter\gdef\csname R@an.sat.surf.Ladrenal.t\endcsname{1.39}
\expandafter\gdef\csname R@an.sat.surf.Ladrenal.tmm\endcsname{1.74}
\expandafter\gdef\csname R@an.sat.surf.Radrenal.pct\endcsname{78.2}
\expandafter\gdef\csname R@an.sat.surf.Radrenal.t\endcsname{1.32}
\expandafter\gdef\csname R@an.sat.surf.Radrenal.tmm\endcsname{1.66}
\expandafter\gdef\csname R@an.sat.surf.liver.pct\endcsname{15.3}
\expandafter\gdef\csname R@an.sat.surf.liver.t\endcsname{7.21}
\expandafter\gdef\csname R@an.sat.surf.liver.tmm\endcsname{8.88}
\makeatother

\input{an_extra_sc15.tex}
\input{an_extra_stockref.tex}
\makeatletter
\expandafter\gdef\csname R@sw.boot.n\endcsname{10000}
\expandafter\gdef\csname R@sw.guard.floor\endcsname{\ensuremath{-1.00}}
\expandafter\gdef\csname R@sw.heart.base.lesion\endcsname{93.29}
\expandafter\gdef\csname R@sw.heart.boot.d\endcsname{\ensuremath{-0.06}}
\expandafter\gdef\csname R@sw.heart.boot.excl\endcsname{0}
\expandafter\gdef\csname R@sw.heart.boot.hi\endcsname{\ensuremath{+0.15}}
\expandafter\gdef\csname R@sw.heart.boot.lo\endcsname{\ensuremath{-0.26}}
\expandafter\gdef\csname R@sw.heart.boot.n\endcsname{12}
\expandafter\gdef\csname R@sw.heart.cls\endcsname{left atrium}
\expandafter\gdef\csname R@sw.heart.d.f1\endcsname{\ensuremath{-0.18}}
\expandafter\gdef\csname R@sw.heart.d.f2\endcsname{\ensuremath{+0.07}}
\expandafter\gdef\csname R@sw.heart.d.f3\endcsname{\ensuremath{-0.16}}
\expandafter\gdef\csname R@sw.heart.d.f4\endcsname{\ensuremath{+0.01}}
\expandafter\gdef\csname R@sw.heart.d.mean\endcsname{\ensuremath{-0.06}}
\expandafter\gdef\csname R@sw.heart.lean.lesion\endcsname{93.23}
\expandafter\gdef\csname R@sw.heart.name\endcsname{MSD Heart}
\expandafter\gdef\csname R@sw.heart.nup\endcsname{2}
\expandafter\gdef\csname R@sw.heart.verdict\endcsname{not confirmed}
\expandafter\gdef\csname R@sw.n\endcsname{2}
\expandafter\gdef\csname R@sw.n.word\endcsname{two}
\expandafter\gdef\csname R@sw.n.word.cap\endcsname{Two}
\expandafter\gdef\csname R@sw.nfold\endcsname{4}
\expandafter\gdef\csname R@sw.ngen\endcsname{0}
\expandafter\gdef\csname R@sw.on\endcsname{1}
\expandafter\gdef\csname R@sw.rep.nallfold\endcsname{0}
\expandafter\gdef\csname R@sw.rep.nconf\endcsname{0}
\expandafter\gdef\csname R@sw.rep.nexcl\endcsname{0}
\expandafter\gdef\csname R@sw.rep.nexclneg\endcsname{0}
\expandafter\gdef\csname R@sw.rep.nexclpos\endcsname{0}
\expandafter\gdef\csname R@sw.rep.nlead\endcsname{1}
\expandafter\gdef\csname R@sw.rep.nrev\endcsname{0}
\expandafter\gdef\csname R@sw.spleen.base.lesion\endcsname{96.86}
\expandafter\gdef\csname R@sw.spleen.boot.d\endcsname{\ensuremath{+0.11}}
\expandafter\gdef\csname R@sw.spleen.boot.excl\endcsname{0}
\expandafter\gdef\csname R@sw.spleen.boot.hi\endcsname{\ensuremath{+0.60}}
\expandafter\gdef\csname R@sw.spleen.boot.lo\endcsname{\ensuremath{-0.50}}
\expandafter\gdef\csname R@sw.spleen.boot.n\endcsname{26}
\expandafter\gdef\csname R@sw.spleen.cls\endcsname{spleen}
\expandafter\gdef\csname R@sw.spleen.d.f1\endcsname{\ensuremath{+0.45}}
\expandafter\gdef\csname R@sw.spleen.d.f2\endcsname{\ensuremath{+0.40}}
\expandafter\gdef\csname R@sw.spleen.d.f3\endcsname{\ensuremath{-0.64}}
\expandafter\gdef\csname R@sw.spleen.d.f4\endcsname{\ensuremath{+0.15}}
\expandafter\gdef\csname R@sw.spleen.d.mean\endcsname{\ensuremath{+0.09}}
\expandafter\gdef\csname R@sw.spleen.lean.lesion\endcsname{96.95}
\expandafter\gdef\csname R@sw.spleen.name\endcsname{MSD Spleen}
\expandafter\gdef\csname R@sw.spleen.nup\endcsname{3}
\expandafter\gdef\csname R@sw.spleen.verdict\endcsname{not confirmed}
\makeatother

\makeatletter
\expandafter\gdef\csname R@an.trace.gpud.baseline.alloc\endcsname{21.6}
\expandafter\gdef\csname R@an.trace.gpud.baseline.dev\endcsname{15.0}
\expandafter\gdef\csname R@an.trace.gpud.baseline.n\endcsname{27}
\expandafter\gdef\csname R@an.trace.gpud.cv_design.alloc\endcsname{333.0}
\expandafter\gdef\csname R@an.trace.gpud.cv_design.dev\endcsname{163.1}
\expandafter\gdef\csname R@an.trace.gpud.cv_design.n\endcsname{349}
\expandafter\gdef\csname R@an.trace.gpud.cv_other.alloc\endcsname{117.9}
\expandafter\gdef\csname R@an.trace.gpud.cv_other.dev\endcsname{58.8}
\expandafter\gdef\csname R@an.trace.gpud.cv_other.n\endcsname{180}
\expandafter\gdef\csname R@an.trace.gpud.design.alloc\endcsname{139.9}
\expandafter\gdef\csname R@an.trace.gpud.design.dev\endcsname{68.3}
\expandafter\gdef\csname R@an.trace.gpud.design.n\endcsname{215}
\expandafter\gdef\csname R@an.trace.gpud.design.share.alloc\endcsname{17}
\expandafter\gdef\csname R@an.trace.gpud.design_late.alloc\endcsname{28.1}
\expandafter\gdef\csname R@an.trace.gpud.design_late.dev\endcsname{13.7}
\expandafter\gdef\csname R@an.trace.gpud.design_late.n\endcsname{32}
\expandafter\gdef\csname R@an.trace.gpud.designdef.alloc\endcsname{140}
\expandafter\gdef\csname R@an.trace.gpud.designdef.runs\endcsname{138}
\expandafter\gdef\csname R@an.trace.gpud.designdef.used\endcsname{68}
\expandafter\gdef\csname R@an.trace.gpud.earlier.alloc\endcsname{14.5}
\expandafter\gdef\csname R@an.trace.gpud.heldout.alloc\endcsname{49.9}
\expandafter\gdef\csname R@an.trace.gpud.heldout.dev\endcsname{24.8}
\expandafter\gdef\csname R@an.trace.gpud.heldout.n\endcsname{58}
\expandafter\gdef\csname R@an.trace.gpud.inflight.alloc\endcsname{0.0}
\expandafter\gdef\csname R@an.trace.gpud.inflight.n\endcsname{0}
\expandafter\gdef\csname R@an.trace.gpud.judge.alloc\endcsname{32.7}
\expandafter\gdef\csname R@an.trace.gpud.judge.dev\endcsname{15.9}
\expandafter\gdef\csname R@an.trace.gpud.judge.n\endcsname{35}
\expandafter\gdef\csname R@an.trace.gpud.lead.controls_ablations.alloc\endcsname{121.0}
\expandafter\gdef\csname R@an.trace.gpud.lead.controls_ablations.dev\endcsname{63.0}
\expandafter\gdef\csname R@an.trace.gpud.lead.controls_ablations.n\endcsname{102}
\expandafter\gdef\csname R@an.trace.gpud.lead.cv.alloc\endcsname{333.0}
\expandafter\gdef\csname R@an.trace.gpud.lead.cv.dev\endcsname{163.1}
\expandafter\gdef\csname R@an.trace.gpud.lead.cv.n\endcsname{349}
\expandafter\gdef\csname R@an.trace.gpud.lead.design_waves.alloc\endcsname{167.9}
\expandafter\gdef\csname R@an.trace.gpud.lead.design_waves.dev\endcsname{81.9}
\expandafter\gdef\csname R@an.trace.gpud.lead.design_waves.n\endcsname{247}
\expandafter\gdef\csname R@an.trace.gpud.lead.heldout_and_unseen.alloc\endcsname{167.8}
\expandafter\gdef\csname R@an.trace.gpud.lead.heldout_and_unseen.dev\endcsname{83.6}
\expandafter\gdef\csname R@an.trace.gpud.lead.heldout_and_unseen.n\endcsname{238}
\expandafter\gdef\csname R@an.trace.gpud.lead.locked.alloc\endcsname{24.1}
\expandafter\gdef\csname R@an.trace.gpud.lead.locked.dev\endcsname{11.8}
\expandafter\gdef\csname R@an.trace.gpud.lead.locked.n\endcsname{24}
\expandafter\gdef\csname R@an.trace.gpud.lead.measurement.alloc\endcsname{14.0}
\expandafter\gdef\csname R@an.trace.gpud.lead.measurement.dev\endcsname{14.0}
\expandafter\gdef\csname R@an.trace.gpud.lead.measurement.n\endcsname{1045}
\expandafter\gdef\csname R@an.trace.gpud.locked.alloc\endcsname{24.1}
\expandafter\gdef\csname R@an.trace.gpud.locked.dev\endcsname{11.8}
\expandafter\gdef\csname R@an.trace.gpud.locked.n\endcsname{24}
\expandafter\gdef\csname R@an.trace.gpud.measurement.alloc\endcsname{14.0}
\expandafter\gdef\csname R@an.trace.gpud.measurement.dev\endcsname{14.0}
\expandafter\gdef\csname R@an.trace.gpud.measurement.n\endcsname{1045}
\expandafter\gdef\csname R@an.trace.gpud.queued.n\endcsname{0}
\expandafter\gdef\csname R@an.trace.gpud.rs_llm.alloc\endcsname{66.7}
\expandafter\gdef\csname R@an.trace.gpud.rs_llm.dev\endcsname{32.1}
\expandafter\gdef\csname R@an.trace.gpud.rs_llm.n\endcsname{40}
\expandafter\gdef\csname R@an.trace.gpud.searchcost.fold5.n\endcsname{14}
\expandafter\gdef\csname R@an.trace.gpud.searchcost.now.alloc\endcsname{155}
\expandafter\gdef\csname R@an.trace.gpud.searchcost.now.runs\endcsname{152}
\expandafter\gdef\csname R@an.trace.gpud.searchcost.now.used\endcsname{75}
\expandafter\gdef\csname R@an.trace.gpud.total.alloc\endcsname{828}
\expandafter\gdef\csname R@an.trace.gpud.total.dev\endcsname{417}
\expandafter\gdef\csname R@an.trace.gpud.total.n\endcsname{2005}
\expandafter\gdef\csname R@an.trace.gpud.total.nfail\endcsname{98}
\expandafter\gdef\csname R@an.trace.on\endcsname{1}
\expandafter\gdef\csname R@an.trace.reg.afterresult\endcsname{none}
\expandafter\gdef\csname R@an.trace.reg.n\endcsname{48}
\expandafter\gdef\csname R@an.trace.reg.nafterresult\endcsname{0}
\expandafter\gdef\csname R@an.trace.reg.npredates\endcsname{10}
\expandafter\gdef\csname R@an.trace.reg.predates\endcsname{w1, rx, r4cv, r8, r9, r15, r16, r23s, r23a, r31s}
\expandafter\gdef\csname R@an.trace.reg.r10.beforeresult\endcsname{yes}
\expandafter\gdef\csname R@an.trace.reg.r10.file\endcsname{PREREG\_R10\_20261017.md}
\expandafter\gdef\csname R@an.trace.reg.r10.firstresult\endcsname{10-18 16:46}
\expandafter\gdef\csname R@an.trace.reg.r10.firstrun\endcsname{10-18 06:28}
\expandafter\gdef\csname R@an.trace.reg.r10.lagh\endcsname{+18.4}
\expandafter\gdef\csname R@an.trace.reg.r10.ledger\endcsname{--}
\expandafter\gdef\csname R@an.trace.reg.r10.reg\endcsname{10-17 12:02}
\expandafter\gdef\csname R@an.trace.reg.r10.src\endcsname{vcs}
\expandafter\gdef\csname R@an.trace.reg.r10.verdict\endcsname{OK}
\expandafter\gdef\csname R@an.trace.reg.r11.file\endcsname{PREREG\_R11\_20261017.md}
\expandafter\gdef\csname R@an.trace.reg.r11.firstrun\endcsname{--}
\expandafter\gdef\csname R@an.trace.reg.r11.ledger\endcsname{--}
\expandafter\gdef\csname R@an.trace.reg.r11.reg\endcsname{10-17 12:47}
\expandafter\gdef\csname R@an.trace.reg.r11.src\endcsname{vcs}
\expandafter\gdef\csname R@an.trace.reg.r11.verdict\endcsname{NO TRAINING RUNS}
\expandafter\gdef\csname R@an.trace.reg.r12.beforeresult\endcsname{yes}
\expandafter\gdef\csname R@an.trace.reg.r12.file\endcsname{PREREG\_R12\_20261019.md}
\expandafter\gdef\csname R@an.trace.reg.r12.firstresult\endcsname{10-20 02:55}
\expandafter\gdef\csname R@an.trace.reg.r12.firstrun\endcsname{10-19 15:11}
\expandafter\gdef\csname R@an.trace.reg.r12.lagh\endcsname{+1.3}
\expandafter\gdef\csname R@an.trace.reg.r12.ledger\endcsname{--}
\expandafter\gdef\csname R@an.trace.reg.r12.reg\endcsname{10-19 13:54}
\expandafter\gdef\csname R@an.trace.reg.r12.src\endcsname{vcs}
\expandafter\gdef\csname R@an.trace.reg.r12.verdict\endcsname{OK}
\expandafter\gdef\csname R@an.trace.reg.r13.beforeresult\endcsname{yes}
\expandafter\gdef\csname R@an.trace.reg.r13.file\endcsname{PREREG\_R13\_20261019.md}
\expandafter\gdef\csname R@an.trace.reg.r13.firstresult\endcsname{10-20 00:15}
\expandafter\gdef\csname R@an.trace.reg.r13.firstrun\endcsname{10-19 15:11}
\expandafter\gdef\csname R@an.trace.reg.r13.lagh\endcsname{+0.3}
\expandafter\gdef\csname R@an.trace.reg.r13.ledger\endcsname{--}
\expandafter\gdef\csname R@an.trace.reg.r13.reg\endcsname{10-19 14:55}
\expandafter\gdef\csname R@an.trace.reg.r13.src\endcsname{vcs}
\expandafter\gdef\csname R@an.trace.reg.r13.verdict\endcsname{OK}
\expandafter\gdef\csname R@an.trace.reg.r14.beforeresult\endcsname{yes}
\expandafter\gdef\csname R@an.trace.reg.r14.file\endcsname{PREREG\_R14\_20261019.md}
\expandafter\gdef\csname R@an.trace.reg.r14.firstresult\endcsname{10-19 22:30}
\expandafter\gdef\csname R@an.trace.reg.r14.firstrun\endcsname{10-19 15:11}
\expandafter\gdef\csname R@an.trace.reg.r14.lagh\endcsname{+0.3}
\expandafter\gdef\csname R@an.trace.reg.r14.ledger\endcsname{--}
\expandafter\gdef\csname R@an.trace.reg.r14.reg\endcsname{10-19 14:55}
\expandafter\gdef\csname R@an.trace.reg.r14.src\endcsname{vcs}
\expandafter\gdef\csname R@an.trace.reg.r14.verdict\endcsname{OK}
\expandafter\gdef\csname R@an.trace.reg.r15.beforeresult\endcsname{yes}
\expandafter\gdef\csname R@an.trace.reg.r15.file\endcsname{PREREG\_R15\_20261020.md}
\expandafter\gdef\csname R@an.trace.reg.r15.firstresult\endcsname{10-20 13:31}
\expandafter\gdef\csname R@an.trace.reg.r15.firstrun\endcsname{10-20 06:33}
\expandafter\gdef\csname R@an.trace.reg.r15.lagh\endcsname{-29.2}
\expandafter\gdef\csname R@an.trace.reg.r15.ledger\endcsname{--}
\expandafter\gdef\csname R@an.trace.reg.r15.reg\endcsname{10-21 11:45}
\expandafter\gdef\csname R@an.trace.reg.r15.src\endcsname{vcs}
\expandafter\gdef\csname R@an.trace.reg.r15.verdict\endcsname{NOT SHOWN}
\expandafter\gdef\csname R@an.trace.reg.r16.beforeresult\endcsname{yes}
\expandafter\gdef\csname R@an.trace.reg.r16.file\endcsname{PREREG\_R16\_20261020.md}
\expandafter\gdef\csname R@an.trace.reg.r16.firstresult\endcsname{10-20 21:08}
\expandafter\gdef\csname R@an.trace.reg.r16.firstrun\endcsname{10-20 11:34}
\expandafter\gdef\csname R@an.trace.reg.r16.lagh\endcsname{-24.2}
\expandafter\gdef\csname R@an.trace.reg.r16.ledger\endcsname{--}
\expandafter\gdef\csname R@an.trace.reg.r16.reg\endcsname{10-21 11:45}
\expandafter\gdef\csname R@an.trace.reg.r16.src\endcsname{vcs}
\expandafter\gdef\csname R@an.trace.reg.r16.verdict\endcsname{NOT SHOWN}
\expandafter\gdef\csname R@an.trace.reg.r17.beforeresult\endcsname{yes}
\expandafter\gdef\csname R@an.trace.reg.r17.file\endcsname{PREREG\_R17\_20261020.md}
\expandafter\gdef\csname R@an.trace.reg.r17.firstresult\endcsname{10-21 20:47}
\expandafter\gdef\csname R@an.trace.reg.r17.firstrun\endcsname{10-21 11:49}
\expandafter\gdef\csname R@an.trace.reg.r17.lagh\endcsname{+0.1}
\expandafter\gdef\csname R@an.trace.reg.r17.ledger\endcsname{--}
\expandafter\gdef\csname R@an.trace.reg.r17.reg\endcsname{10-21 11:45}
\expandafter\gdef\csname R@an.trace.reg.r17.src\endcsname{vcs}
\expandafter\gdef\csname R@an.trace.reg.r17.verdict\endcsname{OK}
\expandafter\gdef\csname R@an.trace.reg.r18.beforeresult\endcsname{yes}
\expandafter\gdef\csname R@an.trace.reg.r18.file\endcsname{PREREG\_R18\_20261021.md}
\expandafter\gdef\csname R@an.trace.reg.r18.firstresult\endcsname{10-22 06:52}
\expandafter\gdef\csname R@an.trace.reg.r18.firstrun\endcsname{10-21 21:04}
\expandafter\gdef\csname R@an.trace.reg.r18.lagh\endcsname{+9.0}
\expandafter\gdef\csname R@an.trace.reg.r18.ledger\endcsname{--}
\expandafter\gdef\csname R@an.trace.reg.r18.reg\endcsname{10-21 12:04}
\expandafter\gdef\csname R@an.trace.reg.r18.src\endcsname{vcs}
\expandafter\gdef\csname R@an.trace.reg.r18.verdict\endcsname{OK}
\expandafter\gdef\csname R@an.trace.reg.r19.beforeresult\endcsname{yes}
\expandafter\gdef\csname R@an.trace.reg.r19.file\endcsname{PREREG\_R19\_20261021.md}
\expandafter\gdef\csname R@an.trace.reg.r19.firstresult\endcsname{10-21 21:04}
\expandafter\gdef\csname R@an.trace.reg.r19.firstrun\endcsname{10-21 12:24}
\expandafter\gdef\csname R@an.trace.reg.r19.lagh\endcsname{+0.3}
\expandafter\gdef\csname R@an.trace.reg.r19.ledger\endcsname{--}
\expandafter\gdef\csname R@an.trace.reg.r19.reg\endcsname{10-21 12:04}
\expandafter\gdef\csname R@an.trace.reg.r19.src\endcsname{vcs}
\expandafter\gdef\csname R@an.trace.reg.r19.verdict\endcsname{OK}
\expandafter\gdef\csname R@an.trace.reg.r20.beforeresult\endcsname{yes}
\expandafter\gdef\csname R@an.trace.reg.r20.file\endcsname{PREREG\_R20\_20261021.md}
\expandafter\gdef\csname R@an.trace.reg.r20.firstresult\endcsname{10-22 06:50}
\expandafter\gdef\csname R@an.trace.reg.r20.firstrun\endcsname{10-21 22:30}
\expandafter\gdef\csname R@an.trace.reg.r20.lagh\endcsname{-0.4}
\expandafter\gdef\csname R@an.trace.reg.r20.ledger\endcsname{--}
\expandafter\gdef\csname R@an.trace.reg.r20.reg\endcsname{10-21 22:51}
\expandafter\gdef\csname R@an.trace.reg.r20.src\endcsname{vcs}
\expandafter\gdef\csname R@an.trace.reg.r20.verdict\endcsname{OK (mtime)}
\expandafter\gdef\csname R@an.trace.reg.r21.file\endcsname{PREREG\_R21\_20261022.md}
\expandafter\gdef\csname R@an.trace.reg.r21.firstrun\endcsname{--}
\expandafter\gdef\csname R@an.trace.reg.r21.ledger\endcsname{--}
\expandafter\gdef\csname R@an.trace.reg.r21.reg\endcsname{10-22 09:52}
\expandafter\gdef\csname R@an.trace.reg.r21.src\endcsname{vcs}
\expandafter\gdef\csname R@an.trace.reg.r21.verdict\endcsname{NO TRAINING RUNS}
\expandafter\gdef\csname R@an.trace.reg.r22.beforeresult\endcsname{yes}
\expandafter\gdef\csname R@an.trace.reg.r22.file\endcsname{PREREG\_R22\_20261022.md}
\expandafter\gdef\csname R@an.trace.reg.r22.firstresult\endcsname{10-23 00:15}
\expandafter\gdef\csname R@an.trace.reg.r22.firstrun\endcsname{10-22 12:13}
\expandafter\gdef\csname R@an.trace.reg.r22.lagh\endcsname{+0.2}
\expandafter\gdef\csname R@an.trace.reg.r22.ledger\endcsname{--}
\expandafter\gdef\csname R@an.trace.reg.r22.reg\endcsname{10-22 12:02}
\expandafter\gdef\csname R@an.trace.reg.r22.src\endcsname{vcs}
\expandafter\gdef\csname R@an.trace.reg.r22.verdict\endcsname{OK}
\expandafter\gdef\csname R@an.trace.reg.r23.beforeresult\endcsname{yes}
\expandafter\gdef\csname R@an.trace.reg.r23.file\endcsname{PREREG\_R23\_20261022.md}
\expandafter\gdef\csname R@an.trace.reg.r23.firstresult\endcsname{10-25 02:27}
\expandafter\gdef\csname R@an.trace.reg.r23.firstrun\endcsname{10-23 00:10}
\expandafter\gdef\csname R@an.trace.reg.r23.lagh\endcsname{+0.3}
\expandafter\gdef\csname R@an.trace.reg.r23.ledger\endcsname{--}
\expandafter\gdef\csname R@an.trace.reg.r23.reg\endcsname{10-22 23:49}
\expandafter\gdef\csname R@an.trace.reg.r23.src\endcsname{vcs}
\expandafter\gdef\csname R@an.trace.reg.r23.verdict\endcsname{OK}
\expandafter\gdef\csname R@an.trace.reg.r23a.beforeresult\endcsname{yes}
\expandafter\gdef\csname R@an.trace.reg.r23a.file\endcsname{PREREG\_R23\_amendment\_s4\_20261024.md}
\expandafter\gdef\csname R@an.trace.reg.r23a.firstresult\endcsname{10-25 02:27}
\expandafter\gdef\csname R@an.trace.reg.r23a.firstrun\endcsname{10-23 00:10}
\expandafter\gdef\csname R@an.trace.reg.r23a.lagh\endcsname{-25.2}
\expandafter\gdef\csname R@an.trace.reg.r23a.ledger\endcsname{--}
\expandafter\gdef\csname R@an.trace.reg.r23a.reg\endcsname{10-24 01:24}
\expandafter\gdef\csname R@an.trace.reg.r23a.src\endcsname{vcs}
\expandafter\gdef\csname R@an.trace.reg.r23a.verdict\endcsname{PREDATES}
\expandafter\gdef\csname R@an.trace.reg.r23s.beforeresult\endcsname{yes}
\expandafter\gdef\csname R@an.trace.reg.r23s.file\endcsname{PREREG\_R23\_scorer\_20261024.md}
\expandafter\gdef\csname R@an.trace.reg.r23s.firstresult\endcsname{10-25 02:27}
\expandafter\gdef\csname R@an.trace.reg.r23s.firstrun\endcsname{10-23 00:10}
\expandafter\gdef\csname R@an.trace.reg.r23s.lagh\endcsname{-25.2}
\expandafter\gdef\csname R@an.trace.reg.r23s.ledger\endcsname{--}
\expandafter\gdef\csname R@an.trace.reg.r23s.reg\endcsname{10-24 01:24}
\expandafter\gdef\csname R@an.trace.reg.r23s.src\endcsname{vcs}
\expandafter\gdef\csname R@an.trace.reg.r23s.verdict\endcsname{PREDATES}
\expandafter\gdef\csname R@an.trace.reg.r24.beforeresult\endcsname{yes}
\expandafter\gdef\csname R@an.trace.reg.r24.file\endcsname{PREREG\_R24\_20261023.md}
\expandafter\gdef\csname R@an.trace.reg.r24.firstresult\endcsname{10-23 18:18}
\expandafter\gdef\csname R@an.trace.reg.r24.firstrun\endcsname{10-23 02:51}
\expandafter\gdef\csname R@an.trace.reg.r24.lagh\endcsname{-0.1}
\expandafter\gdef\csname R@an.trace.reg.r24.ledger\endcsname{--}
\expandafter\gdef\csname R@an.trace.reg.r24.reg\endcsname{10-23 02:56}
\expandafter\gdef\csname R@an.trace.reg.r24.src\endcsname{vcs}
\expandafter\gdef\csname R@an.trace.reg.r24.verdict\endcsname{OK (mtime)}
\expandafter\gdef\csname R@an.trace.reg.r26.beforeresult\endcsname{yes}
\expandafter\gdef\csname R@an.trace.reg.r26.file\endcsname{PREREG\_R26\_20261023.md}
\expandafter\gdef\csname R@an.trace.reg.r26.firstresult\endcsname{10-24 04:36}
\expandafter\gdef\csname R@an.trace.reg.r26.firstrun\endcsname{10-23 20:14}
\expandafter\gdef\csname R@an.trace.reg.r26.lagh\endcsname{+3.2}
\expandafter\gdef\csname R@an.trace.reg.r26.ledger\endcsname{--}
\expandafter\gdef\csname R@an.trace.reg.r26.reg\endcsname{10-23 16:59}
\expandafter\gdef\csname R@an.trace.reg.r26.src\endcsname{vcs}
\expandafter\gdef\csname R@an.trace.reg.r26.verdict\endcsname{OK}
\expandafter\gdef\csname R@an.trace.reg.r27.beforeresult\endcsname{yes}
\expandafter\gdef\csname R@an.trace.reg.r27.file\endcsname{PREREG\_R27\_20261023.md}
\expandafter\gdef\csname R@an.trace.reg.r27.firstresult\endcsname{10-23 19:17}
\expandafter\gdef\csname R@an.trace.reg.r27.firstrun\endcsname{10-23 17:20}
\expandafter\gdef\csname R@an.trace.reg.r27.lagh\endcsname{+0.1}
\expandafter\gdef\csname R@an.trace.reg.r27.ledger\endcsname{--}
\expandafter\gdef\csname R@an.trace.reg.r27.reg\endcsname{10-23 17:14}
\expandafter\gdef\csname R@an.trace.reg.r27.src\endcsname{vcs}
\expandafter\gdef\csname R@an.trace.reg.r27.verdict\endcsname{OK}
\expandafter\gdef\csname R@an.trace.reg.r30.beforeresult\endcsname{yes}
\expandafter\gdef\csname R@an.trace.reg.r30.file\endcsname{PREREG\_R30\_20261024.md}
\expandafter\gdef\csname R@an.trace.reg.r30.firstresult\endcsname{10-25 08:01}
\expandafter\gdef\csname R@an.trace.reg.r30.firstrun\endcsname{10-24 23:29}
\expandafter\gdef\csname R@an.trace.reg.r30.lagh\endcsname{+3.0}
\expandafter\gdef\csname R@an.trace.reg.r30.ledger\endcsname{--}
\expandafter\gdef\csname R@an.trace.reg.r30.reg\endcsname{10-24 20:30}
\expandafter\gdef\csname R@an.trace.reg.r30.src\endcsname{vcs}
\expandafter\gdef\csname R@an.trace.reg.r30.verdict\endcsname{OK}
\expandafter\gdef\csname R@an.trace.reg.r30h.beforeresult\endcsname{yes}
\expandafter\gdef\csname R@an.trace.reg.r30h.file\endcsname{PREREG\_R30\_addendum\_history\_20261024.md}
\expandafter\gdef\csname R@an.trace.reg.r30h.firstresult\endcsname{10-25 08:01}
\expandafter\gdef\csname R@an.trace.reg.r30h.firstrun\endcsname{10-24 23:29}
\expandafter\gdef\csname R@an.trace.reg.r30h.lagh\endcsname{+3.0}
\expandafter\gdef\csname R@an.trace.reg.r30h.ledger\endcsname{--}
\expandafter\gdef\csname R@an.trace.reg.r30h.reg\endcsname{10-24 20:30}
\expandafter\gdef\csname R@an.trace.reg.r30h.src\endcsname{vcs}
\expandafter\gdef\csname R@an.trace.reg.r30h.verdict\endcsname{OK}
\expandafter\gdef\csname R@an.trace.reg.r31.beforeresult\endcsname{yes}
\expandafter\gdef\csname R@an.trace.reg.r31.file\endcsname{PREREG\_R31\_20261025.md}
\expandafter\gdef\csname R@an.trace.reg.r31.firstresult\endcsname{10-25 12:49}
\expandafter\gdef\csname R@an.trace.reg.r31.firstrun\endcsname{10-25 04:24}
\expandafter\gdef\csname R@an.trace.reg.r31.lagh\endcsname{-0.2}
\expandafter\gdef\csname R@an.trace.reg.r31.ledger\endcsname{--}
\expandafter\gdef\csname R@an.trace.reg.r31.reg\endcsname{10-25 04:34}
\expandafter\gdef\csname R@an.trace.reg.r31.src\endcsname{vcs}
\expandafter\gdef\csname R@an.trace.reg.r31.verdict\endcsname{OK (mtime)}
\expandafter\gdef\csname R@an.trace.reg.r31s.beforeresult\endcsname{yes}
\expandafter\gdef\csname R@an.trace.reg.r31s.file\endcsname{PREREG\_R31\_addendum\_sha\_20261025.md}
\expandafter\gdef\csname R@an.trace.reg.r31s.firstresult\endcsname{10-25 12:49}
\expandafter\gdef\csname R@an.trace.reg.r31s.firstrun\endcsname{10-25 04:24}
\expandafter\gdef\csname R@an.trace.reg.r31s.lagh\endcsname{-164.5}
\expandafter\gdef\csname R@an.trace.reg.r31s.ledger\endcsname{--}
\expandafter\gdef\csname R@an.trace.reg.r31s.reg\endcsname{11-01 00:54}
\expandafter\gdef\csname R@an.trace.reg.r31s.src\endcsname{vcs}
\expandafter\gdef\csname R@an.trace.reg.r31s.verdict\endcsname{PREDATES}
\expandafter\gdef\csname R@an.trace.reg.r32.beforeresult\endcsname{yes}
\expandafter\gdef\csname R@an.trace.reg.r32.file\endcsname{PREREG\_R32\_20261101.md}
\expandafter\gdef\csname R@an.trace.reg.r32.firstresult\endcsname{11-01 12:40}
\expandafter\gdef\csname R@an.trace.reg.r32.firstrun\endcsname{11-01 09:22}
\expandafter\gdef\csname R@an.trace.reg.r32.lagh\endcsname{+6.8}
\expandafter\gdef\csname R@an.trace.reg.r32.ledger\endcsname{--}
\expandafter\gdef\csname R@an.trace.reg.r32.reg\endcsname{11-01 02:30}
\expandafter\gdef\csname R@an.trace.reg.r32.src\endcsname{vcs}
\expandafter\gdef\csname R@an.trace.reg.r32.verdict\endcsname{OK}
\expandafter\gdef\csname R@an.trace.reg.r32s.beforeresult\endcsname{yes}
\expandafter\gdef\csname R@an.trace.reg.r32s.file\endcsname{PREREG\_R32\_addendum\_sha\_20261101.md}
\expandafter\gdef\csname R@an.trace.reg.r32s.firstresult\endcsname{11-01 12:40}
\expandafter\gdef\csname R@an.trace.reg.r32s.firstrun\endcsname{11-01 09:22}
\expandafter\gdef\csname R@an.trace.reg.r32s.lagh\endcsname{+6.0}
\expandafter\gdef\csname R@an.trace.reg.r32s.ledger\endcsname{--}
\expandafter\gdef\csname R@an.trace.reg.r32s.reg\endcsname{11-01 03:23}
\expandafter\gdef\csname R@an.trace.reg.r32s.src\endcsname{vcs}
\expandafter\gdef\csname R@an.trace.reg.r32s.verdict\endcsname{OK}
\expandafter\gdef\csname R@an.trace.reg.r33.beforeresult\endcsname{yes}
\expandafter\gdef\csname R@an.trace.reg.r33.file\endcsname{PREREG\_R33\_20261101.md}
\expandafter\gdef\csname R@an.trace.reg.r33.firstresult\endcsname{11-02 03:13}
\expandafter\gdef\csname R@an.trace.reg.r33.firstrun\endcsname{11-01 17:46}
\expandafter\gdef\csname R@an.trace.reg.r33.lagh\endcsname{+1.1}
\expandafter\gdef\csname R@an.trace.reg.r33.ledger\endcsname{--}
\expandafter\gdef\csname R@an.trace.reg.r33.reg\endcsname{11-01 16:38}
\expandafter\gdef\csname R@an.trace.reg.r33.src\endcsname{vcs}
\expandafter\gdef\csname R@an.trace.reg.r33.verdict\endcsname{OK}
\expandafter\gdef\csname R@an.trace.reg.r33a.beforeresult\endcsname{yes}
\expandafter\gdef\csname R@an.trace.reg.r33a.file\endcsname{PREREG\_R33\_amendment\_schedule\_acdc\_20261101.md}
\expandafter\gdef\csname R@an.trace.reg.r33a.firstresult\endcsname{11-02 03:13}
\expandafter\gdef\csname R@an.trace.reg.r33a.firstrun\endcsname{11-01 17:46}
\expandafter\gdef\csname R@an.trace.reg.r33a.lagh\endcsname{+0.4}
\expandafter\gdef\csname R@an.trace.reg.r33a.ledger\endcsname{--}
\expandafter\gdef\csname R@an.trace.reg.r33a.reg\endcsname{11-01 17:22}
\expandafter\gdef\csname R@an.trace.reg.r33a.src\endcsname{vcs}
\expandafter\gdef\csname R@an.trace.reg.r33a.verdict\endcsname{OK}
\expandafter\gdef\csname R@an.trace.reg.r34.beforeresult\endcsname{yes}
\expandafter\gdef\csname R@an.trace.reg.r34.file\endcsname{PREREG\_R34\_20261101.md}
\expandafter\gdef\csname R@an.trace.reg.r34.firstresult\endcsname{11-02 02:20}
\expandafter\gdef\csname R@an.trace.reg.r34.firstrun\endcsname{11-01 17:02}
\expandafter\gdef\csname R@an.trace.reg.r34.lagh\endcsname{+0.4}
\expandafter\gdef\csname R@an.trace.reg.r34.ledger\endcsname{--}
\expandafter\gdef\csname R@an.trace.reg.r34.reg\endcsname{11-01 16:38}
\expandafter\gdef\csname R@an.trace.reg.r34.src\endcsname{vcs}
\expandafter\gdef\csname R@an.trace.reg.r34.verdict\endcsname{OK}
\expandafter\gdef\csname R@an.trace.reg.r35.beforeresult\endcsname{yes}
\expandafter\gdef\csname R@an.trace.reg.r35.file\endcsname{PREREG\_R35\_20261102.md}
\expandafter\gdef\csname R@an.trace.reg.r35.firstresult\endcsname{11-02 21:52}
\expandafter\gdef\csname R@an.trace.reg.r35.firstrun\endcsname{--}
\expandafter\gdef\csname R@an.trace.reg.r35.ledger\endcsname{--}
\expandafter\gdef\csname R@an.trace.reg.r35.reg\endcsname{11-02 22:01}
\expandafter\gdef\csname R@an.trace.reg.r35.src\endcsname{vcs}
\expandafter\gdef\csname R@an.trace.reg.r35.verdict\endcsname{NO RUN STARTED YET}
\expandafter\gdef\csname R@an.trace.reg.r35s.beforeresult\endcsname{not shown}
\expandafter\gdef\csname R@an.trace.reg.r35s.file\endcsname{PREREG\_R35\_addendum\_sha\_20261102.md}
\expandafter\gdef\csname R@an.trace.reg.r35s.firstresult\endcsname{11-02 21:52}
\expandafter\gdef\csname R@an.trace.reg.r35s.firstrun\endcsname{--}
\expandafter\gdef\csname R@an.trace.reg.r35s.ledger\endcsname{--}
\expandafter\gdef\csname R@an.trace.reg.r35s.reg\endcsname{11-02 21:58}
\expandafter\gdef\csname R@an.trace.reg.r35s.src\endcsname{mtime}
\expandafter\gdef\csname R@an.trace.reg.r35s.verdict\endcsname{NO RUN STARTED YET}
\expandafter\gdef\csname R@an.trace.reg.r4cv.beforeresult\endcsname{yes}
\expandafter\gdef\csname R@an.trace.reg.r4cv.file\endcsname{PREREG\_D1D2\_20261014.md}
\expandafter\gdef\csname R@an.trace.reg.r4cv.firstresult\endcsname{10-14 21:36}
\expandafter\gdef\csname R@an.trace.reg.r4cv.firstrun\endcsname{10-14 12:18}
\expandafter\gdef\csname R@an.trace.reg.r4cv.lagh\endcsname{-0.2}
\expandafter\gdef\csname R@an.trace.reg.r4cv.ledger\endcsname{--}
\expandafter\gdef\csname R@an.trace.reg.r4cv.reg\endcsname{10-14 12:29}
\expandafter\gdef\csname R@an.trace.reg.r4cv.src\endcsname{vcs}
\expandafter\gdef\csname R@an.trace.reg.r4cv.verdict\endcsname{NOT SHOWN}
\expandafter\gdef\csname R@an.trace.reg.r4seeds.beforeresult\endcsname{yes}
\expandafter\gdef\csname R@an.trace.reg.r4seeds.file\endcsname{PREREG\_D1D2\_20261014.md}
\expandafter\gdef\csname R@an.trace.reg.r4seeds.firstresult\endcsname{10-23 08:13}
\expandafter\gdef\csname R@an.trace.reg.r4seeds.firstrun\endcsname{10-22 23:43}
\expandafter\gdef\csname R@an.trace.reg.r4seeds.lagh\endcsname{-0.1}
\expandafter\gdef\csname R@an.trace.reg.r4seeds.ledger\endcsname{--}
\expandafter\gdef\csname R@an.trace.reg.r4seeds.reg\endcsname{10-22 23:49}
\expandafter\gdef\csname R@an.trace.reg.r4seeds.src\endcsname{vcs}
\expandafter\gdef\csname R@an.trace.reg.r4seeds.verdict\endcsname{OK (mtime)}
\expandafter\gdef\csname R@an.trace.reg.r5.beforeresult\endcsname{yes}
\expandafter\gdef\csname R@an.trace.reg.r5.file\endcsname{PREREG\_R5\_20261015.md}
\expandafter\gdef\csname R@an.trace.reg.r5.firstresult\endcsname{10-15 19:59}
\expandafter\gdef\csname R@an.trace.reg.r5.firstrun\endcsname{10-15 10:23}
\expandafter\gdef\csname R@an.trace.reg.r5.lagh\endcsname{+0.1}
\expandafter\gdef\csname R@an.trace.reg.r5.ledger\endcsname{--}
\expandafter\gdef\csname R@an.trace.reg.r5.reg\endcsname{10-15 10:16}
\expandafter\gdef\csname R@an.trace.reg.r5.src\endcsname{vcs}
\expandafter\gdef\csname R@an.trace.reg.r5.verdict\endcsname{OK}
\expandafter\gdef\csname R@an.trace.reg.r6.file\endcsname{PREREG\_R6\_20261016.md}
\expandafter\gdef\csname R@an.trace.reg.r6.firstrun\endcsname{--}
\expandafter\gdef\csname R@an.trace.reg.r6.ledger\endcsname{--}
\expandafter\gdef\csname R@an.trace.reg.r6.reg\endcsname{10-16 02:49}
\expandafter\gdef\csname R@an.trace.reg.r6.src\endcsname{vcs}
\expandafter\gdef\csname R@an.trace.reg.r6.verdict\endcsname{NO TRAINING RUNS}
\expandafter\gdef\csname R@an.trace.reg.r8.beforeresult\endcsname{yes}
\expandafter\gdef\csname R@an.trace.reg.r8.file\endcsname{PREREG\_R8\_20261016.md}
\expandafter\gdef\csname R@an.trace.reg.r8.firstresult\endcsname{10-16 23:55}
\expandafter\gdef\csname R@an.trace.reg.r8.firstrun\endcsname{10-16 15:58}
\expandafter\gdef\csname R@an.trace.reg.r8.lagh\endcsname{-0.2}
\expandafter\gdef\csname R@an.trace.reg.r8.ledger\endcsname{--}
\expandafter\gdef\csname R@an.trace.reg.r8.reg\endcsname{10-16 16:11}
\expandafter\gdef\csname R@an.trace.reg.r8.src\endcsname{vcs}
\expandafter\gdef\csname R@an.trace.reg.r8.verdict\endcsname{PREDATES}
\expandafter\gdef\csname R@an.trace.reg.r9.beforeresult\endcsname{yes}
\expandafter\gdef\csname R@an.trace.reg.r9.file\endcsname{PREREG\_R9\_sweep\_20261017.md}
\expandafter\gdef\csname R@an.trace.reg.r9.firstresult\endcsname{10-17 21:03}
\expandafter\gdef\csname R@an.trace.reg.r9.firstrun\endcsname{10-17 11:25}
\expandafter\gdef\csname R@an.trace.reg.r9.lagh\endcsname{-0.2}
\expandafter\gdef\csname R@an.trace.reg.r9.ledger\endcsname{--}
\expandafter\gdef\csname R@an.trace.reg.r9.reg\endcsname{10-17 11:38}
\expandafter\gdef\csname R@an.trace.reg.r9.src\endcsname{vcs}
\expandafter\gdef\csname R@an.trace.reg.r9.verdict\endcsname{NOT SHOWN}
\expandafter\gdef\csname R@an.trace.reg.rh.beforeresult\endcsname{yes}
\expandafter\gdef\csname R@an.trace.reg.rh.file\endcsname{DESIGNS\_20261014.md, PREREG\_D1D2\_20261014.md}
\expandafter\gdef\csname R@an.trace.reg.rh.firstresult\endcsname{10-15 01:05}
\expandafter\gdef\csname R@an.trace.reg.rh.firstrun\endcsname{10-14 16:01}
\expandafter\gdef\csname R@an.trace.reg.rh.lagh\endcsname{+4.0}
\expandafter\gdef\csname R@an.trace.reg.rh.ledger\endcsname{--}
\expandafter\gdef\csname R@an.trace.reg.rh.reg\endcsname{10-14 11:59}
\expandafter\gdef\csname R@an.trace.reg.rh.src\endcsname{earliest of the files}
\expandafter\gdef\csname R@an.trace.reg.rh.verdict\endcsname{OK}
\expandafter\gdef\csname R@an.trace.reg.rx.beforeresult\endcsname{yes}
\expandafter\gdef\csname R@an.trace.reg.rx.file\endcsname{XARCH\_20261013.md}
\expandafter\gdef\csname R@an.trace.reg.rx.firstresult\endcsname{10-14 01:21}
\expandafter\gdef\csname R@an.trace.reg.rx.firstrun\endcsname{10-13 16:20}
\expandafter\gdef\csname R@an.trace.reg.rx.lagh\endcsname{-0.3}
\expandafter\gdef\csname R@an.trace.reg.rx.ledger\endcsname{--}
\expandafter\gdef\csname R@an.trace.reg.rx.reg\endcsname{10-13 16:36}
\expandafter\gdef\csname R@an.trace.reg.rx.src\endcsname{vcs}
\expandafter\gdef\csname R@an.trace.reg.rx.verdict\endcsname{NOT SHOWN}
\expandafter\gdef\csname R@an.trace.reg.w1.beforeresult\endcsname{partly}
\expandafter\gdef\csname R@an.trace.reg.w1.file\endcsname{WAVE1\_REGISTRATION.md}
\expandafter\gdef\csname R@an.trace.reg.w1.firstresult\endcsname{10-04 23:02}
\expandafter\gdef\csname R@an.trace.reg.w1.firstrun\endcsname{10-04 10:14}
\expandafter\gdef\csname R@an.trace.reg.w1.lagh\endcsname{-3.1}
\expandafter\gdef\csname R@an.trace.reg.w1.ledger\endcsname{10-04 12:29}
\expandafter\gdef\csname R@an.trace.reg.w1.reg\endcsname{10-04 13:23}
\expandafter\gdef\csname R@an.trace.reg.w1.src\endcsname{vcs}
\expandafter\gdef\csname R@an.trace.reg.w1.verdict\endcsname{NOT SHOWN}
\expandafter\gdef\csname R@an.trace.reg.w2.beforeresult\endcsname{yes}
\expandafter\gdef\csname R@an.trace.reg.w2.file\endcsname{WAVE2\_REGISTRATION.md}
\expandafter\gdef\csname R@an.trace.reg.w2.firstresult\endcsname{10-05 07:57}
\expandafter\gdef\csname R@an.trace.reg.w2.firstrun\endcsname{10-04 19:30}
\expandafter\gdef\csname R@an.trace.reg.w2.lagh\endcsname{+6.1}
\expandafter\gdef\csname R@an.trace.reg.w2.ledger\endcsname{10-04 12:29}
\expandafter\gdef\csname R@an.trace.reg.w2.reg\endcsname{10-04 13:23}
\expandafter\gdef\csname R@an.trace.reg.w2.src\endcsname{vcs}
\expandafter\gdef\csname R@an.trace.reg.w2.verdict\endcsname{OK}
\expandafter\gdef\csname R@an.trace.reg.w3.beforeresult\endcsname{yes}
\expandafter\gdef\csname R@an.trace.reg.w3.file\endcsname{WAVE3\_REGISTRATION.md}
\expandafter\gdef\csname R@an.trace.reg.w3.firstresult\endcsname{10-07 18:16}
\expandafter\gdef\csname R@an.trace.reg.w3.firstrun\endcsname{10-07 08:40}
\expandafter\gdef\csname R@an.trace.reg.w3.lagh\endcsname{+67.3}
\expandafter\gdef\csname R@an.trace.reg.w3.ledger\endcsname{10-04 12:29}
\expandafter\gdef\csname R@an.trace.reg.w3.reg\endcsname{10-04 13:23}
\expandafter\gdef\csname R@an.trace.reg.w3.src\endcsname{vcs}
\expandafter\gdef\csname R@an.trace.reg.w3.verdict\endcsname{OK}
\expandafter\gdef\csname R@an.trace.reg.w3n.beforeresult\endcsname{yes}
\expandafter\gdef\csname R@an.trace.reg.w3n.file\endcsname{ADDENDUM\_N\_enrichment\_20261005.md}
\expandafter\gdef\csname R@an.trace.reg.w3n.firstresult\endcsname{10-06 15:21}
\expandafter\gdef\csname R@an.trace.reg.w3n.firstrun\endcsname{10-06 01:05}
\expandafter\gdef\csname R@an.trace.reg.w3n.lagh\endcsname{+11.7}
\expandafter\gdef\csname R@an.trace.reg.w3n.ledger\endcsname{10-05 13:20}
\expandafter\gdef\csname R@an.trace.reg.w3n.reg\endcsname{10-05 13:24}
\expandafter\gdef\csname R@an.trace.reg.w3n.src\endcsname{vcs}
\expandafter\gdef\csname R@an.trace.reg.w3n.verdict\endcsname{OK}
\expandafter\gdef\csname R@an.trace.reg.w4.beforeresult\endcsname{yes}
\expandafter\gdef\csname R@an.trace.reg.w4.file\endcsname{ADDENDUM\_O\_wave4\_20261005.md}
\expandafter\gdef\csname R@an.trace.reg.w4.firstresult\endcsname{10-06 19:48}
\expandafter\gdef\csname R@an.trace.reg.w4.firstrun\endcsname{10-06 12:03}
\expandafter\gdef\csname R@an.trace.reg.w4.lagh\endcsname{+21.4}
\expandafter\gdef\csname R@an.trace.reg.w4.ledger\endcsname{10-05 14:37}
\expandafter\gdef\csname R@an.trace.reg.w4.reg\endcsname{10-05 14:37}
\expandafter\gdef\csname R@an.trace.reg.w4.src\endcsname{vcs}
\expandafter\gdef\csname R@an.trace.reg.w4.verdict\endcsname{OK}
\expandafter\gdef\csname R@an.trace.reg.w5.beforeresult\endcsname{yes}
\expandafter\gdef\csname R@an.trace.reg.w5.file\endcsname{ADDENDUM\_AD\_wave5\_screens\_20261008.md}
\expandafter\gdef\csname R@an.trace.reg.w5.firstresult\endcsname{10-09 05:55}
\expandafter\gdef\csname R@an.trace.reg.w5.firstrun\endcsname{10-08 20:41}
\expandafter\gdef\csname R@an.trace.reg.w5.lagh\endcsname{+8.0}
\expandafter\gdef\csname R@an.trace.reg.w5.ledger\endcsname{10-08 12:30}
\expandafter\gdef\csname R@an.trace.reg.w5.reg\endcsname{10-08 12:42}
\expandafter\gdef\csname R@an.trace.reg.w5.src\endcsname{vcs}
\expandafter\gdef\csname R@an.trace.reg.w5.verdict\endcsname{OK}
\expandafter\gdef\csname R@an.trace.reg.w6.beforeresult\endcsname{yes}
\expandafter\gdef\csname R@an.trace.reg.w6.file\endcsname{ADDENDUM\_AF\_wave6\_V6A1r\_20261008.md}
\expandafter\gdef\csname R@an.trace.reg.w6.firstresult\endcsname{10-09 14:22}
\expandafter\gdef\csname R@an.trace.reg.w6.firstrun\endcsname{10-09 00:40}
\expandafter\gdef\csname R@an.trace.reg.w6.lagh\endcsname{+6.8}
\expandafter\gdef\csname R@an.trace.reg.w6.ledger\endcsname{10-08 16:38}
\expandafter\gdef\csname R@an.trace.reg.w6.reg\endcsname{10-08 17:52}
\expandafter\gdef\csname R@an.trace.reg.w6.src\endcsname{vcs}
\expandafter\gdef\csname R@an.trace.reg.w6.verdict\endcsname{OK}
\expandafter\gdef\csname R@an.trace.reg.w7.beforeresult\endcsname{yes}
\expandafter\gdef\csname R@an.trace.reg.w7.file\endcsname{ADDENDUM\_AN\_wave7\_20261009.md}
\expandafter\gdef\csname R@an.trace.reg.w7.firstresult\endcsname{10-10 00:55}
\expandafter\gdef\csname R@an.trace.reg.w7.firstrun\endcsname{10-09 15:49}
\expandafter\gdef\csname R@an.trace.reg.w7.lagh\endcsname{-2.9}
\expandafter\gdef\csname R@an.trace.reg.w7.ledger\endcsname{10-09 15:01}
\expandafter\gdef\csname R@an.trace.reg.w7.reg\endcsname{10-09 18:41}
\expandafter\gdef\csname R@an.trace.reg.w7.src\endcsname{vcs}
\expandafter\gdef\csname R@an.trace.reg.w7.verdict\endcsname{OK (mtime)}
\expandafter\gdef\csname R@an.trace.reg.w8.beforeresult\endcsname{yes}
\expandafter\gdef\csname R@an.trace.reg.w8.file\endcsname{ADDENDUM\_AR\_wave8\_20261010.md}
\expandafter\gdef\csname R@an.trace.reg.w8.firstresult\endcsname{10-10 16:48}
\expandafter\gdef\csname R@an.trace.reg.w8.firstrun\endcsname{10-10 08:06}
\expandafter\gdef\csname R@an.trace.reg.w8.lagh\endcsname{-80.1}
\expandafter\gdef\csname R@an.trace.reg.w8.ledger\endcsname{10-10 07:23}
\expandafter\gdef\csname R@an.trace.reg.w8.reg\endcsname{10-13 16:13}
\expandafter\gdef\csname R@an.trace.reg.w8.src\endcsname{vcs}
\expandafter\gdef\csname R@an.trace.reg.w8.verdict\endcsname{OK (ledger)}
\expandafter\gdef\csname R@an.trace.reg.w9.beforeresult\endcsname{yes}
\expandafter\gdef\csname R@an.trace.reg.w9.file\endcsname{IDEAS\_NOVEL\_20261013.md}
\expandafter\gdef\csname R@an.trace.reg.w9.firstresult\endcsname{10-14 03:05}
\expandafter\gdef\csname R@an.trace.reg.w9.firstrun\endcsname{10-13 18:27}
\expandafter\gdef\csname R@an.trace.reg.w9.lagh\endcsname{+1.3}
\expandafter\gdef\csname R@an.trace.reg.w9.ledger\endcsname{--}
\expandafter\gdef\csname R@an.trace.reg.w9.reg\endcsname{10-13 17:11}
\expandafter\gdef\csname R@an.trace.reg.w9.src\endcsname{vcs}
\expandafter\gdef\csname R@an.trace.reg.w9.verdict\endcsname{OK}
\expandafter\gdef\csname R@an.trace.reg.waves.fail\endcsname{w1}
\expandafter\gdef\csname R@an.trace.reg.waves.n\endcsname{9}
\expandafter\gdef\csname R@an.trace.reg.waves.nok\endcsname{8}
\expandafter\gdef\csname R@an.trace.tab.allbeforeresult\endcsname{1}
\expandafter\gdef\csname R@an.trace.tab.n\endcsname{29}
\expandafter\gdef\csname R@an.trace.tab.nbestbeforerun\endcsname{23}
\expandafter\gdef\csname R@an.trace.tab.nran\endcsname{28}
\expandafter\gdef\csname R@an.trace.tab.nvcsbeforerun\endcsname{17}
\expandafter\gdef\csname R@an.trace.tab.r18.bestres\endcsname{1}
\expandafter\gdef\csname R@an.trace.tab.r18.bestrun\endcsname{1}
\expandafter\gdef\csname R@an.trace.tab.r18.earlier\endcsname{--}
\expandafter\gdef\csname R@an.trace.tab.r18.file\endcsname{PREREG\_R18\_20261021.md}
\expandafter\gdef\csname R@an.trace.tab.r18.result\endcsname{10-22 06:52}
\expandafter\gdef\csname R@an.trace.tab.r18.run\endcsname{10-21 21:04}
\expandafter\gdef\csname R@an.trace.tab.r18.vcs\endcsname{10-21 12:04}
\expandafter\gdef\csname R@an.trace.tab.r18.vcsres\endcsname{1}
\expandafter\gdef\csname R@an.trace.tab.r18.vcsrun\endcsname{1}
\expandafter\gdef\csname R@an.trace.tab.r19.bestres\endcsname{1}
\expandafter\gdef\csname R@an.trace.tab.r19.bestrun\endcsname{1}
\expandafter\gdef\csname R@an.trace.tab.r19.earlier\endcsname{--}
\expandafter\gdef\csname R@an.trace.tab.r19.file\endcsname{PREREG\_R19\_20261021.md}
\expandafter\gdef\csname R@an.trace.tab.r19.result\endcsname{10-21 21:04}
\expandafter\gdef\csname R@an.trace.tab.r19.run\endcsname{10-21 12:24}
\expandafter\gdef\csname R@an.trace.tab.r19.vcs\endcsname{10-21 12:04}
\expandafter\gdef\csname R@an.trace.tab.r19.vcsres\endcsname{1}
\expandafter\gdef\csname R@an.trace.tab.r19.vcsrun\endcsname{1}
\expandafter\gdef\csname R@an.trace.tab.r20.bestres\endcsname{1}
\expandafter\gdef\csname R@an.trace.tab.r20.bestrun\endcsname{1}
\expandafter\gdef\csname R@an.trace.tab.r20.earlier\endcsname{file 10-21 22:11}
\expandafter\gdef\csname R@an.trace.tab.r20.file\endcsname{PREREG\_R20\_20261021.md}
\expandafter\gdef\csname R@an.trace.tab.r20.result\endcsname{10-22 06:50}
\expandafter\gdef\csname R@an.trace.tab.r20.run\endcsname{10-21 22:30}
\expandafter\gdef\csname R@an.trace.tab.r20.vcs\endcsname{10-21 22:51}
\expandafter\gdef\csname R@an.trace.tab.r20.vcsres\endcsname{1}
\expandafter\gdef\csname R@an.trace.tab.r20.vcsrun\endcsname{0}
\expandafter\gdef\csname R@an.trace.tab.r23.bestres\endcsname{1}
\expandafter\gdef\csname R@an.trace.tab.r23.bestrun\endcsname{1}
\expandafter\gdef\csname R@an.trace.tab.r23.earlier\endcsname{--}
\expandafter\gdef\csname R@an.trace.tab.r23.file\endcsname{PREREG\_R23\_20261022.md}
\expandafter\gdef\csname R@an.trace.tab.r23.result\endcsname{10-25 02:27}
\expandafter\gdef\csname R@an.trace.tab.r23.run\endcsname{10-23 00:10}
\expandafter\gdef\csname R@an.trace.tab.r23.vcs\endcsname{10-22 23:49}
\expandafter\gdef\csname R@an.trace.tab.r23.vcsres\endcsname{1}
\expandafter\gdef\csname R@an.trace.tab.r23.vcsrun\endcsname{1}
\expandafter\gdef\csname R@an.trace.tab.r23a.bestres\endcsname{1}
\expandafter\gdef\csname R@an.trace.tab.r23a.bestrun\endcsname{0}
\expandafter\gdef\csname R@an.trace.tab.r23a.earlier\endcsname{file 10-24 01:11}
\expandafter\gdef\csname R@an.trace.tab.r23a.file\endcsname{PREREG\_R23\_amendment\_s4\_20261024.md}
\expandafter\gdef\csname R@an.trace.tab.r23a.result\endcsname{10-25 02:27}
\expandafter\gdef\csname R@an.trace.tab.r23a.run\endcsname{10-23 00:10}
\expandafter\gdef\csname R@an.trace.tab.r23a.vcs\endcsname{10-24 01:24}
\expandafter\gdef\csname R@an.trace.tab.r23a.vcsres\endcsname{1}
\expandafter\gdef\csname R@an.trace.tab.r23a.vcsrun\endcsname{0}
\expandafter\gdef\csname R@an.trace.tab.r23s.bestres\endcsname{1}
\expandafter\gdef\csname R@an.trace.tab.r23s.bestrun\endcsname{0}
\expandafter\gdef\csname R@an.trace.tab.r23s.earlier\endcsname{--}
\expandafter\gdef\csname R@an.trace.tab.r23s.file\endcsname{PREREG\_R23\_scorer\_20261024.md}
\expandafter\gdef\csname R@an.trace.tab.r23s.result\endcsname{10-25 02:27}
\expandafter\gdef\csname R@an.trace.tab.r23s.run\endcsname{10-23 00:10}
\expandafter\gdef\csname R@an.trace.tab.r23s.vcs\endcsname{10-24 01:24}
\expandafter\gdef\csname R@an.trace.tab.r23s.vcsres\endcsname{1}
\expandafter\gdef\csname R@an.trace.tab.r23s.vcsrun\endcsname{0}
\expandafter\gdef\csname R@an.trace.tab.r24.bestres\endcsname{1}
\expandafter\gdef\csname R@an.trace.tab.r24.bestrun\endcsname{1}
\expandafter\gdef\csname R@an.trace.tab.r24.earlier\endcsname{file 10-23 02:40}
\expandafter\gdef\csname R@an.trace.tab.r24.file\endcsname{PREREG\_R24\_20261023.md}
\expandafter\gdef\csname R@an.trace.tab.r24.result\endcsname{10-23 18:18}
\expandafter\gdef\csname R@an.trace.tab.r24.run\endcsname{10-23 02:51}
\expandafter\gdef\csname R@an.trace.tab.r24.vcs\endcsname{10-23 02:56}
\expandafter\gdef\csname R@an.trace.tab.r24.vcsres\endcsname{1}
\expandafter\gdef\csname R@an.trace.tab.r24.vcsrun\endcsname{0}
\expandafter\gdef\csname R@an.trace.tab.r30.bestres\endcsname{1}
\expandafter\gdef\csname R@an.trace.tab.r30.bestrun\endcsname{1}
\expandafter\gdef\csname R@an.trace.tab.r30.earlier\endcsname{--}
\expandafter\gdef\csname R@an.trace.tab.r30.file\endcsname{PREREG\_R30\_20261024.md}
\expandafter\gdef\csname R@an.trace.tab.r30.result\endcsname{10-25 08:01}
\expandafter\gdef\csname R@an.trace.tab.r30.run\endcsname{10-24 23:29}
\expandafter\gdef\csname R@an.trace.tab.r30.vcs\endcsname{10-24 20:30}
\expandafter\gdef\csname R@an.trace.tab.r30.vcsres\endcsname{1}
\expandafter\gdef\csname R@an.trace.tab.r30.vcsrun\endcsname{1}
\expandafter\gdef\csname R@an.trace.tab.r30h.bestres\endcsname{1}
\expandafter\gdef\csname R@an.trace.tab.r30h.bestrun\endcsname{1}
\expandafter\gdef\csname R@an.trace.tab.r30h.earlier\endcsname{file 10-24 20:16}
\expandafter\gdef\csname R@an.trace.tab.r30h.file\endcsname{PREREG\_R30\_addendum\_history\_20261024.md}
\expandafter\gdef\csname R@an.trace.tab.r30h.result\endcsname{10-25 08:01}
\expandafter\gdef\csname R@an.trace.tab.r30h.run\endcsname{10-24 23:29}
\expandafter\gdef\csname R@an.trace.tab.r30h.vcs\endcsname{10-24 20:30}
\expandafter\gdef\csname R@an.trace.tab.r30h.vcsres\endcsname{1}
\expandafter\gdef\csname R@an.trace.tab.r30h.vcsrun\endcsname{1}
\expandafter\gdef\csname R@an.trace.tab.r31.bestres\endcsname{1}
\expandafter\gdef\csname R@an.trace.tab.r31.bestrun\endcsname{1}
\expandafter\gdef\csname R@an.trace.tab.r31.earlier\endcsname{file 10-25 04:10}
\expandafter\gdef\csname R@an.trace.tab.r31.file\endcsname{PREREG\_R31\_20261025.md}
\expandafter\gdef\csname R@an.trace.tab.r31.result\endcsname{10-25 12:49}
\expandafter\gdef\csname R@an.trace.tab.r31.run\endcsname{10-25 04:24}
\expandafter\gdef\csname R@an.trace.tab.r31.vcs\endcsname{10-25 04:34}
\expandafter\gdef\csname R@an.trace.tab.r31.vcsres\endcsname{1}
\expandafter\gdef\csname R@an.trace.tab.r31.vcsrun\endcsname{0}
\expandafter\gdef\csname R@an.trace.tab.r31s.bestres\endcsname{1}
\expandafter\gdef\csname R@an.trace.tab.r31s.bestrun\endcsname{0}
\expandafter\gdef\csname R@an.trace.tab.r31s.earlier\endcsname{file 10-25 04:40}
\expandafter\gdef\csname R@an.trace.tab.r31s.file\endcsname{PREREG\_R31\_addendum\_sha\_20261025.md}
\expandafter\gdef\csname R@an.trace.tab.r31s.result\endcsname{10-25 12:49}
\expandafter\gdef\csname R@an.trace.tab.r31s.run\endcsname{10-25 04:24}
\expandafter\gdef\csname R@an.trace.tab.r31s.vcs\endcsname{11-01 00:54}
\expandafter\gdef\csname R@an.trace.tab.r31s.vcsres\endcsname{0}
\expandafter\gdef\csname R@an.trace.tab.r31s.vcsrun\endcsname{0}
\expandafter\gdef\csname R@an.trace.tab.r32.bestres\endcsname{1}
\expandafter\gdef\csname R@an.trace.tab.r32.bestrun\endcsname{1}
\expandafter\gdef\csname R@an.trace.tab.r32.earlier\endcsname{--}
\expandafter\gdef\csname R@an.trace.tab.r32.file\endcsname{PREREG\_R32\_20261101.md}
\expandafter\gdef\csname R@an.trace.tab.r32.result\endcsname{11-01 12:40}
\expandafter\gdef\csname R@an.trace.tab.r32.run\endcsname{11-01 09:22}
\expandafter\gdef\csname R@an.trace.tab.r32.vcs\endcsname{11-01 02:30}
\expandafter\gdef\csname R@an.trace.tab.r32.vcsres\endcsname{1}
\expandafter\gdef\csname R@an.trace.tab.r32.vcsrun\endcsname{1}
\expandafter\gdef\csname R@an.trace.tab.r32s.bestres\endcsname{1}
\expandafter\gdef\csname R@an.trace.tab.r32s.bestrun\endcsname{1}
\expandafter\gdef\csname R@an.trace.tab.r32s.earlier\endcsname{file 11-01 03:14}
\expandafter\gdef\csname R@an.trace.tab.r32s.file\endcsname{PREREG\_R32\_addendum\_sha\_20261101.md}
\expandafter\gdef\csname R@an.trace.tab.r32s.result\endcsname{11-01 12:40}
\expandafter\gdef\csname R@an.trace.tab.r32s.run\endcsname{11-01 09:22}
\expandafter\gdef\csname R@an.trace.tab.r32s.vcs\endcsname{11-01 03:23}
\expandafter\gdef\csname R@an.trace.tab.r32s.vcsres\endcsname{1}
\expandafter\gdef\csname R@an.trace.tab.r32s.vcsrun\endcsname{1}
\expandafter\gdef\csname R@an.trace.tab.r33.bestres\endcsname{1}
\expandafter\gdef\csname R@an.trace.tab.r33.bestrun\endcsname{1}
\expandafter\gdef\csname R@an.trace.tab.r33.earlier\endcsname{file 11-01 16:27}
\expandafter\gdef\csname R@an.trace.tab.r33.file\endcsname{PREREG\_R33\_20261101.md}
\expandafter\gdef\csname R@an.trace.tab.r33.result\endcsname{11-02 03:13}
\expandafter\gdef\csname R@an.trace.tab.r33.run\endcsname{11-01 17:46}
\expandafter\gdef\csname R@an.trace.tab.r33.vcs\endcsname{11-01 16:38}
\expandafter\gdef\csname R@an.trace.tab.r33.vcsres\endcsname{1}
\expandafter\gdef\csname R@an.trace.tab.r33.vcsrun\endcsname{1}
\expandafter\gdef\csname R@an.trace.tab.r33a.bestres\endcsname{1}
\expandafter\gdef\csname R@an.trace.tab.r33a.bestrun\endcsname{1}
\expandafter\gdef\csname R@an.trace.tab.r33a.earlier\endcsname{file 11-01 17:02}
\expandafter\gdef\csname R@an.trace.tab.r33a.file\endcsname{PREREG\_R33\_amendment\_schedule\_acdc\_20261101.md}
\expandafter\gdef\csname R@an.trace.tab.r33a.result\endcsname{11-02 03:13}
\expandafter\gdef\csname R@an.trace.tab.r33a.run\endcsname{11-01 17:46}
\expandafter\gdef\csname R@an.trace.tab.r33a.vcs\endcsname{11-01 17:22}
\expandafter\gdef\csname R@an.trace.tab.r33a.vcsres\endcsname{1}
\expandafter\gdef\csname R@an.trace.tab.r33a.vcsrun\endcsname{1}
\expandafter\gdef\csname R@an.trace.tab.r34.bestres\endcsname{1}
\expandafter\gdef\csname R@an.trace.tab.r34.bestrun\endcsname{1}
\expandafter\gdef\csname R@an.trace.tab.r34.earlier\endcsname{file 11-01 16:28}
\expandafter\gdef\csname R@an.trace.tab.r34.file\endcsname{PREREG\_R34\_20261101.md}
\expandafter\gdef\csname R@an.trace.tab.r34.result\endcsname{11-02 02:20}
\expandafter\gdef\csname R@an.trace.tab.r34.run\endcsname{11-01 17:02}
\expandafter\gdef\csname R@an.trace.tab.r34.vcs\endcsname{11-01 16:38}
\expandafter\gdef\csname R@an.trace.tab.r34.vcsres\endcsname{1}
\expandafter\gdef\csname R@an.trace.tab.r34.vcsrun\endcsname{1}
\expandafter\gdef\csname R@an.trace.tab.r35.bestres\endcsname{1}
\expandafter\gdef\csname R@an.trace.tab.r35.bestrun\endcsname{--}
\expandafter\gdef\csname R@an.trace.tab.r35.earlier\endcsname{file 11-02 21:18}
\expandafter\gdef\csname R@an.trace.tab.r35.file\endcsname{PREREG\_R35\_20261102.md}
\expandafter\gdef\csname R@an.trace.tab.r35.result\endcsname{11-02 21:52}
\expandafter\gdef\csname R@an.trace.tab.r35.run\endcsname{--}
\expandafter\gdef\csname R@an.trace.tab.r35.vcs\endcsname{11-02 22:01}
\expandafter\gdef\csname R@an.trace.tab.r35.vcsres\endcsname{0}
\expandafter\gdef\csname R@an.trace.tab.r35.vcsrun\endcsname{--}
\expandafter\gdef\csname R@an.trace.tab.r4cv.bestres\endcsname{1}
\expandafter\gdef\csname R@an.trace.tab.r4cv.bestrun\endcsname{0}
\expandafter\gdef\csname R@an.trace.tab.r4cv.earlier\endcsname{--}
\expandafter\gdef\csname R@an.trace.tab.r4cv.file\endcsname{PREREG\_D1D2\_20261014.md, Sec.~3}
\expandafter\gdef\csname R@an.trace.tab.r4cv.result\endcsname{10-14 21:36}
\expandafter\gdef\csname R@an.trace.tab.r4cv.run\endcsname{10-14 12:18}
\expandafter\gdef\csname R@an.trace.tab.r4cv.vcs\endcsname{10-14 12:29}
\expandafter\gdef\csname R@an.trace.tab.r4cv.vcsres\endcsname{1}
\expandafter\gdef\csname R@an.trace.tab.r4cv.vcsrun\endcsname{0}
\expandafter\gdef\csname R@an.trace.tab.r4seeds.bestres\endcsname{1}
\expandafter\gdef\csname R@an.trace.tab.r4seeds.bestrun\endcsname{1}
\expandafter\gdef\csname R@an.trace.tab.r4seeds.earlier\endcsname{file 10-22 23:10}
\expandafter\gdef\csname R@an.trace.tab.r4seeds.file\endcsname{PREREG\_D1D2\_20261014.md, Sec.~6}
\expandafter\gdef\csname R@an.trace.tab.r4seeds.result\endcsname{10-23 08:13}
\expandafter\gdef\csname R@an.trace.tab.r4seeds.run\endcsname{10-22 23:43}
\expandafter\gdef\csname R@an.trace.tab.r4seeds.vcs\endcsname{10-22 23:49}
\expandafter\gdef\csname R@an.trace.tab.r4seeds.vcsres\endcsname{1}
\expandafter\gdef\csname R@an.trace.tab.r4seeds.vcsrun\endcsname{0}
\expandafter\gdef\csname R@an.trace.tab.rx.bestres\endcsname{1}
\expandafter\gdef\csname R@an.trace.tab.rx.bestrun\endcsname{0}
\expandafter\gdef\csname R@an.trace.tab.rx.earlier\endcsname{--}
\expandafter\gdef\csname R@an.trace.tab.rx.file\endcsname{XARCH\_20261013.md}
\expandafter\gdef\csname R@an.trace.tab.rx.result\endcsname{10-14 01:21}
\expandafter\gdef\csname R@an.trace.tab.rx.run\endcsname{10-13 16:20}
\expandafter\gdef\csname R@an.trace.tab.rx.vcs\endcsname{10-13 16:36}
\expandafter\gdef\csname R@an.trace.tab.rx.vcsres\endcsname{1}
\expandafter\gdef\csname R@an.trace.tab.rx.vcsrun\endcsname{0}
\expandafter\gdef\csname R@an.trace.tab.w2.bestres\endcsname{1}
\expandafter\gdef\csname R@an.trace.tab.w2.bestrun\endcsname{1}
\expandafter\gdef\csname R@an.trace.tab.w2.earlier\endcsname{ledger 10-04 12:29}
\expandafter\gdef\csname R@an.trace.tab.w2.file\endcsname{WAVE2\_REGISTRATION.md}
\expandafter\gdef\csname R@an.trace.tab.w2.result\endcsname{10-05 07:57}
\expandafter\gdef\csname R@an.trace.tab.w2.run\endcsname{10-04 19:30}
\expandafter\gdef\csname R@an.trace.tab.w2.vcs\endcsname{10-04 13:23}
\expandafter\gdef\csname R@an.trace.tab.w2.vcsres\endcsname{1}
\expandafter\gdef\csname R@an.trace.tab.w2.vcsrun\endcsname{1}
\expandafter\gdef\csname R@an.trace.tab.w3.bestres\endcsname{1}
\expandafter\gdef\csname R@an.trace.tab.w3.bestrun\endcsname{1}
\expandafter\gdef\csname R@an.trace.tab.w3.earlier\endcsname{ledger 10-04 12:29}
\expandafter\gdef\csname R@an.trace.tab.w3.file\endcsname{WAVE3\_REGISTRATION.md}
\expandafter\gdef\csname R@an.trace.tab.w3.result\endcsname{10-07 18:16}
\expandafter\gdef\csname R@an.trace.tab.w3.run\endcsname{10-07 08:40}
\expandafter\gdef\csname R@an.trace.tab.w3.vcs\endcsname{10-04 13:23}
\expandafter\gdef\csname R@an.trace.tab.w3.vcsres\endcsname{1}
\expandafter\gdef\csname R@an.trace.tab.w3.vcsrun\endcsname{1}
\expandafter\gdef\csname R@an.trace.tab.w3n.bestres\endcsname{1}
\expandafter\gdef\csname R@an.trace.tab.w3n.bestrun\endcsname{1}
\expandafter\gdef\csname R@an.trace.tab.w3n.earlier\endcsname{ledger 10-05 13:20}
\expandafter\gdef\csname R@an.trace.tab.w3n.file\endcsname{ADDENDUM\_N\_enrichment\_20261005.md}
\expandafter\gdef\csname R@an.trace.tab.w3n.result\endcsname{10-06 15:21}
\expandafter\gdef\csname R@an.trace.tab.w3n.run\endcsname{10-06 01:05}
\expandafter\gdef\csname R@an.trace.tab.w3n.vcs\endcsname{10-05 13:24}
\expandafter\gdef\csname R@an.trace.tab.w3n.vcsres\endcsname{1}
\expandafter\gdef\csname R@an.trace.tab.w3n.vcsrun\endcsname{1}
\expandafter\gdef\csname R@an.trace.tab.w4.bestres\endcsname{1}
\expandafter\gdef\csname R@an.trace.tab.w4.bestrun\endcsname{1}
\expandafter\gdef\csname R@an.trace.tab.w4.earlier\endcsname{--}
\expandafter\gdef\csname R@an.trace.tab.w4.file\endcsname{ADDENDUM\_O\_wave4\_20261005.md}
\expandafter\gdef\csname R@an.trace.tab.w4.result\endcsname{10-06 19:48}
\expandafter\gdef\csname R@an.trace.tab.w4.run\endcsname{10-06 12:03}
\expandafter\gdef\csname R@an.trace.tab.w4.vcs\endcsname{10-05 14:37}
\expandafter\gdef\csname R@an.trace.tab.w4.vcsres\endcsname{1}
\expandafter\gdef\csname R@an.trace.tab.w4.vcsrun\endcsname{1}
\expandafter\gdef\csname R@an.trace.tab.w5.bestres\endcsname{1}
\expandafter\gdef\csname R@an.trace.tab.w5.bestrun\endcsname{1}
\expandafter\gdef\csname R@an.trace.tab.w5.earlier\endcsname{ledger 10-08 12:30}
\expandafter\gdef\csname R@an.trace.tab.w5.file\endcsname{ADDENDUM\_AD\_wave5\_screens\_20261008.md}
\expandafter\gdef\csname R@an.trace.tab.w5.result\endcsname{10-09 05:55}
\expandafter\gdef\csname R@an.trace.tab.w5.run\endcsname{10-08 20:41}
\expandafter\gdef\csname R@an.trace.tab.w5.vcs\endcsname{10-08 12:42}
\expandafter\gdef\csname R@an.trace.tab.w5.vcsres\endcsname{1}
\expandafter\gdef\csname R@an.trace.tab.w5.vcsrun\endcsname{1}
\expandafter\gdef\csname R@an.trace.tab.w6.bestres\endcsname{1}
\expandafter\gdef\csname R@an.trace.tab.w6.bestrun\endcsname{1}
\expandafter\gdef\csname R@an.trace.tab.w6.earlier\endcsname{ledger 10-08 16:38}
\expandafter\gdef\csname R@an.trace.tab.w6.file\endcsname{ADDENDUM\_AF\_wave6\_V6A1r\_20261008.md}
\expandafter\gdef\csname R@an.trace.tab.w6.result\endcsname{10-09 14:22}
\expandafter\gdef\csname R@an.trace.tab.w6.run\endcsname{10-09 00:40}
\expandafter\gdef\csname R@an.trace.tab.w6.vcs\endcsname{10-08 17:52}
\expandafter\gdef\csname R@an.trace.tab.w6.vcsres\endcsname{1}
\expandafter\gdef\csname R@an.trace.tab.w6.vcsrun\endcsname{1}
\expandafter\gdef\csname R@an.trace.tab.w7.bestres\endcsname{1}
\expandafter\gdef\csname R@an.trace.tab.w7.bestrun\endcsname{1}
\expandafter\gdef\csname R@an.trace.tab.w7.earlier\endcsname{file 10-09 15:03}
\expandafter\gdef\csname R@an.trace.tab.w7.file\endcsname{ADDENDUM\_AN\_wave7\_20261009.md}
\expandafter\gdef\csname R@an.trace.tab.w7.result\endcsname{10-10 00:55}
\expandafter\gdef\csname R@an.trace.tab.w7.run\endcsname{10-09 15:49}
\expandafter\gdef\csname R@an.trace.tab.w7.vcs\endcsname{10-09 18:41}
\expandafter\gdef\csname R@an.trace.tab.w7.vcsres\endcsname{1}
\expandafter\gdef\csname R@an.trace.tab.w7.vcsrun\endcsname{0}
\expandafter\gdef\csname R@an.trace.tab.w8.bestres\endcsname{1}
\expandafter\gdef\csname R@an.trace.tab.w8.bestrun\endcsname{1}
\expandafter\gdef\csname R@an.trace.tab.w8.earlier\endcsname{ledger 10-10 07:23}
\expandafter\gdef\csname R@an.trace.tab.w8.file\endcsname{ADDENDUM\_AR\_wave8\_20261010.md}
\expandafter\gdef\csname R@an.trace.tab.w8.result\endcsname{10-10 16:48}
\expandafter\gdef\csname R@an.trace.tab.w8.run\endcsname{10-10 08:06}
\expandafter\gdef\csname R@an.trace.tab.w8.vcs\endcsname{10-13 16:13}
\expandafter\gdef\csname R@an.trace.tab.w8.vcsres\endcsname{0}
\expandafter\gdef\csname R@an.trace.tab.w8.vcsrun\endcsname{0}
\expandafter\gdef\csname R@an.trace.tab.w9.bestres\endcsname{1}
\expandafter\gdef\csname R@an.trace.tab.w9.bestrun\endcsname{1}
\expandafter\gdef\csname R@an.trace.tab.w9.earlier\endcsname{--}
\expandafter\gdef\csname R@an.trace.tab.w9.file\endcsname{IDEAS\_NOVEL\_20261013.md}
\expandafter\gdef\csname R@an.trace.tab.w9.result\endcsname{10-14 03:05}
\expandafter\gdef\csname R@an.trace.tab.w9.run\endcsname{10-13 18:27}
\expandafter\gdef\csname R@an.trace.tab.w9.vcs\endcsname{10-13 17:11}
\expandafter\gdef\csname R@an.trace.tab.w9.vcsres\endcsname{1}
\expandafter\gdef\csname R@an.trace.tab.w9.vcsrun\endcsname{1}
\expandafter\gdef\csname R@an.trace.tab.waves.beforeresult\endcsname{1}
\makeatother

\makeatletter
\expandafter\gdef\csname R@v11.L1.aheadall\endcsname{1}
\expandafter\gdef\csname R@v11.V7A1.macs.allunder\endcsname{1}
\expandafter\gdef\csname R@v11.cost.V9R2.acdc.params\endcsname{0.99}
\expandafter\gdef\csname R@v11.cost.V9R2.acdc.step\endcsname{0.88}
\expandafter\gdef\csname R@v11.cost.V9R2.amos.params\endcsname{0.99}
\expandafter\gdef\csname R@v11.cost.V9R2.amos.step\endcsname{0.87}
\expandafter\gdef\csname R@v11.cost.V9R2.kits.params\endcsname{0.99}
\expandafter\gdef\csname R@v11.cost.V9R2.kits.step\endcsname{0.81}
\expandafter\gdef\csname R@v11.cost.Y2.acdc.params\endcsname{1.00}
\expandafter\gdef\csname R@v11.cost.Y2.acdc.step\endcsname{1.16}
\expandafter\gdef\csname R@v11.cost.Y2.amos.params\endcsname{1.00}
\expandafter\gdef\csname R@v11.cost.Y2.amos.step\endcsname{1.14}
\expandafter\gdef\csname R@v11.cost.Y2.kits.params\endcsname{0.99}
\expandafter\gdef\csname R@v11.cost.Y2.kits.step\endcsname{0.95}
\expandafter\gdef\csname R@v11.on\endcsname{1}
\expandafter\gdef\csname R@v11.par.V7A1.vsL1\endcsname{0.30}
\expandafter\gdef\csname R@v11.r3alone.acdc.beyond\endcsname{0}
\expandafter\gdef\csname R@v11.r3alone.acdc.d\endcsname{\ensuremath{+0.04}}
\expandafter\gdef\csname R@v11.r3alone.acdc.ndn\endcsname{1}
\expandafter\gdef\csname R@v11.r3alone.acdc.nup\endcsname{3}
\expandafter\gdef\csname R@v11.r3alone.acdc.verdict\endcsname{UNDERPOWERED}
\expandafter\gdef\csname R@v11.r3alone.amos.beyond\endcsname{0}
\expandafter\gdef\csname R@v11.r3alone.amos.d\endcsname{\ensuremath{+0.21}}
\expandafter\gdef\csname R@v11.r3alone.amos.ndn\endcsname{1}
\expandafter\gdef\csname R@v11.r3alone.amos.nup\endcsname{3}
\expandafter\gdef\csname R@v11.r3alone.amos.verdict\endcsname{CONFIRMED}
\expandafter\gdef\csname R@v11.r3alone.kits.beyond\endcsname{1}
\expandafter\gdef\csname R@v11.r3alone.kits.d\endcsname{\ensuremath{-1.75}}
\expandafter\gdef\csname R@v11.r3alone.kits.ndn\endcsname{4}
\expandafter\gdef\csname R@v11.r3alone.kits.nup\endcsname{0}
\expandafter\gdef\csname R@v11.r3alone.kits.verdict\endcsname{REVERSED}
\expandafter\gdef\csname R@v11.r3alone.rounding\endcsname{1}
\expandafter\gdef\csname R@v11.repl.sel.lost\endcsname{0.34}
\expandafter\gdef\csname R@v11.share.all.hi\endcsname{76}
\expandafter\gdef\csname R@v11.share.all.lo\endcsname{39}
\expandafter\gdef\csname R@v11.share.cv.acdc\endcsname{39}
\expandafter\gdef\csname R@v11.share.cv.amos\endcsname{63}
\expandafter\gdef\csname R@v11.share.hi\endcsname{76}
\expandafter\gdef\csname R@v11.share.lo\endcsname{41}
\makeatother

\makeatletter
\expandafter\gdef\csname R@an.v6.s23.L1.bound\endcsname{86.08}
\expandafter\gdef\csname R@an.v6.s23.L1.bound.gap\endcsname{\ensuremath{+1.03}}
\expandafter\gdef\csname R@an.v6.s23.L1.gross\endcsname{61.08}
\expandafter\gdef\csname R@an.v6.s23.L1.gross.gap\endcsname{\ensuremath{+6.77}}
\expandafter\gdef\csname R@an.v6.s23.L1.share\endcsname{50.7}
\expandafter\gdef\csname R@an.v6.s23.V7A1.bound\endcsname{86.09}
\expandafter\gdef\csname R@an.v6.s23.V7A1.bound.gap\endcsname{\ensuremath{+1.04}}
\expandafter\gdef\csname R@an.v6.s23.V7A1.gross\endcsname{59.09}
\expandafter\gdef\csname R@an.v6.s23.V7A1.gross.gap\endcsname{\ensuremath{+4.79}}
\expandafter\gdef\csname R@an.v6.s23.V7A1.share\endcsname{51.9}
\expandafter\gdef\csname R@an.v6.s23.Y2.bound\endcsname{85.75}
\expandafter\gdef\csname R@an.v6.s23.Y2.bound.gap\endcsname{\ensuremath{+0.69}}
\expandafter\gdef\csname R@an.v6.s23.Y2.gross\endcsname{63.78}
\expandafter\gdef\csname R@an.v6.s23.Y2.gross.gap\endcsname{\ensuremath{+9.48}}
\expandafter\gdef\csname R@an.v6.s23.Y2.share\endcsname{48.3}
\expandafter\gdef\csname R@an.v6.s23.allahead\endcsname{1}
\expandafter\gdef\csname R@an.v6.s23.base.bound\endcsname{85.06}
\expandafter\gdef\csname R@an.v6.s23.base.gross\endcsname{54.30}
\expandafter\gdef\csname R@an.v6.s23.base.share\endcsname{52.9}
\expandafter\gdef\csname R@an.v6.s23.best\endcsname{Y2}
\expandafter\gdef\csname R@an.v6.s23.bound.gap.max\endcsname{1.0}
\expandafter\gdef\csname R@an.v6.s23.bound.gap.min\endcsname{0.7}
\expandafter\gdef\csname R@an.v6.s23.gross.gap.max\endcsname{9.5}
\expandafter\gdef\csname R@an.v6.s23.gross.gap.min\endcsname{4.8}
\expandafter\gdef\csname R@an.v6.s23.n\endcsname{105}
\expandafter\gdef\csname R@an.v6.s23.nbound\endcsname{286}
\expandafter\gdef\csname R@an.v6.s23.on\endcsname{1}
\expandafter\gdef\csname R@an.v6.s23.share.max\endcsname{52.9}
\expandafter\gdef\csname R@an.v6.s23.share.min\endcsname{48.3}
\makeatother

\input{an_extra_v6review.tex}
\makeatletter
\expandafter\gdef\csname R@an.v7.cal.f0.LeanNeXtM.nexcl\endcsname{0}
\expandafter\gdef\csname R@an.v7.cal.f0.LeanNeXtM.npairs\endcsname{3}
\expandafter\gdef\csname R@an.v7.cal.f0.LeanNeXtS.nexcl\endcsname{0}
\expandafter\gdef\csname R@an.v7.cal.f0.LeanNeXtS.npairs\endcsname{3}
\expandafter\gdef\csname R@an.v7.cal.f0.ResEncM.nexcl\endcsname{1}
\expandafter\gdef\csname R@an.v7.cal.f0.ResEncM.npairs\endcsname{3}
\expandafter\gdef\csname R@an.v7.cal.f0.nexcl\endcsname{15}
\expandafter\gdef\csname R@an.v7.cal.f0.nnUNet.nexcl\endcsname{14}
\expandafter\gdef\csname R@an.v7.cal.f0.nnUNet.npairs\endcsname{66}
\expandafter\gdef\csname R@an.v7.cal.f0.npairs\endcsname{75}
\expandafter\gdef\csname R@an.v7.cal.f0.pct\endcsname{20}
\expandafter\gdef\csname R@an.v7.cal.f0.sd\endcsname{0.92}
\expandafter\gdef\csname R@an.v7.cal.f0.width.median\endcsname{2.97}
\expandafter\gdef\csname R@an.v7.cal.kits.cv.V7A1.d\endcsname{\ensuremath{-0.13}}
\expandafter\gdef\csname R@an.v7.cal.kits.cv.V7A1.excl\endcsname{0}
\expandafter\gdef\csname R@an.v7.cal.kits.cv.V7A1.hi\endcsname{\ensuremath{+0.53}}
\expandafter\gdef\csname R@an.v7.cal.kits.cv.V7A1.lo\endcsname{\ensuremath{-0.81}}
\expandafter\gdef\csname R@an.v7.cal.kits.cv.V7A1.width\endcsname{1.34}
\expandafter\gdef\csname R@an.v7.cal.kits.cv.base.d\endcsname{\ensuremath{-0.13}}
\expandafter\gdef\csname R@an.v7.cal.kits.cv.base.excl\endcsname{0}
\expandafter\gdef\csname R@an.v7.cal.kits.cv.base.hi\endcsname{\ensuremath{+0.69}}
\expandafter\gdef\csname R@an.v7.cal.kits.cv.base.lo\endcsname{\ensuremath{-0.96}}
\expandafter\gdef\csname R@an.v7.cal.kits.cv.base.width\endcsname{1.66}
\expandafter\gdef\csname R@an.v7.cal.kits.cv.real.2seed.d\endcsname{\ensuremath{+1.63}}
\expandafter\gdef\csname R@an.v7.cal.kits.cv.real.2seed.excl\endcsname{1}
\expandafter\gdef\csname R@an.v7.cal.kits.cv.real.2seed.hi\endcsname{\ensuremath{+2.27}}
\expandafter\gdef\csname R@an.v7.cal.kits.cv.real.2seed.lo\endcsname{\ensuremath{+1.04}}
\expandafter\gdef\csname R@an.v7.cal.kits.cv.real.2seed.ts.excl\endcsname{1}
\expandafter\gdef\csname R@an.v7.cal.kits.cv.real.2seed.ts.hi\endcsname{\ensuremath{+2.49}}
\expandafter\gdef\csname R@an.v7.cal.kits.cv.real.2seed.ts.lo\endcsname{\ensuremath{+0.78}}
\expandafter\gdef\csname R@an.v7.cal.kits.cv.real.2seed.width\endcsname{1.23}
\expandafter\gdef\csname R@an.v7.cal.kits.cv.real.s0.d\endcsname{\ensuremath{+1.63}}
\expandafter\gdef\csname R@an.v7.cal.kits.cv.real.s0.excl\endcsname{1}
\expandafter\gdef\csname R@an.v7.cal.kits.cv.real.s0.hi\endcsname{\ensuremath{+2.44}}
\expandafter\gdef\csname R@an.v7.cal.kits.cv.real.s0.lo\endcsname{\ensuremath{+0.88}}
\expandafter\gdef\csname R@an.v7.cal.kits.cv.real.s0.ts.excl\endcsname{1}
\expandafter\gdef\csname R@an.v7.cal.kits.cv.real.s0.ts.hi\endcsname{\ensuremath{+2.79}}
\expandafter\gdef\csname R@an.v7.cal.kits.cv.real.s0.ts.lo\endcsname{\ensuremath{+0.48}}
\expandafter\gdef\csname R@an.v7.cal.kits.cv.real.s0.width\endcsname{1.57}
\expandafter\gdef\csname R@an.v7.cal.kits.sigma.V7A1\endcsname{0.22}
\expandafter\gdef\csname R@an.v7.cal.kits.sigma.base\endcsname{0.83}
\expandafter\gdef\csname R@an.v7.cal.null.contains\endcsname{1}
\expandafter\gdef\csname R@an.v7.extr.V7A1.f4\endcsname{11.25}
\expandafter\gdef\csname R@an.v7.extr.V7A1.f4.top4.max.mL\endcsname{353}
\expandafter\gdef\csname R@an.v7.extr.V7A1.f4.top4.min.mL\endcsname{149}
\expandafter\gdef\csname R@an.v7.extr.V7A1.f4.top4.share\endcsname{81}
\expandafter\gdef\csname R@an.v7.extr.V7A1.f4.top4case.max.mL\endcsname{357.6}
\expandafter\gdef\csname R@an.v7.extr.V7A1.f4.top4case.min.mL\endcsname{149.3}
\expandafter\gdef\csname R@an.v7.extr.Y2.f4\endcsname{0.093}
\expandafter\gdef\csname R@an.v7.extr.Y2.f4.dice\endcsname{0.768}
\expandafter\gdef\csname R@an.v7.extr.Y2.f4.dice3\endcsname{0.768}
\expandafter\gdef\csname R@an.v7.extr.Y2.f4.floorbound\endcsname{0.107}
\expandafter\gdef\csname R@an.v7.extr.Y2.f4.ncomp\endcsname{6}
\expandafter\gdef\csname R@an.v7.extr.Y2.f4.recall\endcsname{0.864}
\expandafter\gdef\csname R@an.v7.extr.Y2.f4.recall3\endcsname{0.864}
\expandafter\gdef\csname R@an.v7.extr.Y2.f4.top4.max.mL\endcsname{1.0}
\expandafter\gdef\csname R@an.v7.extr.Y2.f4.top4case.max.mL\endcsname{1}
\expandafter\gdef\csname R@an.v7.extr.Y2.f4.total.mL\endcsname{9.1}
\expandafter\gdef\csname R@an.v7.extr.Y2.f4.volratio\endcsname{0.96}
\expandafter\gdef\csname R@an.v7.extr.f4share\endcsname{65}
\expandafter\gdef\csname R@an.v7.extr.f4share1\endcsname{65.1}
\expandafter\gdef\csname R@an.v7.extr.top4case.marginpct\endcsname{59}
\expandafter\gdef\csname R@an.v7.ho.brats.L1.d\endcsname{\ensuremath{-0.02}}
\expandafter\gdef\csname R@an.v7.ho.brats.L1.excl\endcsname{0}
\expandafter\gdef\csname R@an.v7.ho.brats.L1.hi\endcsname{\ensuremath{+0.95}}
\expandafter\gdef\csname R@an.v7.ho.brats.L1.lo\endcsname{\ensuremath{-1.02}}
\expandafter\gdef\csname R@an.v7.ho.brats.L1.n\endcsname{97}
\expandafter\gdef\csname R@an.v7.ho.brats.L1.npos\endcsname{0}
\expandafter\gdef\csname R@an.v7.ho.brats.L1.nseed\endcsname{1}
\expandafter\gdef\csname R@an.v7.ho.brats.L3.d\endcsname{\ensuremath{+0.07}}
\expandafter\gdef\csname R@an.v7.ho.brats.L3.excl\endcsname{0}
\expandafter\gdef\csname R@an.v7.ho.brats.L3.hi\endcsname{\ensuremath{+1.19}}
\expandafter\gdef\csname R@an.v7.ho.brats.L3.lo\endcsname{\ensuremath{-0.84}}
\expandafter\gdef\csname R@an.v7.ho.brats.L3.n\endcsname{97}
\expandafter\gdef\csname R@an.v7.ho.brats.L3.npos\endcsname{1}
\expandafter\gdef\csname R@an.v7.ho.brats.L3.nseed\endcsname{1}
\expandafter\gdef\csname R@an.v7.ho.brats.V7A1.d\endcsname{\ensuremath{-0.06}}
\expandafter\gdef\csname R@an.v7.ho.brats.V7A1.excl\endcsname{0}
\expandafter\gdef\csname R@an.v7.ho.brats.V7A1.hi\endcsname{\ensuremath{+0.51}}
\expandafter\gdef\csname R@an.v7.ho.brats.V7A1.lo\endcsname{\ensuremath{-0.58}}
\expandafter\gdef\csname R@an.v7.ho.brats.V7A1.n\endcsname{97}
\expandafter\gdef\csname R@an.v7.ho.brats.V7A1.npos\endcsname{1}
\expandafter\gdef\csname R@an.v7.ho.brats.V7A1.npos.word\endcsname{one}
\expandafter\gdef\csname R@an.v7.ho.brats.V7A1.nseed\endcsname{3}
\expandafter\gdef\csname R@an.v7.ho.brats.V7A1.nseed.word\endcsname{three}
\expandafter\gdef\csname R@an.v7.ho.brats.V7C1.d\endcsname{\ensuremath{-0.09}}
\expandafter\gdef\csname R@an.v7.ho.brats.V7C1.excl\endcsname{0}
\expandafter\gdef\csname R@an.v7.ho.brats.V7C1.hi\endcsname{\ensuremath{+0.47}}
\expandafter\gdef\csname R@an.v7.ho.brats.V7C1.lo\endcsname{\ensuremath{-0.66}}
\expandafter\gdef\csname R@an.v7.ho.brats.V7C1.n\endcsname{97}
\expandafter\gdef\csname R@an.v7.ho.brats.V7C1.npos\endcsname{2}
\expandafter\gdef\csname R@an.v7.ho.brats.V7C1.nseed\endcsname{3}
\expandafter\gdef\csname R@an.v7.ho.btcv.L1.d\endcsname{\ensuremath{+0.71}}
\expandafter\gdef\csname R@an.v7.ho.btcv.L1.excl\endcsname{0}
\expandafter\gdef\csname R@an.v7.ho.btcv.L1.hi\endcsname{\ensuremath{+1.84}}
\expandafter\gdef\csname R@an.v7.ho.btcv.L1.lo\endcsname{\ensuremath{-0.38}}
\expandafter\gdef\csname R@an.v7.ho.btcv.L1.n\endcsname{6}
\expandafter\gdef\csname R@an.v7.ho.btcv.L1.npos\endcsname{3}
\expandafter\gdef\csname R@an.v7.ho.btcv.L1.nseed\endcsname{3}
\expandafter\gdef\csname R@an.v7.ho.btcv.L3.d\endcsname{\ensuremath{+0.71}}
\expandafter\gdef\csname R@an.v7.ho.btcv.L3.excl\endcsname{0}
\expandafter\gdef\csname R@an.v7.ho.btcv.L3.hi\endcsname{\ensuremath{+1.54}}
\expandafter\gdef\csname R@an.v7.ho.btcv.L3.lo\endcsname{\ensuremath{-0.33}}
\expandafter\gdef\csname R@an.v7.ho.btcv.L3.n\endcsname{6}
\expandafter\gdef\csname R@an.v7.ho.btcv.L3.npos\endcsname{3}
\expandafter\gdef\csname R@an.v7.ho.btcv.L3.nseed\endcsname{3}
\expandafter\gdef\csname R@an.v7.ho.btcv.V7A1.d\endcsname{\ensuremath{+0.49}}
\expandafter\gdef\csname R@an.v7.ho.btcv.V7A1.excl\endcsname{0}
\expandafter\gdef\csname R@an.v7.ho.btcv.V7A1.hi\endcsname{\ensuremath{+2.07}}
\expandafter\gdef\csname R@an.v7.ho.btcv.V7A1.lo\endcsname{\ensuremath{-1.48}}
\expandafter\gdef\csname R@an.v7.ho.btcv.V7A1.n\endcsname{6}
\expandafter\gdef\csname R@an.v7.ho.btcv.V7A1.npos\endcsname{3}
\expandafter\gdef\csname R@an.v7.ho.btcv.V7A1.npos.word\endcsname{three}
\expandafter\gdef\csname R@an.v7.ho.btcv.V7A1.nseed\endcsname{3}
\expandafter\gdef\csname R@an.v7.ho.btcv.V7A1.nseed.word\endcsname{three}
\expandafter\gdef\csname R@an.v7.ho.btcv.V7C1.d\endcsname{\ensuremath{-0.32}}
\expandafter\gdef\csname R@an.v7.ho.btcv.V7C1.excl\endcsname{0}
\expandafter\gdef\csname R@an.v7.ho.btcv.V7C1.hi\endcsname{\ensuremath{+0.60}}
\expandafter\gdef\csname R@an.v7.ho.btcv.V7C1.lo\endcsname{\ensuremath{-1.37}}
\expandafter\gdef\csname R@an.v7.ho.btcv.V7C1.n\endcsname{6}
\expandafter\gdef\csname R@an.v7.ho.btcv.V7C1.npos\endcsname{1}
\expandafter\gdef\csname R@an.v7.ho.btcv.V7C1.nseed\endcsname{3}
\expandafter\gdef\csname R@an.v7.ho.lits.L1.d\endcsname{\ensuremath{+1.13}}
\expandafter\gdef\csname R@an.v7.ho.lits.L1.excl\endcsname{0}
\expandafter\gdef\csname R@an.v7.ho.lits.L1.hi\endcsname{\ensuremath{+2.67}}
\expandafter\gdef\csname R@an.v7.ho.lits.L1.lo\endcsname{\ensuremath{-0.40}}
\expandafter\gdef\csname R@an.v7.ho.lits.L1.n\endcsname{27}
\expandafter\gdef\csname R@an.v7.ho.lits.L1.npos\endcsname{2}
\expandafter\gdef\csname R@an.v7.ho.lits.L1.nseed\endcsname{3}
\expandafter\gdef\csname R@an.v7.ho.lits.L3.d\endcsname{\ensuremath{+0.50}}
\expandafter\gdef\csname R@an.v7.ho.lits.L3.excl\endcsname{0}
\expandafter\gdef\csname R@an.v7.ho.lits.L3.hi\endcsname{\ensuremath{+1.85}}
\expandafter\gdef\csname R@an.v7.ho.lits.L3.lo\endcsname{\ensuremath{-0.69}}
\expandafter\gdef\csname R@an.v7.ho.lits.L3.n\endcsname{27}
\expandafter\gdef\csname R@an.v7.ho.lits.L3.npos\endcsname{2}
\expandafter\gdef\csname R@an.v7.ho.lits.L3.nseed\endcsname{3}
\expandafter\gdef\csname R@an.v7.ho.lits.V7A1.d\endcsname{\ensuremath{+1.65}}
\expandafter\gdef\csname R@an.v7.ho.lits.V7A1.excl\endcsname{0}
\expandafter\gdef\csname R@an.v7.ho.lits.V7A1.hi\endcsname{\ensuremath{+3.55}}
\expandafter\gdef\csname R@an.v7.ho.lits.V7A1.lo\endcsname{\ensuremath{-0.22}}
\expandafter\gdef\csname R@an.v7.ho.lits.V7A1.n\endcsname{27}
\expandafter\gdef\csname R@an.v7.ho.lits.V7A1.npos\endcsname{2}
\expandafter\gdef\csname R@an.v7.ho.lits.V7A1.npos.word\endcsname{two}
\expandafter\gdef\csname R@an.v7.ho.lits.V7A1.nseed\endcsname{3}
\expandafter\gdef\csname R@an.v7.ho.lits.V7A1.nseed.word\endcsname{three}
\expandafter\gdef\csname R@an.v7.ho.lits.V7C1.d\endcsname{\ensuremath{+0.85}}
\expandafter\gdef\csname R@an.v7.ho.lits.V7C1.excl\endcsname{0}
\expandafter\gdef\csname R@an.v7.ho.lits.V7C1.hi\endcsname{\ensuremath{+2.35}}
\expandafter\gdef\csname R@an.v7.ho.lits.V7C1.lo\endcsname{\ensuremath{-0.36}}
\expandafter\gdef\csname R@an.v7.ho.lits.V7C1.n\endcsname{27}
\expandafter\gdef\csname R@an.v7.ho.lits.V7C1.npos\endcsname{2}
\expandafter\gdef\csname R@an.v7.ho.lits.V7C1.nseed\endcsname{3}
\expandafter\gdef\csname R@an.v7.ho.none.excl\endcsname{1}
\expandafter\gdef\csname R@an.v7.on\endcsname{1}
\expandafter\gdef\csname R@an.v7.pres.D2.duodenum.d\endcsname{\ensuremath{-0.45}}
\expandafter\gdef\csname R@an.v7.pres.D2.duodenum.excl\endcsname{0}
\expandafter\gdef\csname R@an.v7.pres.D2.duodenum.hi\endcsname{\ensuremath{+0.07}}
\expandafter\gdef\csname R@an.v7.pres.D2.duodenum.lo\endcsname{\ensuremath{-0.97}}
\expandafter\gdef\csname R@an.v7.pres.D2.duodenum.n\endcsname{160}
\expandafter\gdef\csname R@an.v7.pres.D2.esophagus.d\endcsname{\ensuremath{-1.05}}
\expandafter\gdef\csname R@an.v7.pres.D2.esophagus.excl\endcsname{1}
\expandafter\gdef\csname R@an.v7.pres.D2.esophagus.hi\endcsname{\ensuremath{-0.37}}
\expandafter\gdef\csname R@an.v7.pres.D2.esophagus.lo\endcsname{\ensuremath{-1.93}}
\expandafter\gdef\csname R@an.v7.pres.D2.esophagus.n\endcsname{159}
\expandafter\gdef\csname R@an.v7.pres.D2.gallbladder.d\endcsname{\ensuremath{-0.80}}
\expandafter\gdef\csname R@an.v7.pres.D2.gallbladder.excl\endcsname{0}
\expandafter\gdef\csname R@an.v7.pres.D2.gallbladder.hi\endcsname{\ensuremath{+0.54}}
\expandafter\gdef\csname R@an.v7.pres.D2.gallbladder.lo\endcsname{\ensuremath{-2.04}}
\expandafter\gdef\csname R@an.v7.pres.D2.gallbladder.n\endcsname{151}
\expandafter\gdef\csname R@an.v7.pres.D2.ladrenal.d\endcsname{\ensuremath{-0.44}}
\expandafter\gdef\csname R@an.v7.pres.D2.ladrenal.excl\endcsname{1}
\expandafter\gdef\csname R@an.v7.pres.D2.ladrenal.hi\endcsname{\ensuremath{-0.02}}
\expandafter\gdef\csname R@an.v7.pres.D2.ladrenal.lo\endcsname{\ensuremath{-0.87}}
\expandafter\gdef\csname R@an.v7.pres.D2.ladrenal.n\endcsname{160}
\expandafter\gdef\csname R@an.v7.pres.D2.liver.d\endcsname{\ensuremath{-0.02}}
\expandafter\gdef\csname R@an.v7.pres.D2.liver.excl\endcsname{0}
\expandafter\gdef\csname R@an.v7.pres.D2.liver.hi\endcsname{\ensuremath{+0.17}}
\expandafter\gdef\csname R@an.v7.pres.D2.liver.lo\endcsname{\ensuremath{-0.16}}
\expandafter\gdef\csname R@an.v7.pres.D2.liver.n\endcsname{160}
\expandafter\gdef\csname R@an.v7.pres.D2.radrenal.d\endcsname{\ensuremath{-0.71}}
\expandafter\gdef\csname R@an.v7.pres.D2.radrenal.excl\endcsname{1}
\expandafter\gdef\csname R@an.v7.pres.D2.radrenal.hi\endcsname{\ensuremath{-0.33}}
\expandafter\gdef\csname R@an.v7.pres.D2.radrenal.lo\endcsname{\ensuremath{-1.10}}
\expandafter\gdef\csname R@an.v7.pres.D2.radrenal.n\endcsname{160}
\expandafter\gdef\csname R@an.v7.pres.V7A1.duodenum.d\endcsname{\ensuremath{+0.42}}
\expandafter\gdef\csname R@an.v7.pres.V7A1.duodenum.excl\endcsname{0}
\expandafter\gdef\csname R@an.v7.pres.V7A1.duodenum.hi\endcsname{\ensuremath{+1.07}}
\expandafter\gdef\csname R@an.v7.pres.V7A1.duodenum.lo\endcsname{\ensuremath{-0.22}}
\expandafter\gdef\csname R@an.v7.pres.V7A1.duodenum.n\endcsname{160}
\expandafter\gdef\csname R@an.v7.pres.V7A1.esophagus.d\endcsname{\ensuremath{-0.59}}
\expandafter\gdef\csname R@an.v7.pres.V7A1.esophagus.excl\endcsname{0}
\expandafter\gdef\csname R@an.v7.pres.V7A1.esophagus.hi\endcsname{\ensuremath{+0.18}}
\expandafter\gdef\csname R@an.v7.pres.V7A1.esophagus.lo\endcsname{\ensuremath{-1.63}}
\expandafter\gdef\csname R@an.v7.pres.V7A1.esophagus.n\endcsname{159}
\expandafter\gdef\csname R@an.v7.pres.V7A1.gallbladder.d\endcsname{\ensuremath{-0.27}}
\expandafter\gdef\csname R@an.v7.pres.V7A1.gallbladder.excl\endcsname{0}
\expandafter\gdef\csname R@an.v7.pres.V7A1.gallbladder.hi\endcsname{\ensuremath{+1.14}}
\expandafter\gdef\csname R@an.v7.pres.V7A1.gallbladder.lo\endcsname{\ensuremath{-1.32}}
\expandafter\gdef\csname R@an.v7.pres.V7A1.gallbladder.n\endcsname{151}
\expandafter\gdef\csname R@an.v7.pres.V7A1.ladrenal.d\endcsname{\ensuremath{-0.06}}
\expandafter\gdef\csname R@an.v7.pres.V7A1.ladrenal.excl\endcsname{0}
\expandafter\gdef\csname R@an.v7.pres.V7A1.ladrenal.hi\endcsname{\ensuremath{+0.34}}
\expandafter\gdef\csname R@an.v7.pres.V7A1.ladrenal.lo\endcsname{\ensuremath{-0.46}}
\expandafter\gdef\csname R@an.v7.pres.V7A1.ladrenal.n\endcsname{160}
\expandafter\gdef\csname R@an.v7.pres.V7A1.liver.d\endcsname{\ensuremath{+0.09}}
\expandafter\gdef\csname R@an.v7.pres.V7A1.liver.excl\endcsname{0}
\expandafter\gdef\csname R@an.v7.pres.V7A1.liver.hi\endcsname{\ensuremath{+0.25}}
\expandafter\gdef\csname R@an.v7.pres.V7A1.liver.lo\endcsname{\ensuremath{-0.05}}
\expandafter\gdef\csname R@an.v7.pres.V7A1.liver.n\endcsname{160}
\expandafter\gdef\csname R@an.v7.pres.V7A1.radrenal.d\endcsname{\ensuremath{+0.09}}
\expandafter\gdef\csname R@an.v7.pres.V7A1.radrenal.excl\endcsname{0}
\expandafter\gdef\csname R@an.v7.pres.V7A1.radrenal.hi\endcsname{\ensuremath{+0.55}}
\expandafter\gdef\csname R@an.v7.pres.V7A1.radrenal.lo\endcsname{\ensuremath{-0.33}}
\expandafter\gdef\csname R@an.v7.pres.V7A1.radrenal.n\endcsname{160}
\expandafter\gdef\csname R@an.v7.pres.V7C1.duodenum.d\endcsname{\ensuremath{-0.74}}
\expandafter\gdef\csname R@an.v7.pres.V7C1.duodenum.excl\endcsname{0}
\expandafter\gdef\csname R@an.v7.pres.V7C1.duodenum.hi\endcsname{\ensuremath{+0.04}}
\expandafter\gdef\csname R@an.v7.pres.V7C1.duodenum.lo\endcsname{\ensuremath{-1.43}}
\expandafter\gdef\csname R@an.v7.pres.V7C1.duodenum.n\endcsname{160}
\expandafter\gdef\csname R@an.v7.pres.V7C1.esophagus.d\endcsname{\ensuremath{-1.31}}
\expandafter\gdef\csname R@an.v7.pres.V7C1.esophagus.excl\endcsname{1}
\expandafter\gdef\csname R@an.v7.pres.V7C1.esophagus.hi\endcsname{\ensuremath{-0.56}}
\expandafter\gdef\csname R@an.v7.pres.V7C1.esophagus.lo\endcsname{\ensuremath{-2.33}}
\expandafter\gdef\csname R@an.v7.pres.V7C1.esophagus.n\endcsname{159}
\expandafter\gdef\csname R@an.v7.pres.V7C1.gallbladder.d\endcsname{\ensuremath{-0.41}}
\expandafter\gdef\csname R@an.v7.pres.V7C1.gallbladder.excl\endcsname{0}
\expandafter\gdef\csname R@an.v7.pres.V7C1.gallbladder.hi\endcsname{\ensuremath{+1.04}}
\expandafter\gdef\csname R@an.v7.pres.V7C1.gallbladder.lo\endcsname{\ensuremath{-1.61}}
\expandafter\gdef\csname R@an.v7.pres.V7C1.gallbladder.n\endcsname{151}
\expandafter\gdef\csname R@an.v7.pres.V7C1.ladrenal.d\endcsname{\ensuremath{-0.67}}
\expandafter\gdef\csname R@an.v7.pres.V7C1.ladrenal.excl\endcsname{1}
\expandafter\gdef\csname R@an.v7.pres.V7C1.ladrenal.hi\endcsname{\ensuremath{-0.20}}
\expandafter\gdef\csname R@an.v7.pres.V7C1.ladrenal.lo\endcsname{\ensuremath{-1.15}}
\expandafter\gdef\csname R@an.v7.pres.V7C1.ladrenal.n\endcsname{160}
\expandafter\gdef\csname R@an.v7.pres.V7C1.liver.d\endcsname{\ensuremath{+0.06}}
\expandafter\gdef\csname R@an.v7.pres.V7C1.liver.excl\endcsname{0}
\expandafter\gdef\csname R@an.v7.pres.V7C1.liver.hi\endcsname{\ensuremath{+0.27}}
\expandafter\gdef\csname R@an.v7.pres.V7C1.liver.lo\endcsname{\ensuremath{-0.09}}
\expandafter\gdef\csname R@an.v7.pres.V7C1.liver.n\endcsname{160}
\expandafter\gdef\csname R@an.v7.pres.V7C1.radrenal.d\endcsname{\ensuremath{-0.75}}
\expandafter\gdef\csname R@an.v7.pres.V7C1.radrenal.excl\endcsname{1}
\expandafter\gdef\csname R@an.v7.pres.V7C1.radrenal.hi\endcsname{\ensuremath{-0.41}}
\expandafter\gdef\csname R@an.v7.pres.V7C1.radrenal.lo\endcsname{\ensuremath{-1.08}}
\expandafter\gdef\csname R@an.v7.pres.V7C1.radrenal.n\endcsname{160}
\expandafter\gdef\csname R@an.v7.tta.acdc.L1.d\endcsname{\ensuremath{+0.37}}
\expandafter\gdef\csname R@an.v7.tta.acdc.L1.excl\endcsname{1}
\expandafter\gdef\csname R@an.v7.tta.acdc.L1.hi\endcsname{\ensuremath{+0.55}}
\expandafter\gdef\csname R@an.v7.tta.acdc.L1.lo\endcsname{\ensuremath{+0.20}}
\expandafter\gdef\csname R@an.v7.tta.acdc.L3.d\endcsname{\ensuremath{+0.33}}
\expandafter\gdef\csname R@an.v7.tta.acdc.L3.excl\endcsname{1}
\expandafter\gdef\csname R@an.v7.tta.acdc.L3.hi\endcsname{\ensuremath{+0.48}}
\expandafter\gdef\csname R@an.v7.tta.acdc.L3.lo\endcsname{\ensuremath{+0.19}}
\expandafter\gdef\csname R@an.v7.tta.acdc.P3c.d\endcsname{\ensuremath{-0.08}}
\expandafter\gdef\csname R@an.v7.tta.acdc.P3c.excl\endcsname{0}
\expandafter\gdef\csname R@an.v7.tta.acdc.P3c.hi\endcsname{\ensuremath{+0.03}}
\expandafter\gdef\csname R@an.v7.tta.acdc.P3c.lo\endcsname{\ensuremath{-0.20}}
\expandafter\gdef\csname R@an.v7.tta.acdc.V7A1.d\endcsname{\ensuremath{+0.08}}
\expandafter\gdef\csname R@an.v7.tta.acdc.V7A1.excl\endcsname{0}
\expandafter\gdef\csname R@an.v7.tta.acdc.V7A1.hi\endcsname{\ensuremath{+0.24}}
\expandafter\gdef\csname R@an.v7.tta.acdc.V7A1.lo\endcsname{\ensuremath{-0.08}}
\expandafter\gdef\csname R@an.v7.tta.acdc.V7C1.d\endcsname{\ensuremath{-0.19}}
\expandafter\gdef\csname R@an.v7.tta.acdc.V7C1.excl\endcsname{1}
\expandafter\gdef\csname R@an.v7.tta.acdc.V7C1.hi\endcsname{\ensuremath{-0.04}}
\expandafter\gdef\csname R@an.v7.tta.acdc.V7C1.lo\endcsname{\ensuremath{-0.34}}
\expandafter\gdef\csname R@an.v7.tta.amos.L1.d\endcsname{\ensuremath{+0.41}}
\expandafter\gdef\csname R@an.v7.tta.amos.L1.excl\endcsname{1}
\expandafter\gdef\csname R@an.v7.tta.amos.L1.hi\endcsname{\ensuremath{+0.64}}
\expandafter\gdef\csname R@an.v7.tta.amos.L1.lo\endcsname{\ensuremath{+0.21}}
\expandafter\gdef\csname R@an.v7.tta.amos.L3.d\endcsname{\ensuremath{+0.19}}
\expandafter\gdef\csname R@an.v7.tta.amos.L3.excl\endcsname{1}
\expandafter\gdef\csname R@an.v7.tta.amos.L3.hi\endcsname{\ensuremath{+0.38}}
\expandafter\gdef\csname R@an.v7.tta.amos.L3.lo\endcsname{\ensuremath{0.00}}
\expandafter\gdef\csname R@an.v7.tta.amos.V7A1.d\endcsname{\ensuremath{+0.12}}
\expandafter\gdef\csname R@an.v7.tta.amos.V7A1.excl\endcsname{0}
\expandafter\gdef\csname R@an.v7.tta.amos.V7A1.hi\endcsname{\ensuremath{+0.33}}
\expandafter\gdef\csname R@an.v7.tta.amos.V7A1.lo\endcsname{\ensuremath{-0.07}}
\expandafter\gdef\csname R@an.v7.tta.amos.V7C1.d\endcsname{\ensuremath{-0.39}}
\expandafter\gdef\csname R@an.v7.tta.amos.V7C1.excl\endcsname{1}
\expandafter\gdef\csname R@an.v7.tta.amos.V7C1.hi\endcsname{\ensuremath{-0.20}}
\expandafter\gdef\csname R@an.v7.tta.amos.V7C1.lo\endcsname{\ensuremath{-0.58}}
\expandafter\gdef\csname R@an.v7.tta.kits.L1.d\endcsname{\ensuremath{+1.49}}
\expandafter\gdef\csname R@an.v7.tta.kits.L1.excl\endcsname{1}
\expandafter\gdef\csname R@an.v7.tta.kits.L1.hi\endcsname{\ensuremath{+2.33}}
\expandafter\gdef\csname R@an.v7.tta.kits.L1.lo\endcsname{\ensuremath{+0.68}}
\expandafter\gdef\csname R@an.v7.tta.kits.L3.d\endcsname{\ensuremath{+0.57}}
\expandafter\gdef\csname R@an.v7.tta.kits.L3.excl\endcsname{0}
\expandafter\gdef\csname R@an.v7.tta.kits.L3.hi\endcsname{\ensuremath{+1.25}}
\expandafter\gdef\csname R@an.v7.tta.kits.L3.lo\endcsname{\ensuremath{-0.08}}
\expandafter\gdef\csname R@an.v7.tta.kits.V7A1.d\endcsname{\ensuremath{+1.06}}
\expandafter\gdef\csname R@an.v7.tta.kits.V7A1.excl\endcsname{1}
\expandafter\gdef\csname R@an.v7.tta.kits.V7A1.hi\endcsname{\ensuremath{+1.74}}
\expandafter\gdef\csname R@an.v7.tta.kits.V7A1.lo\endcsname{\ensuremath{+0.45}}
\expandafter\gdef\csname R@an.v7.tta.kits.V7A1.ts.hi\endcsname{\ensuremath{+2.13}}
\expandafter\gdef\csname R@an.v7.tta.kits.V7A1.ts.lo\endcsname{\ensuremath{0.00}}
\expandafter\gdef\csname R@an.v7.tta.kits.V7A1.ts.lo3\endcsname{0.001}
\expandafter\gdef\csname R@an.v7.tta.kits.V7C1.d\endcsname{\ensuremath{+0.47}}
\expandafter\gdef\csname R@an.v7.tta.kits.V7C1.excl\endcsname{0}
\expandafter\gdef\csname R@an.v7.tta.kits.V7C1.hi\endcsname{\ensuremath{+1.02}}
\expandafter\gdef\csname R@an.v7.tta.kits.V7C1.lo\endcsname{\ensuremath{-0.05}}
\expandafter\gdef\csname R@an.v7w.extr.top4.Y2.max\endcsname{1.0}
\expandafter\gdef\csname R@an.v7w.extr.top4.case.max\endcsname{358}
\expandafter\gdef\csname R@an.v7w.extr.top4.case.min\endcsname{149}
\expandafter\gdef\csname R@an.v7w.extr.top4.marginpct\endcsname{59}
\expandafter\gdef\csname R@an.v7w.ho.counts\endcsname{1}
\expandafter\gdef\csname R@an.v7w.ho.outcome\endcsname{1}
\makeatother

\makeatletter
\expandafter\gdef\csname R@cat.X1.params\endcsname{1.06}
\expandafter\gdef\csname R@cat.X1.stage\endcsname{cross-validated}
\expandafter\gdef\csname R@cat.X1.step\endcsname{0.95}
\expandafter\gdef\csname R@cat.X4.params\endcsname{1.06}
\expandafter\gdef\csname R@cat.X4.stage\endcsname{cross-validated}
\expandafter\gdef\csname R@cat.X4.step\endcsname{1.05}
\makeatother

\input{an_extra_wave.tex}
\InputIfFileExists{supp_keys.tex}{}{}
\InputIfFileExists{v1keys.tex}{}{\typeout{WARNING: v1keys.tex missing; run python3 v1keys.py}}

\makeatletter

\newcommand{\Sref}[1]{\@ifundefined{XS@#1}{\textbf{??}}{\@nameuse{XS@#1}}}
\newcommand{\Mref}[1]{\@ifundefined{XM@#1}{\textbf{??}}{\@nameuse{XM@#1}}}
\makeatother
\InputIfFileExists{results/xref_supp.tex}{}{}
\InputIfFileExists{results/xref_main.tex}{}{}

\newcommand{\cmark}{\ding{51}}
\newcommand{\xmark}{\ding{55}}

\usepackage{colortbl}                    
\usetikzlibrary{shapes.geometric}        
\definecolor{fsagent}{HTML}{2A78D6}      
\definecolor{fsres}{HTML}{EB6834}        
\definecolor{fsneutral}{HTML}{898781}    
\definecolor{fsdefault}{HTML}{52514E}    
\definecolor{hero}{HTML}{1C5CAB}         
\definecolor{herotint}{HTML}{EAF2FB}     
\definecolor{fsink}{HTML}{0B0B0B}
\definecolor{fsinksec}{HTML}{52514E}
\definecolor{fsmuted}{HTML}{898781}
\definecolor{fsgrid}{HTML}{E1E0D9}
\definecolor{fsaxis}{HTML}{C3C2B7}
\definecolor{fsband}{HTML}{F0EFEC}
\definecolor{fsred}{HTML}{E34948}
\definecolor{opres}{HTML}{4A3AA7}
\definecolor{opcnx}{HTML}{1BAF7A}
\definecolor{opplain}{HTML}{E1E0D9}
\tikzset{
  fsop/.style={draw=fsinksec, line width=0.4pt, rounded corners=1pt, inner sep=2pt, align=center, font=\sffamily\scriptsize},
  fsop res/.style={fsop, fill=opres, text=white},
  fsop cnx/.style={fsop, fill=opcnx, text=fsink},
  fsop plain/.style={fsop, fill=opplain, text=fsink},
  fsop trans/.style={fsop, fill=white, text=fsink},
}
\newif\ifresultsfrozen \resultsfrozenfalse
\newcommand{\pend}[1][]{\ifresultsfrozen{\color{fsmuted}--}\else\todo{pending\if\relax\detokenize{#1}\relax\else\ (#1)\fi}\fi}

\DeclareRobustCommand{\fsglyph}[1]{\tikz[baseline=-0.55ex]{#1}}
\DeclareRobustCommand{\lgAgent}{\fsglyph{\fill[fsagent] (0,0) circle (0.6ex);}}
\DeclareRobustCommand{\lgHero}{\fsglyph{\filldraw[fill=fsagent, draw=hero, line width=0.5pt] (0,0) circle (0.75ex);}}

\DeclareRobustCommand{\lgResM}{\fsglyph{\fill[fsres] (-0.52ex,-0.52ex) rectangle (0.52ex,0.52ex);}}
\DeclareRobustCommand{\lgResMatched}{\fsglyph{\fill[fsres] (0,-0.7ex) -- (0.7ex,0) -- (0,0.7ex) -- (-0.7ex,0) -- cycle;}}

\DeclareRobustCommand{\lgSeedZero}{\fsglyph{\draw[fsinksec, line width=0.55pt] (0,0) circle (0.52ex);}}
\DeclareRobustCommand{\lgFresh}{\fsglyph{\fill[fsinksec, opacity=0.6] (0,0) circle (0.32ex);}}
\DeclareRobustCommand{\lgFreshMean}{\fsglyph{\draw[fsinksec, line width=0.35pt] (-0.95ex,0) -- (0.95ex,0); \fill[fsinksec] (0,0) circle (0.6ex);}}

\DeclareRobustCommand{\lgBand}{\fsglyph{\fill[fsband] (-0.95ex,-0.6ex) rectangle (0.95ex,0.6ex);}}
\DeclareRobustCommand{\lgCI}{\fsglyph{\draw[fsinksec, line width=0.55pt] (-0.95ex,0) -- (0.95ex,0);}}

\InputIfFileExists{v12dryrun_keys.tex}{\typeout{WARNING: v12dryrun_keys.tex loaded -- this build carries PLACEHOLDERS, not results}}{}

\providecommand{\archS}{\mbox{R1$+$R3}\xspace}
\providecommand{\archM}{\mbox{LeanNeXt}\xspace}
\providecommand{\archA}{\mbox{R1$+$R2}\xspace}
\providecommand{\archR}{\mbox{LeanNeXt+}\xspace}
\providecommand{\archL}{\mbox{LeanNeXt (wide)}\xspace}
\providecommand{\archRule}{\mbox{LeanNeXt+}\xspace}
\newcommand{\papertitleV}{Can AI Design a Better nnU-Net?\\ A Hypothesis-Driven Architecture Discovery Loop from Failures to Design Rules}

\newcommand{\cPar}[1]{\R{an.arch_profile.ratio.#1.kits.params_vs_default}}
\newcommand{\cMac}[1]{\R{an.arch_profile.ratio.#1.kits.gmac_vs_default}}
\newcommand{\cStep}[1]{\R{an.arch_profile.ratio.#1.kits.step_vs_default}}

\newcommand{\cMem}[1]{\R{an.arch_profile.#1.kits.peak}}
\newcommand{\aPar}[1]{\R{an.arch_profile.#1.kits.params}}
\newcommand{\aMac}[1]{\R{an.arch_profile.#1.kits.gmac}}
\newcommand{\aStep}[1]{\R{an.arch_profile.#1.kits.step}}
\newcommand{\cvD}[2]{\R{v1.cv.#1.#2.d}}

\newcommand{\cvDM}[1]{\resIf{r17.adopt.on}{\R{v1.cv.#1.V7A1cov.d}}{\R{v1.cv.#1.V7A1.d}}}

\newcommand{\cvDK}{\resIf{r20.on}{\R{r20.d2}}{\cvD{kits}{V7A1}}}

\newcommand{\frD}[1]{\R{an.stockref.kits.#1.delta_stock}}

\newcommand{\hoD}[2]{\R{v1.ho.#1.#2.d}}

\makeatletter\expandafter\gdef\csname R@v12.cuts\endcsname{1}\makeatother

\newcommand{\dsub}[1]{\,{\scriptsize(#1)}}

\newcommand{\nWaves}{\ifstate{an.lineage.waves}{8}{eight}{\R{an.lineage.waves}}}
\newcommand{\nIdeas}{\R{an.counts.ideas}}

\newcommand{\nTrained}{\R{an.counts.designs}}

\expandafter\gdef\csname XS@endofbody\endcsname{S13}
\expandafter\gdef\csname XS@eq:grn\endcsname{S1}
\expandafter\gdef\csname XS@fig:s-ablation\endcsname{S5}
\expandafter\gdef\csname XS@fig:s-curves\endcsname{S3}
\expandafter\gdef\csname XS@fig:s-frontier\endcsname{S9}
\expandafter\gdef\csname XS@fig:s-gallery\endcsname{S8}
\expandafter\gdef\csname XS@fig:s-hu\endcsname{S13}
\expandafter\gdef\csname XS@fig:s-lineage\endcsname{S10}
\expandafter\gdef\csname XS@fig:s-medshape\endcsname{S12}
\expandafter\gdef\csname XS@fig:s-organs\endcsname{S7}
\expandafter\gdef\csname XS@fig:s-pareto\endcsname{S1}
\expandafter\gdef\csname XS@fig:s-pctgallery\endcsname{S15}
\expandafter\gdef\csname XS@fig:s-qual\endcsname{S14}
\expandafter\gdef\csname XS@fig:s-r35\endcsname{S6}
\expandafter\gdef\csname XS@fig:s-roadmap\endcsname{S11}
\expandafter\gdef\csname XS@fig:s-screen\endcsname{S4}
\expandafter\gdef\csname XS@fig:s-stagecost\endcsname{S2}
\expandafter\gdef\csname XS@sec:s-ablfig\endcsname{S4.3}
\expandafter\gdef\csname XS@sec:s-arch\endcsname{S2}
\expandafter\gdef\csname XS@sec:s-boundary\endcsname{S4.7}
\expandafter\gdef\csname XS@sec:s-cases\endcsname{S12}
\expandafter\gdef\csname XS@sec:s-curves\endcsname{S4.8}
\expandafter\gdef\csname XS@sec:s-cv\endcsname{S4.1}
\expandafter\gdef\csname XS@sec:s-designs\endcsname{S5}
\expandafter\gdef\csname XS@sec:s-fresh\endcsname{S4.2}
\expandafter\gdef\csname XS@sec:s-frontier\endcsname{S4.9}
\expandafter\gdef\csname XS@sec:s-heldout\endcsname{S4.12}
\expandafter\gdef\csname XS@sec:s-limits\endcsname{S11}
\expandafter\gdef\csname XS@sec:s-locked\endcsname{S4.11}
\expandafter\gdef\csname XS@sec:s-med\endcsname{S4.13}
\expandafter\gdef\csname XS@sec:s-med2\endcsname{S9}
\expandafter\gdef\csname XS@sec:s-mednext\endcsname{S13}
\expandafter\gdef\csname XS@sec:s-panc\endcsname{S10}
\expandafter\gdef\csname XS@sec:s-pctgallery\endcsname{S12.2}
\expandafter\gdef\csname XS@sec:s-protocol\endcsname{S1}
\expandafter\gdef\csname XS@sec:s-qual\endcsname{S12.1}
\expandafter\gdef\csname XS@sec:s-r33\endcsname{S4.5}
\expandafter\gdef\csname XS@sec:s-r34\endcsname{S7.3}
\expandafter\gdef\csname XS@sec:s-r35\endcsname{S4.6}
\expandafter\gdef\csname XS@sec:s-r3off\endcsname{S4.4}
\expandafter\gdef\csname XS@sec:s-r4\endcsname{S7.1}
\expandafter\gdef\csname XS@sec:s-results\endcsname{S4}
\expandafter\gdef\csname XS@sec:s-rs\endcsname{S6}
\expandafter\gdef\csname XS@sec:s-rule4\endcsname{S7}
\expandafter\gdef\csname XS@sec:s-sat\endcsname{S8}
\expandafter\gdef\csname XS@sec:s-splits\endcsname{S1}
\expandafter\gdef\csname XS@sec:s-stepstone\endcsname{S4.10}
\expandafter\gdef\csname XS@sec:s-team\endcsname{S3}
\expandafter\gdef\csname XS@sec:s-wave9\endcsname{S7.2}
\expandafter\gdef\csname XS@tab:s-ablation\endcsname{S19}
\expandafter\gdef\csname XS@tab:s-arch\endcsname{S3}
\expandafter\gdef\csname XS@tab:s-bos\endcsname{S34}
\expandafter\gdef\csname XS@tab:s-boundary\endcsname{S15}
\expandafter\gdef\csname XS@tab:s-compute\endcsname{S10}
\expandafter\gdef\csname XS@tab:s-cost\endcsname{S4}
\expandafter\gdef\csname XS@tab:s-costratio\endcsname{S5}
\expandafter\gdef\csname XS@tab:s-cv\endcsname{S12}
\expandafter\gdef\csname XS@tab:s-designs\endcsname{S24}
\expandafter\gdef\csname XS@tab:s-extrfp\endcsname{S28}
\expandafter\gdef\csname XS@tab:s-fresh\endcsname{S13}
\expandafter\gdef\csname XS@tab:s-funnel\endcsname{S8}
\expandafter\gdef\csname XS@tab:s-heldout\endcsname{S17}
\expandafter\gdef\csname XS@tab:s-hu\endcsname{S33}
\expandafter\gdef\csname XS@tab:s-ledger\endcsname{S36}
\expandafter\gdef\csname XS@tab:s-llm\endcsname{S6}
\expandafter\gdef\csname XS@tab:s-margin\endcsname{S18}
\expandafter\gdef\csname XS@tab:s-med\endcsname{S23}
\expandafter\gdef\csname XS@tab:s-ops\endcsname{S11}
\expandafter\gdef\csname XS@tab:s-panc\endcsname{S35}
\expandafter\gdef\csname XS@tab:s-pub19\endcsname{S14}
\expandafter\gdef\csname XS@tab:s-r31\endcsname{S26}
\expandafter\gdef\csname XS@tab:s-r33\endcsname{S20}
\expandafter\gdef\csname XS@tab:s-r34\endcsname{S29}
\expandafter\gdef\csname XS@tab:s-r35\endcsname{S21}
\expandafter\gdef\csname XS@tab:s-reg\endcsname{S9}
\expandafter\gdef\csname XS@tab:s-roles\endcsname{S7}
\expandafter\gdef\csname XS@tab:s-rs\endcsname{S25}
\expandafter\gdef\csname XS@tab:s-satbnd\endcsname{S31}
\expandafter\gdef\csname XS@tab:s-satdef\endcsname{S30}
\expandafter\gdef\csname XS@tab:s-satkits\endcsname{S32}
\expandafter\gdef\csname XS@tab:s-stages\endcsname{S2}
\expandafter\gdef\csname XS@tab:s-tta\endcsname{S1}
\expandafter\gdef\csname XS@tab:s-unseen\endcsname{S16}
\expandafter\gdef\csname XS@tab:s-wave9\endcsname{S27}
\expandafter\gdef\csname XS@tab:s-waves\endcsname{S22}
}
\usepackage{colortbl}
\definecolor{linkblue}{rgb}{0.21,0.49,0.74}
\usepackage[pagebackref,breaklinks,colorlinks,allcolors=linkblue]{hyperref}

\title{\papertitleV}
\author{Songtao Jiang \quad Yuan Wang \quad Fucheng Niu \quad Hualiang Wang \quad Jian Wu \quad Zuozhu Liu\\[4pt]
Zhejiang University}

\begin{document}
\twocolumn[{\renewcommand\twocolumn[1][]{#1}%
\maketitle

\begin{center}
\includegraphics[width=0.95\textwidth]{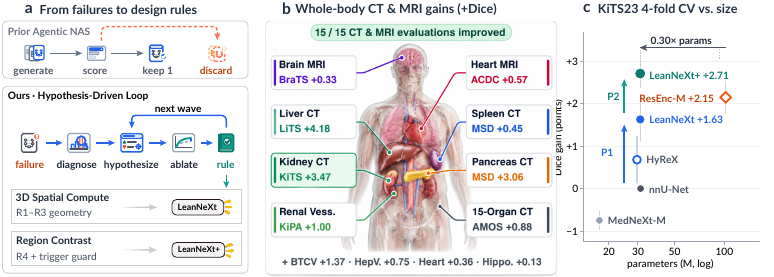}
\captionof{figure}{\textbf{Turning architectural failures into a self-evolving rulebook across whole-body CT and MRI.} (a)~Unlike scalar agentic NAS, our loop diagnoses failures, tests falsifiable hypotheses, and distills general rules. (b)~\archM{} and \archRule{} improve all \R{hyp.ev.rule.n} whole-body CT and MRI evaluations over nnU-Net. (c)~On KiTS23 4-fold CV, Principle~1 yields \archM{} (\R{hyp.bm.kits.V7A1.d} at \R{an.arch_profile.ratio.V7A1.kits.gmac_vs_default}$\times$ MACs) and Principle~2 yields \archR{} (\R{hyp.bm.kits.Y2.d}), surpassing ResEnc-M at \R{fnd.w9.par.vsL1}$\times$ its parameters.}
\label{fig:teaser}
\end{center}
\vspace{2pt}
}]

\begin{abstract}
Neural architecture search and LLM-driven evolution treat evaluation as a scalar fitness black box: winners survive without mechanistic explanation, while failed trials are discarded despite the causal signal their errors carry.
We introduce a \textbf{hypothesis-driven architecture discovery loop} that reframes automated network design from black-box score selection into falsifiable scientific inquiry.
Rather than optimizing task-specific point models, LLM agents use compute profiles and structured error strata to formulate prospective hypotheses, test counterfactual controls under a frozen pipeline, and distill confirmed mechanisms into executable \texttt{trigger}\,$\rightarrow$\,\texttt{action} design rules.
We evaluate this paradigm on \textbf{nnU-Net}, a stringent testbed whose self-configuring pipeline eliminates recipe confounding while its plain U-Net has withstood years of Transformer, state-space, and NAS challengers at matched compute.
Across \nWaves{} waves, our loop uncovers \textbf{two cross-modality design principles of 3D dense prediction}.
First, a \emph{3D surface-to-volume compute principle}: slimming high-resolution decoders exposes a selective geometric regression on thin anatomy ($\rho=\R{med.shape.rho.D2}$), proving 3D boundary fidelity is bottlenecked by mid-encoder depth rather than fine decoder width and unlocking coarse large-kernel mixing.
Second, a \emph{support-confined region-contrast principle}: nested subtype errors track precision rather than recall because local $3^3$ kernels confuse compartments sharing boundary texture unless given mask-confined region distributions and anatomy-conditioned contrast, abolished by constant controls and gated by pre-training triggers.
These principles yield lightweight \textbf{\archM{}} (\cMac{V7A1}$\times$ MACs, \cStep{V7A1}$\times$ step; \cvDK{} 4-fold CV on KiTS23) and trigger-routed \textbf{\archRule{}}, outperforming random sampling, one-shot LLM generation, and score-selected agent search, improving all \R{hyp.ev.rule.n} CT/MRI evaluations over nnU-Net and surpassing $3.3\times$ larger ResEnc-M on \R{hyp.peak.vsL1.nwin}/\R{hyp.peak.vsL1.ntot} benchmarks at \R{fnd.w9.par.vsL1}$\times$ parameters.
These results establish hypothesis-driven rule distillation from failures as a viable paradigm for automated architecture discovery, yielding interpretable design principles that generalize zero-shot across modalities without task-specific search.
\end{abstract}

\section{Introduction}
\label{sec:intro}

\begin{table*}[t]
\centering\footnotesize
\setlength{\tabcolsep}{3.0pt}
\renewcommand{\arraystretch}{1.00}
\caption{\textbf{Validation-score selection vs.\ falsifiable rule discovery.} Prior agentic systems optimize validation scores to emit per-dataset configurations or point models; our discovery loop treats structured failures as diagnostic signal to compile an executable cross-modality rulebook.}
\label{tab:related}
\resizebox{0.92\textwidth}{!}{%
\begin{tabular}{@{}>{\raggedright\arraybackslash}p{2.35cm}>{\raggedright\arraybackslash}p{2.65cm}>{\raggedright\arraybackslash}p{2.75cm}>{\raggedright\arraybackslash}p{2.85cm}>{\raggedright\arraybackslash}p{2.85cm}>{\raggedright\arraybackslash}p{2.85cm}@{}}
\toprule
Paradigm / System & Primary goal & Feedback signal & What agents generate & Experimental role & Final output\\
\midrule
Auto-nnU-Net~\cite{becktepe2025autonnunet} & Per-dataset config & Multi-fidelity val.\ Dice & Hyperparameter/arch.\ configs & Rank candidates (10--1000 ep.) & Task configuration\\
AgentNAS~\cite{jeong2026agentnas} & Per-task model & Proxy val.\ score (30\,\% ep.) & Seed networks \& recipes & Rank top-8 at full budget & Task network + recipe\\
Camyla~\cite{gao2026camyla} & End-to-end paper & Aggregate benchmark score & Candidate architectures & Keep best design per dataset & Per-dataset network\\
\midrule
\rowcolor{herotint}\textbf{Ours} & \textbf{Discover design rules} & \textbf{Profiles + failure patterns} & \textbf{Falsifiable hypotheses} & \textbf{Test prospective predictions} & \textbf{\archM{} / \archRule{}}\\
\bottomrule
\end{tabular}}
\end{table*}

Designing a better neural network for one dataset is not the same as discovering transferable principles of how to design one.
Classical neural architecture search and recent LLM evolution loops~\cite{becktepe2025autonnunet,gao2026camyla,jeong2026agentnas,lu2024aiscientist,novikov2025alphaevolve} rely on \emph{scalar score selection}: candidates are ranked by aggregate validation metrics, winners survive without mechanistic explanation, and lower-scoring trials representing the vast majority of compute are discarded as sunk cost.
Yet in empirical inquiry, controlled failures provide the primary diagnostic signal, because \emph{where and how} an intervention breaks localizes which inductive bias was violated.
\textbf{Can LLM agents move from black-box selection of point architectures to falsifiable discovery of transferable design rules?}

Our core premise is that \textbf{network failures are structured diagnostic signals rather than scalar deficits.}
In 3D dense prediction, aggregate Dice conflates distinct bottlenecks, including misplaced spatial arithmetic, thin-boundary under-capacity, and subtype confusion across compartments sharing local texture.
Scalar ranking obscures these causes, yielding dataset-specific point models that neither transfer zero-shot nor accumulate reusable design knowledge.
We instead formulate a \textbf{hypothesis-driven architecture discovery loop} that executes an \emph{observe--hypothesize--test--distill} cycle: diagnosing compute profiles and error strata, committing \emph{ex ante} to directional predictions (including predicted regressions) and falsification criteria before GPU allocation, running matched counterfactual controls, and distilling confirmed mechanisms into executable \texttt{trigger}\,$\rightarrow$\,\texttt{action} rules.

\begin{figure*}[t]
\centering
\includegraphics[width=0.94\textwidth]{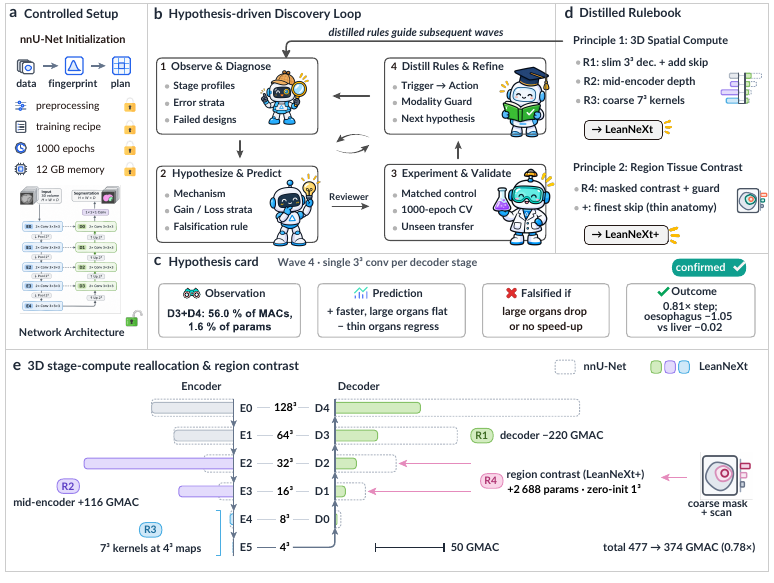}
\caption{\textbf{Hypothesis-driven architecture discovery loop and discovered 3D network topology.}
\textbf{(a)}~Recipe-frozen nnU-Net testbed (Sec.~\ref{sec:loop}).
\textbf{(b)}~Four-stage agentic discovery loop across \nWaves{} waves with reviewer audit (Secs.~\ref{sec:team}--\ref{sec:evidence}).
\textbf{(c)}~Wave-4 hypothesis card committing \emph{ex ante} to directional predictions and falsification criteria (Sec.~\ref{sec:wave4}).
\textbf{(d)}~Distilled cross-modality rulebook (Sec.~\ref{sec:rules}, Tab.~\ref{tab:rules}).
\textbf{(e)}~Stage-wise GMACs on KiTS23 drawn to scale: \archM{} reallocates fine-decoder compute (R1) into mid-encoder depth (R2) and coarse $\R{mr.r3.kits.kmin}^3$ blocks (R3; \R{sc15.base.total.gmac}$\rightarrow$\R{sc15.V7A1.total.gmac}\,GMAC, \R{an.arch_profile.ratio.V7A1.kits.gmac_vs_default}$\times$; Sec.~\ref{sec:roadmap}), while \archRule{} injects masked region contrast at \R{r4.kits.stages} (R4; Sec.~\ref{sec:gains}).}
\label{fig:framework}
\label{fig:arch}
\end{figure*}

We test this discovery loop on \textbf{nnU-Net}~\cite{isensee2021nnunet}, the standard 3D biomedical segmentation framework~\cite{isensee2024revisited}.
nnU-Net poses a uniquely stringent testbed for automated discovery: its self-configuring pipeline freezes preprocessing, patch geometry, and optimization so no gain can come from recipe tuning, while its plain U-Net topology has repeatedly defeated modern Transformer, state-space, and NAS architectures at matched compute~\cite{isensee2024revisited}.
Iterating across \nWaves{} waves under an identical 1000-epoch schedule where only network topology varies (Fig.~\ref{fig:framework}), our loop evaluates \nIdeas{} hypotheses, refutes over 85\,\% via pre-training audits and controlled replications, and distills \textbf{two cross-modality design principles of 3D volumetric representation}:

\textbf{First, a 3D surface-to-volume compute principle governs where spatial arithmetic pays.}
nnU-Net's two finest decoder stages consume \R{mr.r1.kits.gmac}\,\% of MACs for \R{mr.r1.kits.params}\,\% of parameters.
Slimming them saves compute (R1) while inducing a predicted geometric regression on thin, high-surface-to-volume anatomy ($\rho=\R{med.shape.rho.D2}$ on AMOS22).
Turning this failure into a general rule, agents reinvest saved MACs into mid-encoder depth to restore thin boundaries across CT organs and cardiac MRI (R2; $\rho=\R{r35.rho}$, $p=\R{r35.p}$ on AMOS22) and unlock coarse $7^3$ mixing where kernels span the feature map (R3), showing that \emph{where compute sits} outweighs total model scale.

\textbf{Second, a support-confined region-contrast principle resolves the topological blindness of local kernels.}
Across 3D volumetric imaging (solid tumour vs.\ fluid cyst on CT, enhancing glioma rim vs.\ necrotic core on MRI, or dense multi-organ scenes), compartments sharing local $3^3$ boundary texture separate only via region-wide intensity moments and core--rim--host contrast.
Finding that KiTS23 tumour Dice tracks precision rather than recall in cyst-bearing kidneys (Fig.~\ref{fig:cyst}), agents pool mask-confined region distributions and anatomy-conditioned contrast from detached coarse predictions (R4, abolished by constant-input controls) whenever pre-training label topology indicates nested or dense multi-compartment targets.

These principles yield \textbf{two zero-shot models}: lightweight \textbf{\archM{}} (R1--R3; Fig.~\ref{fig:teaser}), gaining \cvDK{} 4-fold CV Dice on KiTS23 (\R{hyp.peak.kits.V7A1.d} on the primary split) at \textbf{\cMac{V7A1}$\times$ MACs and \cStep{V7A1}$\times$ step time}, and trigger-routed \textbf{\archRule{}}, reaching \R{hyp.bm.kits.Y2.d} 4-fold CV (\R{hyp.peak.kits.Rule.d} primary split; \R{hyp.peak.kits.R5.dice} [\R{hyp.peak.kits.R5.d}] with eroded-core pooling) while surpassing $3.3\times$ larger ResEnc-M on \textbf{\R{hyp.peak.vsL1.nwin}/\R{hyp.peak.vsL1.ntot} evaluations at \R{fnd.w9.par.vsL1}$\times$ parameters}.
Our contributions span three levels: (1)~a \emph{hypothesis-driven architecture discovery loop} that turns diagnostic failures into cumulative design memory (\R{hyp.acc.pre.inb}/\R{hyp.acc.pre.n}$\rightarrow$\R{hyp.acc.post.inb}/\R{hyp.acc.post.n} in-budget hits) and outperforms random sampling, one-shot LLM generation, and score-selected agent search within the same system; (2)~an \emph{executable cross-modality rulebook} governing 3D compute allocation and region contrast; and (3)~\archM{} and \archRule{}, improving all \R{hyp.ev.rule.n} whole-body CT and MRI evaluations without task-specific search.

\section{Related Work}
\label{sec:related}

\textbf{3D segmentation backbones inside self-configuring pipelines.}
nnU-Net~\cite{isensee2021nnunet} distills each dataset into a \emph{fingerprint} to configure an \emph{experiment plan} around a plain U-Net~\cite{ronneberger2015unet}, adapting its outer pipeline while keeping block topology fixed.
Under unified benchmarking~\cite{isensee2024revisited}, transformer~\cite{hatamizadeh2022unetr,hatamizadeh2022swin,he2023swinunetrv2,zhou2021nnformer,wald2026primus}, large-kernel~\cite{lee2023uxnet,roy2023mednext}, state-space~\cite{ma2024umamba,xing2024segmamba,wang2024lkmunet}, and hybrid~\cite{chen2023transunet3d,zhou2025nnwnet,sadegheih2025lhunet} backbones rarely beat CNN U-Nets without inflating parameters, memory, and step time, as seen in Residual-Encoder presets~\cite{nnunetresenc,huang2023stunet}, MedNeXt~\cite{roy2023mednext}, and scaling studies~\cite{roy2025mednextv2,huang2025unetscaling}.
Building on slim decoders~\cite{rahman2025effidec3d}, depthwise kernels~\cite{liu2022convnext,woo2023convnextv2,ding2022scaling,ding2024unireplknet,ma2018shufflenetv2}, and design-space principles~\cite{radosavovic2020designing,tan2019efficientnet}, our discovery loop turns failures into falsifiable rules specifying \emph{where} spatial compute and clinical cues belong.

\textbf{Neural architecture search and LLM-driven discovery.}
Classical 3D medical NAS searches fixed spaces under proxy budgets~\cite{zhu2019vnas,yu2020c2fnas,he2021dints,benmeziane2024mednas}, where random sampling stays competitive~\cite{li2020randomnas,yu2019evaluating,yang2020nas} (Sec.~\Sref{sec:s-rs}).
Recent LLM loops evolve code or networks against scalar evaluators~\cite{romera2024funsearch,chen2023evoprompting,nasir2023llmatic,hu2025adas,xu2026asiarch,novikov2025alphaevolve}, self-modify pipelines~\cite{zhang2026dgm,wang2025hgm,srikanth2026aide2}, or automate manuscripts~\cite{lu2024aiscientist,wu2026medai}.
In medical segmentation (Tab.~\ref{tab:related}), Auto-nnU-Net~\cite{becktepe2025autonnunet}, AgentNAS~\cite{jeong2026agentnas}, and Camyla~\cite{gao2026camyla} rank candidates by validation scores to select task-specific configurations or models; our discovery loop instead tests prospective falsifiable predictions to compile an executable zero-shot rulebook.

\section{Hypothesis-Driven Architecture Discovery Loop}
\label{sec:method}

\newcommand{\pCell}[2]{\resIf{hyp.peak.#1.#2.dice}{\R{hyp.peak.#1.#2.dice}\dsub{\R{hyp.peak.#1.#2.d}}}{--}}%
\begin{table*}[t]
\centering\footnotesize
\setlength{\tabcolsep}{4.8pt}
\renewcommand{\arraystretch}{0.98}
\caption{\textbf{\archM{} (lightweight) and \archRule{} (performance) outperform nnU-Net across all \R{hyp.ev.rule.n} evaluations at lower compute.} KiTS23 cost profile and Dice (\%; primary split per row, with 4-fold CV means in Fig.~\ref{fig:forest} and Tab.~\ref{tab:consist}: on 4-fold CV, \archM{} leads ResEnc-1$\times$ on all 3 design sets by \R{r33.p.kits.d}, \R{r33.p.amos.d}, and \R{r33.p.acdc.d}). \textbf{\archRule{}} routes rules zero-shot from pre-training fingerprints ($^\star$R2 off on Heart MRI; $^\dagger$eroded-core reaches \textbf{\R{hyp.peak.kits.R5.dice}} [\R{hyp.peak.kits.R5.d}], Sec.~\ref{sec:wave9}); \textsc{OOM}: $>12$\,GB.}
\label{tab:main}
\resizebox{0.94\textwidth}{!}{%
\begin{tabular}{@{}l ccccc >{\columncolor{herotint}}c >{\columncolor{herotint}}c@{}}
\toprule
 & nnU-Net~\cite{isensee2021nnunet} & MedNeXt-\R{mn.preset}~\cite{roy2023mednext} & ResEnc-1$\times$~\cite{isensee2024revisited} & ResEnc-M~\cite{isensee2024revisited} & HyReX & \textbf{\archM{}} (ours) & \textbf{\archRule{}} (ours)\\
\midrule
\multicolumn{8}{@{}l}{\emph{Cost on the KiTS23 plan}}\\
Parameters (M) & \aPar{base} & \R{mn.par} & \R{r33.cost.kits.params} & \aPar{L1} & \aPar{L3} & \textbf{\aPar{V7A1}} & \textbf{\R{fnd.w9.par}}\\
GMAC ($128^3$) & \aMac{base} & \R{mn.gmac} & \R{r33.cost.kits.gmac} & \aMac{L1} & \aMac{L3} & \textbf{\aMac{V7A1}} & \R{r4.gmac.kits}\\
Memory (GB) & \cMem{base} & \R{mn.mem} & \R{r33.cost.kits.mem} & \cMem{L1} & \cMem{L3} & \cMem{V7A1} & \textbf{\R{fnd.w9.mem}}\\
Step (ms) & \aStep{base} & \R{mn.step} & \R{r33.cost.kits.step} & \aStep{L1} & \aStep{L3} & \textbf{\aStep{V7A1}} & \R{r4.step.kits}\\
\midrule
\multicolumn{8}{@{}l}{\emph{Design benchmarks (Dice, \%)}}\\
KiTS23 & \R{hyp.peak.kits.base.dice} & \pCell{kits}{MedNeXtM} & \pCell{kits}{L1S} & \pCell{kits}{L1} & \pCell{kits}{L3} & \pCell{kits}{V7A1} & \textbf{\R{hyp.peak.kits.Rule.dice}}\dsub{\R{hyp.peak.kits.Rule.d}}$^\dagger$\\
AMOS22 & \R{hyp.peak.amos.base.dice} & \pCell{amos}{MedNeXtM} & \pCell{amos}{L1S} & \pCell{amos}{L1} & \pCell{amos}{L3} & \textbf{\R{hyp.peak.amos.V7A1.dice}}\dsub{\R{hyp.peak.amos.V7A1.d}} & \textbf{\R{hyp.peak.amos.Rule.dice}}\dsub{\R{hyp.peak.amos.Rule.d}}\\
ACDC & \R{hyp.peak.acdc.base.dice} & {\scriptsize\textsc{OOM}} & \pCell{acdc}{L1S} & \pCell{acdc}{L1} & \pCell{acdc}{L3} & \textbf{\R{hyp.peak.acdc.V7A1.dice}}\dsub{\R{hyp.peak.acdc.V7A1.d}} & \textbf{\R{hyp.peak.acdc.Rule.dice}}\dsub{\R{hyp.peak.acdc.Rule.d}}\\
\midrule
\multicolumn{8}{@{}l}{\emph{Locked test sets (held-out evaluation after freezing all rules)}}\\
AMOS22 & \R{hyp.peak.lamos.base.dice} & -- & \pCell{lamos}{L1S} & \pCell{lamos}{L1} & -- & \textbf{\R{hyp.peak.lamos.V7A1.dice}}\dsub{\R{hyp.peak.lamos.V7A1.d}} & \textbf{\R{hyp.peak.lamos.Rule.dice}}\dsub{\R{hyp.peak.lamos.Rule.d}}\\
ACDC & \R{hyp.peak.lacdc.base.dice} & {\scriptsize\textsc{OOM}} & \pCell{lacdc}{L1S} & \pCell{lacdc}{L1} & -- & \textbf{\R{hyp.peak.lacdc.V7A1.dice}}\dsub{\R{hyp.peak.lacdc.V7A1.d}} & \textbf{\R{hyp.peak.lacdc.Rule.dice}}\dsub{\R{hyp.peak.lacdc.Rule.d}}\\
Pancreas & \R{hyp.peak.lpanc.base.dice} & -- & \pCell{lpanc}{L1S} & \pCell{lpanc}{L1} & -- & \textbf{\R{hyp.peak.lpanc.V7A1.dice}}\dsub{\R{hyp.peak.lpanc.V7A1.d}} & \textbf{\R{hyp.peak.lpanc.Rule.dice}}\dsub{\R{hyp.peak.lpanc.Rule.d}}\\
\resIf{hyp.lkipa.on}{KiPA22 & \R{hyp.peak.lkipa.base.dice} & -- & \pCell{lkipa}{L1S} & \pCell{lkipa}{L1} & -- & \textbf{\R{hyp.peak.lkipa.V7A1.dice}}\dsub{\R{hyp.peak.lkipa.V7A1.d}} & \textbf{\R{hyp.peak.lkipa.Rule.dice}}\dsub{\R{hyp.peak.lkipa.Rule.d}}\\}{}%
\midrule
\multicolumn{8}{@{}l}{\emph{Unseen zero-shot transfer benchmarks}}\\
LiTS & \R{hyp.peak.lits.base.dice} & -- & \pCell{lits}{L1S} & \pCell{lits}{L1} & \pCell{lits}{L3} & \textbf{\R{hyp.peak.lits.V7A1.dice}}\dsub{\R{hyp.peak.lits.V7A1.d}} & \textbf{\R{hyp.peak.lits.Rule.dice}}\dsub{\R{hyp.peak.lits.Rule.d}}\\
Pancreas & \R{hyp.peak.panc.base.dice} & -- & \pCell{panc}{L1S} & \pCell{panc}{L1} & \pCell{panc}{L3} & \textbf{\R{hyp.peak.panc.V7A1.dice}}\dsub{\R{hyp.peak.panc.V7A1.d}} & \textbf{\R{hyp.peak.panc.Rule.dice}}\dsub{\R{hyp.peak.panc.Rule.d}}\\
BTCV & \R{hyp.peak.btcv.base.dice} & -- & \pCell{btcv}{L1S} & \pCell{btcv}{L1} & \pCell{btcv}{L3} & \textbf{\R{hyp.peak.btcv.V7A1.dice}}\dsub{\R{hyp.peak.btcv.V7A1.d}} & \textbf{\R{hyp.peak.btcv.Rule.dice}}\dsub{\R{hyp.peak.btcv.Rule.d}}\\
HepaticVessel & \R{hyp.peak.hepv.base.dice} & -- & \pCell{hepv}{L1S} & \pCell{hepv}{L1} & \pCell{hepv}{L3} & \textbf{\R{hyp.peak.hepv.V7A1.dice}}\dsub{\R{hyp.peak.hepv.V7A1.d}} & \textbf{\R{hyp.peak.hepv.Rule.dice}}\dsub{\R{hyp.peak.hepv.Rule.d}}\\
KiPA22 & \R{hyp.peak.kipa22.base.dice} & -- & \pCell{kipa22}{L1S} & \pCell{kipa22}{L1} & \pCell{kipa22}{L3} & \textbf{\R{hyp.peak.kipa22.V7A1.dice}}\dsub{\R{hyp.peak.kipa22.V7A1.d}} & \textbf{\R{hyp.peak.kipa22.Rule.dice}}\dsub{\R{hyp.peak.kipa22.Rule.d}}\\
Spleen & \R{hyp.peak.spleen.base.dice} & -- & \pCell{spleen}{L1S} & \pCell{spleen}{L1} & \pCell{spleen}{L3} & \textbf{\R{hyp.peak.spleen.V7A1.dice}}\dsub{\R{hyp.peak.spleen.V7A1.d}} & \textbf{\R{hyp.peak.spleen.Rule.dice}}\dsub{\R{hyp.peak.spleen.Rule.d}}\\
BraTS (MRI) & \R{hyp.peak.brats.base.dice} & -- & \pCell{brats}{L1S} & \pCell{brats}{L1} & \pCell{brats}{L3} & \textbf{\R{hyp.peak.brats.V7A1.dice}}\dsub{\R{hyp.peak.brats.V7A1.d}} & \textbf{\R{hyp.peak.brats.Rule.dice}}\dsub{\R{hyp.peak.brats.Rule.d}}\\
Heart (MRI) & \R{hyp.peak.heart.base.dice} & -- & \pCell{heart}{L1S} & \pCell{heart}{L1} & \pCell{heart}{L3} & \pCell{heart}{V7A1} & \textbf{\R{hyp.peak.heart.Rule.dice}}\dsub{\R{hyp.peak.heart.Rule.d}}$^\star$\\
\bottomrule
\end{tabular}}
\end{table*}

Whereas agentic architecture search treats evaluation as a scalar oracle and discards lower-scoring trials, our discovery loop operationalizes falsifiable inquiry over neural design spaces (Fig.~\ref{fig:framework}, Tab.~\ref{tab:related}).
Each iteration produces an \emph{ex-ante hypothesis card} committing to directional predictions and a falsification condition before GPU spend, and an \emph{executable design rule} (\texttt{trigger}\,$\rightarrow$\,\texttt{action}) accumulating epistemic memory.

\subsection{Controlled Research Environment}
\label{sec:loop}
\textbf{Freezing the pipeline isolates architectural effects} (Fig.~\ref{fig:framework}a).
In nnU-Net, each dataset's \emph{fingerprint} (geometry and intensity statistics) determines an \emph{experiment plan} (voxel spacing, 3D patch size, and stage strides).
To prevent recipe confounding, our harness locks the fingerprint, plan, preprocessing, splits, optimizer, loss, 1000-epoch schedule, \R{an.arch_profile.cap.gb}\,GB GPU memory cap, and evaluation code (Sec.~\Sref{sec:s-protocol}), leaving only network topology open.
Candidates train for the full 1000 epochs within nnU-Net's parameter and step-time budget, isolating converged capacity from early-stopping noise.

\subsection{From Observations to Falsifiable Hypotheses}
\label{sec:team}
\label{sec:wave4}
\textbf{Proposals commit to directional predictions and falsification criteria before training} (Fig.~\ref{fig:framework}b,c).
\emph{Observe \& Diagnose} combines stage-wise MAC, memory, and step-time profiles with error decompositions across anatomical classes and clinical strata (boundary vs.\ interior, precision vs.\ recall, cyst-bearing vs.\ cyst-free kidneys) and prior wave failures.
\emph{Hypothesize \& Predict} encodes each diagnosis as a \emph{hypothesis card} (Fig.~\ref{fig:framework}c) stating the observation, mechanism, intervention, strata predicted to \emph{improve} and to \emph{regress}, compute bound, and falsification condition.
Four roles powered by \texttt{qwen3.8-flash-next} in a Claude Code scaffold (Tab.~\Sref{tab:s-llm}) run the loop: a \emph{coordinator} writes wave briefs, \emph{ideators} draft hypothesis cards, an independent \emph{reviewer} audits premises and budgets before GPU allocation, and an \emph{experimenter} implements, profiles, and trains approved designs (Sec.~\Sref{sec:s-team}).

\textbf{Turning a predicted regression into the next hypothesis.}
In Wave~4, profiling showed that nnU-Net's two finest decoder stages consume \R{mr.r1.kits.gmac}\,\% of MACs for \R{mr.r1.kits.params}\,\% of parameters.
An ideator slimmed every decoder stage to one $3^3$ conv (\cStep{D2}$\times$ step time), predicting intact large organs alongside thin-organ regressions.
Both held: KiTS23 rose by \cvD{kits}{D2}, whereas AMOS22 lost \R{an.v7.pres.D2.esophagus.d} on oesophagus and \R{an.v7.pres.D2.radrenal.d} on right adrenal vs.\ \R{an.v7.pres.D2.liver.d} on liver, scaling with \emph{boundary share} (fraction of ground-truth organ voxels on its 3D surface; Spearman $\rho=\R{med.shape.rho.D2}$).
That confirmed regression directly seeded Wave~5's hypothesis: reinvesting saved decoder compute into mid-encoder depth.

\subsection{Experimental Validation and Evidence Accumulation}
\label{sec:evidence}

\begin{table*}[t]
\centering\footnotesize
\setlength{\tabcolsep}{3.6pt}
\renewcommand{\arraystretch}{1.02}
\caption{\textbf{Comparison of architecture design paradigms.} Unlike black-box NAS, random grammar sampling, one-shot LLM generation, uniform scaling, or pre-rule agent search, our discovery loop keeps proposals within budget and eliminates negative transfer.}
\label{tab:paradigms}
\resizebox{0.95\textwidth}{!}{%
\begin{tabular}{@{}llccccc@{}}
\toprule
Design paradigm & Representative / pool & In-budget hits & Cost ($\times$ nnU-Net) & KiTS23 $\Delta$Dice & Unseen Pancreas / HepV $\Delta$Dice & Positive transfer\\
\midrule
Differentiable 3D NAS & A3DS DiNTS~\cite{he2021dints} & 0/1 & \R{an.arch_profile.pub19.a3dsdints.vram}\,GB VRAM & $-17.58$ (\R{an.arch_profile.pub19.a3dsdints.kits}\,\%) & -- & 0/3 design tasks\\
Manual 3D Transformers & SwinUNETR-V2~\cite{he2023swinunetrv2} / nnFormer~\cite{zhou2021nnformer} & 0/2 & \R{an.arch_profile.pub.swinunetrv2.vram} / \R{an.arch_profile.pub.nnformer.vram}\,GB VRAM & \R{v1.pub.swinunetrv2.kits.d} / \R{v1.pub.nnformer.kits.d} & -- & 1/3 design tasks\\
Random grammar sampling & \R{v1.rs.n} random draws (same block library) & \R{v1.rs.nfast}/\R{v1.rs.n} & $\ge$\R{v1.rs.step.min}$\times$ step, 14/15 larger & \R{v1.rs.median} med.\ (best \R{v1.rs.best}) & -- & \R{v1.rs.npos}/\R{v1.rs.n} screens\\
One-shot LLM generation & Single-prompt SP1 / SP2 (Claude Opus 5.5) & 0/2 & \R{v1.sp.1.par}$\times$ par., $\ge$\R{v1.rssp.step.min}$\times$ step & \R{v1.sp.1.d} / \R{v1.sp.2.d} (screen) & -- & Over-budget only\\
Human uniform scaling & ResEnc-M~\cite{isensee2024revisited} / MedNeXt-\R{mn.preset}~\cite{roy2023mednext} & 0/2 & \R{r24.L1.params.r1}$\times$ par.\ / \R{mn.step.kits.r}$\times$ step & \R{hyp.peak.kits.L1.d} / \R{hyp.peak.kits.MedNeXtM.d} & \R{hyp.peak.panc.L1.d} / \R{hyp.peak.hepv.L1.d} (ResEnc-M) & 12/15 (ACDC \textsc{OOM})\\
Score-driven agent (same LLM) & Waves~1--3 (\R{hyp.acc.pre.n} models; top: HyReX) & \R{hyp.acc.pre.inb}/\R{hyp.acc.pre.n} & \cMac{L3}$\times$ MACs, \cStep{L3}$\times$ step & \R{hyp.acc.pre.dk.med} med.\ (\R{hyp.peak.kits.L3.d} CV) & \R{hyp.peak.panc.L3.d} / \R{hyp.peak.hepv.L3.d} (\emph{neg.\ transfer}) & 9/11 benchmarks\\
\midrule
\rowcolor{herotint}\textbf{Discovery loop (ours)} & \textbf{Waves~5--10 (\R{hyp.acc.post.n} wave5--8; \archM{} / \archRule{})} & \textbf{\R{hyp.acc.post.inb}/\R{hyp.acc.post.n}} & \textbf{\cMac{V7A1}$\times$ MACs, \cStep{V7A1}$\times$ step} & \textbf{\R{hyp.acc.post.dk.med} med.\ (\R{hyp.peak.kits.Rule.d} CV$^\dagger$)} & \textbf{\R{hyp.peak.panc.Rule.d} / \R{hyp.peak.hepv.Rule.d} (0 neg.\ transfer)} & \textbf{\R{hyp.ev.rule.n}/\R{hyp.ev.rule.n} evaluations}\\
\bottomrule
\end{tabular}}
\end{table*}

\textbf{Controlled ablations, falsification rate, and multi-stage validation.}
Candidates screen on fold~0 of KiTS23~\cite{kits23} and AMOS22~\cite{ji2022amos} (excluded from all reported 4-fold CV estimates), then advance to 4-fold CV, unseen benchmarks, and locked test sets.
Information-mechanism claims require matched counterfactual controls (\eg, replacing region intensity statistics with constants at identical topology and parameter count).
Falsification is the norm: across \nWaves{} waves, agents proposed \nIdeas{} hypotheses; pre-execution reviewer audits rejected \R{an.process.funnel.cut.total} before GPU spend (\R{an.process.funnel.cut.premise} whose premise was contradicted by prior measurements, \R{an.process.funnel.cut.dominated} dominated by simpler designs, and \R{an.process.funnel.cut.infeasible_memory} exceeding memory), and of \nTrained{} full-budget architectures trained, 14 were refuted during fresh-seed replication or ACDC checks (\eg, global tokens X1 trailing \archM{} by \R{med.nwcv.kits.X1.dice.d} vs.\ \cvD{kits}{V7A1}, or $7^3$ blocks without R2 dropping KiTS23 by \R{v11.r3alone.kits.d}), leaving four validated rules (Tab.~\ref{tab:ledger}; full funnel in Tab.~\Sref{tab:s-funnel}).

\begin{table*}[t]
\centering\footnotesize
\setlength{\tabcolsep}{3.4pt}
\renewcommand{\arraystretch}{1.04}
\caption{\textbf{Hypothesis ledger across the discovery loop: pre-training predictions and 4-fold CV outcomes.} Each hypothesis card commits \emph{ex ante} to the anatomical strata expected to gain and to regress; confirmed mechanisms distill into rules R1--R4, whereas refuted hypotheses (\eg, global anatomy tokens) redirect subsequent waves toward calibrated physical cues.}
\label{tab:ledger}
\resizebox{0.96\textwidth}{!}{%
\begin{tabular}{@{}l>{\raggedright\arraybackslash}p{3.40cm}>{\raggedright\arraybackslash}p{3.20cm}>{\raggedright\arraybackslash}p{3.60cm}>{\raggedright\arraybackslash}p{3.90cm}>{\raggedright\arraybackslash}p{2.05cm}@{}}
\toprule
Rule & Observation (diagnosis) & Hypothesis \& intervention & Prediction before training & 4-fold CV / replication outcome & Verdict\\
\midrule
\textbf{R1} & Two finest decoder stages: \R{mr.r1.kits.gmac}\,\% of MACs, \R{mr.r1.kits.params}\,\% of parameters & One $3^3$ conv + additive skip per decoder stage (replacing concat + two $3^3$ convs) & Faster; large organs flat; thin, boundary-dominated organs regress & \cStep{D2}$\times$ step; KiTS23 \cvD{kits}{D2}; oesophagus \R{an.v7.pres.D2.esophagus.d} vs.\ liver \R{an.v7.pres.D2.liver.d} & Confirmed; regression seeds R2\\
\textbf{R2} & R1's per-organ loss grows with boundary share ($\rho=\R{med.shape.rho.D2}$) & Reinvest saved MACs in mid-encoder depth (E2, E3) & Boundary-dominated classes recover; gain rises with boundary share & \R{r19.r2.amos.d} AMOS22, \R{r19.r2.acdc.d} ACDC (\R{fb.r19.r2.sign.n}/\R{fb.r19.r2.sign.n} folds); $\rho=\R{r35.rho}$ ($p=\R{r35.p}$) & Confirmed\\
\textbf{R3} & Dense $7^3$ kernels are costly at fine stages & $7^3$ ConvNeXtV2 blocks at coarsest stages only & Pays where a $7^3$ kernel spans the feature map & Over R1$+$R2: \R{r19.r3inc.kits.d} KiTS23, \R{r19.r3inc.amos.d} AMOS22; without R2: \R{v11.r3alone.kits.d} & Confirmed with R2\\
-- & KiTS23 tumour Dice tracks precision ($r=\R{ins.corr.prec}$); false tumours in cyst-bearing kidneys & Learned global context (anatomy tokens) at every decoder stage & Fewer false-positive tumours and distant components & \R{med.nwcv.kits.X1.dice.d} over nnU-Net, trailing \archM{} (\cvD{kits}{V7A1}) & Refuted; points to calibrated cues\\
\textbf{R4} & Nested-subtype calls flip at $P{=}0.5$ inside the \R{an.hu.bosniak.lo}--\R{an.hu.bosniak.hi}\,HU window & Masked region-level contrast statistics fed to two decoder stages & Subtype precision rises on complex masses; constant inputs stay flat & \R{fnd.w9.kits.dM} over \archM{}; constants \R{r31.ctx.d}; false tumour/mL cyst \R{fnd.w9.slope.M}$\rightarrow$\R{fnd.w9.slope} & Confirmed\\
\bottomrule
\end{tabular}}
\end{table*}

\begin{figure}[t]
\centering
\includegraphics[width=0.92\columnwidth]{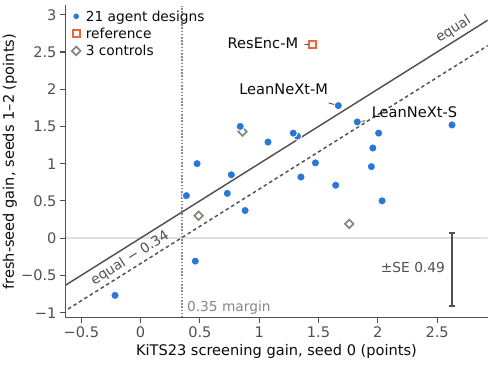}
\caption{\textbf{Single-run screening overstates gains; fresh-seed replication corrects winner's curse.} KiTS23 fold-0 Dice gain over nnU-Net's \R{an.stockref.kits.stock.n} stock runs: screening run (seed~0) vs.\ fresh seeds~1--2 mean. Across the \R{an.replication_all.kits.sel.n} designs admitted by the $+\R{thr.amos}$ screen (filled), fresh seeds regress by \R{r37.mean} points on average (95\,\% CI [\R{r37.lo}, \R{r37.hi}], \R{an.replication_all.kits.sel.ndown}/\R{an.replication_all.kits.sel.n} down; dashed line), whereas \archM{} (LeanNeXt-M) and \archS{} (LeanNeXt-S) hold their gains above the diagonal.}
\label{fig:screen}
\end{figure}

\textbf{Insight 1: Single-seed screening suffers systematic winner's curse; multi-seed and multi-fold verification is essential} (Fig.~\ref{fig:screen}, Tab.~\ref{tab:ledger}).
Because nnU-Net's own run-to-run standard deviation on KiTS23 fold~0 is \R{an.stockref.kits.stock.sd} Dice points across \R{an.stockref.kits.stock.n} identical runs, a null architecture clears a $+\R{thr.amos}$ single-seed screen in \R{an.power_design.trip1_kits_scr}\,\% of draws.
Re-evaluating all \R{an.replication_all.kits.sel.n} screen-admitted agent designs on two fresh seeds reveals a statistically significant selection shrinkage of \R{r37.mean} points (95\,\% CI [\R{r37.lo}, \R{r37.hi}], \R{an.replication_all.kits.sel.ndown}/\R{an.replication_all.kits.sel.n} designs dropping; Fig.~\ref{fig:screen}), and tightening the screening threshold from $+0.35$ to $+1.25$ only worsens mean shrinkage from \R{r37.m.035.shrink} to \R{r37.m.125.shrink} points (Tab.~\Sref{tab:s-margin}).
Consequently, scalar single-run rankers systematically promote lucky noise draws, whereas our loop requires fresh-seed survival, 4-fold cross-validation, and counterfactual mechanism controls (Tab.~\ref{tab:ledger}) before any candidate crystallizes into a rule.

\subsection{From Experimental Evidence to Design Rules}
\label{sec:rules}
\textbf{Distilled rules pair pre-training triggers with architectural actions.}
Computed automatically from the dataset fingerprint and training labels prior to training, each trigger specifies \emph{when} an action applies, bounded by where predicted gains and regressions appeared; the resulting rulebook disciplines later waves and configures unseen datasets zero-shot (Sec.~\ref{sec:rulepred}).

\textbf{Two cross-modality design principles, two zero-shot models} (Fig.~\ref{fig:framework}d,e).
The rulebook crystallizes into two modality-general principles analyzed in Sec.~\ref{sec:analysis} and defines two models (Fig.~\ref{fig:framework}e).
First, the \emph{3D surface-to-volume compute principle}, governed purely by patch grid and 3D boundary geometry, slims each decoder stage to one $3^3$ conv with additive skips (R1), reinvests saved MACs in mid-encoder depth when high-boundary-share anatomy is present (R2), and places $7^3$ ConvNeXtV2~\cite{woo2023convnextv2} blocks at coarse stages where kernels span the map (R3), defining lightweight \textbf{\archM{}} (\cPar{V7A1}$\times$ parameters, \cMac{V7A1}$\times$ MACs, \cStep{V7A1}$\times$ step time).
Second, the \emph{support-confined region-contrast principle} resolves local-kernel blindness when adjacent compartments share $3^3$ boundary texture: whenever pre-training label topology contains multi-compartment scenes (adjacent nested lesion/vessel subtypes or $\ge$10 dense organs; Tab.~\ref{tab:rules}), it pools mask-confined region moments and core--rim--host / anatomy-conditioned contrast from detached coarse predictions into decoder stages via zero-initialized $1^3$ projections (R4, \R{r4.par.extra} parameters; using each plan's native intensity normalization: physical HU quantiles~\cite{silverman2019bosniak} on quantitative CT and z-scored multi-sequence contrast on MRI, plus finest-stage skip retention when maximum boundary share exceeds $0.70$), defining \textbf{\archRule{}}.

\section{Experiments}
\label{sec:experiments}

\subsection{Setup}
\label{sec:setup}

\begin{figure}[t]
\centering
\includegraphics[width=0.90\columnwidth]{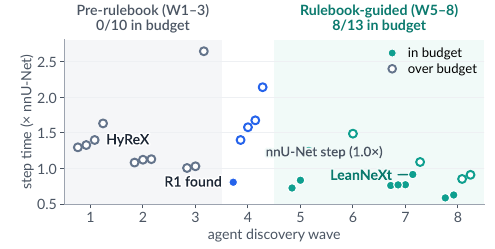}
\caption{\textbf{Distilled rules discipline subsequent agent waves.} KiTS23 step-time ratio by wave (filled: within budget). After R1 in Wave~4, proposals stay within budget at matched screening gain.}
\label{fig:waves}
\end{figure}

\textbf{Benchmarks and baselines.}
Spanning CT and MRI across 12 distinct datasets (\R{hyp.ev.rule.n} evaluation protocols in Tab.~\ref{tab:main}), we develop rules on KiTS23~\cite{kits23}, AMOS22~\cite{ji2022amos}, and ACDC~\cite{bernard2018acdc,isensee2024revisited} (folds~1--4, reserving fold~0 strictly for screening so CV estimates are out-of-sample), evaluate once on held-out locked test cohorts after freezing all rules on AMOS22, ACDC (\R{r18.n} cases each), MSD Pancreas (\R{hyp.lpanc.ncase} cases)\resIf{hyp.lkipa.on}{, and KiPA22 (\R{hyp.lkipa.ncase} cases)}{}, and test zero-shot transfer across eight unseen benchmarks in Tab.~\ref{tab:main} (LiTS~\cite{bilic2023lits}, MSD Pancreas, HepaticVessel, Spleen~\cite{antonelli2022msd,simpson2019msd}, BTCV~\cite{landman2015btcv}, KiPA22~\cite{he2021kipa22}, MSD Heart, and 4-sequence BraTS~\cite{antonelli2022msd}), plus two MRI rule-isolation benchmarks (MSD Hippocampus and 13-organ AMOS22-MRI) in Sec.~\ref{sec:rulepred}.
Inside the identical 1000-epoch, \R{an.arch_profile.cap.gb}\,GB harness, we compare against standard nnU-Net~\cite{isensee2021nnunet}, parameter-matched ResEnc-1$\times$ and scaled ResEnc-M~\cite{isensee2024revisited}, MedNeXt-\R{mn.preset}~\cite{roy2023mednext}, and HyReX, the agents' strongest pre-rule design, scoring Dice, NSD@1mm, and HD95 and verifying under default mirroring and postprocessing.

\subsection{Main Results}
\label{sec:results}

\begin{table*}[t]
\centering\footnotesize
\setlength{\tabcolsep}{2.5pt}
\renewcommand{\arraystretch}{1.02}
\caption{\textbf{Distilled cross-modality design rulebook.} Pre-training triggers (computed prior to training), architectural actions, and isolated 4-fold CV increments (R1 vs.\ nnU-Net, R2 vs.\ R1, R3 vs.\ R1$+$R2, R4 vs.\ \archM{}). Pre-training label topology triggers R4 on multi-compartment scenes (nested lesion/vessel subtypes $C{\ge}2$ with shared boundaries or dense multi-organ scenes $C{\ge}10$, using native CT HU or MRI z-score normalization) and keeps it off on single-target or non-nested tasks.}
\label{tab:rules}
\resizebox{0.96\textwidth}{!}{%
\begin{tabular}{@{}l>{\raggedright\arraybackslash}p{3.00cm}*{7}{c}>{\raggedright\arraybackslash}p{3.40cm}ccc@{}}
\toprule
 & & \multicolumn{3}{c}{Design (CT \& MRI)} & \multicolumn{4}{c}{Unseen (CT \& MRI)} & & \multicolumn{3}{c}{Isolated rule effect ($\Delta$Dice)}\\
\cmidrule(lr){3-5}\cmidrule(lr){6-9}\cmidrule(l){11-13}
Rule & Pre-training trigger & KiTS23 & AMOS22 & ACDC & LiTS & BTCV & Pancreas & BraTS & Architectural action & KiTS23 & AMOS22 & ACDC\\
\midrule
\textbf{R1} & Fine-decoder MAC share (\%) & \R{mr.r1.kits.gmac} & \R{mr.r1.amos.gmac} & \R{mr.r1.acdc.gmac} & \R{mr.r1.lits.gmac} & \R{mr.r1.btcv.gmac} & \R{hyp.trig.panc.r1} & \R{mr.r1.brats.gmac} & Single $3^3$ decoder conv + additive skip & \cvD{kits}{V5B1} & \cvD{amos}{V5B1} & \cvD{acdc}{V5B1}\\
\textbf{R2} & Classes with boundary share $>$\,\R{mr.r2.thr} & \R{mr.r2.kits.nabove}/\R{mr.r2.kits.nclass} & \R{mr.r2.amos.nabove}/\R{mr.r2.amos.nclass} & \R{mr.r2.acdc.nabove}/\R{mr.r2.acdc.nclass} & \R{mr.r2.lits.nabove}/\R{mr.r2.lits.nclass} & \R{mr.r2.btcv.nabove}/\R{mr.r2.btcv.nclass} & \R{hyp.trig.panc.r2} & \R{mr.r2.brats.nabove}/\R{mr.r2.brats.nclass} & Reinvest in encoder depth (4, 3 residual blocks at E2, E3) & \R{r19.r2.kits.d} & \R{r19.r2.amos.d} & \R{r19.r2.acdc.d}\\
\textbf{R3} & Coarsest feature-map size & \R{mr.r3.kits.map} & \R{mr.r3.amos.map} & \R{mr.r3.acdc.map} & \R{mr.r3.lits.map} & \R{mr.r3.btcv.map} & \R{hyp.trig.panc.r3} & \R{mr.r3.brats.map} & Four $7^3$ ConvNeXtV2 blocks at coarsest stages (\R{mr.r3.relmac.coarse}$\times$ dense MACs) & \R{r19.r3inc.kits.d} & \R{r19.r3inc.amos.d} & \R{r19.r3inc.acdc.d}\\
\textbf{R4} & Multi-compartment topology ($C$) & Yes (3) & Yes (15) & Yes (3) & Off & Yes (13) & Off & Yes (3) & Masked region contrast \& anatomy-conditioned moments at decoder & \R{fnd.w9.kits.dM} & \R{fnd.w9.amos.dM} & --\\
\bottomrule
\end{tabular}}
\end{table*}

\begin{figure*}[t]
\centering
\includegraphics[width=0.92\textwidth]{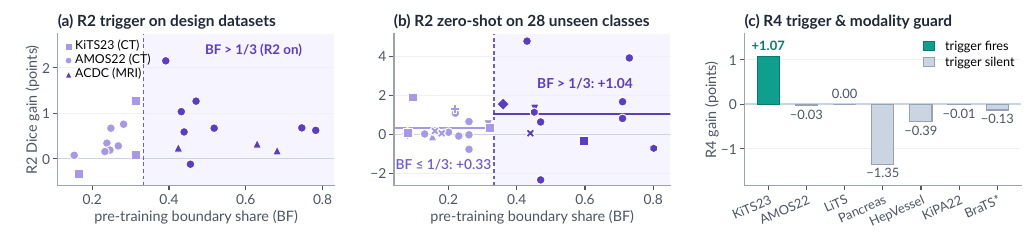}
\caption{\textbf{Pre-training triggers predict where each rule pays across CT and MRI.} (a)~Design datasets: per-class R2 gain vs.\ boundary share (threshold \R{r35.thr}; AMOS22 $\rho=\R{r35.rho}$, $p=\R{r35.p}$). (b)~Across \R{hyp.trig.r2.nclass} unseen CT/MRI classes, R2 adds \R{hyp.trig.r2.above.d} above threshold vs.\ \R{hyp.trig.r2.below.d} below. (c)~CT HU-band R4 trigger: positive on nested cyst--tumour pathology (KiTS23), silent on non-nested CT tasks (forcing CT HU bands on uncalibrated BraTS MRI: $-0.13$, whereas native MRI z-score contrast gains $+0.33$).}
\label{fig:triggers}
\label{fig:r2class}
\end{figure*}

\textbf{\archM{} and \archRule{} beat nnU-Net on all \R{hyp.ev.rule.n} evaluations and surpass $3.3\times$ larger ResEnc-M on \R{hyp.peak.vsL1.nwin}/\R{hyp.peak.vsL1.ntot} rows at \R{fnd.w9.par.vsL1}$\times$ parameters} (Tab.~\ref{tab:main}, Fig.~\ref{fig:forest}).
Across all 4 CV folds (8 paired runs on KiTS23), \archM{} beats parameter-matched ResEnc-1$\times$ on all three design datasets (\R{r33.p.kits.d} on KiTS23 [\R{r33.mean.kits.V7A1} vs.\ \R{r33.mean.kits.L1S}], \R{r33.p.amos.d} on AMOS22 [4/4 folds], and \R{r33.p.acdc.d} on ACDC), while on the primary splits of Tab.~\ref{tab:main}, zero-shot \archRule{} beats ResEnc-1$\times$ on \R{hyp.peak.vsL1S.nwin}/\R{hyp.peak.vsL1S.ntot} rows, gaining \R{hyp.peak.kits.Rule.d} on KiTS23 (\R{hyp.peak.kits.Rule.dice} vs.\ ResEnc-M's \R{hyp.peak.kits.L1.dice}; eroded-core reaches \R{hyp.peak.kits.R5.dice}, \R{hyp.peak.kits.R5.d}), \R{hyp.peak.heart.Rule.d} on Heart MRI, and \R{hyp.peak.other.min}--\R{hyp.peak.other.max} across the remaining rows (\eg, \R{hyp.peak.kipa22.Rule.dice} KiPA22, \R{hyp.peak.btcv.Rule.dice} BTCV, \R{hyp.peak.brats.Rule.dice} BraTS).
Across 4-fold CV, \archRule{} boosts KiTS23 NSD@1mm by \R{hyp.bm.kits.Y2.dnsd1} points (HD95 \R{hyp.bm.kits.Y2.dhd95}\,mm), while \archM{} lifts NSD@1mm across CT and MRI on LiTS (\R{hyp.bm.lits.V7A1.dnsd1}), Pancreas (\R{hyp.bm.panc.V7A1.dnsd1}), BraTS (\R{hyp.bm.brats.V7A1.dnsd1}; HD95 \R{hyp.bm.brats.V7A1.dhd95}\,mm), HepaticVessel (\R{hyp.bm.hepv.V7A1.dnsd1}), KiPA22 (\R{hyp.bm.kipa22.V7A1.dnsd1}), and Spleen (\R{hyp.bm.spleen.V7A1.dnsd1}; HD95 \R{hyp.bm.spleen.V7A1.dhd95}\,mm).

\textbf{Efficiency vs.\ scaled and pre-rule backbones.}
\archM{} uses \cMac{V7A1}$\times$ nnU-Net's MACs on KiTS23 (\R{an.arch_profile.ratio.V7A1.amos.gmac_vs_default}$\times$ on AMOS22/ACDC), cuts peak memory (\cMem{V7A1} vs.\ \cMem{base}\,GB), and speeds inference to \R{an.gpu_measures.infer.V7A1.kits.ratio}--\R{an.gpu_measures.infer.V7A1.amos.ratio}$\times$ (Tabs.~\Sref{tab:s-cost}--\Sref{tab:s-pub19}).
Conversely, ResEnc-M takes \R{r24.L1.params.r1}$\times$ parameters and \cStep{L1}$\times$ step time, MedNeXt-\R{mn.preset} takes \R{mn.step.kits.r}--\R{mn.step.amos.r}$\times$ step time while trailing on KiTS23 (Sec.~\Sref{sec:s-mednext}), and pre-rule HyReX gains \cvD{kits}{L3} at \cStep{L3}$\times$ step time versus \archM{}'s \cvD{kits}{V7A1} at \cStep{V7A1}$\times$.
\resIf{r32.on}{Under default mirroring and postprocessing, \archM{} retains its lead on KiTS23 (\R{r32.kits.V7A1.d2}\ifstate{r32.kits.V7A1.nup2}{4}{, 4/4 folds}{}), AMOS22 (\R{r32.amos.V7A1.d}), ACDC (\R{r32.acdc.V7A1.d}), and locked AMOS22/ACDC (\R{r32.lamos.V7A1.d}, \R{r32.lacdc.V7A1.d}).}{}

\subsection{Why the Discovery Loop Outperforms Prior Paradigms}
\label{sec:paradigms}
\textbf{Evaluating six design paradigms inside the identical harness isolates what hypothesis-driven rule distillation buys} (Tab.~\ref{tab:paradigms}, Fig.~\ref{fig:waves}).
\emph{(i)~Random grammar sampling}: drawing \R{v1.rs.n} valid architectures uniformly from the loop's block grammar yields \R{v1.rs.nfast}/\R{v1.rs.n} models within nnU-Net's step budget ($\ge$\R{v1.rs.step.min}$\times$ step) and a median KiTS23 $\Delta$Dice of \R{v1.rs.median} (best \R{v1.rs.best}), whereas our hypothesis-driven designs (\R{v1.ag.n} models) achieve a median of \R{v1.ag.median} (best \R{v1.ag.best}, \R{v1.ag.npos}/\R{v1.ag.n} positive, and \R{v1.ag.nfast} faster than nnU-Net).
\emph{(ii)~One-shot LLM generation}: single-prompt designs (SP1, SP2) buy accuracy (\R{v1.sp.1.d}, \R{v1.sp.2.d}) only by inflating parameters to \R{v1.sp.1.par}$\times$ nnU-Net ($\ge$\R{v1.rssp.step.min}$\times$ step).
\emph{(iii)~Proxy NAS and manual backbones}: differentiable 3D NAS (DiNTS~\cite{he2021dints}) consumes \R{an.arch_profile.pub19.a3dsdints.vram}\,GB VRAM yet trails on KiTS23 (\R{an.arch_profile.pub19.a3dsdints.kits}\,\%); Transformers (SwinUNETR-V2~\cite{he2023swinunetrv2}, nnFormer~\cite{zhou2021nnformer}) degrade KiTS23 (\R{v1.pub.swinunetrv2.kits.d}, \R{v1.pub.nnformer.kits.d}) and AMOS22 (\R{v1.pub.swinunetrv2.amos.d}, \R{v1.pub.nnformer.amos.d}); MedNeXt-\R{mn.preset} runs out of memory on ACDC; and uniform ResEnc-M scaling requires $3.3\times$ parameters yet drops on HepaticVessel (\R{hyp.peak.hepv.L1.d}) and BraTS (\R{hyp.peak.brats.L1.d}), losing to \archRule{} on \R{hyp.peak.vsL1.nwin}/\R{hyp.peak.vsL1.ntot} benchmarks.
\emph{(iv)~Within-system ablation (score-selected agent search vs.\ rule distillation)}: Waves~1--3 use the \emph{same} \texttt{qwen3.8-flash-next} team, scaffold, block library, and 1000-epoch evaluator, but rank candidates by validation Dice prior to rule distillation. Without failure attribution and rule constraints, \R{hyp.acc.pre.inb}/\R{hyp.acc.pre.n} proposals in Waves~1--3 met budget (median step \R{hyp.acc.pre.step.med}$\times$), and their top score-selected model (HyReX, \cStep{L3}$\times$ step) suffers negative transfer on unseen Pancreas (\R{hyp.peak.panc.L3.d}) and HepaticVessel (\R{hyp.peak.hepv.L3.d}); after rule distillation (Waves~5--8), \R{hyp.acc.post.inb}/\R{hyp.acc.post.n} proposals meet budget (median \R{hyp.acc.post.step.med}$\times$, gain \R{hyp.acc.pre.dk.med} vs.\ \R{hyp.acc.post.dk.med}; Fig.~\ref{fig:waves}), and \archRule{} gains on all \R{hyp.ev.rule.n}/\R{hyp.ev.rule.n} evaluations (\R{hyp.peak.panc.Rule.d} Pancreas, \R{hyp.peak.hepv.Rule.d} HepaticVessel), while profile-anchored cost modeling cuts step prediction error from \R{an.lineage.pred.w5.mean_abs}\,\% (Wave~5) to \R{an.lineage.pred.w7.mean_abs}--\R{an.lineage.pred.w8.mean_abs}\,\% (Waves~7--8).

\begin{figure*}[t]
\centering
\includegraphics[width=0.95\textwidth]{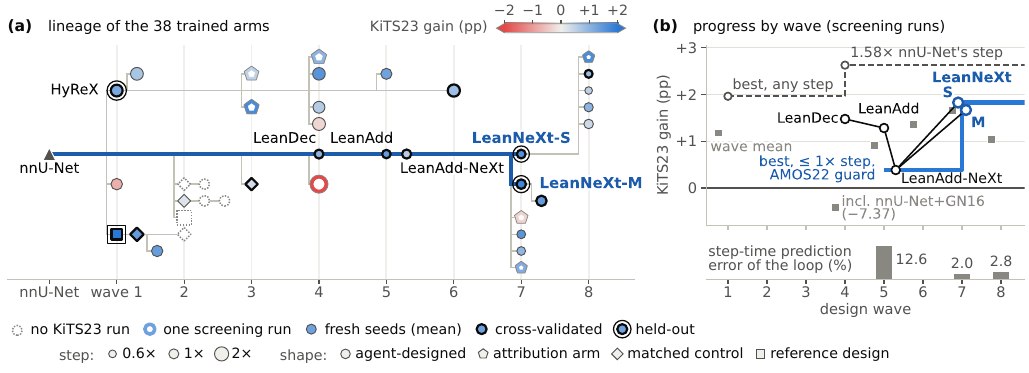}
\caption{\textbf{Evolutionary lineage of the \R{an.counts.trained} full-budget architectures and stepping-stone trajectory across waves.} \textbf{(a)}~Lineage tree by wave, connecting each child architecture to its parent (fill: KiTS23 $\Delta$Dice on fresh seeds; marker area: step time; blue path: $\text{nnU-Net}\rightarrow\text{LeanDec}\rightarrow\text{LeanAdd}\rightarrow\text{LeanAdd-NeXt}\rightarrow\archS{}/\archM{}$). \textbf{(b)}~Best in-budget KiTS23 gain meeting the AMOS22 guard (solid blue) vs.\ unconstrained top score at $1.58\times$ step time (dashed grey), alongside the rapid drop in step-time prediction error from \R{an.lineage.pred.w5.mean_abs}\,\% to \R{an.lineage.pred.w7.mean_abs}--\R{an.lineage.pred.w8.mean_abs}\,\% once agents anchor cost models on measured stage profiles.}
\label{fig:lineage}
\end{figure*}

\textbf{Insight 2: Score-greedy selection prunes the critical stepping stones that rule-guided discovery preserves} (Fig.~\ref{fig:lineage}).
Tracing the complete phylogenetic tree of trained architectures (Fig.~\ref{fig:lineage}a) reveals why black-box keep-the-best evolution stalls: \emph{neither precursor that yielded our final rules was the highest-scoring model of its wave}.
In Wave~4, LeanDec (D2) trailed the wave's top-scoring design, a heavy-decoder HyReX variant (C1), on both KiTS23 (\R{an.stockref.kits.D2.delta_stock} vs.\ \R{an.stockref.kits.C1.delta_stock} on fresh seeds) and AMOS22 (\R{v1.amos0.D2.d} vs.\ \R{v1.amos0.C1.d}), yet LeanDec was the \emph{only} candidate to break below nnU-Net's step time (\cStep{D2}$\times$).
The heavy-decoder winner proved an evolutionary dead end, whereas LeanDec's diagnostic thin-organ regression seeded the entire lean branch.
Similarly, in Wave~5, adding coarse $7^3$ blocks in LeanAdd-NeXt (V5B2) lowered KiTS23 fresh-seed Dice relative to its parent LeanAdd (\R{an.stockref.kits.V5B2.delta_stock} vs.\ \R{an.stockref.kits.V5B1.delta_stock}) while recovering AMOS22 (\R{v1.amos0.V5B2.d} vs.\ \R{v1.amos0.V5B1.d})---an intermediate trade-off that greedy KiTS23 ranking would discard, but which directly parented \archM{} in Wave~7 once combined with R2's mid-encoder depth (Fig.~\ref{fig:lineage}b).

\section{Analysis: What the Loop Discovered}
\label{sec:analysis}
We now dissect the two cross-modality principles into rules R1--R4, test whether their pre-training triggers predict gains on unseen CT and MRI tasks, and evaluate consistency across all folds and runs.

\subsection{Principle 1: 3D Surface-to-Volume Compute}
\label{sec:roadmap}

\begin{figure*}[t]
\centering
\includegraphics[width=0.92\textwidth]{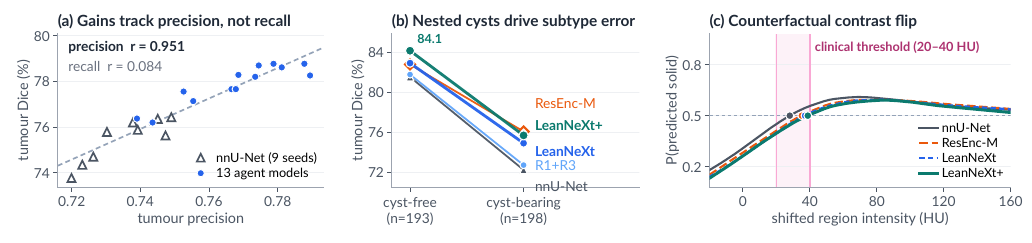}
\caption{\textbf{Principle~2: local kernels miss region-level tissue contrast on nested pathology.} (a)~Across \R{ins.corr.n} checkpoints, KiTS23 tumour Dice tracks precision ($r=\R{ins.corr.prec}$), not recall ($r=\R{ins.corr.rec}$). (b)~Networks drop \R{ins.cyst.pen.L1}--\R{ins.cyst.pen.base} points on cyst-bearing vs.\ cyst-free kidneys; \archRule{} gains \R{hyp.cyst.gain.bear.Y2} on cyst-bearing cases. (c)~Counterfactual intensity probe: every network flips at $P{=}0.5$ inside the \R{an.hu.bosniak.lo}--\R{an.hu.bosniak.hi}\,HU Bosniak threshold~\cite{silverman2019bosniak}.}
\label{fig:cyst}
\end{figure*}

\begin{figure}[t]
\centering
\includegraphics[width=0.94\columnwidth]{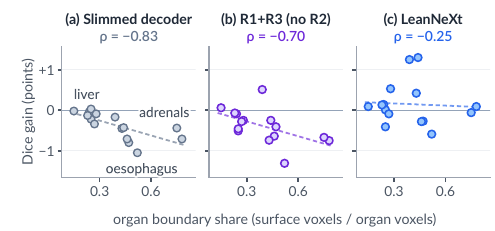}
\caption{\textbf{Decoder slimming penalizes high-boundary-share organs; encoder depth recovers them.} AMOS22 per-organ Dice change over nnU-Net (folds~1--4) across \R{med.amos.norg} organs.}
\label{fig:medshape}
\end{figure}

\textbf{Reallocating decoder arithmetic into mid-encoder depth eliminates the thin-organ penalty and unlocks coarse large kernels} (Tab.~\ref{tab:rules}).
R1 alone lifts KiTS23 by \cvD{kits}{V5B1} (\R{v1.cv.kits.V5B1.npos}/\R{v1.cv.kits.V5B1.n} folds) at \cStep{V5B1}$\times$ step time, but causes an organ-specific AMOS22 drop scaling with boundary share ($\rho=\R{med.shape.rho.D2}$; Fig.~\ref{fig:medshape}a).
Reinvesting saved compute in mid-encoder depth (R2) adds \R{r19.r2.amos.d} on AMOS22 and \R{r19.r2.acdc.d} on ACDC over R1 (\R{fb.r19.r2.sign.n}/\R{fb.r19.r2.sign.n} folds), concentrating gains above the \R{r35.thr} boundary-share threshold (\R{r35.c.amos.above} vs.\ \R{r35.c.amos.below} on AMOS22, $\rho=\R{r35.rho}$, $p=\R{r35.p}$; Figs.~\ref{fig:triggers}a,~\ref{fig:qual}C) and flattening \archM{}'s slope ($\rho=\R{med.shape.rho.V7A1}$; Fig.~\ref{fig:medshape}c).
\resIf{v11.on}{R2 also enables R3: adding $7^3$ blocks without R2 drops KiTS23 by \R{v11.r3alone.kits.d}, whereas R3 atop R1$+$R2 improves all three benchmarks at \cStep{V7A1}$\times$ step time.}{}

\begin{table}[t]
\centering\scriptsize
\setlength{\tabcolsep}{2.4pt}
\renewcommand{\arraystretch}{1.02}
\caption{\textbf{Where compute sits matters more than total size.} KiTS23 matched-cost ablations vs.\ nnU-Net.}
\label{tab:factorial}
\resizebox{0.94\columnwidth}{!}{%
\begin{tabular}{@{}lcccc@{}}
\toprule
Network & Par. & GMAC & Step & $\Delta$Dice\\
\midrule
\rowcolor{herotint}Rules R1$+$R3 & \cPar{V7C1} & \cMac{V7C1} & \cStep{V7C1} & \frD{V7C1}\\
nnU-Net, width $\times$0.70 & \R{v1.abc.7.params} & \R{v1.abc.7.gmac} & \R{v1.abc.7.step} & \resIf{r30.on}{\R{r30.ab.7.d}}{\R{v1.ab.7.kits.dstock}}\\
nnU-Net, width $\times$0.82 & \R{v1.abc.11.params} & \R{v1.abc.11.gmac} & \R{v1.abc.11.step} & \resIf{r30.on}{\R{r30.ab.11.d}}{\R{v1.ab.11.kits.dstock}}\\
EffiDec3D-style decoder & \R{v1.abc.8.params} & \R{v1.abc.8.gmac} & \R{v1.abc.8.step} & \resIf{r30.on}{\R{r30.ab.8.d}}{\R{v1.ab.8.kits.dstock}}\\
\bottomrule
\end{tabular}}
\end{table}

\textbf{Gains stem from targeted stage allocation rather than uniform scaling, slim decoders, or wider receptive fields} (Tab.~\ref{tab:factorial}).
Uniformly narrowing nnU-Net to match R1$+$R3 (\archM{} without R2's mid-encoder depth) in MACs or parameters degrades accuracy (\R{r30.ab.7.d}, \R{r30.ab.11.d}), and EffiDec3D-style channel slimming~\cite{rahman2025effidec3d} at matched MACs stays flat (\R{r30.ab.8.d}) versus \frD{V7C1} for R1$+$R3.
Measuring the 90\,\% effective receptive field radius confirms that \archM{} matches nnU-Net's spatial extent (\R{med.erf.V7A1.r90} vs.\ \R{med.erf.base.r90} voxels; Sec.~\Sref{sec:s-med}), showing coarse $7^3$ blocks strengthen channel--spatial mixing rather than raw field size.
Likewise, against ResEnc's uniform depth schedule scaled to nnU-Net's parameter budget, \archM{} leads by \R{r33.p.amos.d} on AMOS22\ifstate{r33.p.amos.nup}{4}{ (4/4 folds)}{}, \R{r33.p.kits.d} on KiTS23, and \R{r33.p.acdc.d} on ACDC (Sec.~\Sref{sec:s-r33}).
Direct GPU operator profiling (Tab.~\Sref{tab:s-ops}) explains why: removing the second $3^3$ convolution at the two finest decoder stages cuts \R{an.lineage.stagecost.dec_fine.share.min}--\R{an.lineage.stagecost.dec_fine.share.max}\,\% of step time for $\le$\R{an.lineage.stagecost.dec_fine.pshare.max}\,\% of parameters, whereas a depthwise $7^3$ kernel costs \R{op.rel.dw7.s1}$\times$ a dense $3^3$ conv at fine resolution (E1) but only \R{op.rel.dw7.s4}$\times$ at coarse maps (E4).

\begin{table*}[t]
\centering\footnotesize
\setlength{\tabcolsep}{4.0pt}
\renewcommand{\arraystretch}{1.03}
\caption{\textbf{Stage-by-stage architecture specification and arithmetic reallocation on the KiTS23 plan ($128^3$ patch), alongside the $2\times 2$ $\text{R2}\times\text{R3}$ factorial on top of R1.} \textbf{(a)}~Encoder stages E0--E5 and decoder stages D0--D4: parameter count $P$ (M) and forward multiply-adds $G$ (GMAC, deep supervision off). R1 cuts decoder compute from \R{s1.arch.base.dec.gmac} to \R{s1.arch.V7C1.dec.gmac}\,GMAC (\R{s1.arch.V7C1.dec.gmac_pct}\,\% of nnU-Net's decoder); \archM{} reinvests freed parameters into residual depth at E2--E3 (R2) and $7^3$ ConvNeXtV2 blocks at E4--E5 (R3), matching nnU-Net's parameters (\aPar{V7A1} vs.\ \aPar{base}\,M) at \cMac{V7A1}$\times$ MACs. \textbf{(b)}~4-fold CV $\Delta$Dice (points) across the four arms that switch R2 and R3 on top of R1: R3 without R2 degrades KiTS23 (\cvD{kits}{V5B2}), whereas R2 unlocks R3 across all three design benchmarks.}
\label{tab:archspec}
\resizebox{0.94\textwidth}{!}{%
\begin{tabular}{@{}lcc lrr lrr lrr@{}}
\multicolumn{12}{@{}l}{\textbf{(a) Stage-by-stage topology, parameters $P$ (M), and multiply-adds $G$ (GMAC) on KiTS23}}\\
\toprule
 & & & \multicolumn{3}{c}{nnU-Net (default)} & \multicolumn{3}{c}{LeanNeXt-S (R1$+$R3)} & \multicolumn{3}{c}{\textbf{\archM{}} (R1$+$R2$+$R3)}\\
\cmidrule(lr){4-6}\cmidrule(lr){7-9}\cmidrule(l){10-12}
Stage & Map & Ch. & Blocks & $P$ & $G$ & Blocks & $P$ & $G$ & Blocks & $P$ & $G$\\
\midrule
E0 & 128$^3$ & 32 & 2$\times$conv & 0.03 & 59.8 & 2$\times$conv & 0.03 & 59.8 & 2$\times$conv & 0.03 & 59.8 \\
E1 & 64$^3$ & 64 & 2$\times$conv & 0.17 & 43.5 & 2$\times$conv & 0.17 & 43.5 & 2$\times$conv & 0.17 & 43.5 \\
E2 & 32$^3$ & 128 & 2$\times$conv & 0.66 & 21.7 & 2$\times$conv & 0.66 & 21.7 & 4$\times$res & 3.33 & 109.0 \\
E3 & 16$^3$ & 256 & 2$\times$conv & 2.66 & 10.9 & 2$\times$conv & 2.66 & 10.9 & 3$\times$res & 9.77 & 40.0 \\
E4 & 8$^3$ & 320 & 2$\times$conv & 4.98 & 2.5 & 2$\times$conv & 4.98 & 2.5 & conv + 4$\times$NeXt$_{7}$ & 5.13 & 2.6 \\
E5 & 4$^3$ & 320 & 2$\times$conv & 5.53 & 0.4 & conv + 4$\times$NeXt$_{7}$ & 5.68 & 0.4 & conv + 4$\times$NeXt$_{7}$ & 5.68 & 0.4 \\
\midrule
D0 & 8$^3$ & 320 & cat, 2$\times$conv & 9.12 & 4.3 & add, 1$\times$conv & 3.59 & 1.5 & add, 1$\times$conv & 3.59 & 1.5 \\
D1 & 16$^3$ & 256 & cat, 2$\times$conv & 5.97 & 22.1 & add, 1$\times$conv & 2.43 & 7.6 & add, 1$\times$conv & 2.43 & 7.6 \\
D2 & 32$^3$ & 128 & cat, 2$\times$conv & 1.59 & 44.6 & add, 1$\times$conv & 0.71 & 15.6 & add, 1$\times$conv & 0.71 & 15.6 \\
D3 & 64$^3$ & 64 & cat, 2$\times$conv & 0.40 & 89.1 & add, 1$\times$conv & 0.18 & 31.1 & add, 1$\times$conv & 0.18 & 31.1 \\
D4 & 128$^3$ & 32 & cat, 2$\times$conv & 0.10 & 178.2 & add, 1$\times$conv & 0.04 & 62.3 & add, 1$\times$conv & 0.04 & 62.3 \\
heads & 128$^3$ & 3 & 1$^3$ conv & 0.00 & 0.2 & 1$^3$ conv & 0.00 & 0.2 & 1$^3$ conv & 0.00 & 0.2 \\
\midrule
total &  &  & & 31.20 & 477.3 & & 21.11 & 257.0 & & 31.04 & 373.5 \\
 
\bottomrule
\end{tabular}}
\par\medskip
\resizebox{0.90\textwidth}{!}{%
\begin{tabular}{@{}lcc ccc ccc@{}}
\multicolumn{9}{@{}l}{\textbf{(b) $2\times 2$ factorial of mid-encoder depth (R2) $\times$ coarse $7^3$ mixing (R3) atop slim decoder (R1)}}\\
\toprule
Architecture arm & R2 (E2--E3 depth) & R3 (coarse $7^3$) & Params & GMAC & Step & KiTS23 $\Delta$Dice & AMOS22 $\Delta$Dice & ACDC $\Delta$Dice\\
\midrule
LeanAdd (R1 alone) & -- & -- & \cPar{V5B1} & \cMac{V5B1} & \cStep{V5B1} & \cvD{kits}{V5B1} & \cvD{amos}{V5B1} & \cvD{acdc}{V5B1}\\
LeanAdd$+$R2 (\archA{}) & \checkmark & -- & \R{r19.cost.V9R2.params} & \R{r19.cost.V9R2.gmac} & \R{r19.cost.V9R2.step} & \cvD{kits}{V9R2} & \cvD{amos}{V9R2} & \cvD{acdc}{V9R2}\\
LeanAdd-NeXt (R1$+$R3) & -- & \checkmark & \cPar{V5B2} & \cMac{V5B2} & \cStep{V5B2} & \cvD{kits}{V5B2} & \cvD{amos}{V5B2} & \cvD{acdc}{V5B2}\\
\rowcolor{herotint}\textbf{\archM{}} (R1$+$R2$+$R3) & \checkmark & \checkmark & \textbf{\cPar{V7A1}} & \textbf{\cMac{V7A1}} & \textbf{\cStep{V7A1}} & \textbf{\cvD{kits}{V7A1}} & \textbf{\cvD{amos}{V7A1}} & \textbf{\cvD{acdc}{V7A1}}\\
\bottomrule
\end{tabular}}
\end{table*}

\textbf{Insight 3: Mid-encoder depth and coarse large-kernel mixing interact synergistically rather than additively} (Tab.~\ref{tab:archspec}).
Inspecting the exact stage-wise budget (Tab.~\ref{tab:archspec}a) clarifies why R1--R3 form a coupled system: nnU-Net's decoder consumes \R{s1.arch.base.dec.gmac}\,GMAC (\R{s1.arch.base.dec.params}\,M parameters), with D3--D4 alone burning $267.3$\,GMAC on $64^3$--$128^3$ grids.
R1's single-conv additive decoder slashes total decoder cost by $65\,\%$ to \R{s1.arch.V7C1.dec.gmac}\,GMAC (\R{s1.arch.V7C1.dec.params}\,M parameters), freeing $10.23$\,M parameters that R2 reinvests into E2 ($4\times\text{res}$, $109.0$\,GMAC at $32^3$) and E3 ($3\times\text{res}$, $40.0$\,GMAC at $16^3$) while R3 places $7^3$ depthwise blocks at E4--E5 for only $3.0$\,GMAC total.
Crucially, the $2\times 2$ factorial over folds~1--4 (Tab.~\ref{tab:archspec}b) demonstrates a strong positive interaction between R2 and R3: attaching $7^3$ blocks to E4--E5 \emph{without} R2's mid-encoder depth (LeanAdd-NeXt) degrades KiTS23 from \cvD{kits}{V5B1} to \cvD{kits}{V5B2} (\R{v11.r3alone.kits.d}), because shallow two-conv encoder features at E2--E3 lack the semantic abstraction required for global $7^3$ channel--spatial mixing.
Once R2 deepens E2--E3 (lifting AMOS22 by \R{r19.r2.amos.d} and ACDC by \R{r19.r2.acdc.d} over R1), the same coarse $7^3$ blocks reverse sign and boost all three design benchmarks simultaneously (\cvD{kits}{V7A1} KiTS23, \cvD{amos}{V7A1} AMOS22, \cvD{acdc}{V7A1} ACDC).

\subsection{Principle 2: Support-Confined Region Contrast}
\label{sec:gains}
\label{sec:wave9}

\resIf{qual.on}{%
\begin{figure}[t]
\centering
\newcommand{\qw}{0.215}
\newcommand{\qaspA}{1.1519}%
\newcommand{\qaspB}{1.1765}%
\newcommand{\qaspC}{0.6724}%
\newlength{\qpw}\setlength{\qpw}{\qw\columnwidth}%
\newcommand{\qp}[1]{\IfFileExists{figs/v13_qual_#1.pdf}{\includegraphics[width=\qpw]{figs/v13_qual_#1.pdf}}{\fbox{\parbox[c][\qpw]{\qpw}{\centering #1}}}}
\newcommand{\qh}[1]{\makebox[\qpw][c]{\tiny #1}}
\newcommand{\ql}[1]{\makebox[\qpw][c]{\tiny #1}}
\newcommand{\qr}[2]{\raisebox{\dimexpr #1\qpw/2 - 0.04\qpw\relax}{\scriptsize\textbf{#2}}}
\setlength{\tabcolsep}{1.5pt}
\renewcommand{\arraystretch}{0.85}
\begin{tabular}{@{}c@{\hskip 2pt}cccc@{}}
 & \qh{CT \& ref.} & \qh{nnU-Net} & \qh{\archM{}} & \qh{\archR{}}\\
\qr{\qaspA}{A} & \qp{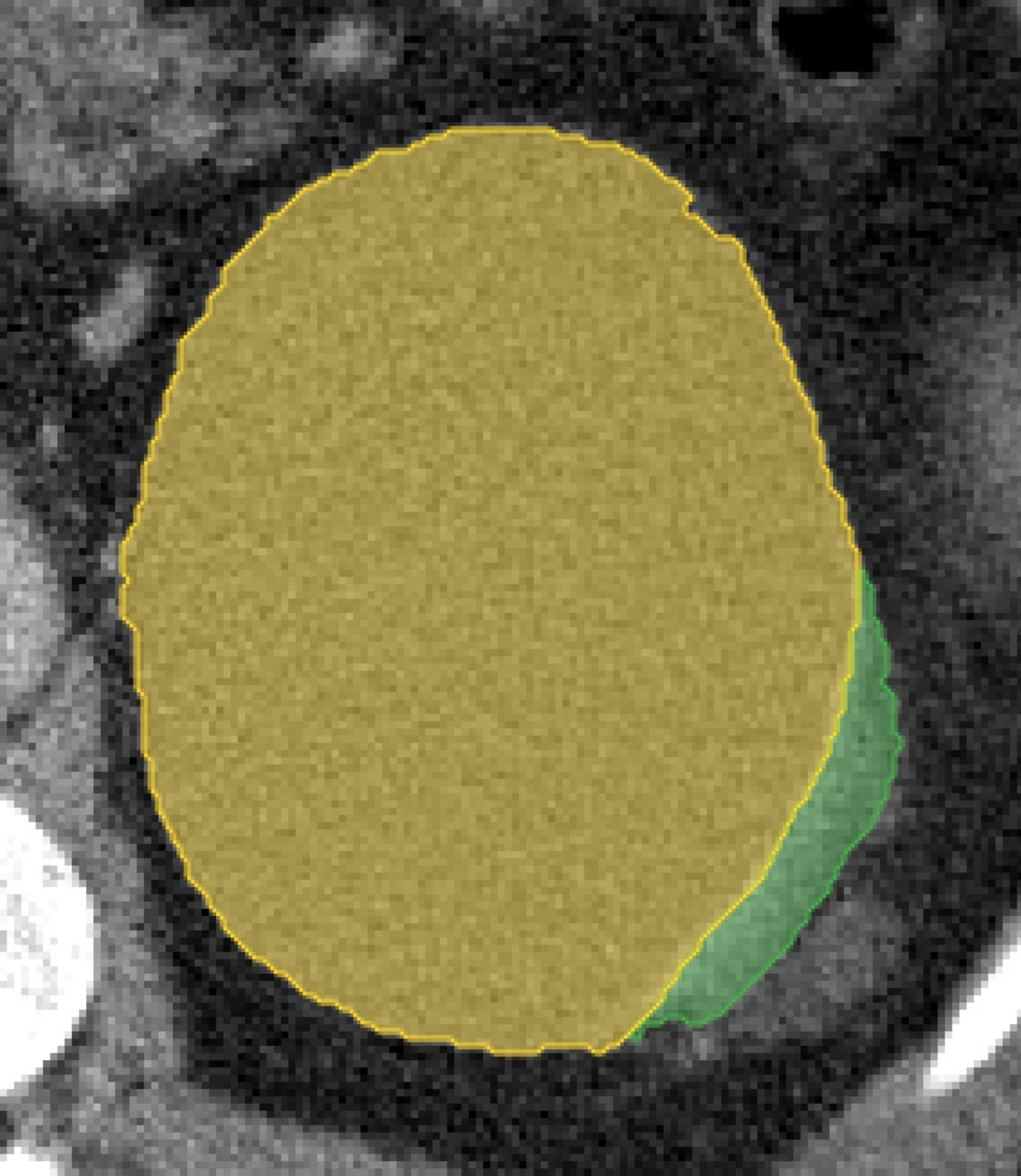} & \qp{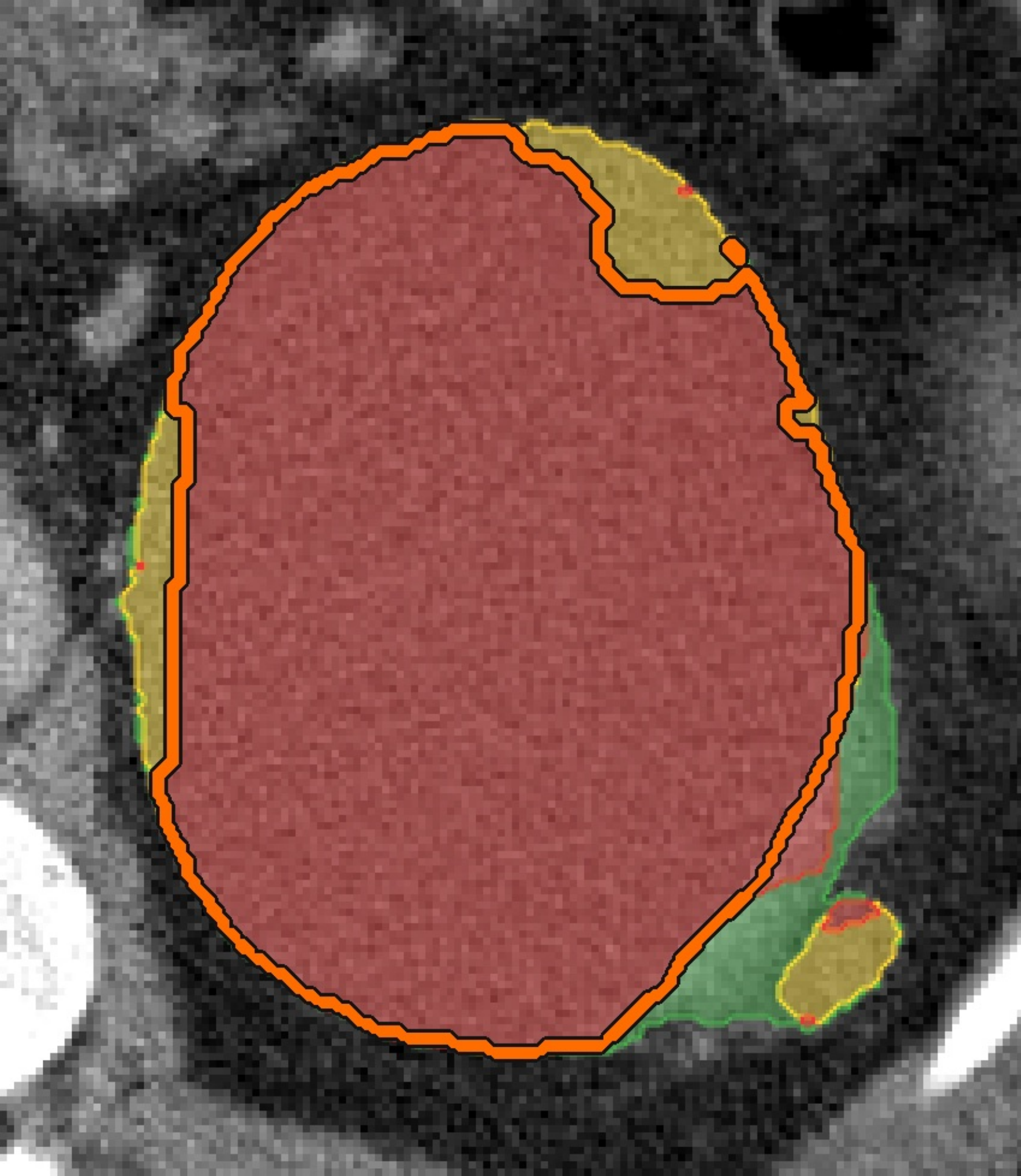} & \qp{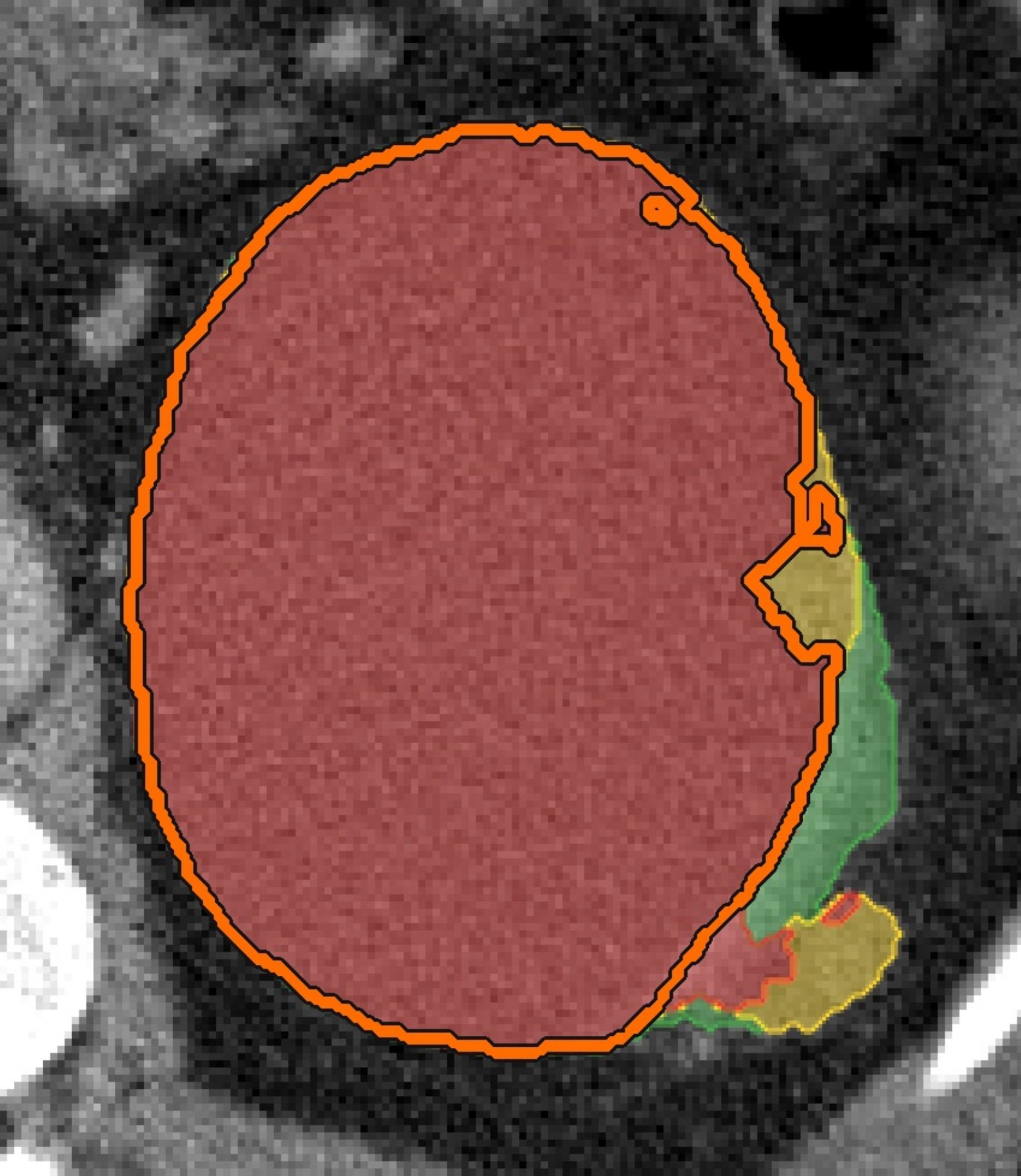} & \qp{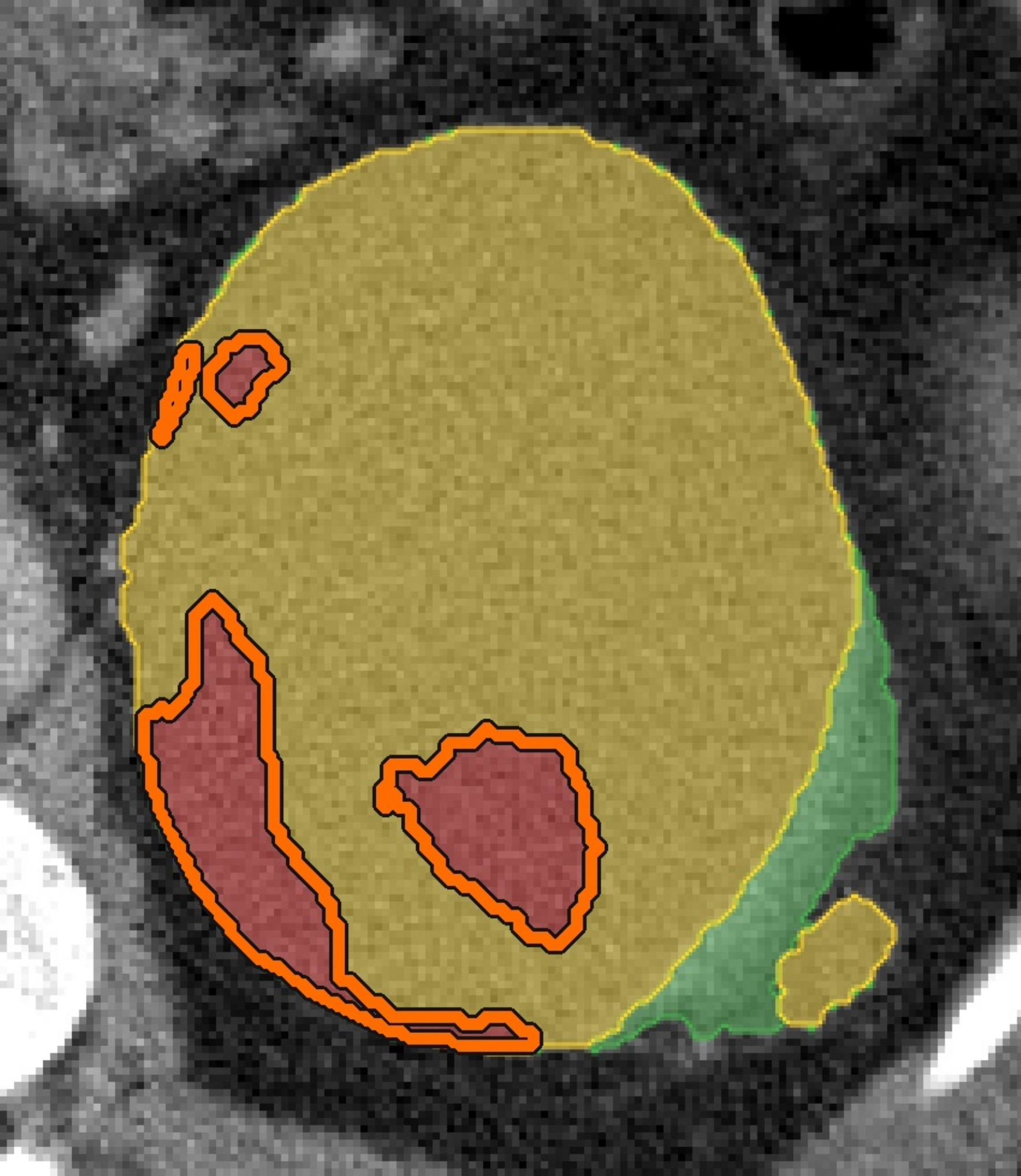}\\
 & \ql{KiTS \R{qual.A.case}} & \ql{\R{qual.A.fp.base}mL\,|\,\R{qual.A.dice.base}} & \ql{\R{qual.A.fp.V7A1}mL\,|\,\R{qual.A.dice.V7A1}} & \ql{\R{qual.A.fp.Y2}mL\,|\,\R{qual.A.dice.Y2}}\\[1.5pt]
\qr{\qaspB}{B} & \qp{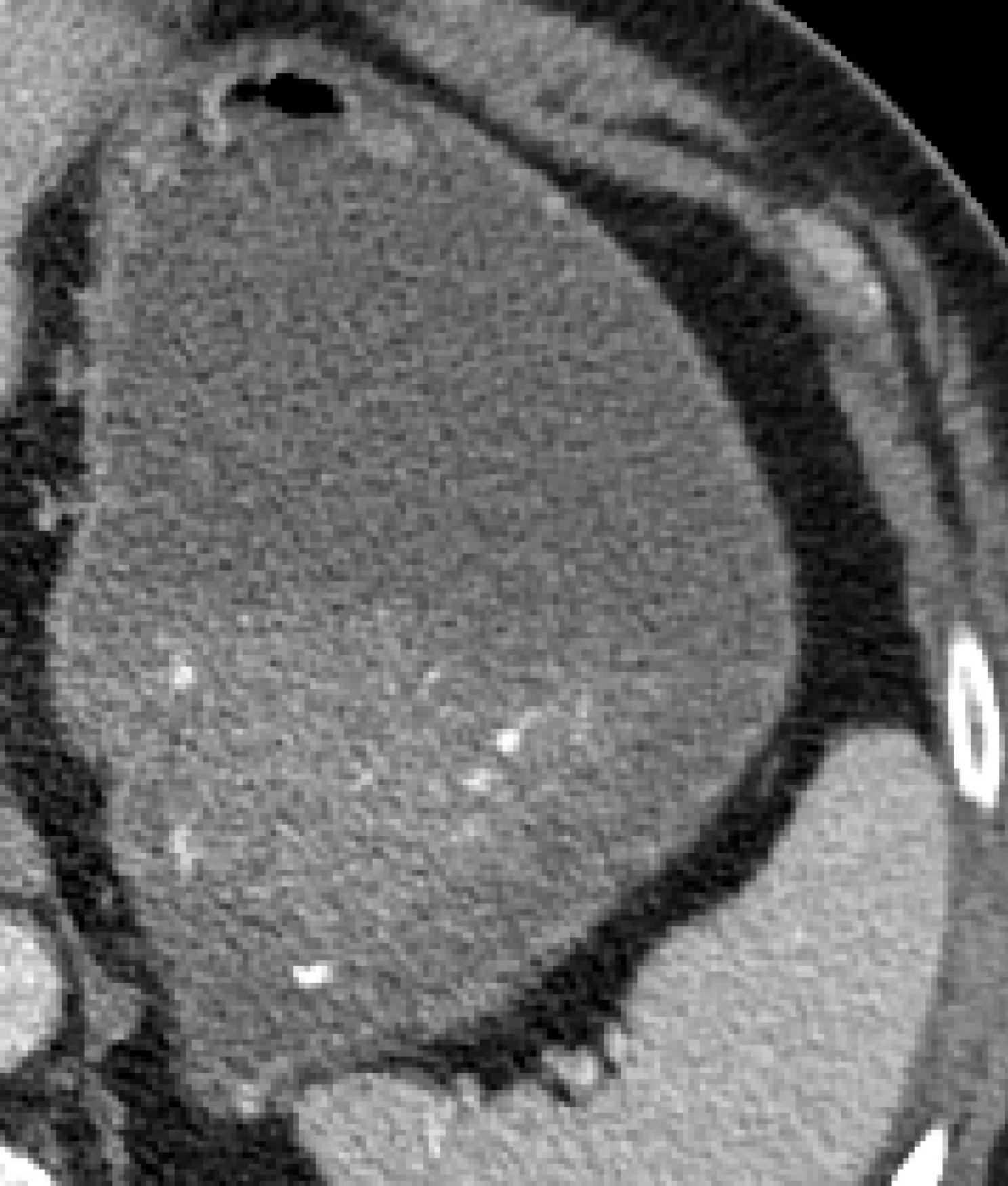} & \qp{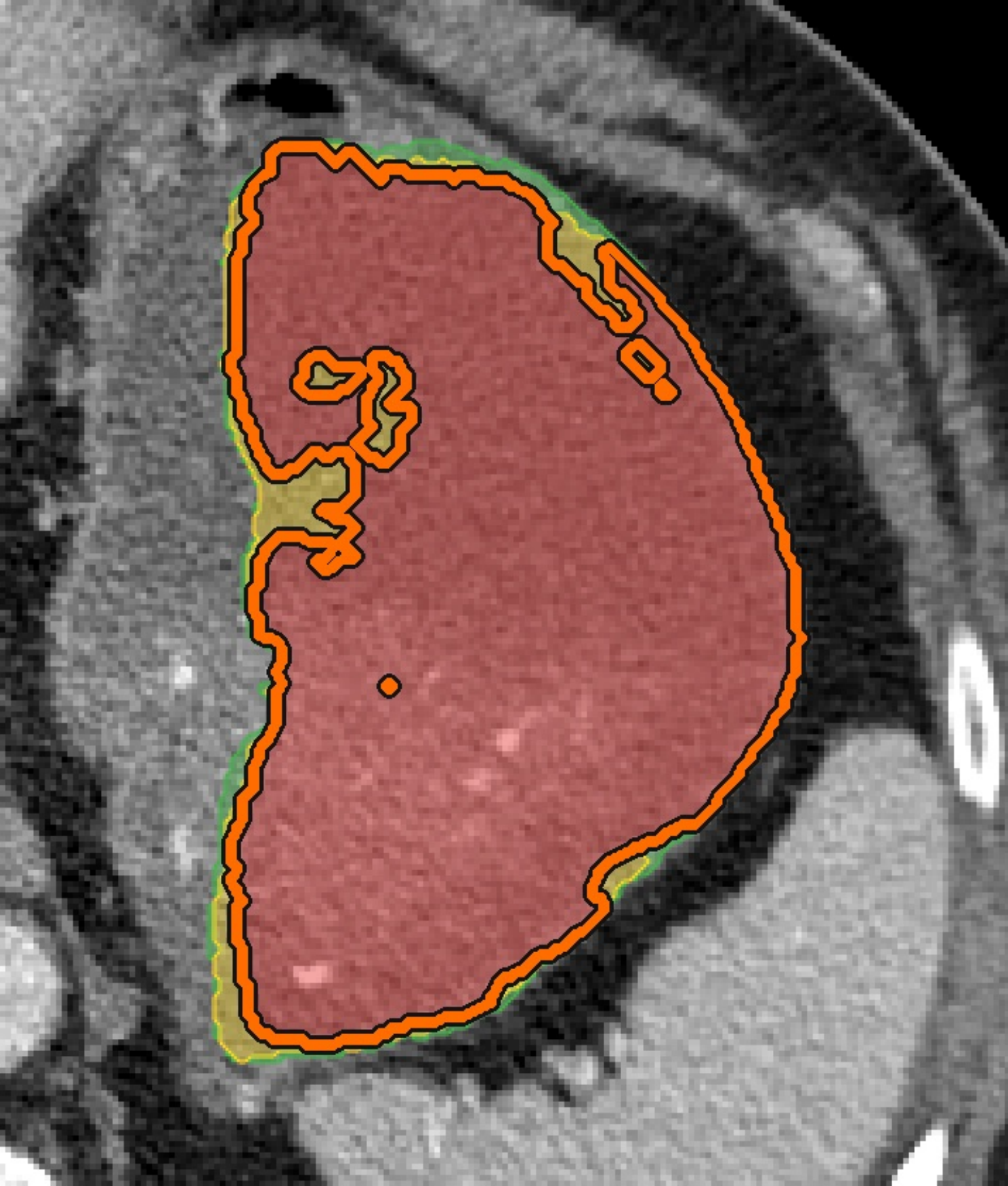} & \qp{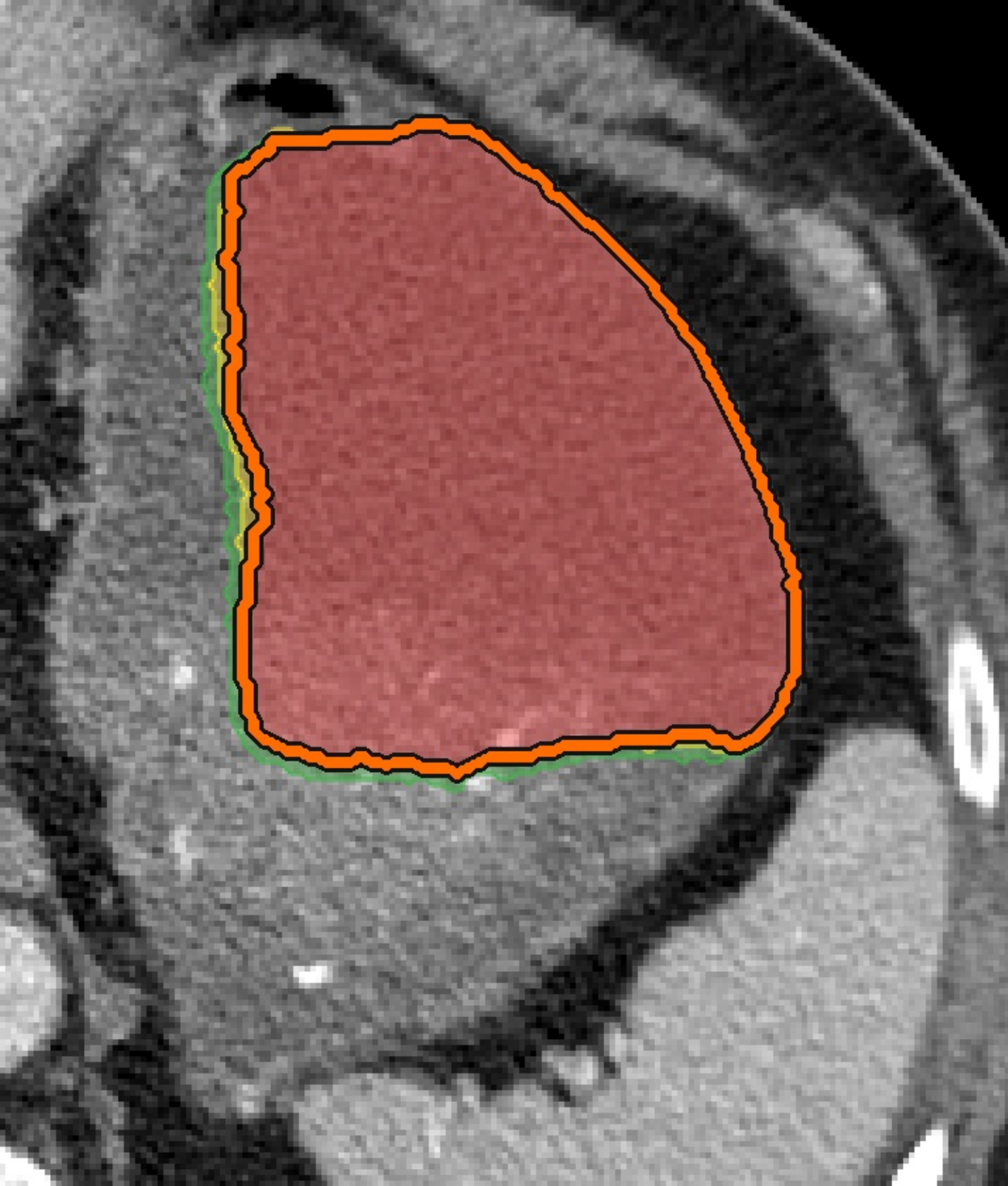} & \qp{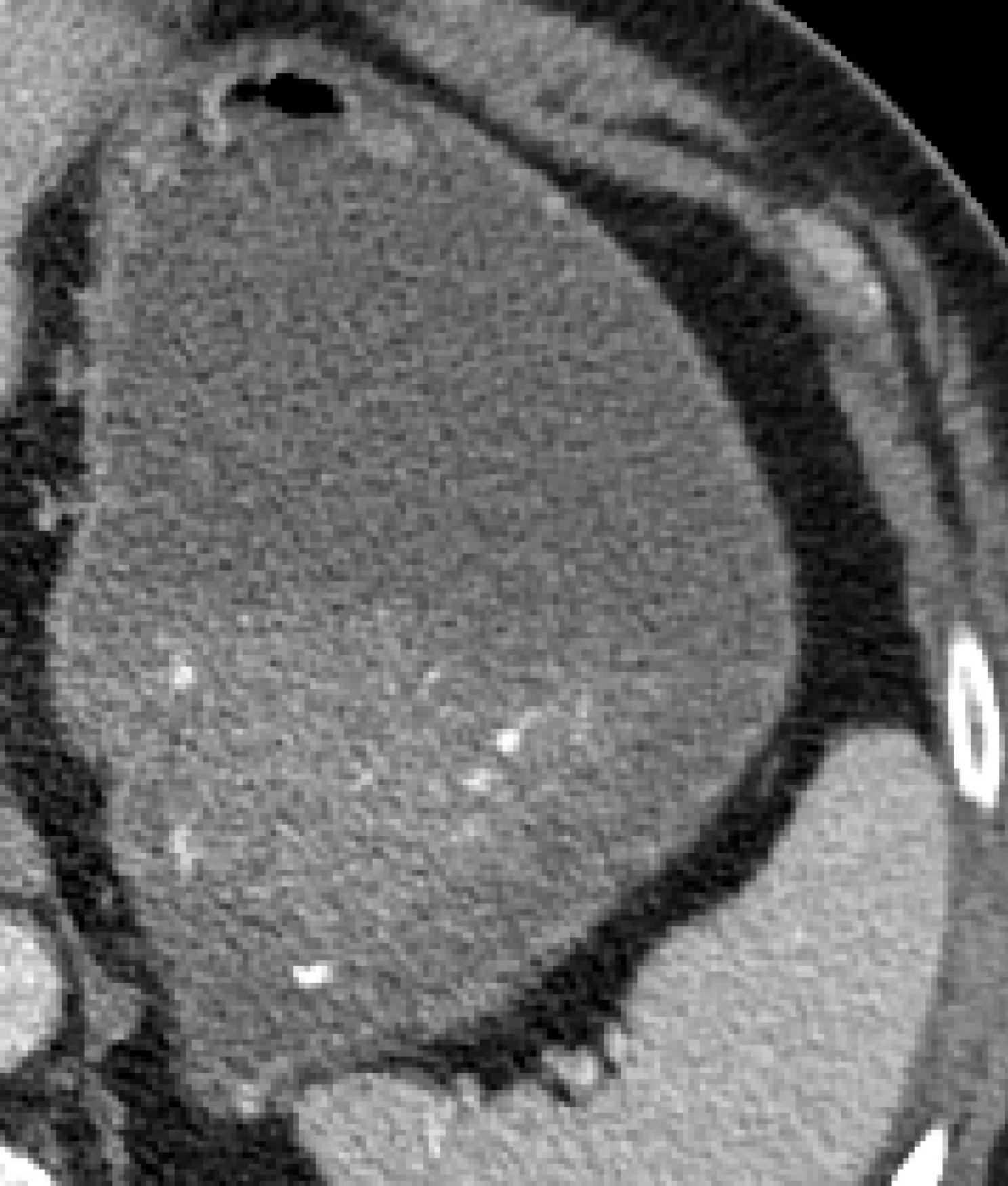}\\
 & \ql{KiTS \R{qual.B.case}} & \ql{\R{qual.B.extr.base}mL\,|\,\R{qual.B.dice.base}} & \ql{\R{qual.B.extr.V7A1}mL\,|\,\R{qual.B.dice.V7A1}} & \ql{\R{qual.B.extr.Y2}mL\,|\,\R{qual.B.dice.Y2}}\\[1.5pt]
 & \qh{CT \& ref.} & \qh{nnU-Net} & \qh{Trimmed dec.} & \qh{\archM{}}\\
\qr{\qaspC}{C} & \qp{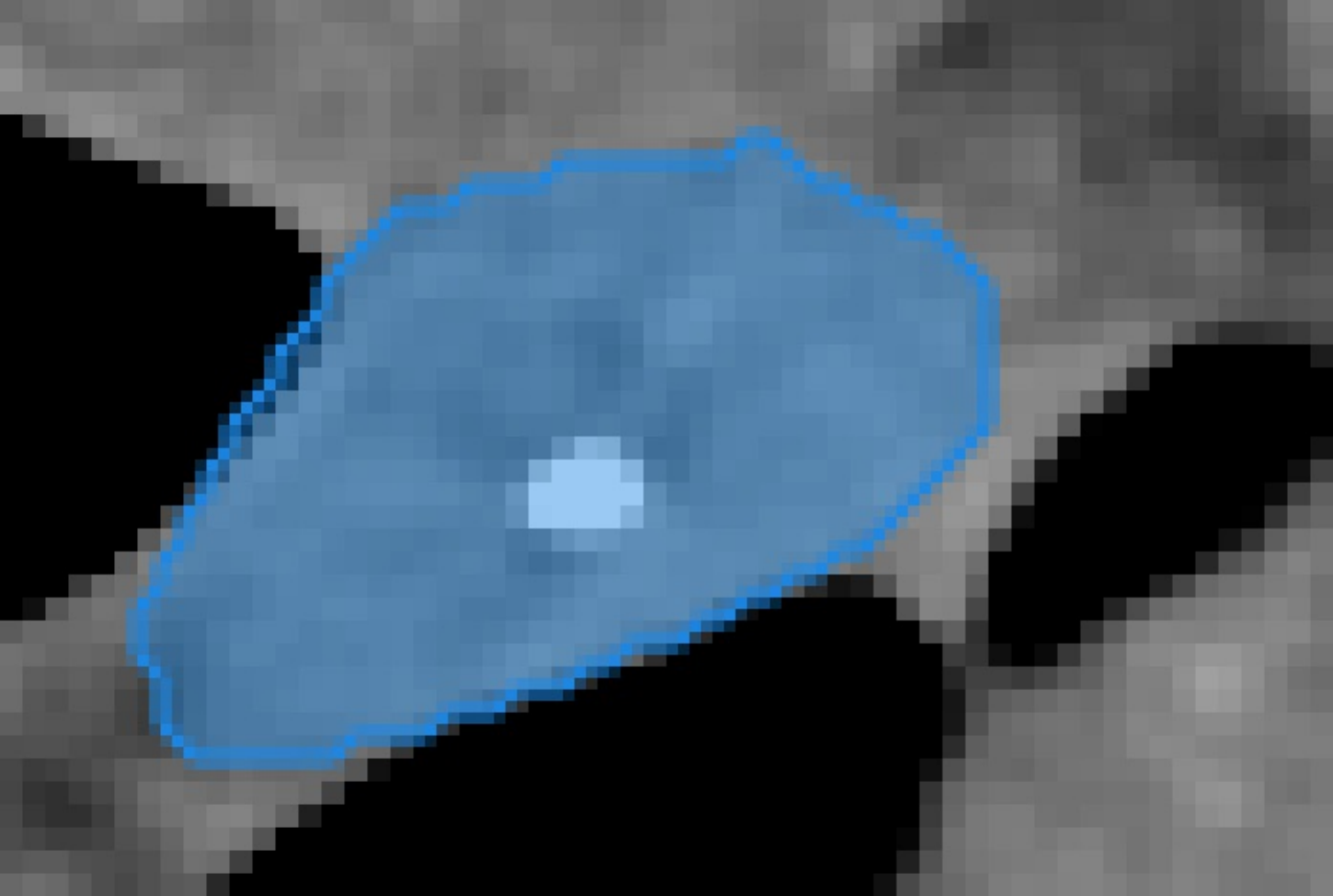} & \qp{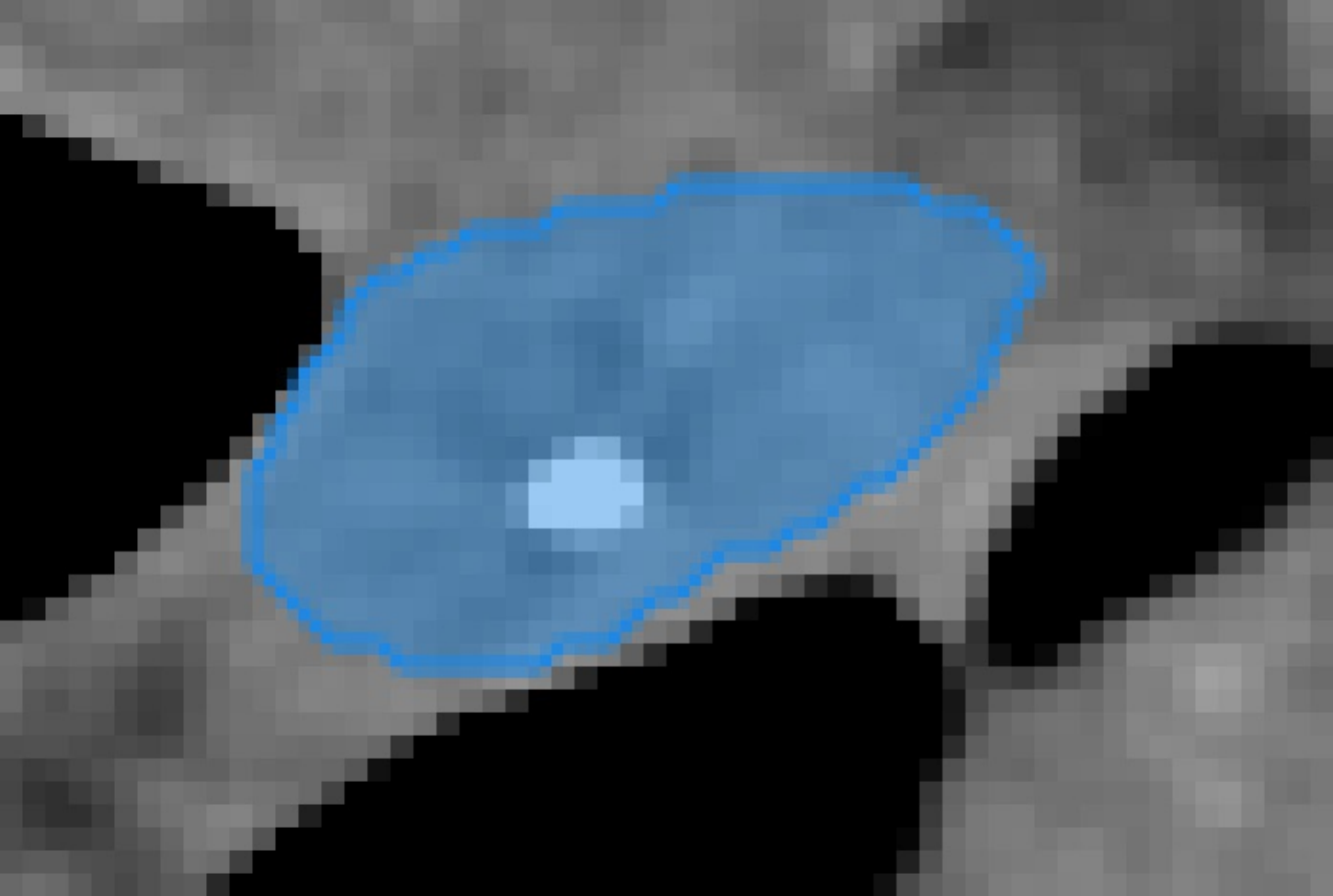} & \qp{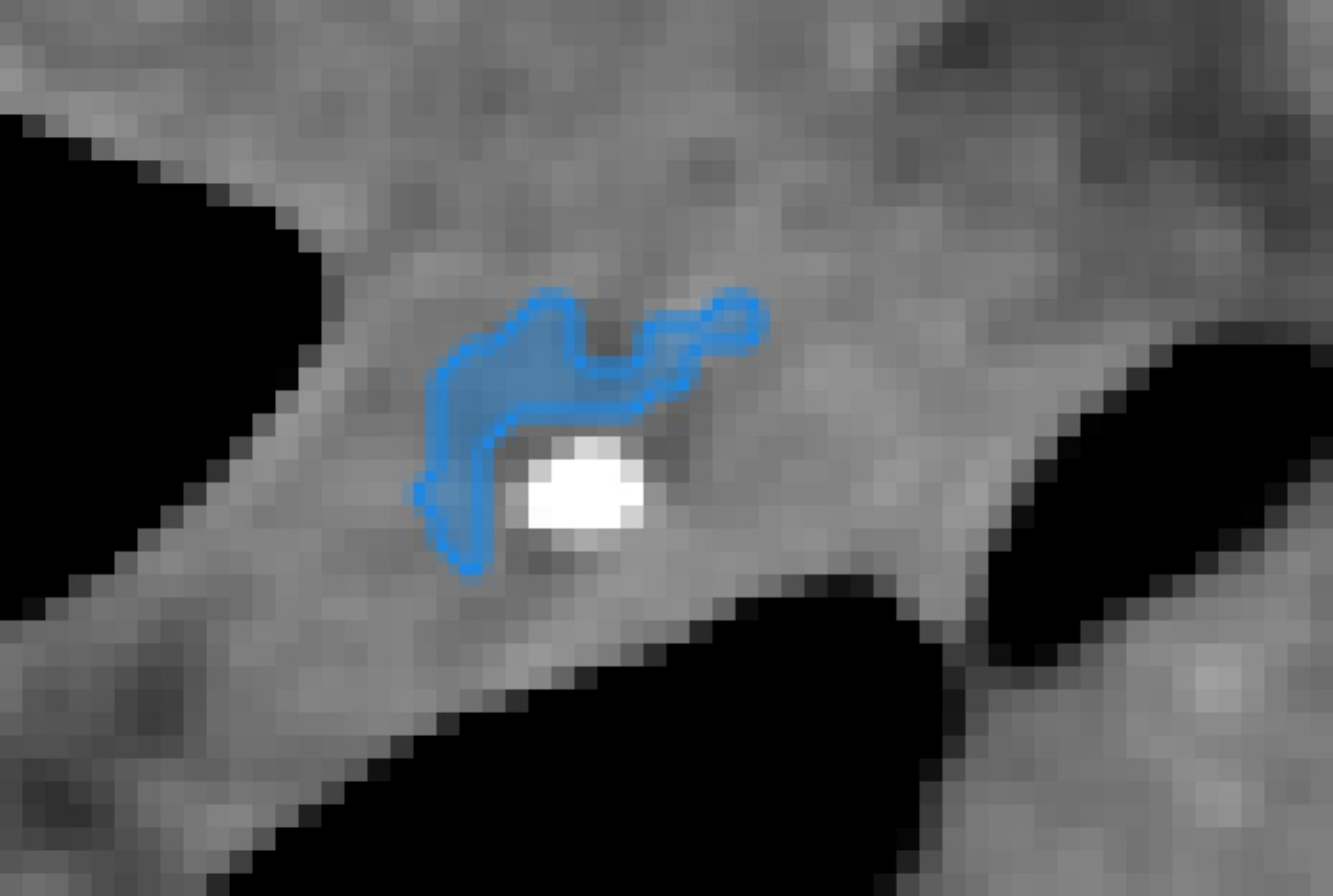} & \qp{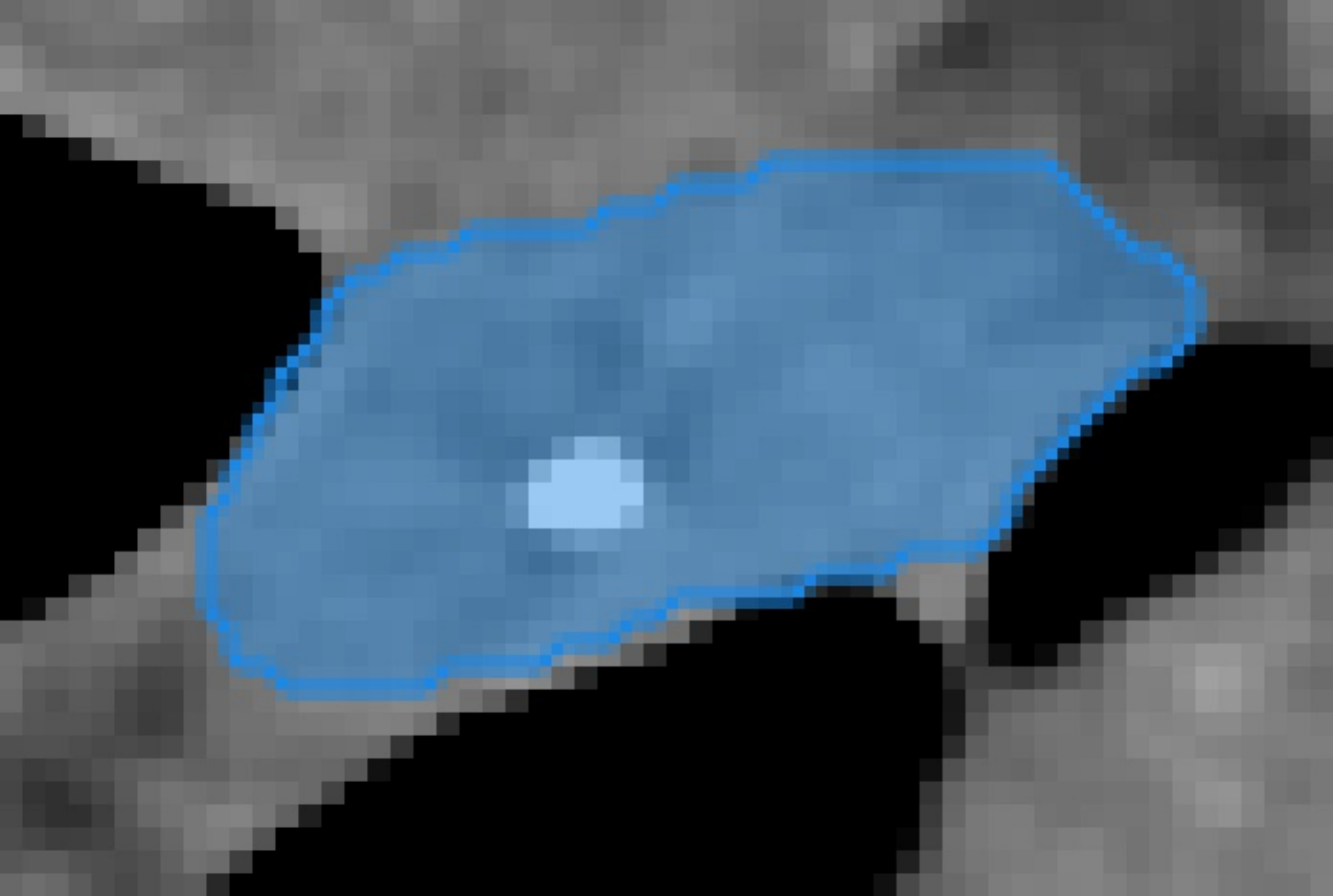}\\
 & \ql{AMOS \R{qual.C.case}} & \ql{\R{qual.C.dice.base}} & \ql{\R{qual.C.dice.D2}} & \ql{\R{qual.C.dice.V7A1}}\\
\end{tabular}
\caption{\textbf{Qualitative comparison on nested-contrast and thin-boundary cases.} (A,~B)~\archRule{} eliminates false-positive tumour predictions on a fluid cyst and an adjacent structure (FP mL \,|\, tumour Dice \%). (C)~\archM{} recovers a thin organ (\R{qual.C.organ} Dice \%) lost by decoder slimming (Sec.~\Sref{sec:s-pctgallery}).}
\label{fig:qual}
\end{figure}
}{}

\textbf{Subtyping failures expose a topological support mismatch between local $3^3$ kernels and region-level tissue contrast} (Fig.~\ref{fig:cyst}).
Whenever nested compartments share local $3^3$ boundary texture, bounded convolutions miss region-wide intensity moments while global attention leaks across boundaries.
On KiTS23, across \R{ins.corr.n} checkpoints (\R{ins.corr.narch} architectures), tumour Dice tracks precision ($r=\R{ins.corr.prec}$) rather than recall ($r=\R{ins.corr.rec}$; Fig.~\ref{fig:cyst}a): nnU-Net drops from \R{ins.cyst.free.base} on cyst-free to \R{ins.cyst.bear.base} on cyst-bearing kidneys (Fig.~\ref{fig:cyst}b; \R{ins.share.deficit.bear}\,\% of tumour error), and shifting lesion attenuation flips every network's call ($P{=}0.5$) at \R{an.hu.flip.min}--\R{an.hu.flip.max}\,HU inside the \R{an.hu.bosniak.lo}--\R{an.hu.bosniak.hi}\,HU clinical threshold~\cite{silverman2019bosniak} (Fig.~\ref{fig:cyst}c, Sec.~\Sref{sec:s-med2}).

\resIf{r31.on}{%
\begin{table}[t]
\centering\footnotesize
\setlength{\tabcolsep}{4pt}
\caption{\textbf{Constants remove R4's gain} (KiTS23 4-fold CV; matched topology and parameters with constant inputs).}
\label{tab:r4ctl}
\resizebox{0.90\columnwidth}{!}{%
\begin{tabular}{@{}lcc@{}}
\toprule
Contrast & $\Delta$Dice & Folds up\\
\midrule
Control $-$ \archM{} & \R{r31.ctx.d} & --\\
\archR{} $-$ control & \textbf{\R{r31.d}} [\R{r31.lo}, \R{r31.hi}] & \R{r31.nup}/\R{r31.nfold}\\
\archR{} $-$ \archM{} & \textbf{\R{fnd.w9.kits.dM}} & --\\
\bottomrule
\end{tabular}}
\end{table}
}{}%

\textbf{Mask-confined region contrast resolves subtype confusion; constant controls yield zero gain} (Tab.~\ref{tab:r4ctl}).
Unconstrained feature pooling (OCRNet~\cite{yuan2020ocr}, ACFNet~\cite{zhang2019acfnet}) trails \archM{} (\R{med.nwcv.kits.X1.dice.d}, \R{med.nwcv.kits.X4.dice.d} vs.\ \cvD{kits}{V7A1}).
Instead, R4 diffuses mask-confined intensity moments and contrast quantiles (\R{r4.bands}\,HU on CT) over \R{r4.steps} non-leaking $3^3$ steps inside the detached coarse prediction at \R{r4.kits.stages}, lifting KiTS23 CV by \R{fnd.w9.kits.dM} over \archM{} (\R{hyp.bm.kits.Y2.d} vs.\ nnU-Net, 4/4 folds).
\resIf{r31.on}{Constant-input controls abolish the effect (\R{r31.ctx.d}), isolating \archRule{}'s \R{r31.d} gain ([\R{r31.lo}, \R{r31.hi}], \R{r31.nup}/\R{r31.nfold} folds) to physical contrast, while cutting false-positive tumour per mL of cyst from \R{fnd.w9.slope.M} to \R{fnd.w9.slope}\,mL and extra-renal false positives from \R{an.extrfp.V7A1.mL} to \R{an.extrfp.Y2.mL}\,mL/case (Fig.~\ref{fig:qual}).}{}

\textbf{Anatomical error decomposition and open-ended refinement.}
Decomposing residual errors across \R{an.sat.amos.npairs} AMOS22 organ pairs and \R{med.kits.n} KiTS23 cases (Tabs.~\Sref{tab:s-med}, \Sref{tab:s-satdef}--\Sref{tab:s-satkits}) separates two regimes: on multi-organ AMOS22, \R{an.sat.amos.boundary.share.min}--\R{an.sat.amos.boundary.share.max}\,\% of deficit is sub-voxel boundary displacement ($e\approx\R{an.sat.e.base.vox}$ voxels) while deeper encoders cut absent-organ hallucinations (\R{med.abs.base.all}$\rightarrow$\R{med.abs.V7A1.all}); on nested KiTS23 pathology, \R{an.v6.kits.wo50.min}--\R{an.v6.kits.wo50.max}\,\% of tumour deficit is whole-object confusion, where \archRule{} leads on the gross-error stratum (\R{an.v6.s23.Y2.gross} vs.\ nnU-Net's \R{an.v6.s23.base.gross}) and passes all pre-registered clauses (Tab.~\Sref{tab:s-wave9}).
These residual boundary errors (\R{an.v6.kits.wo50.min}\,\% on \R{an.med2.cyst.n} complex masses) directly seed next-wave refinements (\R{hyp.next.kits.H1.dM}, \R{hyp.next.kits.J1.dM}, \R{hyp.next.kits.J2.dM}, \R{hyp.next.kits.Y3.dM} over \archM{} on KiTS23 CV): partial-volume unmixing lifts sub-5\,mL tumour Dice (\R{hyp.r5.base.small5.dice}$\rightarrow$\R{hyp.r5.J1.small5.dice}, \R{hyp.r5.J1.small5.d}), eroded-core pooling reaches \R{hyp.peak.kits.R5.dice} (\R{hyp.peak.kits.R5.d}) on KiTS23, and anatomy-conditioned contrast lifts 4-fold AMOS22 to \R{hyp.y13.amos.cv.dice} (\R{hyp.y13.amos.cv.dL1S} vs.\ ResEnc-1$\times$), KiPA22 to \R{hyp.peak.kipa22.Rule.dice} (\R{hyp.peak.kipa22.Rule.d}), BTCV to \R{hyp.peak.btcv.Rule.dice} (\R{hyp.peak.btcv.Rule.d}; 3-seed \R{hyp.y13.btcv.mean}), and HepaticVessel to \R{hyp.peak.hepv.Rule.dice} (\R{hyp.peak.hepv.Rule.d}).

\begin{table}[!t]
\centering\scriptsize
\setlength{\tabcolsep}{4.5pt}
\renewcommand{\arraystretch}{1.02}
\caption{\textbf{All-split overlap and boundary gains.} \archM{} minus nnU-Net across all folds/runs.}
\label{tab:consist}
\resizebox{0.88\columnwidth}{!}{%
\begin{tabular}{@{}llccc@{}}
\toprule
 & Benchmark & $\Delta$Dice & $\Delta$NSD@1mm & $\Delta$NSD@2mm\\
\midrule
\multirow{3}{*}{Design} & KiTS23 & \R{hyp.bm.kits.V7A1.d} & \R{hyp.bm.kits.V7A1.dnsd1} & \R{hyp.bm.kits.V7A1.dnsd2}\\
 & AMOS22 & \R{hyp.bm.amos.V7A1.d} & \R{hyp.bm.amos.V7A1.dnsd1} & \R{hyp.bm.amos.V7A1.dnsd2}\\
 & ACDC & \R{hyp.bm.acdc.V7A1.d} & \R{hyp.bm.acdc.V7A1.dnsd1} & \R{hyp.bm.acdc.V7A1.dnsd2}\\
\midrule
\multirow{\resIf{hyp.lkipa.on}{4}{3}}{*}{Locked} & AMOS22 & \R{hyp.bm.lamos.V7A1.d} & \R{hyp.bm.lamos.V7A1.dnsd1} & \R{hyp.bm.lamos.V7A1.dnsd2}\\
 & ACDC & \R{hyp.bm.lacdc.V7A1.d} & \R{hyp.bm.lacdc.V7A1.dnsd1} & \R{hyp.bm.lacdc.V7A1.dnsd2}\\
 & Pancreas & \R{hyp.bm.lpanc.V7A1.d} & \R{hyp.bm.lpanc.V7A1.dnsd1} & \R{hyp.bm.lpanc.V7A1.dnsd2}\\
\resIf{hyp.lkipa.on}{ & KiPA22 & \R{hyp.bm.lkipa.V7A1.d} & \R{hyp.bm.lkipa.V7A1.dnsd1} & \R{hyp.bm.lkipa.V7A1.dnsd2}\\}{}%
\midrule
\multirow{7}{*}{Unseen} & LiTS & \R{hyp.bm.lits.V7A1.d} & \R{hyp.bm.lits.V7A1.dnsd1} & \R{hyp.bm.lits.V7A1.dnsd2}\\
 & Pancreas & \R{hyp.bm.panc.V7A1.d} & \R{hyp.bm.panc.V7A1.dnsd1} & \R{hyp.bm.panc.V7A1.dnsd2}\\
 & BTCV & \R{hyp.bm.btcv.V7A1.d} & \R{hyp.bm.btcv.V7A1.dnsd1} & \R{hyp.bm.btcv.V7A1.dnsd2}\\
 & BraTS (MRI) & \R{hyp.bm.brats.V7A1.d} & \R{hyp.bm.brats.V7A1.dnsd1} & \R{hyp.bm.brats.V7A1.dnsd2}\\
 & HepaticVessel & \R{hyp.bm.hepv.V7A1.d} & \R{hyp.bm.hepv.V7A1.dnsd1} & \R{hyp.bm.hepv.V7A1.dnsd2}\\
 & KiPA22 & \R{hyp.bm.kipa22.V7A1.d} & \R{hyp.bm.kipa22.V7A1.dnsd1} & \R{hyp.bm.kipa22.V7A1.dnsd2}\\
 & Spleen & \R{hyp.bm.spleen.V7A1.d} & \R{hyp.bm.spleen.V7A1.dnsd1} & \R{hyp.bm.spleen.V7A1.dnsd2}\\
\bottomrule
\end{tabular}}
\end{table}

\begin{figure*}[t]
\centering
\includegraphics[width=0.95\textwidth]{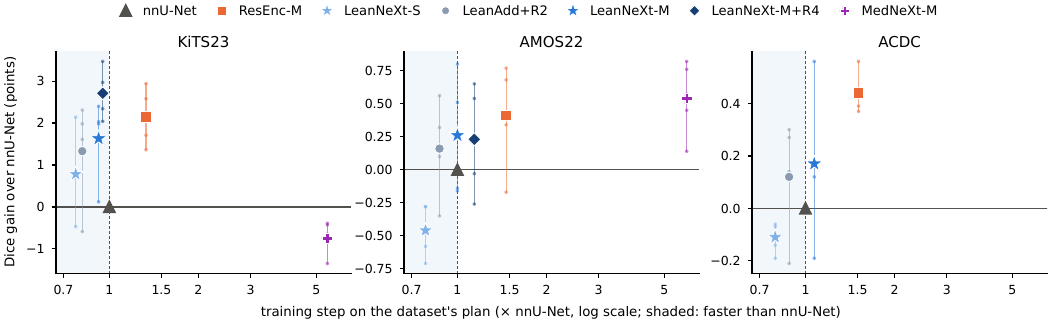}
\caption{\textbf{Cross-validated accuracy--speed Pareto frontier across all three design datasets.} 4-fold CV $\Delta$Dice over nnU-Net (large markers: 4-fold mean; small dots: individual folds~1--4) vs.\ profiled training step ratio on each dataset's native plan (shaded region: faster than nnU-Net). \archM{} (LeanNeXt-M) and \archRule{} (LeanNeXt-M+R4) occupy the upper-left Pareto frontier inside the $\le 1\times$ step budget, whereas ResEnc-M ($1.34$--$1.50\times$ step, $3.3\times$ parameters) and MedNeXt-M ($5.45$--$5.95\times$ step; OOM on ACDC) lie far to the right.}
\label{fig:pareto}
\end{figure*}

\subsection{Do the Rules Guide New Designs?}
\label{sec:controls}
\label{sec:rulepred}
\textbf{Pre-training triggers predict rule transfer across unseen CT and MRI benchmarks} (Fig.~\ref{fig:triggers}).
Across all \R{hyp.trig.r2.nclass} classes of the \R{hyp.trig.r2.nds} unseen CT and MRI benchmarks, classes above R2's boundary-share threshold gain \R{hyp.trig.r2.above.d} Dice when R2 joins R1$+$R3 (\R{hyp.trig.r2.above.nup}/\R{hyp.trig.r2.nabove} positive) vs.\ \R{hyp.trig.r2.below.d} on the \R{hyp.trig.r2.nbelow} below (Fig.~\ref{fig:triggers}b).
Across MRI, beyond ACDC (\cvDM{acdc} CV, \R{r18.acdc.d} locked), R2 stays off on compact Heart left-atrium MRI (boundary share \R{hyp.mri.heart.bf}\,$<$\,\R{mr.r2.thr}), where R1$+$R3 gains up to \R{hyp.mri.heart.S.peak} (\R{hyp.mri.heart.S.dM} vs.\ R2); R1 lifts Hippocampus MRI by \R{hyp.mri.hippo.r1.d} (\R{hyp.mri.hippo.r1.nup}/4 folds); \archM{} lifts 13-organ AMOS22-MRI by \R{hyp.mri.amos13.M.d} (\R{hyp.mri.amos13.M.dL1} vs.\ ResEnc-M); and on BraTS (\hoD{brats}{Y2} if CT bands forced), \archRule{} hits \R{hyp.peak.brats.Rule.dice} (\R{hyp.peak.brats.Rule.d}, \R{hyp.peak.brats.Rule.dL1} vs.\ ResEnc-M\resIf{hyp.mri.brats.Y7.on}{; 3-run \R{hyp.mri.brats.Y7.d}}{}).
CT HU-band R4 fires on KiTS23 (\R{fnd.w9.kits.dM}) and stays silent on the \R{hyp.trig.r4.nsilent} non-nested CT tasks (forcing it: \R{hyp.trig.r4.silent.mean}; Fig.~\ref{fig:triggers}c): zero-shot rulebook routing averages \R{hyp.rule.all7.d} across all \R{hyp.rule.all7.n} multi-class, lesion, and MRI benchmarks (\R{hyp.rule.all7.npos}/\R{hyp.rule.all7.n} up), beating R1$+$R3 (\R{hyp.rule.all7.S.d}), R1--R3 (\R{hyp.rule.all7.M.d}), and unselective R4 (\R{hyp.rule.ct5.rule.d} vs.\ \R{hyp.rule.ct5.Y2.d}).

\begin{figure}[!t]
\centering
\includegraphics[width=0.92\columnwidth]{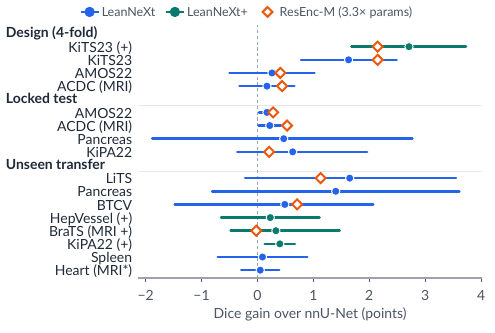}
\caption{\textbf{Positive mean Dice gains across all \R{hyp.ev.rule.n} CT and MRI evaluations.} Mean paired $\Delta$Dice and 95\,\% CIs vs.\ nnU-Net (diamonds: ResEnc-M).}
\label{fig:forest}
\end{figure}

\subsection{Consistency Across Folds and Runs}
\label{sec:heldout}
\textbf{Across all folds and seeds, \archRule{} improves all \R{hyp.ev.rule.n}/\R{hyp.ev.rule.n} evaluations (benchmark sign test $p=3.1\times 10^{-5}$), with \archM{} ahead on \R{hyp.split.npos}/\R{hyp.split.n} paired splits} (Fig.~\ref{fig:forest}, Tab.~\ref{tab:consist}) and positive $\Delta$NSD@1/2\,mm on every row.
Mean 4-fold CV gains are \cvDK{} on KiTS23 (\archR{} \R{hyp.bm.kits.Y2.d}), \cvDM{amos} on AMOS22, and \cvDM{acdc} on ACDC; locked test gains are \R{r18.amos.d} (AMOS22), \R{r18.acdc.d} (ACDC), \resIf{hyp.lkipa.on}{\R{hyp.bm.lpanc.V7A1.d} (Pancreas; cancer \R{hyp.bm.lpanc.V7A1.cancer.d}), and \R{hyp.bm.lkipa.V7A1.d} (KiPA22; tumour \R{hyp.bm.lkipa.V7A1.tumour.d})}{and \R{hyp.bm.lpanc.V7A1.d} (Pancreas; cancer \R{hyp.bm.lpanc.V7A1.cancer.d})}, winning \R{hyp.lock.npos}/\R{hyp.lock.n} independent runs ($p=\R{hyp.lock.p}$).
Unseen gains concentrate on complex lesions and thin anatomy (\R{hyp.bm.lits.V7A1.d} LiTS [tumour \R{hyp.bm.lits.V7A1.tumour.d}], \R{hyp.bm.panc.V7A1.d} Pancreas [cancer \R{hyp.bm.panc.V7A1.cancer.d}], \R{hyp.bm.btcv.V7A1.d} BTCV [pancreas \R{hyp.bm.btcv.V7A1.panc.d}], \R{hyp.mri.brats.M.d} BraTS, \R{hyp.bm.kipa22.Y3.d} KiPA22 [tumour \R{hyp.bm.kipa22.Y3.tumour.d}; HepV tumour \R{hyp.bm.hepv.Rule.tumour.d}]), while saturated single-organ masks stay non-degraded (\R{hyp.bm.spleen.V7A1.d} Spleen at $97.35\,\%$, \R{hyp.mri.heart.S.d} Heart): \R{hyp.cls.lesion.n} unseen lesions average \R{hyp.cls.lesion.mean} (\R{hyp.cls.lesion.npos}/\R{hyp.cls.lesion.n} up) vs.\ \R{hyp.cls.organ.mean} across \R{hyp.cls.organ.n} organs.
Two seeds/fold on KiTS23 cut run SD fourfold (\R{an.v7.cal.kits.sigma.base}$\rightarrow$\R{an.v7.cal.kits.sigma.V7A1}), with \archR{} ahead of \archM{} by \R{r34.ii.d} (\R{r34.ii.k}/\R{r34.ii.n} pairs).

\begin{table}[t]
\centering\scriptsize
\setlength{\tabcolsep}{2.8pt}
\renewcommand{\arraystretch}{1.05}
\caption{\textbf{Two-seed 4-fold CV verification on KiTS23 ($8$ paired runs per contrast).} Mean $\Delta$Dice across folds~1--4 and seeds~0--1, 95\,\% interval incorporating both case bootstrap and run-to-run training variance, winning fold--seed pairs $k/n$ (exact one-sided sign test $p$), and per-seed means.}
\label{tab:r34}
\resizebox{0.96\columnwidth}{!}{%
\begin{tabular}{@{}lccccc@{}}
\toprule
Paired contrast (KiTS23) & $\Delta$Dice & 95\,\% CI & $k/n$ ($p$) & Seed 0 & Seed 1\\
\midrule
\archM{} $-$ nnU-Net & \textbf{\R{r20.d2}} & [\R{r20.d2.lo}, \R{r20.d2.hi}] & 8/8 (0.0039) & \cvD{kits}{V7A1} & +1.63\\
\archR{} $-$ \archM{} & \textbf{\R{r34.ii.d}} & [\R{r34.ii.lo}, \R{r34.ii.hi}] & \R{r34.ii.k}/\R{r34.ii.n} (\R{r34.ii.p}) & \R{r34.ii.s0.d} & \R{r34.ii.s1.d}\\
\archR{} $-$ nnU-Net & \textbf{\R{r34.iiiY2.d}} & [\R{r34.iiiY2.lo}, \R{r34.iiiY2.hi}] & \R{r34.iiiY2.k}/\R{r34.iiiY2.n} (\R{r34.iiiY2.p}) & \R{r34.iiiY2.s0.d} & \R{r34.iiiY2.s1.d}\\
\archR{} $-$ ResEnc-M & \textbf{\R{r34.i.d}} & [\R{r34.i.lo}, \R{r34.i.hi}] & \R{r34.i.k}/\R{r34.i.n} (\R{r34.i.p}) & \R{r34.i.s0.d} & \R{r34.i.s1.d}\\
ResEnc-M $-$ nnU-Net & \R{r34.iiiL1.d} & [\R{r34.iiiL1.lo}, \R{r34.iiiL1.hi}] & \R{r34.iiiL1.k}/\R{r34.iiiL1.n} (\R{r34.iiiL1.p}) & \R{r34.iiiL1.s0.d} & \R{r34.iiiL1.s1.d}\\
\bottomrule
\end{tabular}}
\end{table}

\textbf{Insight 4: Targeted rules compress run-to-run training variance fourfold and shift the Pareto frontier without capacity inflation} (Fig.~\ref{fig:pareto}, Tab.~\ref{tab:r34}).
Running two independent seeds on every fold of KiTS23 ($8$ runs per architecture; Tab.~\ref{tab:r34}) uncovers a striking stability advantage: whereas default nnU-Net exhibits a within-fold run-to-run standard deviation of \R{an.v7.cal.kits.sigma.base} Dice points, \archM{} reduces run-to-run SD nearly fourfold to \R{an.v7.cal.kits.sigma.V7A1} points, winning all $8/8$ fold--seed pairs over nnU-Net (\R{r20.d2}, 95\,\% CI [\R{r20.d2.lo}, \R{r20.d2.hi}]).
Adding R4's zero-initialized region contrast (\archR{}) further improves over \archM{} in \R{r34.ii.k}/\R{r34.ii.n} fold--seed pairs (\R{r34.ii.d}, $p=\R{r34.ii.p}$), beats nnU-Net in \R{r34.iiiY2.k}/\R{r34.iiiY2.n} pairs (\R{r34.iiiY2.d}, $p=\R{r34.iiiY2.p}$), and leads $3.3\times$ larger ResEnc-M in \R{r34.i.k}/\R{r34.i.n} pairs (\R{r34.i.d}; \R{r34.i.s0.d} on seed~0 and \R{r34.i.s1.d} on seed~1).
Across all three design datasets (Fig.~\ref{fig:pareto}), \archM{} and \archRule{} establish a new accuracy--speed Pareto frontier inside the $\le 1\times$ step-time regime, demonstrating that falsifiable architectural rules achieve what brute-force parameter scaling and 5$\times$ slower depthwise backbones cannot.

\section{Conclusion}
\label{sec:conclusion}
We introduced a hypothesis-driven architecture discovery loop that reframes automated neural design from black-box score selection into falsifiable scientific inquiry: LLM agents treat failures as diagnostic evidence, test counterfactual predictions under a frozen pipeline, and distill confirmed mechanisms into executable \texttt{trigger}\,$\rightarrow$\,\texttt{action} rules.
Uncovering cross-modality 3D principles for surface-to-volume compute (\archM{}) and support-confined region contrast (\archRule{}), the discovery loop outperforms random sampling, one-shot LLM generation, and score-selected agent search, improving all \R{hyp.ev.rule.n} CT/MRI evaluations over nnU-Net and surpassing $3.3\times$ larger ResEnc-M on \R{hyp.peak.vsL1.nwin}/\R{hyp.peak.vsL1.ntot} benchmarks at \R{fnd.w9.par.vsL1}$\times$ parameters.

\label{endofbody}

{
    \small
    \hyphenation{EvoPrompting ConvNeXt ConvNeXts nnFormer MedNeXt IJCAI}

}
\end{document}